%% file: main.tex
\documentclass[11pt, a4paper, logo, onecolumn, nonumbering]{googledeepmind}

\pdftrailerid{redacted}


\input{preamble.tex}
\title{\centering Video models are zero-shot learners~and~reasoners} 
\usepackage[numbers, sort&compress]{natbib}
\correspondingauthor{my-email@google.com}

\reportnumber{} %

\author[1]{Thaddäus~Wiedemer\textsuperscript{*}}
\author[1]{Yuxuan~Li}
\author[1]{Paul~Vicol}
\author[1]{Shixiang~Shane~Gu}
\author[1]{Nick~Matarese}
\author[1]{Kevin~Swersky}
\author[1]{Been~Kim}
\author[1]{Priyank~Jaini\textsuperscript{*}}
\author[1]{Robert~Geirhos\textsuperscript{*}}

\affil[]{Google DeepMind}

\newcommand{\websiteurl}{https://video-zero-shot.github.io/}
\input{sections/00_abstract}

\begin{document}

\newpage
\maketitle

\input{sections/01_intro}
\input{sections/02_methods}
\input{sections/03_qualitative}
\input{sections/04_quantitative}
\input{sections/05_discussion}

\section*{Acknowledgements}
We would like to thank Oyvind Tafjord, Mike Mozer, Katherine Hermann, Andrew Lampinen, Viorica Patraucean, Shiry Ginosar, Ross Goroshin, Abhijit Ogale, Claire Cui, Kun Zhang for helpful feedback/discussions, Anish Nangia for Vertex insights, Tuan Anh Le for suggesting the Omniglot evaluation, Medhini Narasimhan and Pieter-Jan Kindermans for Veo insights and discussion, Micromelon Robotics for permission to use the image in \cref{fig:robot_navigation}, Shelly Sheynin for permission to use the EMU Edit dataset, David Fleet and Jon Shlens for their support, and the Veo team for developing an incredible model.

\section*{Author contributions}
Leads: TW, PJ, RG. Project idea: RG. Core contributors: YL, PV. Partial contributor: SG. \cref{fig:doodle}: NM. Advisors: KS, BK.

\bibliographystyle{abbrvnat}
\bibliography{refs}

\newpage
\appendix

\section*{Appendix}
\input{sections/99_appendix}

\end{document}

%% file: preamble.tex
\usepackage{kantlipsum, lipsum}
\usepackage{dsfont}
\usepackage{multirow}
\usepackage{array}
\usepackage{amssymb} %
\usepackage{pifont} %
\usepackage{makecell}
\usepackage{algorithm}
\usepackage{algpseudocode}
\usepackage{amsmath}
\usepackage{multicol}
\usepackage{float}
\usepackage{siunitx}
\usepackage{tabularx}
\usepackage{booktabs}

\usepackage{xcolor}
\definecolor{GoogleBlue}{HTML}{1A73E8}
\definecolor{GoogleBlueLight}{HTML}{e8f0fe}
\definecolor{GoogleBlueDark}{HTML}{174ea6}
\definecolor{GoogleRed}{HTML}{D93025}
\definecolor{GoogleGray}{HTML}{767676}
\definecolor{figtextcolor}{HTML}{333333}
\definecolor{perceptioncol}{HTML}{1A73E8}
\definecolor{modelingcol}{HTML}{925BCF}
\definecolor{manipulationcol}{HTML}{C440A7}
\definecolor{reasoningcol}{HTML}{DA2F77}

\usepackage{subcaption}
\usepackage{opensans}
\DeclareCaptionFont{subfigcapsize}{\fontsize{8pt}{8pt}\selectfont}  %
\newcommand{\veo}{Veo}
\newcommand{\veotwo}{Veo~2}  %
\newcommand{\veothree}{Veo~3}  %
\newcommand{\geminipro}{Gemini~2.5~Pro}  %
\newcommand{\nanobanana}{Nano~Banana}  %

\newcommand{\threefigwidth}{0.325}
\newcommand{\fourfigwidth}{0.24}
\usepackage{xurl}
\usepackage[colorlinks=true, allcolors=GoogleBlue]{hyperref}
\usepackage[capitalise]{cleveref}
\crefname{section}{Sec.}{Secs.}  %
\crefname{subsection}{Sec.}{Secs.}  %
\crefname{subsubsection}{Sec.}{Secs.}  %
\crefname{appendix}{App.}{Apps.}  %

\usepackage{graphicx}
\graphicspath{{figures/}}

\usepackage[breakable]{tcolorbox}
\newcommand{\callout}[1]{%
    \begin{center}
    \begin{tcolorbox}[colframe=GoogleBlue, colback=GoogleBlueLight, coltext=GoogleBlue!30!black, width=\columnwidth, left=2mm, right=2mm, top=1mm, bottom=1mm, boxrule=.5mm]
        #1
    \end{tcolorbox}
    \end{center}
}
\newcounter{takeaway}
\newcommand{\takeaway}[1]{%
    \stepcounter{takeaway}
    \callout{\textbf{Takeaway\;\thetakeaway}\hspace{1em}#1}
}

\usepackage{tikz}
\usetikzlibrary{calc}
\newlength{\tikzwidth}
\newlength{\boxpad}
\newlength{\boxgap}
\newlength{\boxtextheight}

%% file: sections/00_abstract.tex
\begin{abstract}
The remarkable zero-shot capabilities of Large Language Models (LLMs) have propelled natural language processing from task-specific models to unified, generalist foundation models.
This transformation emerged from simple primitives: large, generative models trained on web-scale data.
Curiously, the same primitives apply to today's generative video models. Could video models be on a trajectory towards general-purpose \emph{vision} understanding, much like LLMs developed general-purpose \emph{language} understanding? We demonstrate that Veo~3 can solve a broad variety of tasks it wasn't explicitly trained for:
segmenting objects, detecting edges, editing images, understanding physical properties, recognizing object affordances, simulating tool use, and more.
These abilities to perceive, model, and manipulate the visual world enable early forms of visual reasoning like maze and symmetry solving.
Veo's emergent zero-shot capabilities indicate that video models are on a path to becoming unified, generalist vision foundation models.

Project page: \url{\websiteurl}
\end{abstract}

%% file: sections/01_intro.tex
\section{Introduction}
\label{sec:intro}
We believe that video models will become unifying, general-purpose foundation models for machine vision just like large language models (LLMs) have become foundation models for natural language processing (NLP).
Within the last few years, NLP underwent a radical transformation: from task-specific, bespoke models (e.g., one model for translation, another one for question-answering, yet another one for summarization) to LLMs as unified foundation models. Today's LLMs are capable of general-purpose language understanding, which enables a single model to tackle a wide variety of tasks including coding~\citep{jiang2024survey}, math~\citep{comanici2025gemini}, creative writing~\citep{wang2024weaver}, summarization, translation~\citep{zhu2023multilingual}, and deep research~\citep{lu2024aiscientist, schmidgall2025agent}. These abilities started to emerge from simple primitives: training large, generative models on web-scale datasets \citep[e.g][]{brown2020language,wei2022emergent}. As a result, LLMs are increasingly able to solve novel tasks through few-shot in-context learning~\citep{brown2020language,dong2022survey} and zero-shot learning~\citep{kojima2022large}. Zero-shot learning here means that prompting a model with a task instruction replaces the need for fine-tuning or adding task-specific inference heads.

Machine vision today in many ways resembles the state of NLP a few years ago: There are excellent task-specific models like ``Segment Anything''~\citep{kirillov2023segment,ravi2024sam} for segmentation or YOLO variants for object detection~\citep{redmon2016you,khanam2024yolov11}. While attempts to unify some vision tasks exist~\citep{acuaviva2025generation,zamir2018taskonomy,lin2025realgeneral,li2025visualcloze,wang2023images,xie2025test,lin2024pixwizard,xiao2025omnigen,le2025one,molad2023dreamix,ravishankar2025scaling}, no existing model can solve \emph{any} problem just by prompting. However, the exact same primitives that enabled zero-shot learning in NLP also apply to today's generative video models---large-scale training with a generative objective (text/video continuation) on web-scale data~\citep{yang2024video}. In this article, we therefore ask: Do video models develop general-purpose \emph{vision} understanding, similar to how LLMs developed general-purpose \emph{language} understanding? We answer this question in the affirmative:

\begin{enumerate}
\item Analyzing 18,384 generated videos across 62 qualitative and 7 quantitative tasks, we report that \veothree{} can solve a wide range of tasks that it was neither trained nor adapted for.
\item Based on its ability to \textbf{\textcolor{perceptioncol}{perceive}}, \textbf{\textcolor{modelingcol}{model}}, and \textbf{\textcolor{manipulationcol}{manipulate}} the visual world, \veothree{} shows early forms of ``chain-of-frames (CoF)'' \textbf{\textcolor{reasoningcol}{visual reasoning}} like maze and symmetry solving.
\item While task-specific bespoke models still outperform a zero-shot video model, we observe a substantial and consistent performance improvement from \veotwo{} to \veothree{}, indicating a rapid advancement in the capabilities of video models.
\end{enumerate}

%% file: sections/02_methods.tex
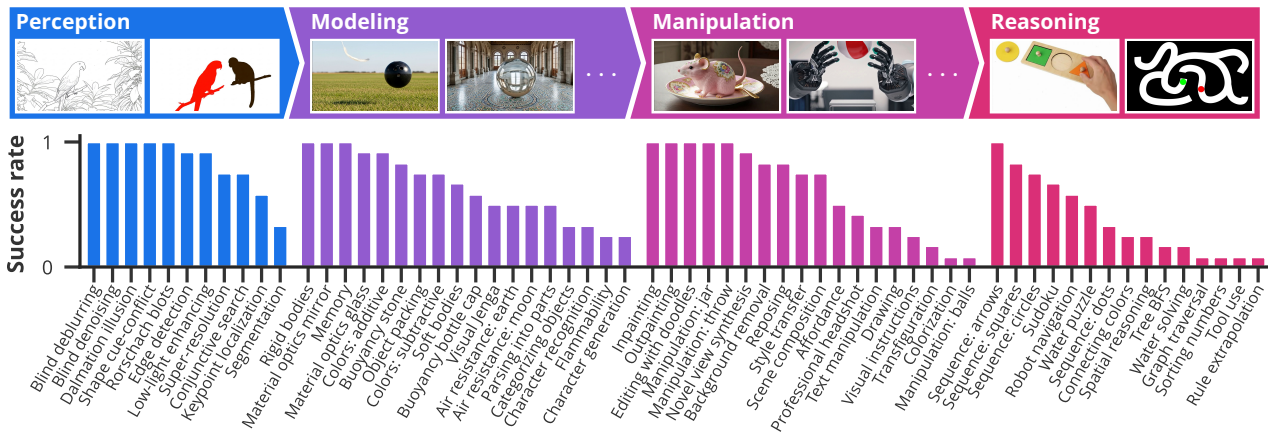
\begin{figure}[t]
    \centering
    \resizebox{\linewidth}{!}{%
    \begin{tikzpicture}
        \newlength{\boxheight};
        \setlength{\boxheight}{42.4375pt}
        \newlength{\boxtip}
        \setlength{\boxtip}{6pt}
        \newlength{\percepmodel}
        \setlength{\percepmodel}{111.5pt}
        \newlength{\modelmanip}
        \setlength{\modelmanip}{240.5pt}
        \newlength{\manipreason}
        \setlength{\manipreason}{368.31595pt}

        \coordinate (B1sw) at (0,0);
        \coordinate (B1nw) at (0, \boxheight);
        \coordinate (B1ne) at (\percepmodel - \boxgap - \boxtip, \boxheight);
        \coordinate (B1e) at (\percepmodel - \boxgap, 0.5\boxheight);
        \coordinate (B1se) at (\percepmodel - \boxgap - \boxtip, 0);
        \fill[perceptioncol] (B1sw) -- (B1nw) -- (B1ne) -- (B1e) -- (B1se) -- cycle;
        \coordinate (B2sw) at ($(B1se) + (\boxgap, 0)$);
        \coordinate (B2w) at ($(B1e) + (\boxgap, 0)$);
        \coordinate (B2nw) at ($(B1ne) + (\boxgap, 0)$);
        \coordinate (B2ne) at (\modelmanip - \boxgap - \boxtip, \boxheight);
        \coordinate (B2e) at (\modelmanip - \boxgap, 0.5\boxheight);
        \coordinate (B2se) at (\modelmanip - \boxgap - \boxtip, 0);
        \fill[modelingcol] (B2sw) -- (B2w) -- (B2nw) -- (B2ne) -- (B2e) -- (B2se) -- cycle;
        \coordinate (B3sw) at ($(B2se) + (\boxgap, 0)$);
        \coordinate (B3w) at ($(B2e) + (\boxgap, 0)$);
        \coordinate (B3nw) at ($(B2ne) + (\boxgap, 0)$);
        \coordinate (B3ne) at (\manipreason - \boxgap - \boxtip, \boxheight);
        \coordinate (B3e) at (\manipreason - \boxgap, 0.5\boxheight);
        \coordinate (B3se) at (\manipreason - \boxgap - \boxtip, 0);
        \fill[manipulationcol] (B3sw) -- (B3w) -- (B3nw) -- (B3ne) -- (B3e) -- (B3se) -- cycle;
        \coordinate (B4sw) at ($(B3se) + (\boxgap, 0)$);
        \coordinate (B4w) at ($(B3e) + (\boxgap, 0)$);
        \coordinate (B4nw) at ($(B3ne) + (\boxgap, 0)$);
        \coordinate (B4ne) at (\tikzwidth, \boxheight);
        \coordinate (B4se) at (\tikzwidth, 0);
        \fill[reasoningcol] (B4sw) -- (B4w) -- (B4nw) -- (B4ne) -- (B4se) -- cycle;

        \tikzset{
            boxcaption/.style={
                font=\fontsize{8}{8}\opensans\bfseries,
                text=white,
                anchor=west
            }
        }
        \coordinate (B1) at (B1nw);
        \coordinate (B2) at ($(B2nw) + (\boxtip, 0)$);
        \coordinate (B3) at ($(B3nw) + (\boxtip, 0)$);
        \coordinate (B4) at ($(B4nw) + (\boxtip, 0)$);
        \node[boxcaption] at ($(B1) + (-1.5pt, -0.5\boxtextheight)$) {Perception};
        \node[boxcaption] at ($(B2) + (-1.5pt, -0.5\boxtextheight)$) {Modeling};
        \node[boxcaption] at ($(B3) + (-1.5pt, -0.5\boxtextheight)$) {Manipulation};
        \node[boxcaption] at ($(B4) + (-1.5pt, -0.5\boxtextheight)$) {Reasoning};

        \newlength{\imagewidth}
        \setlength{\imagewidth}{47pt}
        \tikzset{
            exampleframe/.style={
                anchor=north west,
                inner sep=0pt,
                draw,
                white,
                line width=2pt
            }
        }
        \node[exampleframe] at ($(B1) + (\boxpad, -\boxtextheight)$) (i1a) {
                \includegraphics[width=\imagewidth]{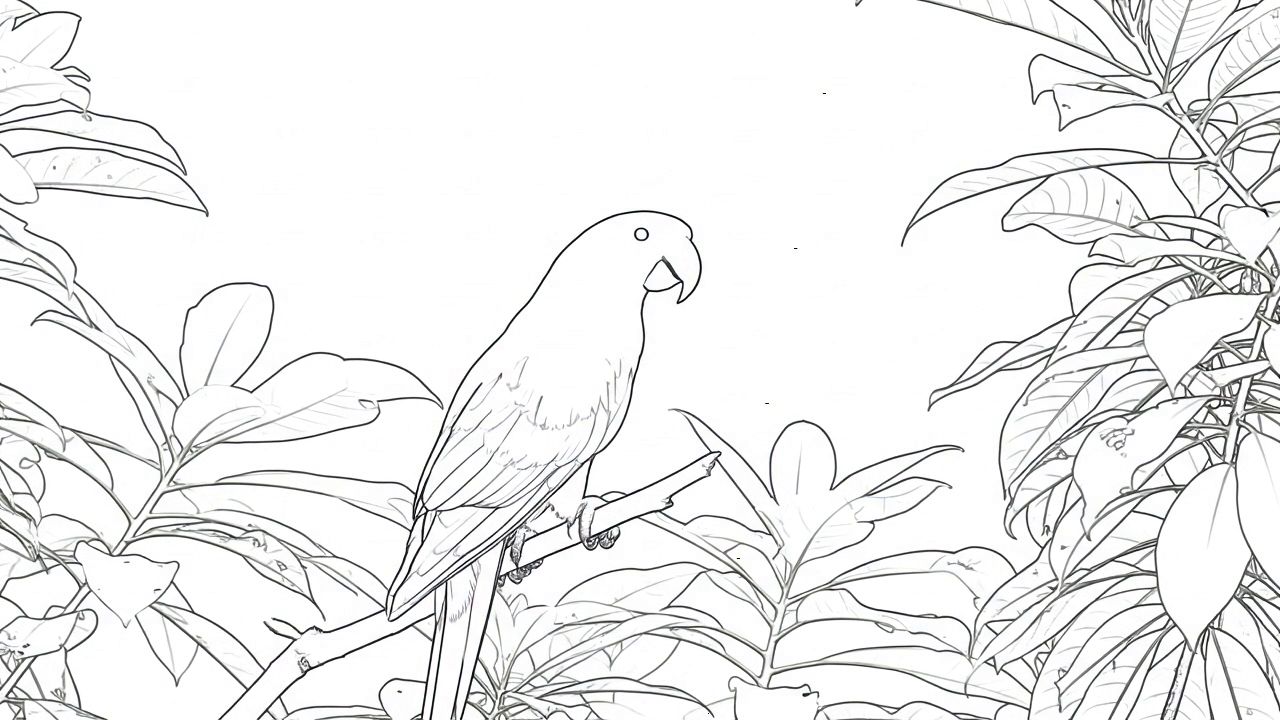}
        };
        \node[exampleframe] at ($(B1) + (\imagewidth + 3\boxpad, -\boxtextheight)$) (i1b) {
                \includegraphics[width=\imagewidth]{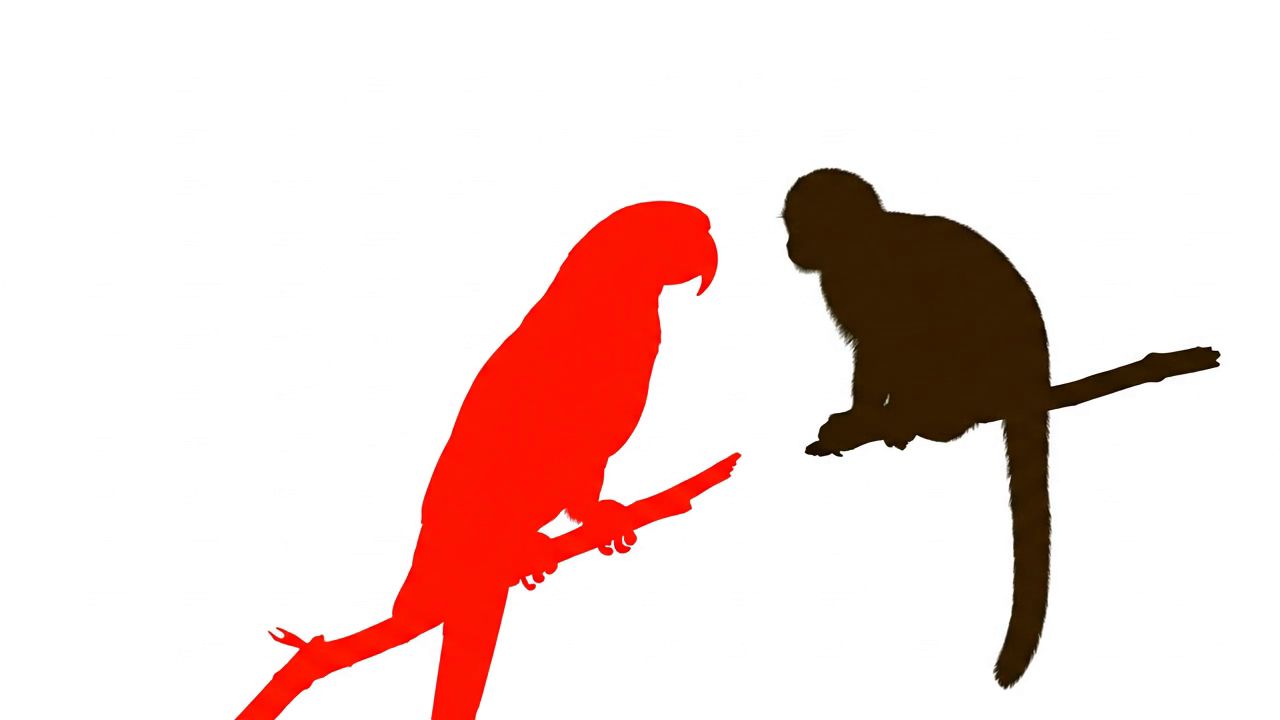}
        };
        \node[exampleframe] at ($(B2) + (\boxpad, -\boxtextheight)$) (i2a) {
                \includegraphics[width=\imagewidth]{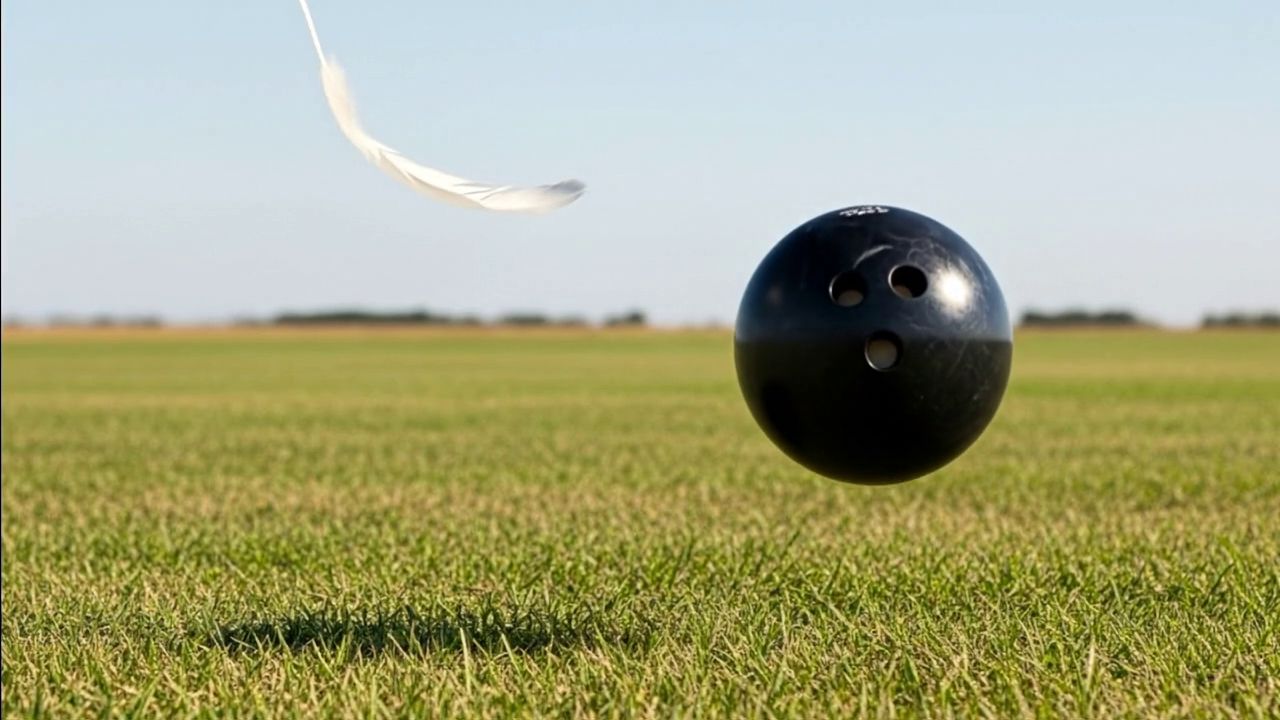}
        };
        \node[exampleframe] at ($(B2) + (\imagewidth + 3\boxpad, -\boxtextheight)$) (i2b) {
                \includegraphics[width=\imagewidth]{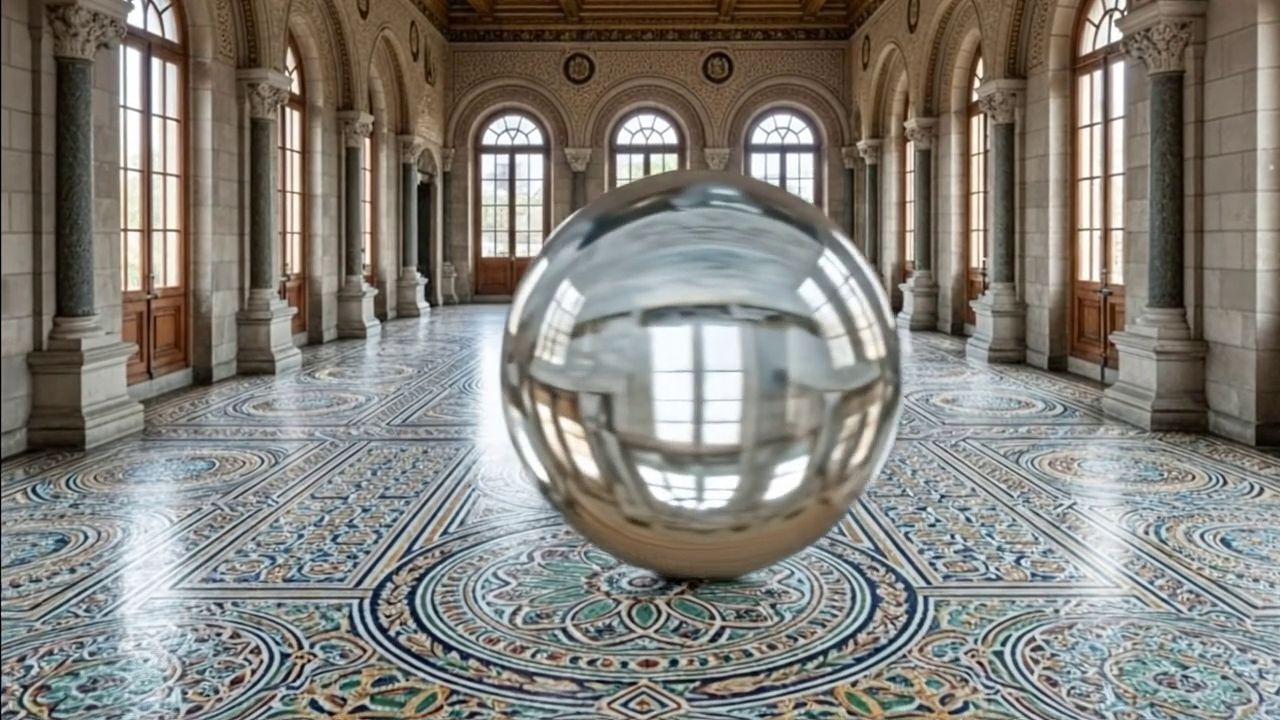}
        };
        \node[anchor=south west, white] at ($(B2) + (2\imagewidth + 4\boxpad, 10pt - \boxheight)$) () {$\cdots$};
        \node[exampleframe] at ($(B3) + (\boxpad, -\boxtextheight)$) (i3a) {
                \includegraphics[width=\imagewidth]{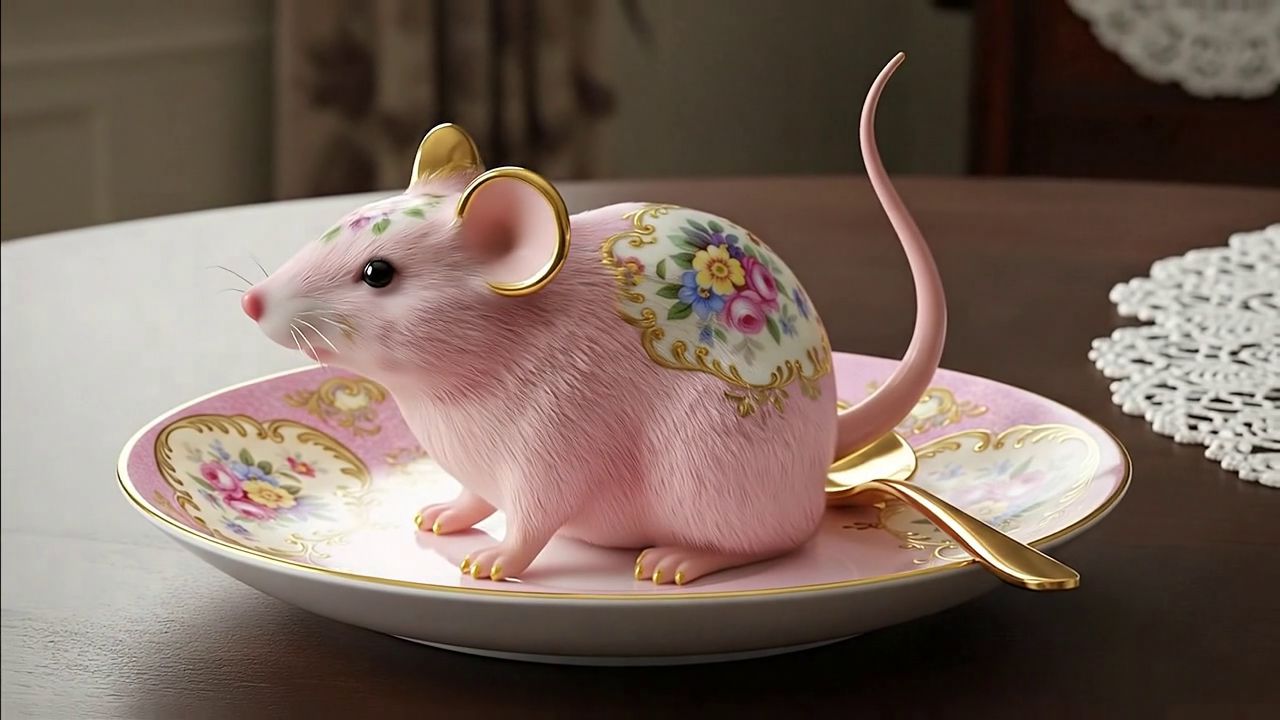}
        };
        \node[exampleframe] at ($(B3) + (\imagewidth + 3\boxpad, -\boxtextheight)$) (i3b) {
                \includegraphics[width=\imagewidth]{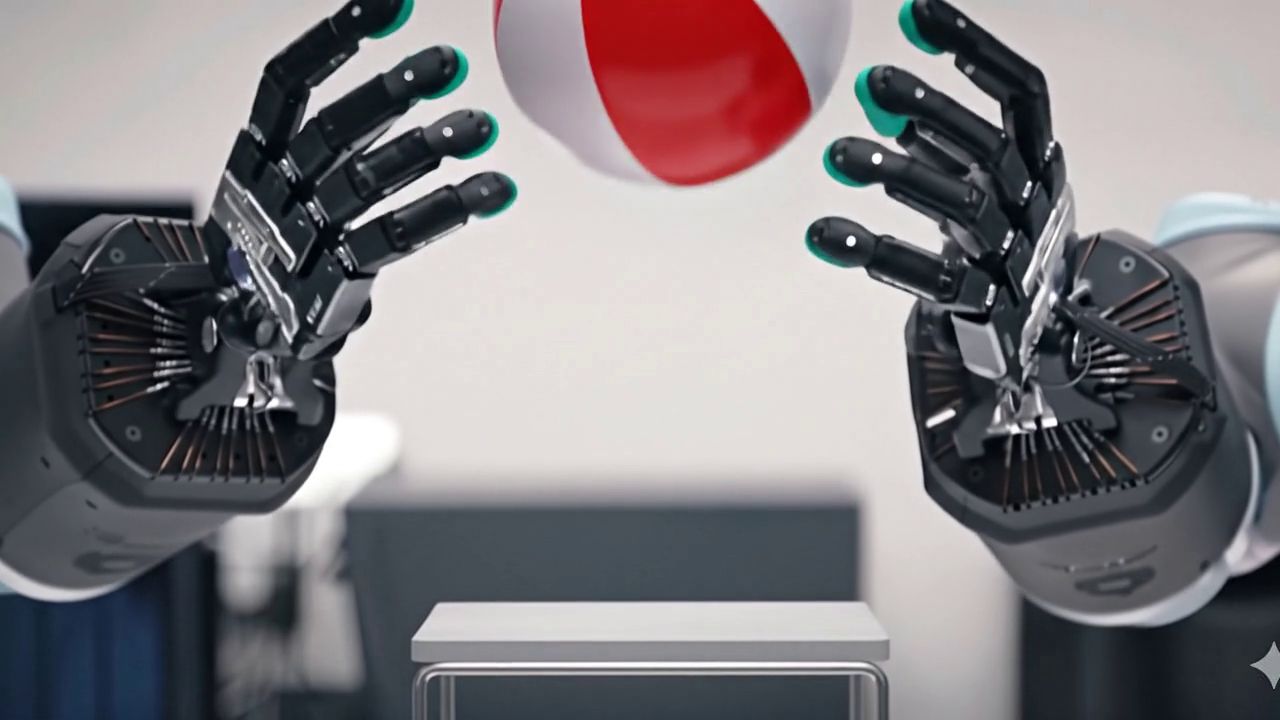}
        };
        \node[anchor=south west, white] at ($(B3) + (2\imagewidth + 4\boxpad, 10pt - \boxheight)$) () {$\cdots$};
        \node[exampleframe] at ($(B4) + (\boxpad, -\boxtextheight)$) (i4a) {
                \includegraphics[width=\imagewidth]{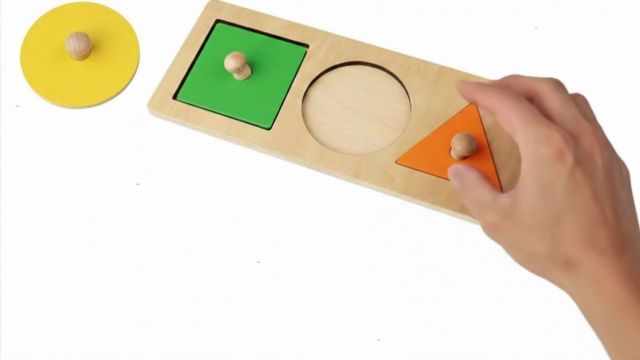}
        };
        \node[exampleframe] at ($(B4) + (\imagewidth + 3\boxpad, -\boxtextheight)$) (i4b) {
                \includegraphics[width=\imagewidth]{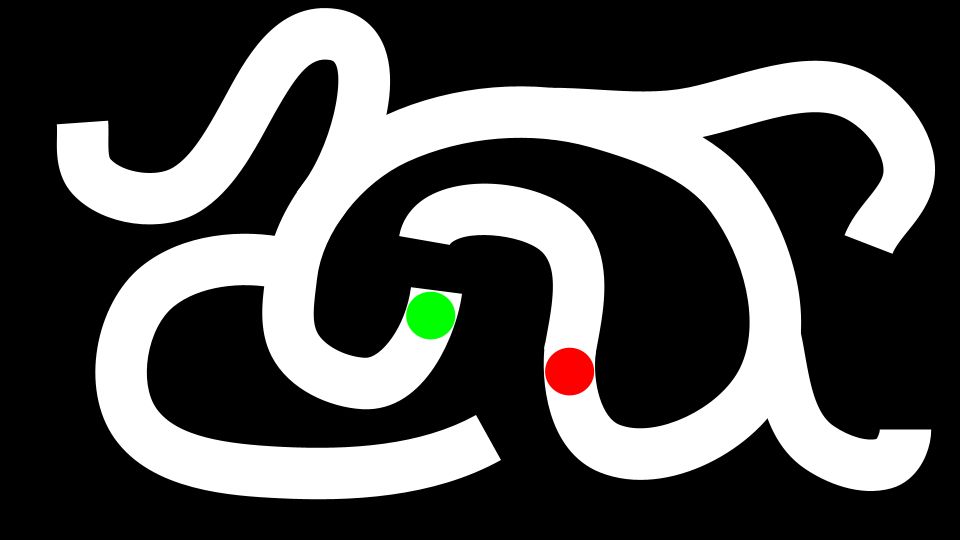}
        };

    \end{tikzpicture}
    }
    \\\vspace{5pt}
    \includegraphics[width=\linewidth]{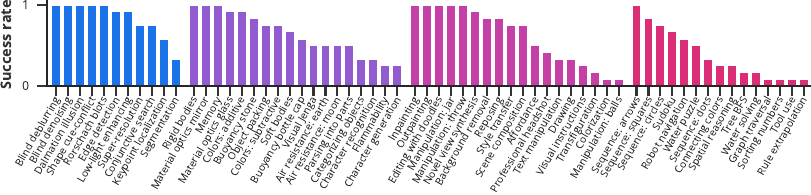}
    \caption{\textbf{A qualitative overview of Veo 3's zero-shot abilities}. The plot shows \veothree{}'s success rate across 12 samples as a rough estimate of model performance on 62~tasks across the vision stack. Tasks are described in \cref{sec:qualitative_results} and shown in \Cref{app:qualitative}. Find videos of all tasks on our \href{\websiteurl}{project page}.
    }
    \label{fig:overview}
\end{figure}

\section{Methods}
\label{sec:methods}
\paragraph{Approach and motivation} Our method is simple: We prompt \veo{}. This minimalist strategy is intentional, as it mirrors the transformation of NLP from task-specific fine-tuning or training to prompting a capable foundation model~\citep{wei2022chain,zhou2022large,liu2023pre}. Here, we adopt the same philosophy to explore the capabilities of \veothree{} as a general-purpose vision model.

\takeaway{In NLP, prompting replaced task-specific training or adaptation. A similar paradigm shift is on the horizon in machine vision, facilitated by video models.}

\paragraph{Video generation} For each task, we query the publicly available \veotwo{} or \veothree{} models via Google Cloud's Vertex AI API. We prompt the model with an initial input image (which the model uses as the first frame) and a text instruction. The models then generate a 16:9 video at 720p resolution, 24 FPS, for a duration of 8s. \veothree{} has model ID \texttt{\small veo-3.0-generate-preview} and \veotwo{} model ID \texttt{\small veo-2.0-generate-001}. According to the Vertex documentation~\citep{GoogleCloudVertexAIPromptRewriter}, the API uses an LLM-based prompt rewriter. This means that for some tasks, the solution is likely to come from the LLM instead of the video (e.g., \cref{fig:sudoku}: Sudoku). We treat the system (rewriter and video generator) as a single black-box entity. However, to isolate the video model's reasoning abilities, we verified that a standalone LLM (Gemini 2.5 Pro~\citep{comanici2025gemini}) could not reliably solve key tasks (\cref{fig:robot_navigation}: Robot navigation, \cref{subsec:maze}: Maze solving, \cref{subsec:visual_symmetry}: Visual symmetry) from the input image alone.

\paragraph{Why Veo?} The core argument of this paper---that video models are zero-shot learners and reasoners---can be supported by demonstrating success on \emph{any} sufficiently capable model. We choose Veo because it has consistently ranked high on \texttt{text2video} and \texttt{image2video} leaderboards~\citep{lmarena_leaderboard}. Unless noted otherwise, our figures are generated with \veothree{}. To provide a sense of how rapidly performance is improving, our quantitative analyses compare \veothree{} with its predecessor, \veotwo{}, released roughly within half a year of each other: \veotwo{} was announced in December 2024 and released in April 2025~\citep{Veo2Announcement, Veo2Launch}, while \veothree{} was announced in May 2025 and released in July 2025~\citep{Veo3Announcement, Veo3Launch}.

%% file: sections/03_qualitative.tex
\section{Qualitative results: Sparks of visual intelligence?}
\label{sec:qualitative_results}
\begin{figure}[t]
    \centering
    \resizebox{\linewidth}{!}{%
    \begin{tikzpicture}
        \newlength{\quadrantw}
        \pgfmathsetlength{\quadrantw}{\tikzwidth}
        \newlength{\imw}
        \pgfmathsetlength{\imw}{(\quadrantw - 2\boxpad - 6\boxgap) / 6}
        \newlength{\imh}
        \pgfmathsetlength{\imh}{\imw / 16 * 9}
        \newlength{\quadranth}
        \pgfmathsetlength{\quadranth}{\imh + \boxpad + \boxtextheight}
        \setlength{\boxtextheight}{11pt}

        \tikzset{
            boxcaption/.style={
                font=\fontsize{8}{8}\opensans,
                text=white,
                anchor=west
            }
        }
        \tikzset{
            exampleframe/.style={
                anchor=north west,
                inner sep=0pt,
            }
        }
        \coordinate (B1) at ($(0, 4\quadranth + 6\boxgap)$);
        \fill[perceptioncol] (B1) -- ++(\quadrantw, 0) -- ++(0, -\quadranth) -- ++(-\quadrantw, 0);
        \node[boxcaption] at ($(B1) + (-1.5pt, -6pt)$) () {\textbf{Perception} -- superresolution};
        \node[exampleframe] at ($(B1) + (\boxpad, -\boxtextheight)$) () {
                \includegraphics[width=\imw]{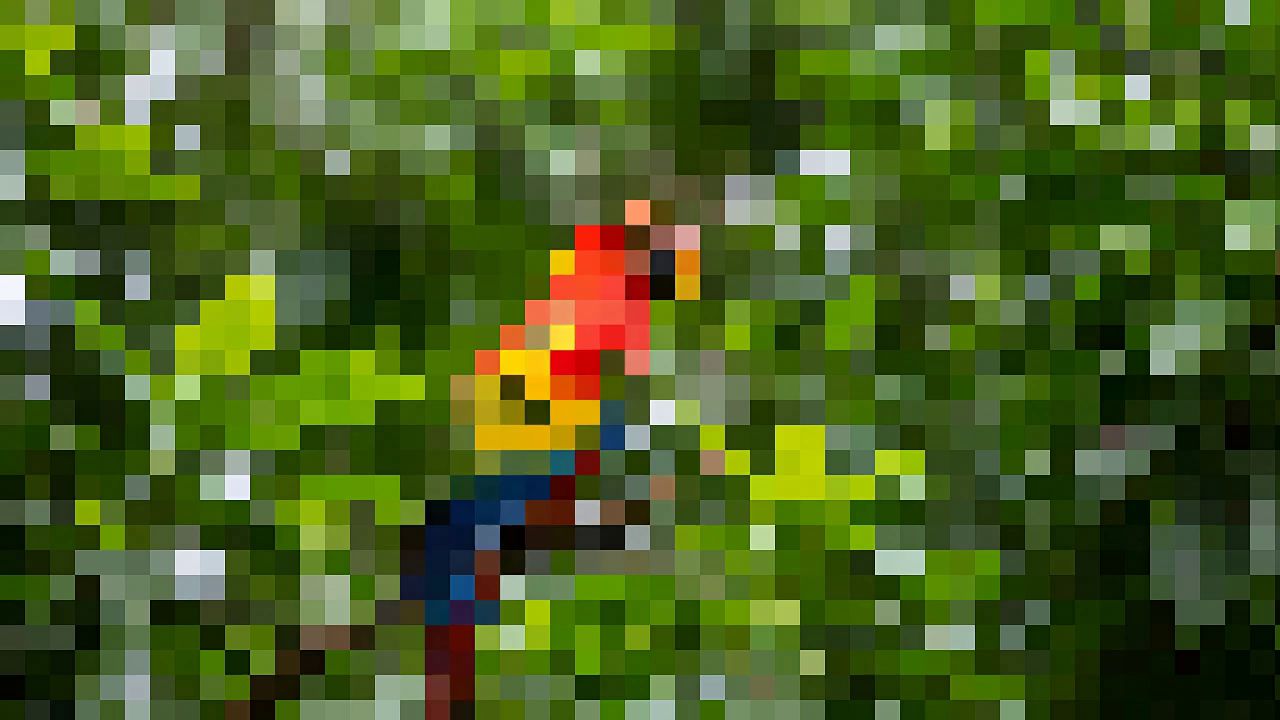}
        };
        \node[exampleframe] at ($(B1) + (\boxpad + \imw + \boxgap, -\boxtextheight)$) () {
                \includegraphics[width=\imw]{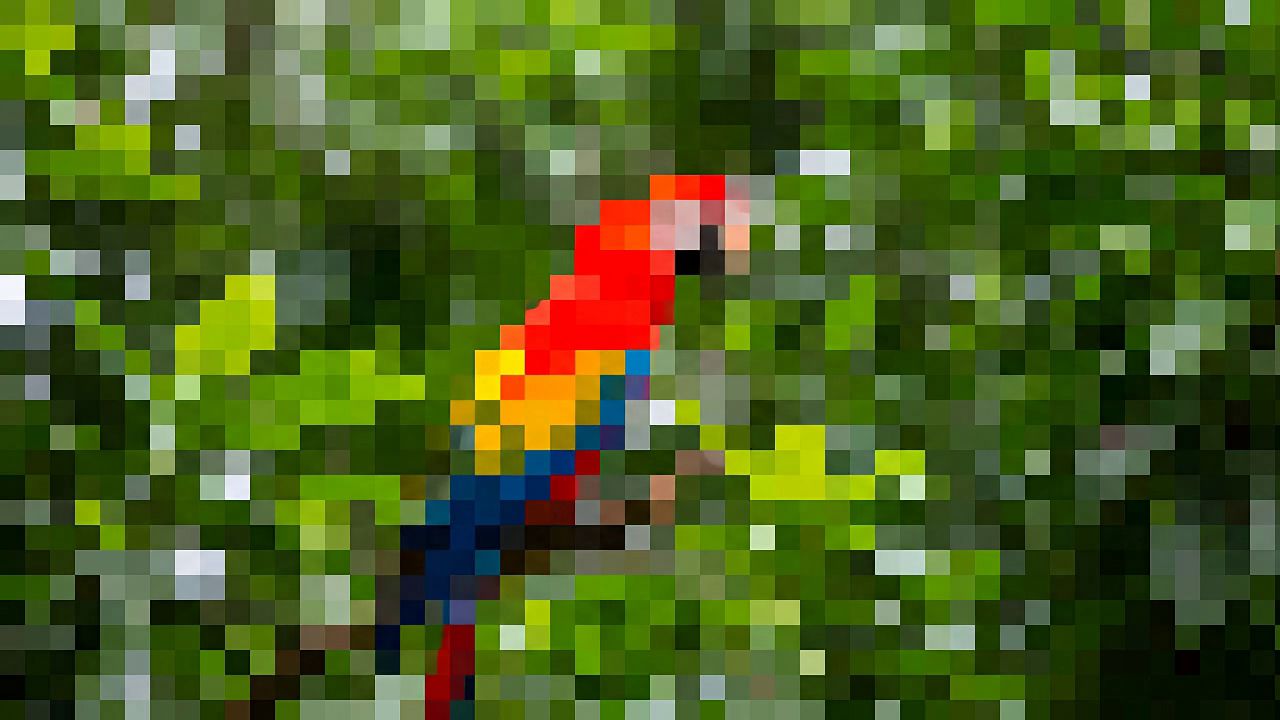}
        };
        \node[exampleframe] at ($(B1) + (\boxpad + 2\imw + 2\boxgap, -\boxtextheight)$) () {
                \includegraphics[width=\imw]{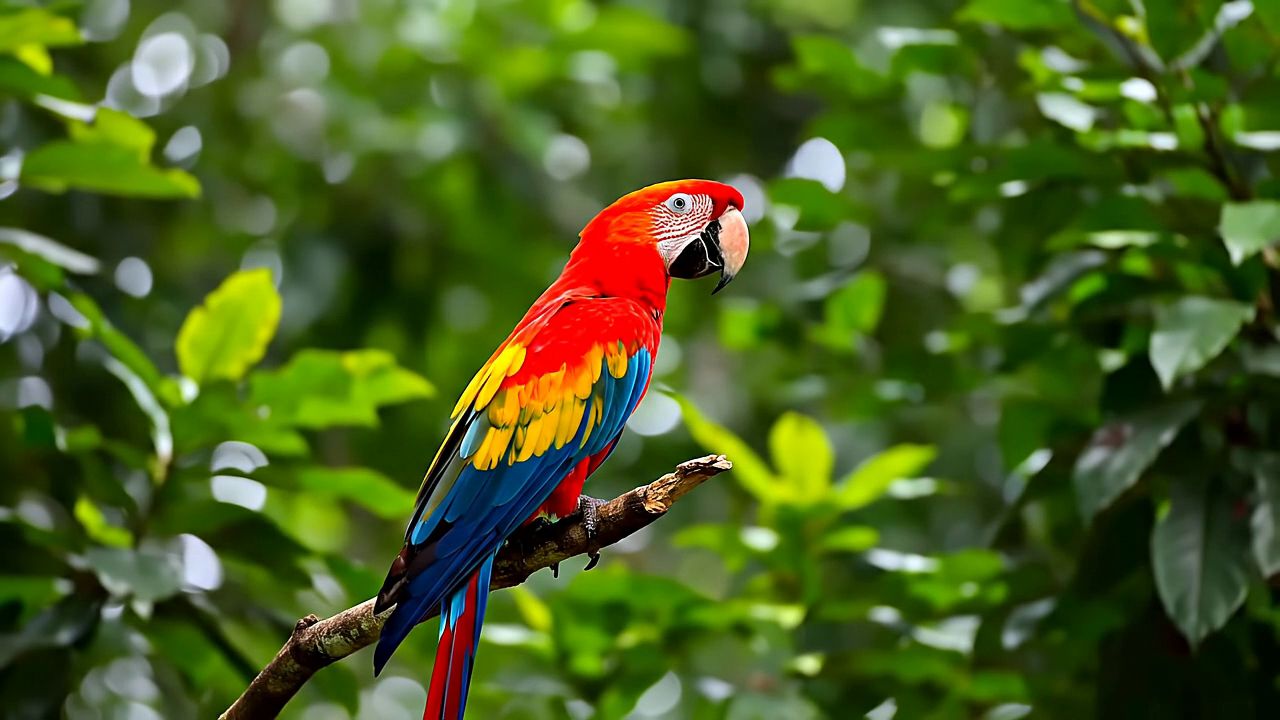}
        };
        \node[boxcaption] at ($(B1) + (\boxpad + 3\imw + 4\boxgap -1.5pt, -6pt)$) () {\textbf{Perception} -- conjunctive search / binding problem};
        \node[exampleframe] at ($(B1) + (\boxpad + 3\imw + 4\boxgap, -\boxtextheight)$) () {
                \includegraphics[width=\imw]{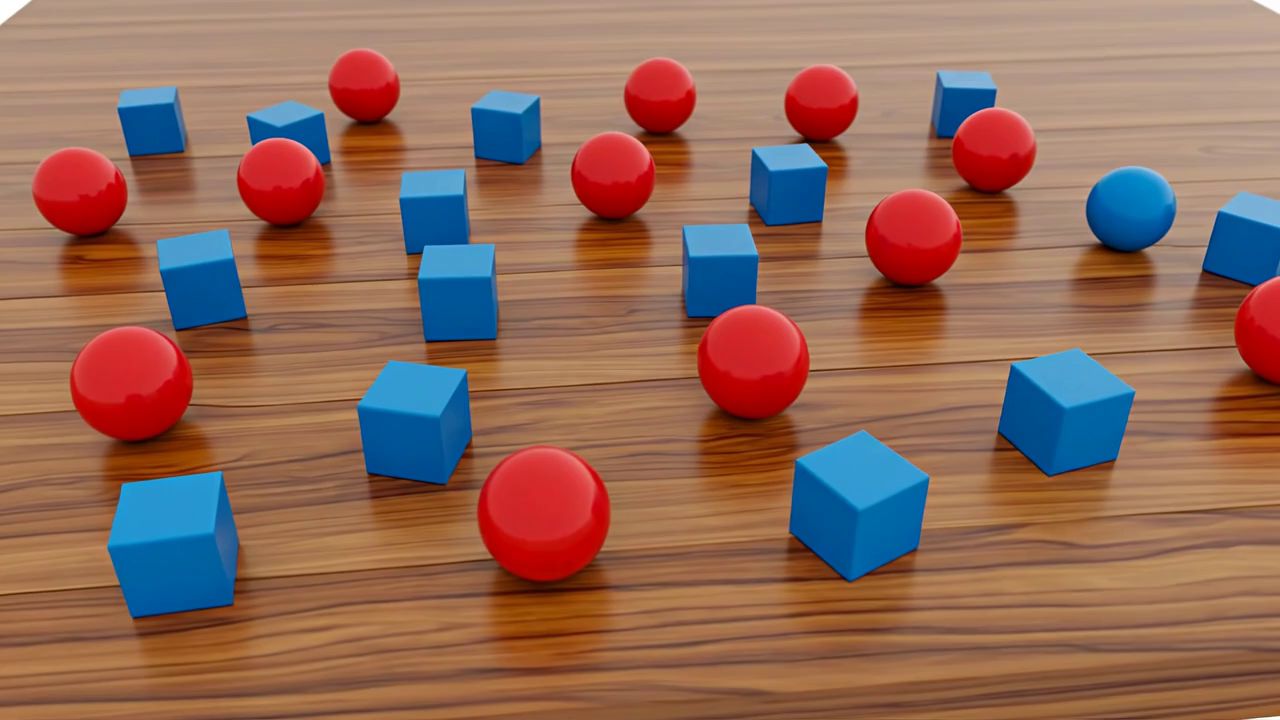}
        };
        \node[exampleframe] at ($(B1) + (\boxpad + 4\imw + 5\boxgap, -\boxtextheight)$) () {
                \includegraphics[width=\imw]{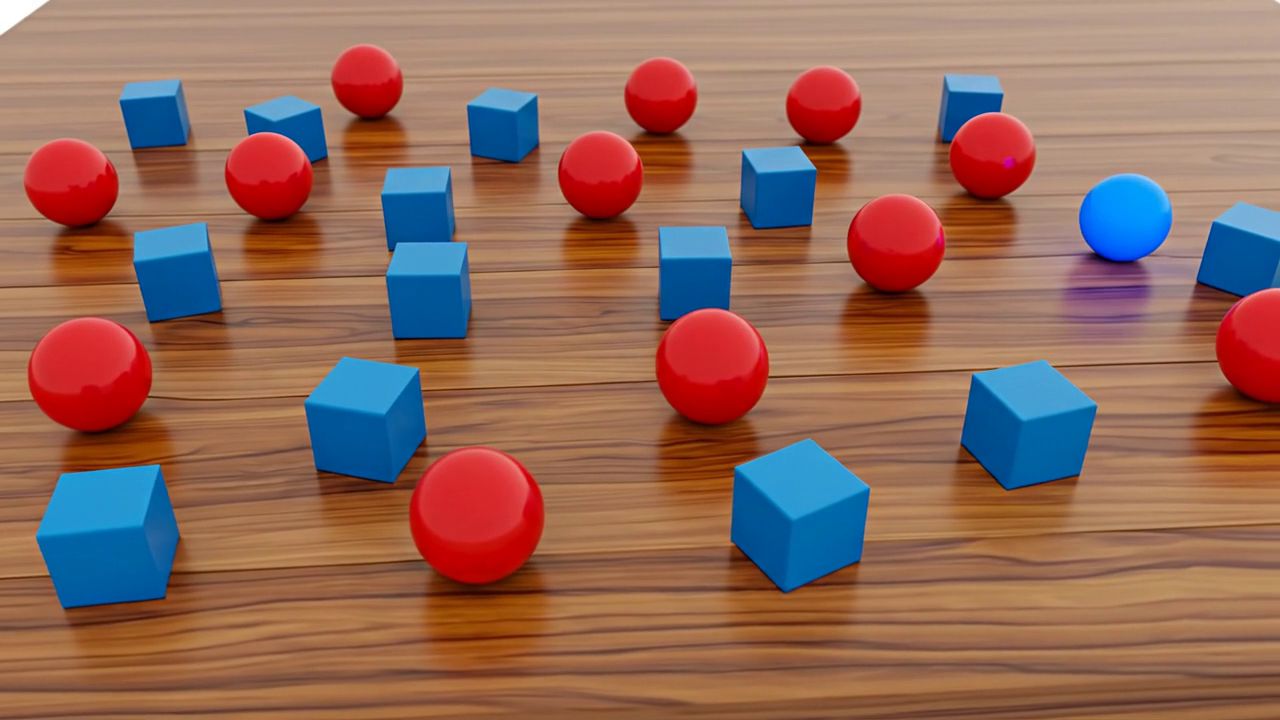}
        };
        \node[exampleframe] at ($(B1) + ((\boxpad + 5\imw + 6\boxgap, -\boxtextheight)$) () {
                \includegraphics[width=\imw]{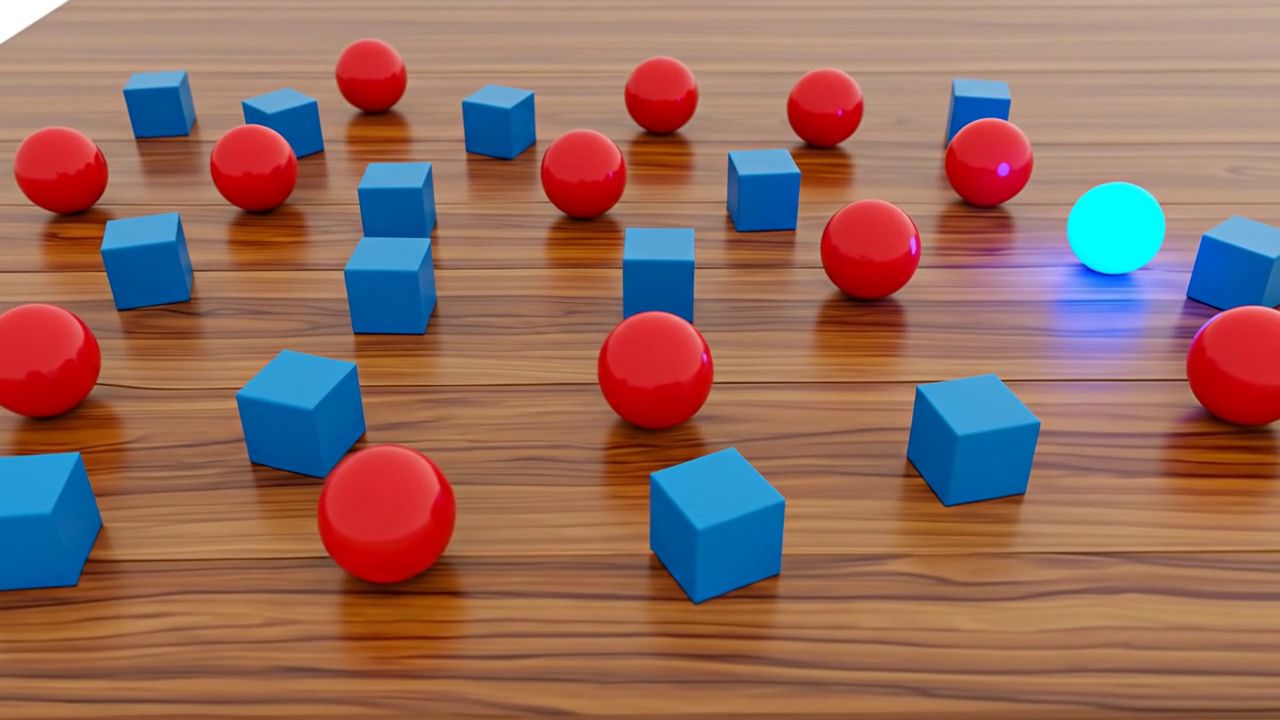}
        };

        \coordinate (B2) at ($(0, 3\quadranth + 4\boxgap)$);
        \fill[modelingcol] (B2) -- ++(\quadrantw, 0) -- ++(0, -\quadranth) -- ++(-\quadrantw, 0);
        \node[boxcaption] at ($(B2) + (-1.5pt, -6pt)$) () {\textbf{Modeling} -- buoyancy};
        \node[exampleframe] at ($(B2) + (\boxpad, -\boxtextheight)$) () {
                \includegraphics[width=\imw]{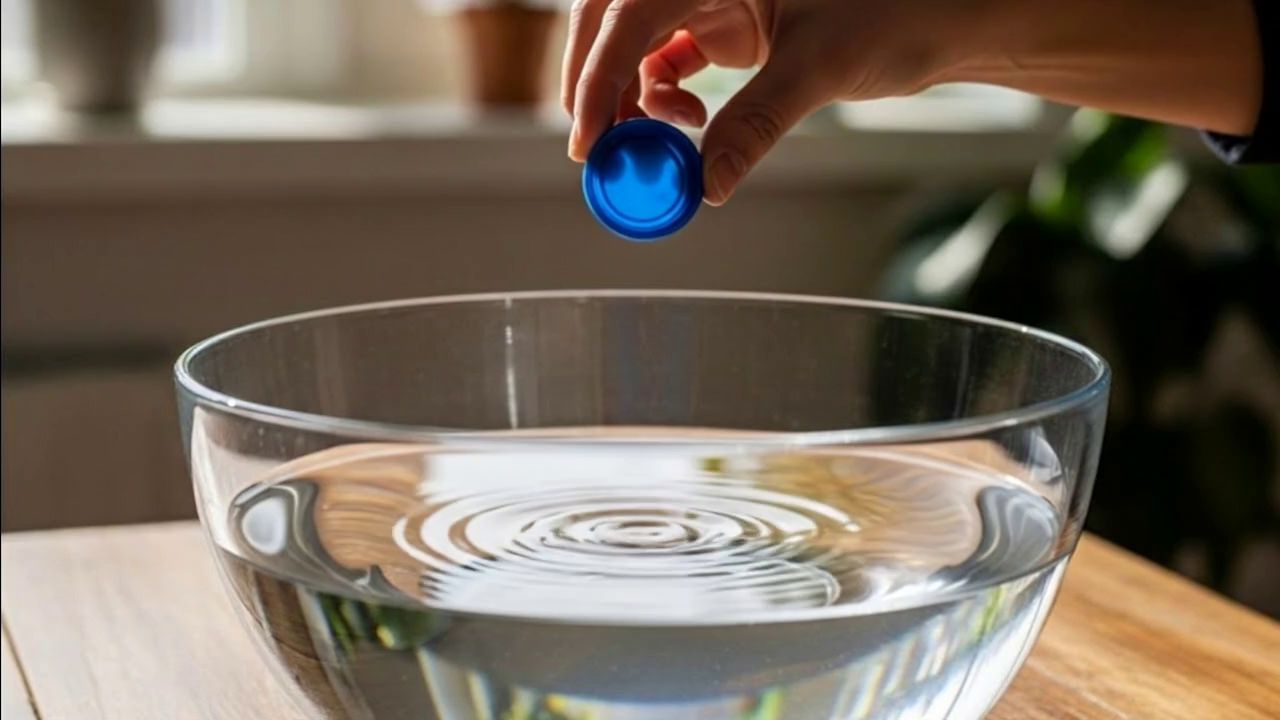}
        };
        \node[exampleframe] at ($(B2) + (\boxpad + \imw + \boxgap, -\boxtextheight)$) () {
                \includegraphics[width=\imw]{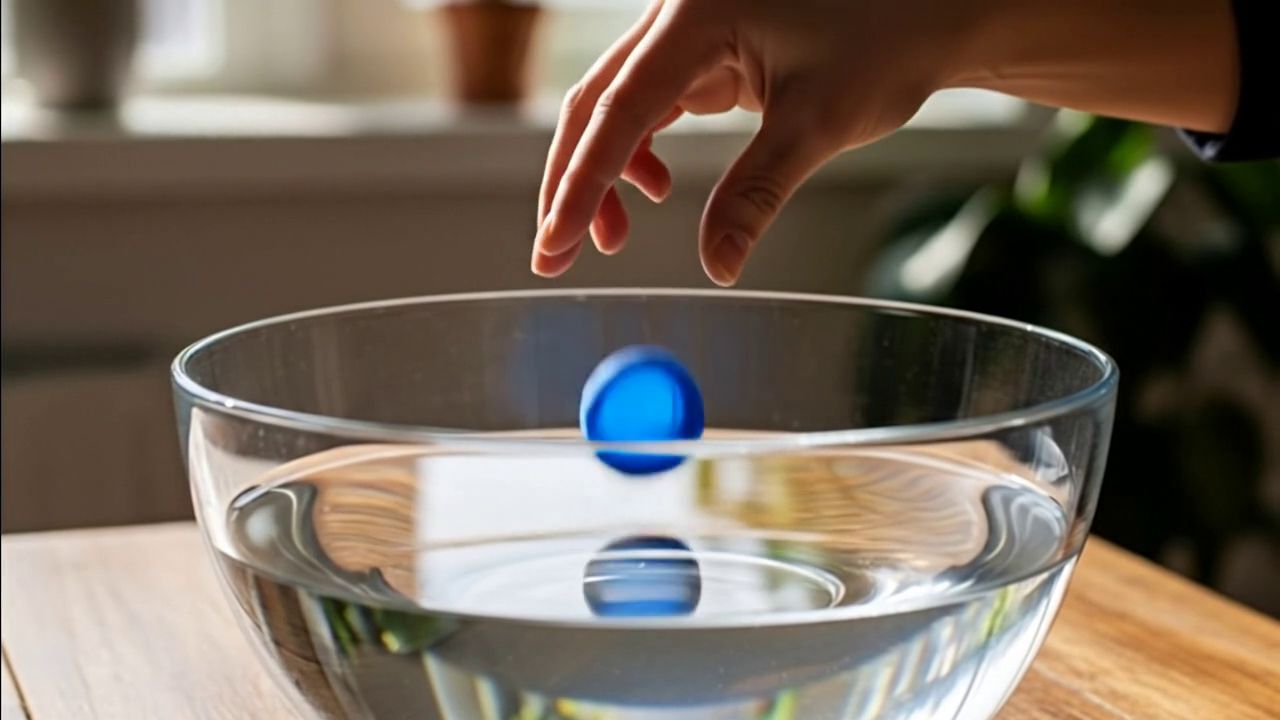}
        };
        \node[exampleframe] at ($(B2) + (\boxpad + 2\imw + 2\boxgap, -\boxtextheight)$) () {
                \includegraphics[width=\imw]{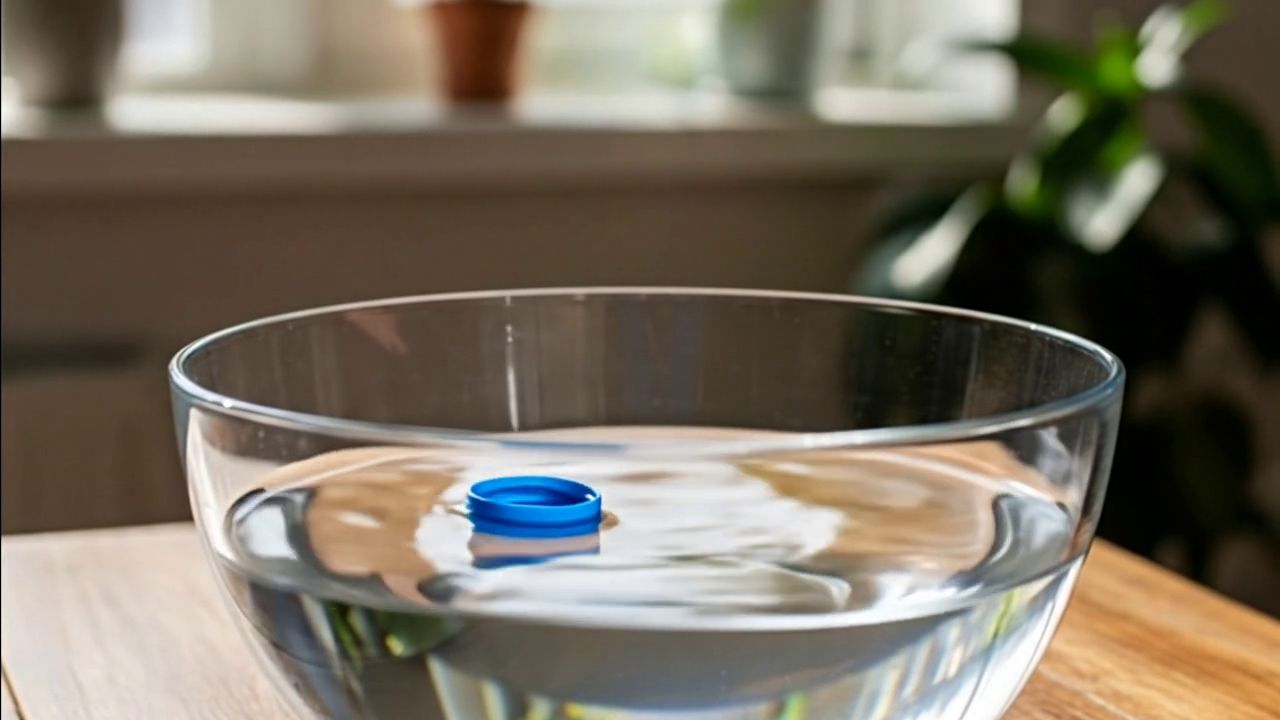}
        };
        \node[boxcaption] at ($(B2) + (\boxpad + 3\imw + 4\boxgap -1.5pt, -6pt)$) () {\textbf{Modeling} -- memory of world states};
        \node[exampleframe] at ($(B2) + (\boxpad + 3\imw + 4\boxgap, -\boxtextheight)$) () {
                \includegraphics[width=\imw]{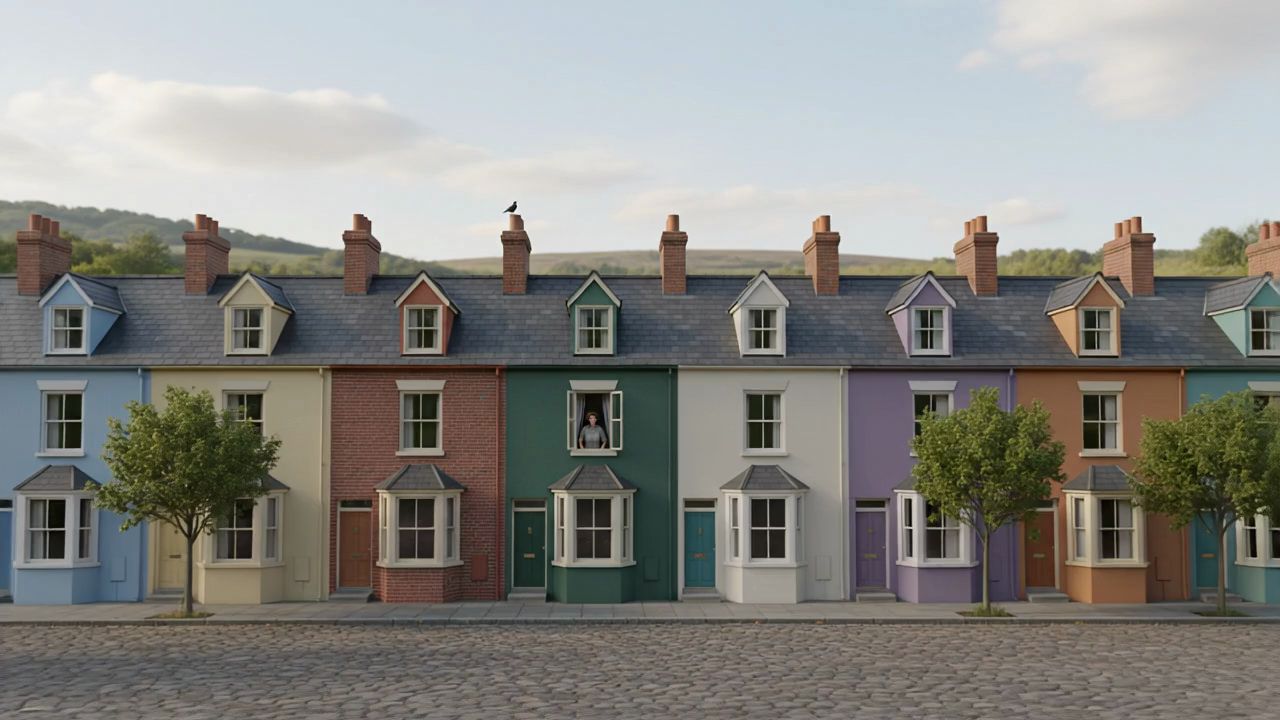}
        };
        \node[exampleframe] at ($(B2) + (\boxpad + 4\imw + 5\boxgap, -\boxtextheight)$) () {
                \includegraphics[width=\imw]{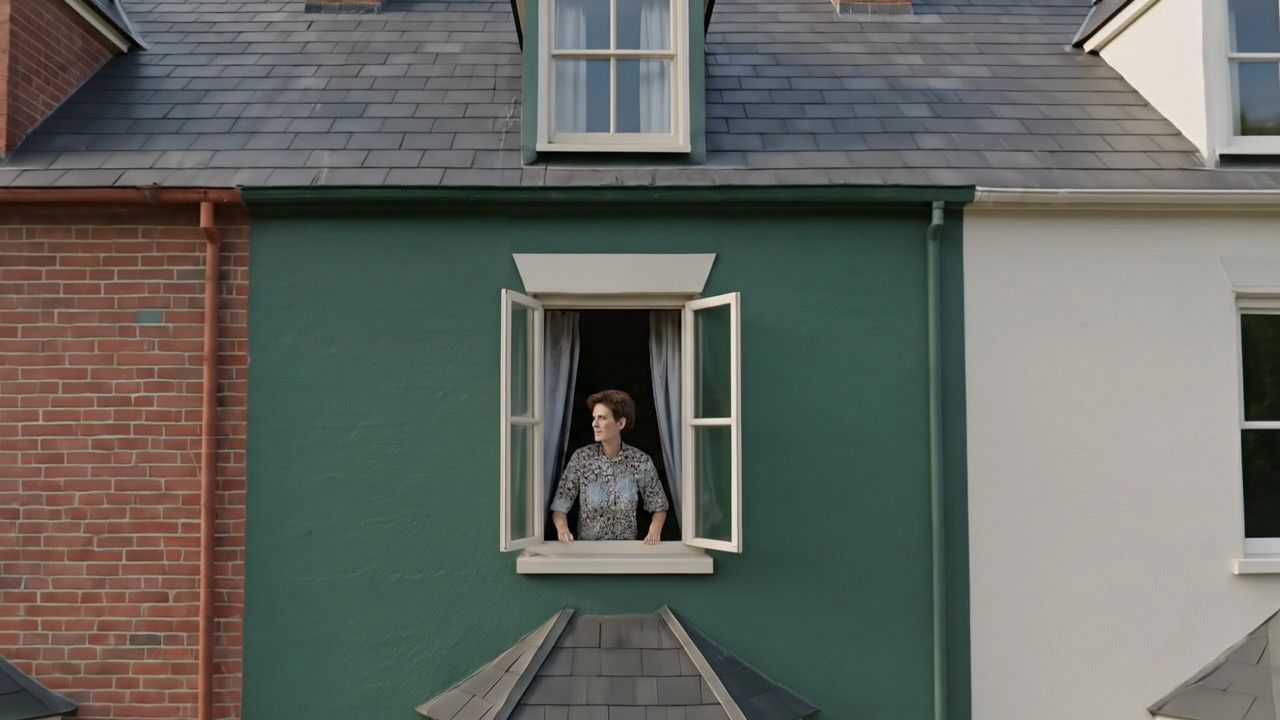}
        };
        \node[exampleframe] at ($(B2) + ((\boxpad + 5\imw + 6\boxgap, -\boxtextheight)$) () {
                \includegraphics[width=\imw]{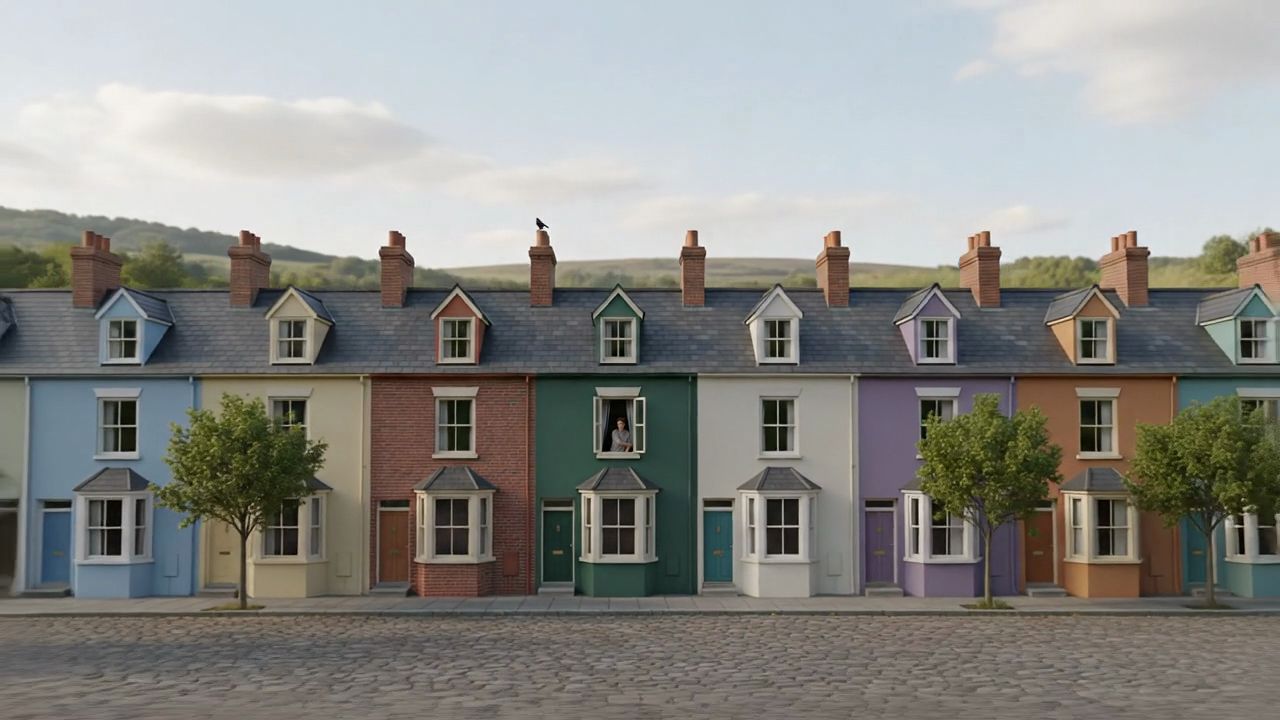}
        };

        \coordinate (B3) at ($(0, 2\quadranth + 2\boxgap)$);
        \fill[manipulationcol] (B3) -- ++(\quadrantw, 0) -- ++(0, -\quadranth) -- ++(-\quadrantw, 0);
        \node[boxcaption] at ($(B3) + (-1.5pt, -6pt)$) () {\textbf{Manipulation} -- 3D-aware reposing};
        \node[exampleframe] at ($(B3) + (\boxpad, -\boxtextheight)$) () {
                \includegraphics[width=\imw]{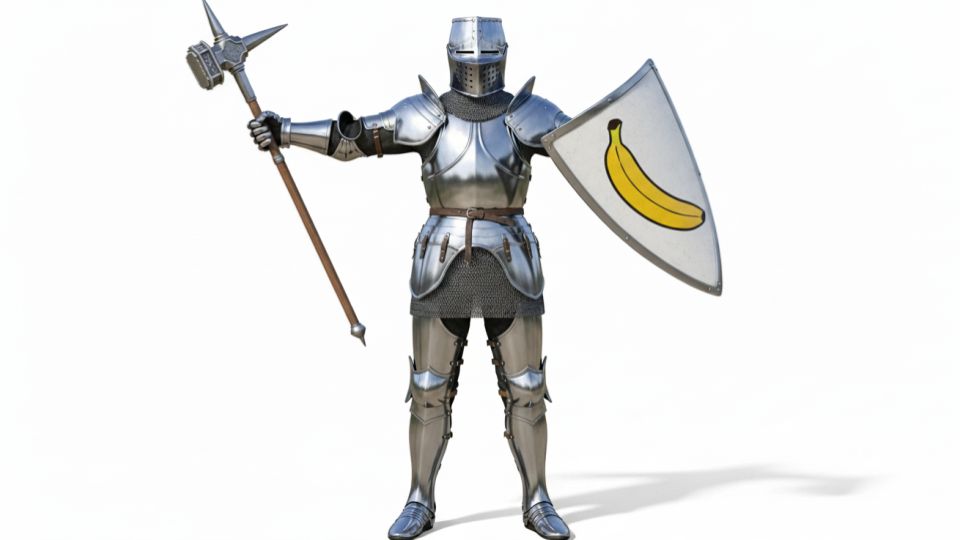}
        };
        \node[exampleframe] at ($(B3) + (\boxpad + \imw + \boxgap, -\boxtextheight)$) () {
                \includegraphics[width=\imw]{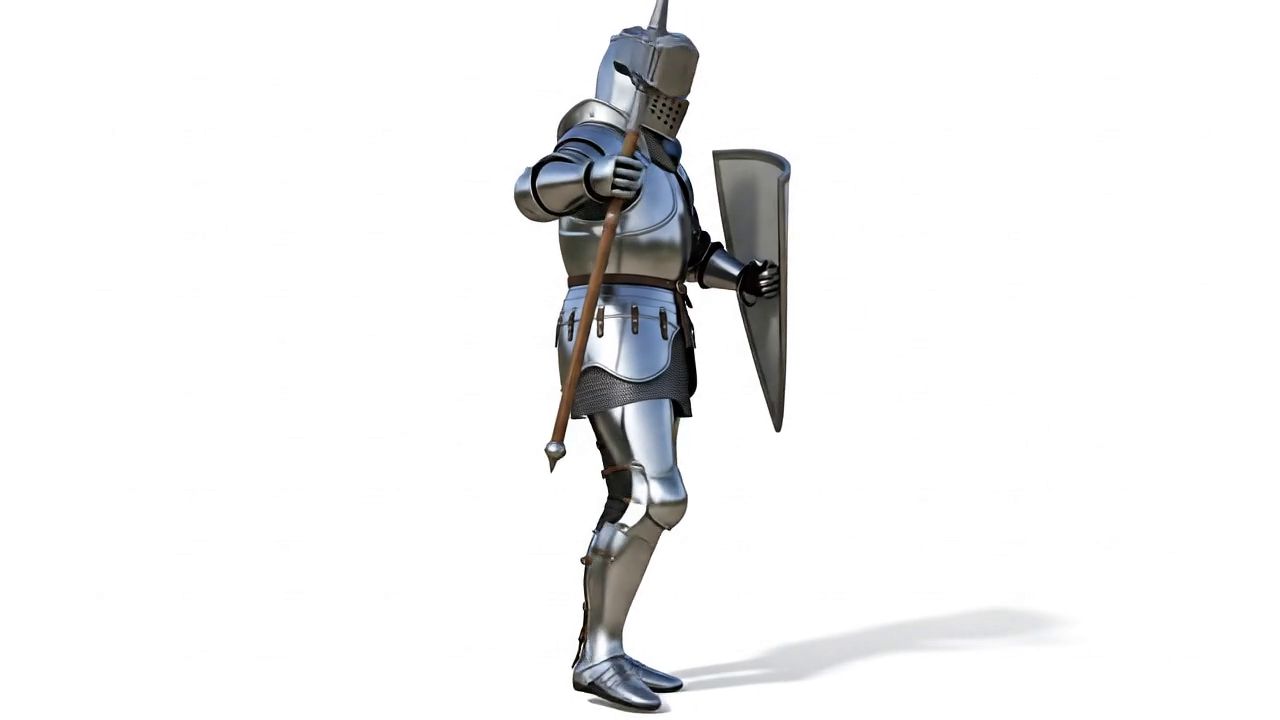}
        };
        \node[exampleframe] at ($(B3) + (\boxpad + 2\imw + 2\boxgap, -\boxtextheight)$) () {
                \includegraphics[width=\imw]{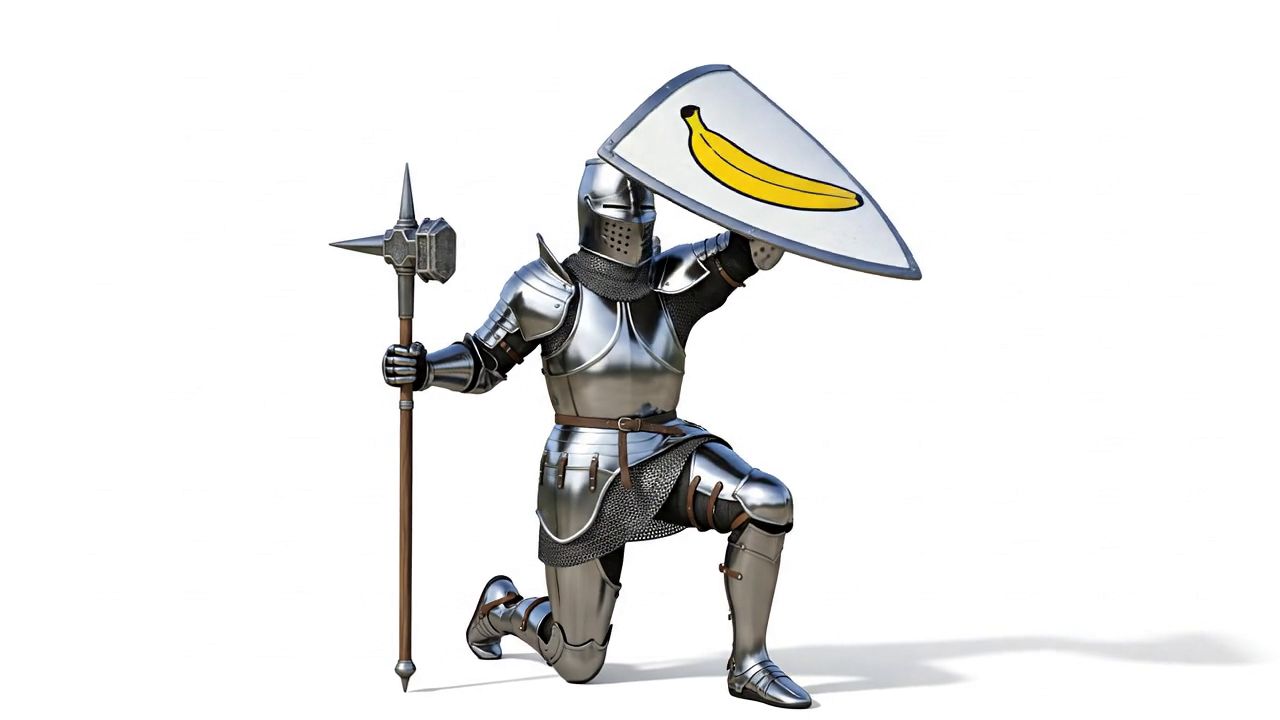}
        };
        \node[boxcaption] at ($(B3) + (\boxpad + 3\imw + 4\boxgap -1.5pt, -6pt)$) () {\textbf{Manipulation} -- jar opening};
        \node[exampleframe] at ($(B3) + (\boxpad + 3\imw + 4\boxgap, -\boxtextheight)$) () {
                \includegraphics[width=\imw]{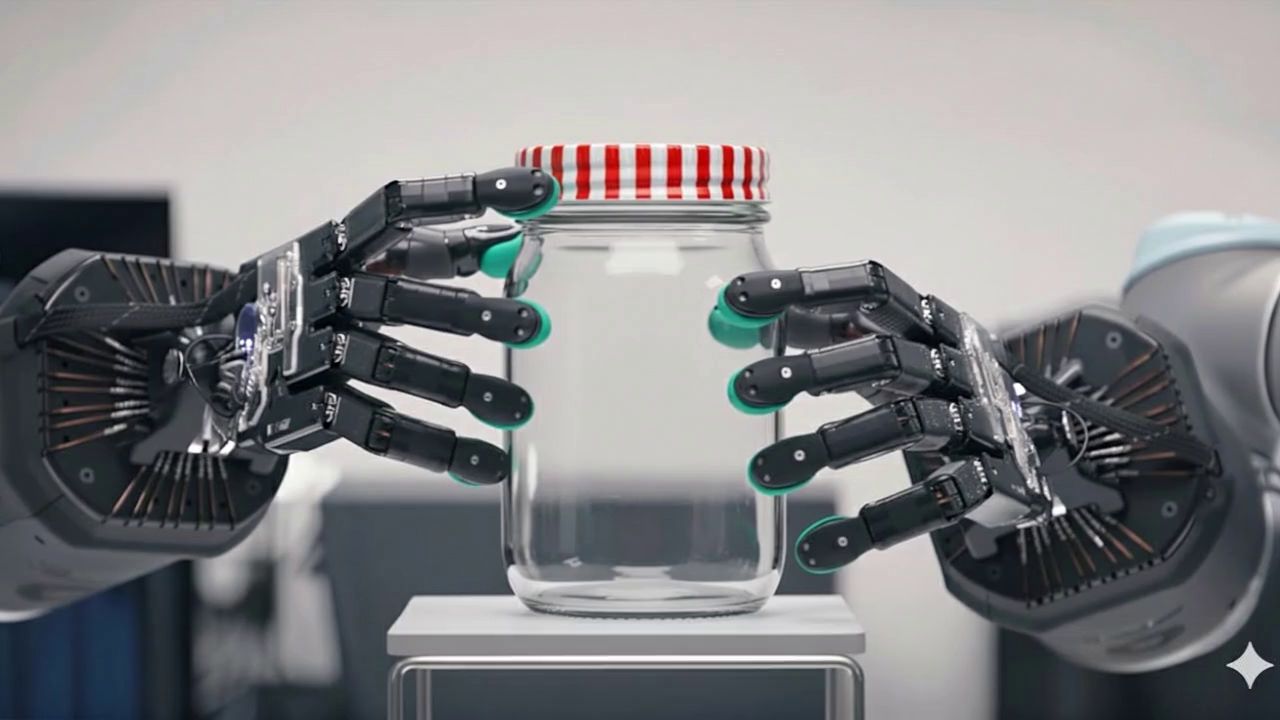}
        };
        \node[exampleframe] at ($(B3) + (\boxpad + 4\imw + 5\boxgap, -\boxtextheight)$) () {
                \includegraphics[width=\imw]{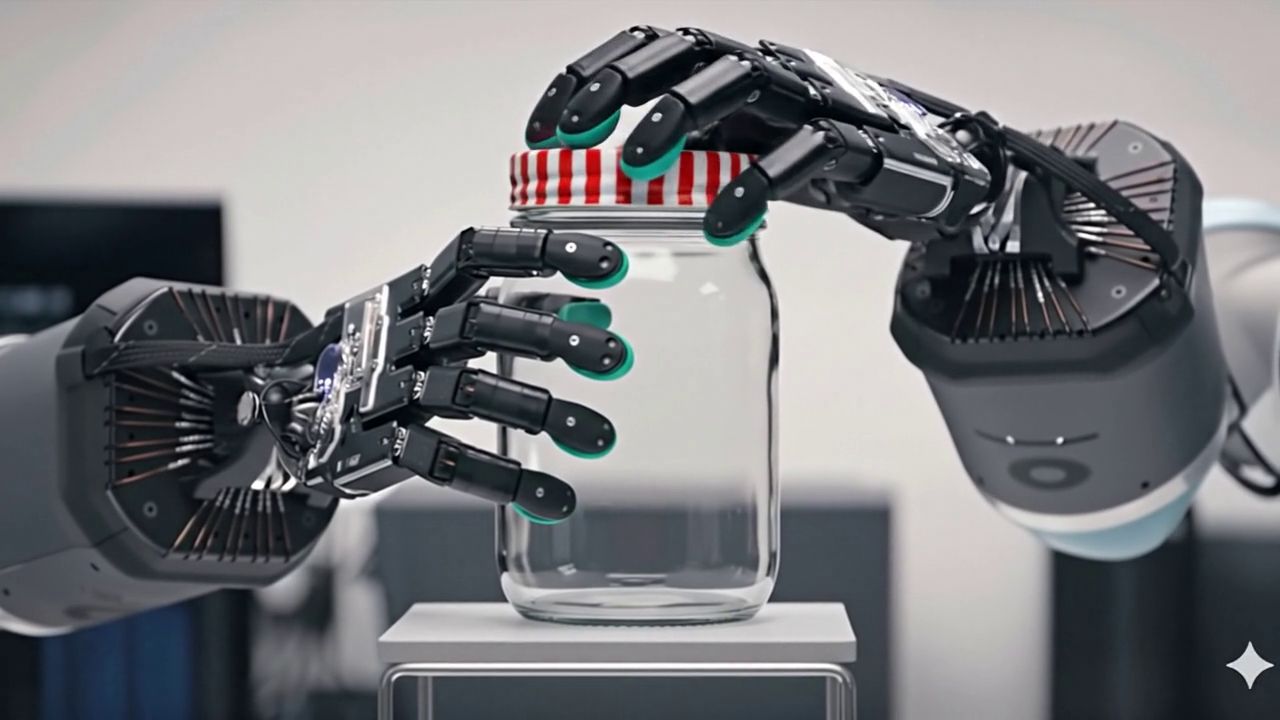}
        };
        \node[exampleframe] at ($(B3) + ((\boxpad + 5\imw + 6\boxgap, -\boxtextheight)$) () {
                \includegraphics[width=\imw]{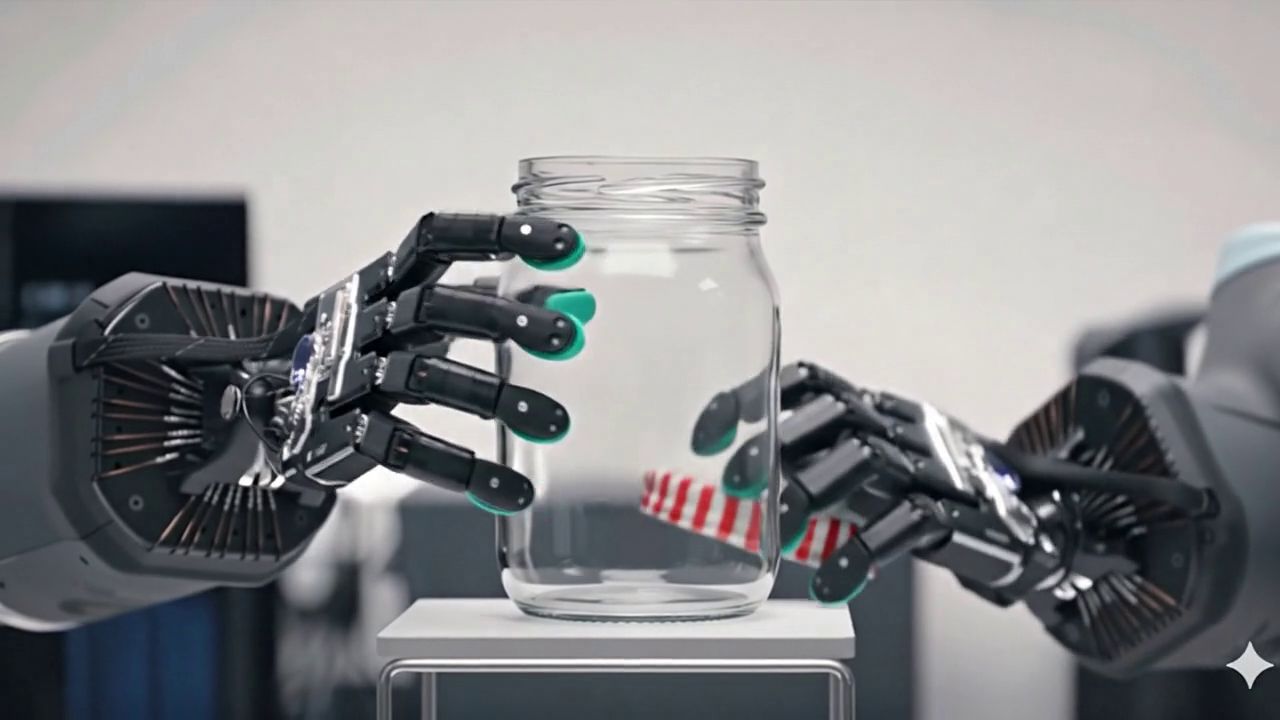}
        };

        \coordinate (B4) at (0, \quadranth);
        \fill[reasoningcol] (B4) -- ++(\quadrantw, 0) -- ++(0, -\quadranth) -- ++(-\quadrantw, 0);
        \node[boxcaption] at ($(B4) + (-1.5pt, -6pt)$) () {\textbf{Reasoning} -- robot navigation};
        \node[exampleframe] at ($(B4) + (\boxpad, -\boxtextheight)$) () {
                \includegraphics[width=\imw]{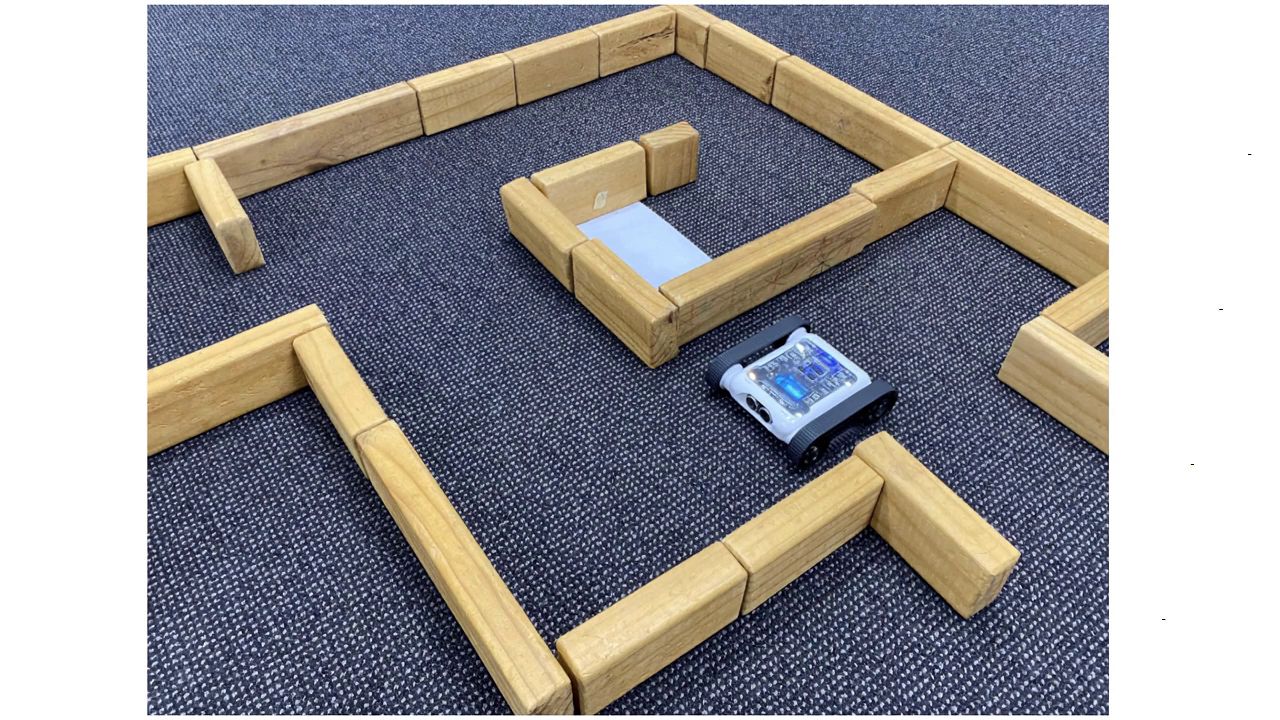}
        };
        \node[exampleframe] at ($(B4) + (\boxpad + \imw + \boxgap, -\boxtextheight)$) () {
                \includegraphics[width=\imw]{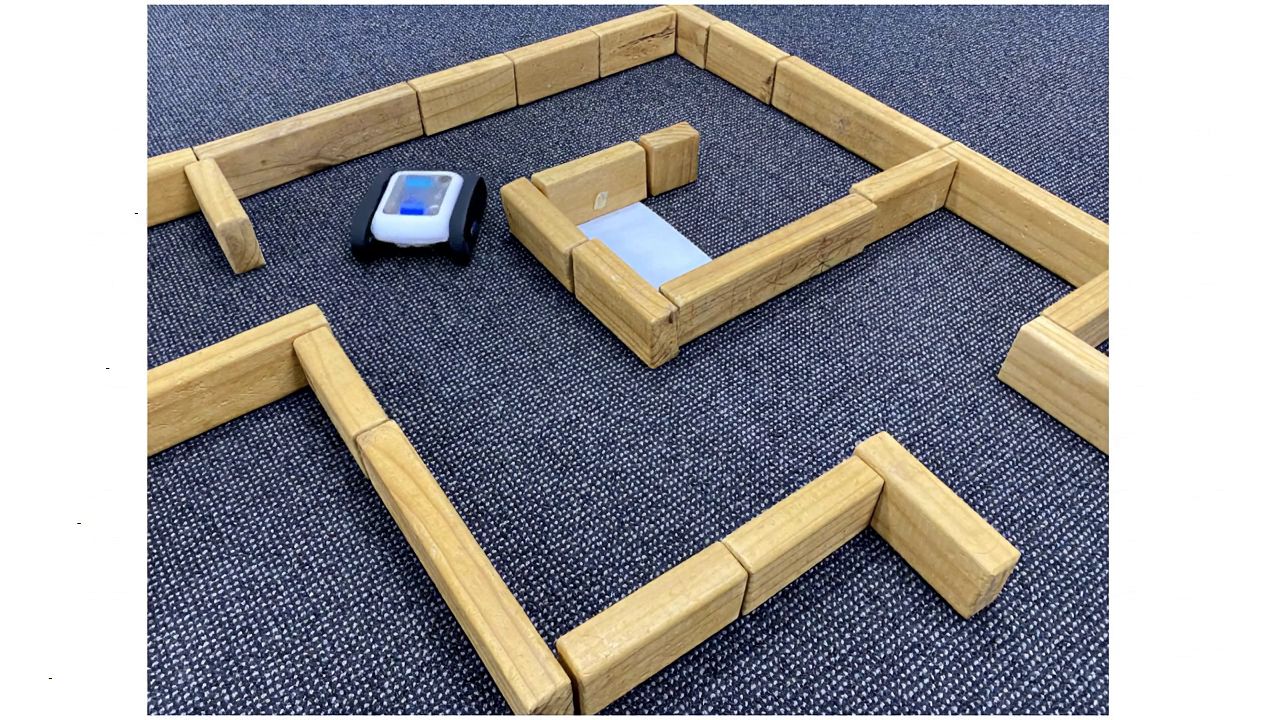}
        };
        \node[exampleframe] at ($(B4) + (\boxpad + 2\imw + 2\boxgap, -\boxtextheight)$) () {
                \includegraphics[width=\imw]{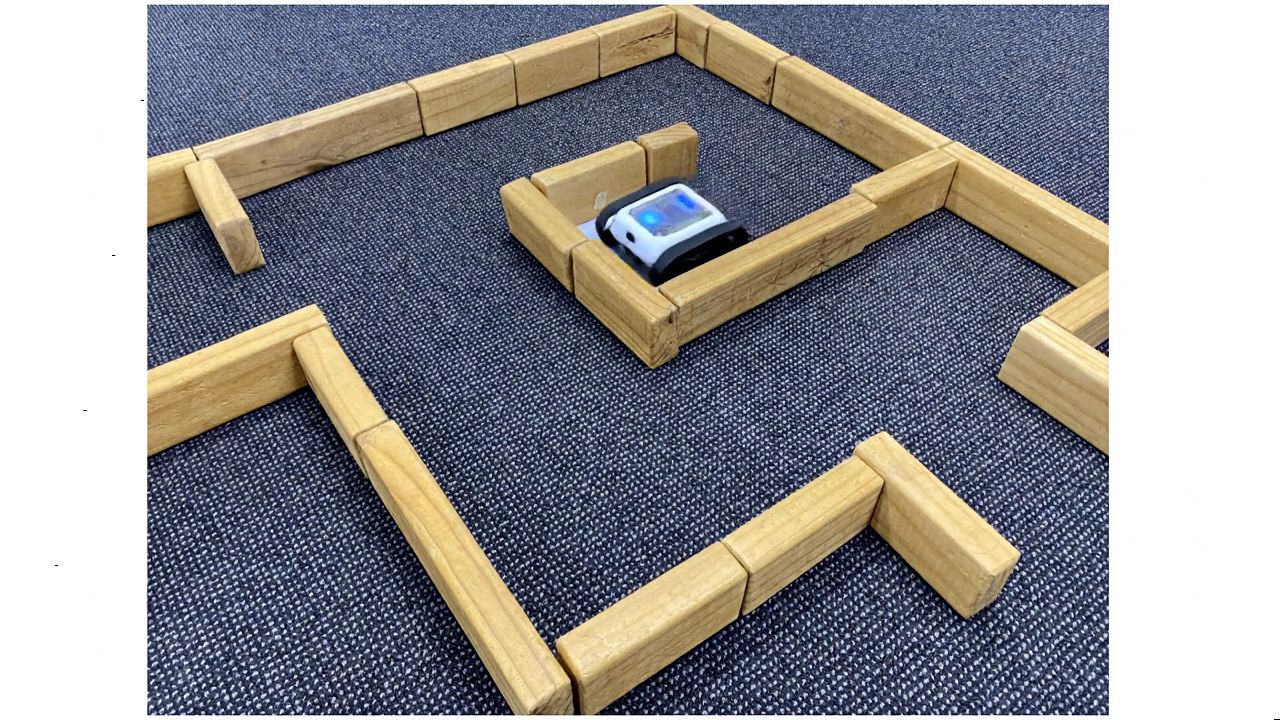}
        };
        \node[boxcaption] at ($(B4) + (\boxpad + 3\imw + 4\boxgap -1.5pt, -6pt)$) () {\textbf{Reasoning} -- rule extrapolation};
        \node[exampleframe] at ($(B4) + (\boxpad + 3\imw + 4\boxgap, -\boxtextheight)$) () {
                \includegraphics[width=\imw]{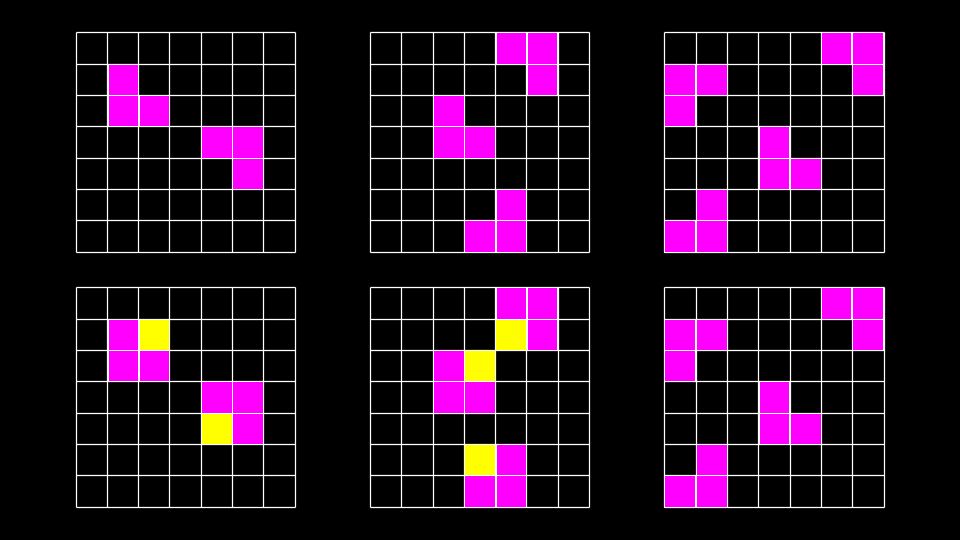}
        };
        \node[exampleframe] at ($(B4) + (\boxpad + 4\imw + 5\boxgap, -\boxtextheight)$) () {
                \includegraphics[width=\imw]{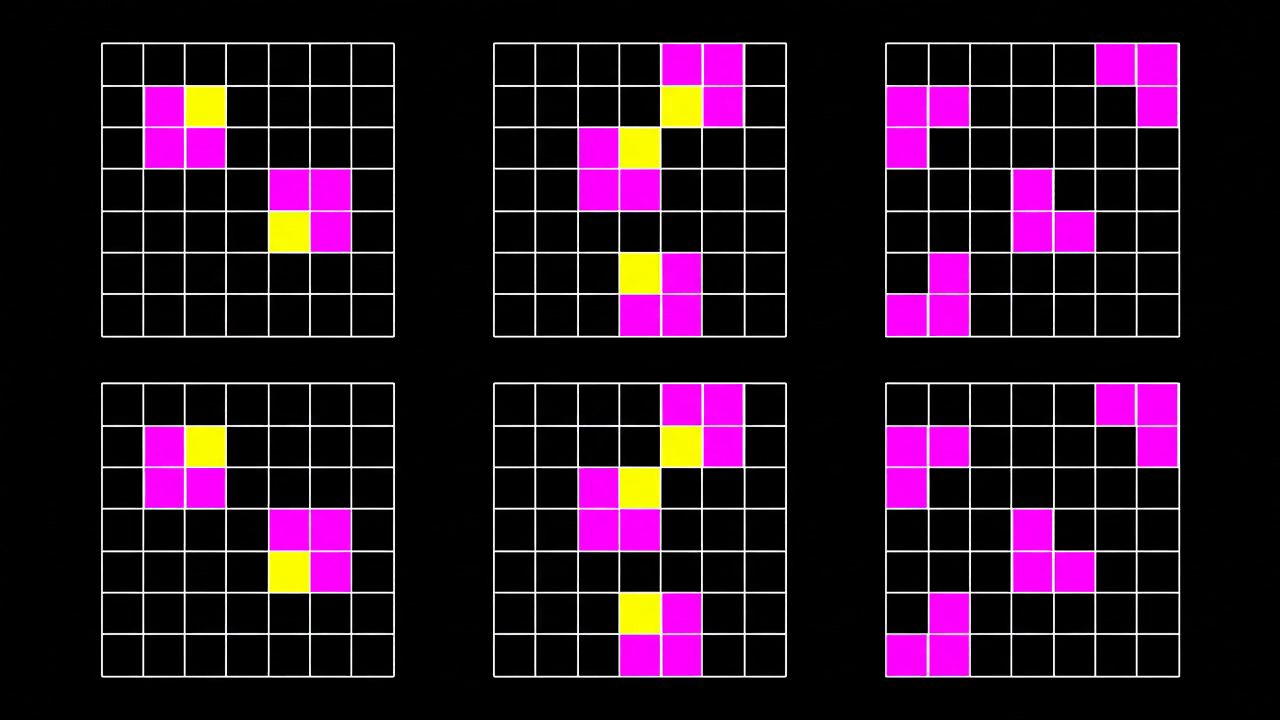}
        };
        \node[exampleframe] at ($(B4) + ((\boxpad + 5\imw + 6\boxgap, -\boxtextheight)$) () {
                \includegraphics[width=\imw]{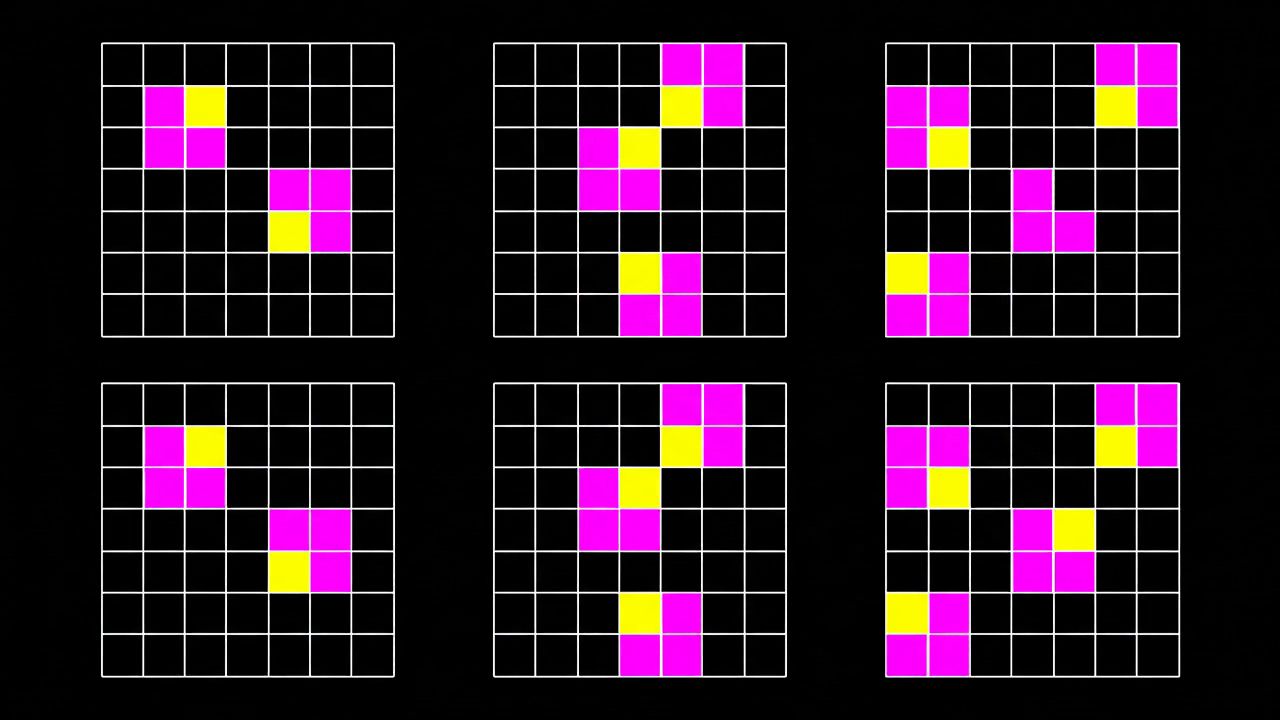}
        };
    \end{tikzpicture}
    }
    \caption{\textbf{\veothree{} zero-shot learning and reasoning examples}. From classic \textcolor{perceptioncol}{\textbf{perceptual}} tasks (superresolution, visual search) to \textcolor{modelingcol}{\textbf{modeling}} (buoyancy, memory of world states after zooming in), \textcolor{manipulationcol}{\textbf{manipulation}} (pose editing, simulating dexterous manipulation) and \textcolor{reasoningcol}{\textbf{visual reasoning}} (navigation, rule extrapolation): \veothree{} can zero-shot solve a host of visual tasks that are specified as an input image and a text prompt. Examples are shown in \Cref{app:qualitative}; videos of all tasks are on our \href{\websiteurl}{project page}.}
    \label{fig:qualitative-examples}
\end{figure}

We begin with a comprehensive, qualitative investigation across visual tasks to assess the potential of video models as visual foundation models. We organize our findings into four hierarchical capabilities, each building on the preceding ones (c.f. ~\cref{fig:overview} and \cref{fig:qualitative-examples}): 
\begin{enumerate}
    \item \textcolor{perceptioncol}{\textbf{Perception}} as a foundational ability to understand visual information.
    \item \textcolor{modelingcol}{\textbf{Modeling}}, which builds on the perception of objects to form a model of the visual world.
    \item \textcolor{manipulationcol}{\textbf{Manipulation}}, which meaningfully alters the perceived and modeled world.
    \item \textcolor{reasoningcol}{\textbf{Reasoning}} across space and time over a sequence of manipulation steps. 
\end{enumerate}
\noindent
While capability boundaries often overlap, this hierarchy provides a framework for understanding the emergent abilities of video models. For example, solving a maze (see \cref{fig:maze_mouse,subsec:maze}) requires perceiving the maze, modeling its state (walls vs.\ floor), and finally manipulating an object (a mouse, a circle) to move from start to finish.

For each task in this section, we prompt \veothree{} twelve times and report the \emph{success rate} in the caption. The \emph{success rate} is defined as the fraction of generated videos that solved the task (as determined by the authors). A success rate greater than 0 suggests that the model possesses the \emph{ability} to solve the task, while a success rate closer to 1 indicates that the task is solved \emph{reliably} irrespective of the random seed. While not a substitute for the systematic quantification we perform in \cref{sec:quantitative_results}, this provides a ballpark estimate of the model's capabilities.

\paragraph{\textcolor{perceptioncol}{Perception}}
Computer vision has historically relied on a suite of specialized models for tasks like segmentation~\citep{kirillov2023segment,ravi2024sam}, object detection~\citep{redmon2016you,khanam2024yolov11}, and edge detection~\citep{xie2015holistically}. While some backbones can be adapted or fine-tuned for other tasks, training-free transfer to novel tasks is rare, limiting generalization. As we show here, this is changing with large video models.

Without any task-specific training, \veothree{} can perform a range of classic computer vision tasks, including edge detection~(\cref{fig:edge_detection}), segmentation~(\cref{fig:segmentation}), keypoint localization~(\cref{fig:keypoint_localization}), super-resolution~(\cref{fig:superresolution}), blind deblurring (\cref{fig:deblurring}), denoising~(\cref{fig:denoising}) and low-light enhancing~(\cref{fig:low_light}).  These emergent abilities extend to more complex perceptual tasks like conjunctive search~(\cref{fig:conjunctive_search}) and interpreting ambiguous images such as the classic dalmatian illusion~(\cref{fig:dalmatian}), the cat shape in a texture-shape cue conflict image~(\cref{fig:texture-shape}), and colored blots from the Rorschach test~(\cref{fig:rorschach}). Apart from denoising---the classic diffusion objective---none of these tasks are explicitly trained for in video models.

\takeaway{\veothree{} shows emergent zero-shot perceptual abilities well beyond the training task. Just like LLMs replaced task-specific NLP models, video models will likely replace most bespoke models in computer vision---once they become sufficiently cheap and reliable.}

\paragraph{\textcolor{modelingcol}{Modeling: intuitive physics \& world models}} Based on their \emph{perception} of the visual world, video models are starting to \emph{model} it, too. Modeling the world and the principles that govern it (e.g., laws of physics) is a critical step toward successful prediction and action. Several works have investigated and quantified intuitive physics in deep models~\citep[e.g.,][]{riochet2018intphys,bear2021physion,weihs2022benchmarking,jassim2023grasp,tung2024physionpp,bansal2024videophy,cherian2024llmphy,meng2024world,kang2024far,motamed2025generative,liu2025generative,schulze2025visual,zhang2025morpheus,garrido2025intuitive}. Here, we investigate an exemplary subset of tasks from these works. \veo{}'s grasp of physics is demonstrated by its ability to model various physical properties, like flammability~(\cref{fig:material_properties}), rigid and soft body dynamics and their surface interactions~(\cref{fig:physics-body}) and air resistance affecting falling objects~(\cref{fig:gravity_air_resistance}), and buoyancy~(\cref{fig:water_floating}). As illustrated by the Visual Jenga task~\citep{bhattad2025visual}, \veo{} can remove objects from a scene in a physically plausible order~(\cref{fig:visual-jenga}) and understands which objects fit into a backpack~(\cref{fig:fitting_objects}). Furthermore, it correctly generates some optical phenomenona like refraction and reflection~(\cref{fig:material_optics}) and additive/subtractive color mixing~(\cref{fig:color_mixing}). Beyond physical characteristics, \veo{} understands some abstract relationships which is an important aspect of modeling the world. For example, \veo{} can distinguish categories like toys from other objects like a laptop~(\cref{fig:categorizing_objects}). On samples inspired by the Omniglot dataset~\citep{lake2015human}, we demonstrate \veo{}'s ability to recognize patterns, generate variations thereof, and parse larger wholes into parts~(\cref{fig:omniglot}). Lastly, \veo{} maintains a memory of the world state across time and camera movements within the video context~(\cref{fig:working_memory}).

\paragraph{\textcolor{manipulationcol}{Manipulation: editing \& imagination}}
Based on its ability to \emph{perceive} objects and \emph{model} their relation to each other and the world, \veo{} can meaningfully \emph{manipulate} the visual world, too. This enables \veothree{} to perform a variety of zero-shot image editing tasks like background removal~(\cref{fig:background_removal}), style transfer~(\cref{fig:style_transfer}), colorization~(\cref{fig:colorization}), inpainting~(\cref{fig:inpainting}), and outpainting~(\cref{fig:outpainting}). Furthermore, it can manipulate text elements~(\cref{fig:text_manipulation}), and edit images based on doodle instructions~(\cref{fig:doodle}). \veo{}'s understanding of 3D world enables it to compose scenes from individual components~(\cref{fig:scene_composition}), generate novel views of objects and characters~(\cref{fig:view_synthesis,fig:reposing}), smoothly transform one object into another~(\cref{fig:transfiguration}), or change the perspective, lighting, and appearance to turn a selfie into a professional photograph~(\cref{fig:linkedin_headshot}).

This ability to plausibly modify a scene allows it to imagine complex interactions, simulate dexterous object manipulation~(\cref{fig:hand-dexterity}; note that we do not test actual robots as e.g.\ \citep{assran2025v} do), interpreting object affordances~(\cref{fig:affordance}), demonstrating how to draw a shape~(\cref{fig:drawing}) and roll a burrito~(\cref{fig:visual-explanation}). Overall, video models can meaningfully manipulate and simulate aspects of the (digital) visual world.

\paragraph{\textcolor{reasoningcol}{Visual reasoning across time and space}}
\emph{Perception}, \emph{modeling}, and \emph{manipulation} all integrate to tackle \emph{visual reasoning}. While language models manipulate human-invented symbols, video models can apply changes across the dimensions of the real world: time and space. Since these changes are applied frame-by-frame in a generated video, this parallels chain-of-thought in LLMs and could therefore be called \emph{chain-of-frames}, or CoF for short. In the language domain, chain-of-thought enabled models to tackle reasoning problems~\citep{wei2022chain}. Similarly, chain-of-frames (a.k.a.\ video generation) might enable video models to solve challenging visual problems that require step-by-step reasoning across time and space.

We see early signs of this ability in tasks such as generating a valid graph traversal~(\cref{fig:graph_traversal}), performing visual breadth-first search on a tree~(\cref{fig:tree_bfs}), completing visual sequences~(\cref{fig:raven_logic}), connecting matching colors~(\cref{fig:connecting_colors}), fitting shapes into holes~(\cref{fig:shape-fitting}), and sorting numbers~(\cref{fig:number_sort}). Furthermore, \veo{} can use tools to accomplish a visual task~(\cref{fig:tool-use}) and solve simple Sudokus~(\cref{fig:sudoku}) or visual puzzles~(\cref{fig:water-puzzle}). Finally, it can solve mazes and navigation tasks~(\cref{fig:maze_mouse,fig:robot_navigation,subsec:maze}) and extrapolate rules from visual examples~(\cref{fig:arc-agi}). While not always perfect, the model's ability to solve such problems in a zero-shot manner points to exciting possibilities for more advanced visual reasoning and planning in future, with more capable video models.

\takeaway{Frame-by-frame video generation parallels chain-of-thought in language models. Just like chain-of-thought (CoT) enables language models to reason with symbols, a ``chain-of-frames'' (CoF) enables video models to reason across time and space.}

\paragraph{Summary}
Taken together, the qualitative examples from this section indicate that a capable video model like \veothree{} possesses strong zero-shot learning abilities. While the results are not always perfect, the model consistently demonstrates the capacity to solve a wide variety of tasks for which it was not explicitly trained.

%% file: sections/04_quantitative.tex
\section{Quantitative results}
\label{sec:quantitative_results}
The previous section offered a qualitative exploration of video model capabilities. In this section, we add a quantitative assessment for seven tasks.
As in \cref{sec:qualitative_results}, we consider different facets of visual understanding: For \textcolor{perceptioncol}{\textbf{perception}}, we assess \veo{} on edge detection and segmentation. For \textcolor{manipulationcol}{\textbf{manipulation}}, we examine image editing and object extraction performance. Finally, we evaluate \textcolor{reasoningcol}{\textbf{reasoning}} abilities through maze solving, visual symmetry, and visual analogies.
We do not include evaluations for \textcolor{modelingcol}{\textbf{modeling}}, since this area is well addressed by recent benchmarks, see~\cref{sec:qualitative_results}.

We evaluate performance separately for the \emph{best frame} and the \emph{last frame} (where applicable). \emph{Best frame} reports the best performance across any frame in the generated videos. This indicates the performance ceiling, but the optimal frame is not known a priori. Therefore, we also report performance on the \emph{last frame} of each video, which may underestimate a model's ability but has the practical advantage that the frame is predetermined. This distinction is important because \veo{} tends to continue animating a scene even after task completion, potentially reducing last frame performance.

Where applicable, we use the state-of-the-art image editing model \nanobanana{}~\citep{NanoBanana2025} as a reference. 
As a general trend, we observe a large performance increase from \veotwo{} to \veothree{}, often matching or exceeding \nanobanana{}'s performance. Performance tends to improve substantially from $k=1$ to $k=10$ attempts, indicating that a good solution can be found in a reasonable number of tries. While image models are excellent zero-shot learners, too \citep{clark2023text,jaini2023intriguing, burgert2022peekaboo, khachatryan2023text2video}, video models are the more general framework because of their ability to process both temporal and spatial information.

\subsection{\textcolor{perceptioncol}{\textbf{Perception}}: Edge detection}
\label{subsec:edge}
\begin{figure}[t]
    \centering
    \begin{subfigure}[t]{\threefigwidth\textwidth}
        \caption{\opensans Original image}
        \fbox{\includegraphics[width=\linewidth]{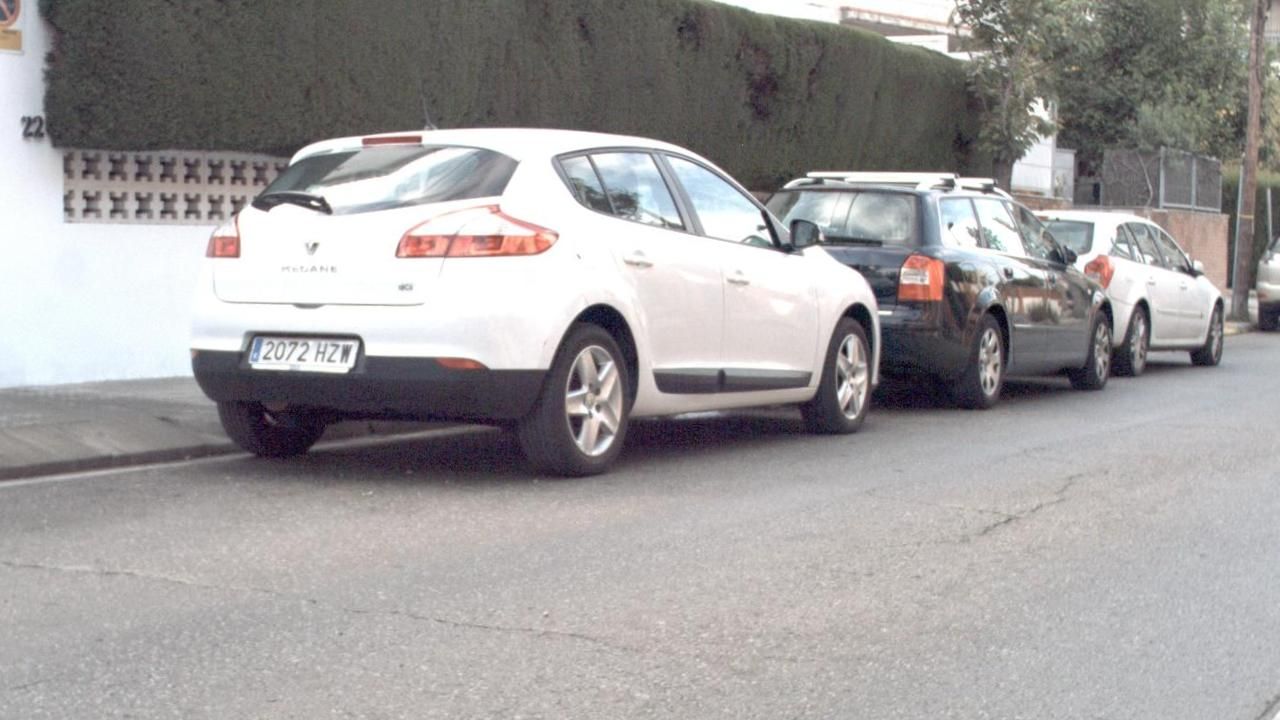}}
    \end{subfigure}
    \hfill
    \begin{subfigure}[t]{\threefigwidth\textwidth}
        \caption{\opensans Generated edge map (\veothree{})}
        \fbox{\includegraphics[width=\linewidth]{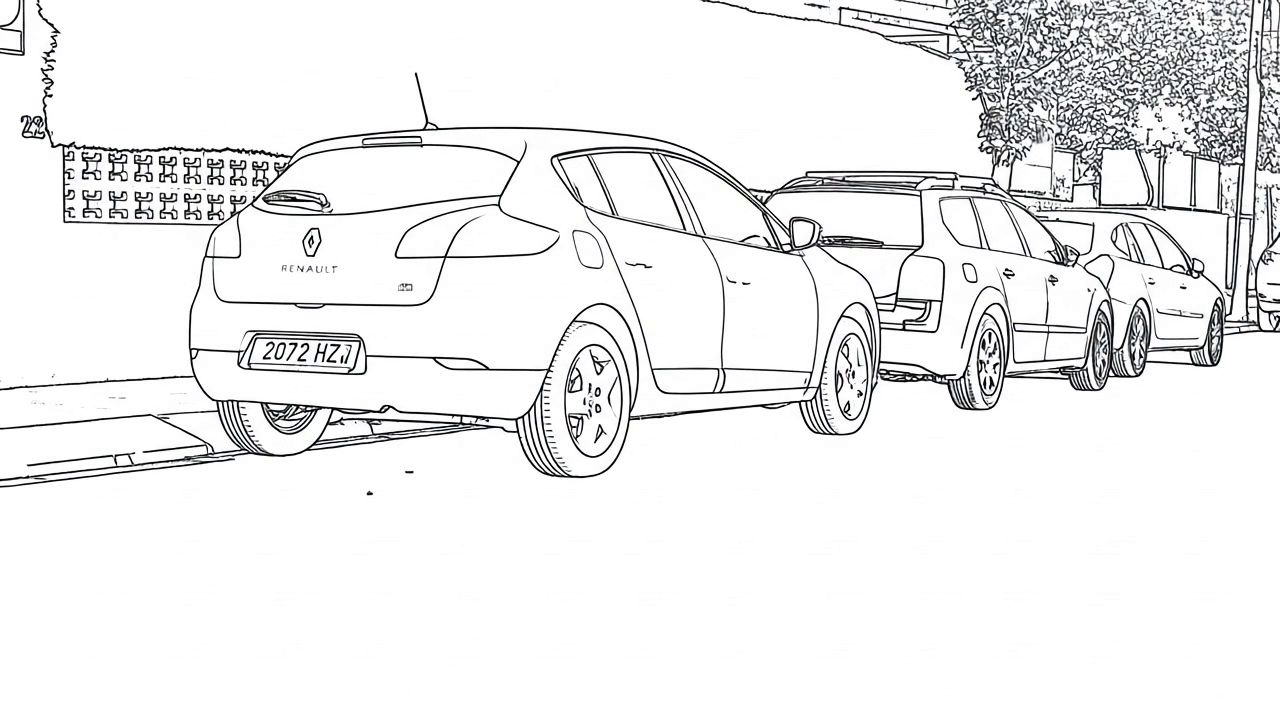}}
    \end{subfigure}
    \hfill
    \begin{subfigure}[t]{\threefigwidth\textwidth}
        \caption{\opensans Ground-truth edge map}
        \fbox{\includegraphics[width=\linewidth]{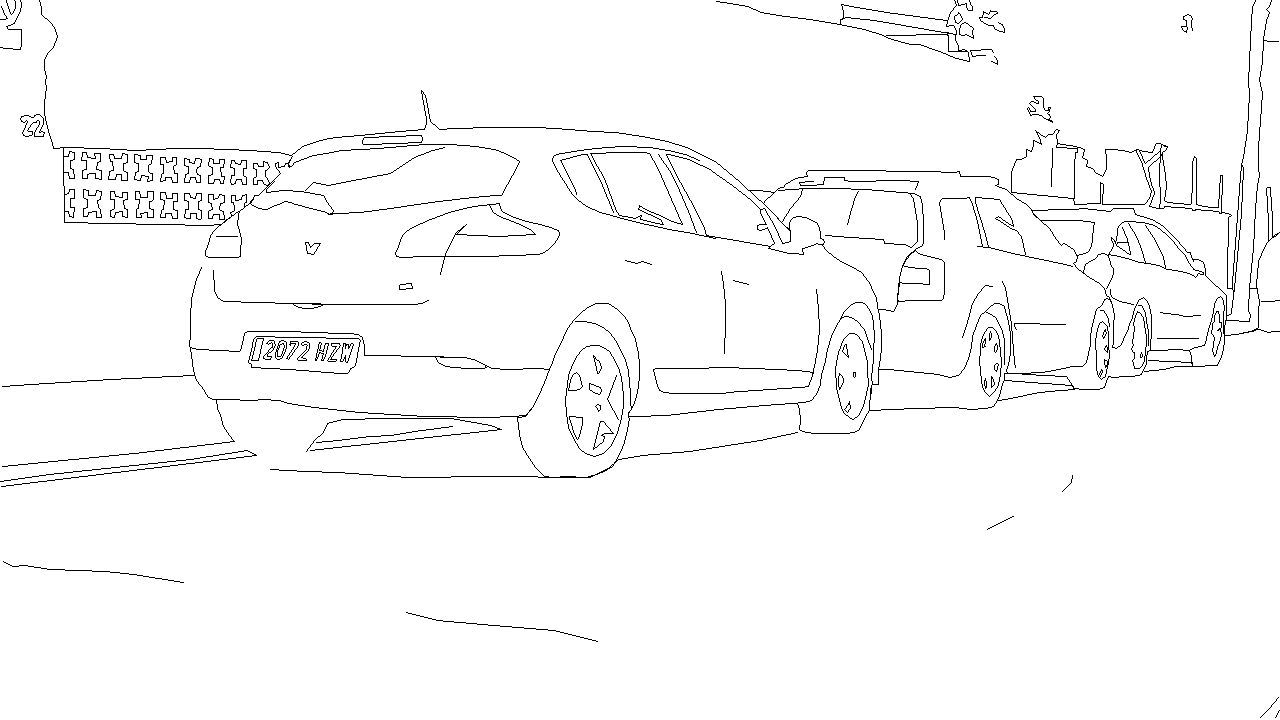}}
    \end{subfigure}
    \vspace{.5em}\newline
    \includegraphics[width=\linewidth]{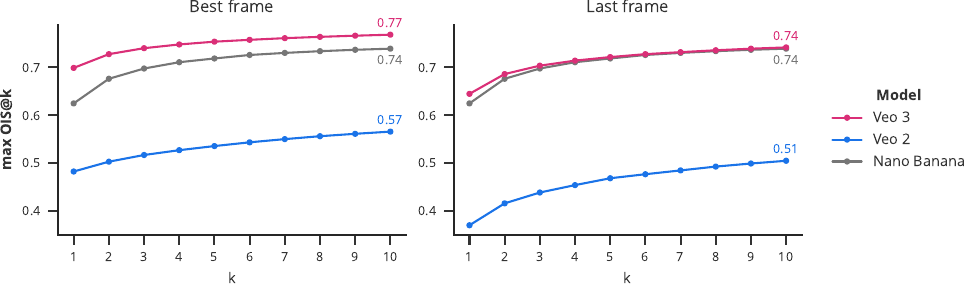}
    \caption{
        \textbf{Edge detection} on all 50 test images from BIPEDv2~\citep{soria2020dexined, soria2023dexined_ext}. We generate 10 videos per sample and report best performance over $k$ attempts as a function of $k$.
        Prompt: \textit{``All edges in this image become more salient by transforming into black outlines. Then, all objects fade away [...]''} Details \& full prompt: \cref{app:edge}.
    }
    \label{fig:edge-detection-results}
\end{figure}
Despite not being trained for it, \veothree{} can be prompted to detect, and therefore \textcolor{perceptioncol}{\textbf{perceive}}, edges. \cref{fig:edge-detection-results} details edge detection performance (measured by OIS; details and prompt in \Cref{app:edge}) for \veotwo{} and \veothree{}. While \veothree{} (0.77 pass@10) is not on par with task-specific SOTA (0.90)~\citep{soria2023dexined_ext}, its performance is remarkable for two reasons: First, it is zero-shot. Second, many of \veothree{}'s edge maps are more detailed than the ground truth. For example, \veothree{} correctly outlines foliage and tire profiles; this hurts performance on the dataset but seems more indicative of a dataset limitation than a model limitation (the annotators understandably did not bother to trace each and every edge).

\subsection{\textcolor{perceptioncol}{\textbf{Perception}}: Segmentation}
\label{subsec:segmentation}
\begin{figure}[t]
    \centering
        \begin{subfigure}[t]{0.24\textwidth}
        \caption{\opensans Original image}
        \fbox{\includegraphics[width=\linewidth]{figures-compressed/segmentation_segmentation-visualization-input_20250911-082317.jpg}}
    \end{subfigure}
    \hfill
    \begin{subfigure}[t]{0.24\textwidth}
        \caption{\opensans Generated frame (\veothree{})}
        \fbox{\includegraphics[width=\linewidth]{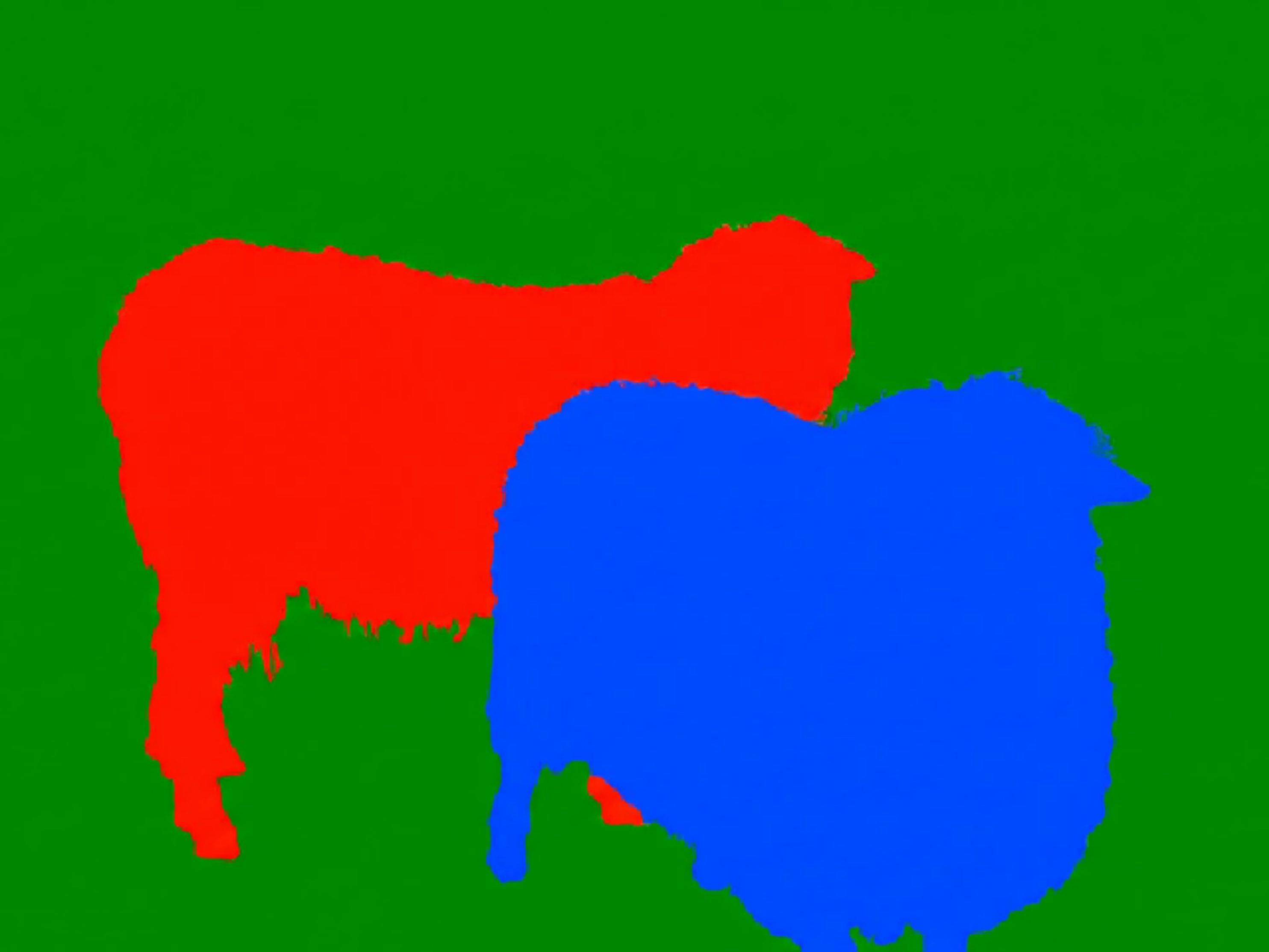}}
    \end{subfigure}
    \hfill
    \begin{subfigure}[t]{0.24\textwidth}
        \caption{\opensans Extracted masks}
        \fbox{\includegraphics[width=\linewidth]{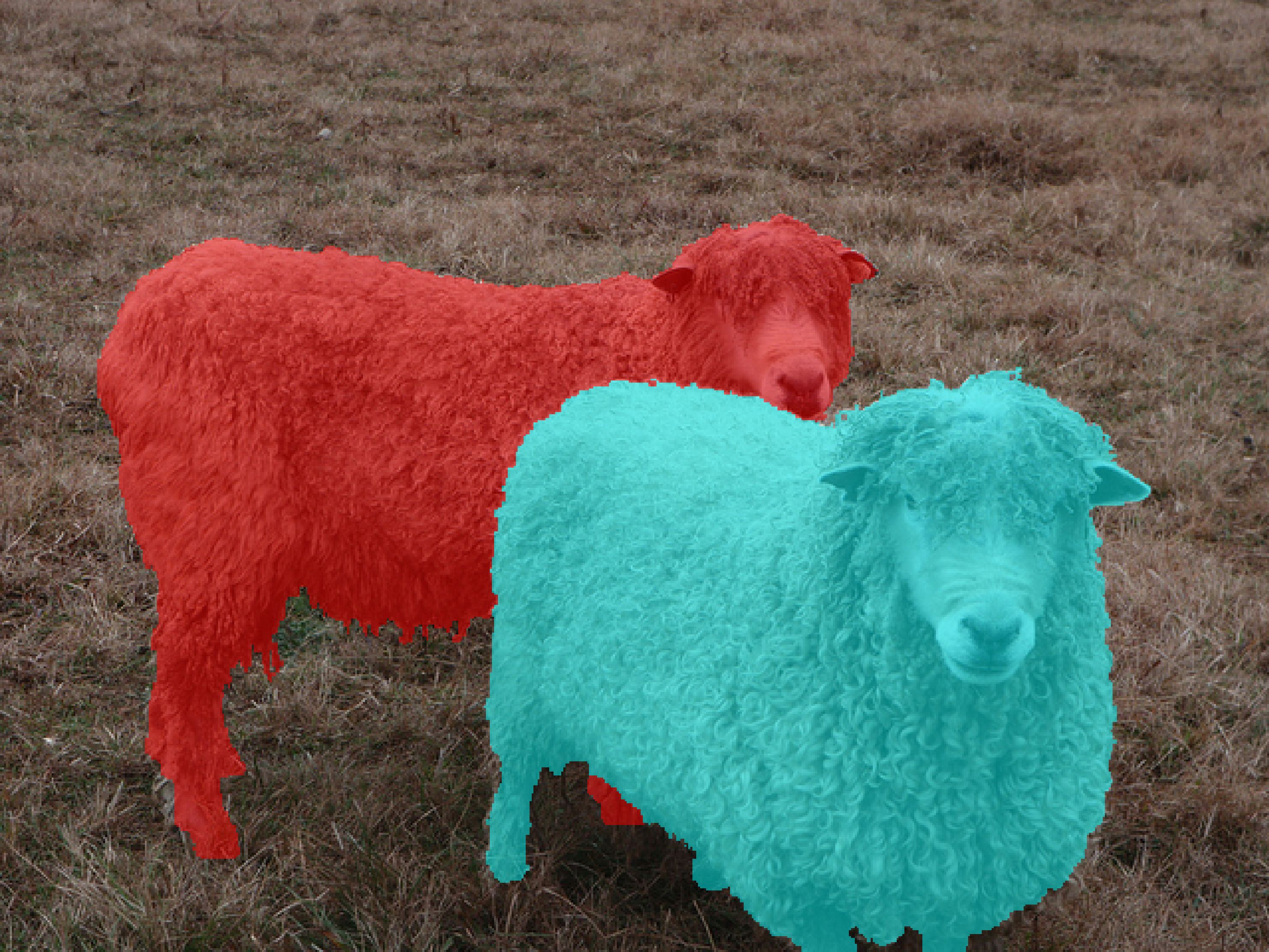}}
    \end{subfigure}
    \hfill
    \begin{subfigure}[t]{0.24\textwidth}
        \caption{\opensans Ground-truth masks}
        \fbox{\includegraphics[width=\linewidth]{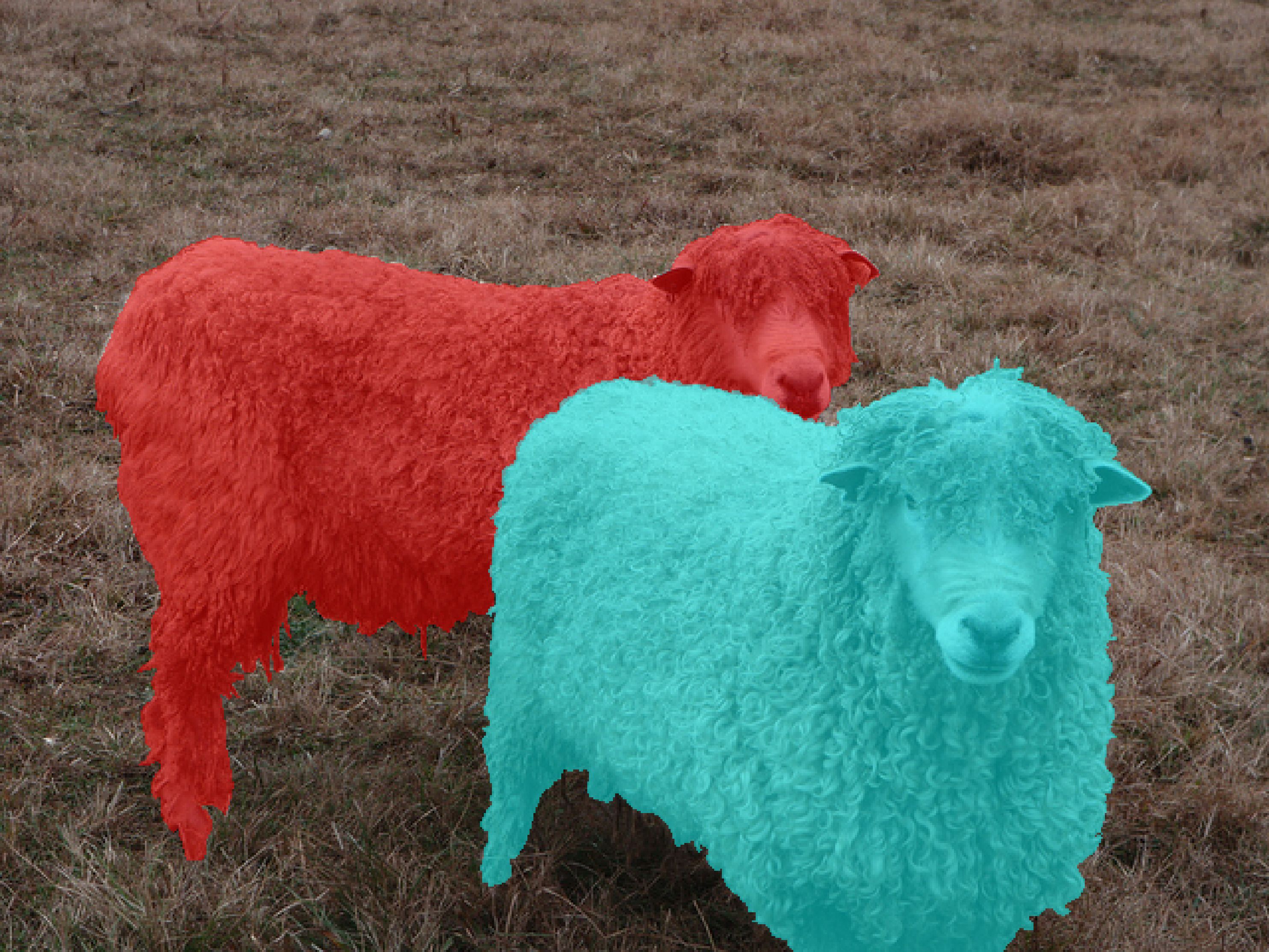}}
    \end{subfigure}
    \vspace{0.5em}\newline
    \includegraphics[width=\linewidth]{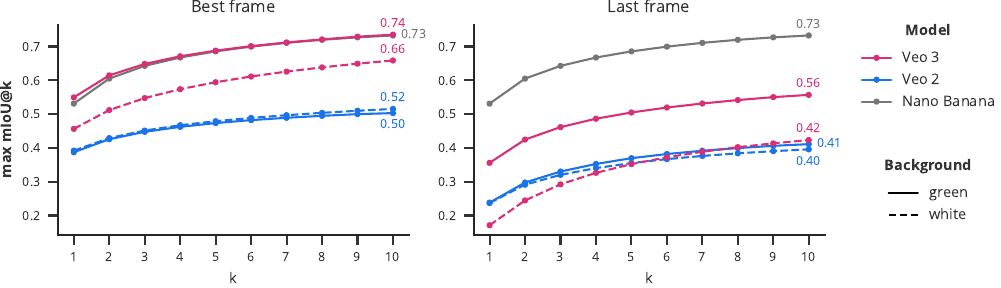}
    \caption{
        \textbf{Class-agnostic instance segmentation} on a subset of 50 easy images (1-3 large objects) from LVIS~\citep{gupta2019lvis}.
        Prompt: \textit{``[...] each distinct entity is overlaid in a different flat color [...] the background fades to \{white, green\} [...]''} Details \& full prompt: \cref{app:segmentation}.
    }
    \label{fig:segmentation-results}
\end{figure}

Instance segmentation requires delineating (i.e., \textcolor{perceptioncol}{\textbf{perceiving}}) distinct objects in an image.
Contrary to classic instance segmentation or promptable segmentation, we prompt models to segment all objects in a scene, without specifying an object category or location. We report mean Intersection over Union (mIoU); experiment details are in \cref{app:segmentation}. As shown in \cref{fig:segmentation-results}, \veothree{} achieves an mIoU of 0.74 (best frame pass@10), comparable to \nanobanana{}'s 0.73. Naturally, \veothree{} lacks behind the performance of a bespoke model like SAMv2~\citep{ravi2024sam}, but nevertheless shows remarkable zero-shot segmentation abilities. Interestingly, the prompt really matters: \veo{} consistently performs better with a green background than a white one (0.74 vs.\ 0.66 best frame pass@10); possibly due to the widespread use of green screens. See also \cref{app:prompt-sensitivity} for prompting best practices.

\subsection{\textcolor{manipulationcol}{\textbf{Manipulation}}: Object extraction}
\label{subsec:counting}
\begin{figure}[t]
    \centering
    \includegraphics[width=\linewidth]{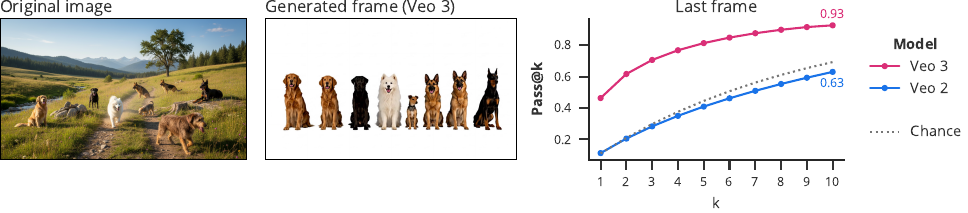}
    \caption{\textbf{Object extraction} on an animal dataset. Prompt: \textit{``The background changes to white [...] all animals line up in a row [...]''} Details \& full prompt: \Cref{app:counting}.}
    \label{fig:counting}
\end{figure}
Can \veo{} perceive and extract (i.e., \textcolor{manipulationcol}{\textbf{manipulate}}) all objects in a scene? We test this using a simple dataset depicting between one and nine animals (details in \cref{app:counting}). \veo{} is asked to extract and line up all animals in a row (in some sense, a visual ``tally''), with white space between them. To assess whether the number of extracted animals is correct, we count connected components in the last frame. \cref{fig:counting} shows a sample and the results. While \veotwo{} performs around chance, \veothree{} achieves up to 93\% pass@10. Given the simplicity of the task, a perfect model should easily achieve 100\% accuracy.

\subsection{\textcolor{manipulationcol}{\textbf{Manipulation}}: Image editing}
\label{subsec:editing}
\begin{figure}[t]
    \centering
    \includegraphics[width=\linewidth]{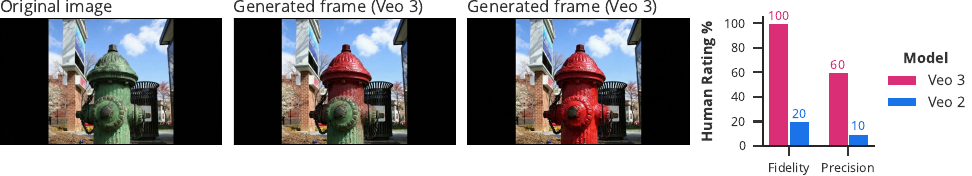}
    \caption{\textbf{Image editing} on a subset of Emu-edit~\cite{sheynin2024emu}. Prompt: \textit{``Create a smooth, static animation that slowly changes the color of the fire hydrant to red. [...]''} Details \& full prompt: \cref{app:editing}.}
    \label{fig:image-editing}
\end{figure}

Image editing requires \textcolor{manipulationcol}{\textbf{manipulating}} images according to a text instruction (e.g., adding/removing objects or changing their appearance). We evaluate whether \veo{} can edit images on a random subset of 30 samples from Emu-edit~\cite{sheynin2024emu}.
\veo{} has a strong bias for animated scenes and might introduce unintended changes (e.g., camera movement, animating people).
We therefore ask three human raters to evaluate \emph{fidelity} (correct edit) and \emph{precision} (correct edit without unintended changes).
An example edit and results are shown in \cref{fig:image-editing}. We find that \veothree{} especially excels in preserving details and textures across edits. If unintended changes such as camera movement or animating people can be controlled better, video models could become highly capable 3D-aware image and video editors (see also~\citep{molad2023dreamix,sun2024diffusion,yu2025veggie,rotstein2025pathways}).

\begin{figure}[t]
    \centering
    \includegraphics[width=\linewidth]{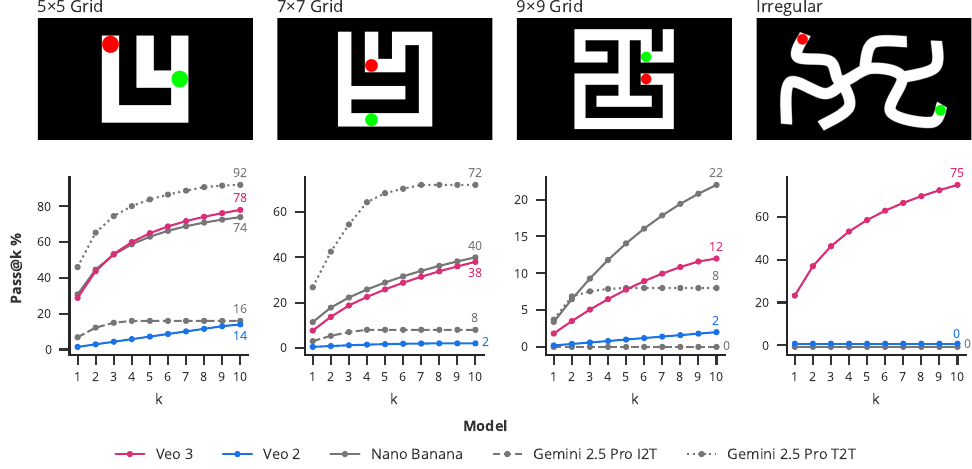}
    \caption{
        \textbf{Maze solving}. Mazes of various sizes with start (red) and goal (green) locations.
        Prompt: \emph{``[...] The red circle slides smoothly along the white path, stopping perfectly on the green circle [...]''} Details \& full prompt: \cref{app:maze}.
        Veo~2 struggles to solve even small sizes, mostly due to illegal moves early in the generation. \veothree{} performs much better and benefits from multiple attempts. For comparison, we evaluate \nanobanana{} and also show \geminipro{}'s performance on mazes presented as images (I2T) or ASCII text (T2T).
    }
    \label{fig:maze}
\end{figure}

\subsection{\textcolor{reasoningcol}{\textbf{Reasoning}}: Maze solving}
\label{subsec:maze}
Maze solving tests a model's ability to plan a path in a constrained environment, a key aspect of \textcolor{reasoningcol}{\textbf{reasoning}}. In our setup, a red circle needs to navigate to a target location (green circle) without crossing any walls. We automatically verify path correctness (details in \Cref{app:maze}) and present results for different mazes in \cref{fig:maze}.
\veothree{} shows zero-shot maze solving abilities, significantly outperforming \veotwo{} which often produces illegal moves. For instance, in the 5$\times$5 grids, \veothree{} achieves a $\text{pass@10}$ rate of 78\% compared to \veotwo{}'s 14\%. The consistent performance gap highlights the advancing reasoning capabilities of the newer model. While \nanobanana{} matches or surpasses \veothree{}'s performance on rectangular mazes, it fails to solve irregular mazes entirely.
Similarly, \geminipro{} outperforms \veothree{} on small mazes when given an ASCII representation of the maze (T2T), but falls behind on 9$\times$9 mazes, and generally struggles when the maze is represented as image (as opposed to text) input.
Both comparisons highlight the advantages of solving a visual task step-by-step in a visual medium.

\subsection{\textcolor{reasoningcol}{\textbf{Reasoning}}: Visual symmetry solving}
\label{subsec:visual_symmetry}
\begin{figure}[t]
    \centering
    \includegraphics[width=\textwidth]{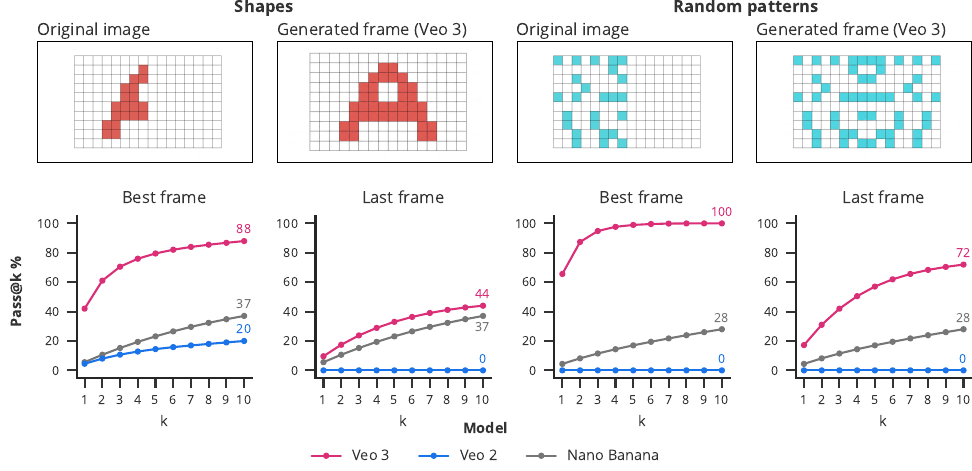}
    \caption{
        \textbf{Visual symmetry}. 
        Prompt: \emph{``Instantly reflect this pattern along the central, vertical axis while keeping the existing colored pattern without modification. [...]''}
        A model has to color all cells correctly to pass. Details \& full prompt: \cref{app:symmetry}.
    }
    \label{fig:symmetry}
\end{figure}

Completing a pattern to be symmetrical assesses the ability to understand and apply spatial \textcolor{reasoningcol}{\textbf{reasoning}}. We create a custom dataset of shapes (e.g., heart, letters) and random patterns (see \cref{app:symmetry} for details). \cref{fig:symmetry} shows that \veothree{} outperforms \veotwo{} and \nanobanana{} by a large margin.
We also use this task to systematically analyze how different prompts affect performance in \cref{app:prompt-sensitivity}: The pass@1 difference between best and worst prompt is 40 percentage points on the shape split and 64 percentage points on the random split.

\subsection{\textcolor{reasoningcol}{\textbf{Reasoning}}: Visual analogy completion}
\label{subsec:visual_analogy}
\begin{figure}[t]
    \centering
    \includegraphics[width=\linewidth]{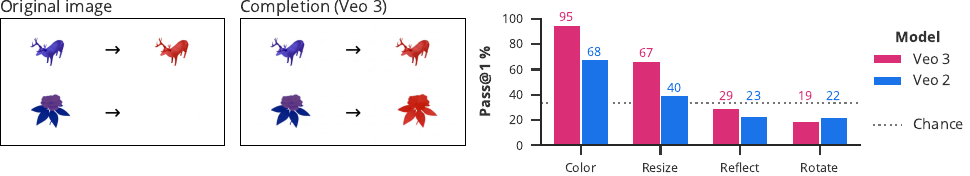}
    \caption{\textbf{Visual analogy solving} on four transformations à 50 samples from KiVA~\citep{yiu2024kiva}. Prompt: \textit{``[...] generate the missing object in the lower right region and solve the visual analogy. [...]''} Pass@1 is evaluated on the last frame, results up to $\text{pass@10}$ can be found in \cref{app:visual_analogy}.}
    \label{fig:visual-analogy-kiva}
\end{figure}
Visual analogies test a model's ability to understand transformations and relationships between objects, a form of abstract \textcolor{reasoningcol}{\textbf{reasoning}}. Concretely, we prompt the model to fill the missing quadrant of a 2$\times$2 grid to complete the analogy (e.g., A is to B as C is to ?). We evaluate the correctness of the generated infill for four transformation types from KiVA~\citep{yiu2024kiva}, see \cref{app:visual_analogy} for details. The results are summarized in \cref{fig:visual-analogy-kiva}. While \veotwo{} struggles to understand any analogies, \veothree{} correctly completes examples for \emph{color} and \emph{resize}. However, both models perform below chance (0.33) on \emph{reflect} and \emph{rotate} analogies, indicating an erroneous, systematic bias.

\takeaway{While far from perfect, Veo 3---building on its ability to perceive, model and manipulate objects---shows emergent visual reasoning abilities.}

%% file: sections/05_discussion.tex
\section{Discussion}
\paragraph{Summary}
A breakthrough in machine vision started the deep learning revolution in 2012~\citep{krizhevsky2012imagenet}, but in recent years, natural language processing has seen the most rapid progress. This was driven by the rise of general-purpose LLMs, whose ability to solve novel tasks in a zero-shot fashion has led them to replace most task-specific models in NLP. We here make the case that machine vision is on the cusp of a similar paradigm shift, enabled by emergent abilities of large-scale video models. Our core finding is that \veothree{} can solve a wide range of tasks in a zero-shot manner, spanning the full vision stack from \textcolor{perceptioncol}{\textbf{perception}} to \textcolor{modelingcol}{\textbf{modeling}}, \textcolor{manipulationcol}{\textbf{manipulation}} and even early forms of \textcolor{reasoningcol}{\textbf{visual reasoning}}. While its performance is not yet perfect, the massive and consistent improvement from \veotwo{} to \veothree{} indicates that video models will become general-purpose foundation models for vision, just as LLMs have for language.

\paragraph{Performance is a lower bound}
Tasks can be represented in a myriad of ways; a maze, for example, can be presented as a black-and-white grid, a video game, or a photorealistic scene, with the prompt requesting a solution in the form of a line, a moving object, or a glowing path. Moreover, visually, a maze could be represented as a black-and-white grid, a Pac-Man game, or a photorealistic top-down view of an apartment. This has three implications: First, prompt engineering---including the \textit{visual} prompt a.k.a.\ starting frame---is as important for visual tasks as it is for LLMs (see also \cref{app:prompt-sensitivity} and \citep{lampinen2024broader} for a discussion). Second, we must distinguish between a model's task performance and its underlying ability (i.e., competence) to solve that task \citep{firestone2020performance,geirhos2020shortcut}. Third, as a consequence, the model performance reported here with a given visual and textual prompt should be considered a lower bound on the model's true capabilities. This also holds for the tasks that we report as failure cases in \cref{app:failure_cases}, such as providing visual instructions to fold laundry (\cref{fig:laundry-folding}), planning to move a sofa between rooms separated by a small door (\cref{fig:piano-mover}), or certain visual puzzles (\cref{fig:spot-difference}).

\paragraph{Video generation is expensive, but costs tend to fall}
While generating a video is currently more expensive than running a bespoke, task-specific model, the economics of general-purpose models are on a predictable trajectory. Epoch~AI~\citep{epoch2025llminferencepricetrends} estimates that LLM inference costs are falling by a factor of 9$\times$ to 900$\times$ per year for a given performance level. In NLP, early generalist models were also considered prohibitively expensive (``GPT-3's size makes it challenging to deploy'' \citep[][p.~8]{brown2020language}). Nevertheless, rapidly falling inference costs, combined with the appeal of generalist models, have replaced most task-specific language models. If NLP is a guide, the same trend will play out in vision.

\paragraph{Jack of many trades, master of few?}
For many tasks, \veothree{}'s performance is below state of the art of specialized models. This mirrors the early days of LLMs; GPT-3 reported performance well below fine-tuned models on many tasks \citep[][cf.\ Tables 3.1, 3.3, 3.4, 3.5]{brown2020language}). This did not stop language models from becoming foundation models, and we don't believe it will stop video models from becoming vision foundation models for two reasons. First, the step-change in performance from \veotwo{} to \veothree{} is evidence of rapid progress over time. Second, our scaling results from \cref{sec:quantitative_results} show $\text{pass@10}$ to be consistently higher than $\text{pass@1}$ with no signs of a plateau. Therefore, inference-time scaling methods \citep[e.g.][]{wang2022self,madaan2023self,jaech2024openai,snell2024scaling} in combination with the standard optimization toolkit like post-training with automatic verifiers are likely to boost performance. For the tasks we test here,  \veothree{} is akin to a pre-trained language model that has yet to undergo instruction tuning or RLHF \citep{ouyang2022training,peng2023instruction}.

\paragraph{Outlook} This is an exciting time for vision. Seeing NLP's recent transformation from task-specific to generalist models, it is conceivable that the same transformation will happen in machine vision through video models (a ``GPT-3 moment for vision''), enabled by their emergent ability to perform a broad variety of tasks in a zero-shot fashion, from perception to visual reasoning.

%% file: sections/99_appendix.tex
\input{sections/03a_qualitative_figures}

\section{Quantitative results: experimental details}
\label{app: quant_results}

\begin{table}[H]
    \centering
    \caption{\textbf{Video count breakdown for quantitative tasks}. For segmentation, $2\times1$ splits indicate one test set with two different background color prompts (white/green). For the prompt sensitivity study on the symmetry task (\cref{app:prompt-sensitivity}), $2\times10$ splits indicate 2 splits (random/shape) across 10 tested prompt variations. For the qualitative tasks, we additionally generated 744 videos (62 tasks $\times$ 12 samples).}
    \label{tab:quantitative_tasks}
    \sisetup{group-digits = false} %
    \begin{tabular}{l r r r r S[table-format=4.0]} 
        \toprule
        \textbf{Task} & \textbf{Splits} & \textbf{Imgs/split} & \textbf{Pass@} & \textbf{Video models} & \textbf{Total videos} \\
        \midrule
        Edge & 1 & 50 & 10 & 2 & 1000 \\
        Segmentation & 2 $\times$ 1 & 50 & 10 & 2 & 2000 \\
        Object extraction & 1 & 54 & 10 & 2 & 1080 \\
        Editing & 1 & 30 & 1 & 2 & 60 \\
        Maze & 2 $\times$ 4 & 50 & 10 & 2 & 8000 \\
        Symmetry & 2 & 25 & 10 & 2 & 1000 \\
        Symmetry prompt analysis & 2 $\times$ 10 & 25 & 1 & 1 & 500 \\
        Analogy & 4 & 50 & 10 & 2 & 4000 \\
        \midrule
        Total &  &  &  & & 17640\\
        \bottomrule
    \end{tabular}
\label{tab:video_count_breakdown}
\end{table}

\subsection{\textcolor{perceptioncol}{\textbf{Perception}}: Edge detection}
\label{app:edge}
We provide details for the image editing task in \Cref{subsec:edge}. 

\begin{figure}
    \centering
    \includegraphics[width=\linewidth]{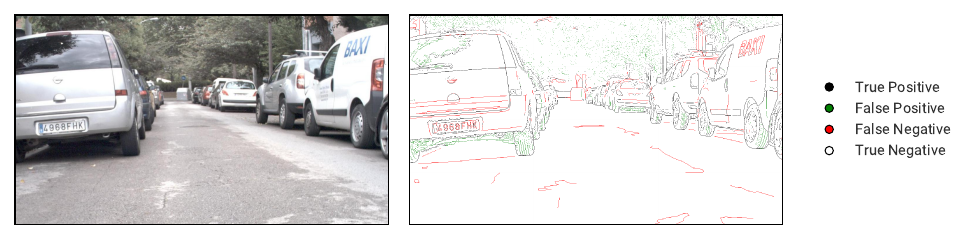}
    \caption{\textbf{Graded \veothree{} edge map}. While false negatives reflect genuine oversights of \veothree{} (e.g., cracks in the road, lettering on the car), many false positives correspond to actual image details that seem to be erroneously excluded from the ground truth (e.g., the outline of the trees, the reflection in the car window, and the tire profiles).}
    \label{fig:edge-detection-graded}
\end{figure}

\paragraph{Evaluation}
As is standard in the literature, we refine and binarize predicted edges and allow for small local shifts compared to the ground truth~\citep{DollarICCV13edges, DollarARXIV14edges, ZitnickECCV14edgeBoxes, leng2024superedge}.
Concretely, we use non-maximum suppression, then binarize with one of 16 evenly-spaced thresholds, then thin the binary edge map.
At each threshold, we find the optimal mapping between predicted and ground-truth edge pixels within a radius of 0.75\% of the image diagonal (around 11 pixels).
\cref{fig:edge-detection-graded} shows an example rating of a \veothree{}-generated edge map.
We report the best OIS over $k$ attempts (optimal image scale; the maximum $F_1$-score over all thresholds) for the best/last frame.

\paragraph{Dataset}
We used all 50 test images from BIPEDv2~\citep{soria2020dexined, soria2023dexined_ext}.

\paragraph{Models \& prompts}
We tested \veothree{} \textcolor{GoogleGray}{\small\texttt{veo-3.0-generate-preview}} and \veotwo{} \textcolor{GoogleGray}{\small\texttt{veo-2.0-\allowbreak{}generate-\allowbreak{}preview-001}} through the Vertex~AI API. We also tested \nanobanana{} \textcolor{GoogleGray}{\small\texttt{gemini-2.5-\allowbreak{}flash-\allowbreak{}image-\allowbreak{}preview}} through Google~AI~Studio.

\begin{tcolorbox}[colback=gray!20, arc=5pt, boxrule=0pt]
\textbf{\veo{}}\\
\itshape
All edges in this image become more salient by transforming into black outlines. Then, all objects fade away, with just the edges remaining on a white background. Static camera perspective, no zoom or pan.
\end{tcolorbox}

\begin{tcolorbox}[colback=gray!20, arc=5pt, boxrule=0pt]
\textbf{\nanobanana{}}\\
\itshape
Outline all edges in the image in black, make everything else white.
\end{tcolorbox}

\paragraph{Sampling}
We generated 10 videos per sample with a fixed prompt.

\subsection{\textcolor{perceptioncol}{\textbf{Perception}}: Segmentation}
\label{app:segmentation}
We provide details for the image editing task in \Cref{subsec:segmentation}. 

\paragraph{Evaluation} 
Since the model is free to choose any colors for segmentation masks, we first determine the number and hue of each mask by considering the hue-difference histogram between the original image and the extracted frame. We smooth the histogram with \texttt{scipy.ndimage.gaussian\_filter1d} with a standard deviation of 2. Peaks with a minimum height of 10\% of the maximum and at least 10 hue steps apart are considered to correspond to predicted segmentation masks.
We then map each pixel to the mask with the closest hue.

Contrary to classic instance segmentation~\citep{gupta2019lvis} or promptable segmentation~\citep{kirillov2023segment, ravi2024sam}, our prompts do not specify a class or list of possible classes, a location prior (e.g., point or bounding box), or the number of instances in the image.
This also means that mapping between predictions and annotated instances is established.
Instead, we pair each ground-truth mask (including the background) with the predicted mask with the highest IoU (intersection over union), if any.
We report mIoU as the average IoU over all pairs (excluding the background).

\paragraph{Dataset}
We evaluated on 50 randomly chosen test images from LVIS~\citep{gupta2019lvis} that contain one to three objects, each with at least 5000 pixels.

\paragraph{Models \& prompts}
We tested \veothree{} \textcolor{GoogleGray}{\small\texttt{veo-3.0-generate-preview}} and \veotwo{} \textcolor{GoogleGray}{\small\texttt{veo-2.0-\allowbreak{}generate-\allowbreak{}preview-001}} through the Vertex~AI API. We also tested \nanobanana{} \textcolor{GoogleGray}{\small\texttt{gemini-2.5-\allowbreak{}flash-\allowbreak{}image-\allowbreak{}preview}} through Google~AI~Studio.

\begin{tcolorbox}[colback=gray!20, arc=5pt, boxrule=0pt]
\textbf{\veo{}}\\
\itshape
Create an animation of instance segmentation being performed on this photograph: each distinct entity is overlaid in a different flat color.\\

Scene:
\begin{itemize}
    \item The animation starts from the provided, unaltered photograph.
    \item The scene in the photograph is static and doesn't move.
    \item First, the background fades to \{white, green\}.
    \item Then, the first entity is covered by a flat color, perfectly preserving its silhouette.
    \item Then the second entity, too, is covered by a different flat color, perfectly preserving its silhouette.
    \item One by one, each entity is covered by a different flat color.
    \item Finally, all entities are covered with different colors.
\end{itemize}
\vspace{\baselineskip}
Camera:
\begin{itemize}
    \item Static shot without camera movement.
    \item No pan.
    \item No rotation.
    \item No zoom.
    \item No glitches or artifacts.
\end{itemize}
\end{tcolorbox}

\begin{tcolorbox}[colback=gray!20, arc=5pt, boxrule=0pt]
\textbf{\nanobanana{}}\\
Perform instance segmentation on this image: Mask each distinct entity in a different opaque flat color that only preserves the silhouette and turn the background green.
\end{tcolorbox}

\paragraph{Sampling}
We generated 10 videos per sample and prompt.

\subsection{\textcolor{manipulationcol}{\textbf{Manipulation}}: Object extraction}
\label{app:counting}
We provide details for the image editing task in \Cref{subsec:counting}. 

\paragraph{Evaluation} We extract the last frame from each generated video; the resulting image is converted to greyscale, a binary mask with threshold 200 is applied, and the number of connected components is extracted using \texttt{scipy.ndimage.label}, resulting in the count estimate. We also report the chance baseline which can be calculated as: $\mathsf{random-chance} = 1 - (1-p)^k$ where $p$ is the probability to get the count correct via guessing (here: $p=\frac{1}{9}$) and $k \in [1, 10]$.

\paragraph{Dataset} We generated an animal counting dataset using \nanobanana{}. Starting from a white 16:9 image, we used the following prompt, where \texttt{number} is in $[1, 9]$ and \texttt{animal} is in [`dog', `elephant', `cat', `brown bear', `horse', `rabbit', `raccoon']. We manually evaluated the generated dataset for correctness; the resulting dataset has 54 images (exactly 6 per count).

\begin{tcolorbox}[colback=gray!20, arc=5pt, boxrule=0pt]
\textbf{\nanobanana{}}\\
\itshape
Exchange the white space with a realistic photograph of: exactly \{number\} \{animal\}, outside, not overlapping, in a natural landscape.
\end{tcolorbox}

\paragraph{Models \& prompts}
We tested \veothree{} \textcolor{GoogleGray}{\small\texttt{veo-3.0-generate-preview}} and \veotwo{} \textcolor{GoogleGray}{\small\texttt{veo-2.0-\allowbreak{}generate-\allowbreak{}preview-001}} through the Vertex~AI API.

\begin{tcolorbox}[colback=gray!20, arc=5pt, boxrule=0pt]
\textbf{\veo{}}\\
\itshape
The background changes to white. Then:
\begin{itemize}
    \item If there is just a single animal: the animal sits in the middle of the image, looking straight at the camera.
    \item If there are multiple animals: all animals line up in a row, with ample white space between them.
\end{itemize}
\end{tcolorbox}

\paragraph{Sampling} We generated 10 videos per sample with a fixed prompt.

\subsection{\textcolor{manipulationcol}{\textbf{Manipulation}}: Image editing}
\label{app:editing}
We provide details for the image editing task in \Cref{subsec:editing}. 

\paragraph{Evaluation} We perform a human study with three human raters to evaluate \textit{fidelity} (correct edit) and \textit{precision} (correct edit with no unintended changes like zooming). 

\paragraph{Dataset}
We used a random sample of 30 images from the test set of the Emu-edit dataset~\cite{sheynin2024emu}.

\paragraph{Models \& prompts}
We tested \veothree{} \textcolor{GoogleGray}{\small\texttt{veo-3.0-generate-preview}} and \veotwo{} \textcolor{GoogleGray}{\small\texttt{veo-2.0-\allowbreak{}generate-\allowbreak{}preview-001}} through the Vertex~AI API.

\begin{tcolorbox}[colback=gray!20, arc=5pt, boxrule=0pt]
\textbf{\veo{}}\\
\itshape
Create a smooth, static animation that slowly \{image specific edit direction\}. Do not change anything else. No zoom, no pan, no dolly.
\end{tcolorbox}

\paragraph{Sampling} For each image, we generated two samples and use the first sample for human rating.

\subsection{\textcolor{reasoningcol}{\textbf{Reasoning}}: Maze solving}
\label{app:maze}
We provide details for the image editing task in \Cref{subsec:maze}.

\paragraph{Evaluation}
Our evaluation process is tailored to the model type. For \veo{}, we analyze the generated video frame-by-frame, extracting the path taken by the agent (red circle). We check for any invalid moves, such as jumping over walls, clipping through boundaries, or any alteration of the goal's position. We report the success rate as the fraction of $k$ attempts where the agent successfully reaches the goal (green circle) without any illegal moves.

For \nanobanana{}, which generates the full path in one edit, we assess whether the drawn path connects the start and end points (allowing for minor discontinuities) and crucially, whether it intersects with any maze walls or goes off the valid path.

For \geminipro{} with a maze input as an image (I2T) or as ASCII (T2T), we check whether the series of grid positions represents an uninterrupted path from the start position to the goal.

\paragraph{Dataset}
For rectangular mazes, we generated 50 random mazes per size using \texttt{maze-dataset}~\citep{mazedataset}, but replacing the square start and end with circles and swapping their colors.
We also drew 10 irregular mazes by hand and flipped/rotated them to obtain 40 unique samples.

\paragraph{Models \& prompts}
We tested \veothree{} \textcolor{GoogleGray}{\small\texttt{veo-3.0-generate-preview}} and \veotwo{} \textcolor{GoogleGray}{\small\texttt{veo-2.0-\allowbreak{}generate-\allowbreak{}preview-001}} through the Vertex~AI API. We also tested \nanobanana{} \textcolor{GoogleGray}{\small\texttt{gemini-2.5-\allowbreak{}flash-\allowbreak{}image-\allowbreak{}preview}} and \geminipro{} \textcolor{GoogleGray}{\small\texttt{gemini-2.5-pro}} through Google~AI~Studio.

\begin{tcolorbox}[colback=gray!20, arc=5pt, boxrule=0pt]
\textbf{\veo{}}\\
\itshape
Create a 2D animation based on the provided image of a maze. The red circle slides smoothly along the white path, stopping perfectly on the green circle. The red circle never slides or crosses into the black areas of the maze. The camera is a static, top-down view showing the entire maze.\\

Maze:
\begin{itemize}
  \item The maze paths are white, the walls are black.
  \item The red circle moves to the goal position, represented by a green circle.
  \item The red circle slides smoothly along the white path.
  \item The red circle never slides or crosses into the black areas of the maze.
  \item The red circle stops perfectly on the green circle.
\end{itemize}
\vspace{\baselineskip}
Scene:
\begin{itemize}
  \item No change in scene composition.
  \item No change in the layout of the maze.
  \item The red circle travels along the white path without speeding up or slowing down.
\end{itemize}
\vspace{\baselineskip}
Camera:
\begin{itemize}
  \item Static camera.
  \item No zoom.
  \item No pan.
  \item No glitches, noise, or artifacts.
\end{itemize}
\end{tcolorbox}

\begin{tcolorbox}[colback=gray!20, arc=5pt, boxrule=0pt]
\textbf{\geminipro{} I2T}\\
\textcolor{black!70}{\textsc{system}}\\
\itshape
Think step by step as needed and output in xml format:\\
$<$think$>$thinking process$<$/think$>$\\
$<$final\_answer$>$final answer$<$/final\_answer$>$\\

\textcolor{black!70}{\textsc{user}}\\
The following image shows a maze, represented by colored squares:
\begin{itemize}
  \item Black squares represent walls and cannot be passed through.
  \item White squares are empty and can be passed through.
  \item The red square is the starting point.
  \item The green square is the end point.
\end{itemize}
\vspace{\baselineskip}
Please solve the maze by providing a path from the starting point to the end point. The path should be provided as a list of coordinates of each step, where each coordinate is a (row, col) tuple, and row, col are 0-based indices. Consider the origin (0, 0) to be the top-left corner. Overall, the path should be provided in the format of [(row1, col1), (row2, col2), ...].\\

A valid path must:
\begin{itemize}
  \item Start at the starting point (the red square).
  \item End at the end point (the green square).
  \item Avoid the walls (the black squares).
  \item Pass only through empty space (the white squares).
  \item Move one square at a time.
  \item Only move up, down, left, and right, not diagonally.
\end{itemize}

Correct your answer if you spot any errors.\\

Here is the maze image: \{image\}
\end{tcolorbox}

\begin{tcolorbox}[colback=gray!20, arc=5pt, boxrule=0pt]
\textbf{\nanobanana{}}\\
\itshape
Mark the correct path from the red to the green circle through the maze in blue.
\end{tcolorbox}

\begin{tcolorbox}[colback=gray!20, arc=5pt, boxrule=0pt]
\textbf{\geminipro{} T2T}\\
\textcolor{black!70}{\textsc{system}}\\
\itshape
Think step by step as needed and output in xml format:\\
$<$think$>$thinking process$<$/think$>$\\
$<$final\_answer$>$final answer$<$/final\_answer$>$\\

\textcolor{black!70}{\textsc{user}}\\
The following is an ASCII-representation of a maze:
\begin{itemize}
  \item `\#' represents walls which cannot be passed through.
  \item ` ' represents empty spaces that can be passed through.
  \item `S' is the starting point.
  \item `E' is the end point.
\end{itemize}
\vspace{\baselineskip}
Please solve the maze by providing a path from the starting point to the end point. The path should be provided as a list of coordinates of each step, where each coordinate is a (row, col) tuple, and row, col are 0-based indices. Consider the origin (0, 0) to be the top-left corner. Overall, the path should be provided in the format of [(row1, col1), (row2, col2), ...].\\

A valid path must:
\begin{itemize}
  \item Start at the starting point `S'.
  \item End at the end point `E'.
  \item Avoid the walls `\#'.
  \item Pass only through empty space ` '.
  \item Move one square at a time.
  \item Only move up, down, left, and right, not diagonally.
\end{itemize}

Correct your answer if you spot any errors.\\

Here is the maze in ASCII format: \{maze\}
\end{tcolorbox}

\paragraph{Sampling} We generated 10 videos per sample with a fixed prompt. Note that for Gemini 2.5 Pro I2T, we represented the maze as a grid where the red and green positions are marked as squares (not circles) to make the setup grid-like (i.e., a matrix with cells), since this might be easier for a language model.

\subsection{\textcolor{reasoningcol}{\textbf{Reasoning}}: Visual symmetry solving}
\label{app:symmetry}
We provide details for the visual symmetry task in \cref{subsec:visual_symmetry}. 

\paragraph{Evaluation} We prompt \veo{} with input images containing a 10$\times$16 grid where a pattern is drawn on the left half. The goal is to complete the pattern on the empty right half so that the final pattern is symmetrical along the central vertical axis.

We compare \veo{}'s best-frame and last-frame solutions with the ground-truth symmetrical grid and compute the number of incorrectly-colored cells. A cell is determined as incorrectly-colored if the average color across pixels in the cell is perceptually distinct from the ground-truth average color in the matching cell. We compute perceptual color differences of the average cell color in the CIELAB color space, with a difference threshold of 15.0. In \cref{fig:symmetry}, we report the percentage of attempts in which the best or last frame solution has zero incorrect cells for $k=1$.

\paragraph{Dataset}
We created a synthetic grid coloring image dataset to evaluate visual symmetry.  We generated 25 samples using common symmetrical symbols, objects and shapes such as english letters (e.g., A, H, M, X), geometric shapes (e.g., square, triangle), symmetrical objects (e.g., wineglass, balloon; together, the \textit{shape} condition). We also generated 25 samples consisting of randomly-colored cells (the \textit{random} condition).

\paragraph{Models \& prompts}
We tested \veothree{} \textcolor{GoogleGray}{\small\texttt{veo-3.0-generate-preview}} and \veotwo{} \textcolor{GoogleGray}{\small\texttt{veo-2.0-\allowbreak{}generate-\allowbreak{}preview-001}} through the Vertex~AI API. We also tested \nanobanana{} \textcolor{GoogleGray}{\small\texttt{gemini-2.5-\allowbreak{}flash-\allowbreak{}image-\allowbreak{}preview}} through Google~AI~Studio.

\begin{tcolorbox}[colback=gray!20, arc=5pt, boxrule=0pt]
\textbf{\veo{}}\\
\itshape
Instantly reflect this pattern along the central, vertical axis while keeping the existing colored pattern without modification. Static camera perspective, no zoom or pan.
\end{tcolorbox}

\paragraph{Sampling} We generated 10 videos per sample with a fixed prompt.

\subsection{\textcolor{reasoningcol}{\textbf{Reasoning}}: Visual analogy completion}
\label{app:visual_analogy}
We provide details for the visual analogy task in \Cref{subsec:visual_analogy}. 

\begin{figure}
    \centering
    \includegraphics[width=\linewidth]{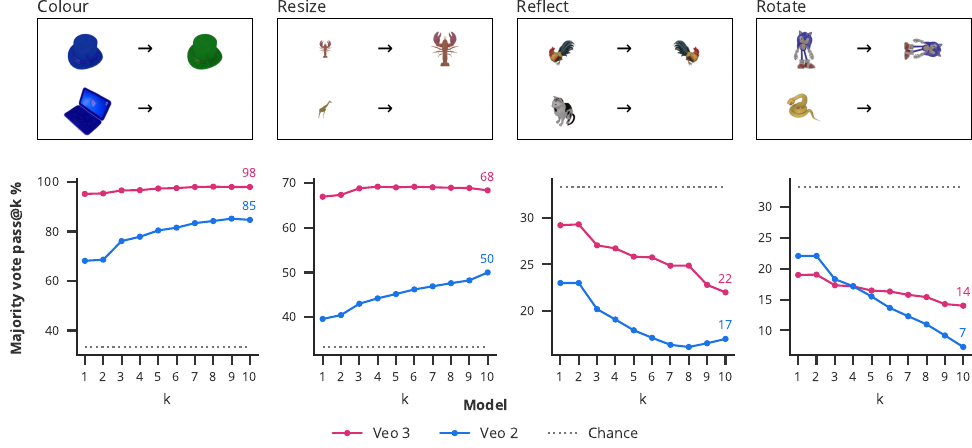}
    \caption{\textbf{Visual analogy performance over 10 attempts}. In contrast to other plots in this paper, we here report not the best performance over $k$ attempts, but instead the performance when choosing the \emph{majority vote} from $k$ attempts. As a result, performance is not necessarily monotonic in $k$. In fact, for \emph{reflect} and \emph{rotate}, performance \emph{decreases} with $k$, indicating that both models have systematic, erroneous biases. In the case of \veothree{}, the model tends to perform reflections and rotations, but not along the same axis as shown in the image. \veotwo{} simply tends to copy the object without applying any transformation.}
    \label{fig:analogy-k10}
\end{figure}

\paragraph{Evaluation} We prompt \veo{} to solve visual analogies with an input image showing a reference object pair and a test object. The object images are sourced from the Kid-inspired Visual Analogies benchmark~\citep[KiVA,][]{yiu2024kiva}. Consistent with the multi-choice format in the KiVA benchmark, we evaluated \veo{}'s generation by cropping out the generated target object in the lower-right region of the last frame and compare \veo{}'s generated object with three candidate object choices using an autorater (see details below).

In \cref{fig:visual-analogy-kiva}, we report the pass@1 accuracy across different conditions for both \veotwo{} and \veothree{} for $k=1$. \cref{fig:analogy-k10} shows performance for $k=10$.

\paragraph{Dataset}
We used the test trials and choice images from the KiVA benchmark~\citep{yiu2024kiva}.

\paragraph{Models \& prompts}
We tested \veothree{} \textcolor{GoogleGray}{\small\texttt{veo-3.0-generate-preview}} and \veotwo{} \textcolor{GoogleGray}{\small\texttt{veo-2.0-\allowbreak{}generate-\allowbreak{}preview-001}} through the Vertex~AI API.

We used \geminipro{} \textcolor{GoogleGray}{\small\texttt{gemini-2.5-pro}} through Google~AI~Studio to identify which image choice \veo{}'s generation is most similar with. To enhance the autorater's image comparison accuracy for this task, Gemini is prompted with privileged information about the values in the dataset conditions (see below for the full autorater prompt). If no object is visible in the lower-right region of \veo{}'s generated last frame or if the generated object is of a different object type, we randomly sampled one of three choices as \veo{}'s choice. In pilot experiments, we found that the Gemini-assisted autorater's ratings achieve above 88\% agreement with expert human ratings by the authors on 25 samples within each conditions.

Note that in the prompt, words in \{~\} are updated based on the test condition of the current generation (one of \emph{color}, \emph{resize}, \emph{reflect}, and \emph{rotate}) to provide more information of the feature name and values to direct the image comparison. Image choice orders are shuffled for each prompt.

\begin{tcolorbox}[colback=gray!20, arc=5pt, boxrule=0pt]
\textbf{\veo{}}\\
\itshape
Create a smooth animation to generate the missing object in the lower right region and solve the visual analogy. The original three objects must remain still. Static shot, no zoom no pan no dolly.
\end{tcolorbox}

\begin{tcolorbox}[colback=gray!20, arc=5pt, boxrule=0pt]
\textbf{\geminipro{} autorater}\\
\textcolor{black!70}{\textsc{system}}\\
\itshape
You are an expert visual judge. You will be presented with a "target image" and three "choice images" labeled A, B, and C. Your goal is to identify the choice image that is most visually similar to the target image. \\ \\
Follow these steps:
\begin{enumerate}
    \item Analyze each provided image and describe the objects shown. Focus on the object \{color\}. That is, if the objects appear \{green\}, \{blue\}, or \{red\}.
    \item Determine if the primary object in the target image is of the same general category or type as the objects in the choice images. For example, if the target image shows a dog, and the choices show a cat, the object types are considered different. If no object is visible in the target image, the object type is considered to be mismatched. 
    \item If the object type matches between the target image and the choice images, identify the choice that is most visually similar to the target image in terms of the object \{color\}.
\end{enumerate} 
Provide a brief justification for your choice, explaining why it is the best match and why the others are less suitable. Conclude your response with the final answer on a new line in the format: \\
``Final Answer: [answer]'' \\
where ``answer'' is one of (``A'', ``B'', ``C'', or ``different object type''). Do not use markdown format for the final answer line.\\

\normalfont
\textcolor{black!70}{\textsc{user}}\\
\itshape
Please evaluate the following images.\\
--- TARGET IMAGE --- \\
\{target object image\} \\
--- CHOICE IMAGES --- \\
CHOICE A: \{image choice\} CHOICE B: \{image choice\} CHOICE C: \{image choice\} \\
Which choice image is most similar to the target image?
\end{tcolorbox}

\paragraph{Sampling} We generated 10 videos per sample with a fixed prompt.

\newpage

\section{Prompting best practices}
\label{app:prompt-sensitivity}

\begin{table}[ht]
\small
\centering
\caption{\textbf{Prompt sensitivity study on the visual symmetry task}. We report best frame pass@1~\% and the average number of incorrectly-colored cells across 25 samples on each split (shape/random).}
\begin{tabularx}{\textwidth}{r >{\raggedright\arraybackslash}X r r r r}
\toprule
& & \multicolumn{2}{c}{\textbf{Pass@1}} & \multicolumn{2}{c}{\textbf{Avg incorrect cells}}\\
\cmidrule(lr){3-4} \cmidrule(lr){5-6}
\textbf{No.} & \textbf{Prompt} & \textbf{Shape} & \textbf{Random} & \textbf{Shape} & \textbf{Random} \\
\midrule
1 & Instantly reflect this pattern along the central, vertical axis while keeping the existing colored pattern without modification. & 48 & 68 & 4.16 & 7.00 \\
2 & Instantly reflect this pattern along the central, vertical axis while keeping the existing colored pattern without modification. Static camera perspective, no zoom or pan. & 42 & 65 & 5.00 & 3.52\\
3 & Instantly reflect this pattern along the central, vertical axis while keeping the existing colored pattern without modification. The result needs to be mirror-symmetrical along the vertical axis. Static camera perspective, no zoom or pan. & 36 & 52 & 6.28 & 9.04 \\
4 & One by one, cells in the right half of the grid are filled in to complete the pattern. The pattern is mirror-symmetrical along the central vertical line. Static shot; no zoom, no pan, no dolly. & 32 & 12 & 10.76 & 14.08 \\
5 & Reflect this pattern along the central, vertical axis. & 28 & 40 & 9.76 & 4.52 \\
6 & An animation showing the left half of the grid being mirrored onto the right half to create a symmetrical pattern. Static shot; no zoom, no pan, no dolly. & 24 & 12 & 10.96 & 16.32 \\
7 & You're a master symmetry solver. Your task is to fill the cells on the right side of the grid to mirror the pattern on the left, such that it's symmetrical along the vertical axis. & 24 & 8 & 9.20 & 17.72 \\
8 & Fill color in the appropriate cells on the right side of the grid to complete the pattern. The final image should be symmetrical along the central vertical line. Static shot, no zoom no pan no dolly. & 13 & 9 & 10.30 & 14.74 \\
9 & Create a static, smooth, realistic animation completing the pattern in the image by filling the grid on the right hand side. Do not change anything else. No zoom, no pan. & 12 & 4 & 14.88 & 21.00 \\
10 & A timelapse of a professional pixel artist drawing a symmetrical pattern onto a white canvas. Static shot; no zoom, no pan, no dolly. & 8 & 20 & 14.20 & 12.64 \\
\bottomrule
\end{tabularx}
\label{tab:symmetry_prompt_comparison}
\end{table}

The results in \cref{sec:qualitative_results,sec:quantitative_results} are best-effort estimates of \veo{}'s performance using carefully chosen prompts. Generally, performance varies greatly with the exact task description provided in the prompt, as illustrated by a prompt sensitivity study on the visual symmetry task in \cref{tab:symmetry_prompt_comparison}. Here are best practices from this sensitivity analysis and our other experiments:

\begin{itemize}
    \item \textbf{Remove ambiguity}. Tasks can be solved in a variety of ways, and natural language descriptions tend to leave a lot of room for interpretation. The goal should be formulated clearly, e.g., saying ``symmetrical along the central, vertical axis'', rather than just ``symmetrical''.
    \item \textbf{Specify what shouldn't change}. \veo{} has a tendency to change any part of the input to create interesting, dynamic scenes. Including not only a positive task description, but also specifying what \emph{not} to change can help mitigate this, e.g., ``keep the existing colored pattern without modification''.
    \item \textbf{Providing an outlet}. As mentioned above, \veo{} has a strong prior to keep things moving. Providing a ``motion outlet'' in the form of, e.g., a spinning ball can help keep the rest of the scene static.
    \item \textbf{Let the model decide when its done}. The motion prior also means that \veo{} often keeps modifying the scene, even after solving the task. Providing a visual indicator, e.g., ``add a glowing red dot once the goal is reached'' allows for easy extraction of the solution from the generated video.
    \item \textbf{Scene and camera controls}. Phrases like ``static camera, no zoom, no pan, no dolly'' can help keeping the scene static, e.g., for image-to-image tasks.
    \item \textbf{Speed control}. Some tasks like maze solving benefit from being solved step-by-step. For other tasks, especially image-to-image tasks, specifying instant changes can help avoid artifacts.
    \item \textbf{Realism}. \veo{} was trained to generate plausible, realistic-looking videos. Translating an abstract task into a realistic setting (including, but not limited to editing the original image to depict realistic, 3D scenes rather than abstract shapes) can improve generation results. A similar effect was observed in \citep{jeong2025diffusion}, and we expect \emph{visual prompt engineering} to emerge as a powerful tool for video models.
\end{itemize}

\newpage
\section{Failure cases}
\label{app:failure_cases}

\begin{figure}[H]
    \centering
    \begin{subfigure}[t]{\threefigwidth\textwidth}
        \centering
        \fbox{\includegraphics[width=\linewidth]{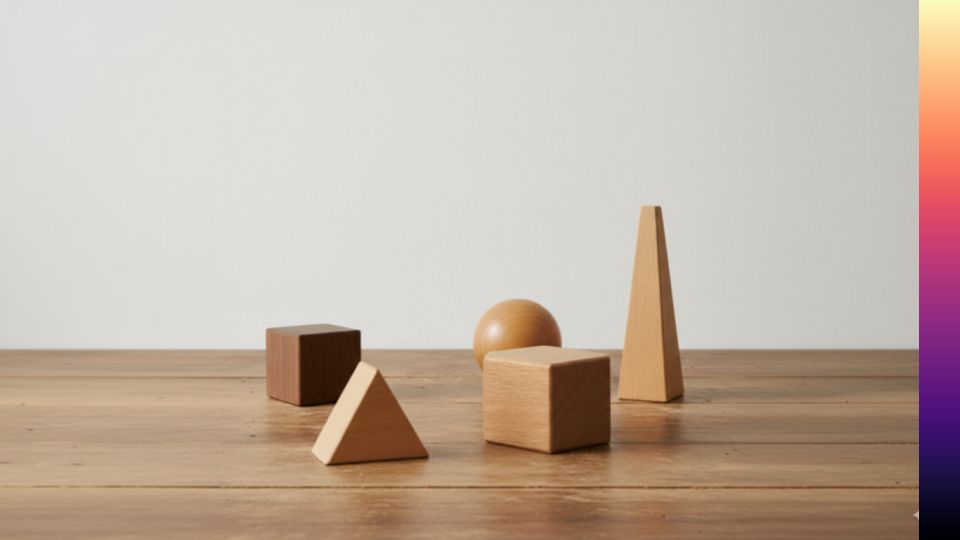}}
    \end{subfigure}
    \hfill
    \begin{subfigure}[t]{\threefigwidth\textwidth}
        \centering
        \fbox{\includegraphics[width=\linewidth]{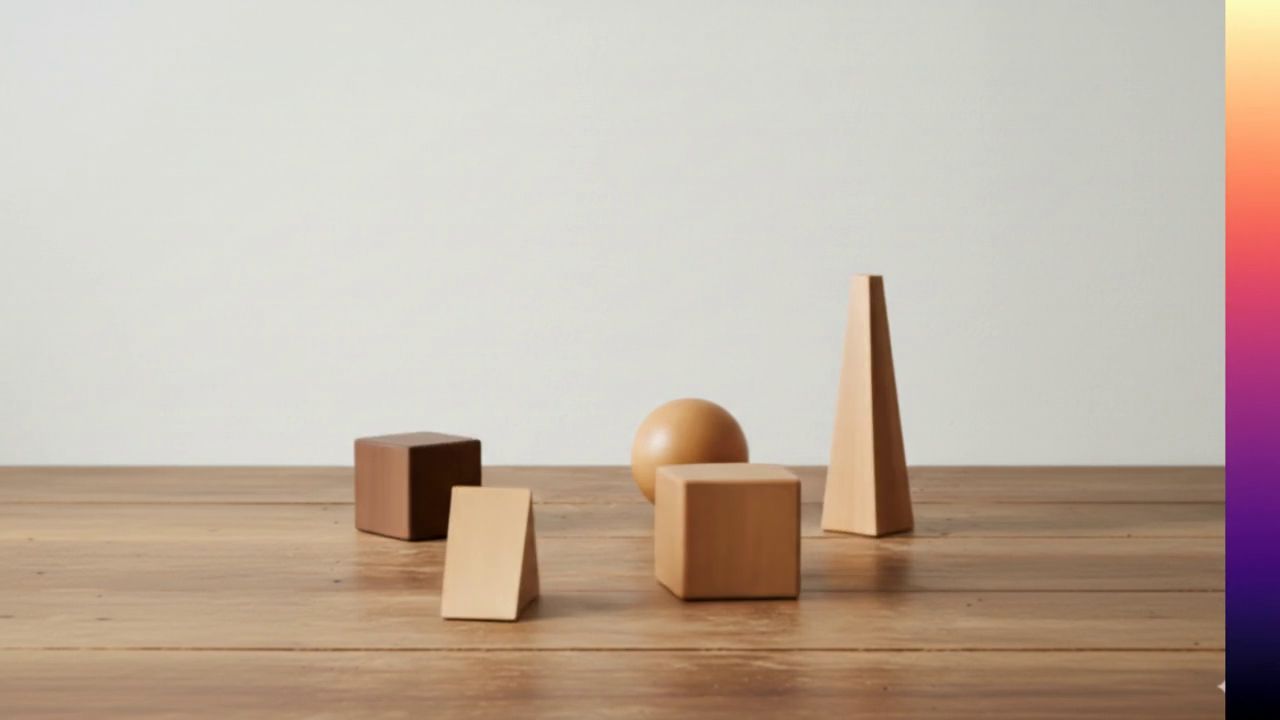}}
    \end{subfigure}
    \hfill    
    \begin{subfigure}[t]{\threefigwidth\textwidth}
        \centering
        \fbox{\includegraphics[width=\linewidth]{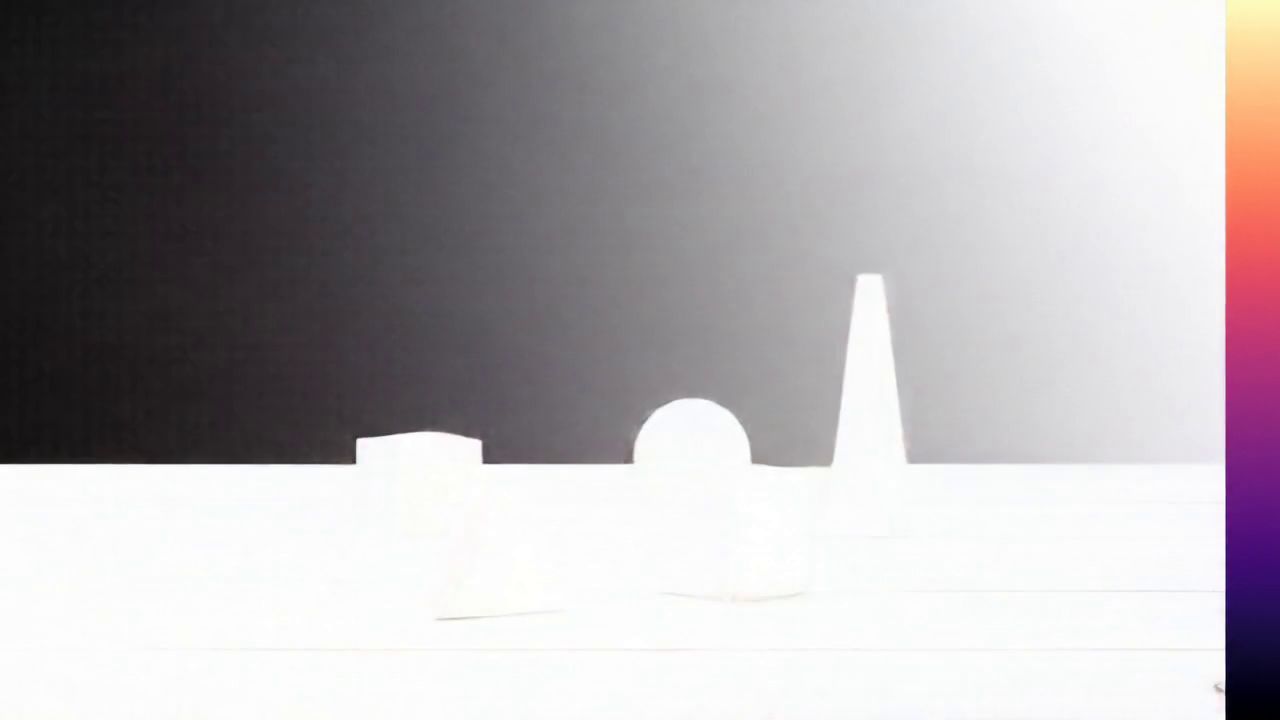}}
    \end{subfigure}
    \caption{\textbf{Monocular depth estimation}. Prompt: \textit{``The image transitions to a depth-map of the scene: Darker colors represent pixels further from the camera, lighter colors represent pixels closer to the camera. The exact color map to use is provided on the right side of the image. Static scene, no pan, no zoom, no dolly.''} Failure: \veothree{} seems generally unable to color pixels by depth beyond a binary foreground/background mapping and specifically struggles with using a provided color map.}
    \label{fig:depth}
\end{figure}

\begin{figure}[H]
    \centering
    \begin{subfigure}[t]{\threefigwidth\textwidth}
        \centering
        \fbox{\includegraphics[width=\linewidth]{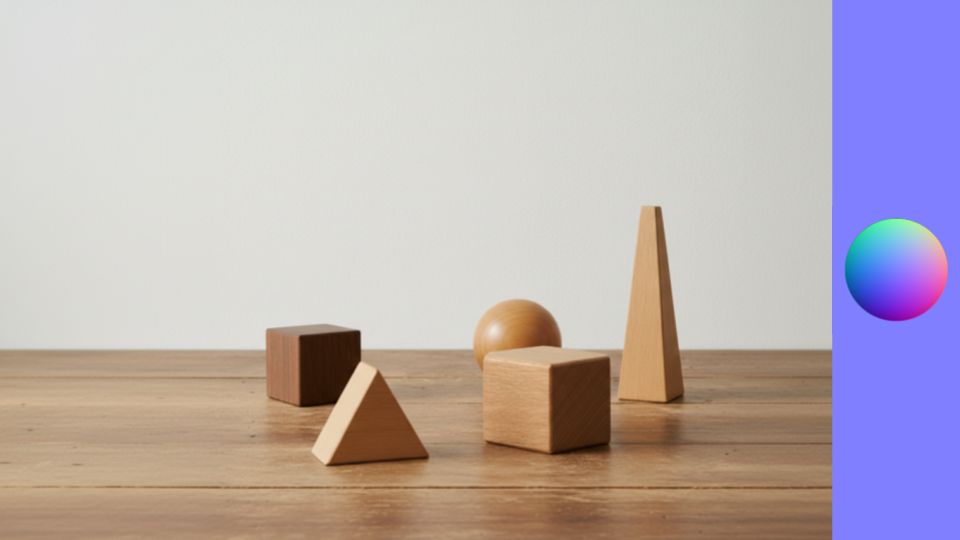}}
    \end{subfigure}
    \hfill
    \begin{subfigure}[t]{\threefigwidth\textwidth}
        \centering
        \fbox{\includegraphics[width=\linewidth]{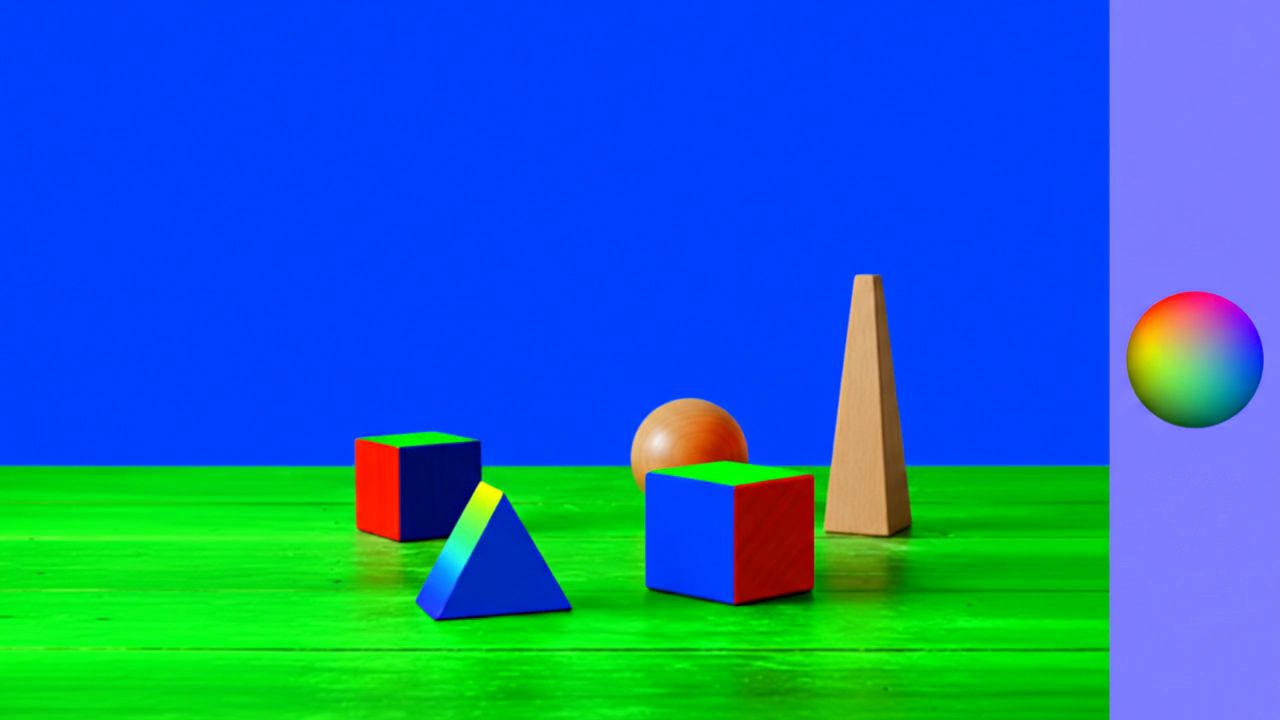}}
    \end{subfigure}
    \hfill    
    \begin{subfigure}[t]{\threefigwidth\textwidth}
        \centering
        \fbox{\includegraphics[width=\linewidth]{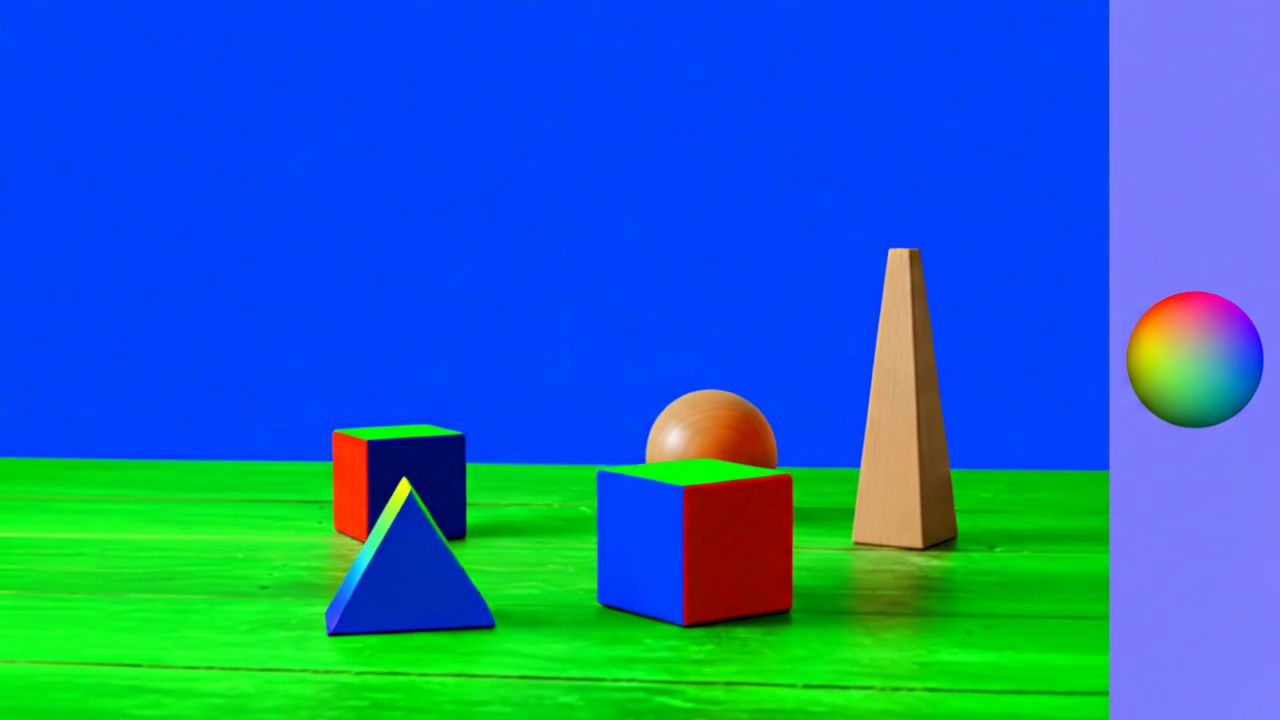}}
    \end{subfigure}
    \caption{\textbf{Monocular surface normal estimation}. Prompt: \textit{``The image transitions to a surface-normal map of the scene: the red/green/blue color channel specify the direction of the surface-normal at each point, as illustrated on the right side of the image on a sphere. Static scene, no pan, no zoom, no dolly.''} Failure: While \veothree{} shows some promise in coloring surfaces according to their orientation (e.g., the cube in the front), coloration is inconsistent (compare the two cubes) and doesn't correctly interpolate colors (e.g., for the slope on the triangle).}
    \label{fig:normals}
\end{figure}

\begin{figure}[H]
    \centering
    \begin{subfigure}[t]{\threefigwidth\textwidth}
        \centering
        \fbox{\includegraphics[width=\linewidth]{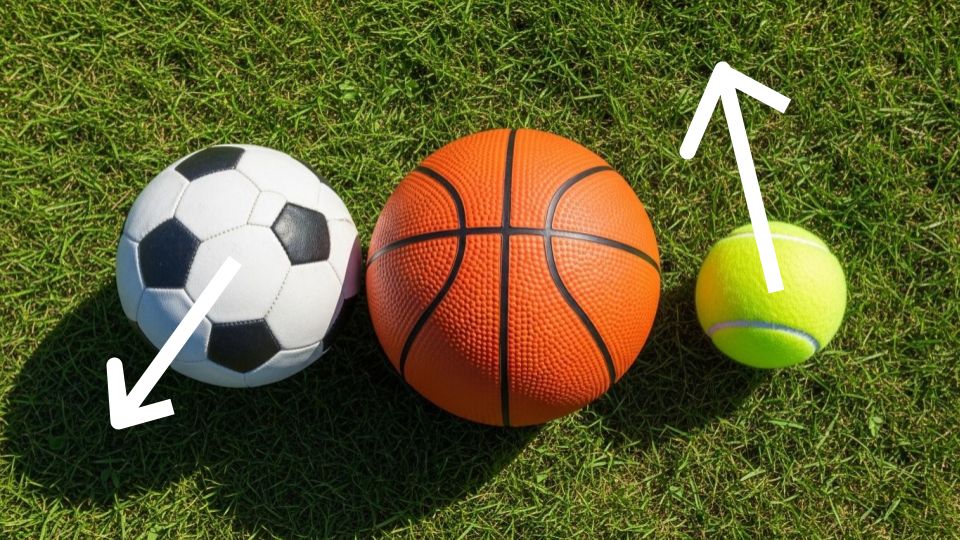}}
    \end{subfigure}
    \hfill
    \begin{subfigure}[t]{\threefigwidth\textwidth}
        \centering
        \fbox{\includegraphics[width=\linewidth]{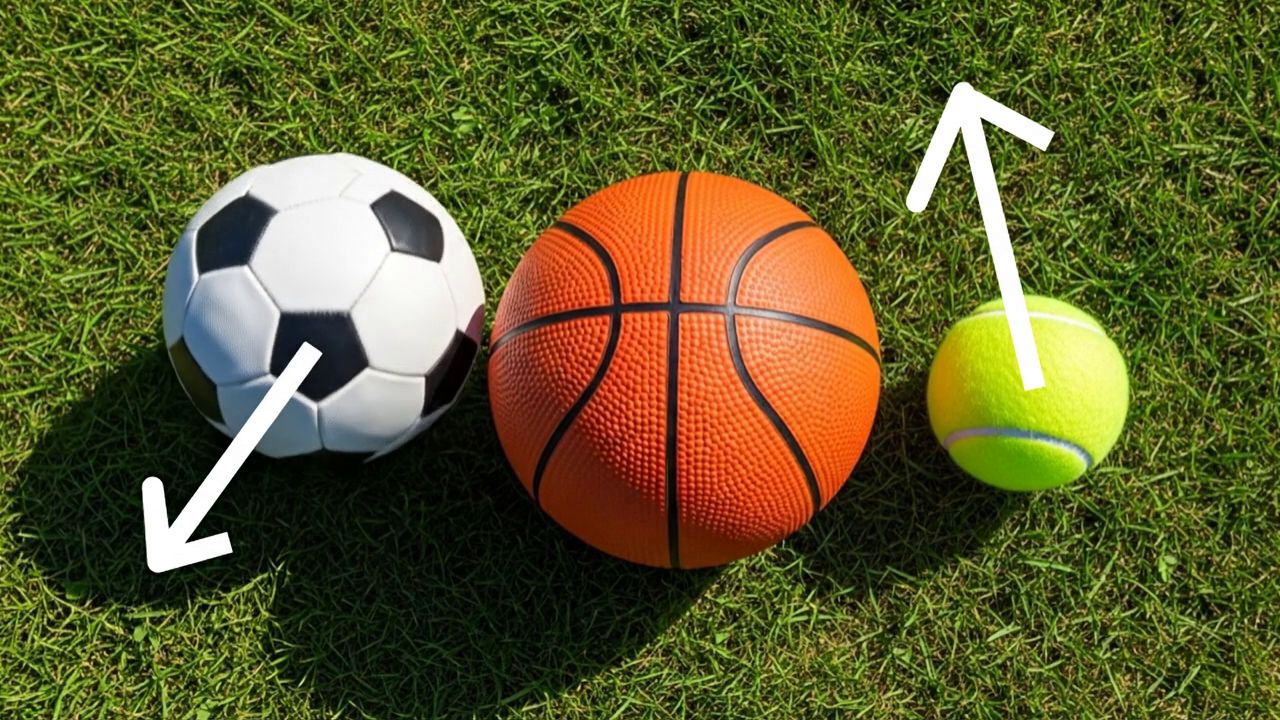}}
    \end{subfigure}
    \hfill    
    \begin{subfigure}[t]{\threefigwidth\textwidth}
        \centering
        \fbox{\includegraphics[width=\linewidth]{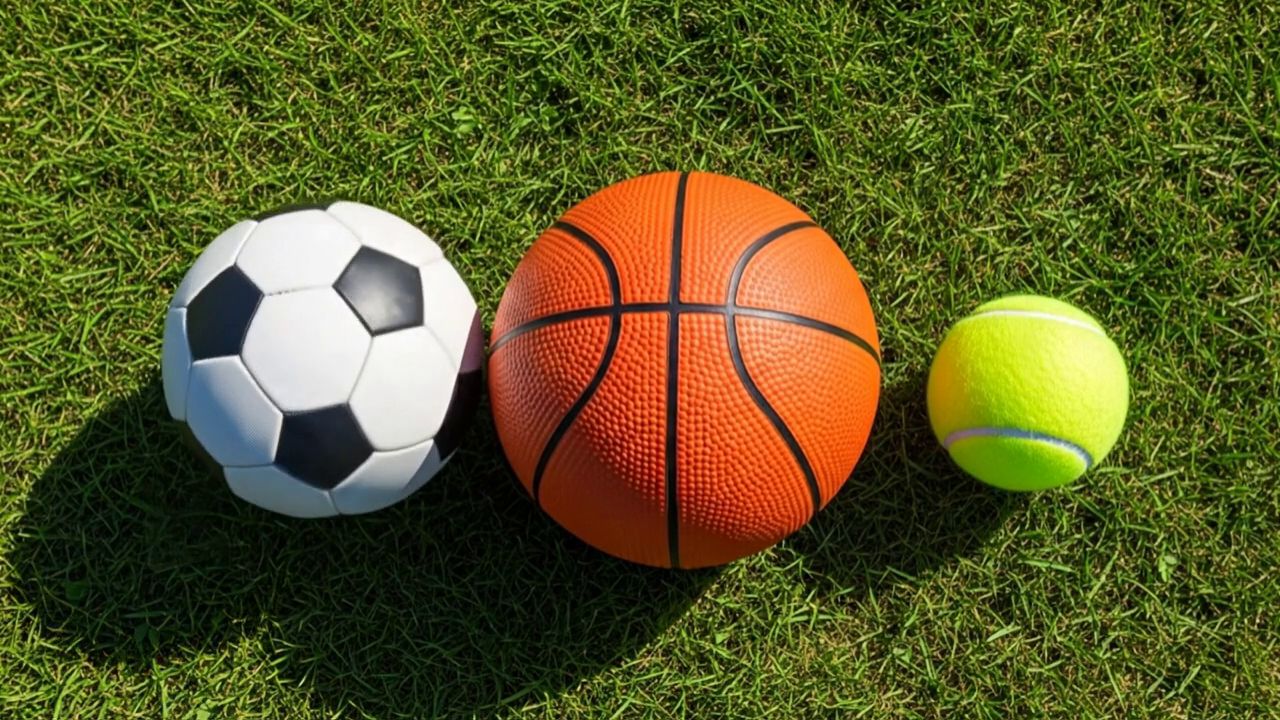}}
    \end{subfigure}

    \vspace{0.2cm} %

    \centering
    \begin{subfigure}[t]{\threefigwidth\textwidth}
        \centering
        \fbox{\includegraphics[width=\linewidth]{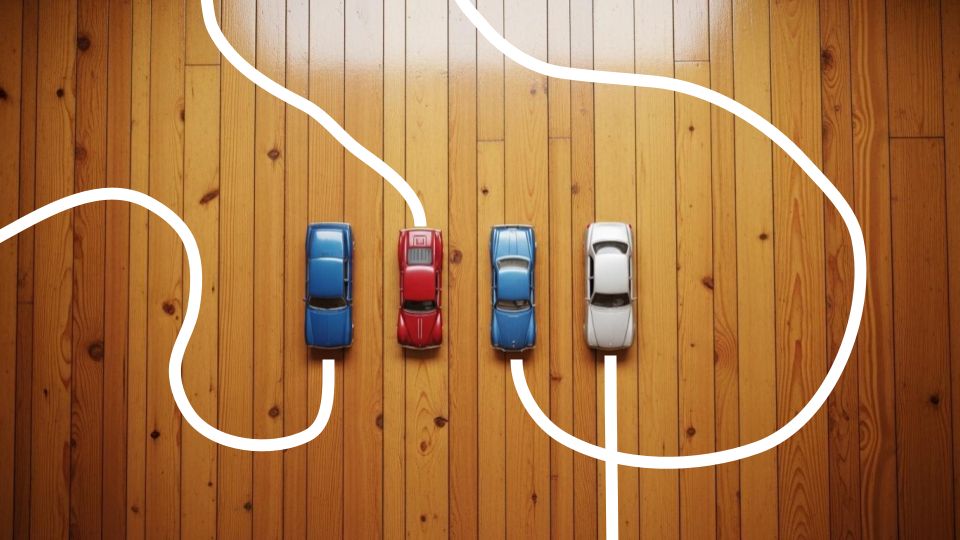}}
    \end{subfigure}
    \hfill
    \begin{subfigure}[t]{\threefigwidth\textwidth}
        \centering
        \fbox{\includegraphics[width=\linewidth]{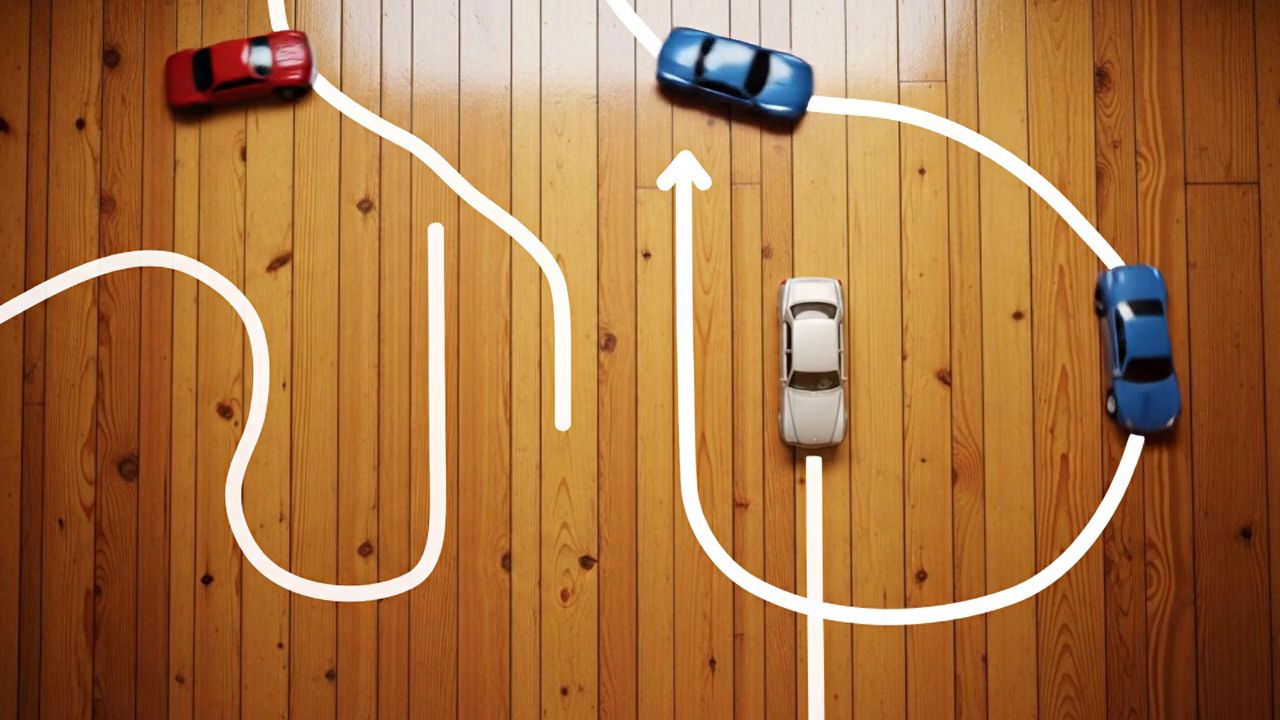}}
    \end{subfigure}
    \hfill    
    \begin{subfigure}[t]{\threefigwidth\textwidth}
        \centering
        \fbox{\includegraphics[width=\linewidth]{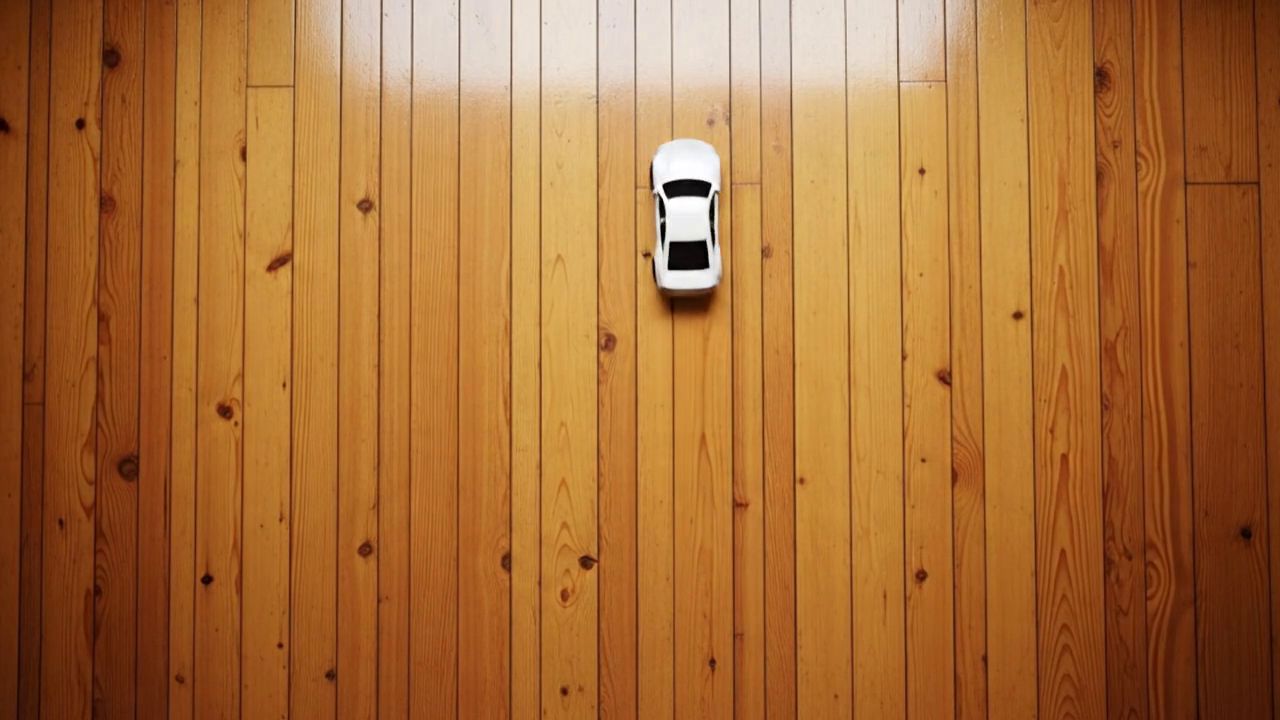}}
    \end{subfigure}
    \caption{\textbf{Force \& motion prompting}, inspired by~\citep{gillman2025forcepromptingvideogeneration, geng2024motionprompting}.
    \textbf{Force prompting} (top). Prompt: \textit{``The balls move in the direction indicated by the arrows. Balls without an arrow don't move. Static scene, no pan, no zoom, no dolly.''}
    \textbf{Motion trajectory prompting} (bottom). Prompt: \textit{``Each car drives out of the frame following the indicated trajectory. Static camera, no zoom, no pan, no dolly.''}
    Failure: \veothree{} seems unable to follow force/motion annotations with any consistency. Providing annotations for the first frame and letting the model remove them before generating the scene in motion does not work, either.
    }
    \label{fig:force-motion-prompting}
\end{figure}

\begin{figure}[H]
    \centering
    \begin{subfigure}[t]{\threefigwidth\textwidth}
        \centering
        \fbox{\includegraphics[width=\linewidth]{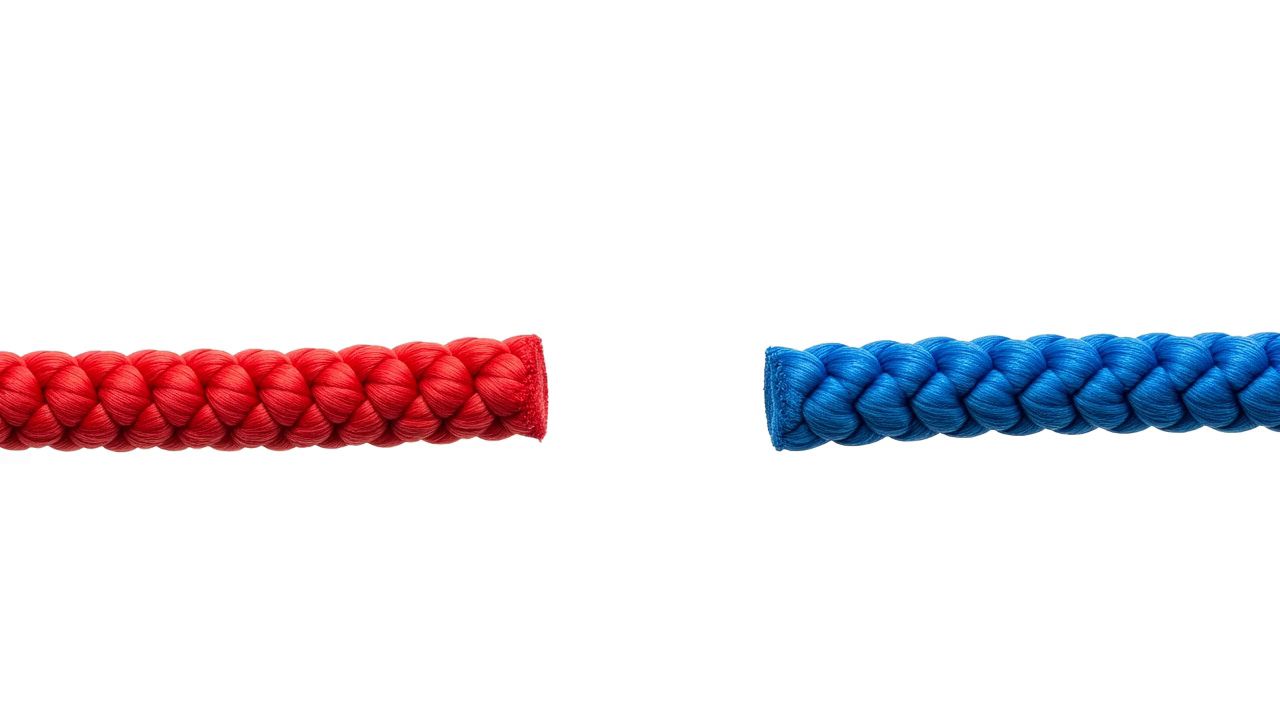}}
    \end{subfigure}
    \hfill
    \begin{subfigure}[t]{\threefigwidth\textwidth}
        \centering
        \fbox{\includegraphics[width=\linewidth]{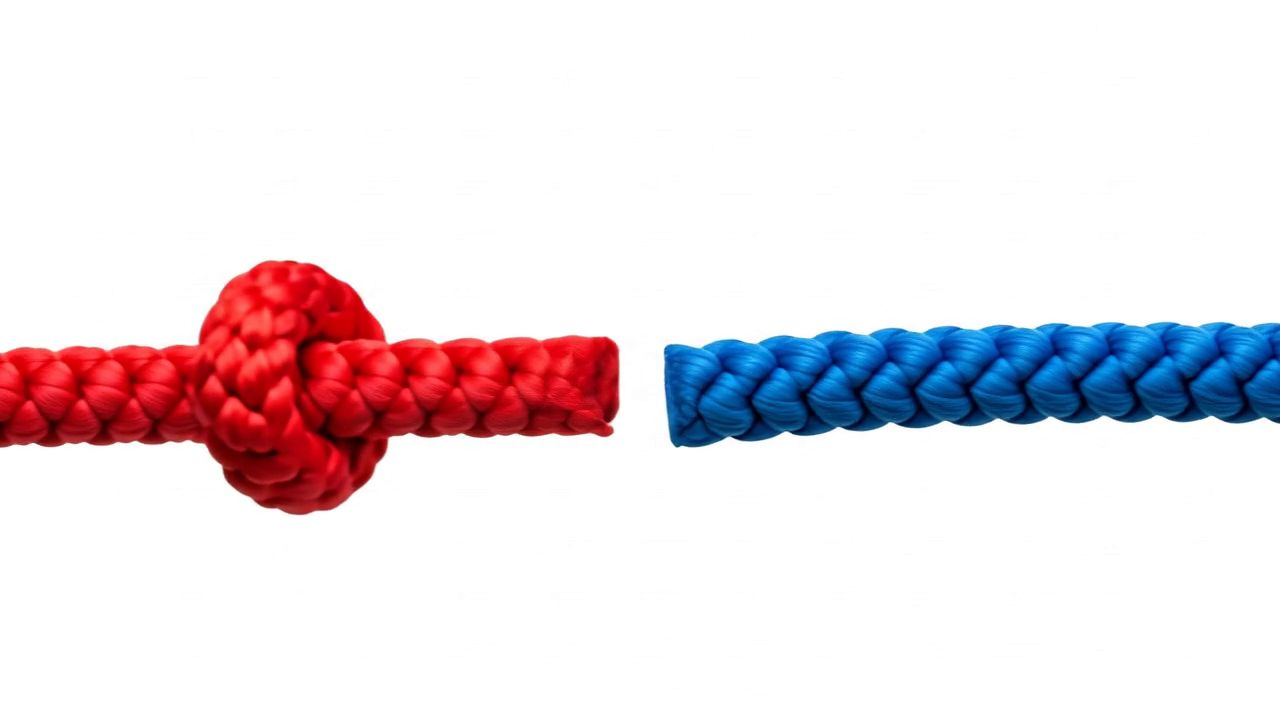}}
    \end{subfigure}
    \hfill    
    \begin{subfigure}[t]{\threefigwidth\textwidth}
        \centering
        \fbox{\includegraphics[width=\linewidth]{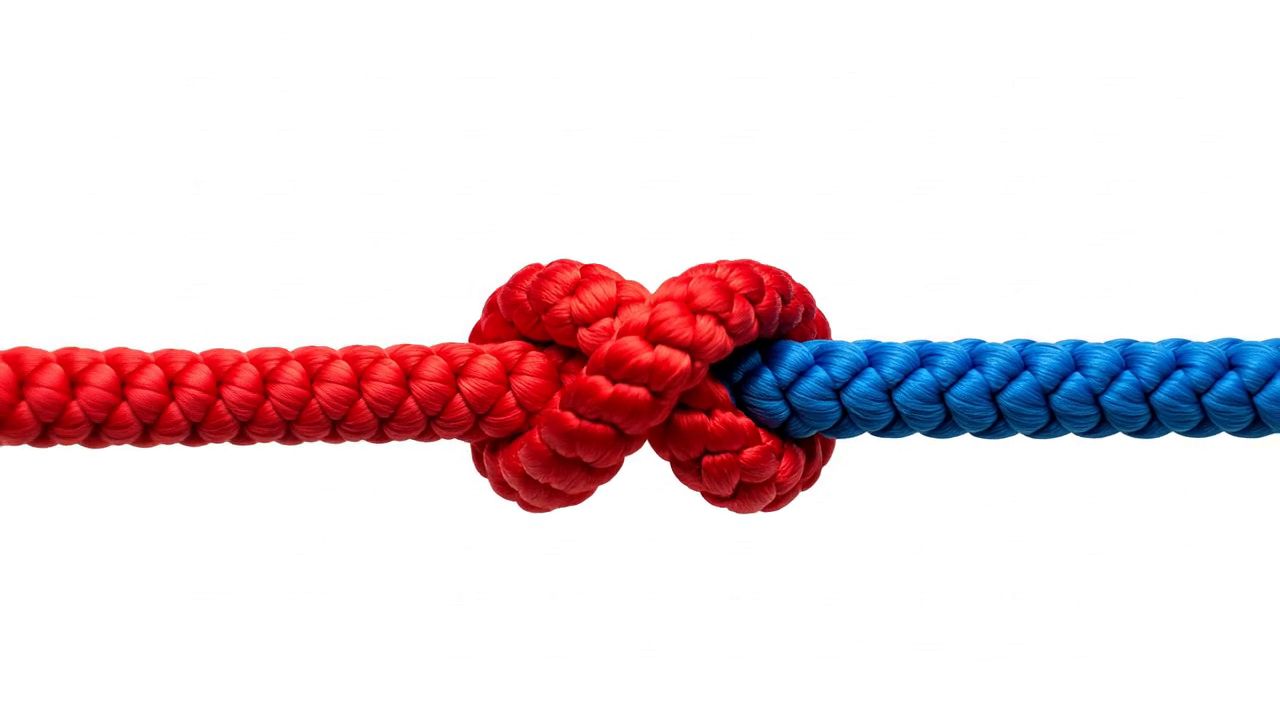}}
    \end{subfigure}
    \caption{\textbf{Tying the knot}. Prompt: \textit{``A knot is tied connecting these two rope ends.''} Failure: physics violation, impossible rope movement.}
    \label{fig:knots}
\end{figure}

\begin{figure}[H]
    \centering
    \begin{subfigure}[t]{\threefigwidth\textwidth}
        \centering
        \fbox{\includegraphics[width=\linewidth]{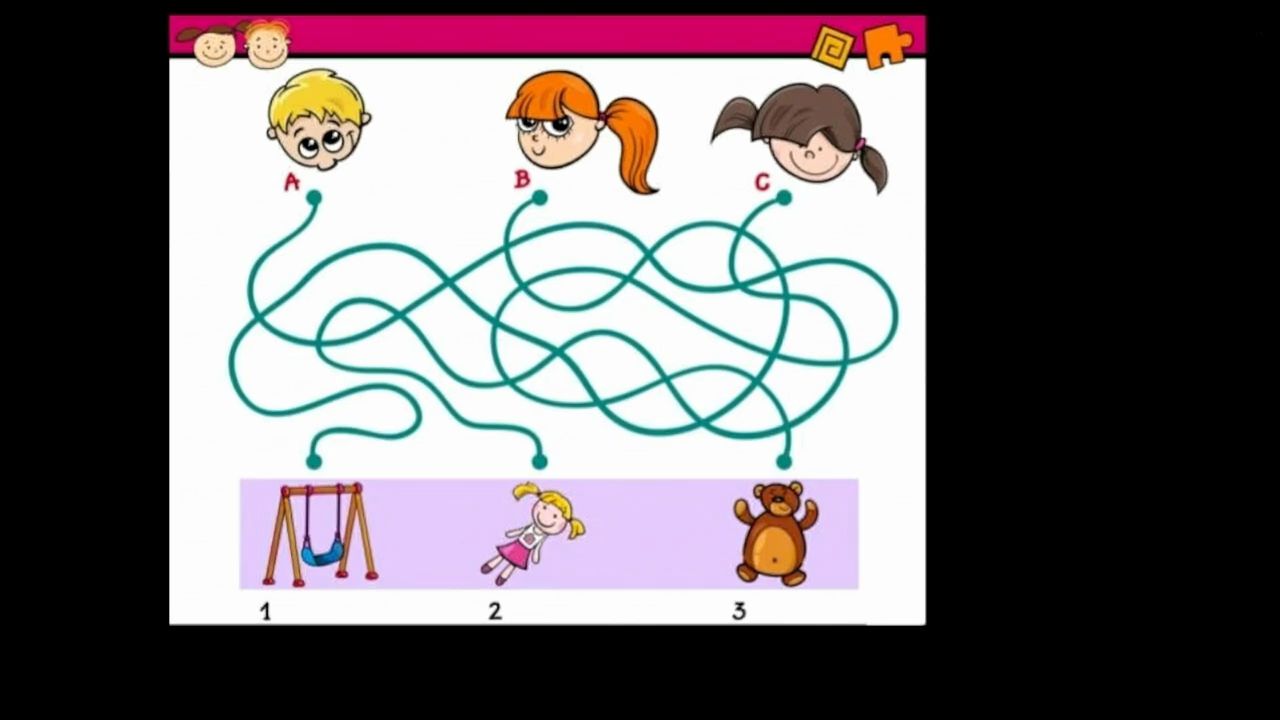}}
    \end{subfigure}
    \hfill
    \begin{subfigure}[t]{\threefigwidth\textwidth}
        \centering
        \fbox{\includegraphics[width=\linewidth]{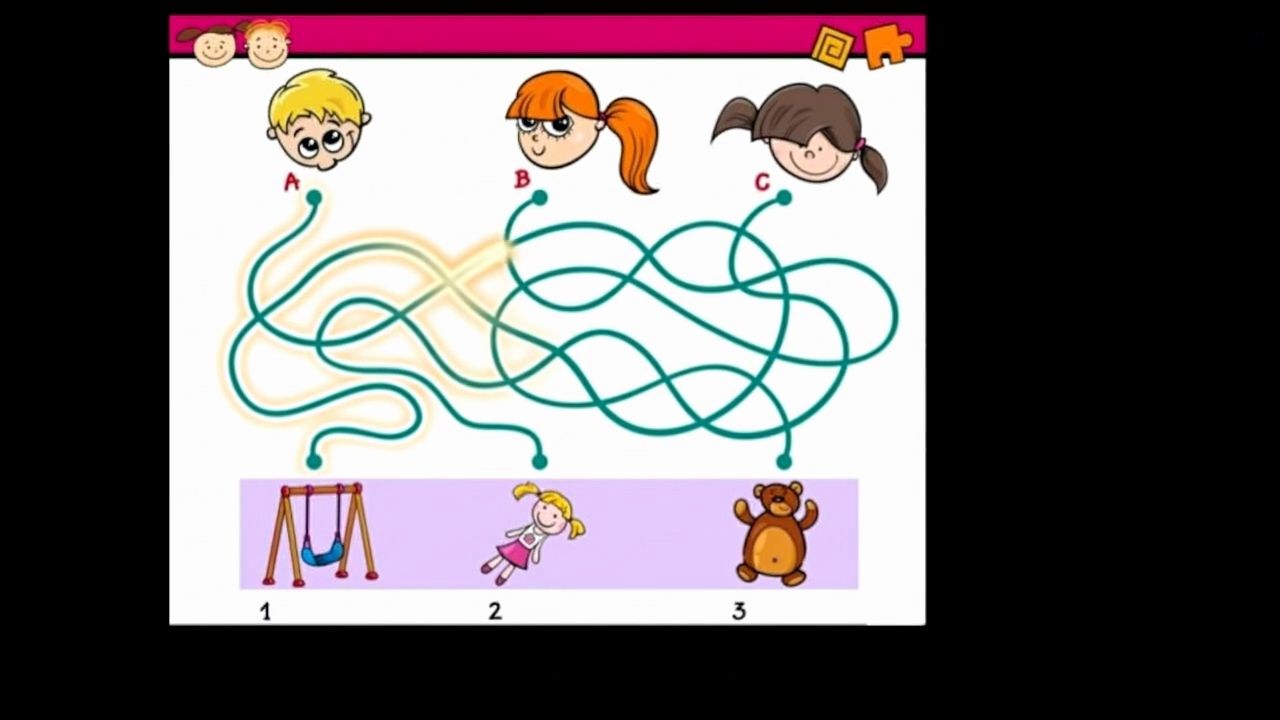}}
    \end{subfigure}
    \hfill    
    \begin{subfigure}[t]{\threefigwidth\textwidth}
        \centering
        \fbox{\includegraphics[width=\linewidth]{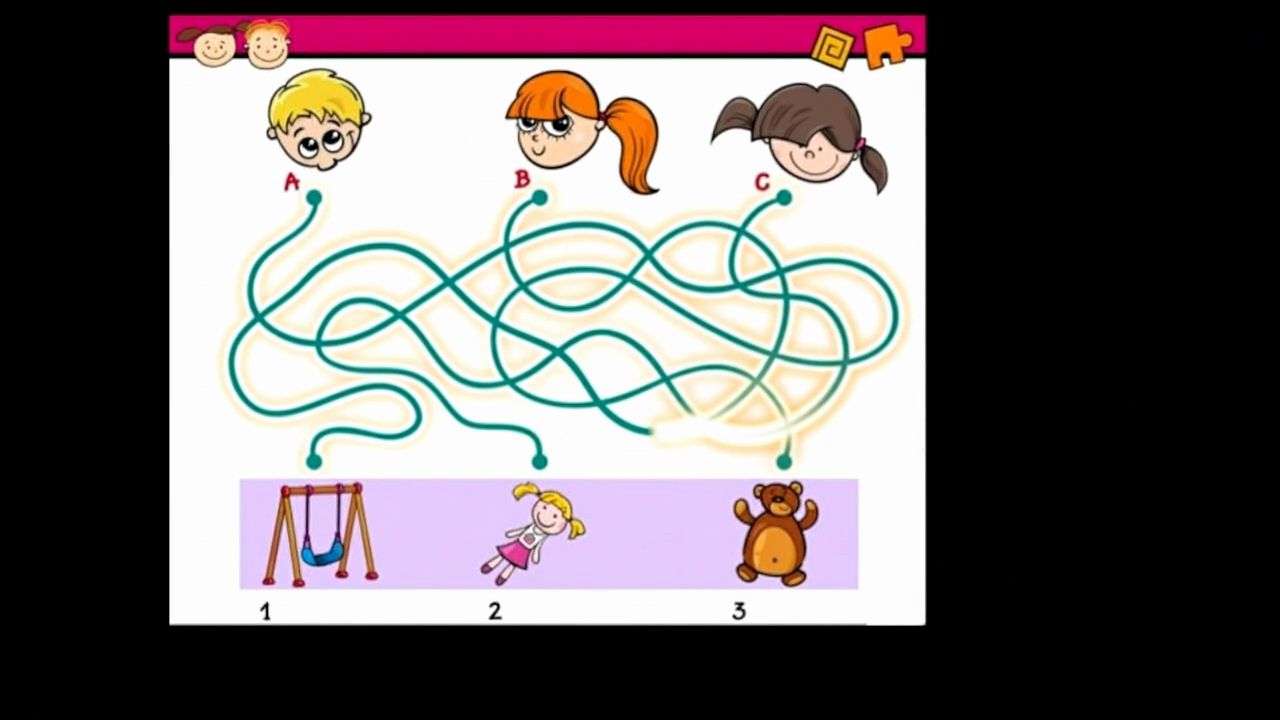}}
    \end{subfigure}
    \caption{\textbf{Connect the path puzzle}. Prompt: \textit{``The path connecting the boy to the object starts glowing slowly. Nothing else changes. No zoom, no pan, no dolly."} Failure: hallucinations, lighting up of all paths.}
    \label{fig:connect-path}
\end{figure}

\begin{figure}[H]
    \centering
    \begin{subfigure}[t]{\threefigwidth\textwidth}
        \centering
        \fbox{\includegraphics[width=\linewidth]{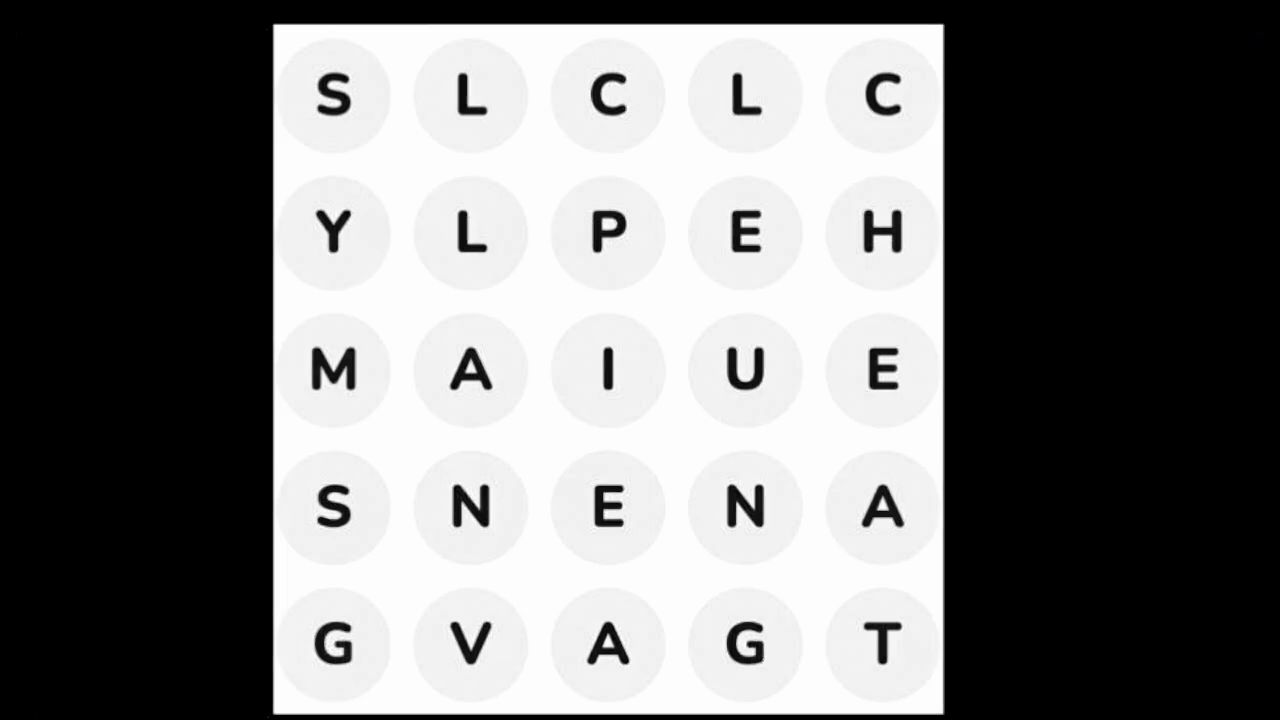}}
    \end{subfigure}
    \hfill
    \begin{subfigure}[t]{\threefigwidth\textwidth}
        \centering
        \fbox{\includegraphics[width=\linewidth]{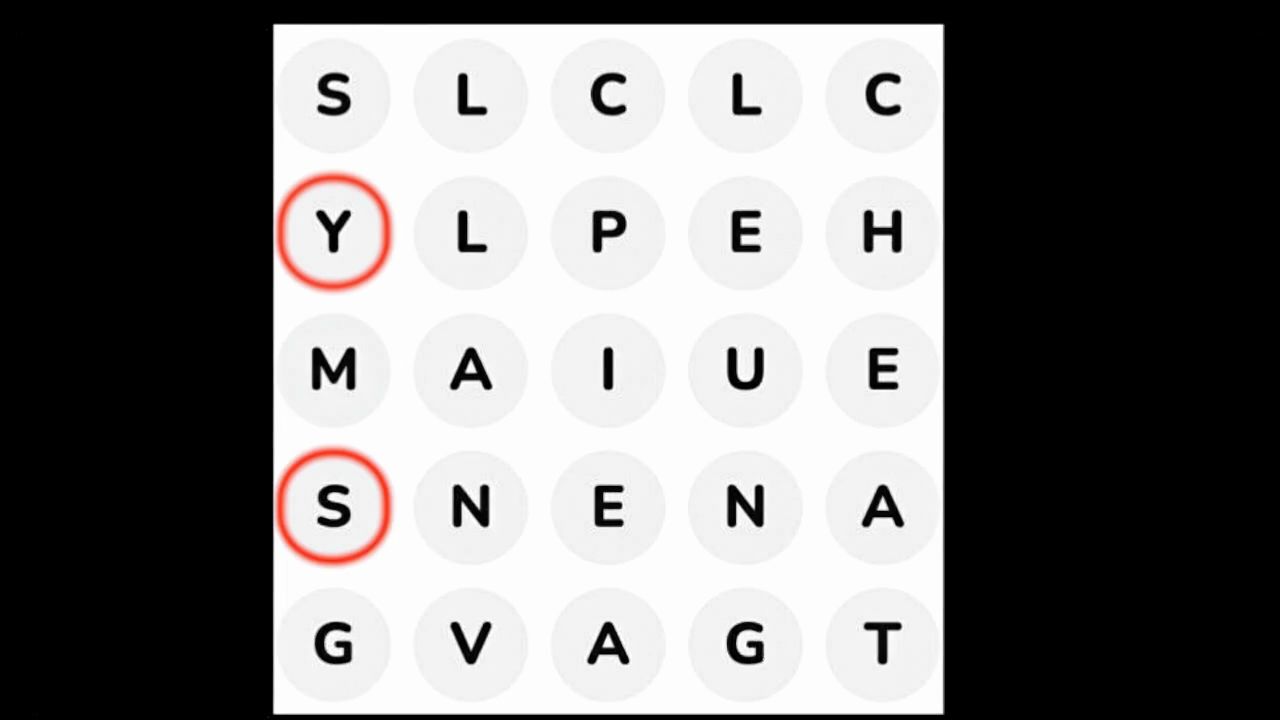}}
    \end{subfigure}
    \hfill    
    \begin{subfigure}[t]{\threefigwidth\textwidth}
        \centering
        \fbox{\includegraphics[width=\linewidth]{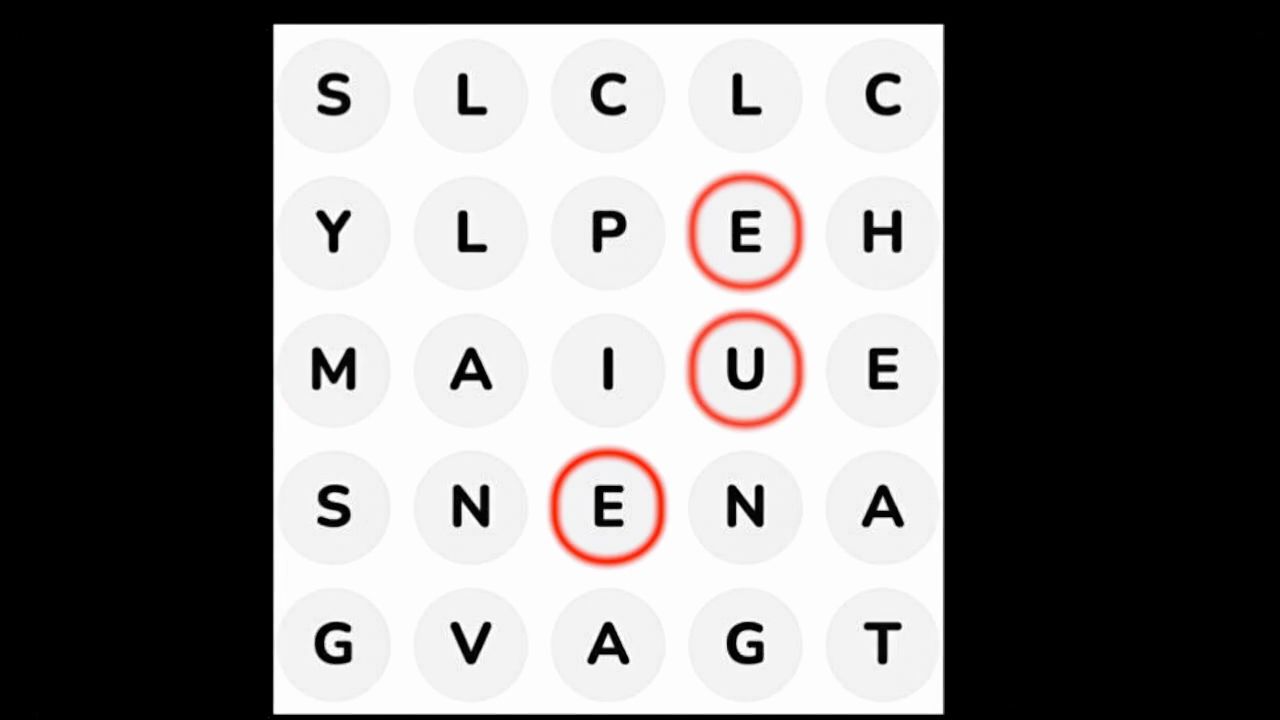}}
    \end{subfigure}
    \caption{\textbf{Five letter word search}. Prompt: \textit{``Generate a static video animation using the provided letter grid. The task is to highlight the only 5-letter English word CHEAT, which may be oriented in any direction (horizontally, vertically, or diagonally). The animation should consist of a semi-transparent red rectangle with rounded corners smoothly fading into view, perfectly encapsulating the five letters of the word. The rectangle should have a subtle, soft glow. Do not change anything else in the image. The camera must remain locked in place with no movement. No zoom, no pan, no dolly."} Failure: does not recognize words; highlights individual letters randomly.}
    \label{fig:crossword}
\end{figure}

\begin{figure}[H]
    \centering
    \begin{subfigure}[t]{\threefigwidth\textwidth}
        \centering
        \fbox{\includegraphics[width=\linewidth]{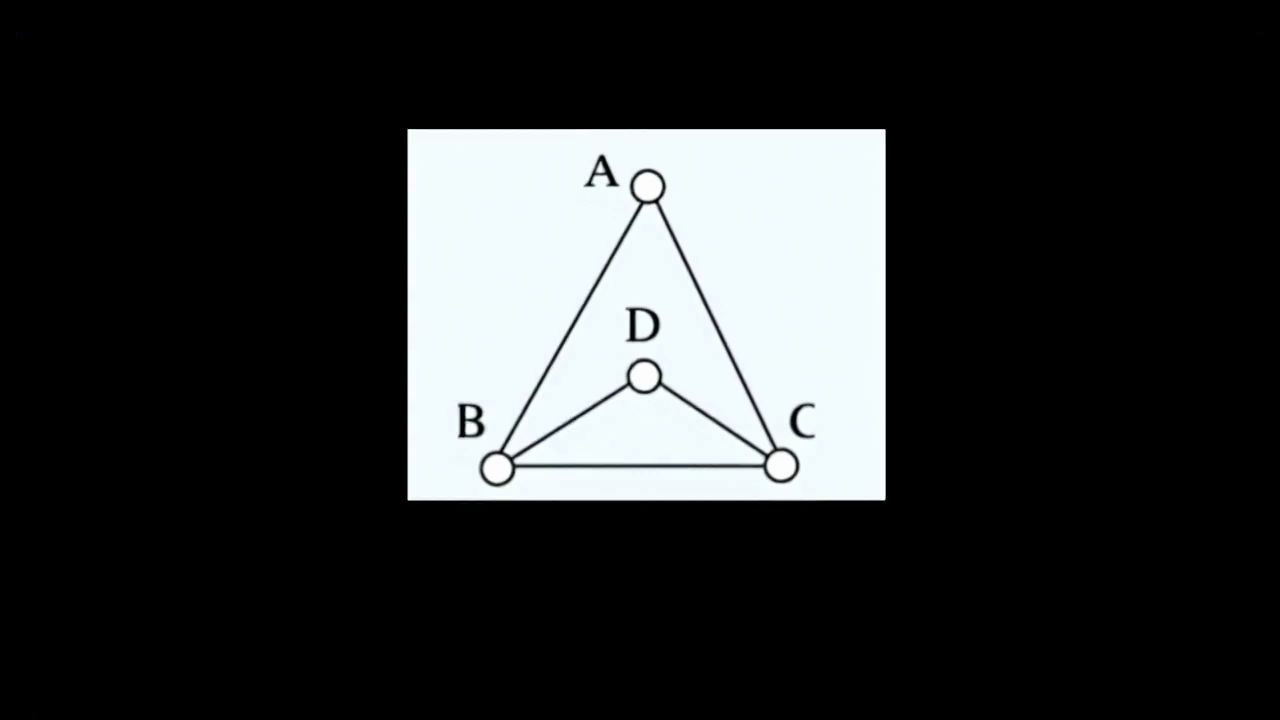}}
    \end{subfigure}
    \hfill
    \begin{subfigure}[t]{\threefigwidth\textwidth}
        \centering
        \fbox{\includegraphics[width=\linewidth]{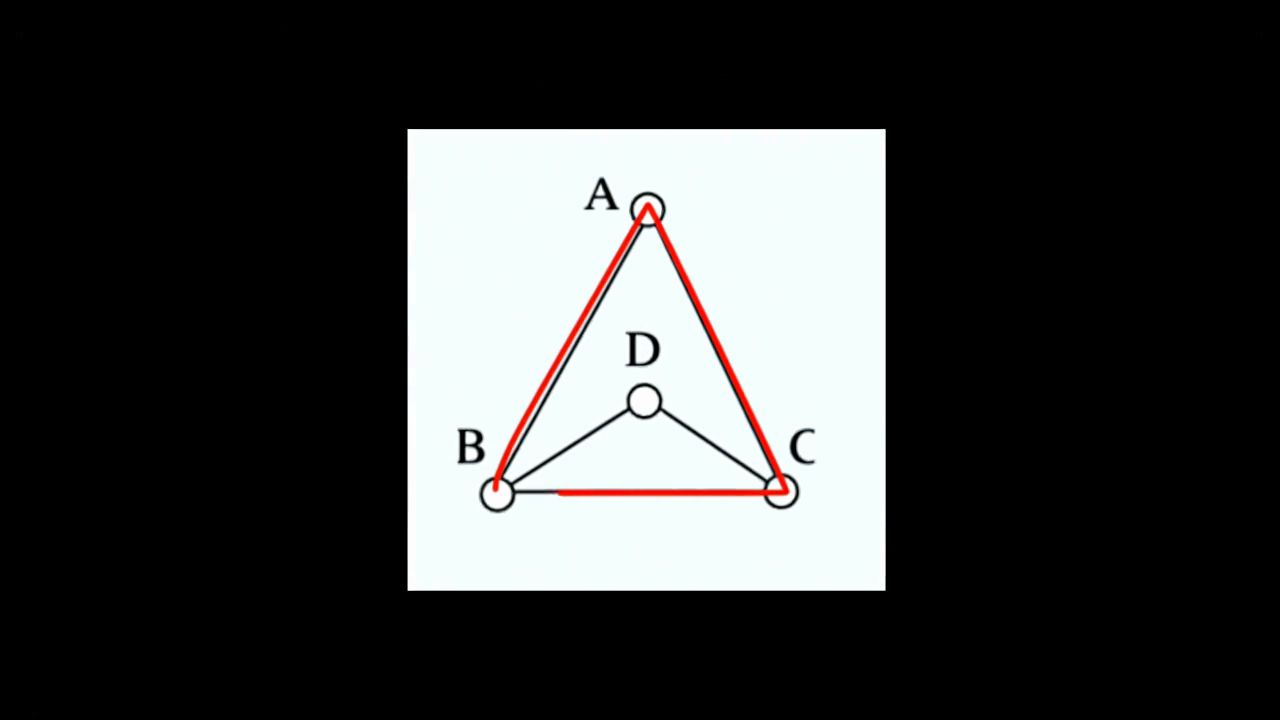}}
    \end{subfigure}
    \hfill    
    \begin{subfigure}[t]{\threefigwidth\textwidth}
        \centering
        \fbox{\includegraphics[width=\linewidth]{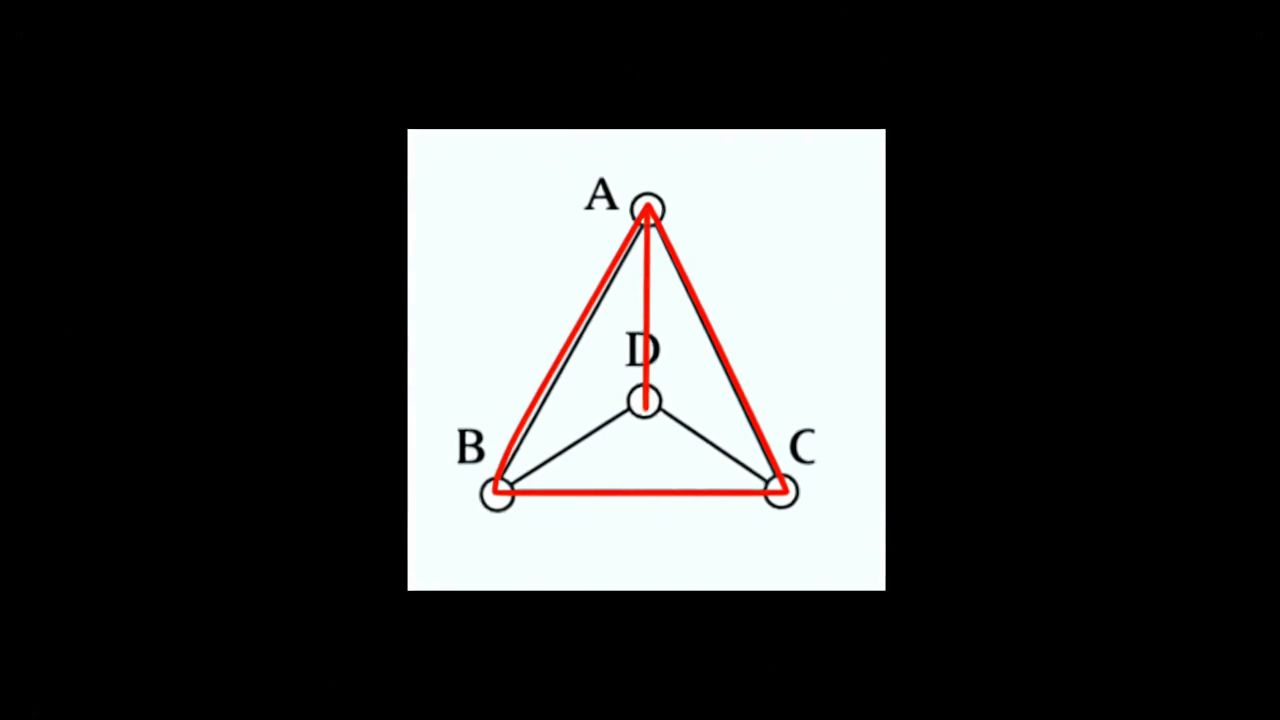}}
    \end{subfigure}
    \caption{\textbf{Eulerian path}. Prompt: \textit{``Create a smooth animation where a red pen traces all existing edges in a continuous path without lifting the pen. All edges need to be traced. Do not visit any edge twice and do not lift the pen. No zoom, no pan."} Failure: does not trace the edges exactly, traces non-existent edges.}
    \label{fig:eulerian-path}
\end{figure}

\begin{figure}[H]
    \centering
    \begin{subfigure}[t]{\threefigwidth\textwidth}
        \centering
        \fbox{\includegraphics[width=\linewidth]{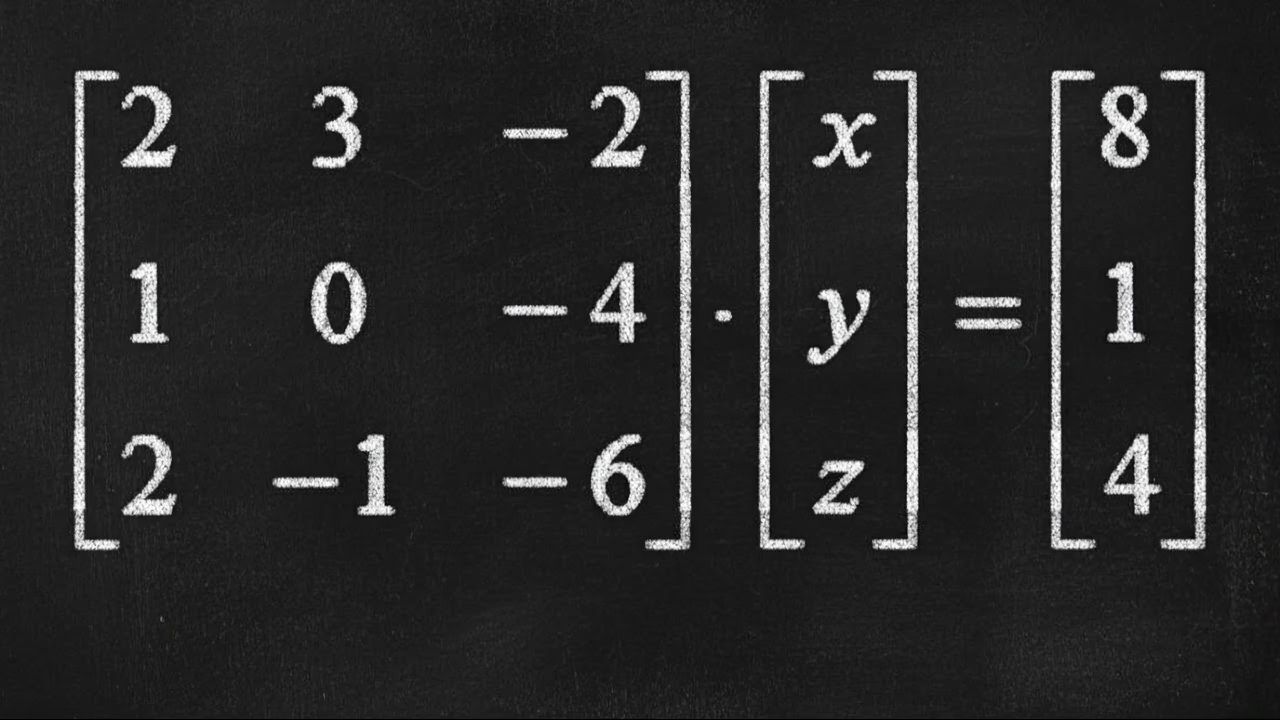}}
    \end{subfigure}
    \hfill
    \begin{subfigure}[t]{\threefigwidth\textwidth}
        \centering
        \fbox{\includegraphics[width=\linewidth]{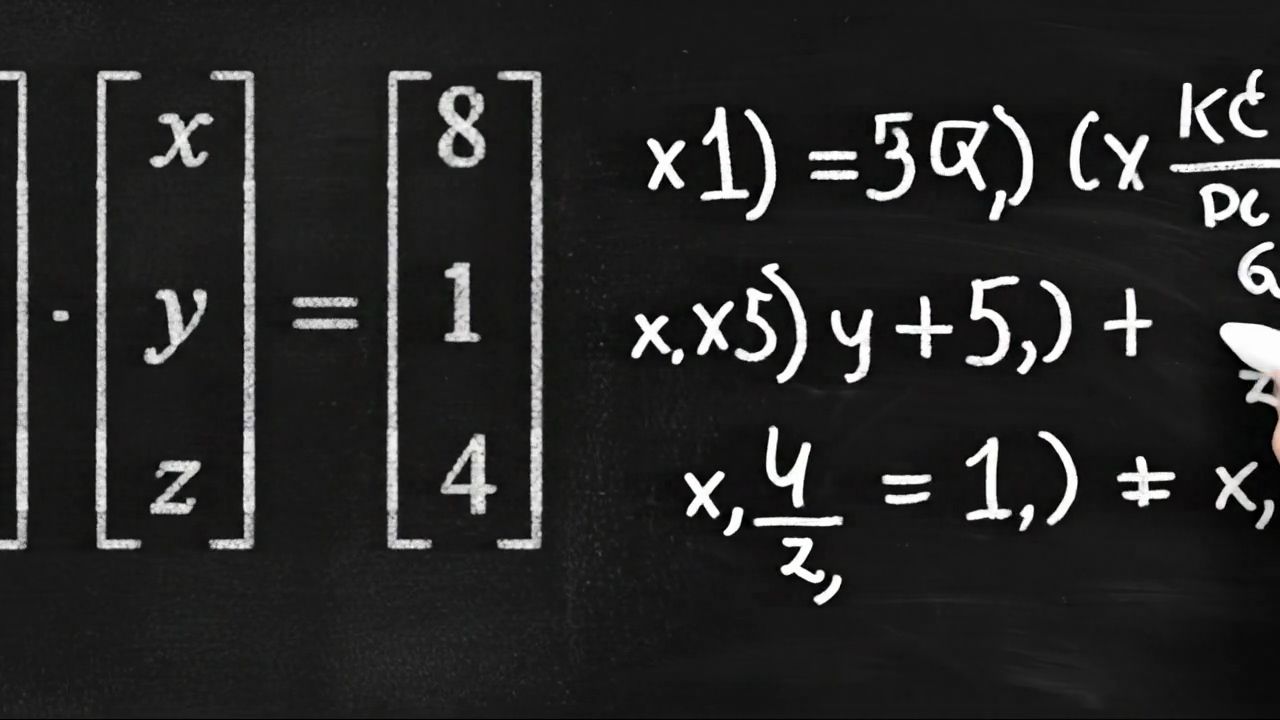}}
    \end{subfigure}
    \hfill    
    \begin{subfigure}[t]{\threefigwidth\textwidth}
        \centering
        \fbox{\includegraphics[width=\linewidth]{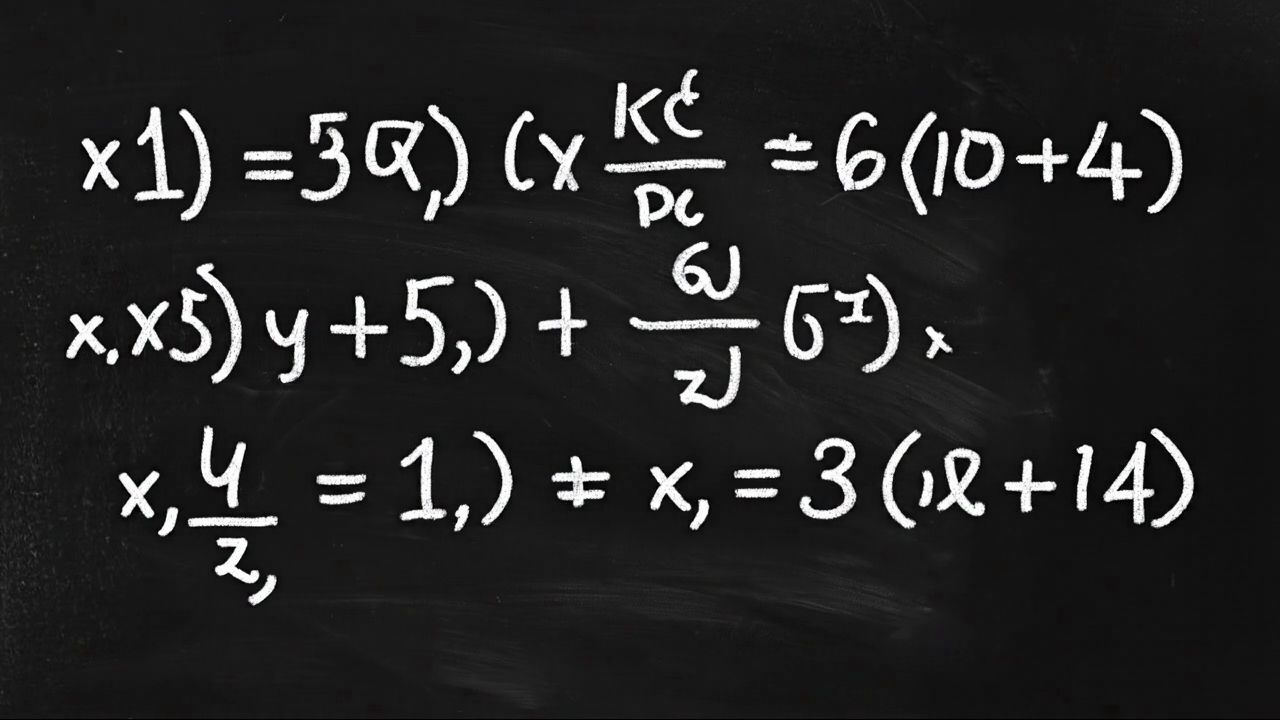}}
    \end{subfigure}
    \caption{\textbf{Solving system of linear equations}. Prompt: \textit{``A hand appears and solves the set of linear equations. It replaces the x, y, z matrix with their correct values that solves the equation. Do not change anything else."} Failure: hallucinations with text on the blackboard.}
    \label{fig:linear-equations}
\end{figure}

\begin{figure}[H]
    \centering
    \begin{subfigure}[t]{\fourfigwidth\textwidth}
        \centering
        \fbox{\includegraphics[width=\linewidth]{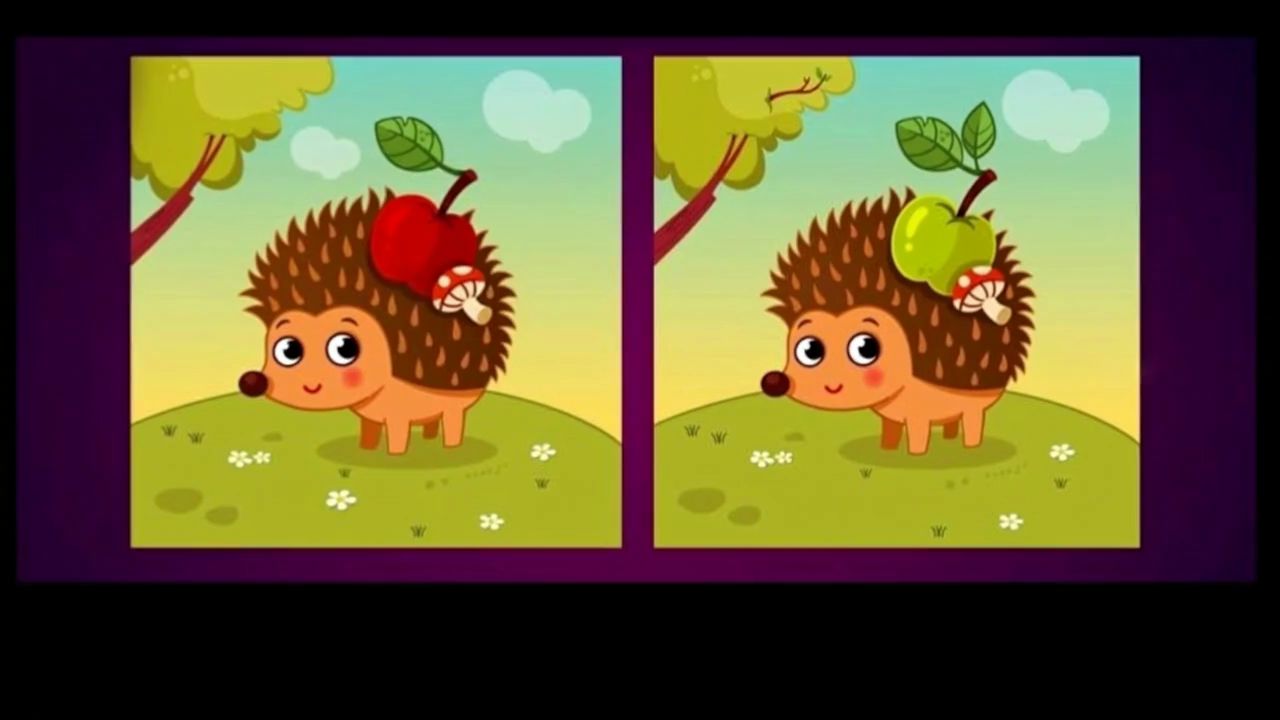}}
    \end{subfigure}
    \hfill
    \begin{subfigure}[t]{\fourfigwidth\textwidth}
        \centering
        \fbox{\includegraphics[width=\linewidth]{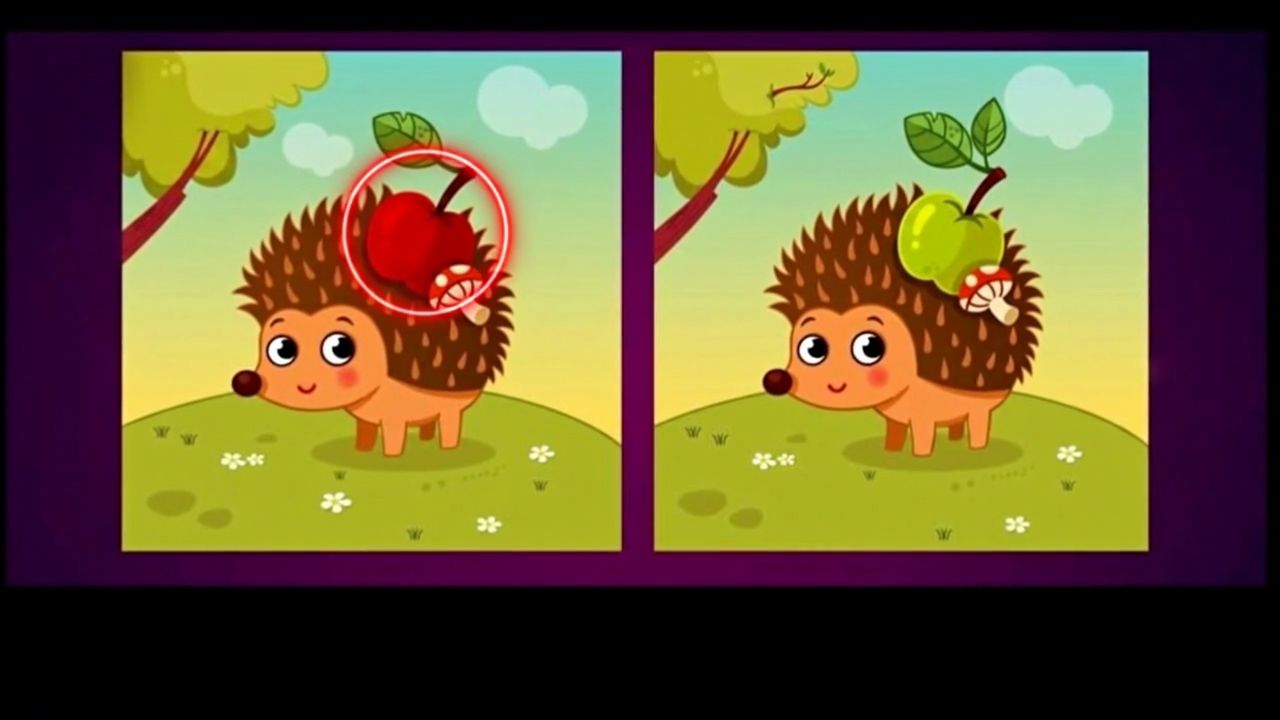}}
    \end{subfigure}
    \hfill    
    \begin{subfigure}[t]{\fourfigwidth\textwidth}
        \centering
        \fbox{\includegraphics[width=\linewidth]{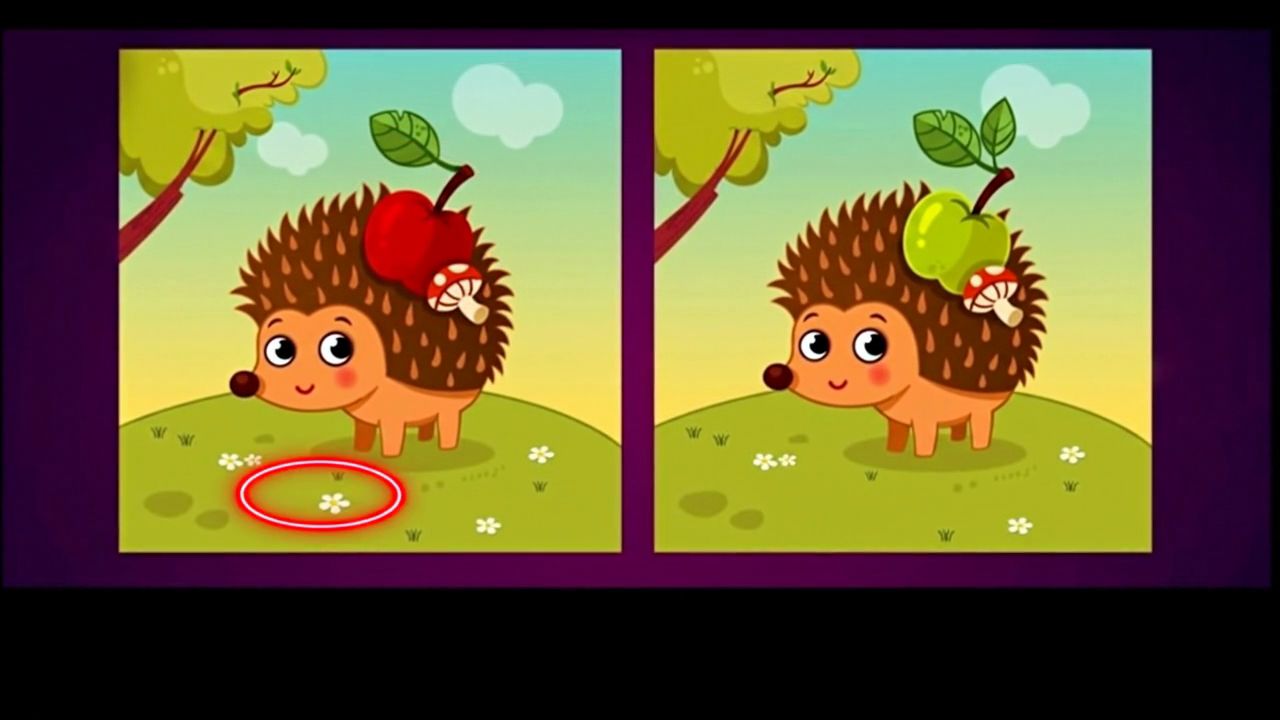}}
    \end{subfigure}
    \hfill    
    \begin{subfigure}[t]{\fourfigwidth\textwidth}
        \centering
        \fbox{\includegraphics[width=\linewidth]{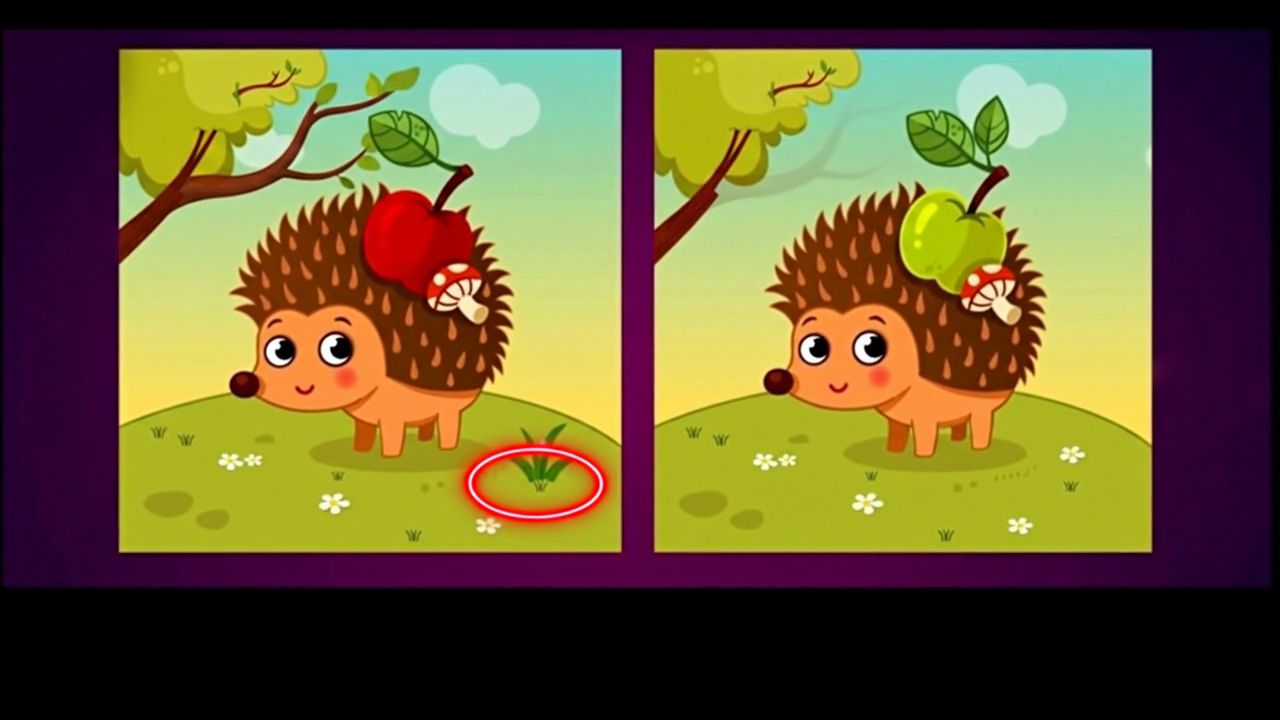}}
    \end{subfigure}
    \caption{\textbf{Spot the difference}. Prompt: \textit{``There are two images. The left image is different from the right image in 5 spots. Create a static, realistic, smooth animation where a cursor appears and points at each place where the left image is different from the right image. The cursor points one by one and only on the left image. Do not change anything in the right image. No pan. No zoom. No movement. Keep the image static."} Failure: does not identify all the differences. Hallucinates differences.}
    \label{fig:spot-difference}
\end{figure}

\begin{figure}[H]
    \centering
    \begin{subfigure}[t]{\threefigwidth\textwidth}
        \centering
        \fbox{\includegraphics[width=\linewidth]{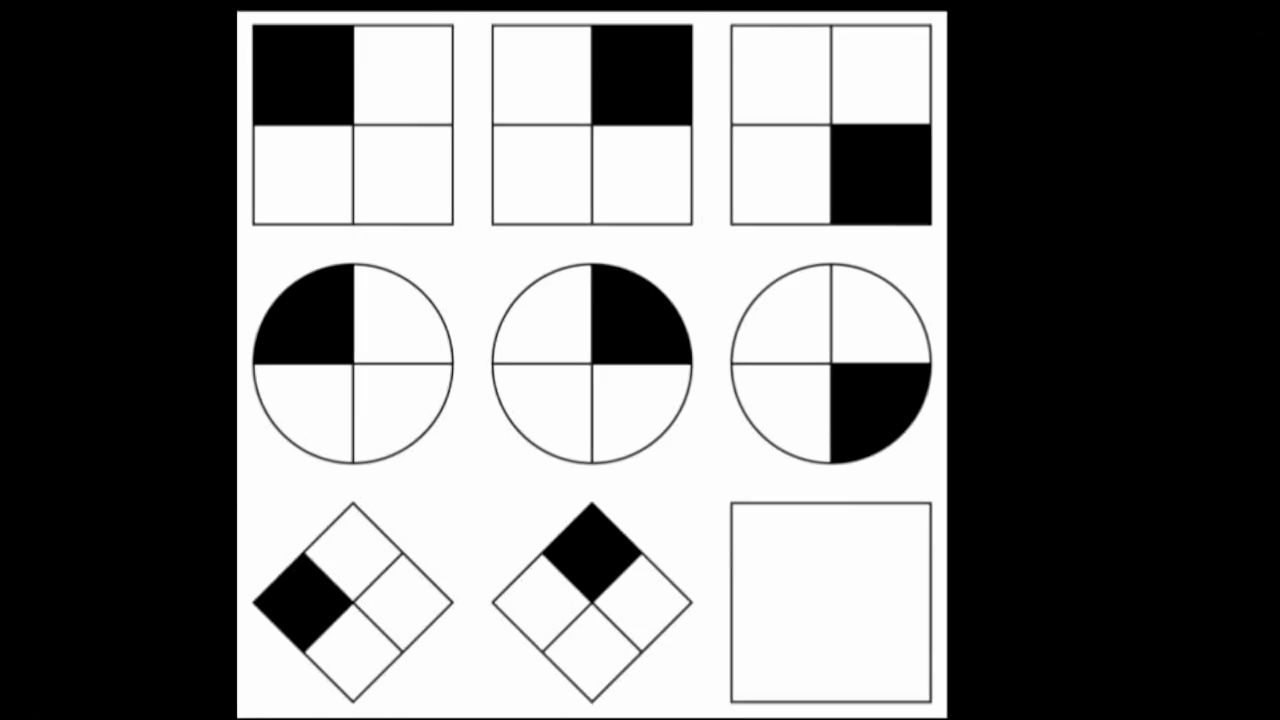}}
    \end{subfigure}
    \hfill
    \begin{subfigure}[t]{\threefigwidth\textwidth}
        \centering
        \fbox{\includegraphics[width=\linewidth]{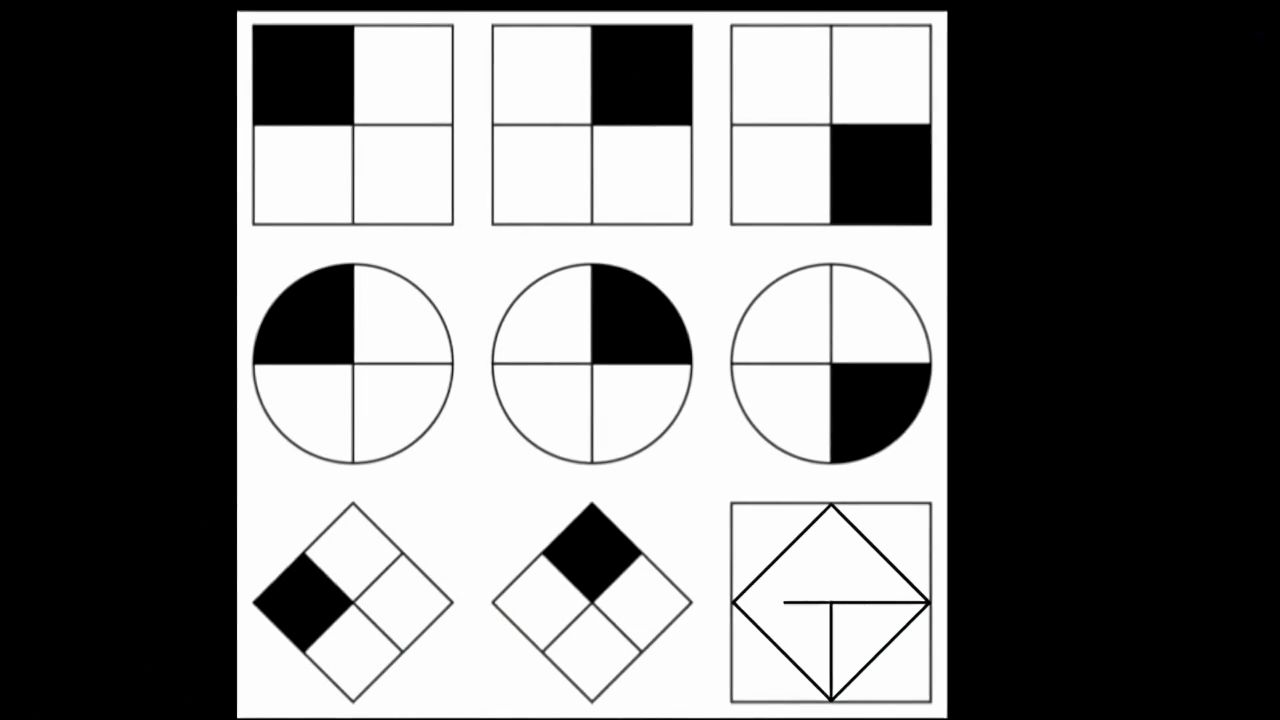}}
    \end{subfigure}
    \hfill    
    \begin{subfigure}[t]{\threefigwidth\textwidth}
        \centering
        \fbox{\includegraphics[width=\linewidth]{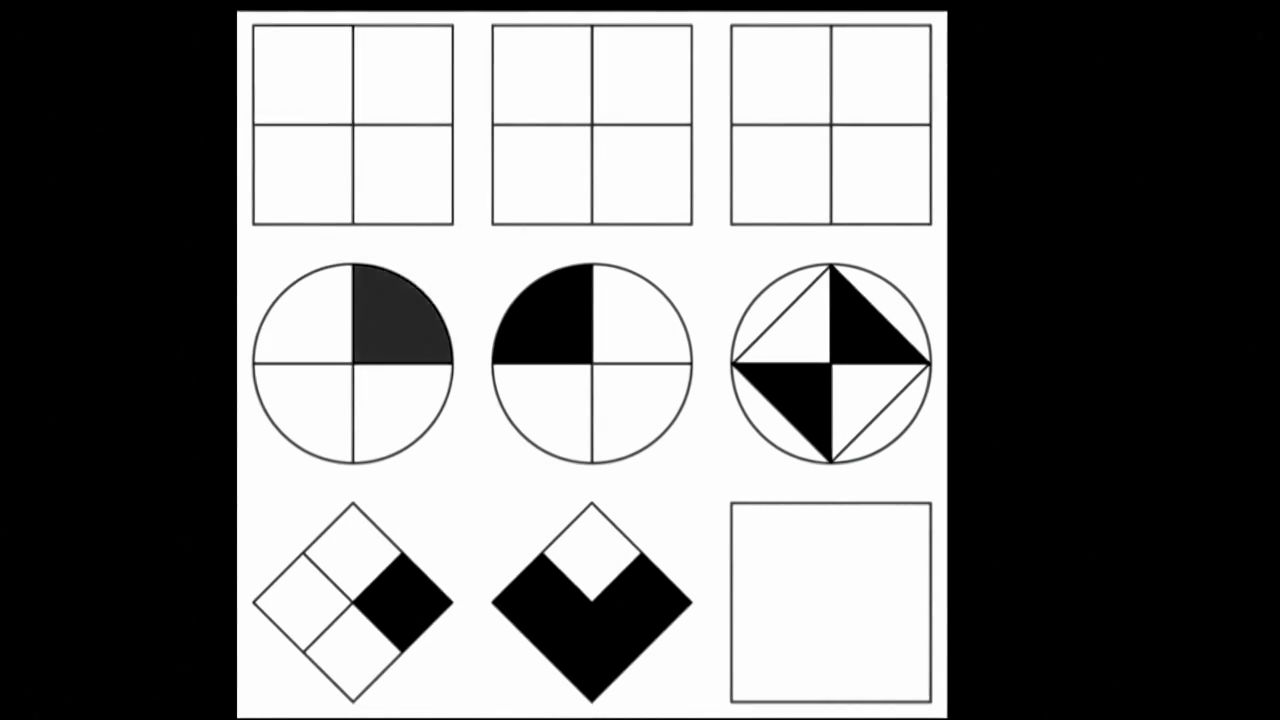}}
    \end{subfigure}
    \caption{\textbf{Visual IQ test}. Prompt: \textit{``Create a static, smooth, animation that solves the puzzle in the given image. The correct pattern should appear at the bottom right to solve the puzzle. Do not change anything else in the picture. No zoom, no pan, no dolly"} Failure: incorrect figure pattern.}
    \label{fig:visual_iq}
\end{figure}

\begin{figure}[H]
    \centering
    \begin{subfigure}[t]{\fourfigwidth\textwidth}
        \centering
        \fbox{\includegraphics[width=\linewidth]{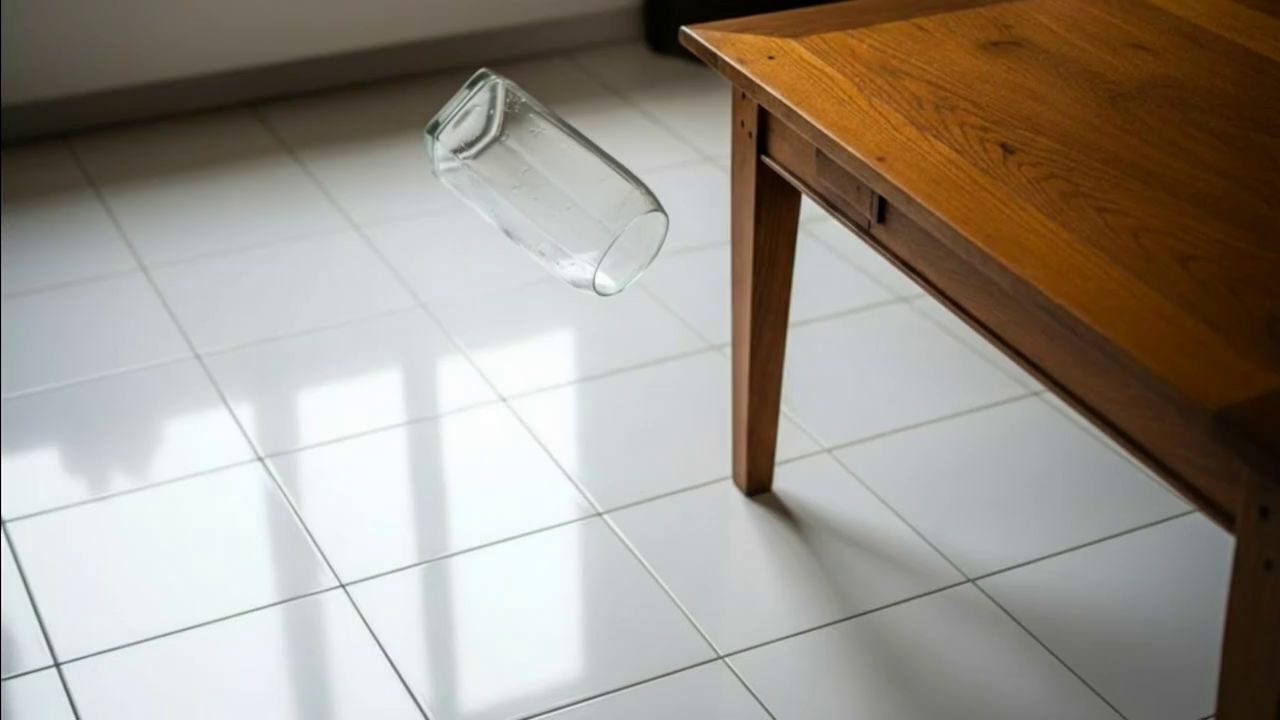}}
    \end{subfigure}
    \hfill
    \begin{subfigure}[t]{\fourfigwidth\textwidth}
        \centering
        \fbox{\includegraphics[width=\linewidth]{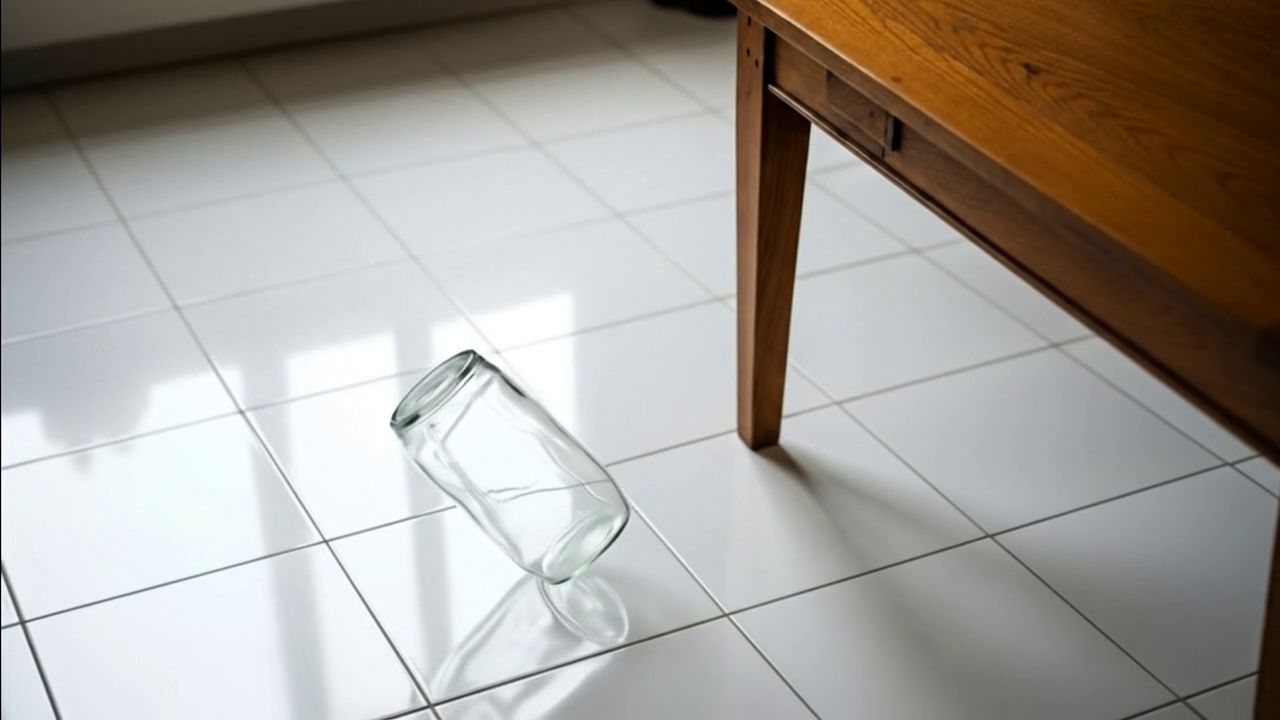}}
    \end{subfigure}
    \hfill
    \begin{subfigure}[t]{\fourfigwidth\textwidth}
        \centering
        \fbox{\includegraphics[width=\linewidth]{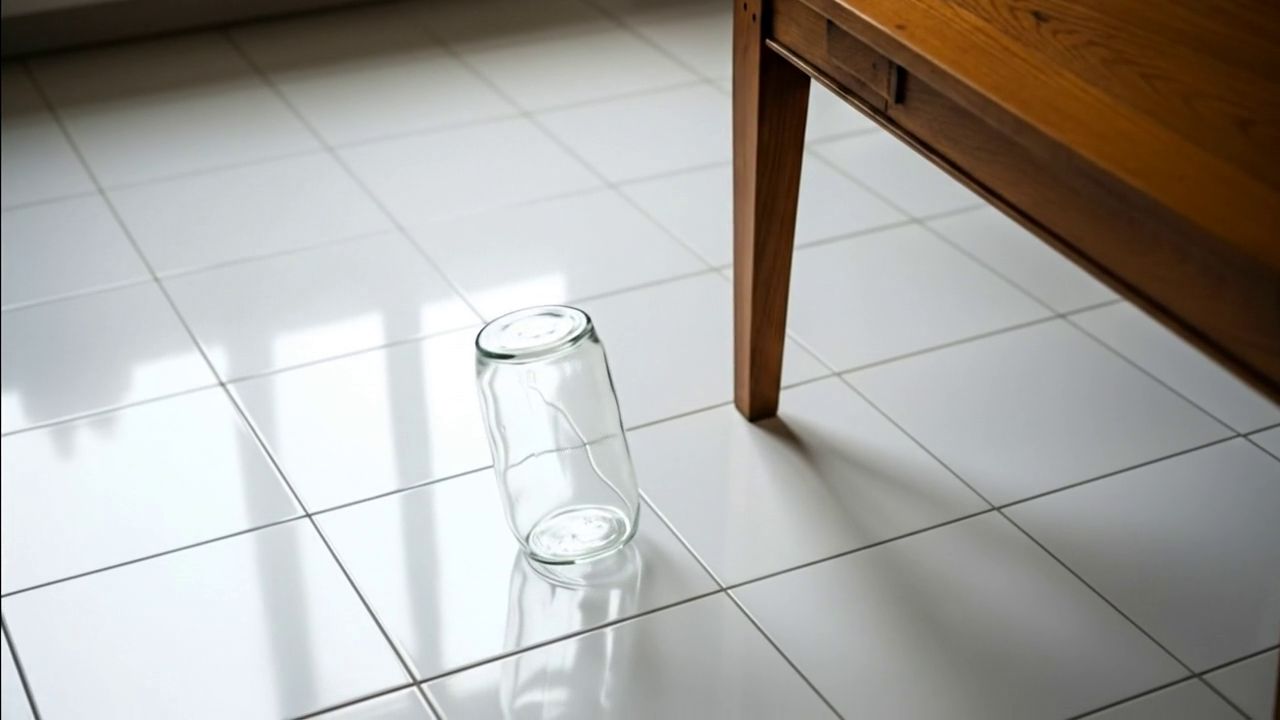}}
    \end{subfigure}
    \hfill
    \begin{subfigure}[t]{\fourfigwidth\textwidth}
        \centering
        \fbox{\includegraphics[width=\linewidth]{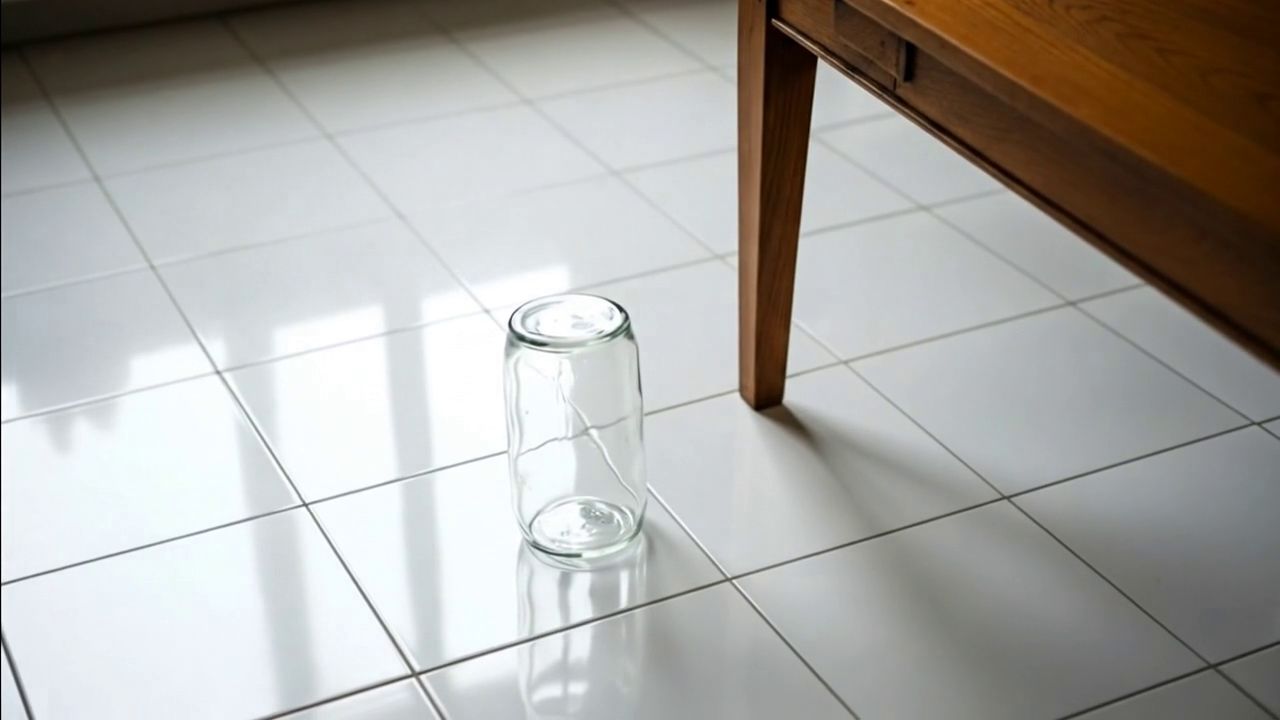}}
    \end{subfigure}
    \caption{\textbf{Glass falling}. Prompt: \textit{``The object falls. Static camera, no pan, no zoom, no dolly.''} Failure: physics violation, glass does not break, and orients itself to be vertical after landing on the floor.}
    \label{fig:glass_falling}
\end{figure}

\begin{figure}[H]
    \centering
    \begin{subfigure}[t]{\fourfigwidth\textwidth}
        \centering
        \fbox{\includegraphics[width=\linewidth]{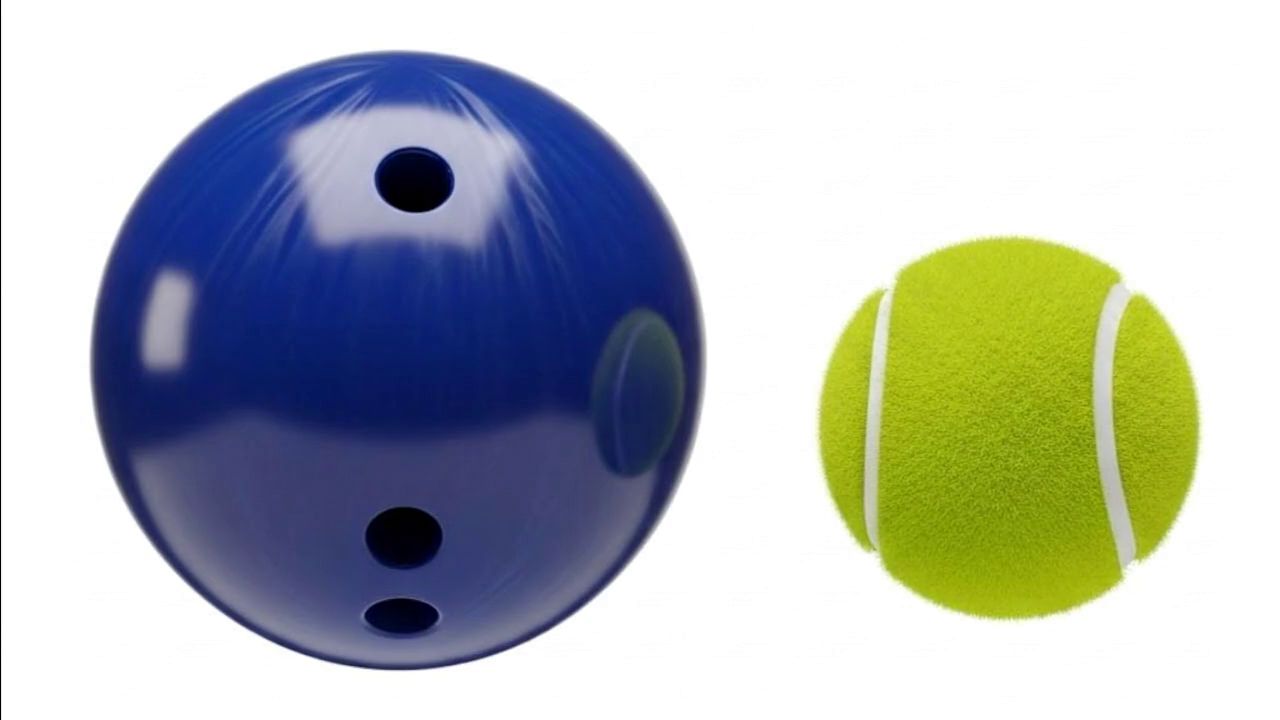}}
    \end{subfigure}
    \hfill
    \begin{subfigure}[t]{\fourfigwidth\textwidth}
        \centering
        \fbox{\includegraphics[width=\linewidth]{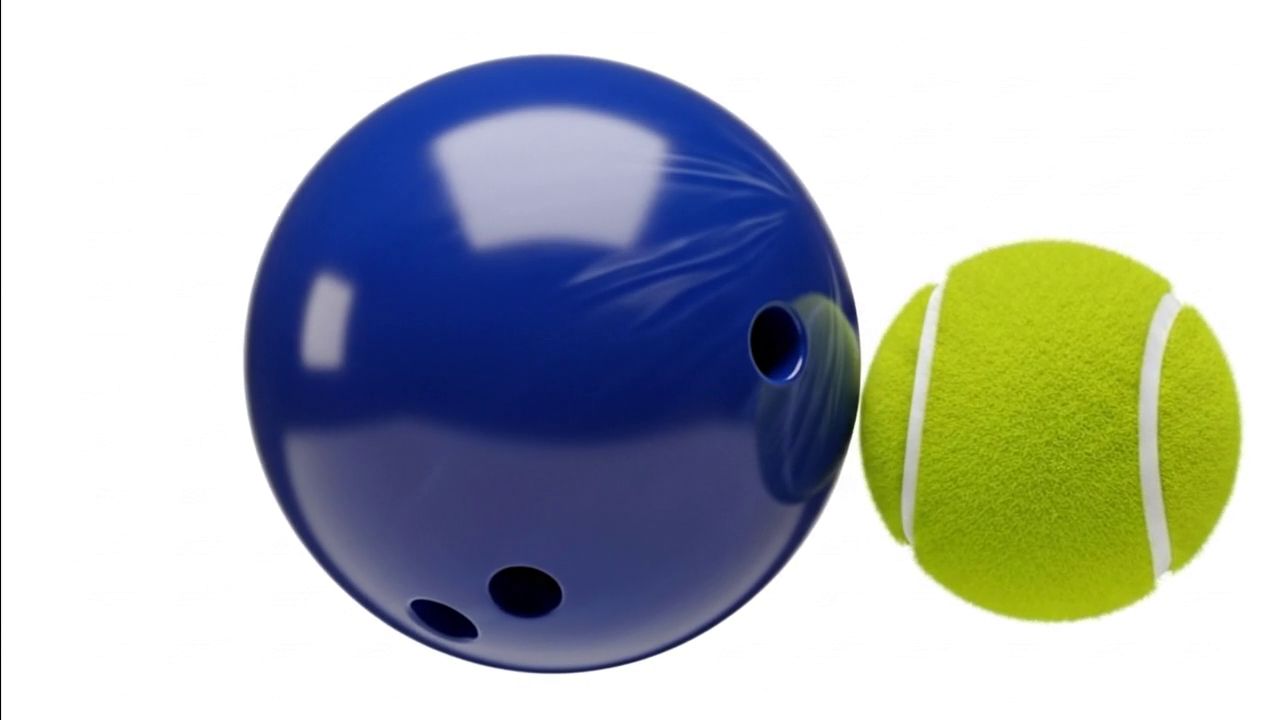}}
    \end{subfigure}
    \hfill
    \begin{subfigure}[t]{\fourfigwidth\textwidth}
        \centering
        \fbox{\includegraphics[width=\linewidth]{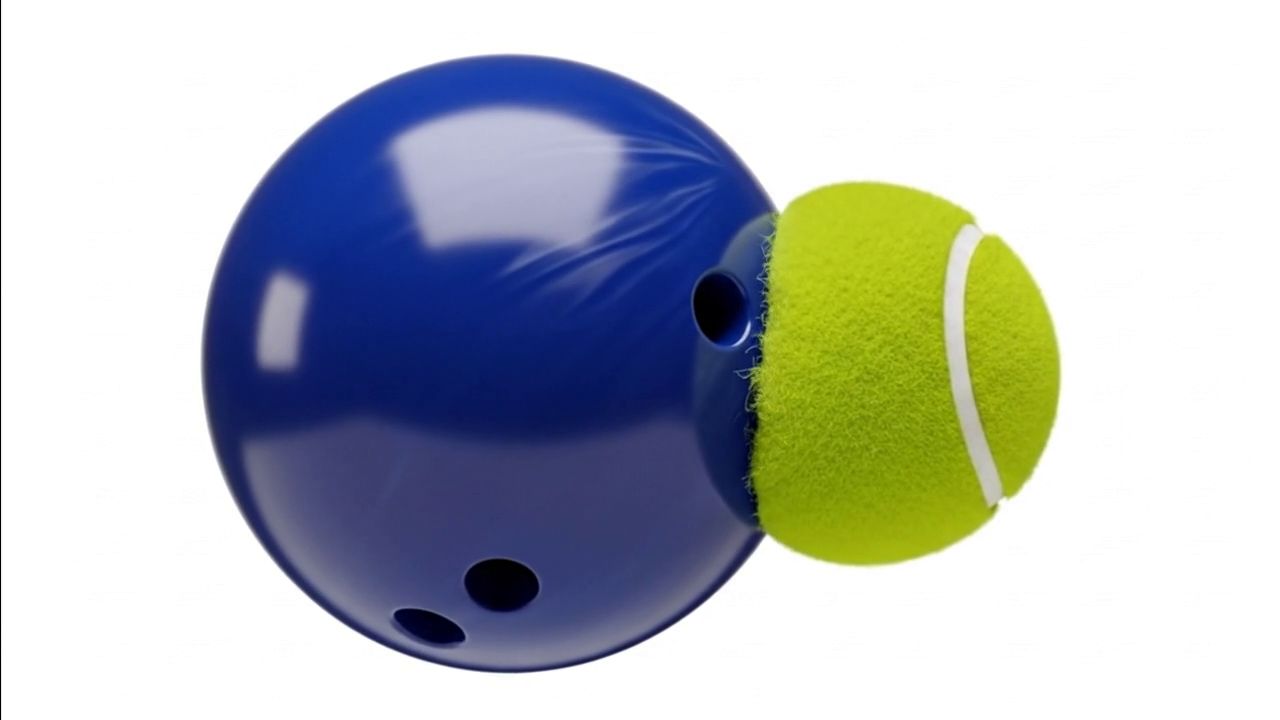}}
    \end{subfigure}
    \hfill
    \begin{subfigure}[t]{\fourfigwidth\textwidth}
        \centering
        \fbox{\includegraphics[width=\linewidth]{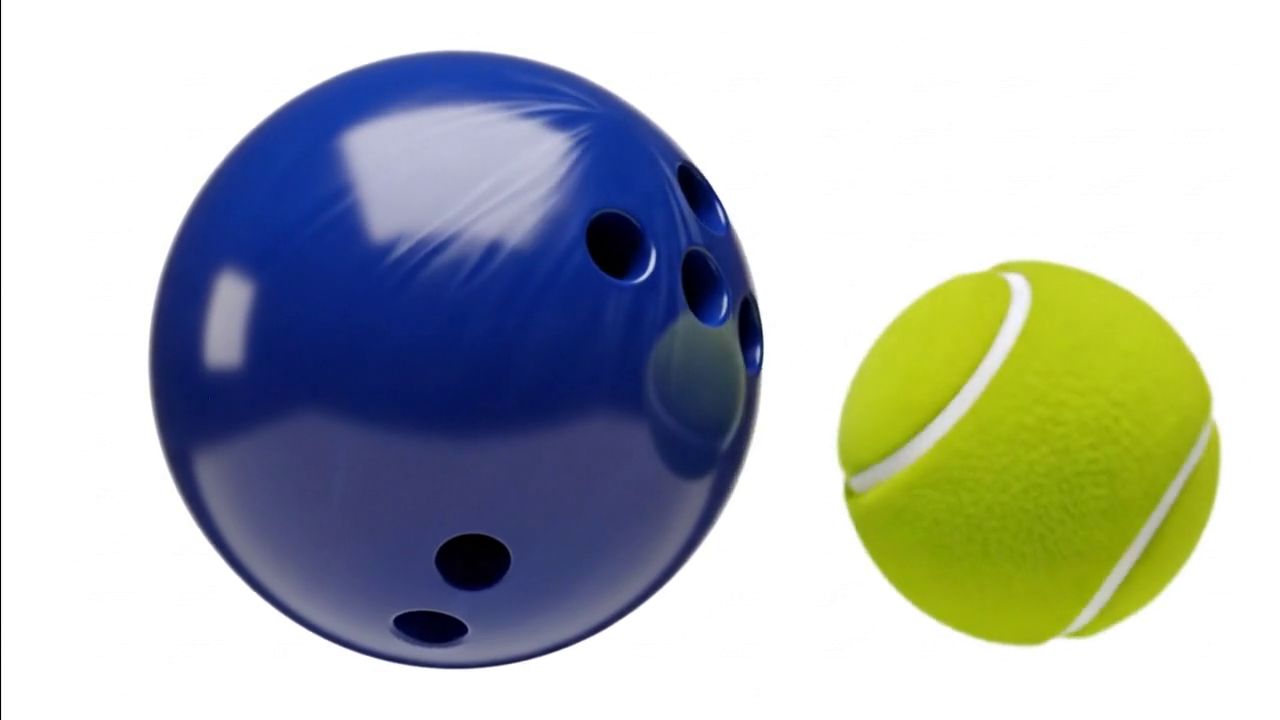}}
    \end{subfigure}
    \caption{\textbf{Collisions}. Prompt: \textit{``The two objects collide in slow motion. Static camera, no pan, no zoom, no dolly.''} Failure: not physically plausible, the objects pause at the moment of impact and then are pushed together by an invisible force.}
    \label{fig:collisions}
\end{figure}

\begin{figure}[H]
    \centering
    \begin{subfigure}[t]{\threefigwidth\textwidth}
        \centering
        \fbox{\includegraphics[width=\linewidth]{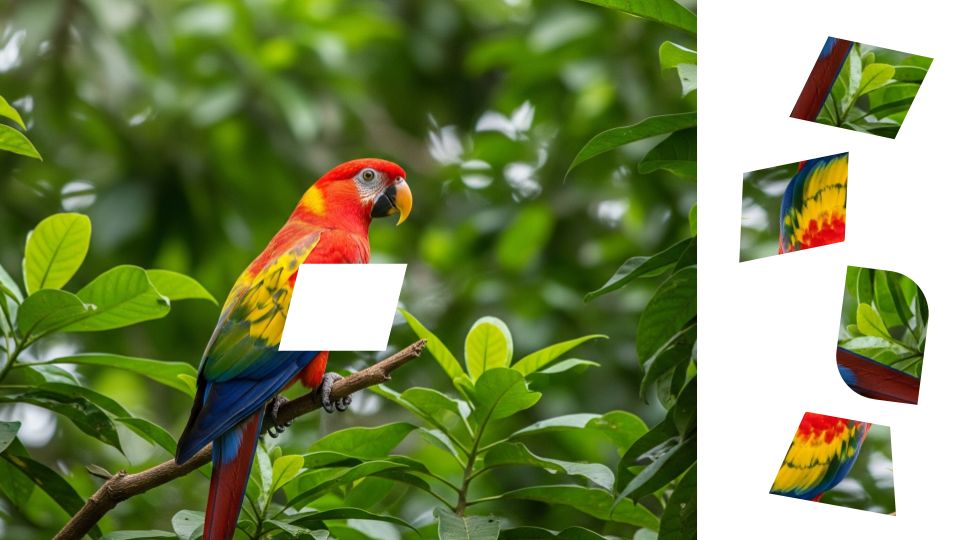}}
    \end{subfigure}
    \hfill
    \begin{subfigure}[t]{\threefigwidth\textwidth}
        \centering
        \fbox{\includegraphics[width=\linewidth]{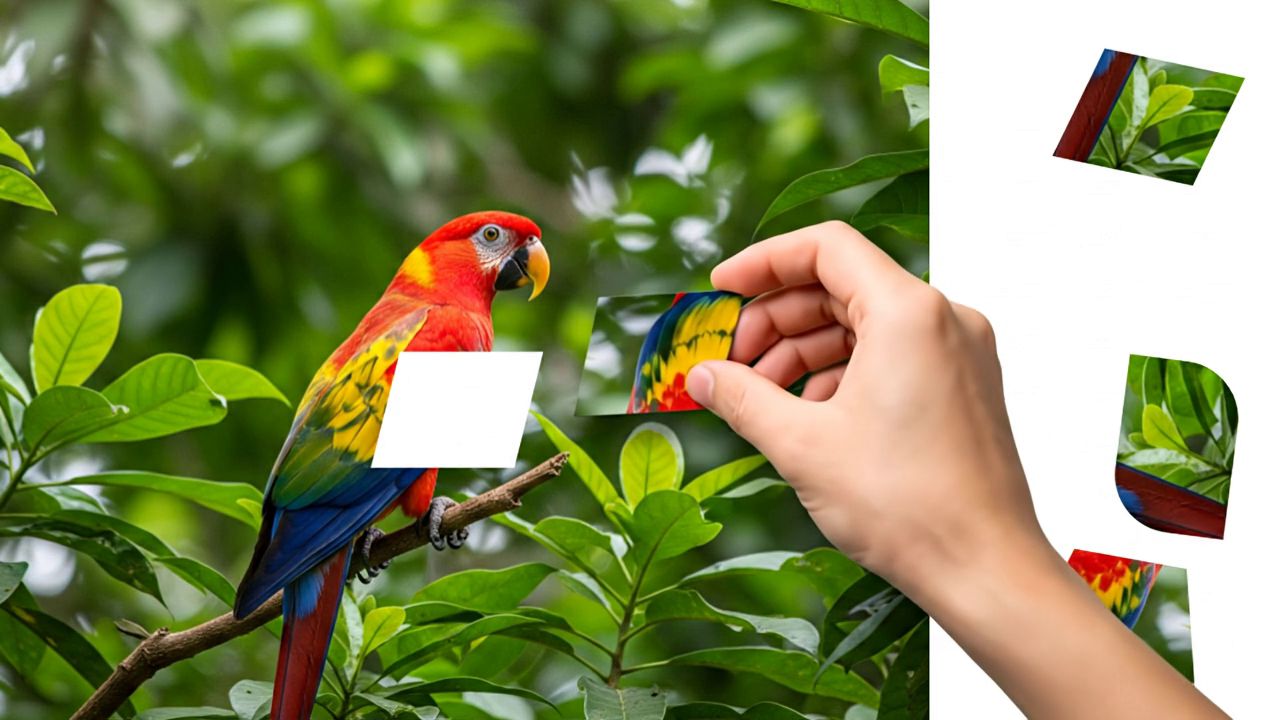}}
    \end{subfigure}
    \hfill    
    \begin{subfigure}[t]{\threefigwidth\textwidth}
        \centering
        \fbox{\includegraphics[width=\linewidth]{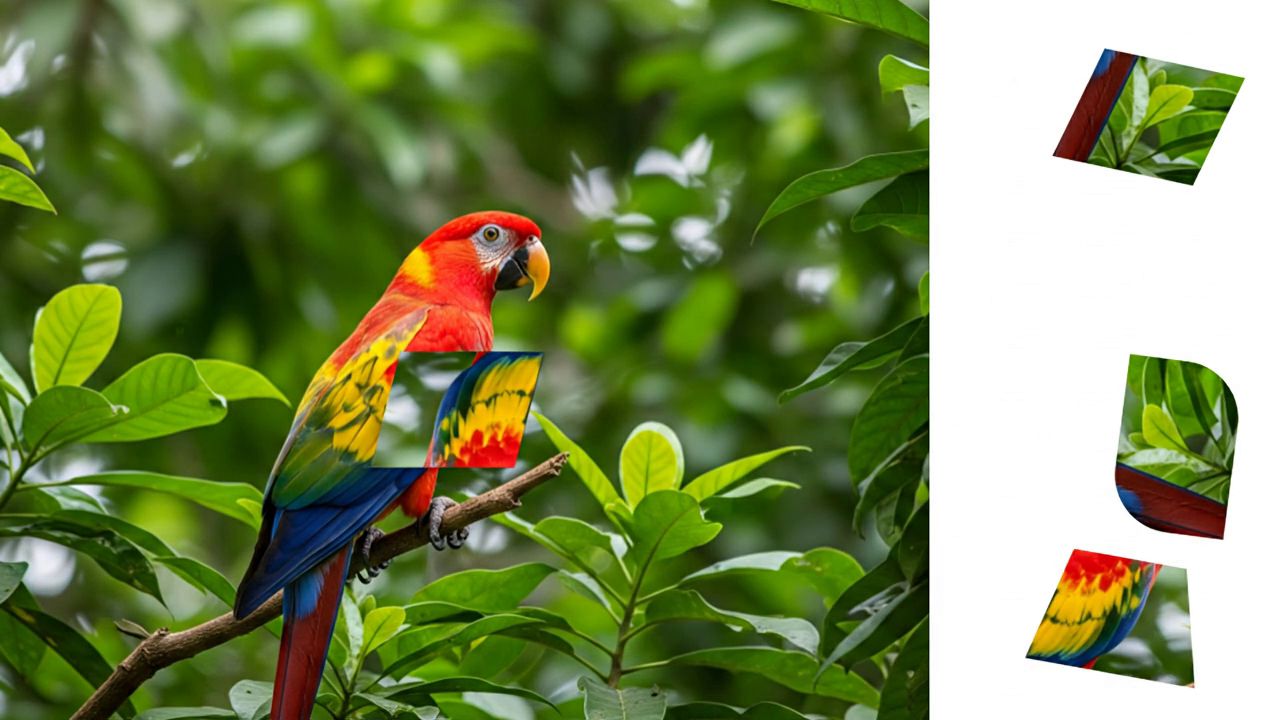}}
    \end{subfigure}
    
    \vspace{0.2cm} %

    \begin{subfigure}[t]{\threefigwidth\textwidth}
        \centering
        \fbox{\includegraphics[width=\linewidth]{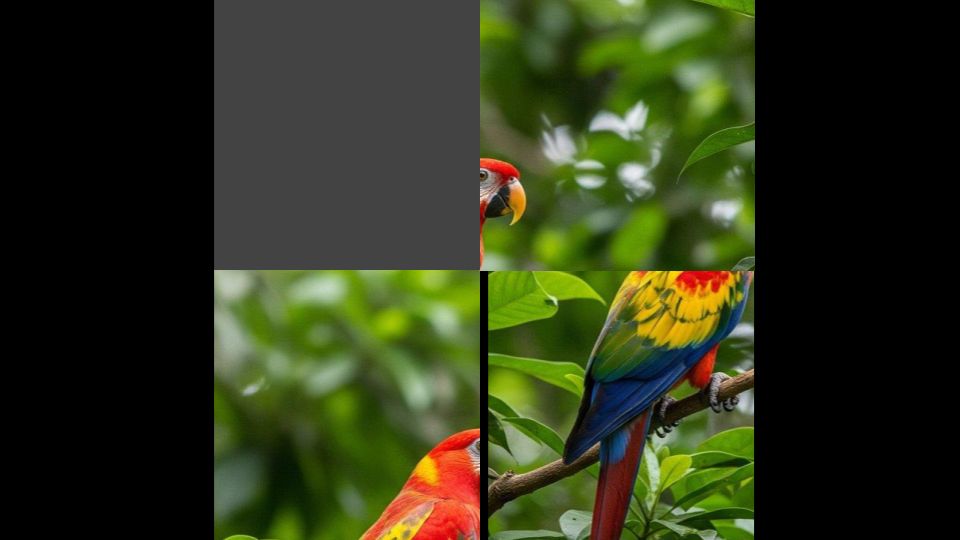}}
    \end{subfigure}
    \hfill
    \begin{subfigure}[t]{\threefigwidth\textwidth}
        \centering
        \fbox{\includegraphics[width=\linewidth]{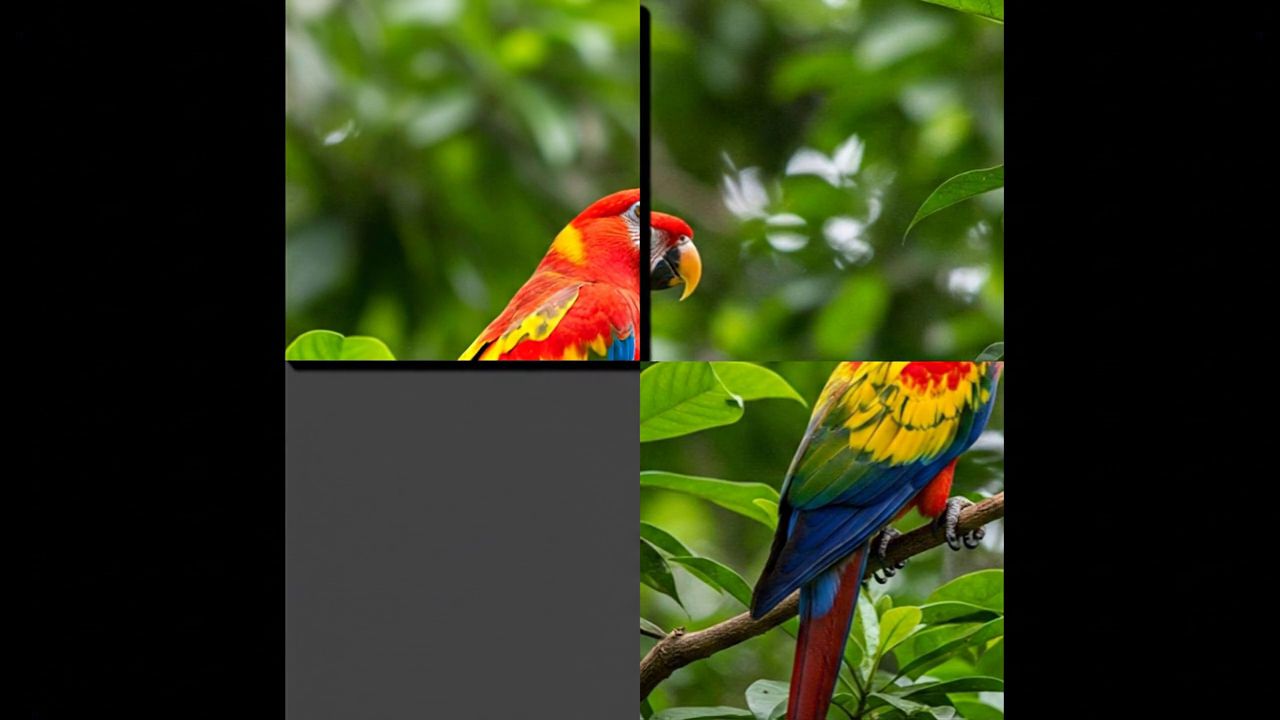}}
    \end{subfigure}
    \hfill    
    \begin{subfigure}[t]{\threefigwidth\textwidth}
        \centering
        \fbox{\includegraphics[width=\linewidth]{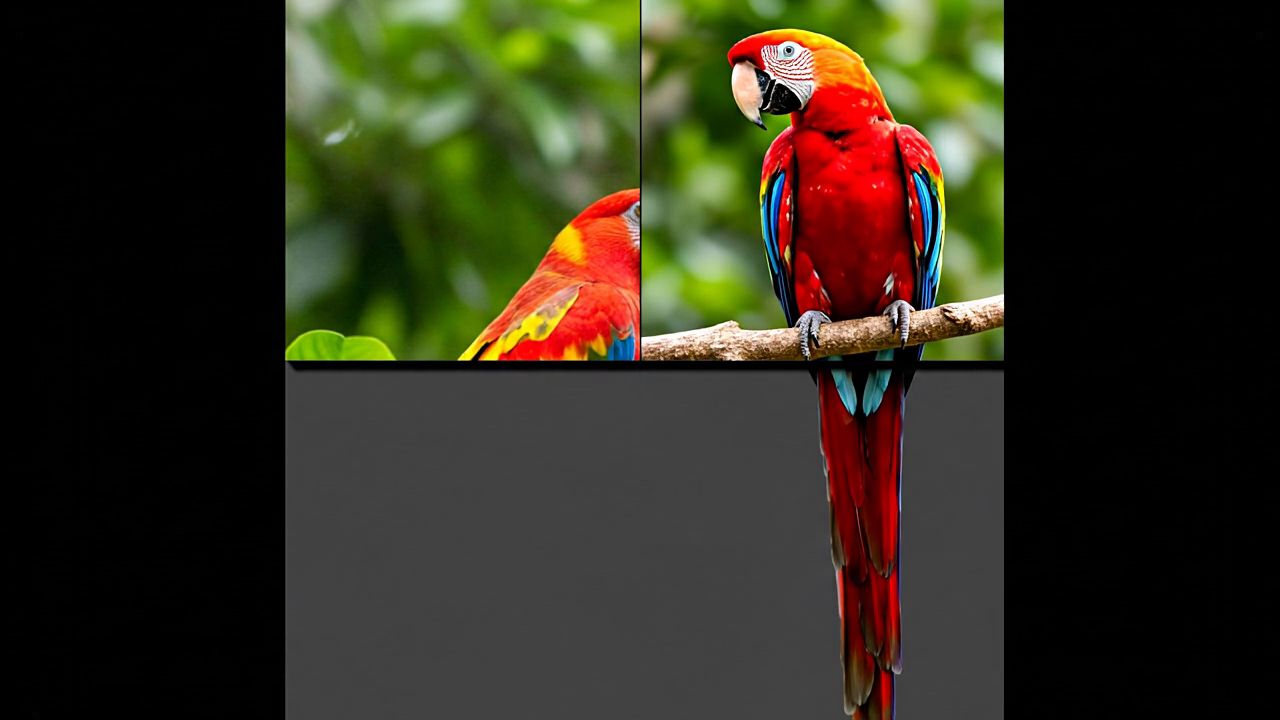}}
    \end{subfigure}

    \vspace{0.2cm} %
    
    \begin{subfigure}[t]{\threefigwidth\textwidth}
        \centering
        \fbox{\includegraphics[width=\linewidth]{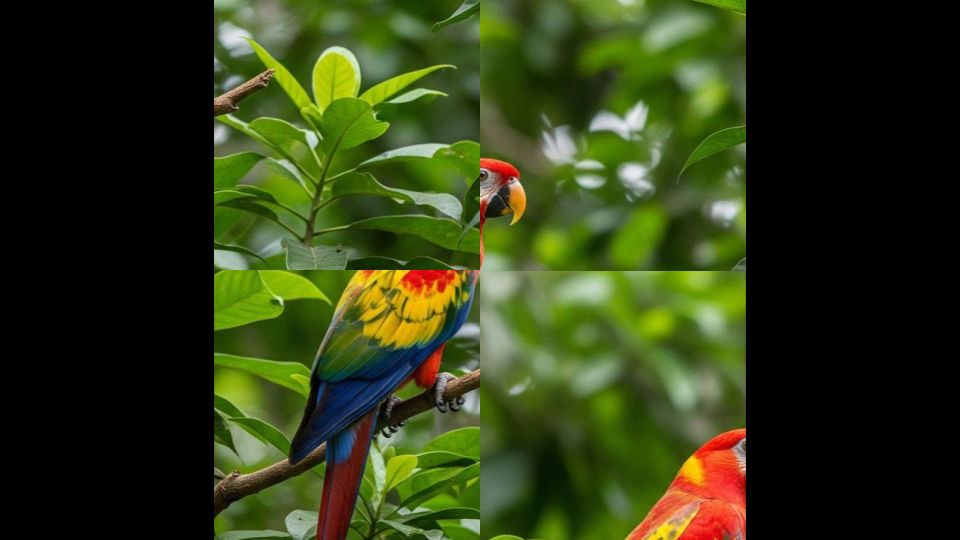}}
    \end{subfigure}
    \hfill
    \begin{subfigure}[t]{\threefigwidth\textwidth}
        \centering
        \fbox{\includegraphics[width=\linewidth]{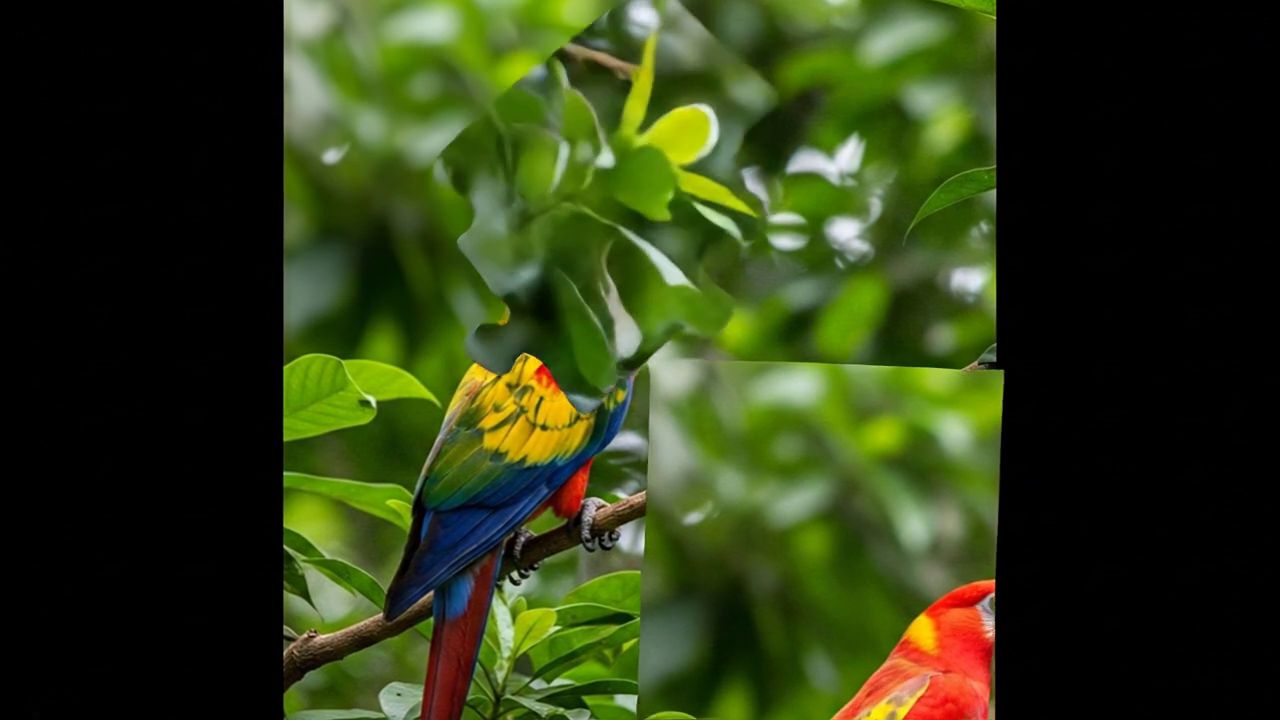}}
    \end{subfigure}
    \hfill    
    \begin{subfigure}[t]{\threefigwidth\textwidth}
        \centering
        \fbox{\includegraphics[width=\linewidth]{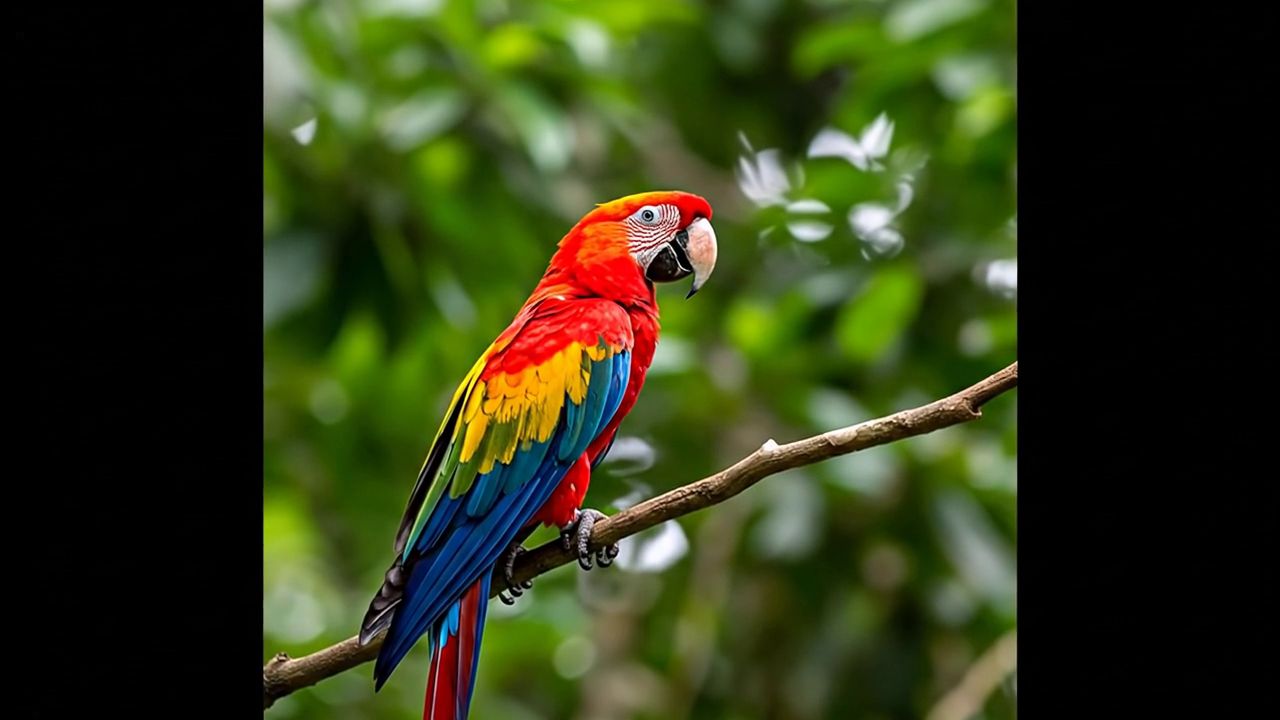}}
    \end{subfigure}
    \caption{\textbf{Tiling puzzles}.
    \textbf{Jigsaw puzzle} (top). Prompt: \textit{``A hand takes the fitting puzzle piece from the right, rotates it to be in the correct orientation, then puts it into the hole, completing the puzzle. Static scene, no pan, no zoom, no dolly.''} Failure: wrong piece orientation.
    \textbf{Sliding puzzle} (middle). Prompt: \textit{``Slide the pieces of this sliding puzzle around one-at-a-time until all edges align.''} Failure: doesn't maintain piece integrity while sliding, hallucinates new pieces.
    \textbf{Scrambled puzzle} (bottom). Prompt: \textit{``Unscramble this image.''} Failure: image details are inconsistent with original pieces.
    }
    \label{fig:tiling-puzzles}
\end{figure}

\begin{figure}[H]
    \centering
    \begin{subfigure}[t]{\fourfigwidth\textwidth}
        \centering
        \fbox{\includegraphics[width=\linewidth]{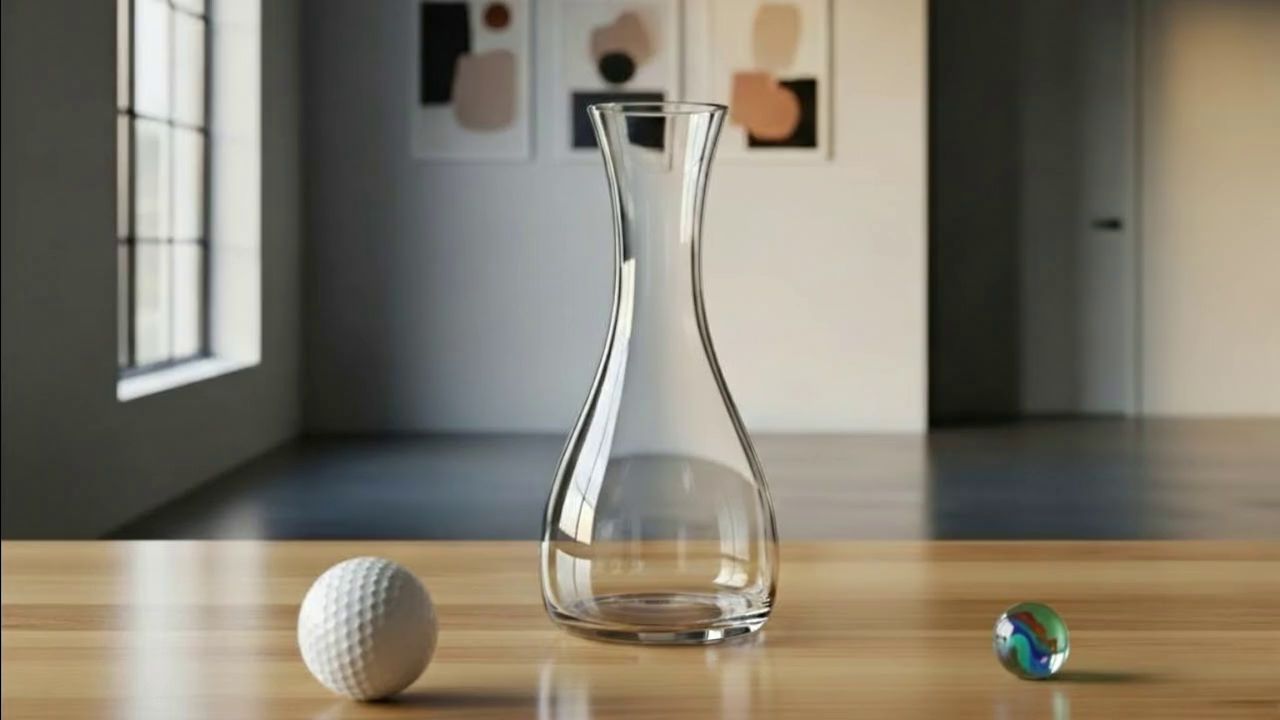}}
    \end{subfigure}
    \hfill
    \begin{subfigure}[t]{\fourfigwidth\textwidth}
        \centering
        \fbox{\includegraphics[width=\linewidth]{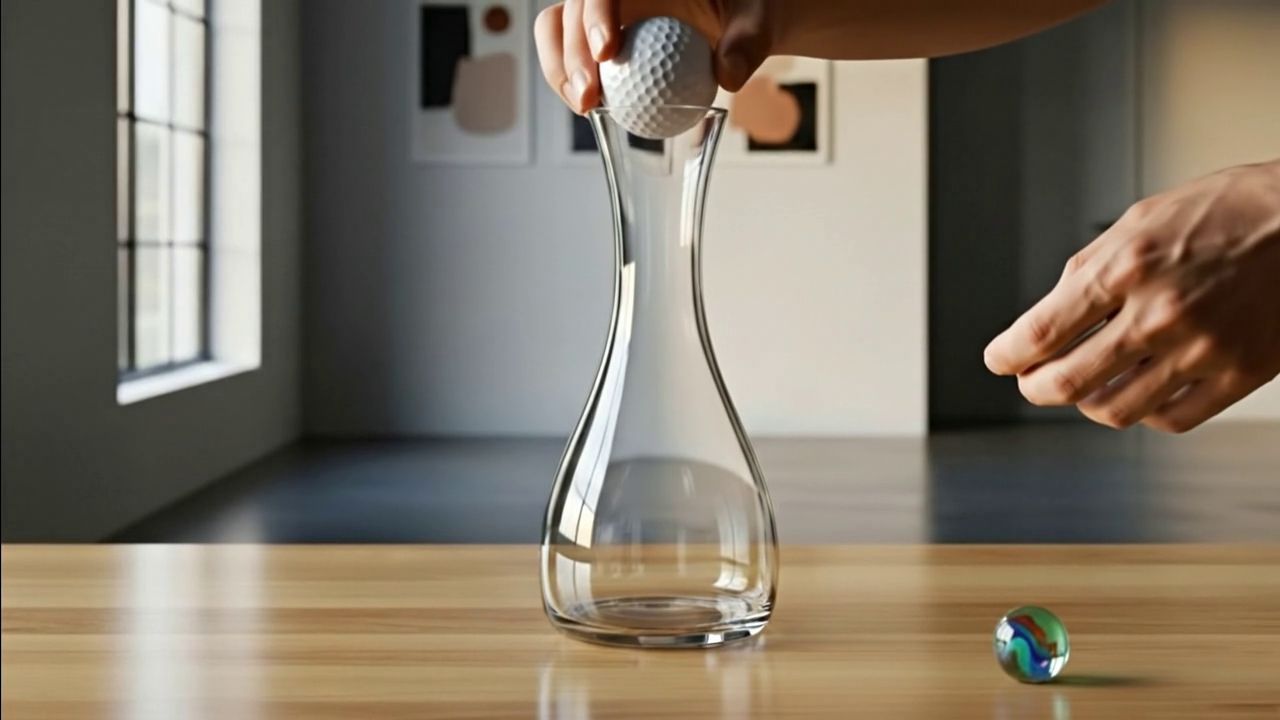}}
    \end{subfigure}
    \hfill
    \begin{subfigure}[t]{\fourfigwidth\textwidth}
        \centering
        \fbox{\includegraphics[width=\linewidth]{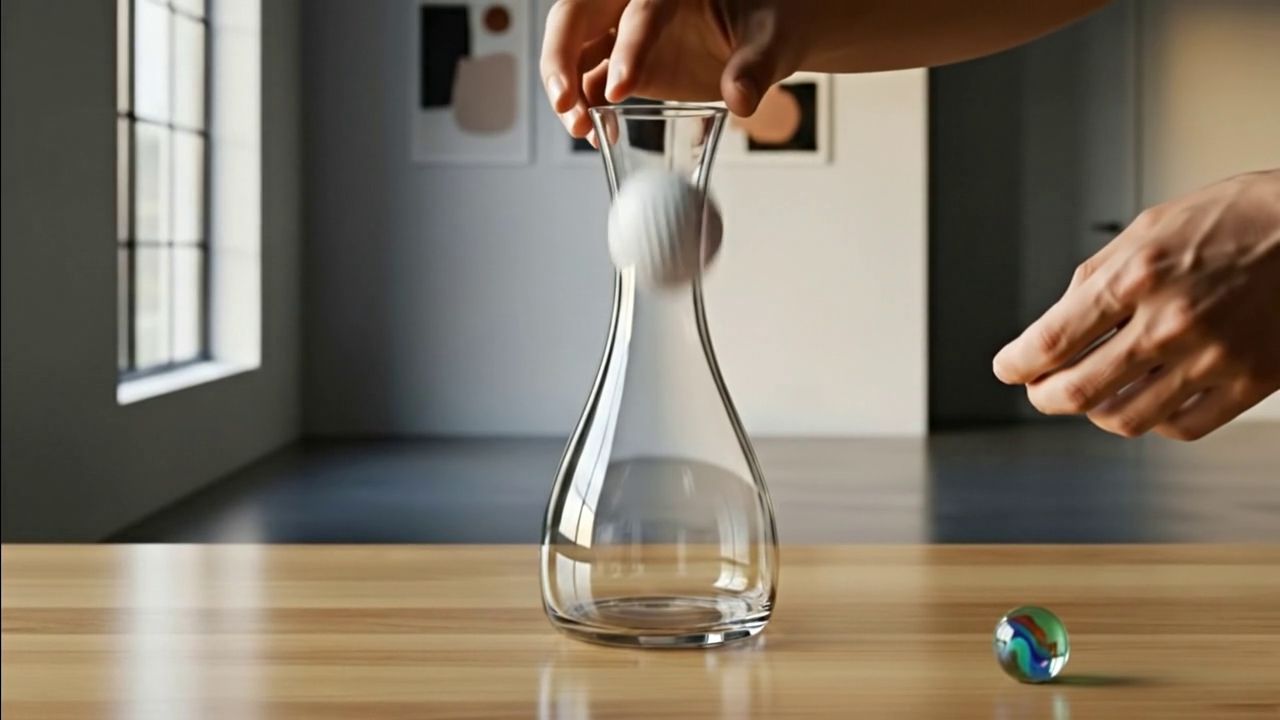}}
    \end{subfigure}
    \hfill
    \begin{subfigure}[t]{\fourfigwidth\textwidth}
        \centering
        \fbox{\includegraphics[width=\linewidth]{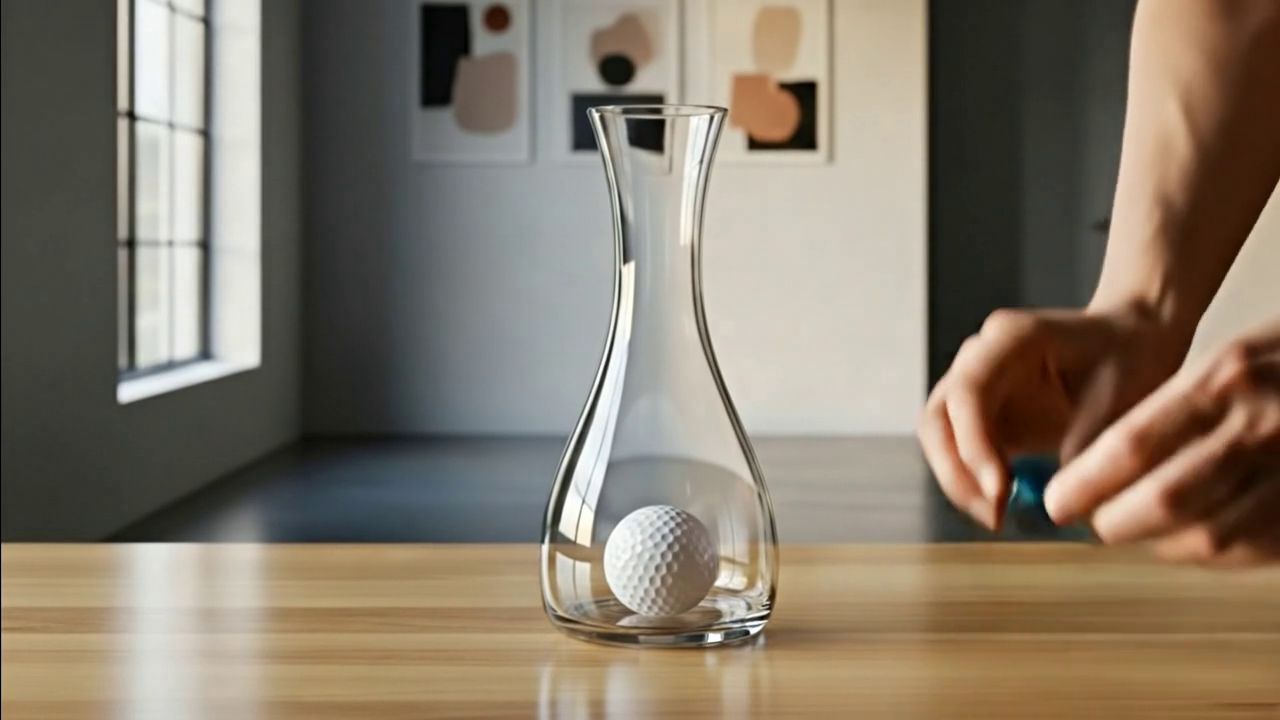}}
    \end{subfigure}
    \caption{\textbf{Bottleneck}. Prompt: \textit{``A person tries to put the golf ball in the vase. Static camera, no pan, no zoom, no dolly..''} Failure: not physically plausible, golf ball is too large to pass through the bottleneck of the vase.}
    \label{fig:bottleneck}
\end{figure}

\begin{figure}[H]
    \centering
    \begin{subfigure}[t]{\fourfigwidth\textwidth}
        \centering
        \fbox{\includegraphics[width=\linewidth]{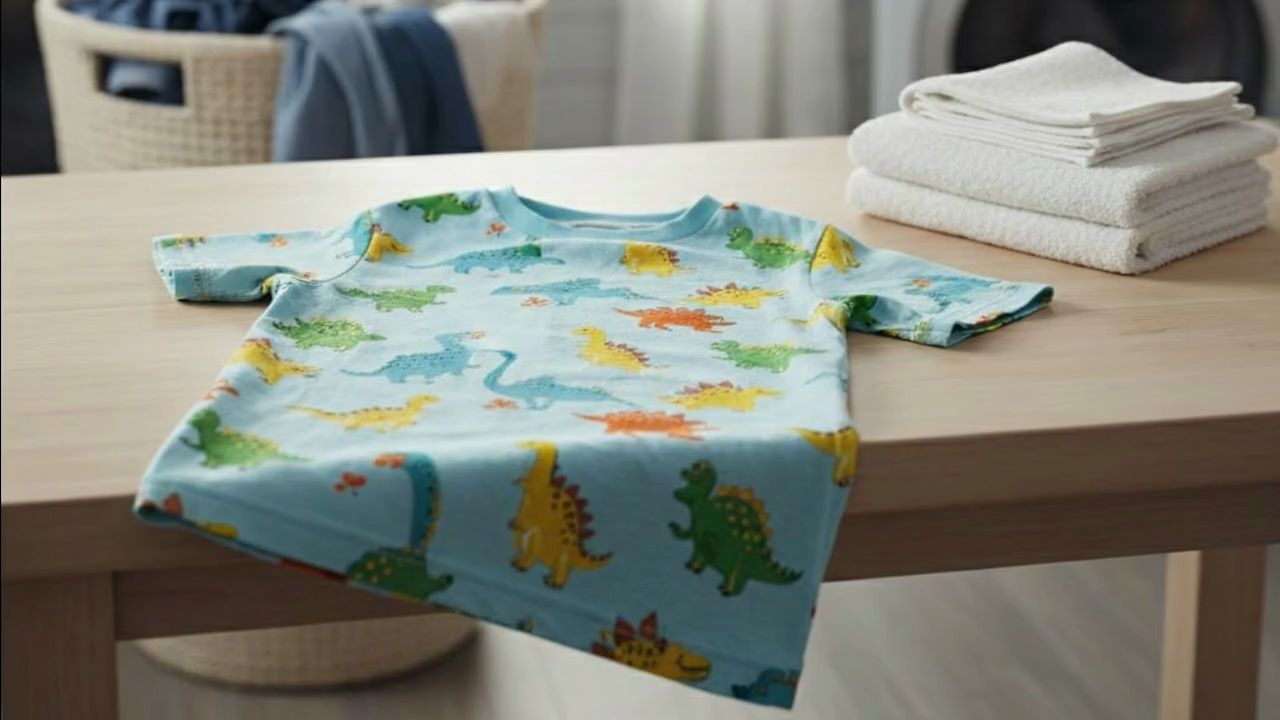}}
    \end{subfigure}
    \hfill
    \begin{subfigure}[t]{\fourfigwidth\textwidth}
        \centering
        \fbox{\includegraphics[width=\linewidth]{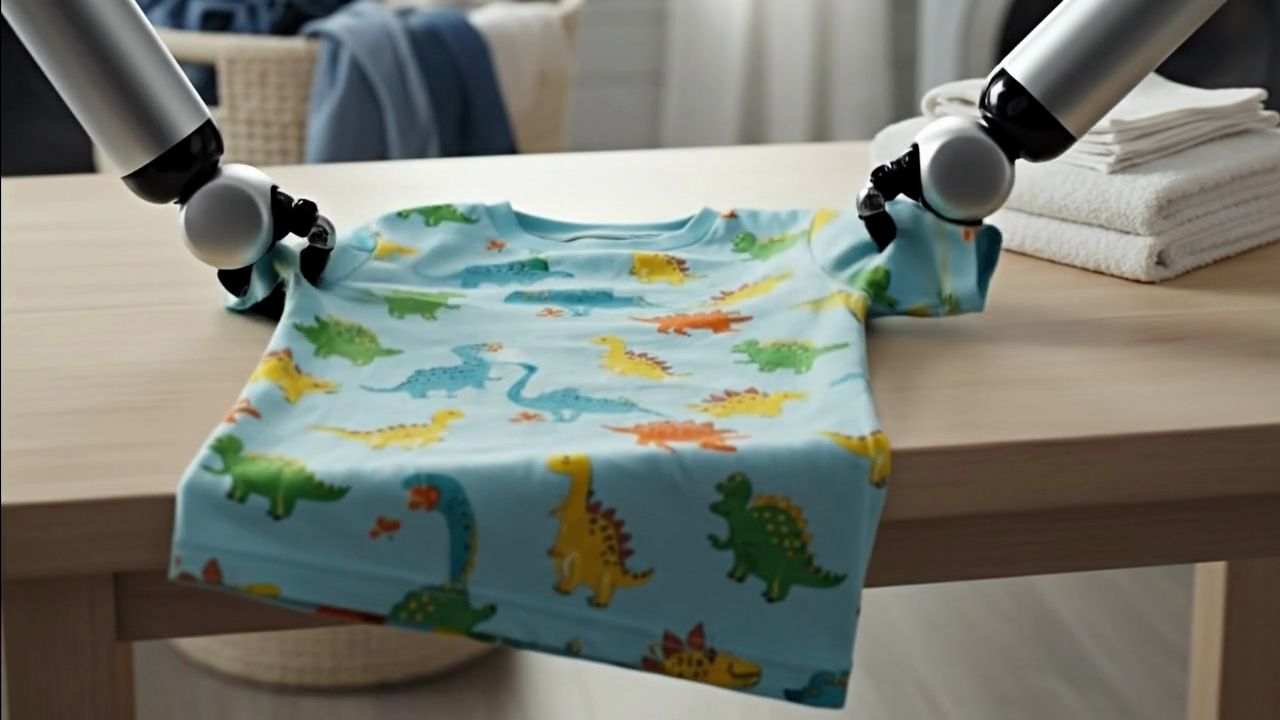}}
    \end{subfigure}
    \hfill    
    \begin{subfigure}[t]{\fourfigwidth\textwidth}
        \centering
        \fbox{\includegraphics[width=\linewidth]{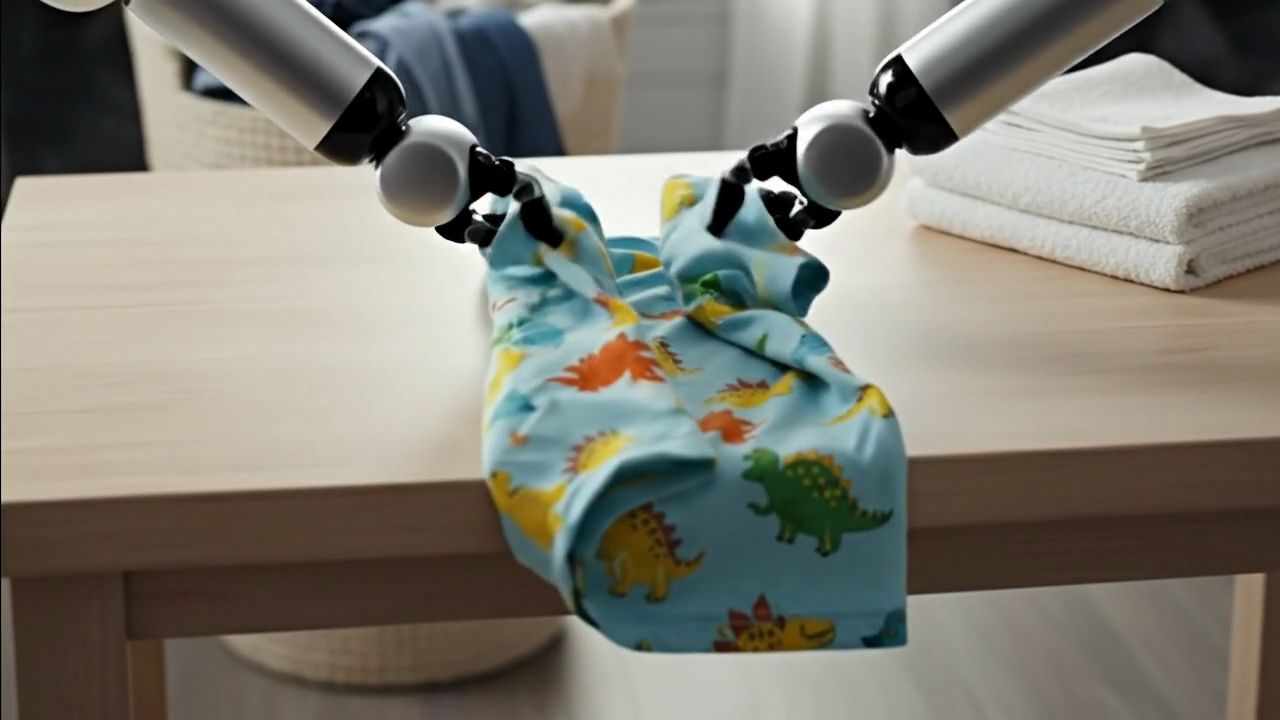}}
    \end{subfigure}
    \hfill    
    \begin{subfigure}[t]{\fourfigwidth\textwidth}
        \centering
        \fbox{\includegraphics[width=\linewidth]{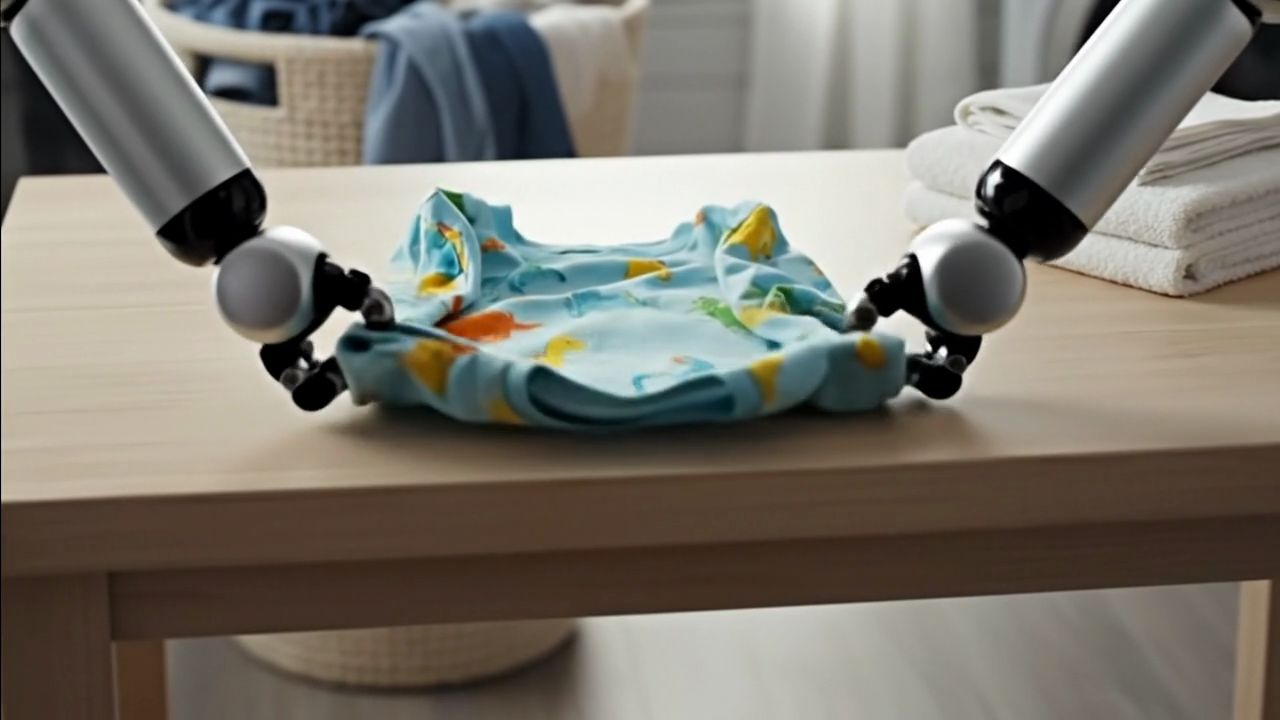}}
    \end{subfigure}
    \caption{\textbf{Laundry folding}. Prompt: \textit{``Generate a video of two metal robotic arms properly folding the t-shirt on the table.} Failure: physics violation, implausible folding movements.}
    \label{fig:laundry-folding}
\end{figure}

\begin{figure}[H]
    \centering
    \begin{subfigure}[t]{\threefigwidth\textwidth}
        \centering
        \fbox{\includegraphics[width=\linewidth]{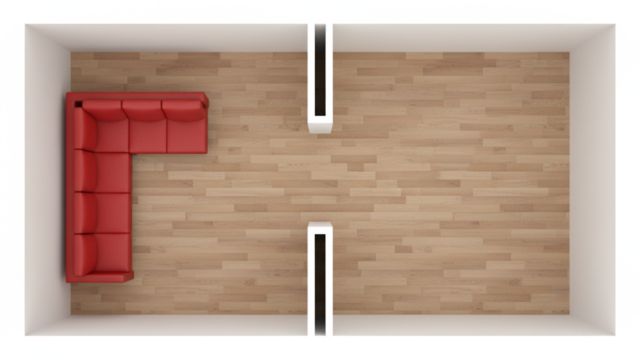}}
    \end{subfigure}
    \hfill
    \begin{subfigure}[t]{\threefigwidth\textwidth}
        \centering
        \fbox{\includegraphics[width=\linewidth]{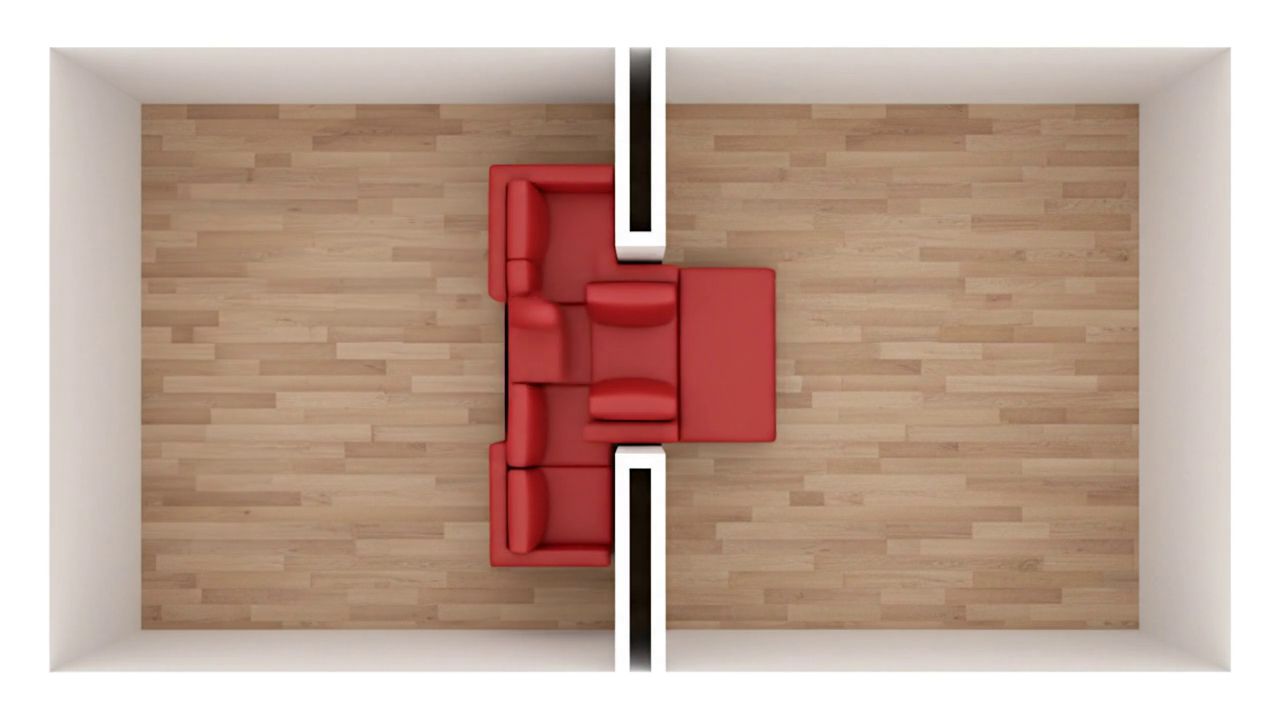}}
    \end{subfigure}
    \hfill    
    \begin{subfigure}[t]{\threefigwidth\textwidth}
        \centering
        \fbox{\includegraphics[width=\linewidth]{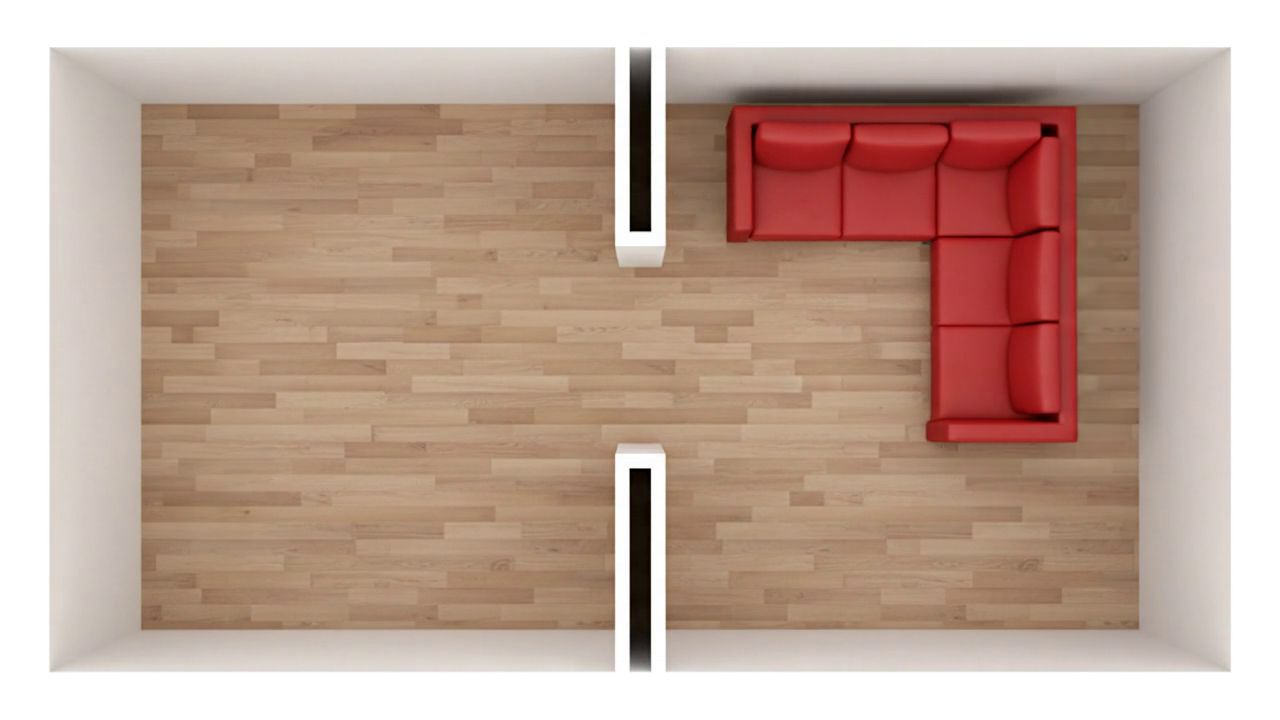}}
    \end{subfigure}
    \caption{\textbf{Motion planning}; inspired by the piano mover's problem. Prompt: \textit{``The red couch slides from the left room over into the right room, skillfully maneuvering to fit through the doorways without bumping into the walls. The walls are fixed: they don't shift or disappear, and no new walls are introduced. Static camera, no pan, no zoom, no dolly.''} Failure: violating rigid-body integrity, not keeping to permissible transformations (rotation, translation).}
    \label{fig:piano-mover}
\end{figure}

\section{LLM use}
Gemini 2.5 Flash and Gemini 2.5 Pro ~\citep{comanici2025gemini} were used for brainstorming task ideas, suggesting related work that we might have otherwise missed, coding support, and to polish human writing.

\section{Image sources}
Where not stated in the figure caption, images were obtained as follows.
\begin{itemize}
    \item \cref{fig:background_removal,fig:edge_detection,fig:segmentation,fig:keypoint_localization,fig:colorization,fig:style_transfer,fig:inpainting,fig:outpainting,fig:superresolution,fig:deblurring,fig:denoising,fig:doodle,fig:text_manipulation,fig:tiling-puzzles}: The original macaw image was generated with Gemini and, depending on the figure, subsequently modified by the authors (e.g., conversion to grayscale, adding noise, adding the monkey with \nanobanana{}).
    \item \cref{fig:low_light}: The input image was obtained from \href{https://huggingface.co/datasets/okhater/lolv2-real}{here} (Apache 2.0 license) based on the LOLv2 dataset \citep{yang2021sparse} and randomly selected. The image was slightly cropped to fit a 16:9 aspect ratio.
    \item \cref{fig:conjunctive_search,fig:color_mixing,fig:visual-explanation,fig:working_memory,fig:physics-body,fig:gravity_air_resistance,fig:water_floating,fig:material_properties,fig:material_optics,fig:color_mixing,fig:fitting_objects,fig:categorizing_objects,fig:scene_composition,fig:transfiguration,fig:drawing,fig:glass_falling,fig:collisions,fig:bottleneck,fig:reposing,fig:tool-use,fig:shape-fitting,fig:linear-equations,fig:laundry-folding,fig:knots,fig:piano-mover}: generated with Gemini.
    \item \cref{fig:visual-jenga}: The input image was obtained from \href{https://pixnio.com/food-and-drink/coffee/coffee-cup-apple-pencil-book}{here} (CC0 license).
    \item \cref{fig:omniglot}: hand drawn by us, inspired by Fig.~1 of the Omniglot paper~\cite{lake2015human}.
    \item \cref{fig:view_synthesis}: sample from Objaverse~\cite{deitke2023objaverse}
    \item \cref{fig:graph_traversal,fig:tree_bfs,fig:raven_logic,fig:connecting_colors,fig:number_sort,fig:maze_mouse,fig:sudoku}: created by us.
    \item \cref{fig:water-puzzle,fig:connect-path,fig:crossword,fig:spot-difference}: original image from Reddit.
    \item \cref{fig:arc-agi}: hand drawn by us, inspired by ARC-AGI~\citep{chollet2019measure}.
    \item \cref{fig:edge-detection-graded}: sample from BIPEDv2~\citep{soria2020dexined, soria2023dexined_ext}.
    \item \cref{fig:depth,fig:normals,fig:force-motion-prompting}: generated with Gemini, then annotated by us.
    \item \cref{fig:eulerian-path,fig:visual_iq}: hand drawn by us. Inspired by original images from Reddit.
    \item \cref{fig:hand-dexterity,fig:affordance}: The robot hands are extracted from a frame in \href{https://www.youtube.com/watch?v=rWHk4ht-boM}{this video} and were subsequently adapted with \nanobanana{}. The hands holding Baoding balls were obtained from \href{https://www.ebay.com/itm/324085657759}{here}.
\end{itemize}

%% file: sections/03a_qualitative_figures.tex
\section{Qualitative results:\\ \textcolor{perceptioncol}{\textbf{Perception}}, \textcolor{modelingcol}{\textbf{Modeling}}, \textcolor{manipulationcol}{\textbf{Manipulation}}, \textcolor{reasoningcol}{\textbf{Reasoning}}}
\label{app:qualitative}

\subsection{\textcolor{perceptioncol}{\textbf{Perception}}}
\label{app:_perception_qualitative}

\begin{figure}[H]
    \centering
    \begin{subfigure}[t]{\threefigwidth\textwidth}
        \centering
        \fbox{\includegraphics[width=\linewidth]{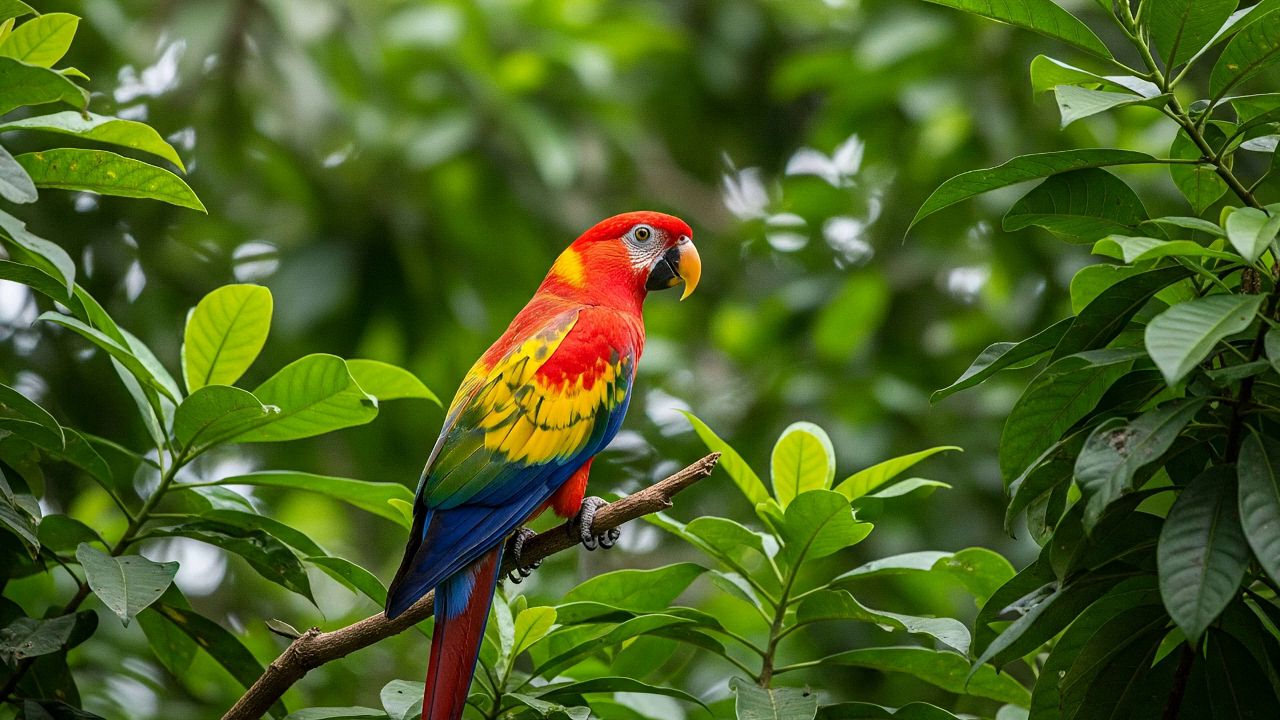}}
    \end{subfigure}
    \hfill
    \begin{subfigure}[t]{\threefigwidth\textwidth}
        \centering
        \fbox{\includegraphics[width=\linewidth]{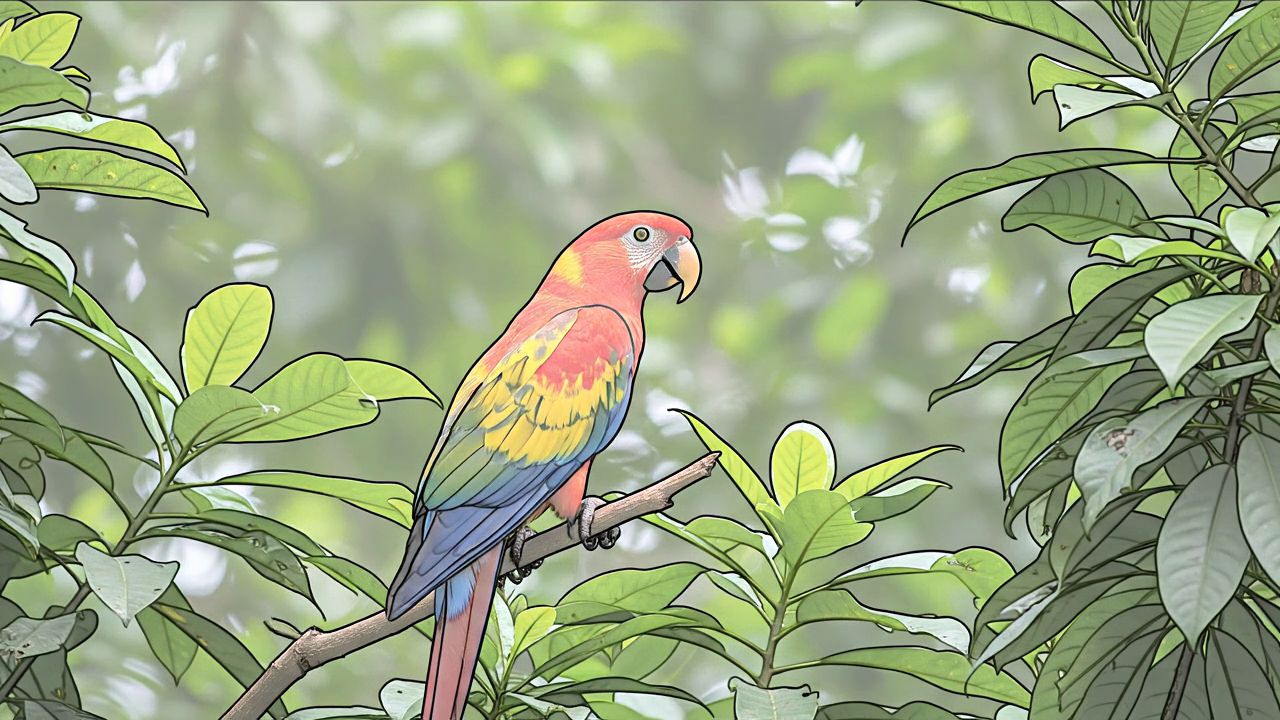}}
    \end{subfigure}
    \hfill    
    \begin{subfigure}[t]{\threefigwidth\textwidth}
        \centering
        \fbox{\includegraphics[width=\linewidth]{figures-compressed/edge-detection_frame_0110.jpg}}
    \end{subfigure}
    \caption{\textbf{Edge detection}. Prompt: \textit{``All edges in this image become more salient by transforming into black outlines. Then, all objects fade away, with just the edges remaining on a white background. Static camera perspective, no zoom or pan.''} Success rate:~0.92.}
    \label{fig:edge_detection}
\end{figure}

\begin{figure}[H]
    \centering
    \begin{subfigure}[t]{\threefigwidth\textwidth}
        \centering
        \fbox{\includegraphics[width=\linewidth]{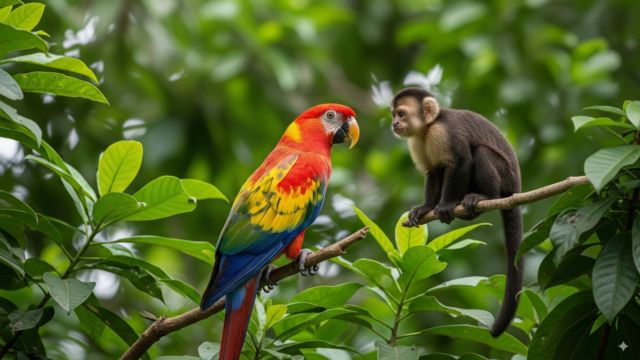}}
    \end{subfigure}
    \hfill
    \begin{subfigure}[t]{\threefigwidth\textwidth}
        \centering
        \fbox{\includegraphics[width=\linewidth]{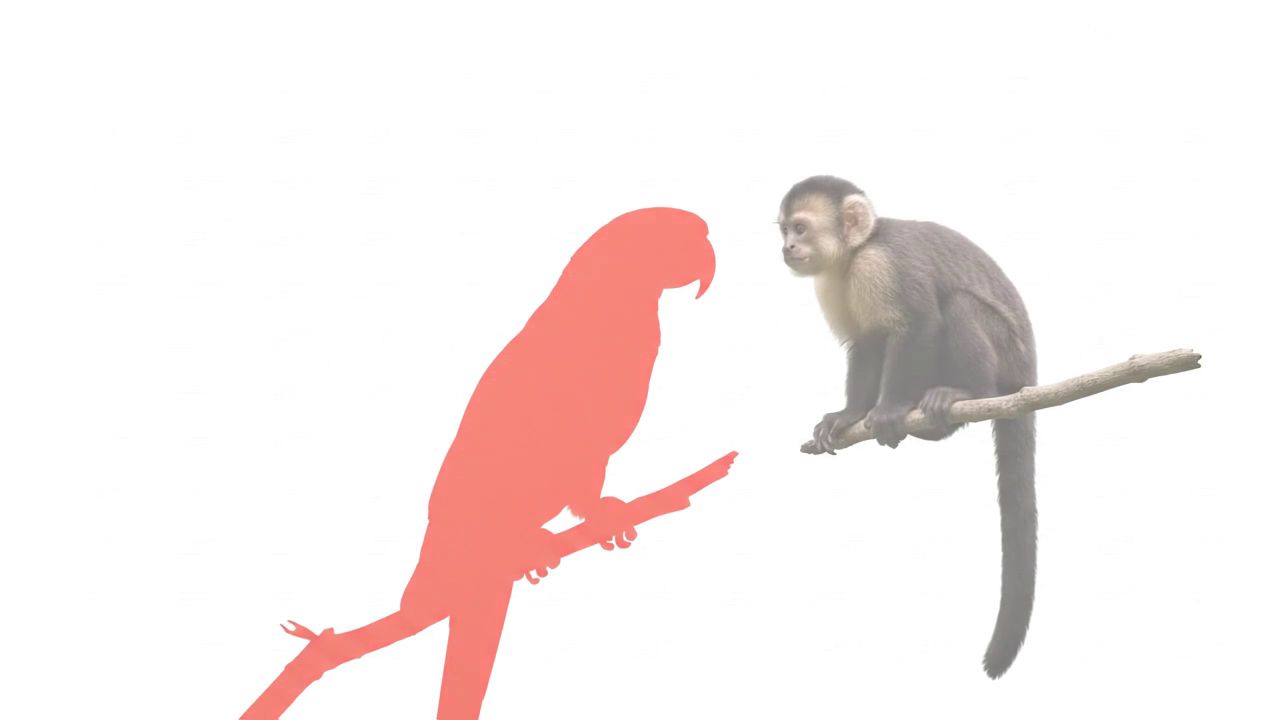}}
    \end{subfigure}
    \hfill    
    \begin{subfigure}[t]{\threefigwidth\textwidth}
        \centering
        \fbox{\includegraphics[width=\linewidth]{figures-compressed/segmentation_macaw-monkey-2_segmentation_frame140.jpg}}
    \end{subfigure}
    \caption{\textbf{Segmentation}. Prompt: \textit{``Create an animation of instance segmentation being performed on this photograph: each distinct entity is overlaid in a different flat color [...]''} (full prompt:~\cref{app:segmentation}). Success rate:~0.33.}
    \label{fig:segmentation}
\end{figure}

\begin{figure}[H]
    \centering
    \begin{subfigure}[t]{\threefigwidth\textwidth}
        \centering
        \fbox{\includegraphics[width=\linewidth]{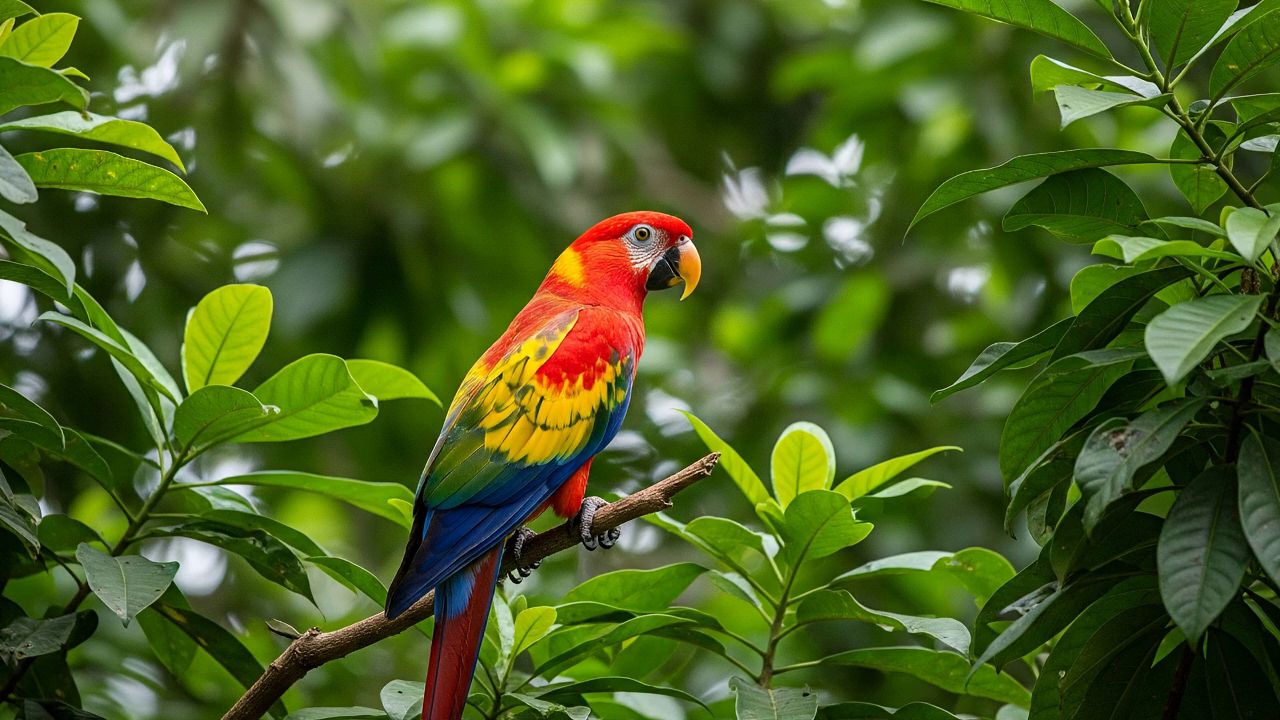}}
    \end{subfigure}
    \hfill
    \begin{subfigure}[t]{\threefigwidth\textwidth}
        \centering
        \fbox{\includegraphics[width=\linewidth]{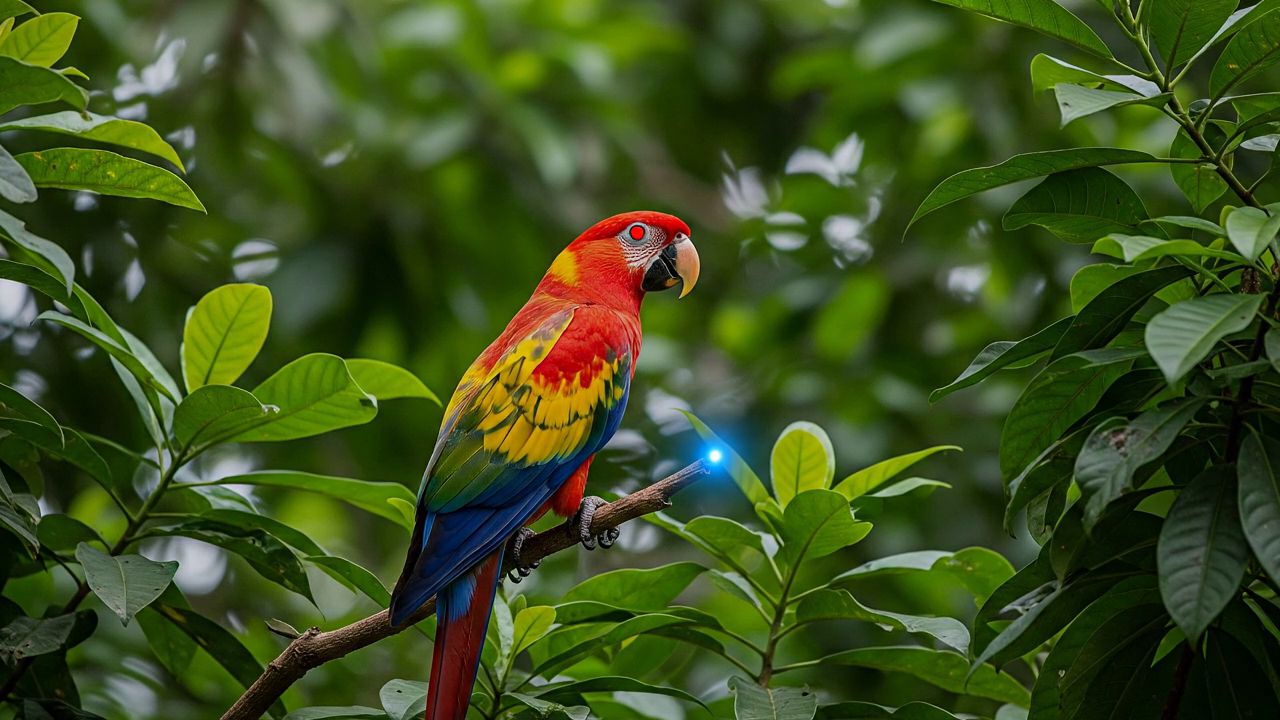}}
    \end{subfigure}
    \hfill    
    \begin{subfigure}[t]{\threefigwidth\textwidth}
        \centering
        \fbox{\includegraphics[width=\linewidth]{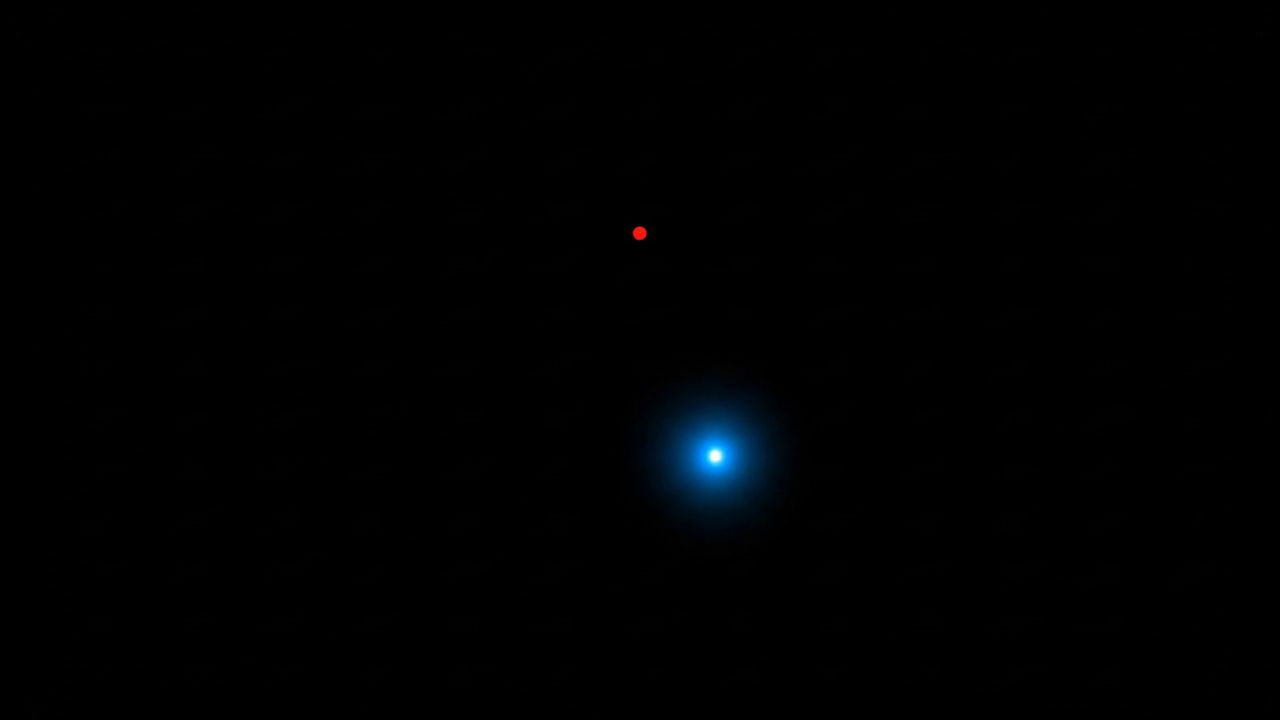}}
    \end{subfigure}
    \caption{\textbf{Keypoint localization}. Prompt: \textit{``Add a bright blue dot at the tip of the branch on which the macaw is sitting. The macaw’s eye turns bright red. Everything else turns pitch black. Static camera perspective, no zoom or pan.''} Success rate:~0.58.}
    \label{fig:keypoint_localization}
\end{figure}

\begin{figure}[H]
    \centering
    \begin{subfigure}[t]{\threefigwidth\textwidth}
        \centering
        \fbox{\includegraphics[width=\linewidth]{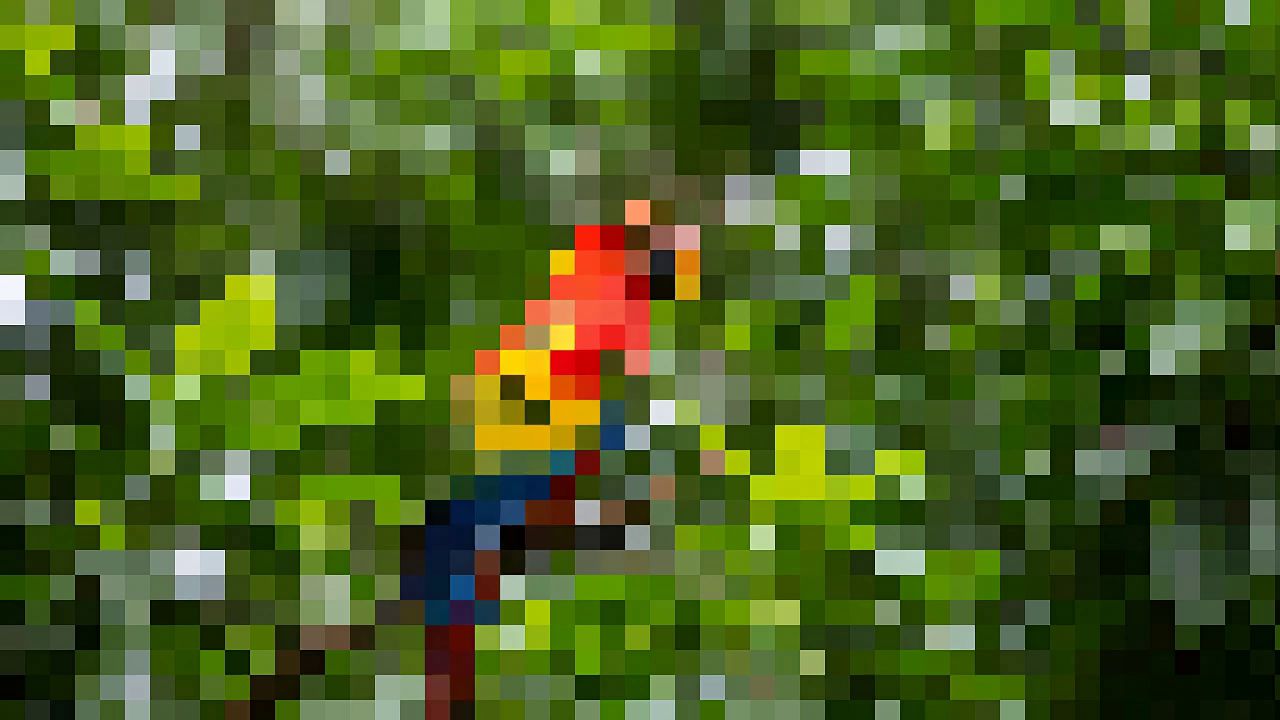}}
    \end{subfigure}
    \hfill
    \begin{subfigure}[t]{\threefigwidth\textwidth}
        \centering
        \fbox{\includegraphics[width=\linewidth]{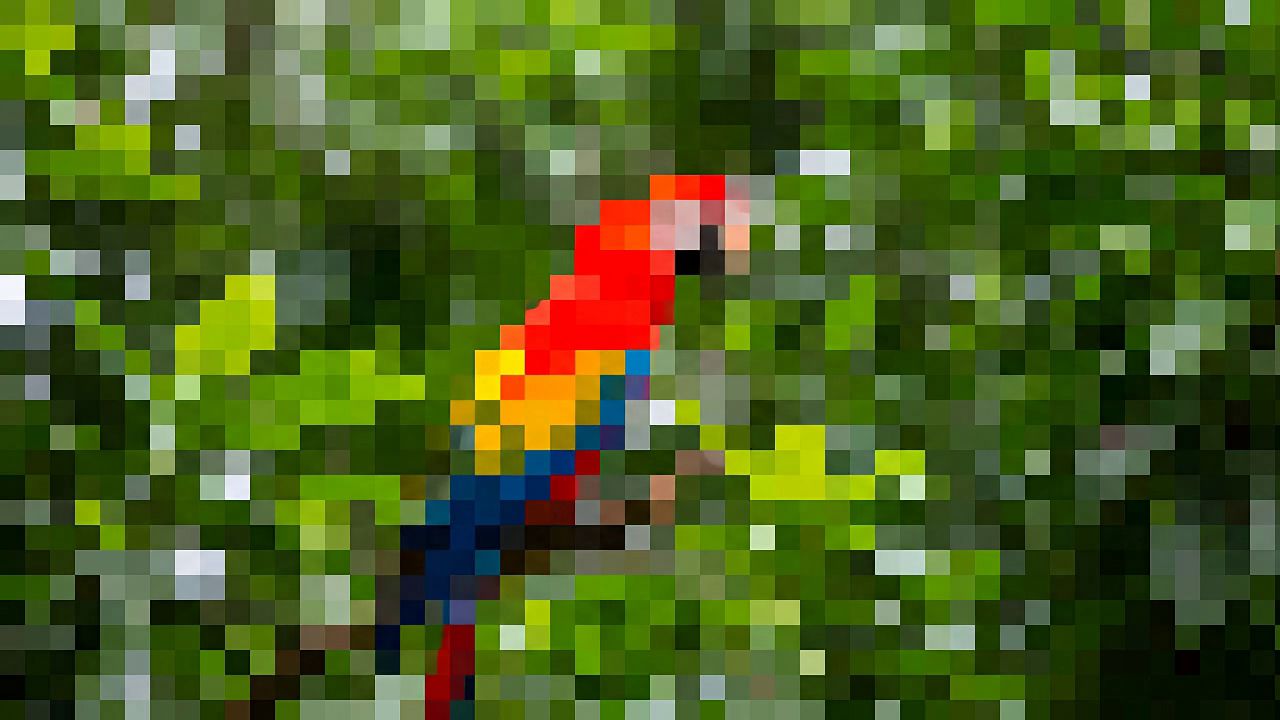}}
    \end{subfigure}
    \hfill    
    \begin{subfigure}[t]{\threefigwidth\textwidth}
        \centering
        \fbox{\includegraphics[width=\linewidth]{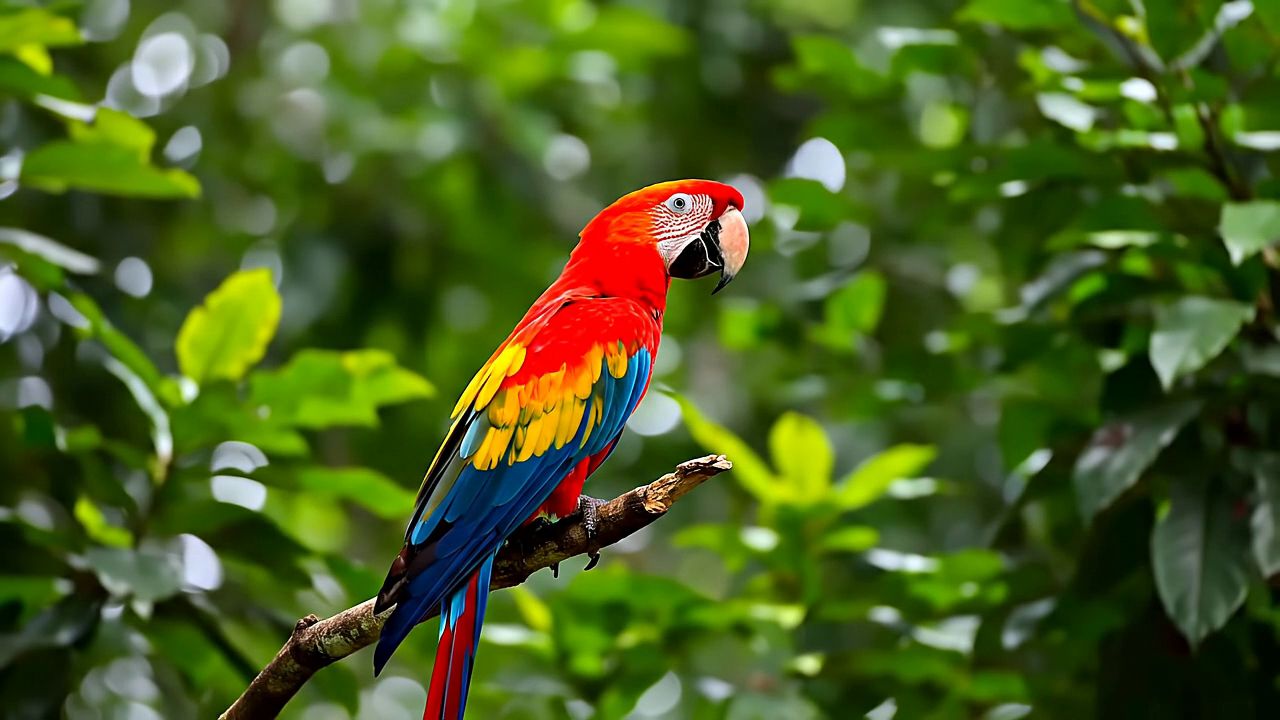}}
    \end{subfigure}
    \caption{\textbf{Super-resolution}. Prompt: \textit{``Perform superresolution on this image. Static camera perspective, no zoom or pan.''} Success rate:~0.75.}
    \label{fig:superresolution}
\end{figure}

\begin{figure}[H]
    \centering
    \begin{subfigure}[t]{\threefigwidth\textwidth}
        \centering
        \fbox{\includegraphics[width=\linewidth]{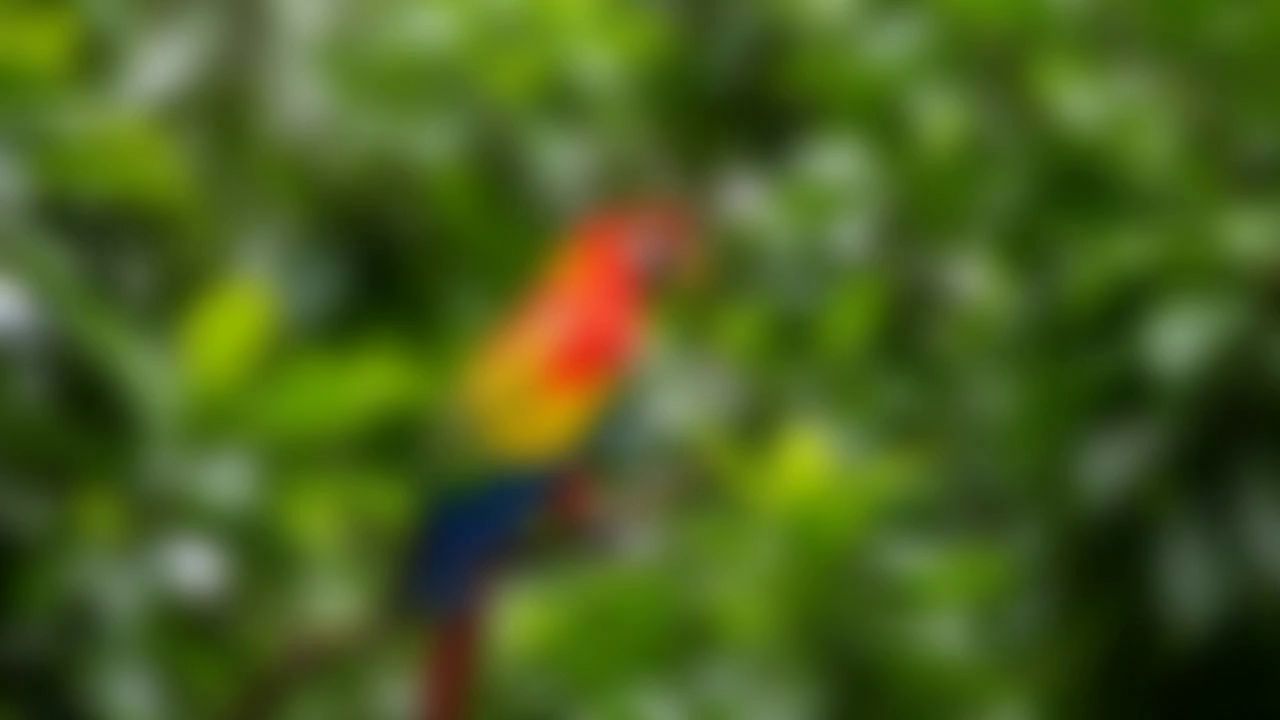}}
    \end{subfigure}
    \hfill
    \begin{subfigure}[t]{\threefigwidth\textwidth}
        \centering
        \fbox{\includegraphics[width=\linewidth]{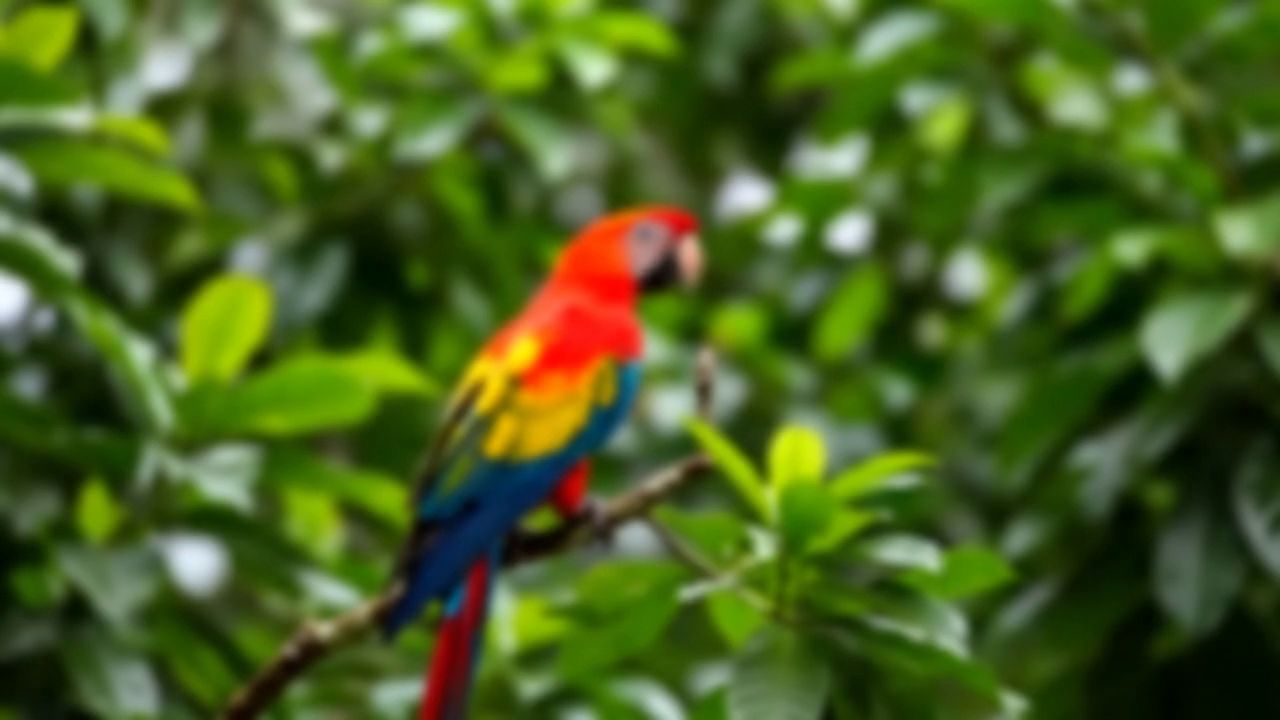}}
    \end{subfigure}
    \hfill    
    \begin{subfigure}[t]{\threefigwidth\textwidth}
        \centering
        \fbox{\includegraphics[width=\linewidth]{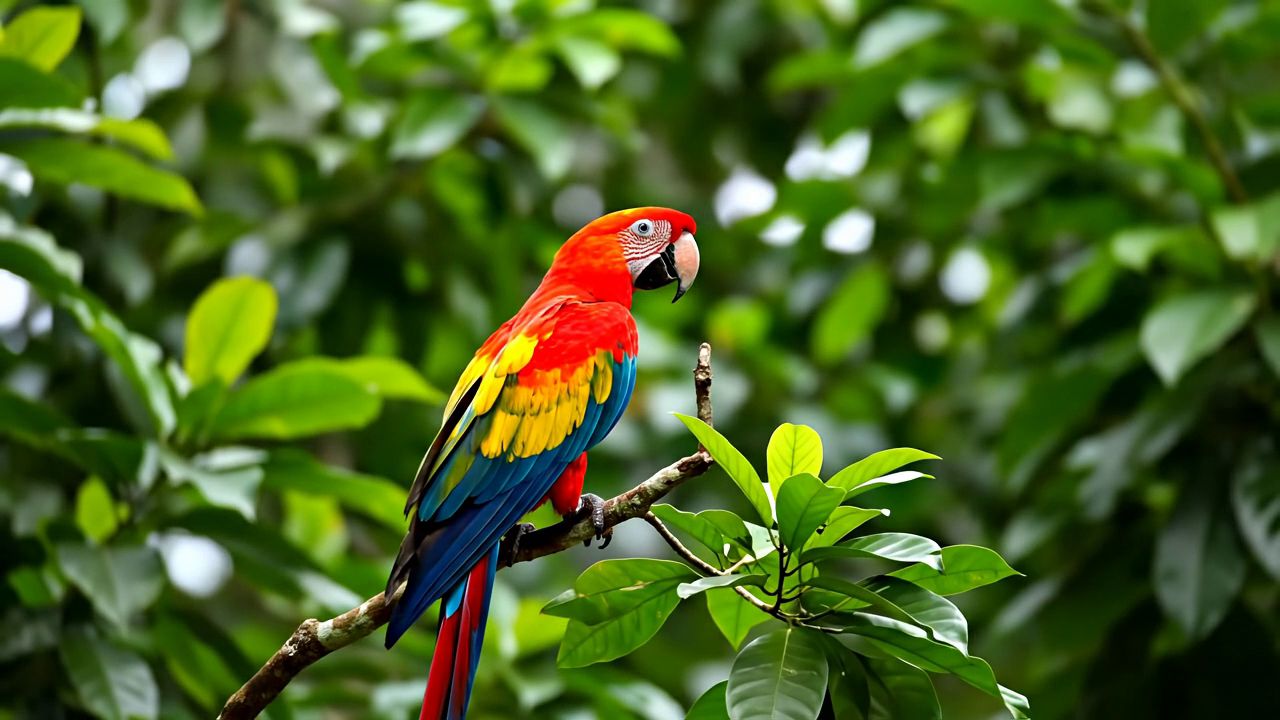}}
    \end{subfigure}
    \caption{\textbf{Blind deblurring}. Prompt: \textit{``Unblur image including background. Static camera perspective, no zoom or pan.''} Success rate:~1.0.}
    \label{fig:deblurring}
\end{figure}

\begin{figure}[H]
    \centering
    \begin{subfigure}[t]{\threefigwidth\textwidth}
        \centering
        \fbox{\includegraphics[width=\linewidth]{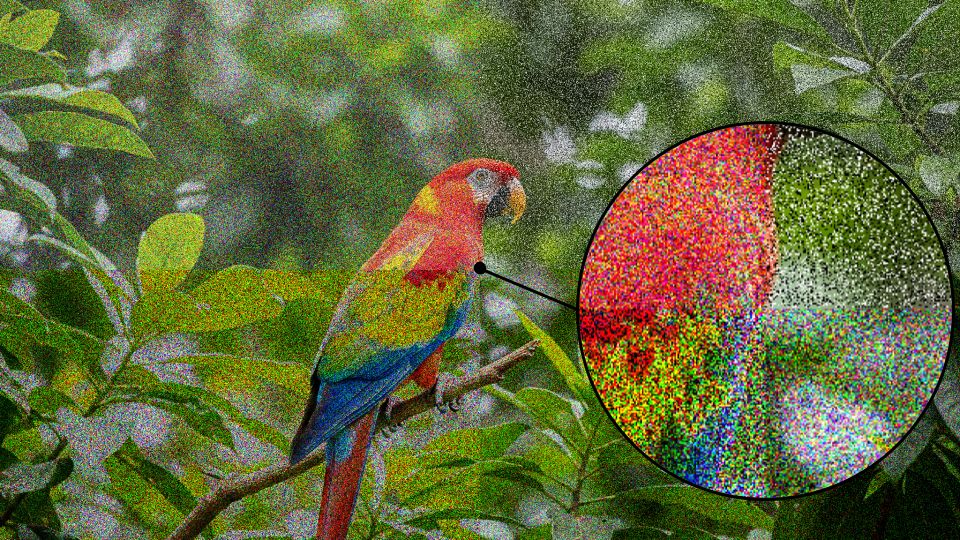}}
    \end{subfigure}
    \hfill
    \begin{subfigure}[t]{\threefigwidth\textwidth}
        \centering
        \fbox{\includegraphics[width=\linewidth]{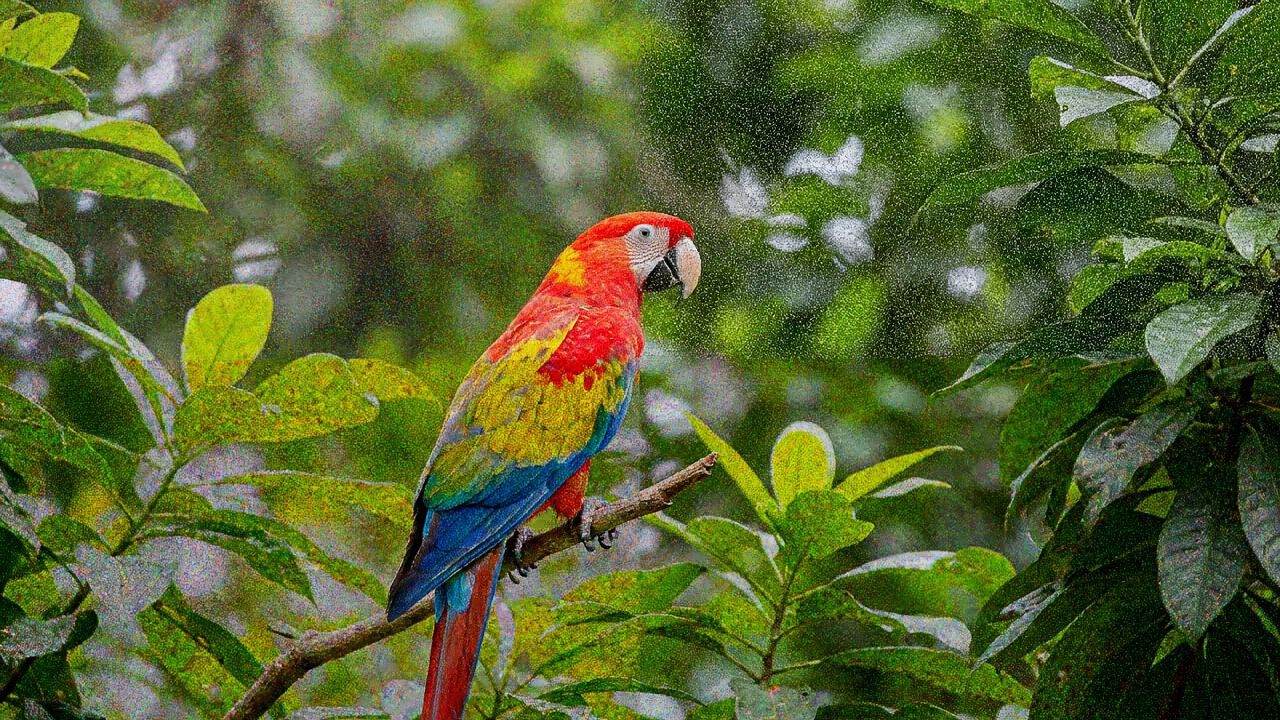}}
    \end{subfigure}
    \hfill    
    \begin{subfigure}[t]{\threefigwidth\textwidth}
        \centering
        \fbox{\includegraphics[width=\linewidth]{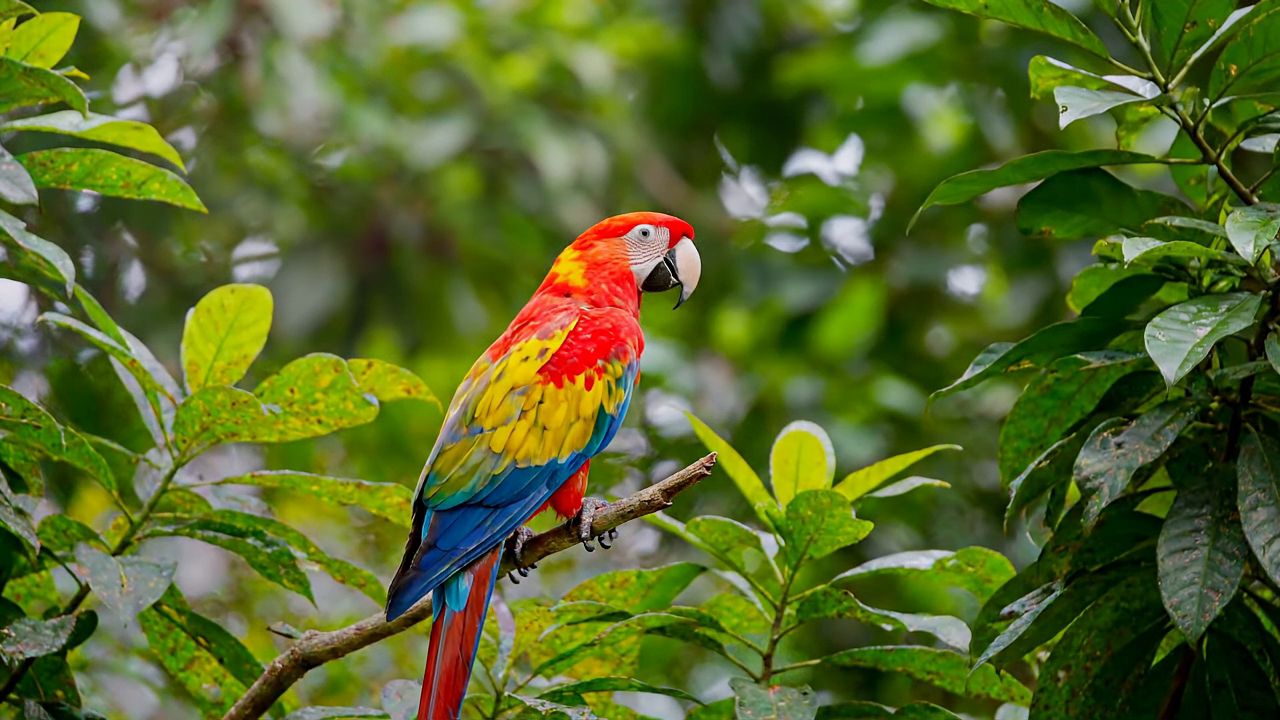}}
    \end{subfigure}
    \caption{\textbf{Blind denoising}. Each quadrant was corrupted with a different type of noise. Clockwise from top left: Gaussian noise, salt-and-pepper noise, speckle noise, shot noise. Prompt: \textit{``Remove the noise from this image. Static camera perspective, no zoom or pan.''} Success rate:~1.0.}
    \label{fig:denoising}
\end{figure}

\begin{figure}[H]
    \centering
    \begin{subfigure}[t]{\threefigwidth\textwidth}
        \centering
        \fbox{\includegraphics[width=\linewidth]{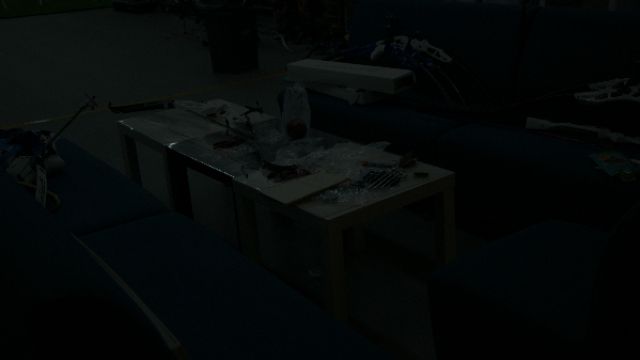}}
        \caption{Original low-light image}
    \end{subfigure}
    \hfill
    \begin{subfigure}[t]{\threefigwidth\textwidth}
        \centering
        \fbox{\includegraphics[width=\linewidth]{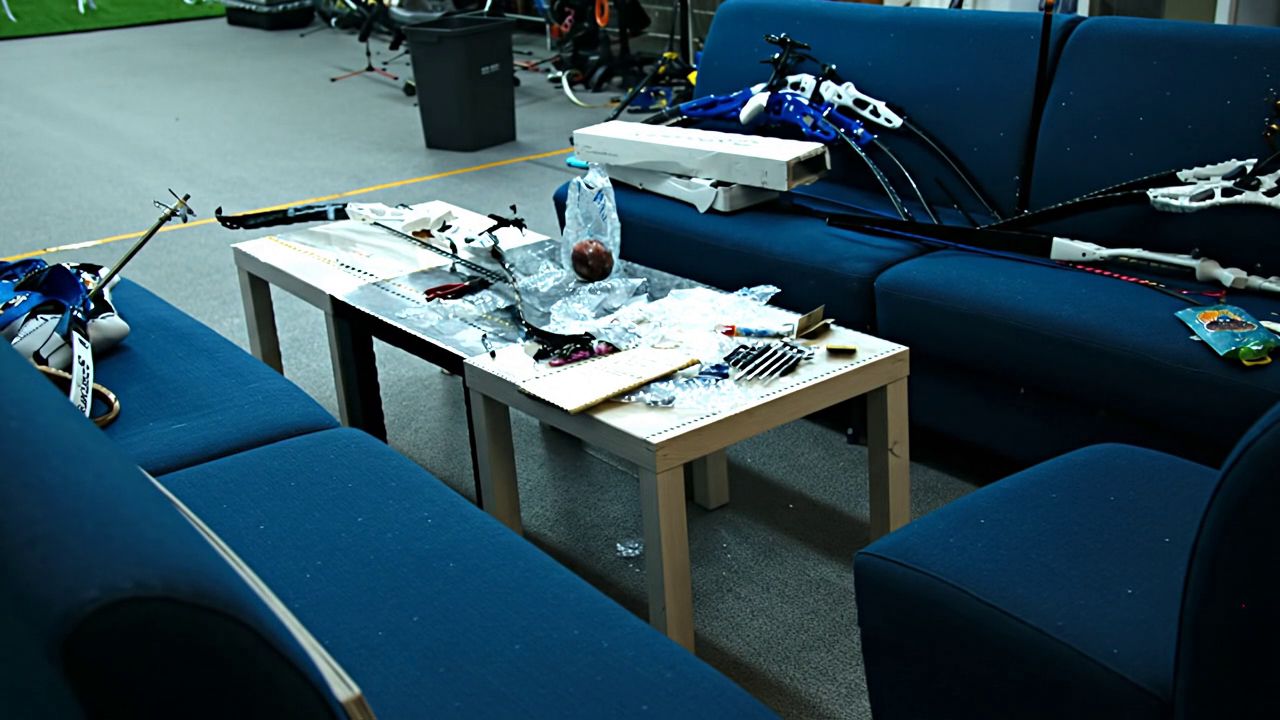}}
        \caption{\veothree{}-generated lit image}
    \end{subfigure}
    \hfill
    \begin{subfigure}[t]{\threefigwidth\textwidth}
        \centering
        \fbox{\includegraphics[width=\linewidth]{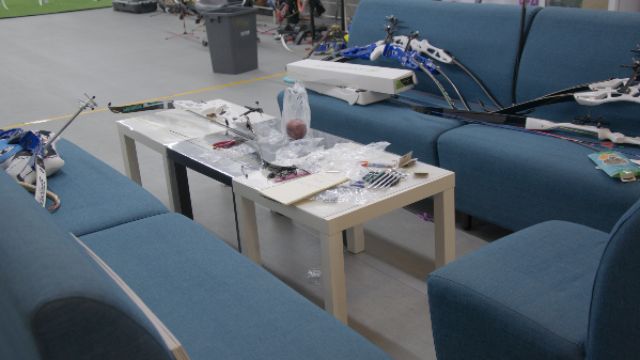}}
        \caption{Ground-truth lit image}
    \end{subfigure}
    \caption{\textbf{Low-light enhancing}. Prompt: \textit{``Fully restore the light in this image. Static camera perspective, no zoom or pan.''} Success rate:~0.92. Image and ground-truth source: LOLv2 dataset~\citep{yang2021sparse}.}
    \label{fig:low_light}
\end{figure}

\begin{figure}[H]
    \centering
    \begin{subfigure}[t]{\threefigwidth\textwidth}
        \centering
        \fbox{\includegraphics[width=\linewidth]{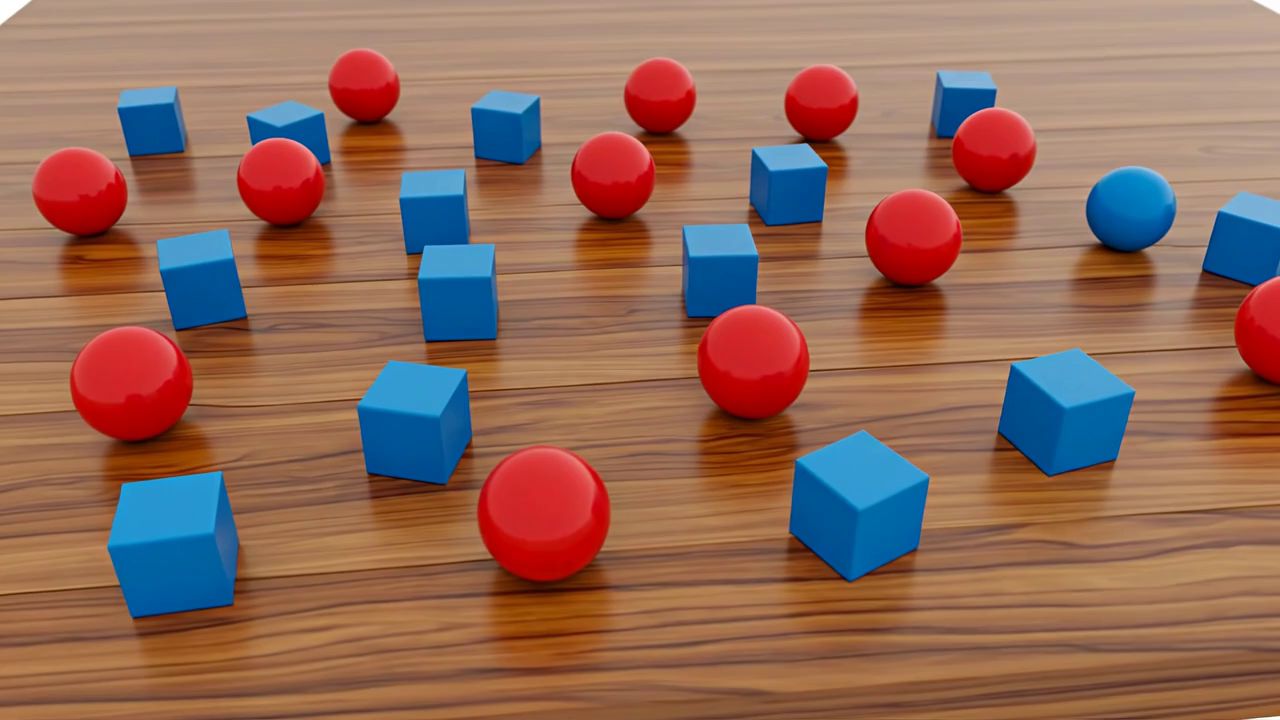}}
    \end{subfigure}
    \hfill
    \begin{subfigure}[t]{\threefigwidth\textwidth}
        \centering
        \fbox{\includegraphics[width=\linewidth]{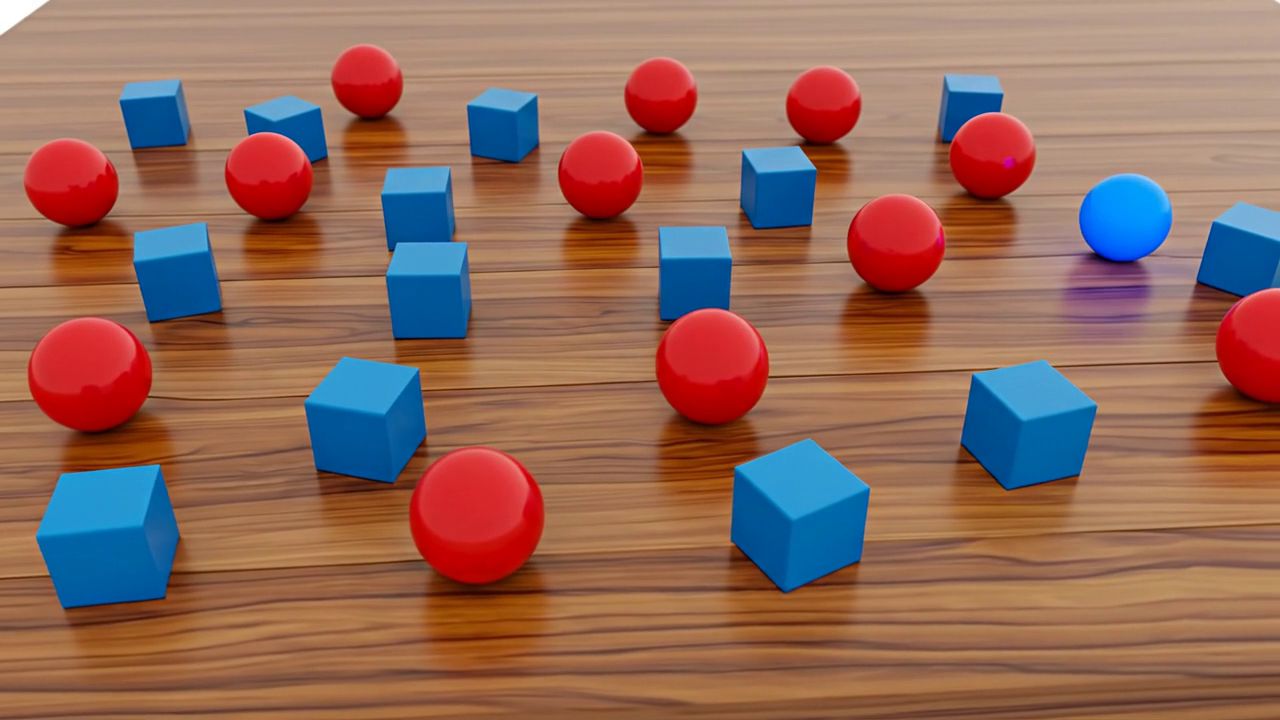}}
    \end{subfigure}
    \hfill    
    \begin{subfigure}[t]{\threefigwidth\textwidth}
        \centering
        \fbox{\includegraphics[width=\linewidth]{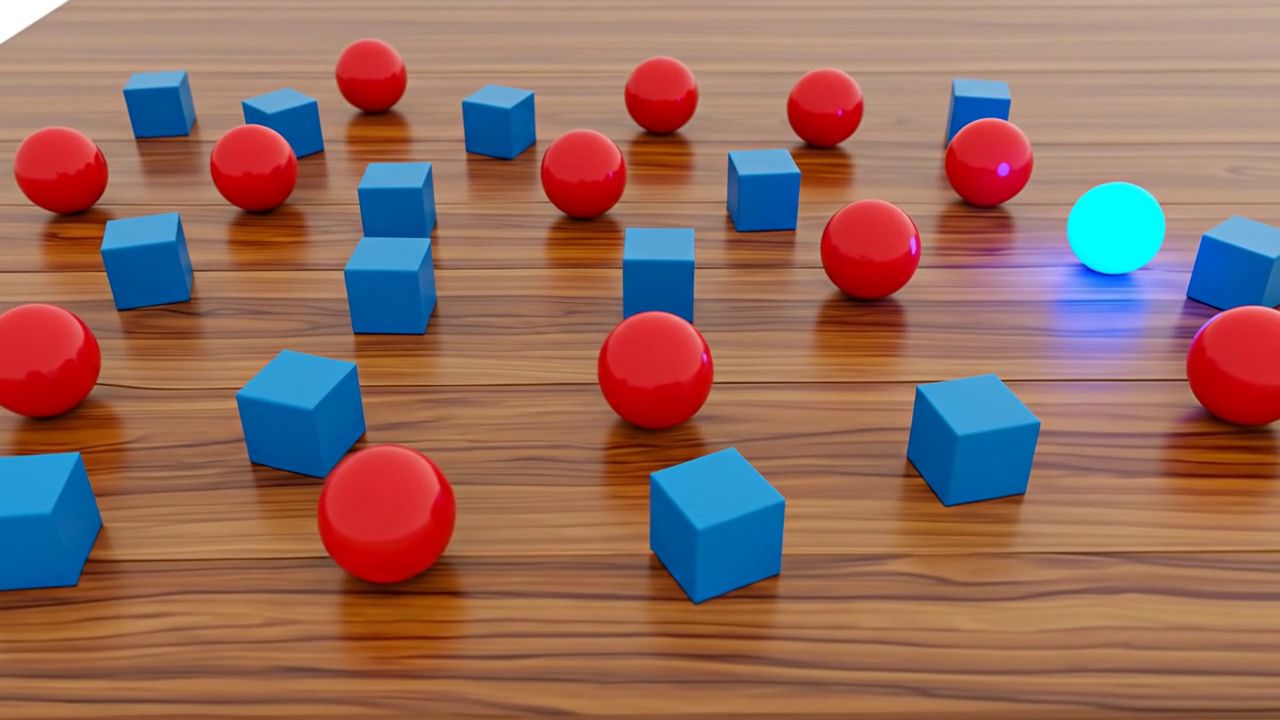}}
    \end{subfigure}
    \caption{\textbf{Conjunctive search / binding problem}. Prompt: \textit{``The blue ball instantly begins to glow. Static camera perspective, no zoom no pan no movement no dolly no rotation.''} Success rate:~0.75. Inspiration: \cite{campbell2024understanding}.}
    \label{fig:conjunctive_search}
\end{figure}

\begin{figure}[H]
    \centering
    \begin{subfigure}[t]{\threefigwidth\textwidth}
        \centering
        \fbox{\includegraphics[width=\linewidth]{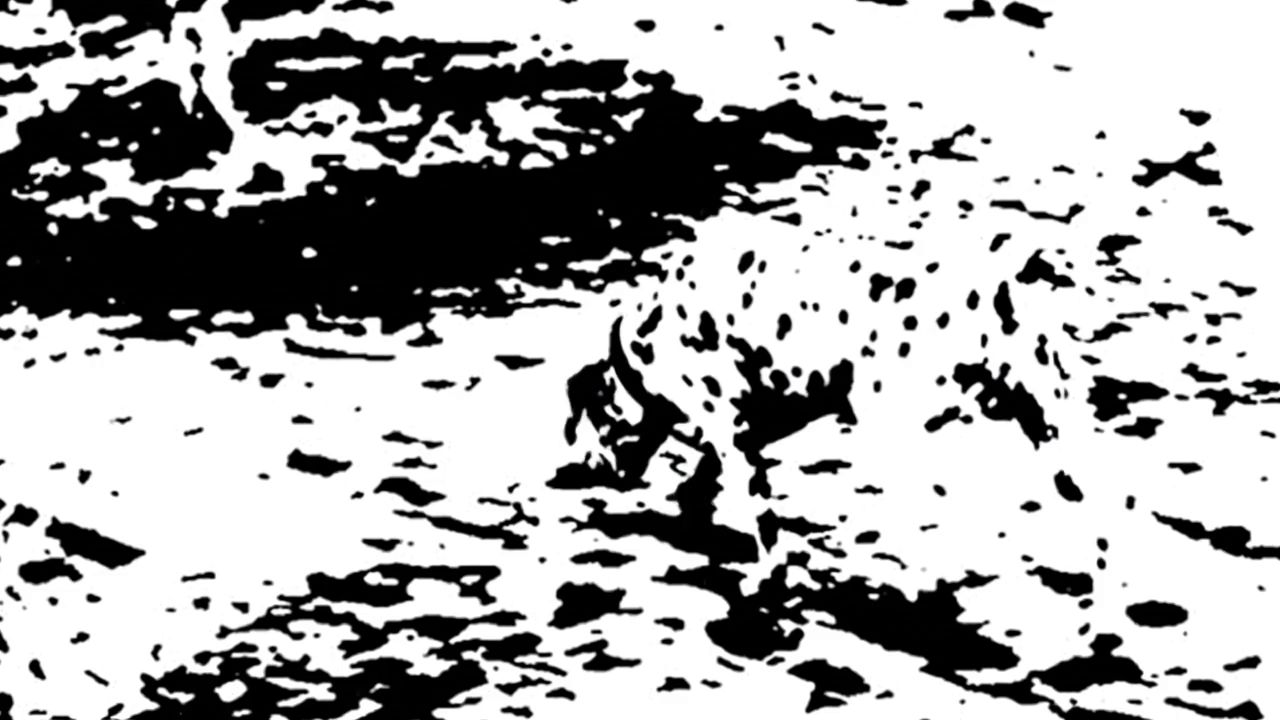}}
    \end{subfigure}
    \hfill
    \begin{subfigure}[t]{\threefigwidth\textwidth}
        \centering
        \fbox{\includegraphics[width=\linewidth]{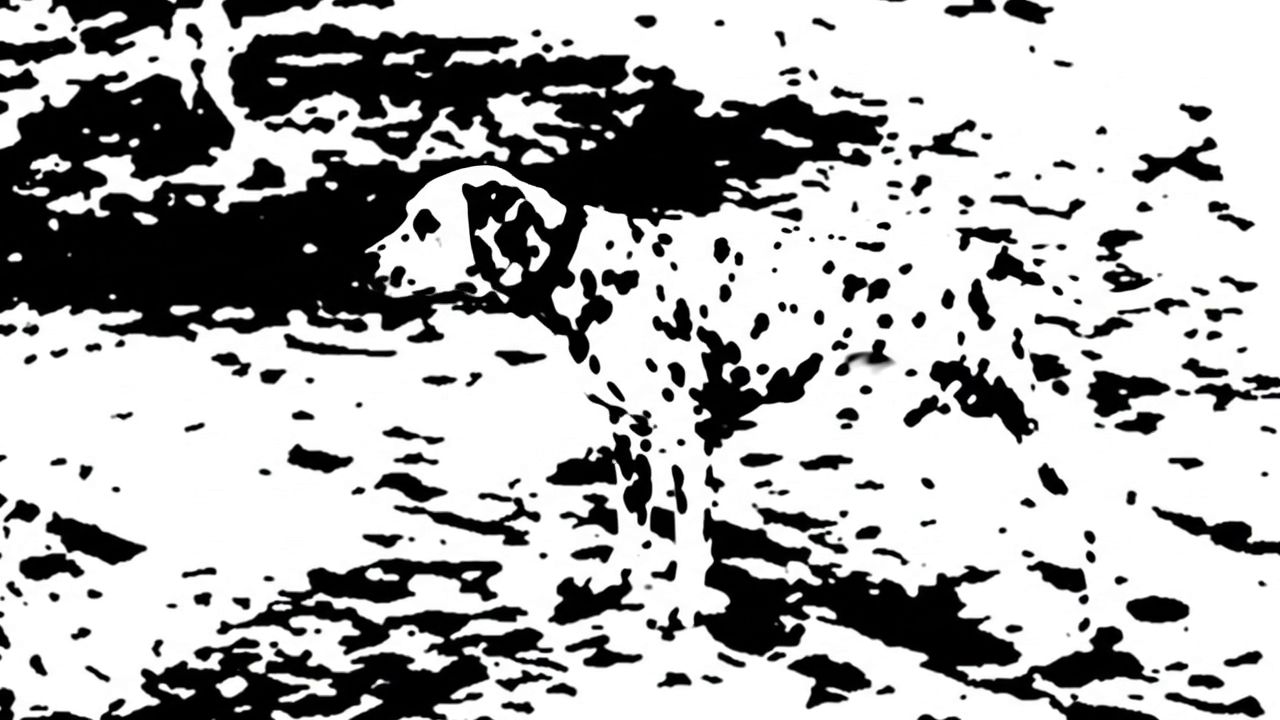}}
    \end{subfigure}
    \hfill    
    \begin{subfigure}[t]{\threefigwidth\textwidth}
        \centering
        \fbox{\includegraphics[width=\linewidth]{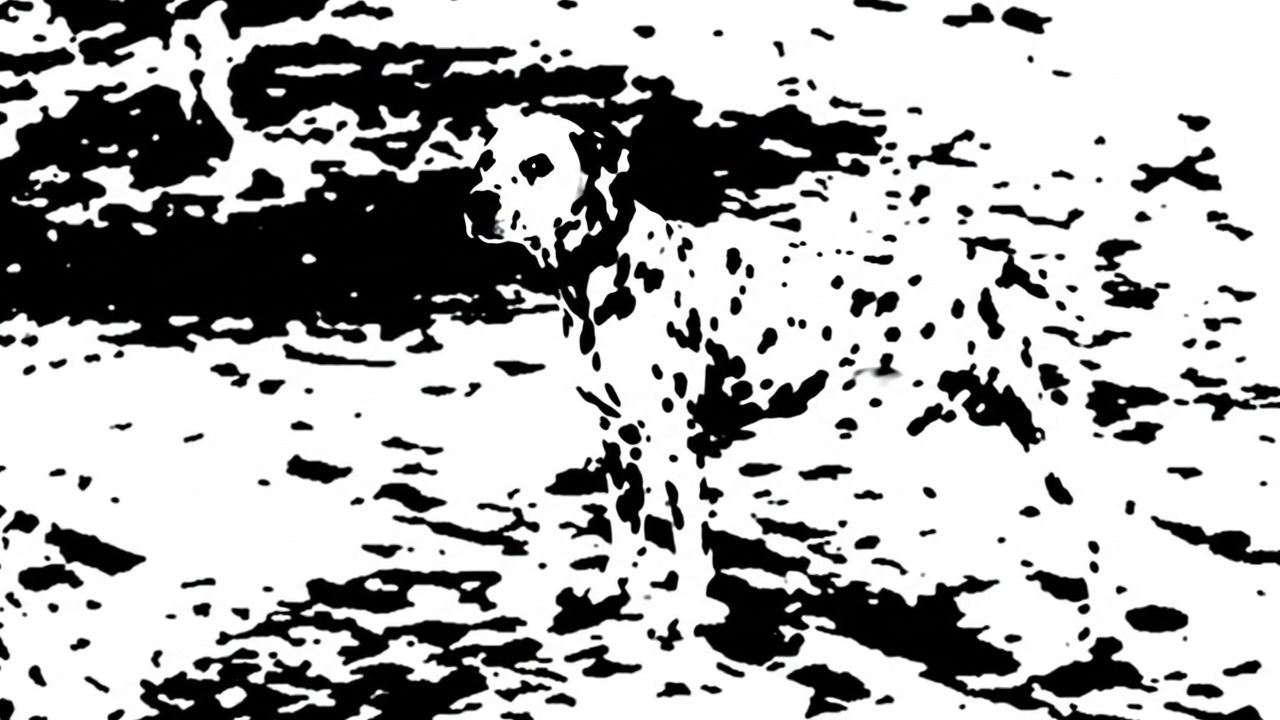}}
    \end{subfigure}
    \caption{\textbf{Dalmatian illusion understanding}. Prompt: \textit{``Static camera perspective.''} Success rate:~1.0. Image credit: \cite{kames1965dalmatian,gregory1970intelligent}.}
    \label{fig:dalmatian}
\end{figure}

\begin{figure}[H]
    \centering
    \begin{subfigure}[t]{\threefigwidth\textwidth}
        \centering
        \fbox{\includegraphics[width=\linewidth]{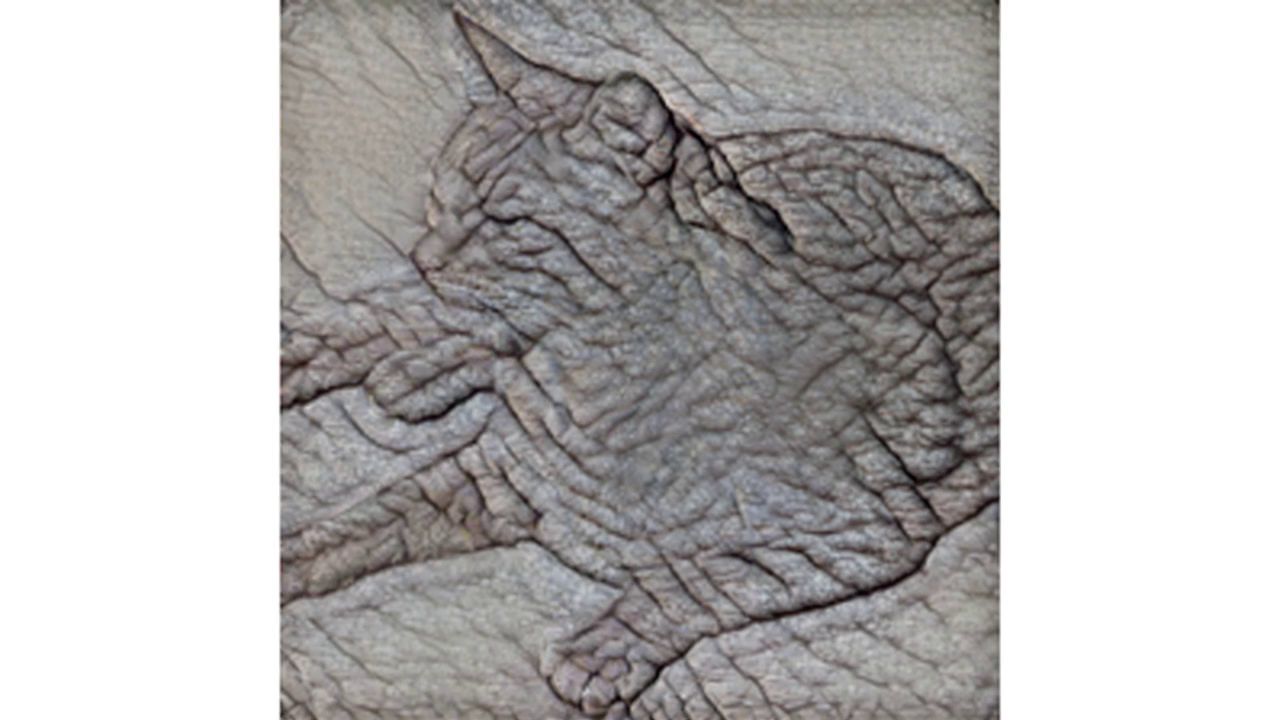}}
    \end{subfigure}
    \hfill
    \begin{subfigure}[t]{\threefigwidth\textwidth}
        \centering
        \fbox{\includegraphics[width=\linewidth]{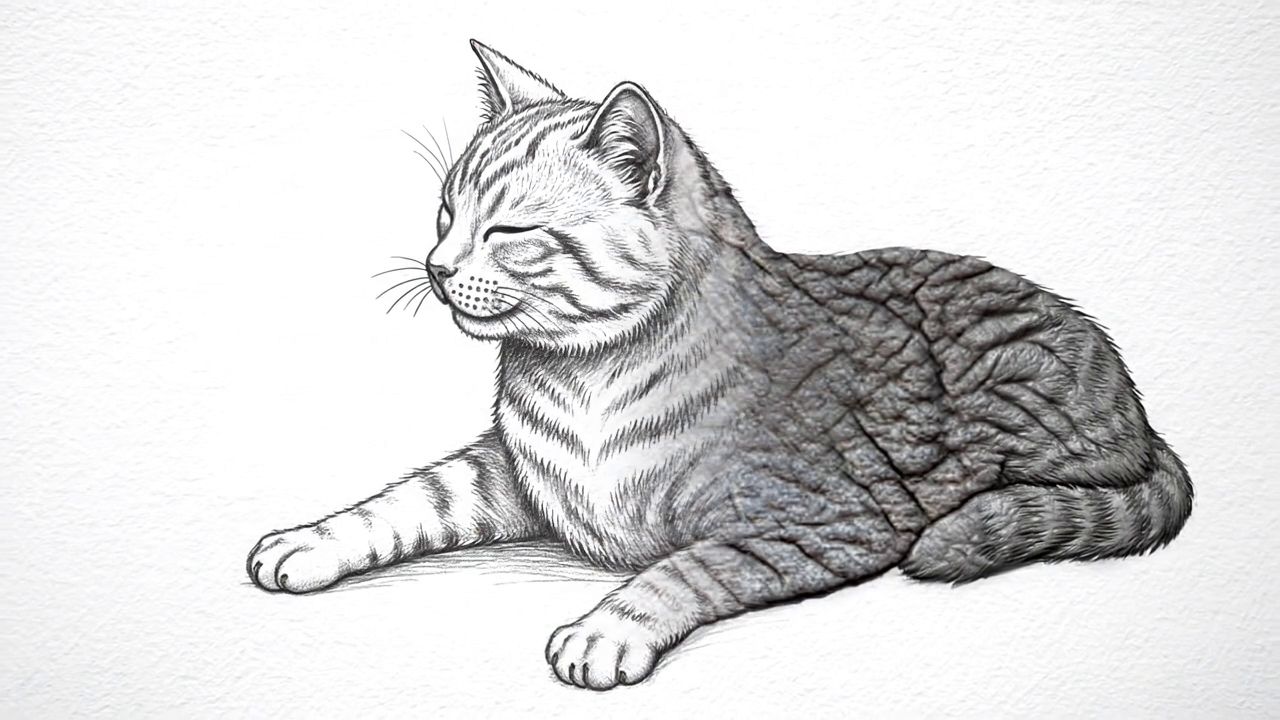}}
    \end{subfigure}
    \hfill    
    \begin{subfigure}[t]{\threefigwidth\textwidth}
        \centering
        \fbox{\includegraphics[width=\linewidth]{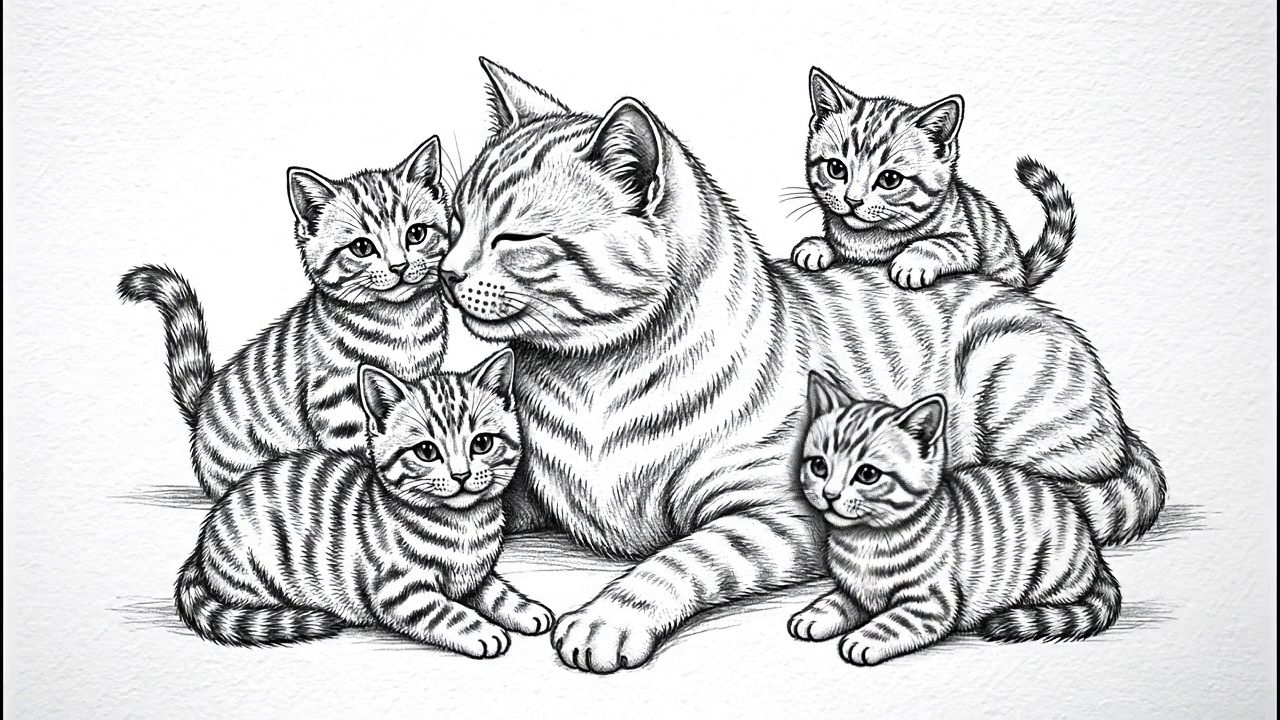}}
    \end{subfigure}
    \caption{\textbf{Shape (cue-conflict) understanding}. Prompt: \textit{``Transform the animal in this image into a sketch of the animal surrounded by its family.''} Success rate:~1.0. Image credit: \cite{geirhos2019imagenet}.}
    \label{fig:texture-shape}
\end{figure}

\begin{figure}[H]
    \centering
    \begin{subfigure}[t]{\threefigwidth\textwidth}
        \centering
        \fbox{\includegraphics[width=\linewidth]{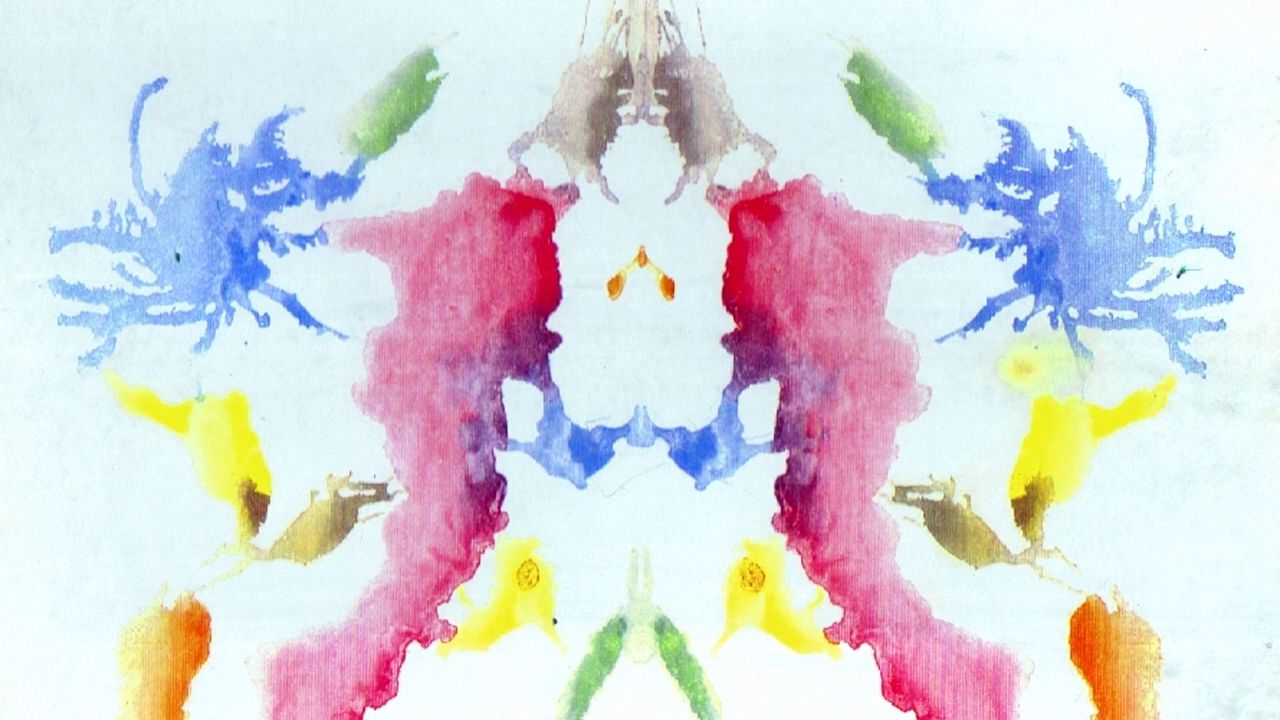}}
    \end{subfigure}
    \hfill
    \begin{subfigure}[t]{\threefigwidth\textwidth}
        \centering
        \fbox{\includegraphics[width=\linewidth]{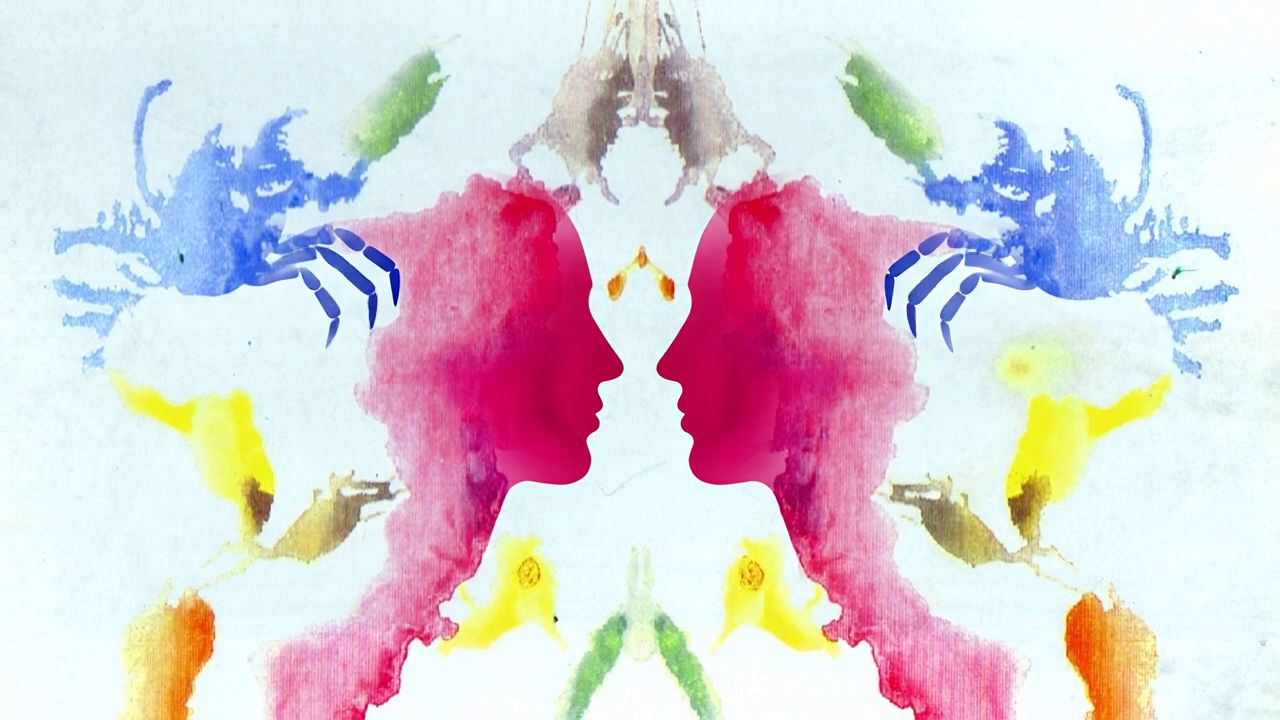}}
    \end{subfigure}
    \hfill    
    \begin{subfigure}[t]{\threefigwidth\textwidth}
        \centering
        \fbox{\includegraphics[width=\linewidth]{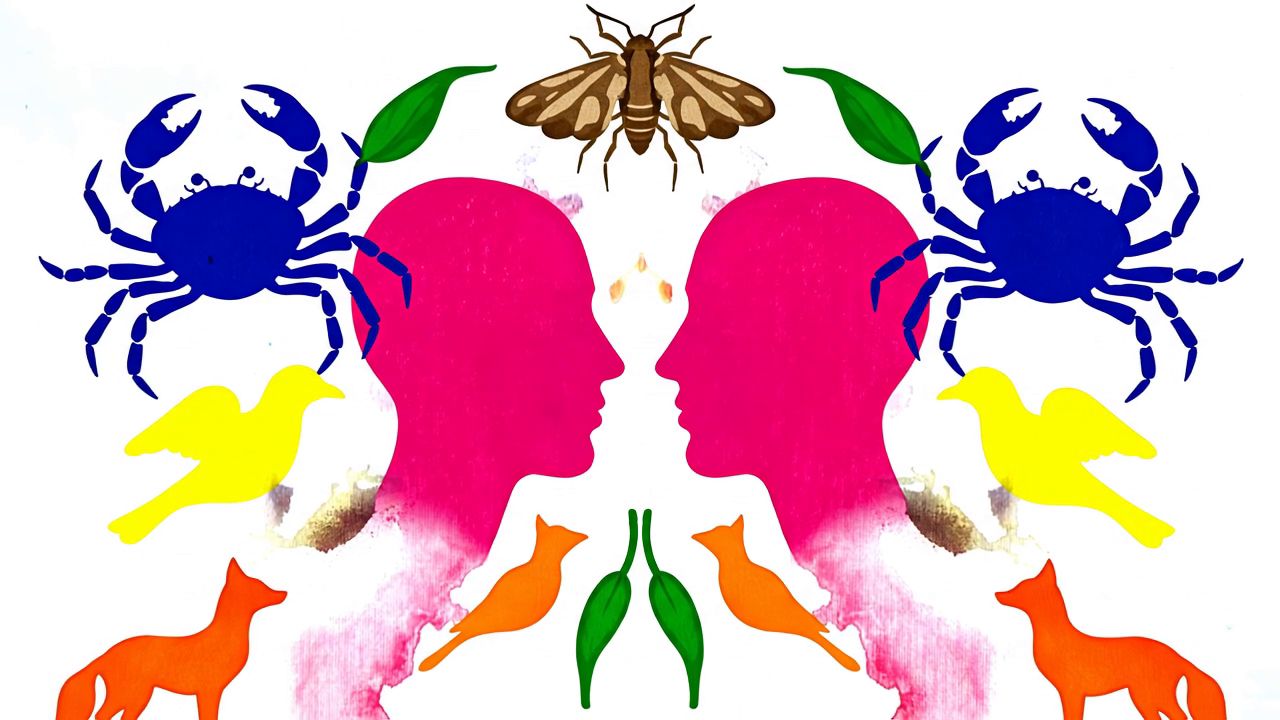}}
    \end{subfigure}
    \caption{\textbf{Rorschach blot interpretation}. Prompt: \textit{``The patterns transform into objects.''} Success rate: undefined (1.0 for prompt following). Image credit: H.\ Rorschach, public domain via \href{https://upload.wikimedia.org/wikipedia/commons/3/32/Rorschach_blot_10.jpg}{wikipedia}.}
    \label{fig:rorschach}
\end{figure}

\subsection{\textcolor{modelingcol}{\textbf{Modeling}}}
\label{app:_modeling_qualitative}

\begin{figure}[H]
    \centering
    \begin{subfigure}[t]{\threefigwidth\textwidth}
        \centering
        \fbox{\includegraphics[width=\linewidth]{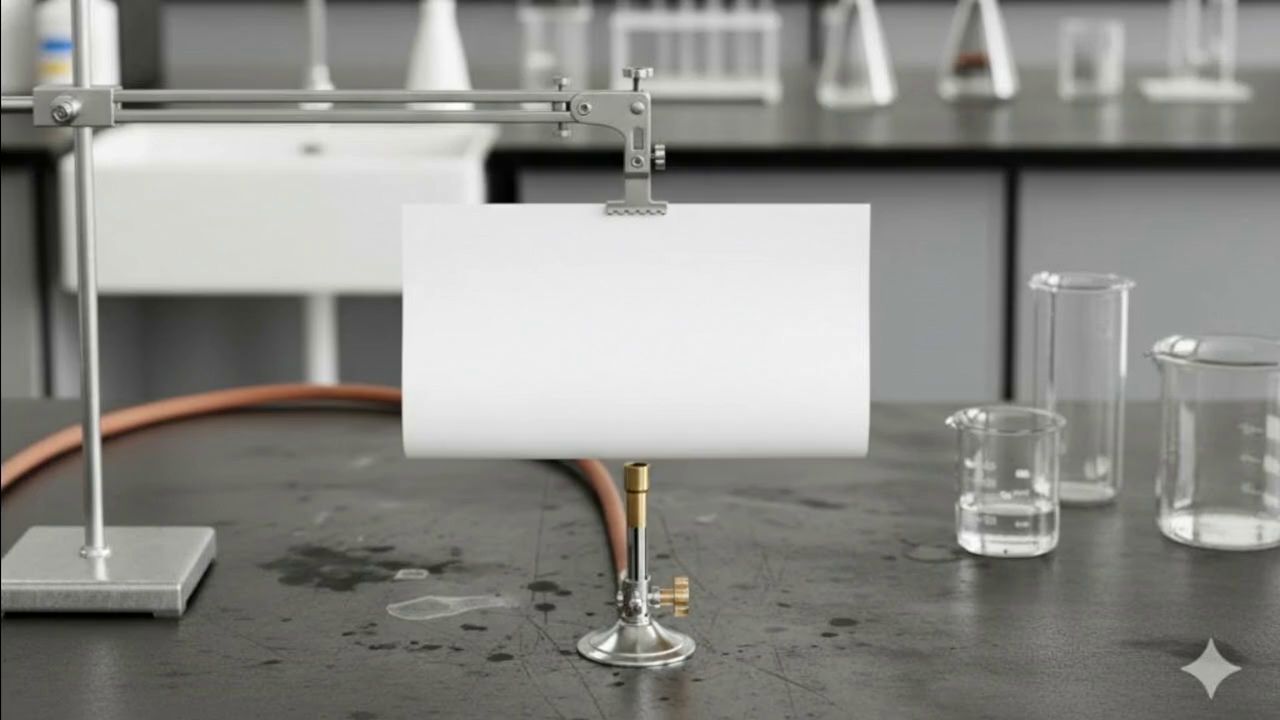}}
    \end{subfigure}
    \hfill
    \begin{subfigure}[t]{\threefigwidth\textwidth}
        \centering
        \fbox{\includegraphics[width=\linewidth]{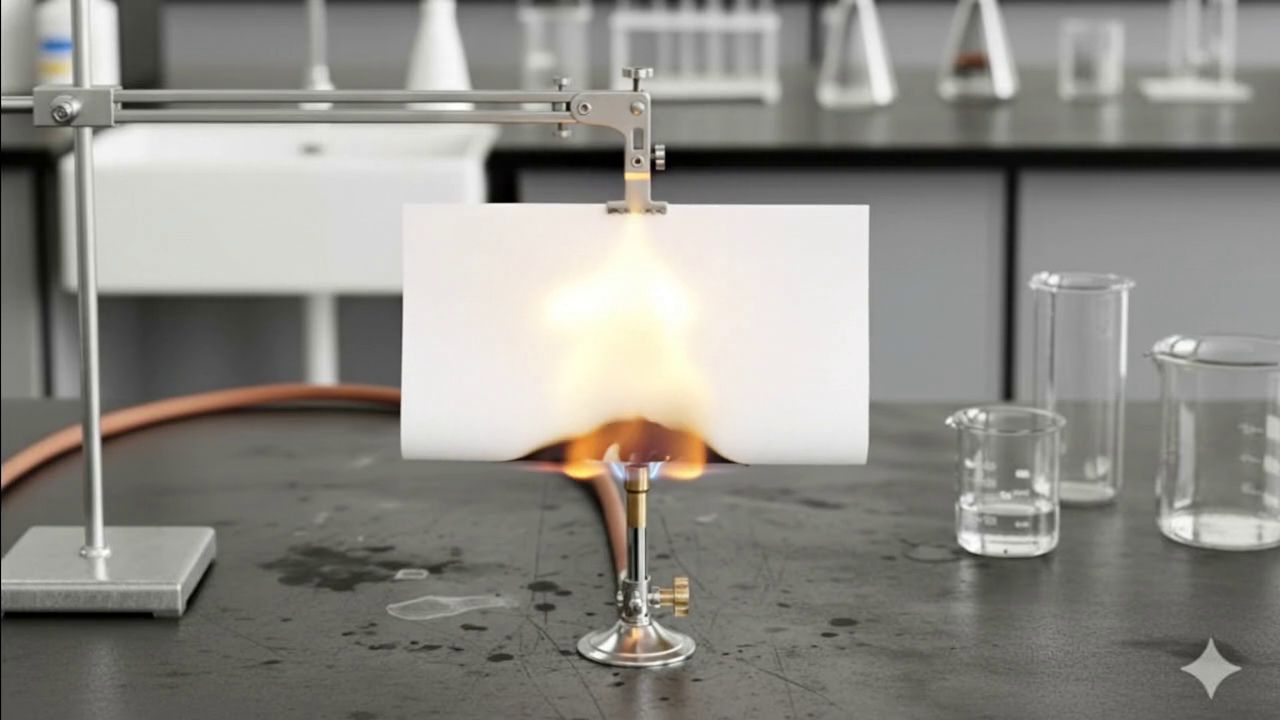}}
    \end{subfigure}
    \hfill    
    \begin{subfigure}[t]{\threefigwidth\textwidth}
        \centering
        \fbox{\includegraphics[width=\linewidth]{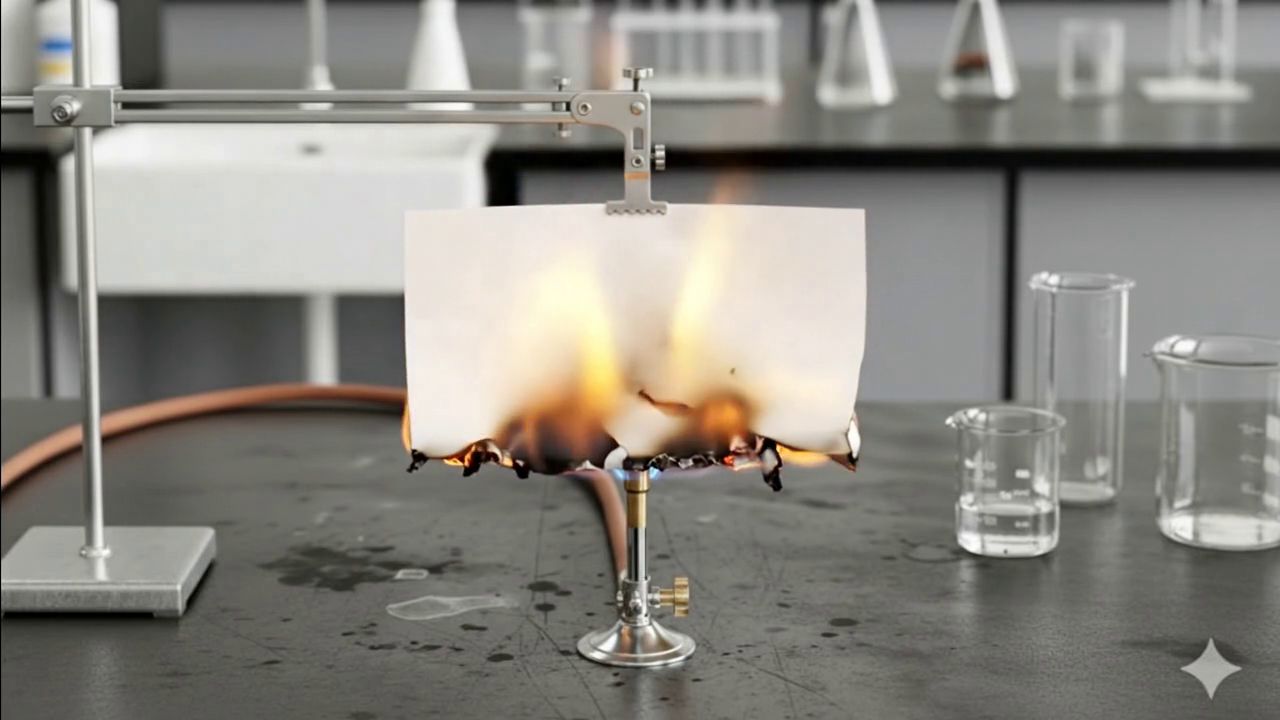}}
    \end{subfigure}
    \caption{\textbf{Material properties.}
    Prompt: \textit{``The bunsen burner at the bottom turns on. Sped up time lapse. Static camera, no pan, no zoom, no dolly.''} Success rate:~0.25.}
    \label{fig:material_properties}
\end{figure}

\begin{figure}[H]
    \centering
    \begin{subfigure}[t]{\threefigwidth\textwidth}
        \centering
        \fbox{\includegraphics[width=\linewidth]{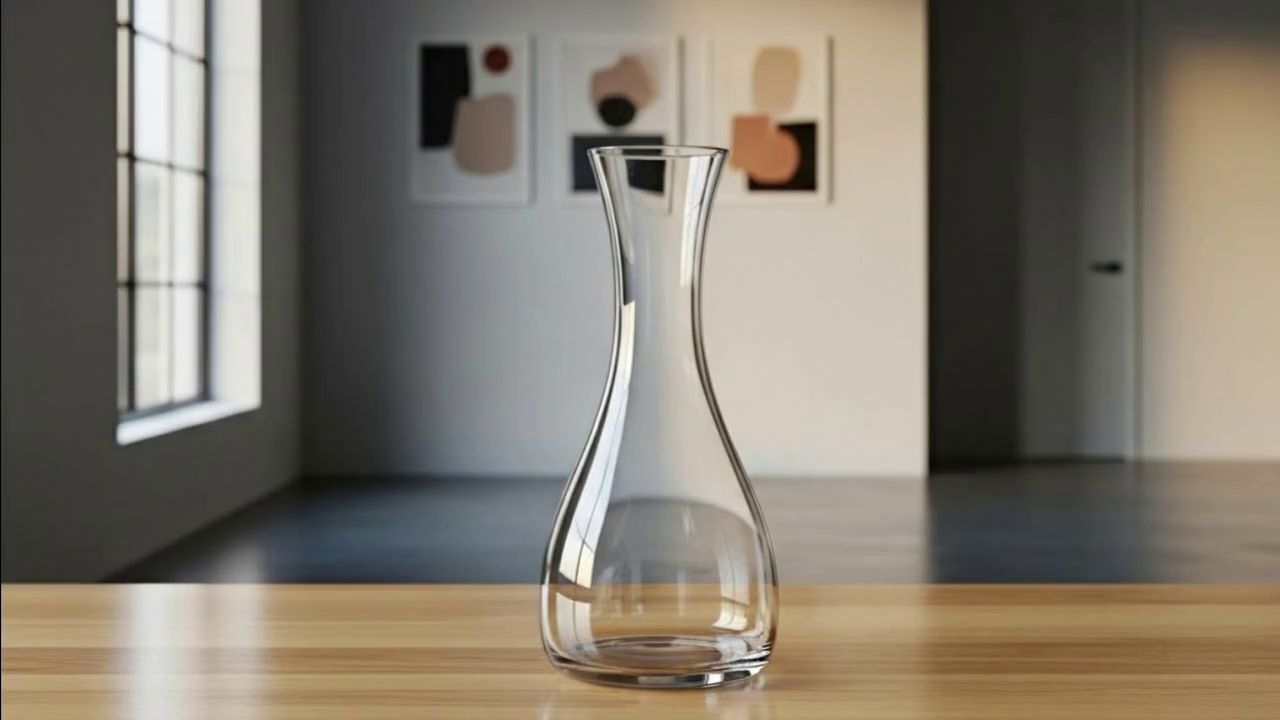}}
    \end{subfigure}
    \hfill
    \begin{subfigure}[t]{\threefigwidth\textwidth}
        \centering
        \fbox{\includegraphics[width=\linewidth]{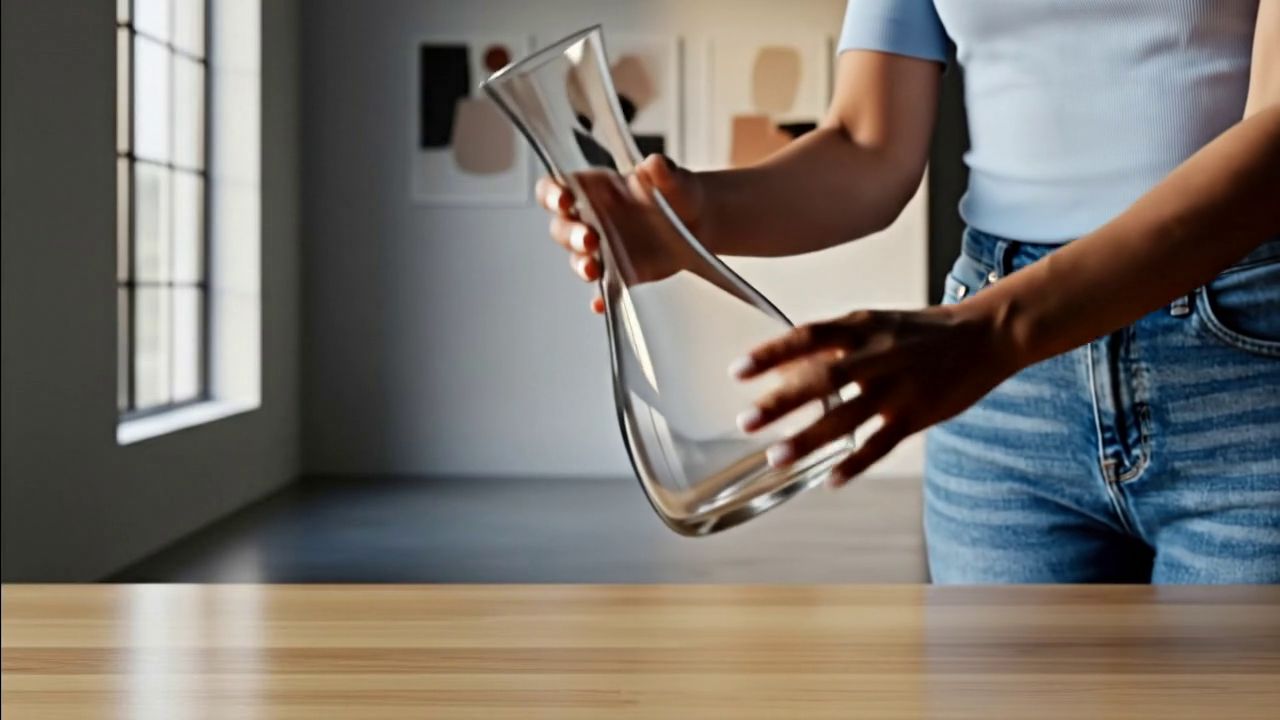}}
    \end{subfigure}
    \hfill    
    \begin{subfigure}[t]{\threefigwidth\textwidth}
        \centering
        \fbox{\includegraphics[width=\linewidth]{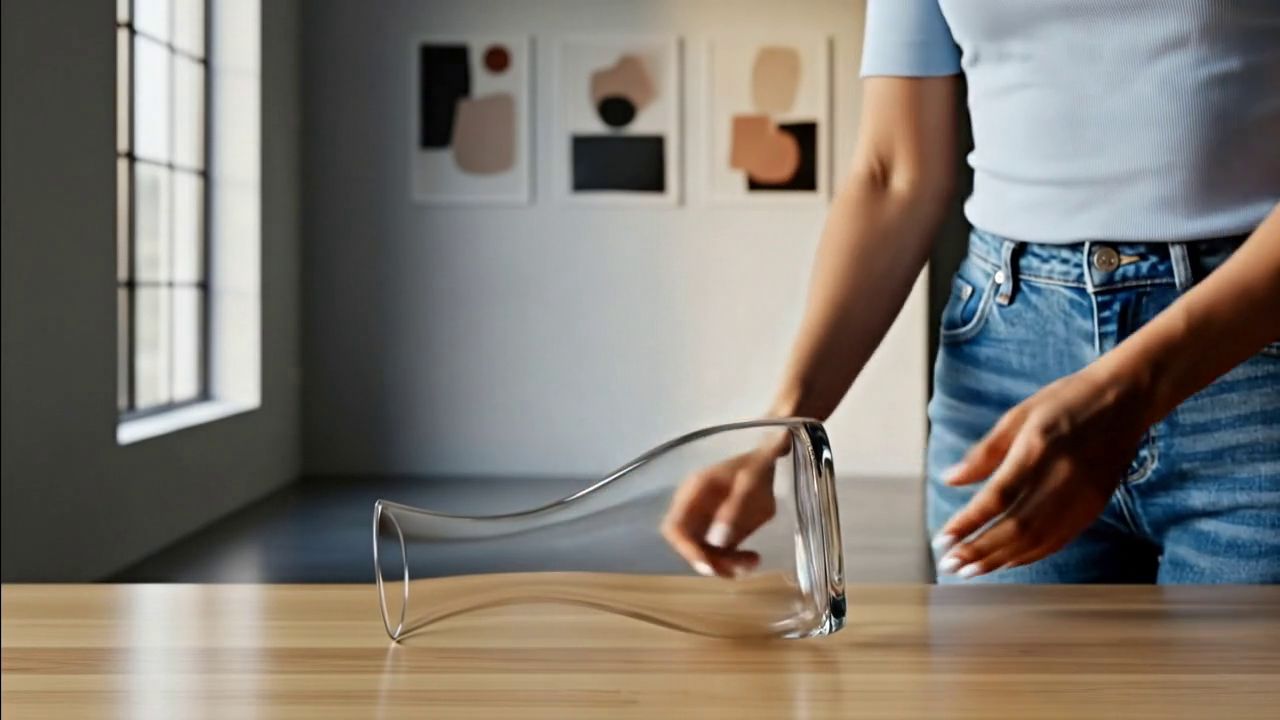}}
    \end{subfigure}

    \vspace{0.2cm}

    \begin{subfigure}[t]{\threefigwidth\textwidth}
        \centering
        \fbox{\includegraphics[width=\linewidth]{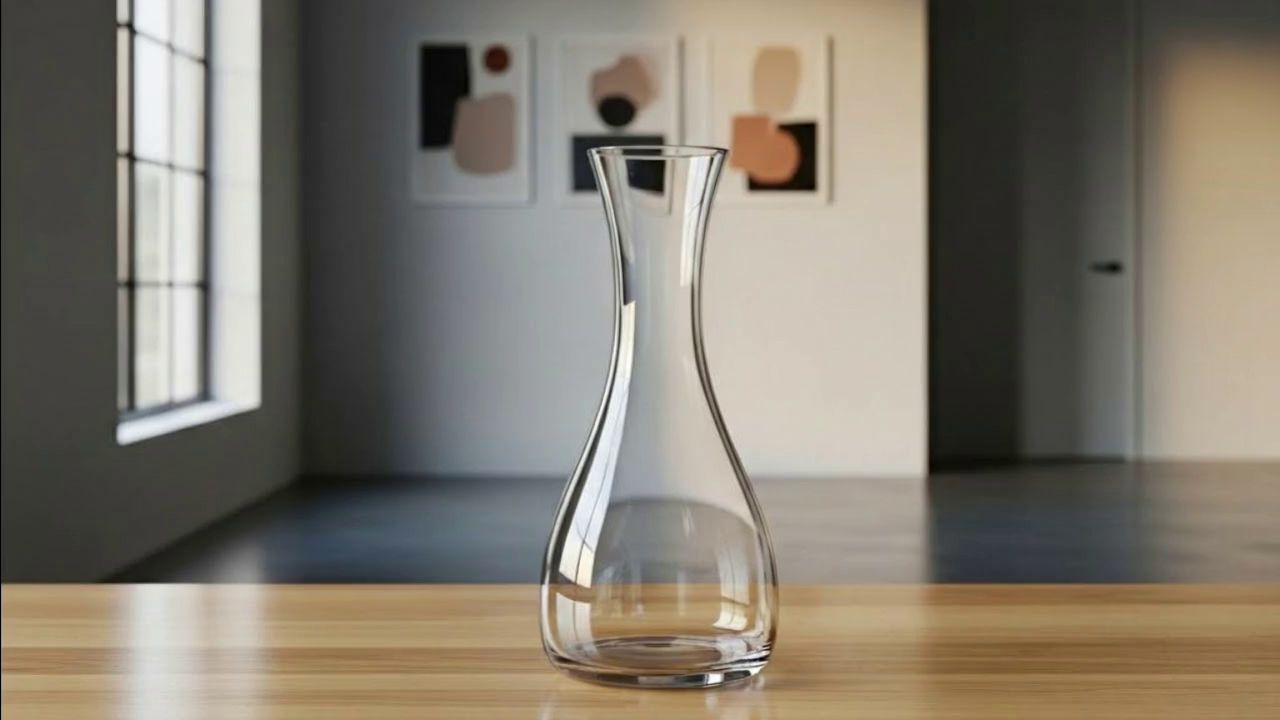}}
    \end{subfigure}
    \hfill
    \begin{subfigure}[t]{\threefigwidth\textwidth}
        \centering
        \fbox{\includegraphics[width=\linewidth]{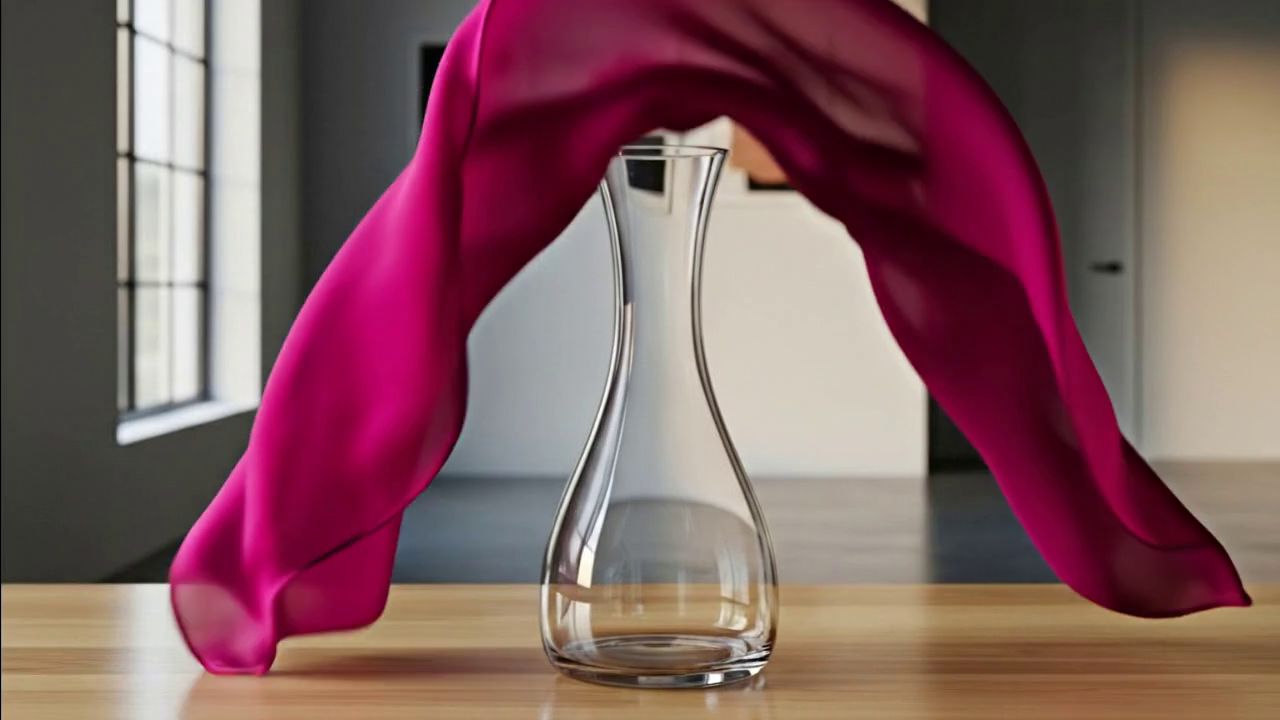}}
    \end{subfigure}
    \hfill    
    \begin{subfigure}[t]{\threefigwidth\textwidth}
        \centering
        \fbox{\includegraphics[width=\linewidth]{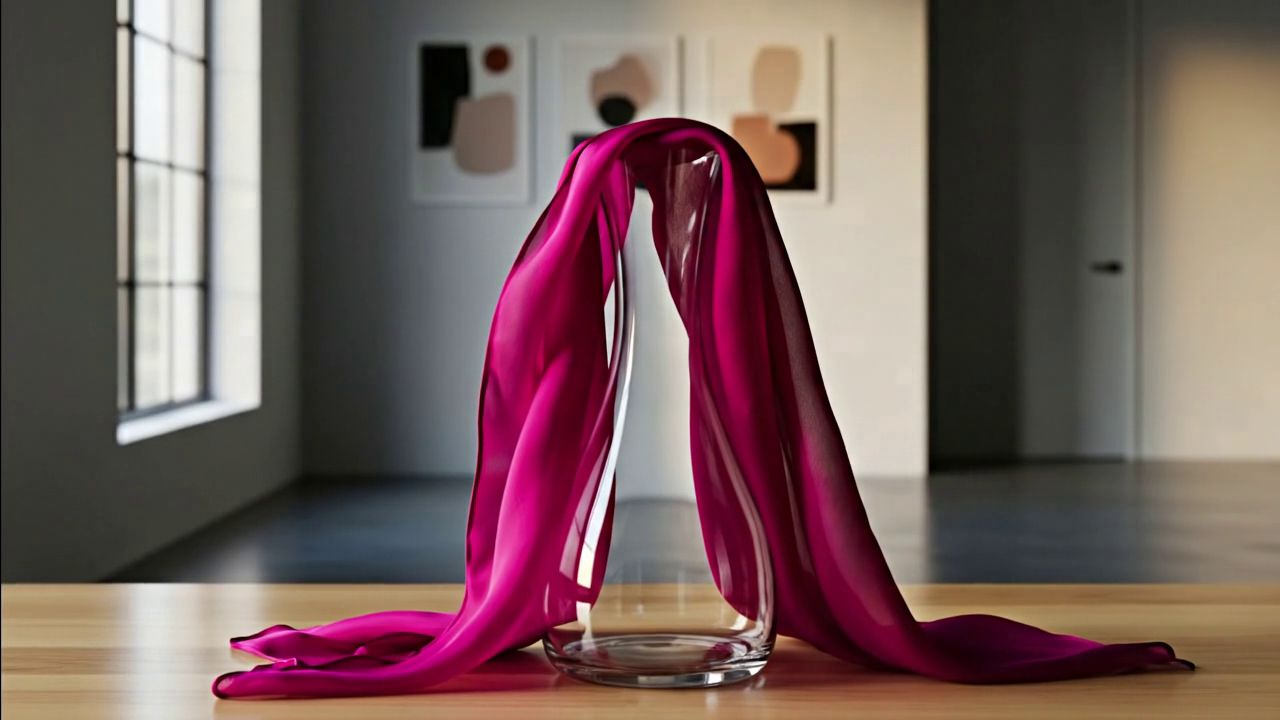}}
    \end{subfigure}
    \caption{\textbf{Physics body transform}.
    \textbf{Rigid body} (top). Prompt: \textit{``A person picks up the vase and puts it back on the table in a sideways orientation. Static camera, no pan, no zoom, no dolly.''} Success rate:~1.0.
    \textbf{Soft body} (bottom). Prompt: \textit{``A person drapes a thin silk scarf over the vase. Static camera, no pan, no zoom, no dolly.''} Success rate:~0.67.}
    \label{fig:physics-body}
\end{figure}

\begin{figure}[H]
    \centering
    \begin{subfigure}[t]{\threefigwidth\textwidth}
        \centering
        \fbox{\includegraphics[width=\linewidth]{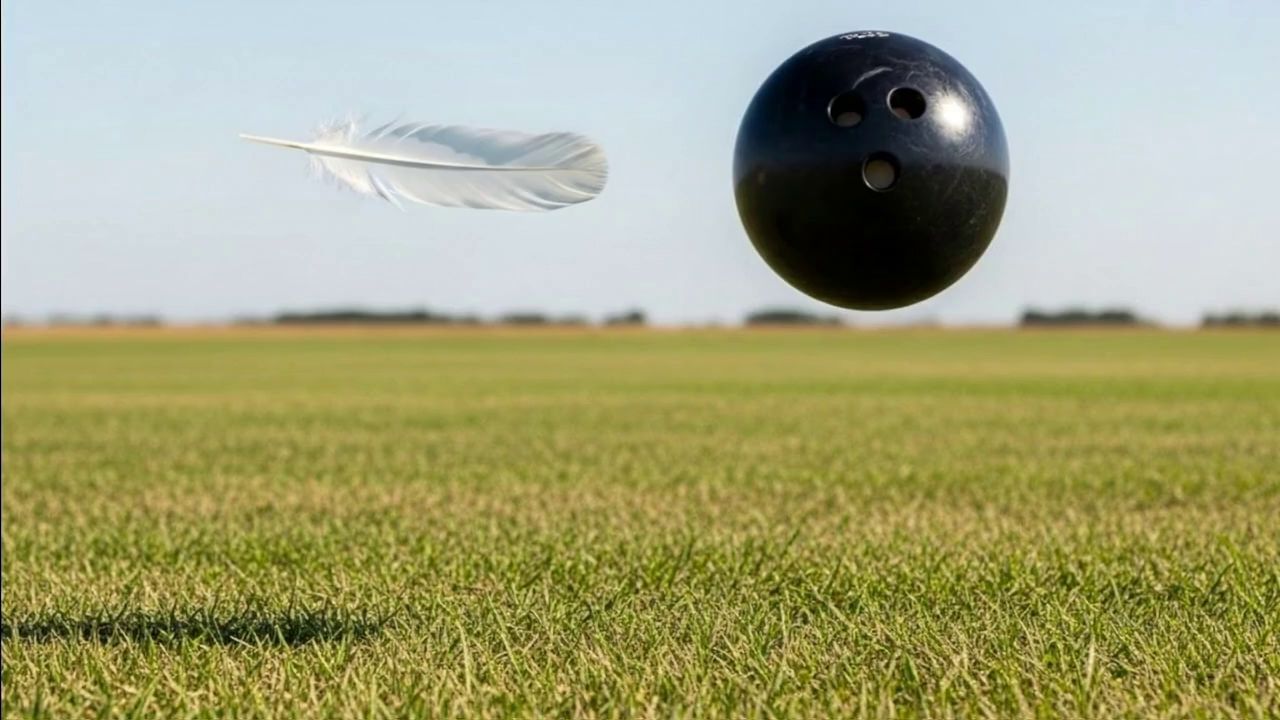}}
    \end{subfigure}
    \hfill
    \begin{subfigure}[t]{\threefigwidth\textwidth}
        \centering
        \fbox{\includegraphics[width=\linewidth]{figures-compressed/gravity_air_resistance_Objects_Falling_Earth_Gravity_frame060.jpg}}
    \end{subfigure}
    \hfill
    \begin{subfigure}[t]{\threefigwidth\textwidth}
        \centering
        \fbox{\includegraphics[width=\linewidth]{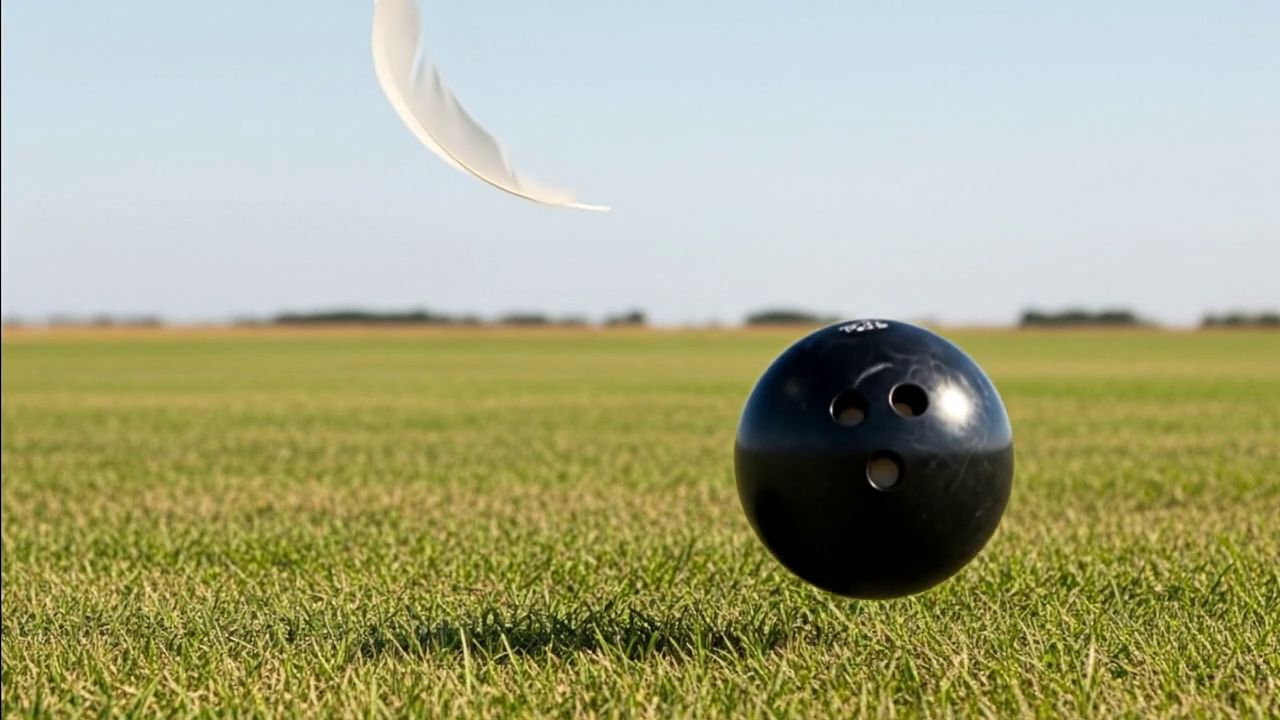}}
    \end{subfigure}
    
    \vspace{0.2cm}
    
    \begin{subfigure}[t]{\threefigwidth\textwidth}
        \centering
        \fbox{\includegraphics[width=\linewidth]{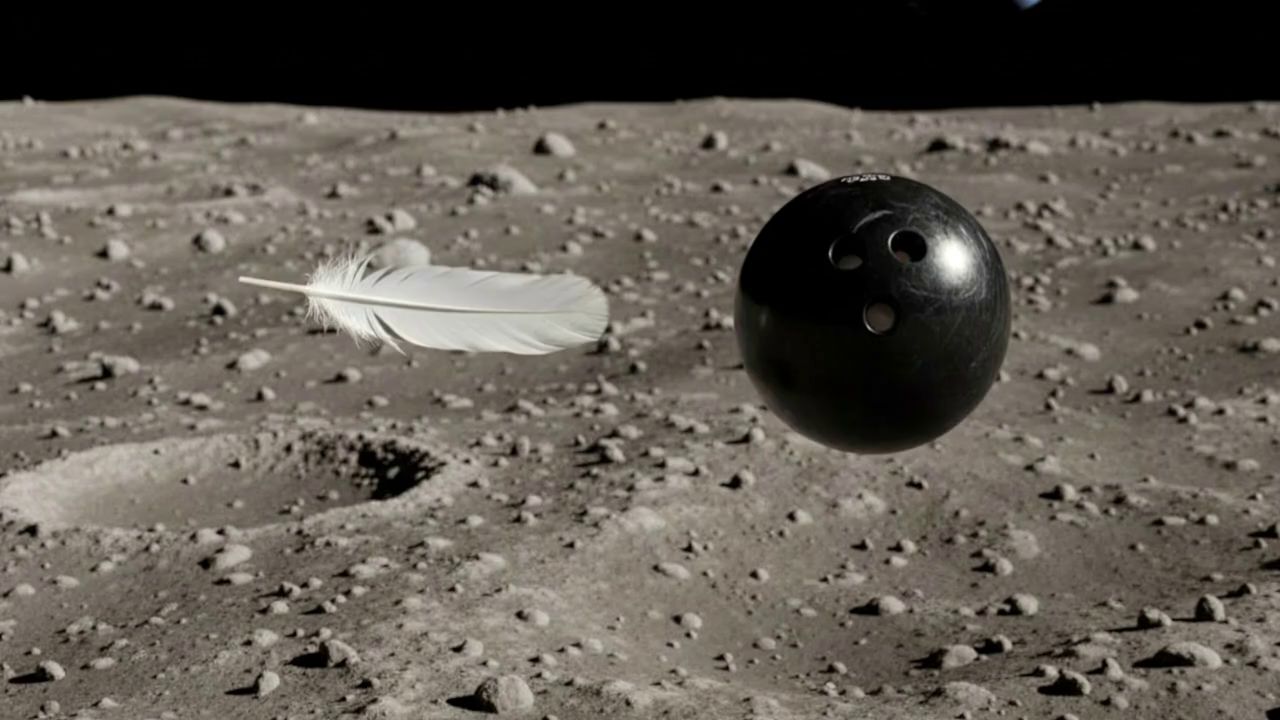}}
    \end{subfigure}
    \hfill
    \begin{subfigure}[t]{\threefigwidth\textwidth}
        \centering
        \fbox{\includegraphics[width=\linewidth]{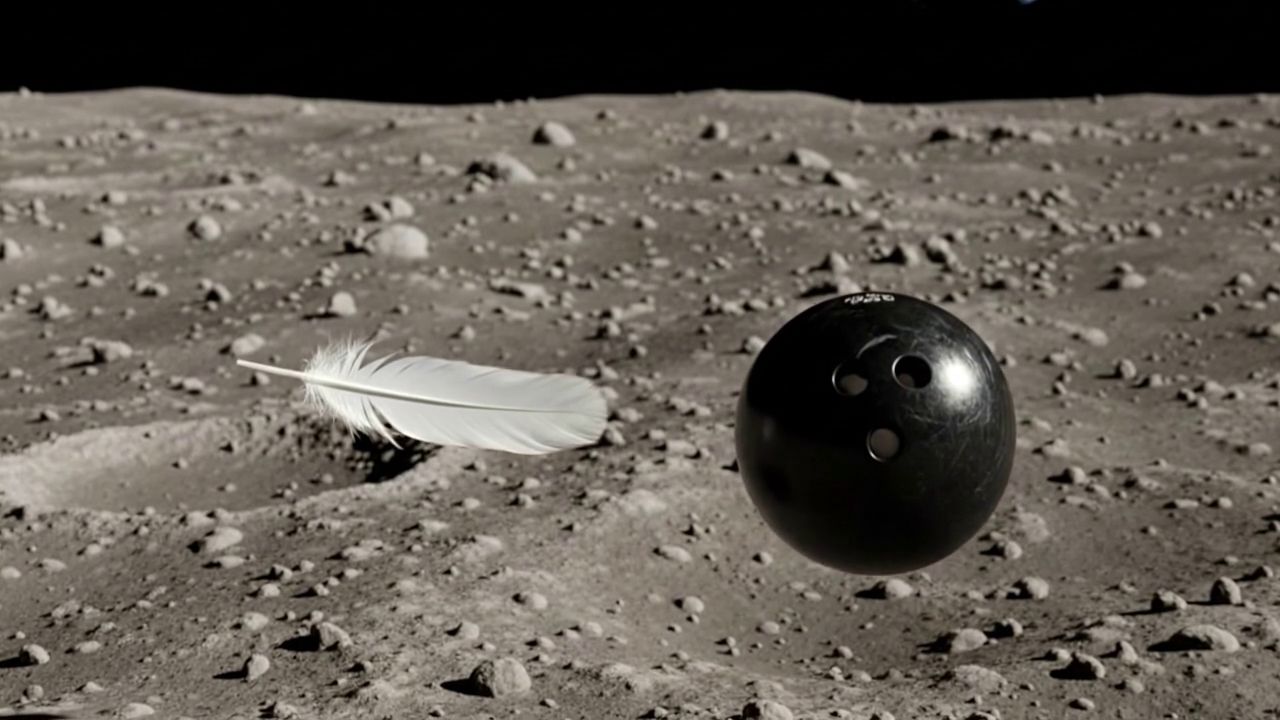}}
    \end{subfigure}
    \hfill
    \begin{subfigure}[t]{\threefigwidth\textwidth}
        \centering
        \fbox{\includegraphics[width=\linewidth]{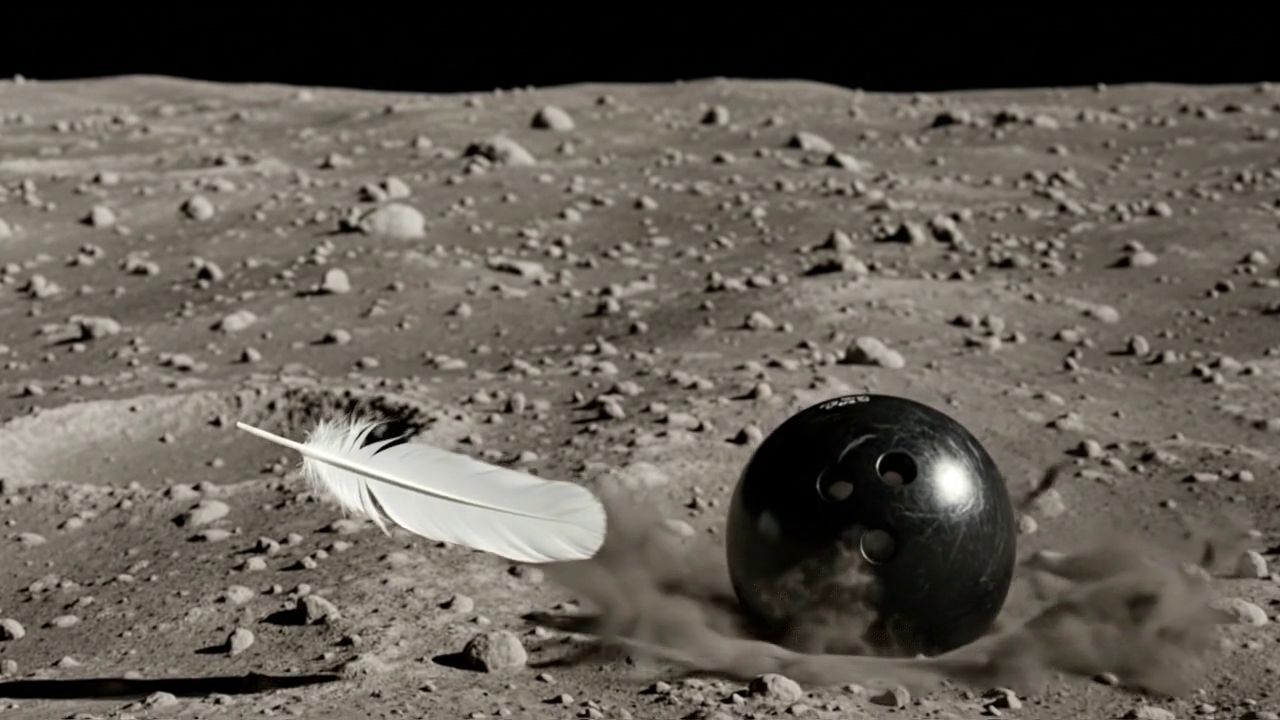}}
    \end{subfigure}
    \caption{\textbf{Gravity and air resistance}.
    \textbf{On earth} (top). Prompt: \textit{``The objects fall due to gravity. Static camera, no pan, no zoom, no dolly.''} Success rate: 0.5.
    \textbf{On the moon} (bottom). Prompt: \textit{``The objects fall down on the moon due to gravity. Static camera, no pan, no zoom, no dolly.''} Success rate: 0.5.}
    \label{fig:gravity_air_resistance}
\end{figure}

\begin{figure}[H]
    \centering
    \begin{subfigure}[t]{\threefigwidth\textwidth}
        \centering
        \fbox{\includegraphics[width=\linewidth]{figures-compressed/water_floating_Water_Bottlecap_frame000.jpg}}
    \end{subfigure}
    \hfill
    \begin{subfigure}[t]{\threefigwidth\textwidth}
        \centering
        \fbox{\includegraphics[width=\linewidth]{figures-compressed/water_floating_Water_Bottlecap_frame112.jpg}}
    \end{subfigure}
    \hfill
    \begin{subfigure}[t]{\threefigwidth\textwidth}
        \centering
        \fbox{\includegraphics[width=\linewidth]{figures-compressed/water_floating_Water_Bottlecap_frame191.jpg}}
    \end{subfigure}

    \vspace{0.2cm}
    
    \begin{subfigure}[t]{\threefigwidth\textwidth}
        \centering
        \fbox{\includegraphics[width=\linewidth]{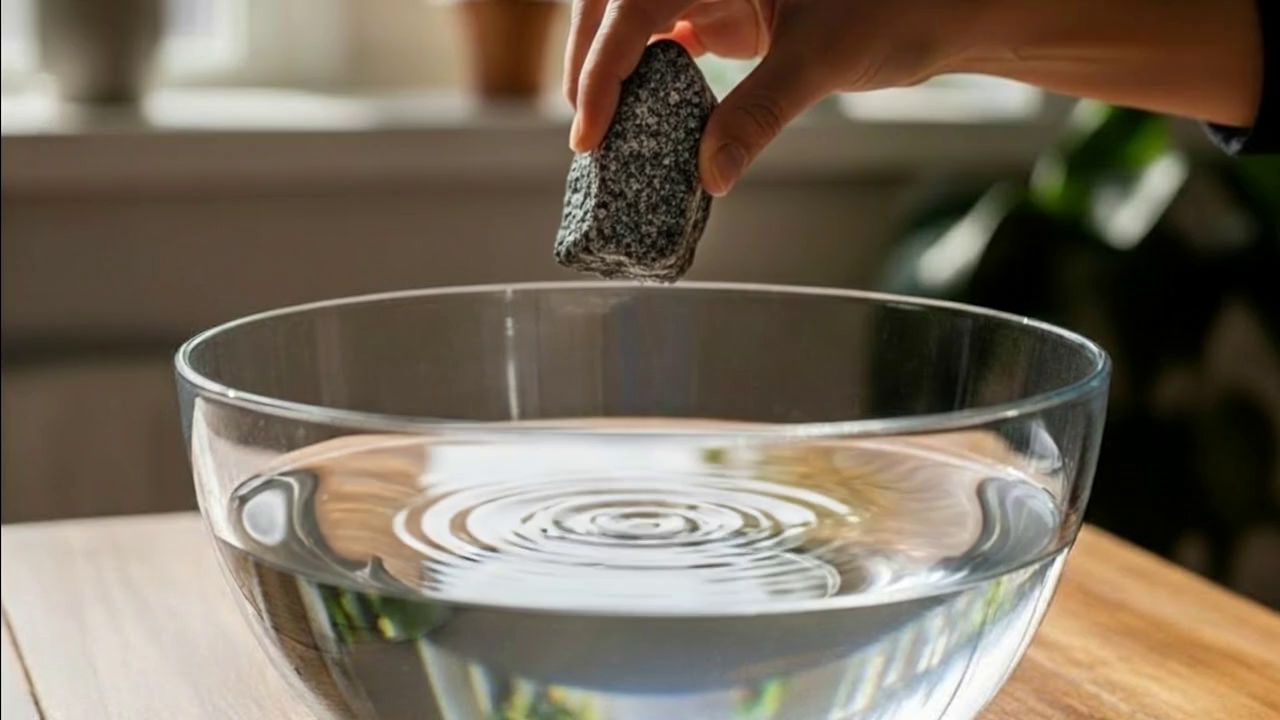}}
    \end{subfigure}
    \hfill
    \begin{subfigure}[t]{\threefigwidth\textwidth}
        \centering
        \fbox{\includegraphics[width=\linewidth]{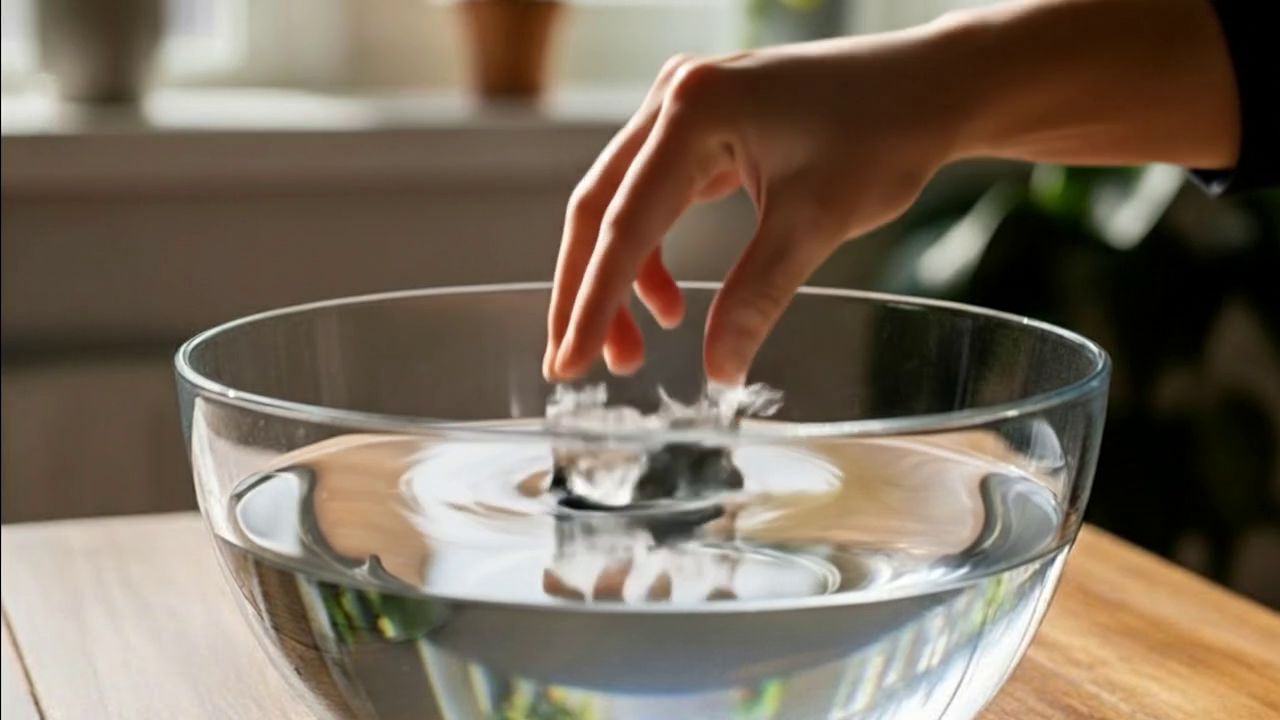}}
    \end{subfigure}
    \hfill
    \begin{subfigure}[t]{\threefigwidth\textwidth}
        \centering
        \fbox{\includegraphics[width=\linewidth]{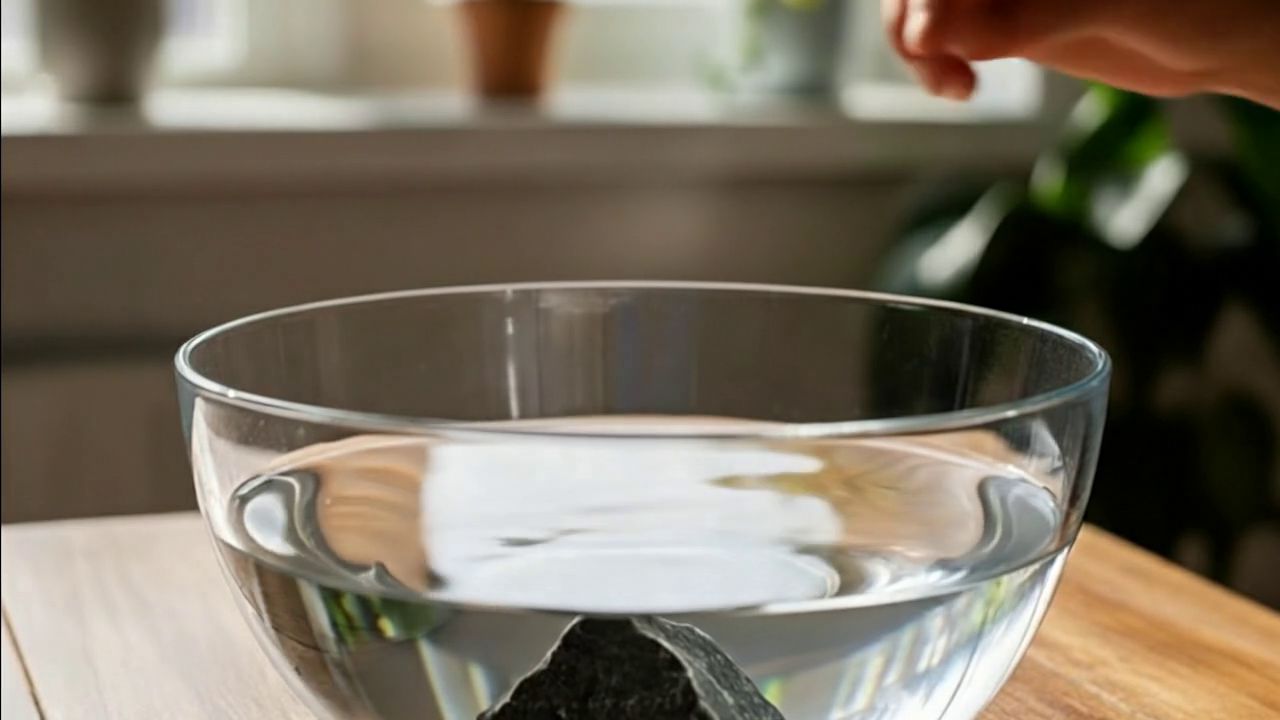}}
    \end{subfigure}
    \caption{\textbf{Buoyancy.}
    Prompt: \textit{``The hand lets go of the object. Static camera, no pan, no zoom, no dolly.''} Success rate (bottle cap): 0.58; success rate (rock): 0.83.}
    \label{fig:water_floating}
\end{figure}

\begin{figure}[H]
    \centering
    \begin{subfigure}[t]{0.195\textwidth}
        \centering
        \fbox{\includegraphics[width=\linewidth]{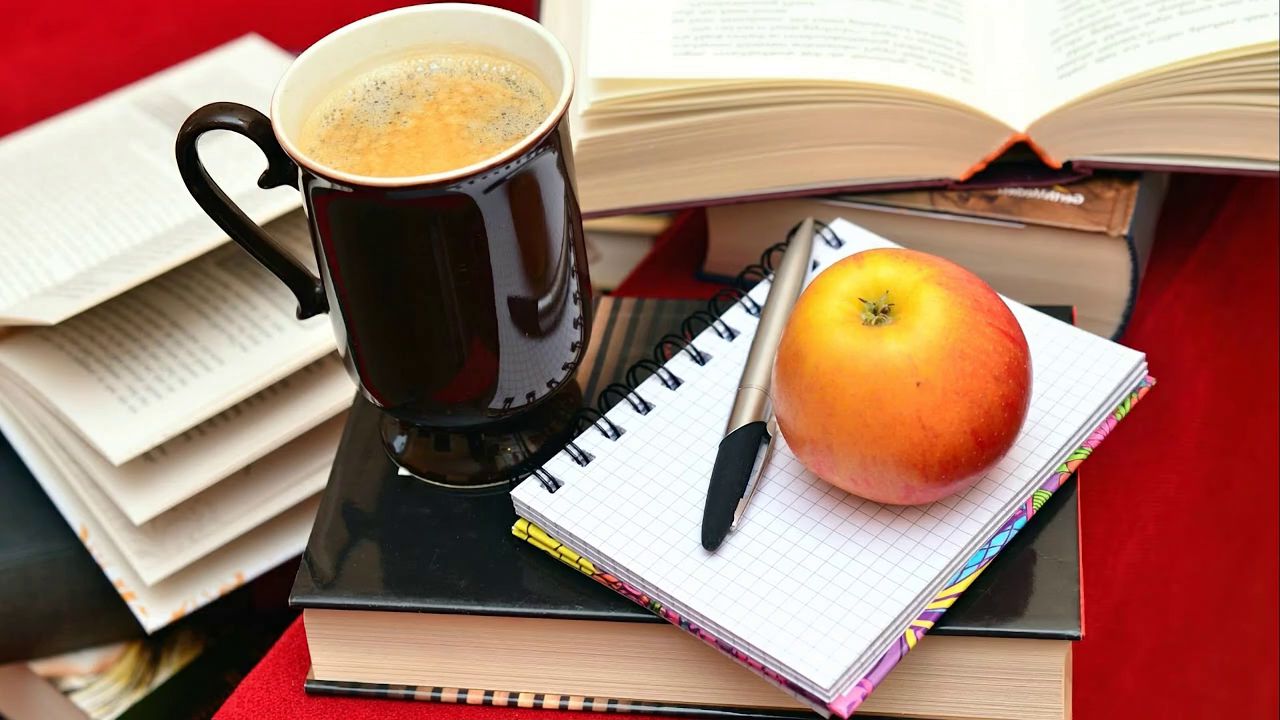}}
    \end{subfigure}
    \hfill
    \begin{subfigure}[t]{0.195\textwidth}
        \centering
        \fbox{\includegraphics[width=\linewidth]{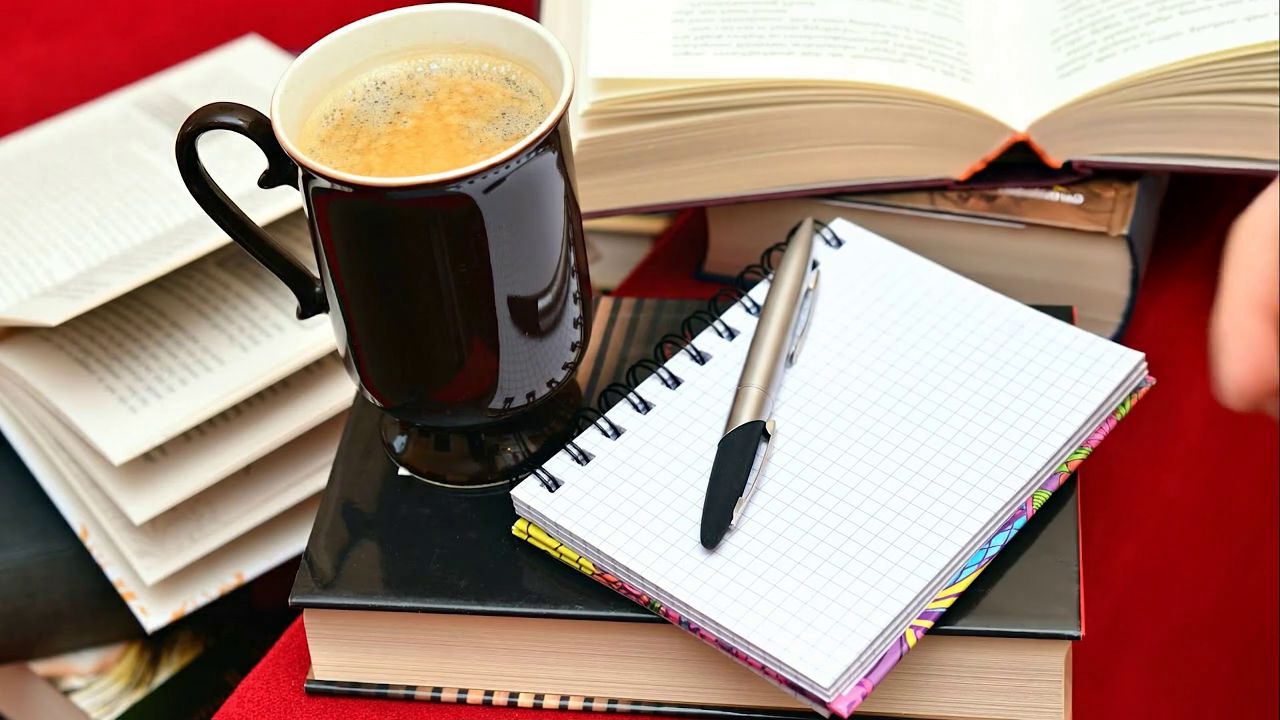}}
    \end{subfigure}
    \hfill
    \begin{subfigure}[t]{0.195\textwidth}
        \centering
        \fbox{\includegraphics[width=\linewidth]{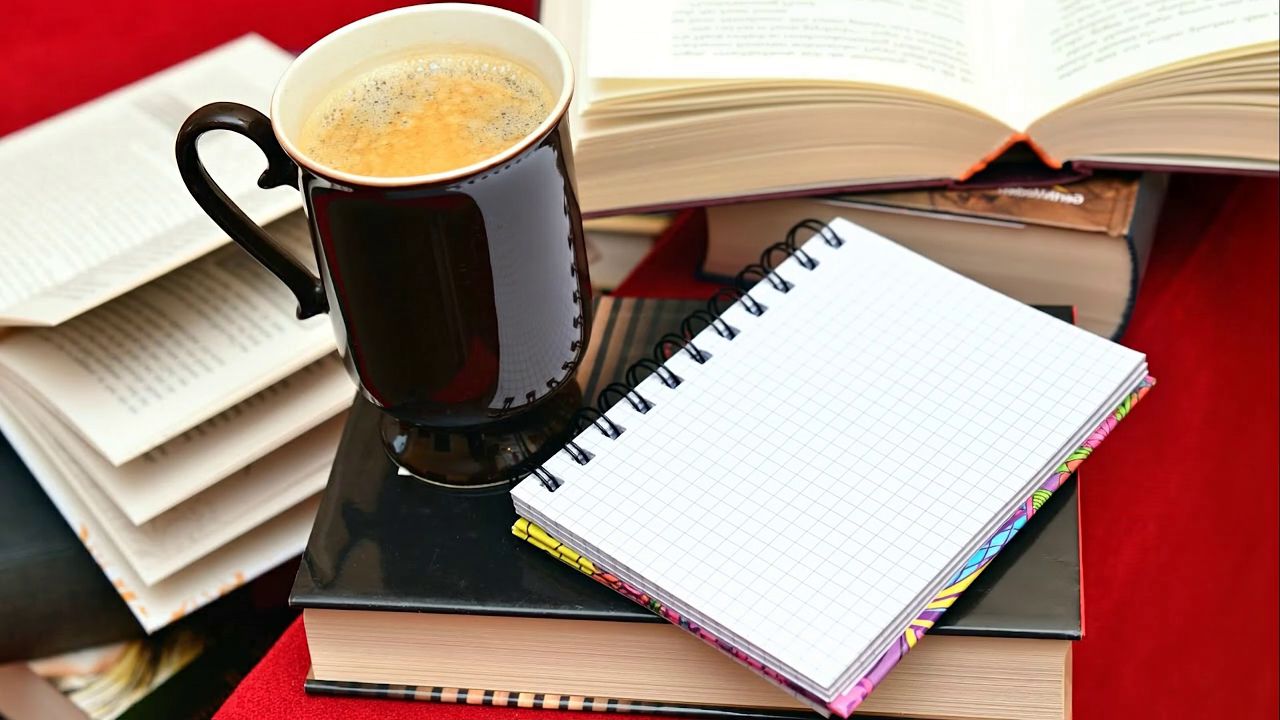}}
    \end{subfigure}
    \begin{subfigure}[t]{0.195\textwidth}
        \centering
        \fbox{\includegraphics[width=\linewidth]{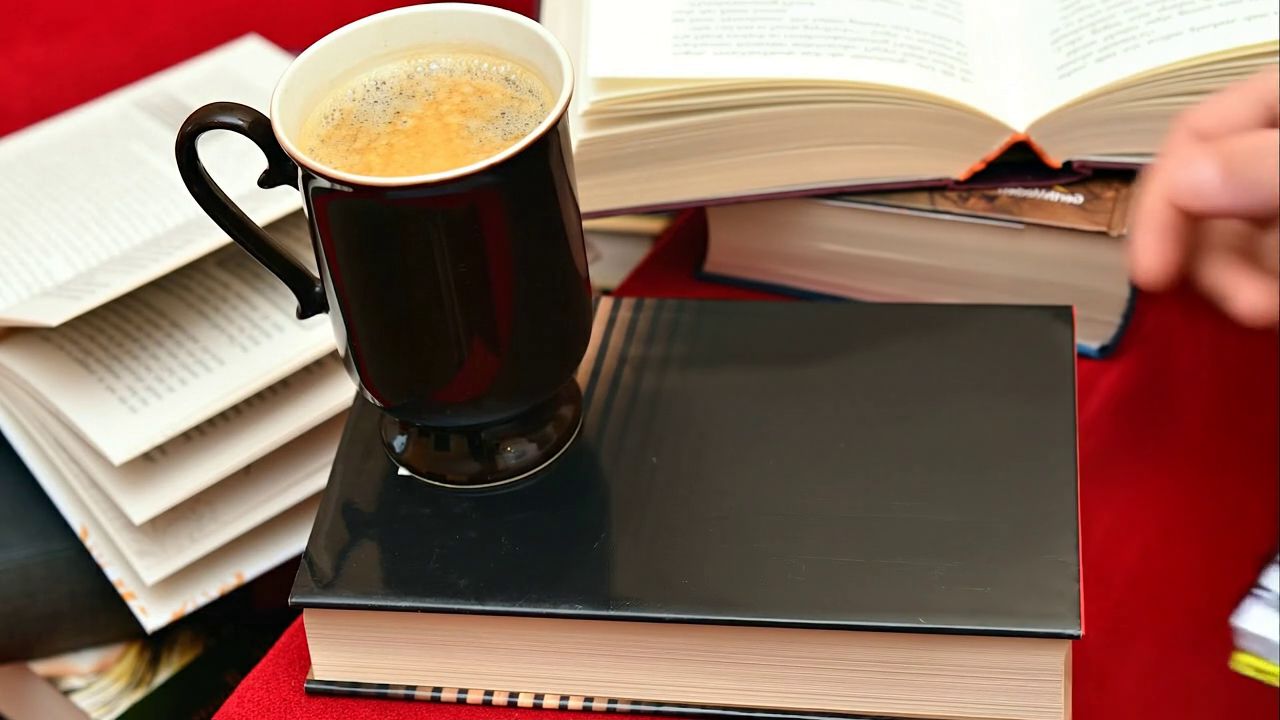}}
    \end{subfigure}
    \begin{subfigure}[t]{0.195\textwidth}
        \centering
        \fbox{\includegraphics[width=\linewidth]{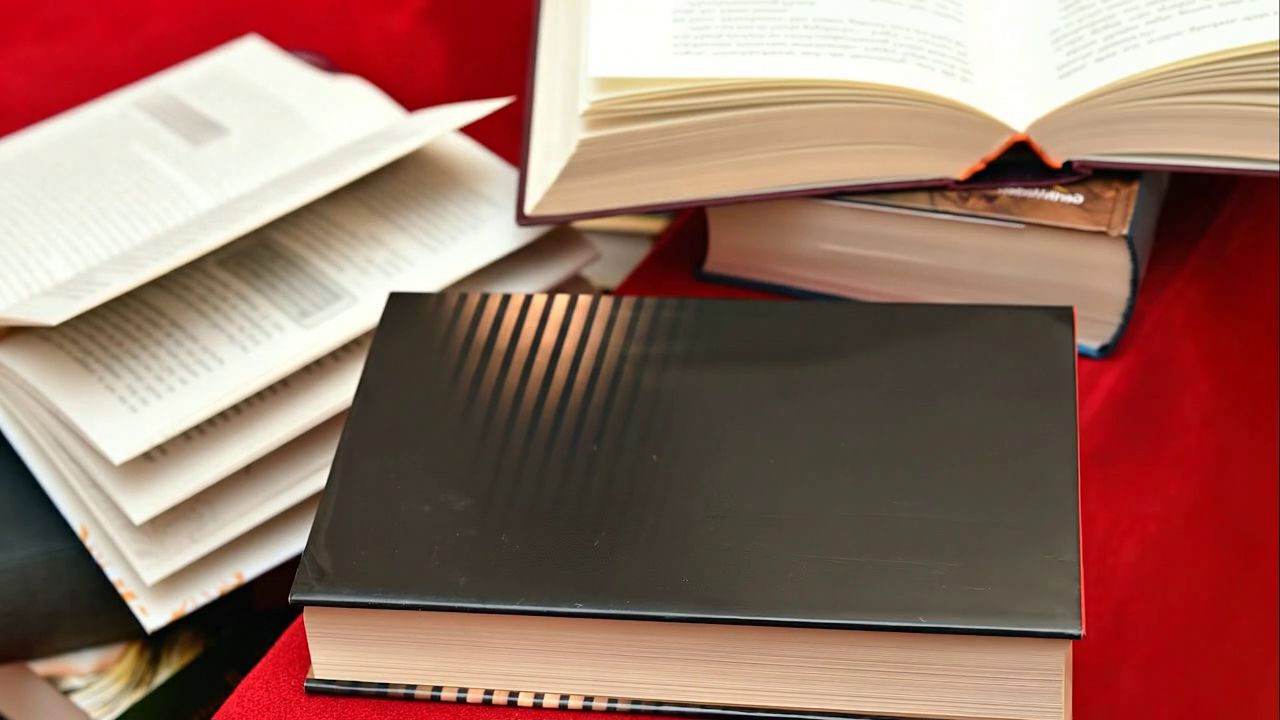}}
    \end{subfigure}
    \caption{\textbf{Visual Jenga}, inspired by~\cite{bhattad2025visual}.
    Prompt: \textit{``A hand quickly removes each of the items in this image, one at a time.''} Success rate, based on removal of at least three objects:~0.5.}
    \label{fig:visual-jenga}
\end{figure}

\begin{figure}[H]
    \centering
    \begin{subfigure}[t]{\threefigwidth\textwidth}
        \centering
        \fbox{\includegraphics[width=\linewidth]{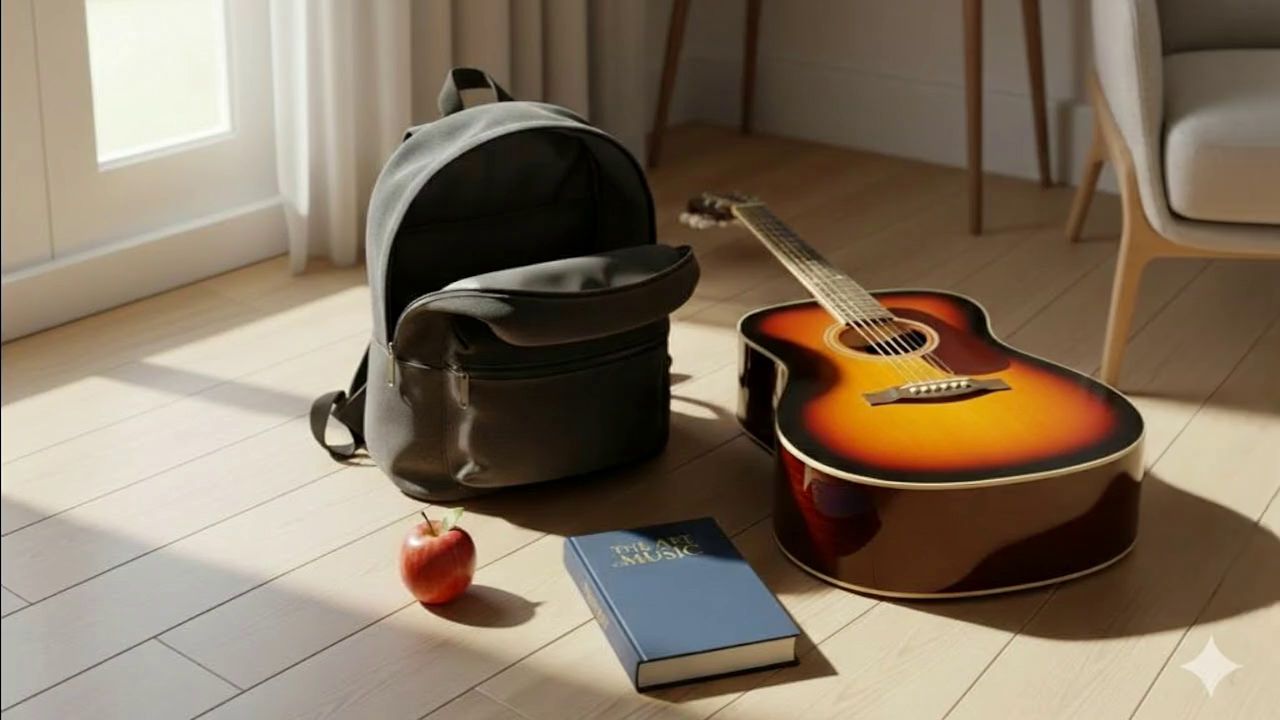}}
    \end{subfigure}
    \hfill
    \begin{subfigure}[t]{\threefigwidth\textwidth}
        \centering
        \fbox{\includegraphics[width=\linewidth]{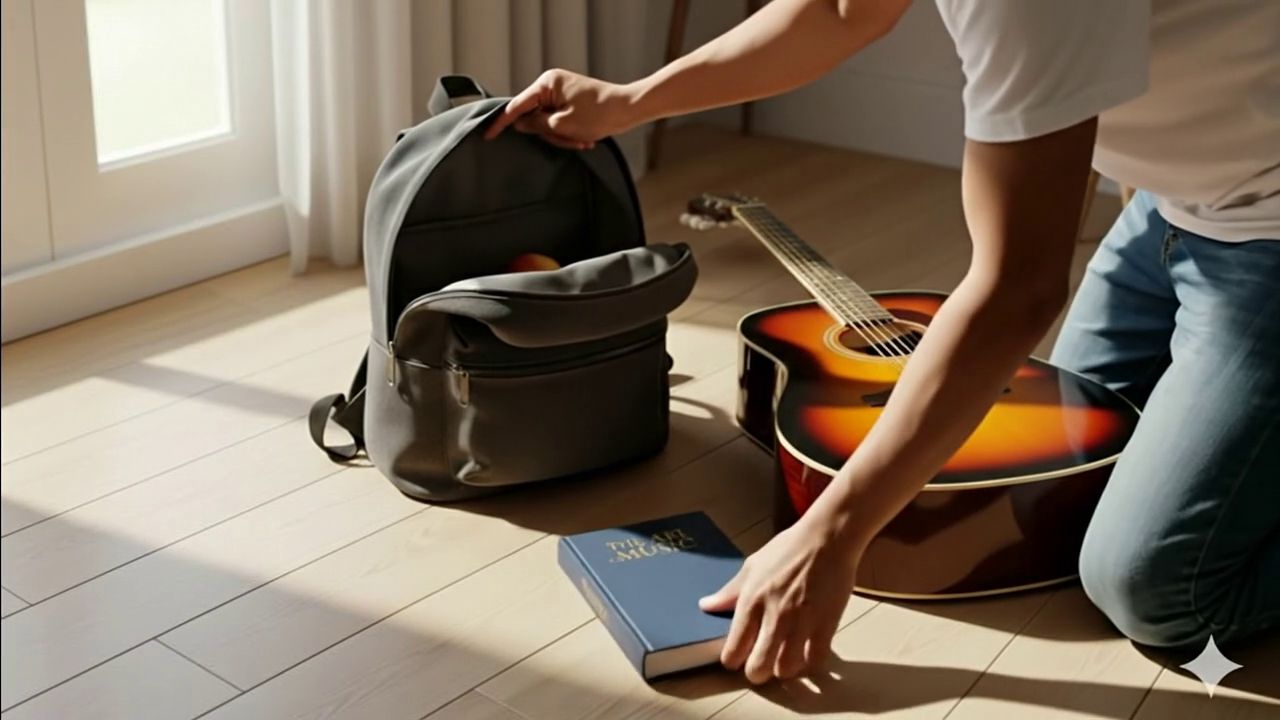}}
    \end{subfigure}
    \hfill
    \begin{subfigure}[t]{\threefigwidth\textwidth}
        \centering
        \fbox{\includegraphics[width=\linewidth]{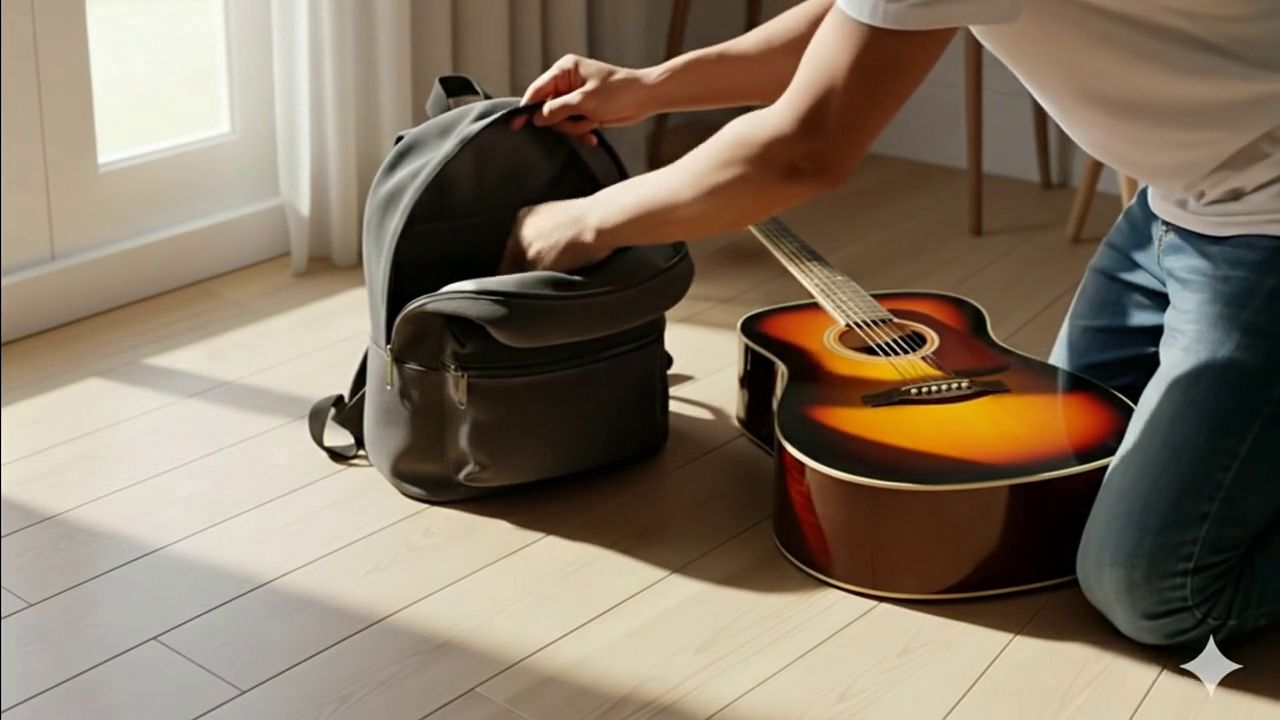}}
    \end{subfigure}
    \caption{\textbf{Object packing}.
    Prompt: \textit{``A person puts all the objects that can fit in the backpack inside of it. Static camera, no pan, no zoom, no dolly.''} Success rate: 0.75.}
    \label{fig:fitting_objects}
\end{figure}

\begin{figure}[H]
    \centering
    \begin{subfigure}[t]{\threefigwidth\textwidth}
        \centering
        \fbox{\includegraphics[width=\linewidth]{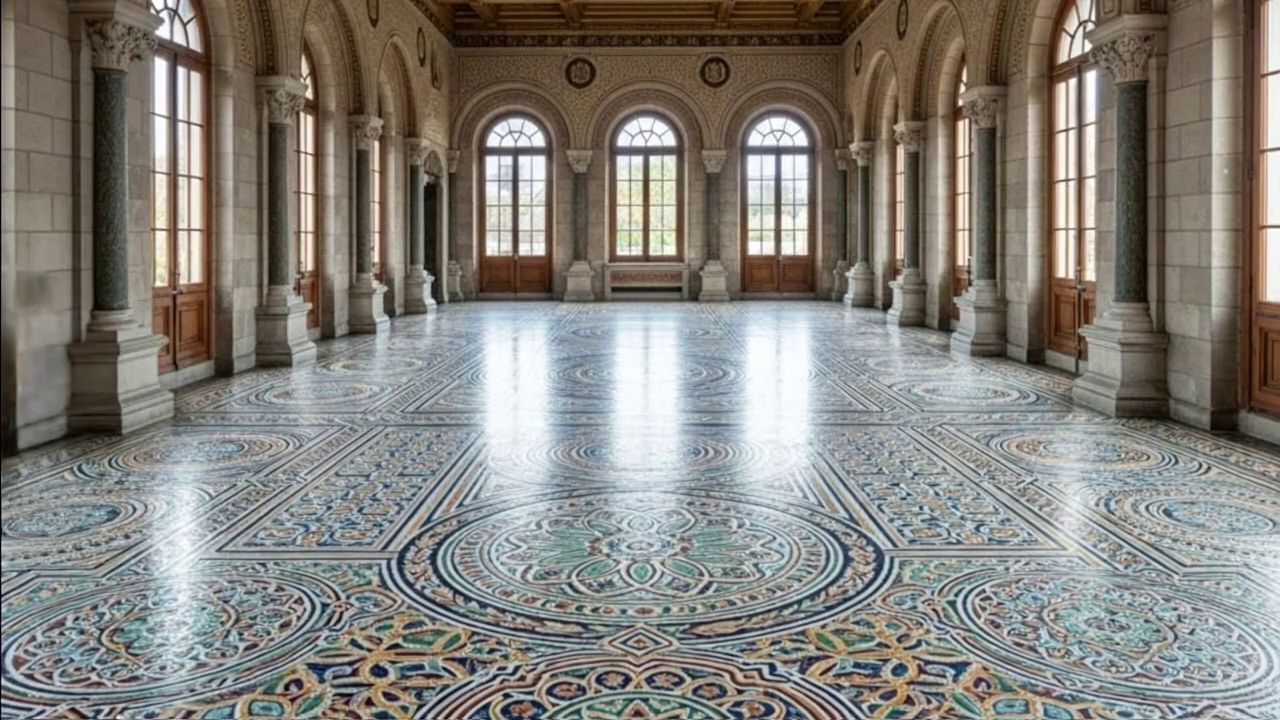}}
    \end{subfigure}
    \hfill
    \begin{subfigure}[t]{\threefigwidth\textwidth}
        \centering
        \fbox{\includegraphics[width=\linewidth]{figures-compressed/material_optics_Room_Glass_Sphere_frame080.jpg}}
    \end{subfigure}
    \hfill    
    \begin{subfigure}[t]{\threefigwidth\textwidth}
        \centering
        \fbox{\includegraphics[width=\linewidth]{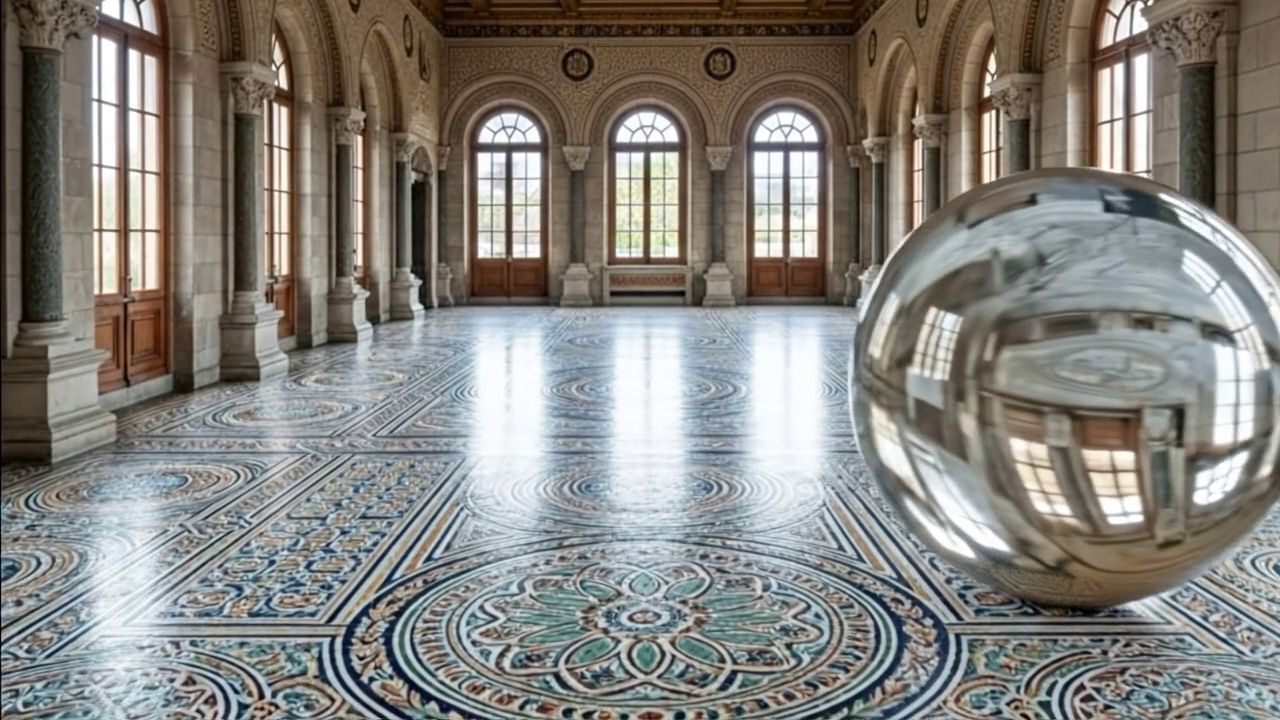}}
    \end{subfigure}

    \vspace{0.2cm}
    
    \begin{subfigure}[t]{\threefigwidth\textwidth}
        \centering
        \fbox{\includegraphics[width=\linewidth]{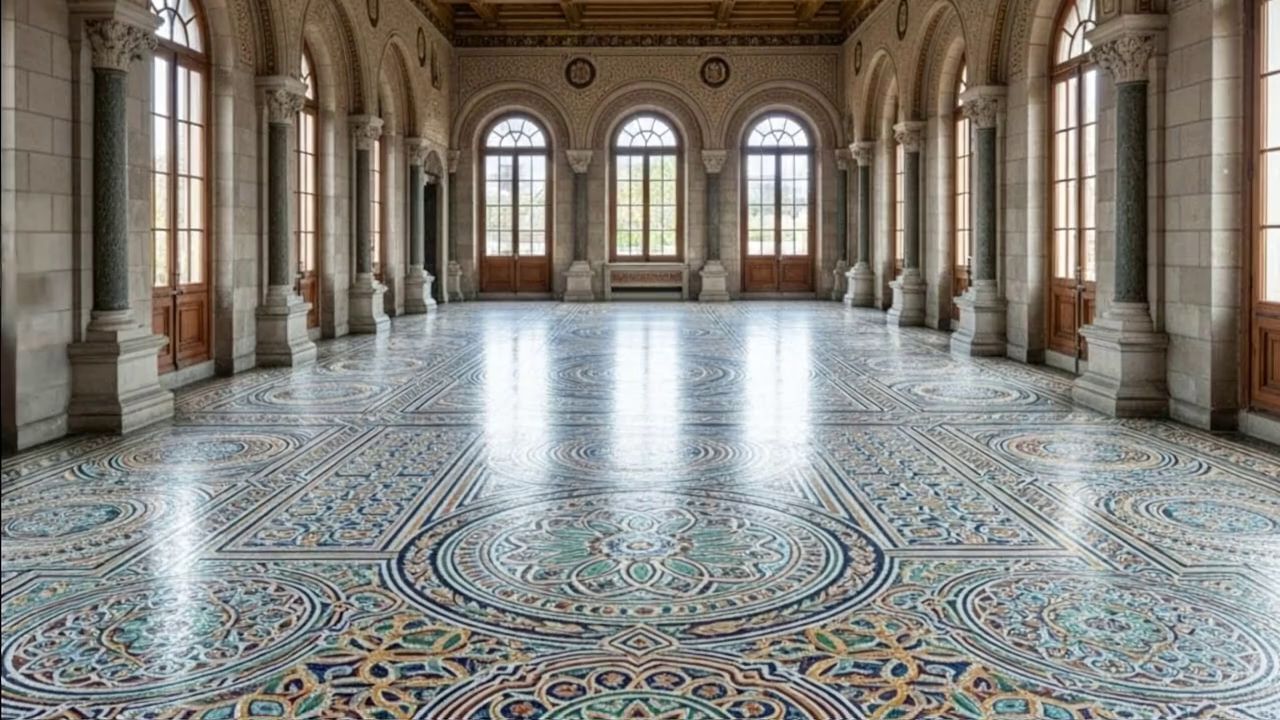}}
    \end{subfigure}
    \hfill
    \begin{subfigure}[t]{\threefigwidth\textwidth}
        \centering
        \fbox{\includegraphics[width=\linewidth]{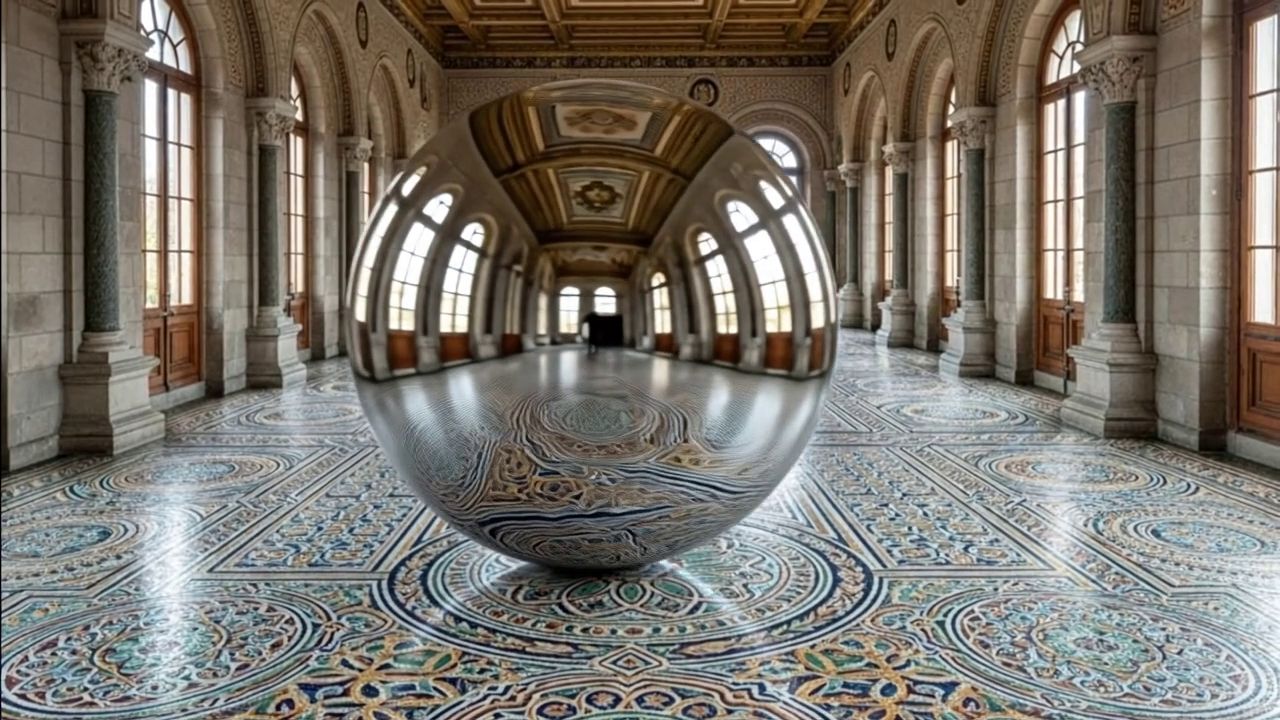}}
    \end{subfigure}
    \hfill    
    \begin{subfigure}[t]{\threefigwidth\textwidth}
        \centering
        \fbox{\includegraphics[width=\linewidth]{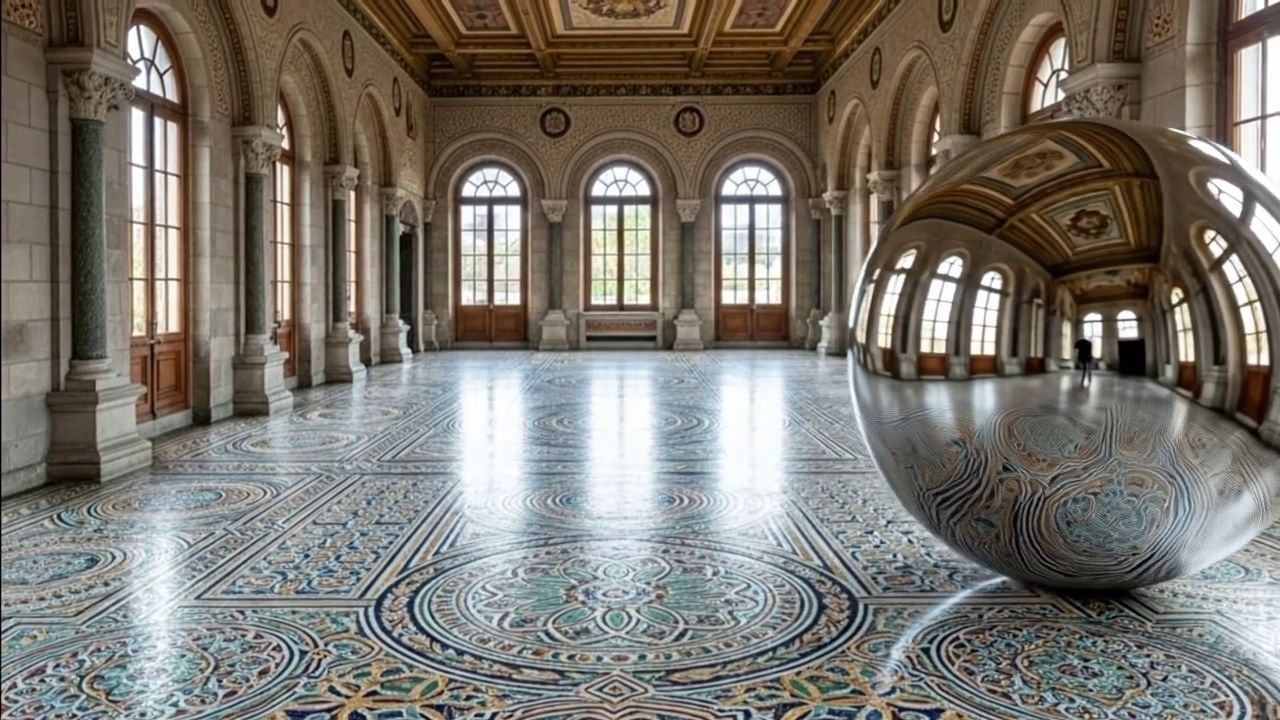}}
    \end{subfigure}
    \caption{\textbf{Material optics}.
    \textbf{Glass} (top). Prompt: \textit{``A giant glass sphere rolls through the room. Static camera, no pan, no zoom, no dolly.''} Note that the image through the glass sphere is inverted. Success rate:~0.92.
    \textbf{Mirror} (bottom). Prompt: \textit{``A giant mirror-polish metal sphere rolls through the room. Static camera, no pan, no zoom, no dolly.''} Note that the image reflected off the sphere is not inverted). Success rate:~1.0.}
    \label{fig:material_optics}
\end{figure}

\begin{figure}[H]
    \centering
    \begin{subfigure}[t]{\threefigwidth\textwidth}
        \centering
        \fbox{\includegraphics[width=\linewidth]{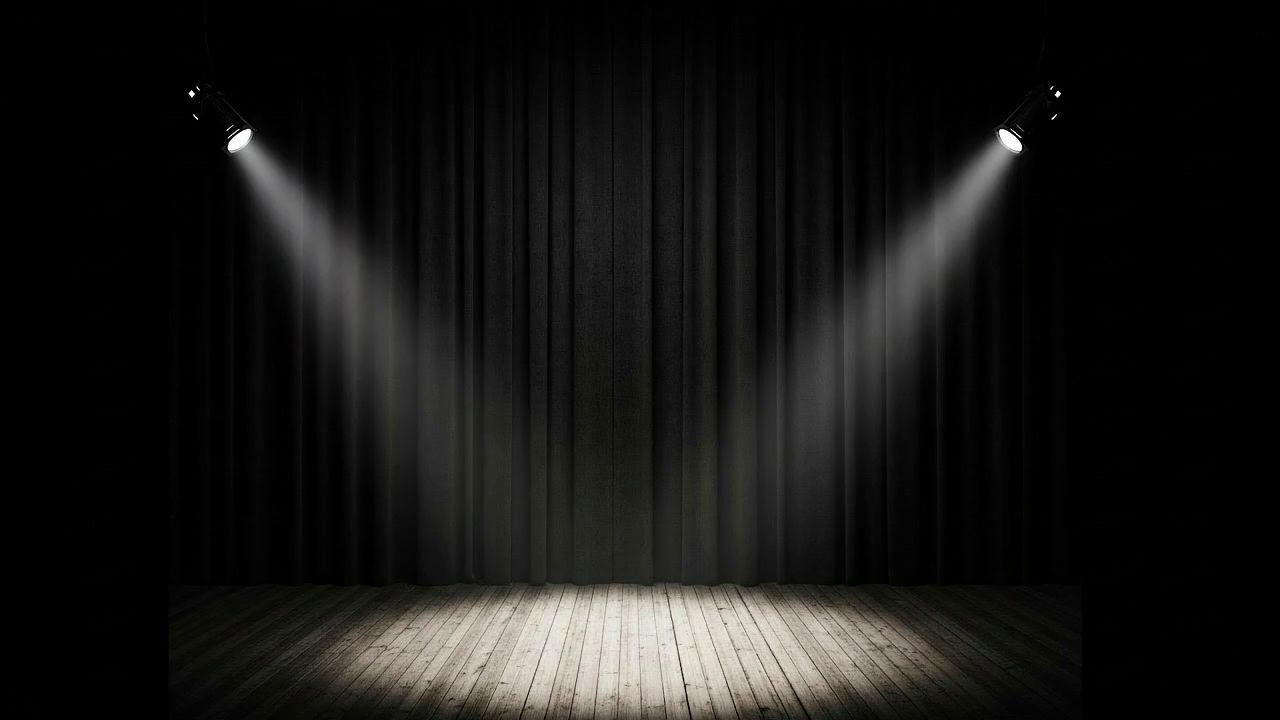}}
    \end{subfigure}
    \hfill
    \begin{subfigure}[t]{\threefigwidth\textwidth}
        \centering
        \fbox{\includegraphics[width=\linewidth]{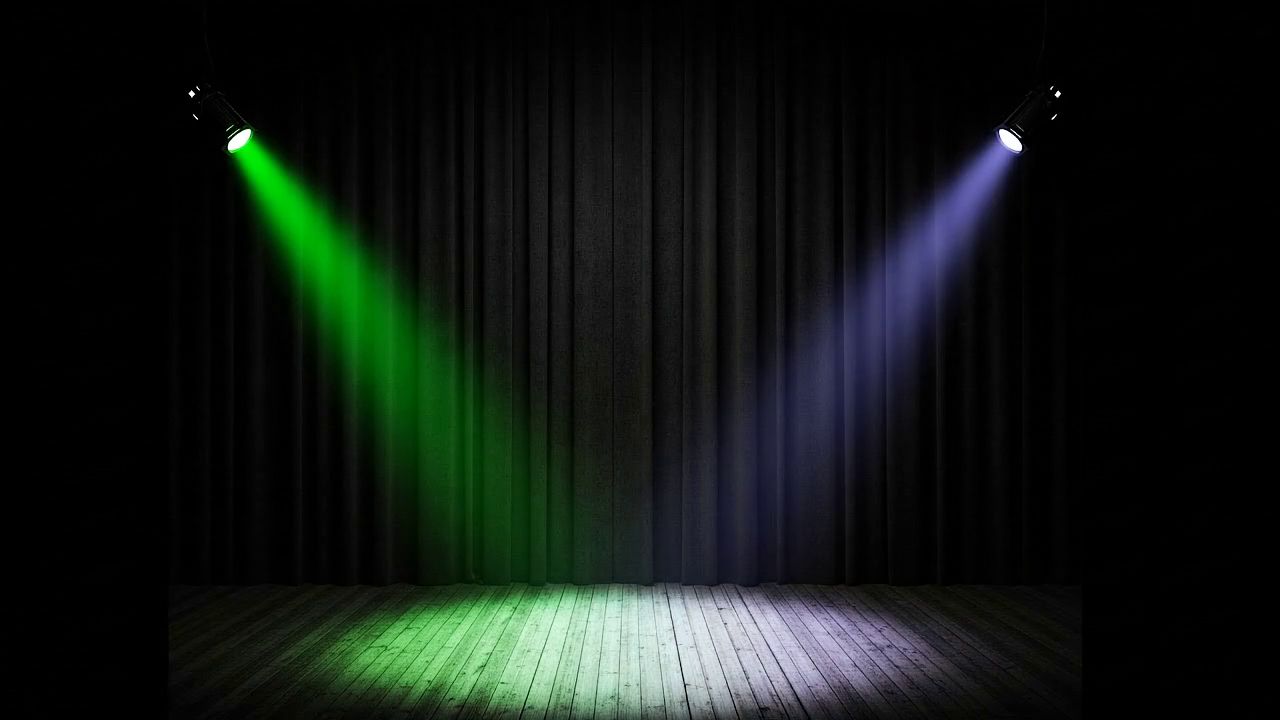}}
    \end{subfigure}
    \hfill    
    \begin{subfigure}[t]{\threefigwidth\textwidth}
        \centering
        \fbox{\includegraphics[width=\linewidth]{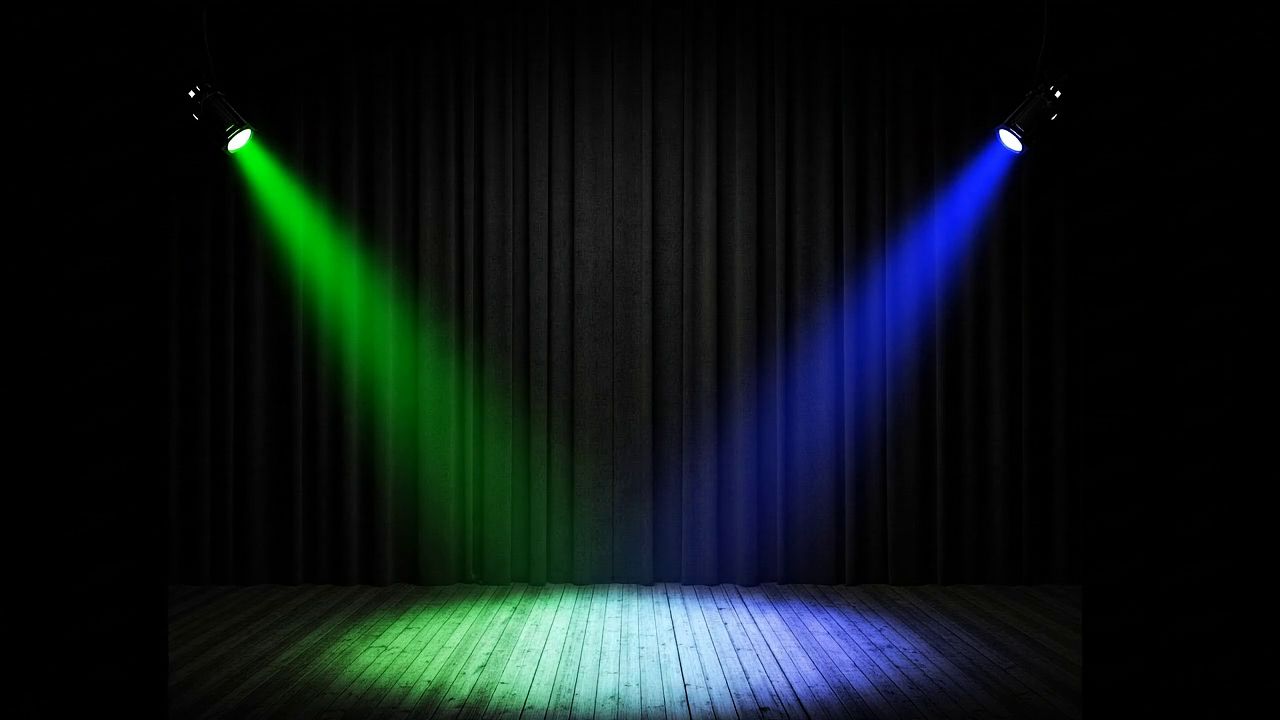}}
    \end{subfigure}

    \vspace{0.2cm}

    \begin{subfigure}[t]{\threefigwidth\textwidth}
        \centering
        \fbox{\includegraphics[width=\linewidth]{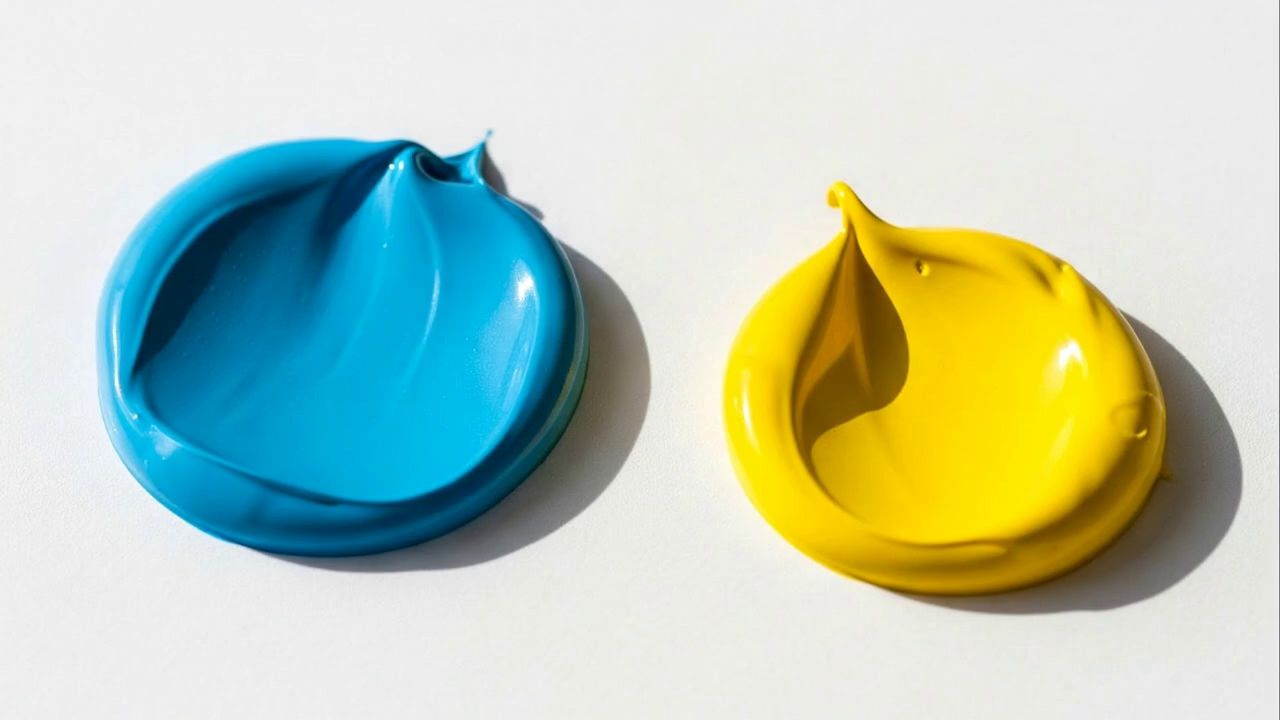}}
    \end{subfigure}
    \hfill
    \begin{subfigure}[t]{\threefigwidth\textwidth}
        \centering
        \fbox{\includegraphics[width=\linewidth]{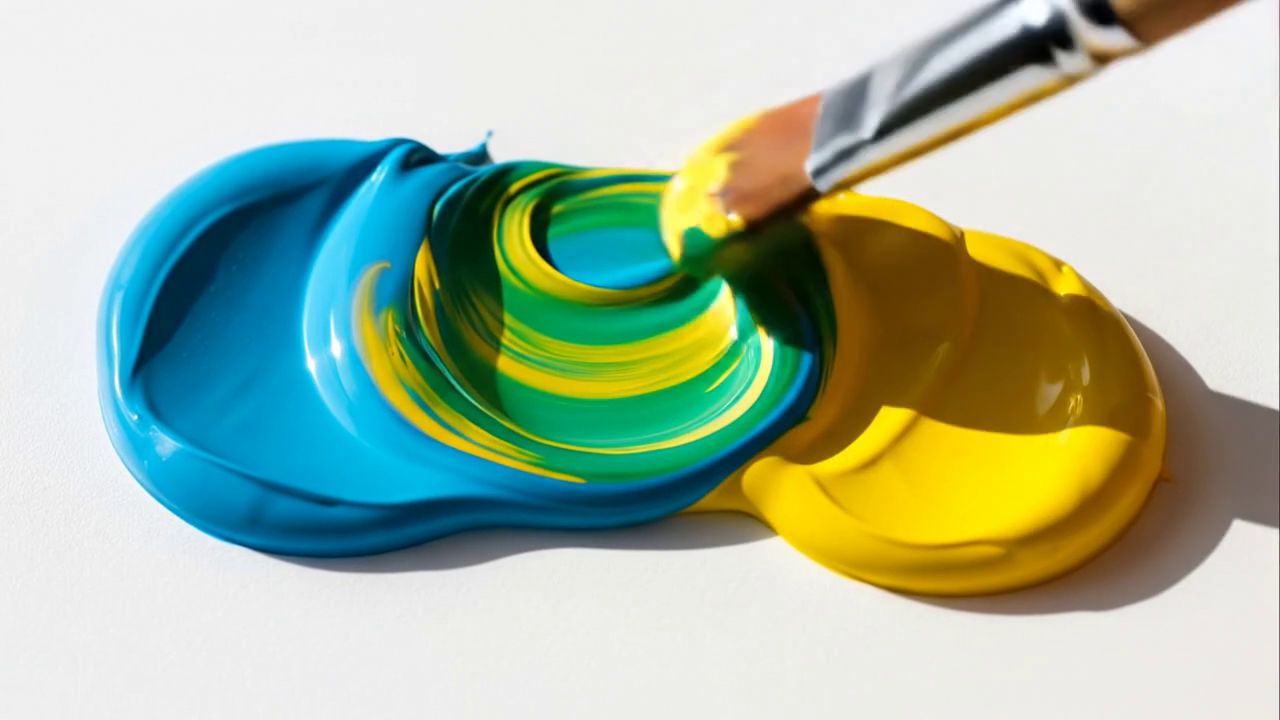}}
    \end{subfigure}
    \hfill    
    \begin{subfigure}[t]{\threefigwidth\textwidth}
        \centering
        \fbox{\includegraphics[width=\linewidth]{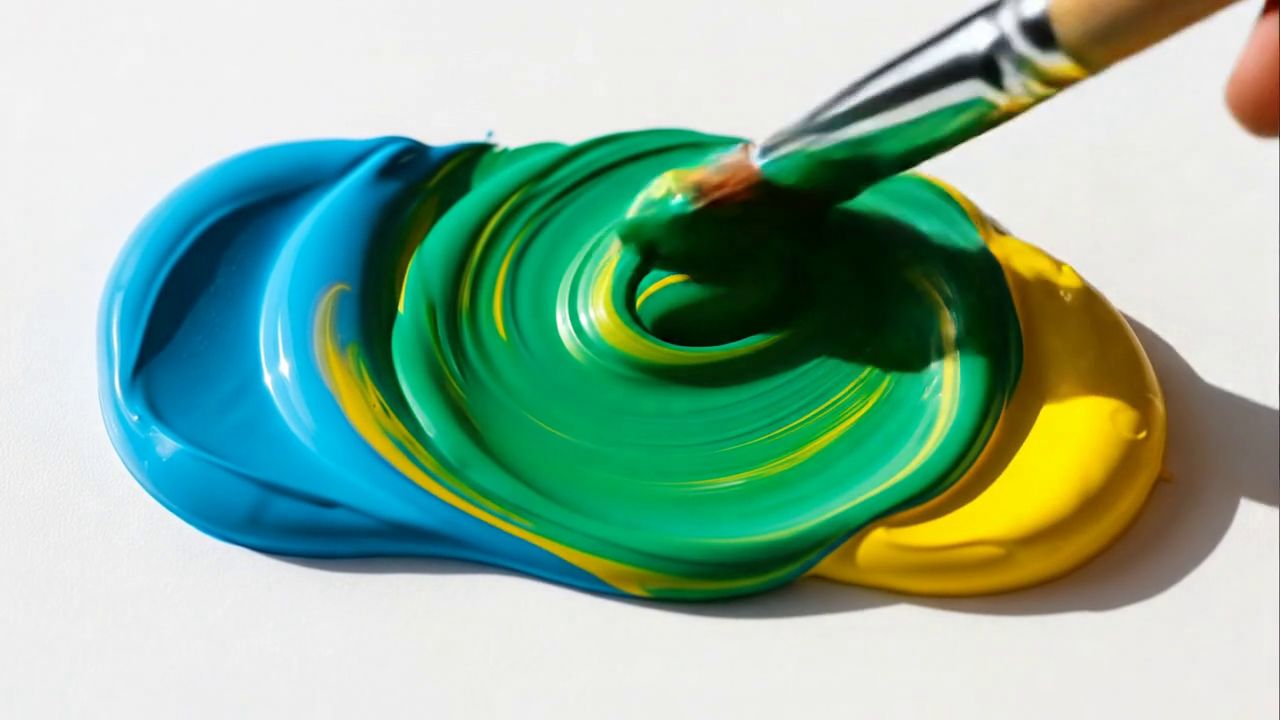}}
    \end{subfigure}
    \caption{\textbf{Color mixing}.
    \textbf{Additive} (lights, top). Prompt: \textit{``The spotlight on the left changes color to green, and the spotlight on the right changes color to blue.''} Success rate:~0.92.
    \textbf{Subtractive} (paints, bottom). Prompt: \textit{``A paintbrush mixes these colors together thoroughly until they blend completely. Static camera, no pan, no zoom.''} Success rate:~0.75.}
    \label{fig:color_mixing}
\end{figure}

\begin{figure}[H]
    \centering
    \begin{subfigure}[t]{\threefigwidth\textwidth}
        \centering
        \fbox{\includegraphics[width=\linewidth]{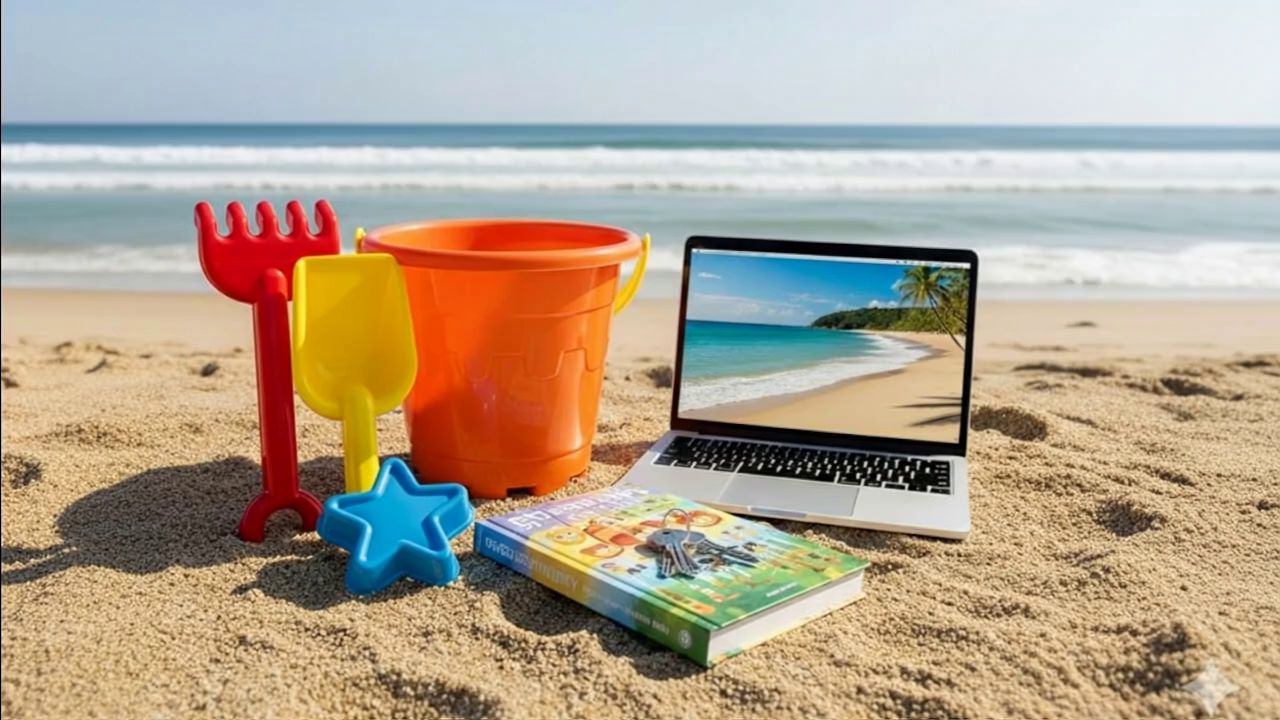}}
    \end{subfigure}
    \hfill
    \begin{subfigure}[t]{\threefigwidth\textwidth}
        \centering
        \fbox{\includegraphics[width=\linewidth]{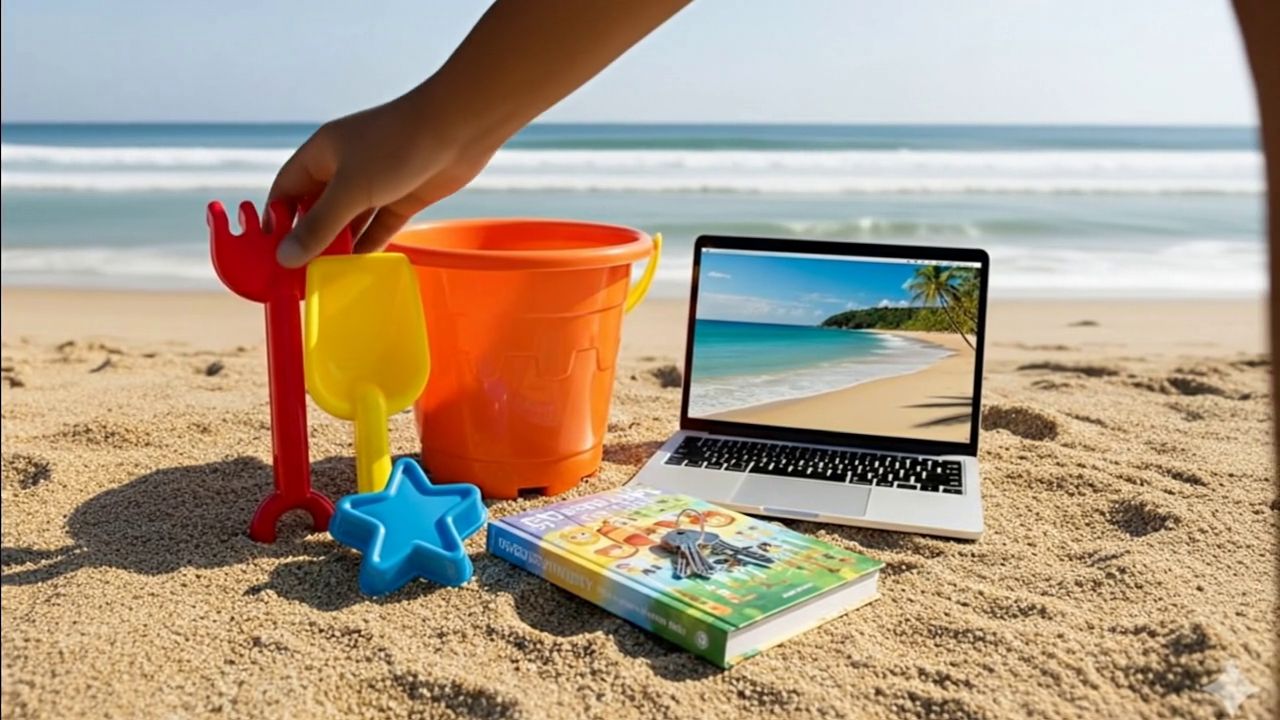}}
    \end{subfigure}
    \hfill    
    \begin{subfigure}[t]{\threefigwidth\textwidth}
        \centering
        \fbox{\includegraphics[width=\linewidth]{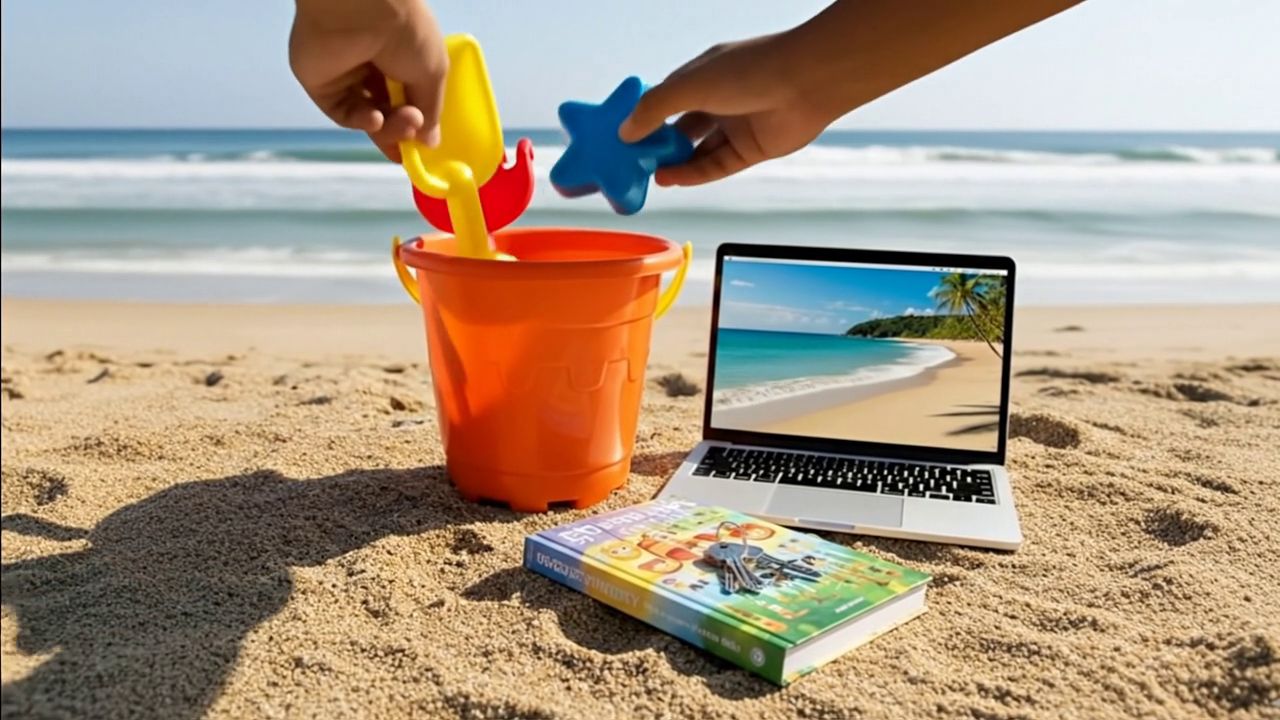}}
    \end{subfigure}
    \caption{\textbf{Categorizing objects.}
    Prompt: \textit{``A person puts all the kids toys in the bucket. Static camera, no pan, no zoom, no dolly.''} Success rate:~0.33.}
    \label{fig:categorizing_objects}
\end{figure}

\begin{figure}[H]
    \centering
    \begin{subfigure}[t]{\threefigwidth\textwidth}
        \centering
        \fbox{\includegraphics[width=\linewidth]{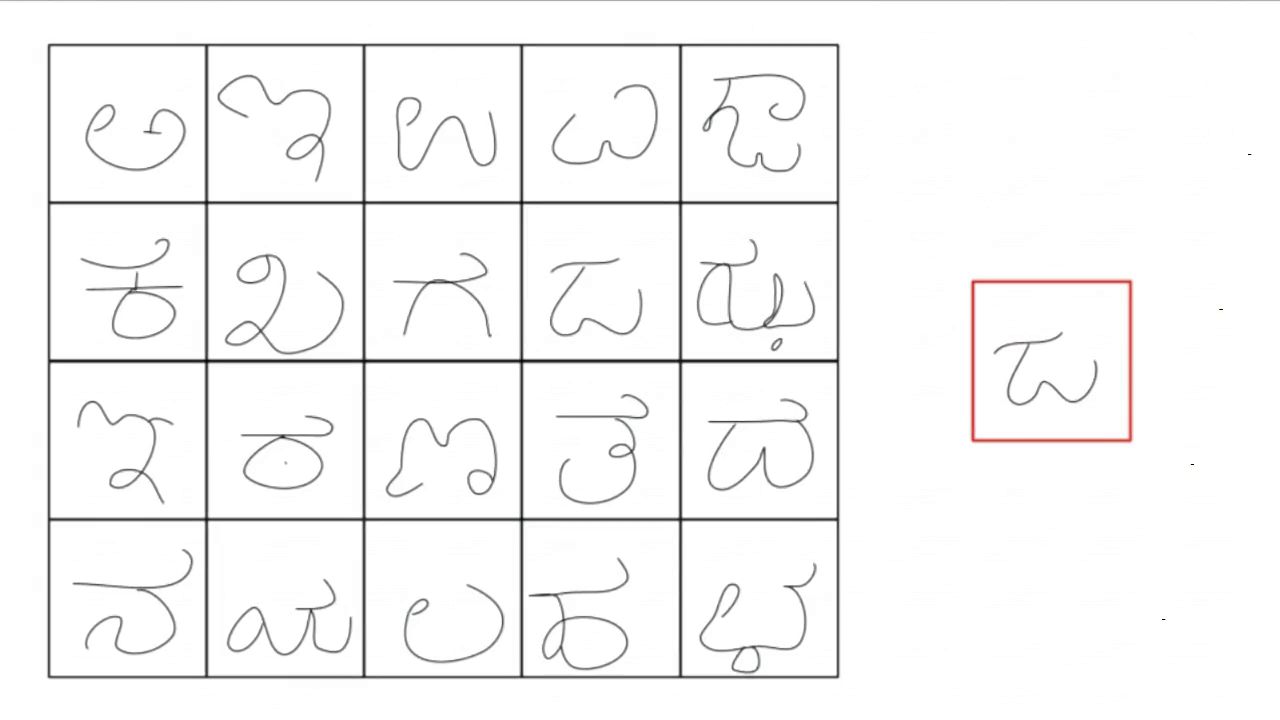}}
    \end{subfigure}
    \hfill
    \begin{subfigure}[t]{\threefigwidth\textwidth}
        \centering
        \fbox{\includegraphics[width=\linewidth]{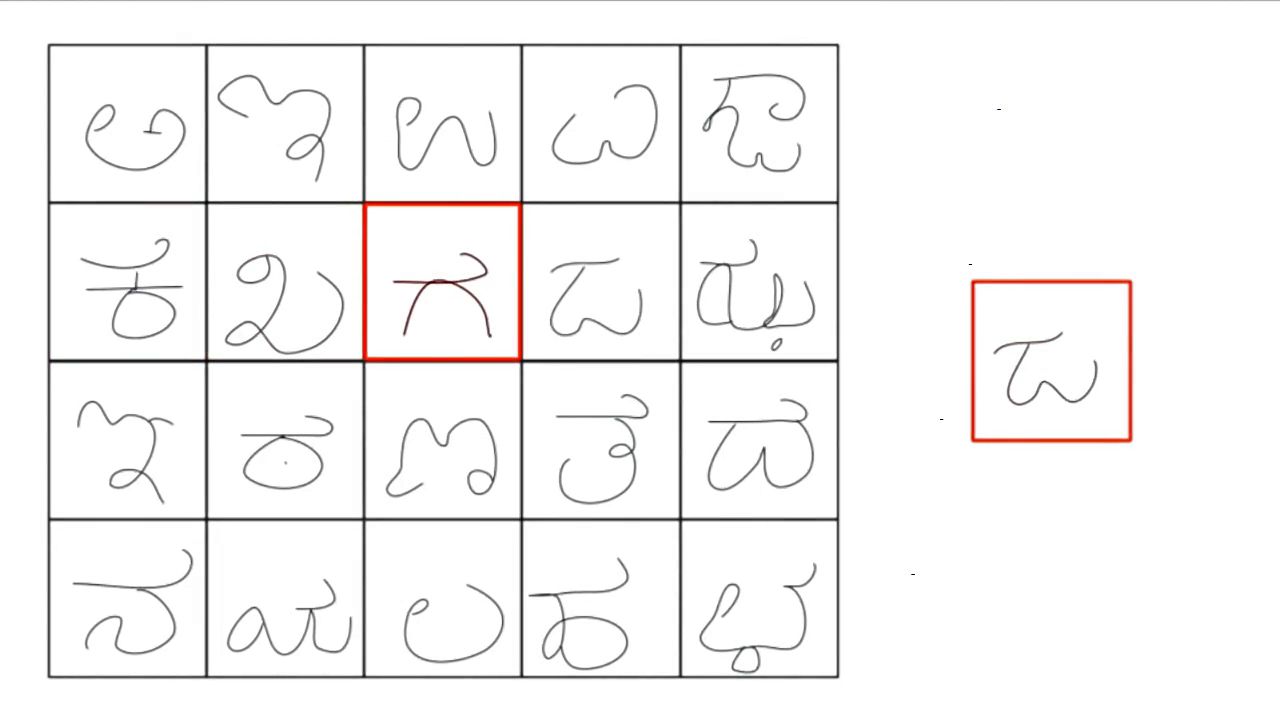}}
    \end{subfigure}
    \hfill    
    \begin{subfigure}[t]{\threefigwidth\textwidth}
        \centering
        \fbox{\includegraphics[width=\linewidth]{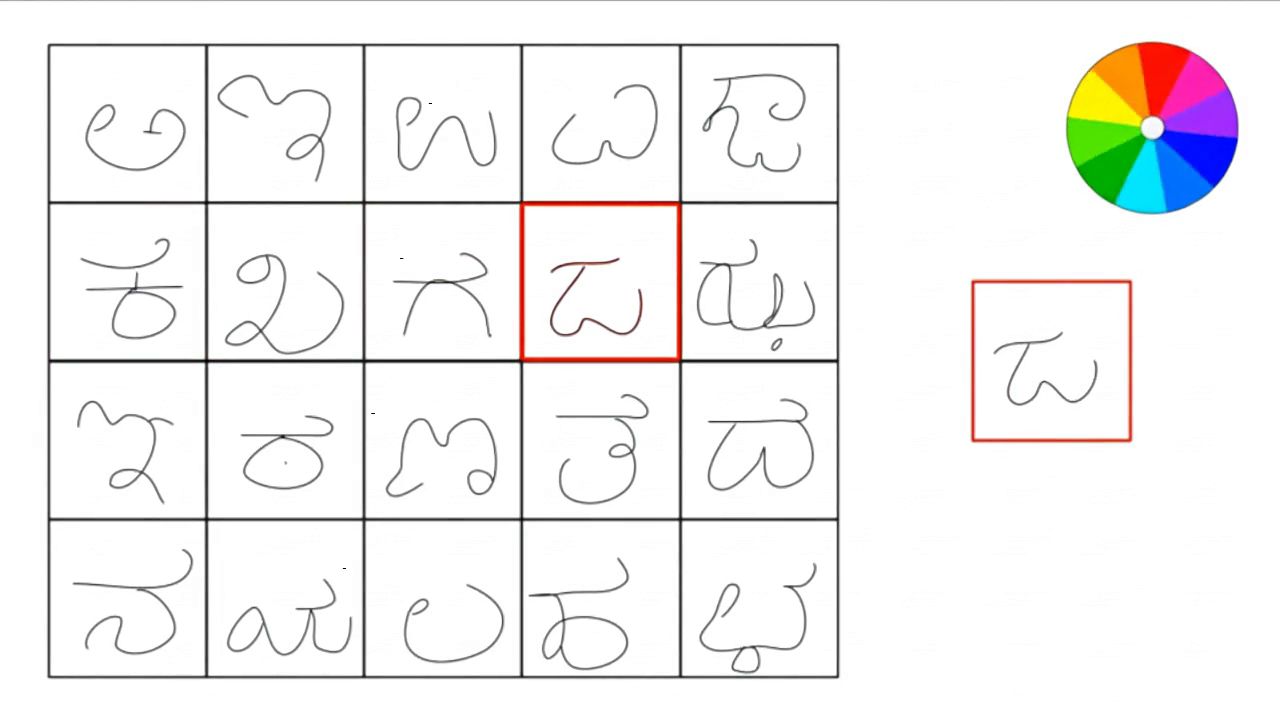}}
    \end{subfigure}
    
    \vspace{0.2cm} %
    
    \centering
    \begin{subfigure}[t]{\threefigwidth\textwidth}
        \centering
        \fbox{\includegraphics[width=\linewidth]{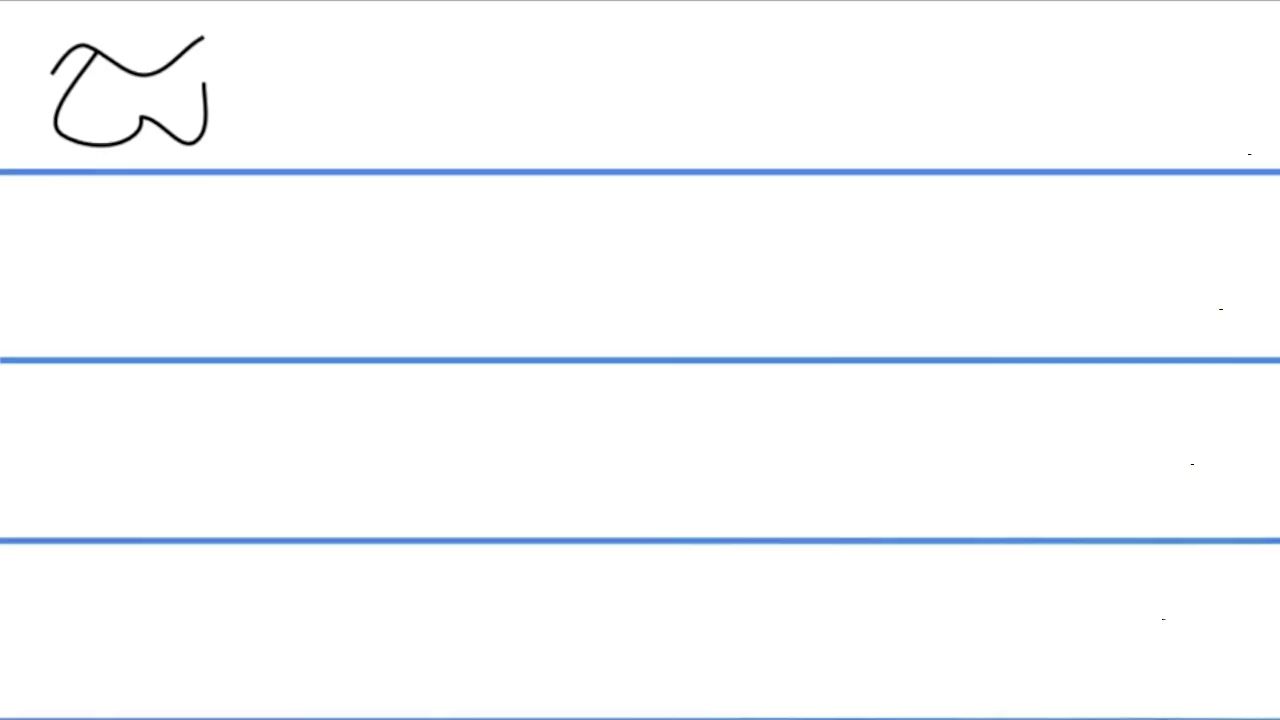}}
    \end{subfigure}
    \hfill
    \begin{subfigure}[t]{\threefigwidth\textwidth}
        \centering
        \fbox{\includegraphics[width=\linewidth]{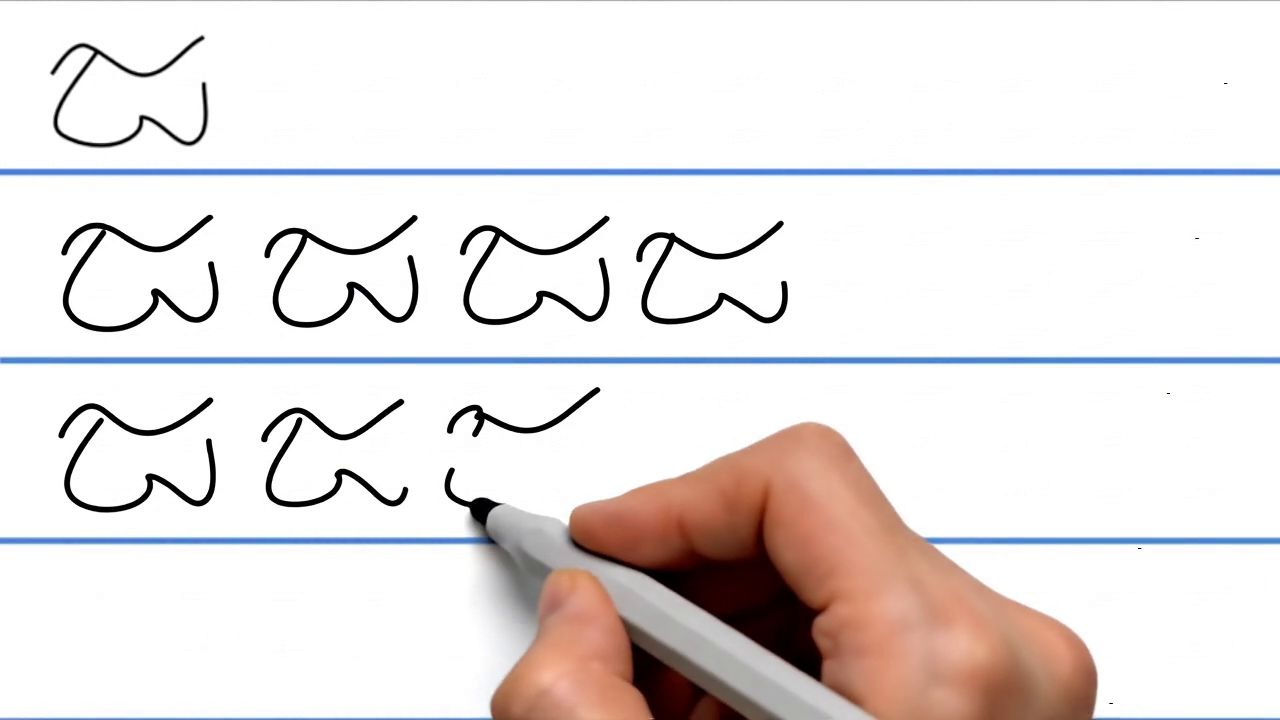}}
    \end{subfigure}
    \hfill    
    \begin{subfigure}[t]{\threefigwidth\textwidth}
        \centering
        \fbox{\includegraphics[width=\linewidth]{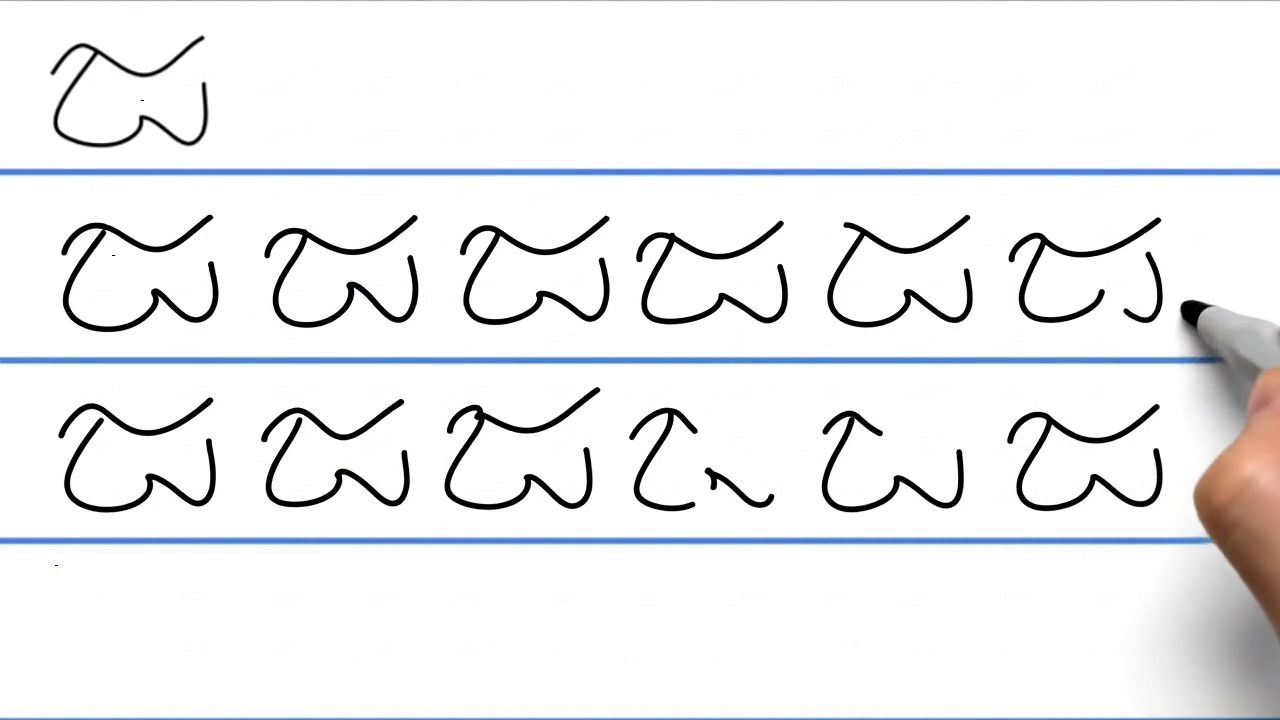}}
    \end{subfigure}
    
    \vspace{0.2cm} %
    
    \centering
    \begin{subfigure}[t]{\threefigwidth\textwidth}
        \centering
        \fbox{\includegraphics[width=\linewidth]{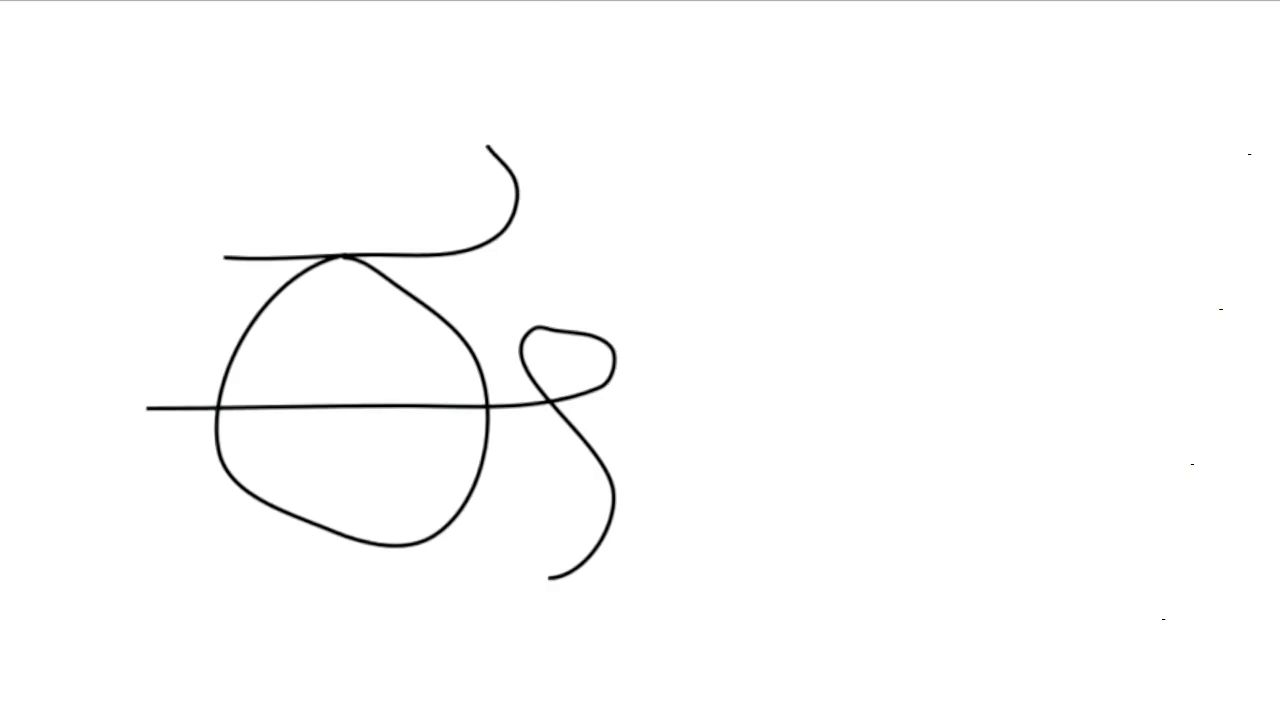}}
    \end{subfigure}
    \hfill
    \begin{subfigure}[t]{\threefigwidth\textwidth}
        \centering
        \fbox{\includegraphics[width=\linewidth]{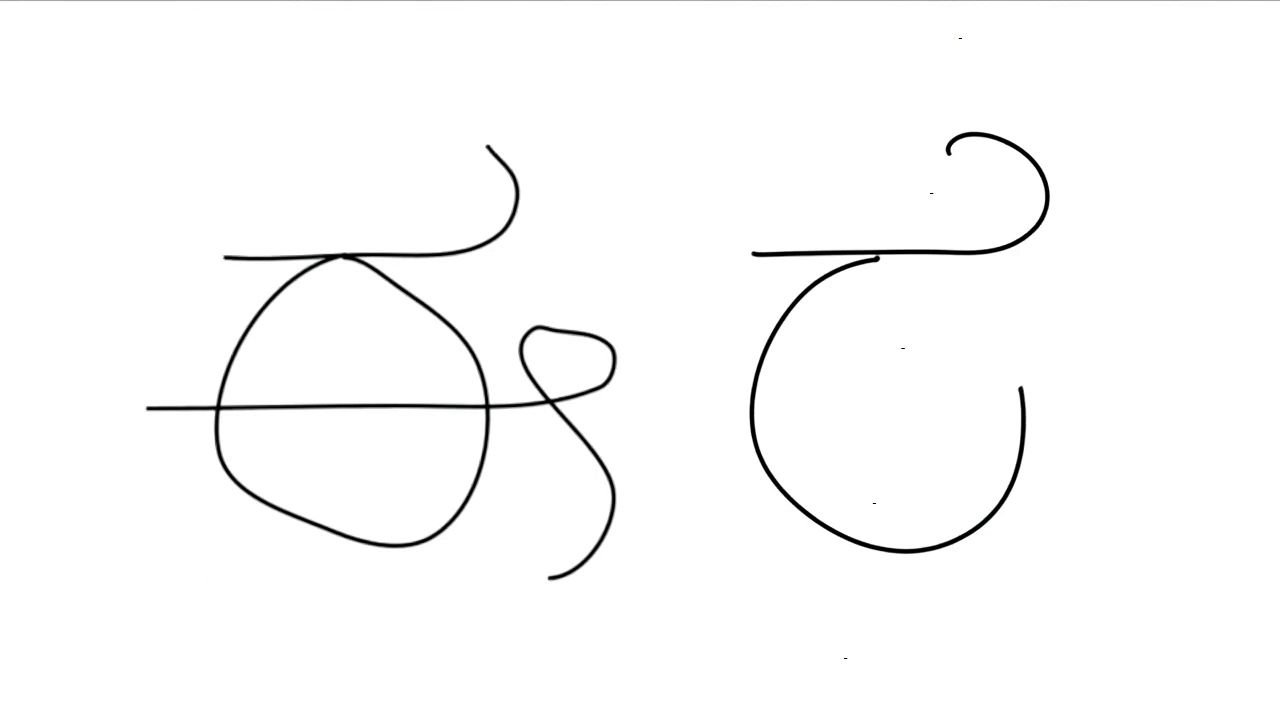}}
    \end{subfigure}
    \hfill    
    \begin{subfigure}[t]{\threefigwidth\textwidth}
        \centering
        \fbox{\includegraphics[width=\linewidth]{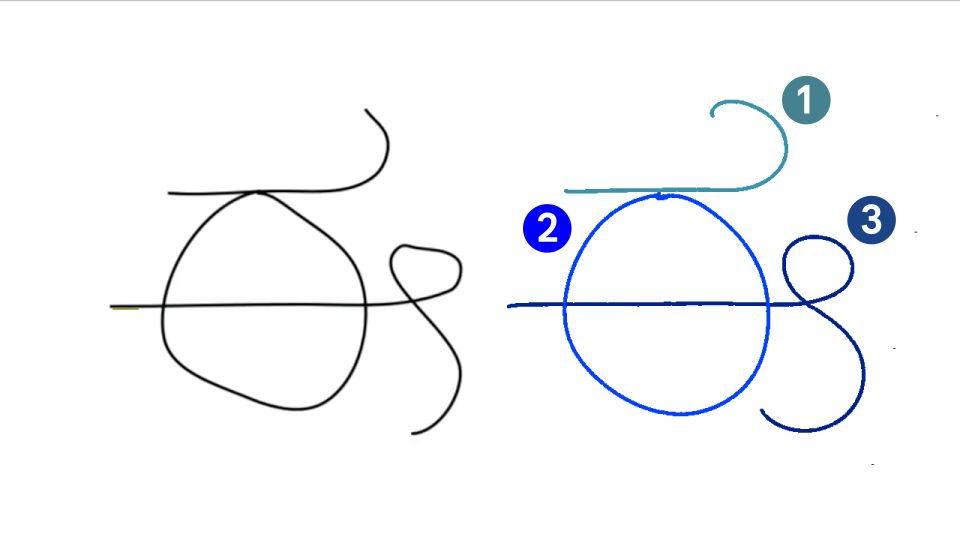}}
    \end{subfigure}
    \caption{\textbf{Character recognition, generation, and parsing}, inspired by the Omniglot dataset~\citep{lake2015human}.
    \textbf{Recognition} (top). Prompt: \textit{``The background of the grid cell with the same symbol as the one indicated on the right turns red. All other grid cells remain unchanged. After that, a spinning color wheel appears in the top right corner.''} (Note: \veothree{} has a prior to keep things moving, which is detrimental for tasks where the solution is obtained in an early frame. We observe that a `motion outlet', such as a color wheel, can indicate task completion and `freeze' the solution.) Success rate:~0.33.
    \textbf{Generation of variations} (middle). Prompt: \textit{``The page is filled line-by-line with hand-written practice variations of the symbol.''} Success rate:~0.25.
    \textbf{Parsing into parts} (bottom). Color and numbers in final frame are added post-hoc to show stroke order. Prompt: \textit{``Stroke-by-stroke, a replica of the symbol is drawn on the right.''} Success rate:~0.5.}
    \label{fig:omniglot}
\end{figure}

\begin{figure}[H]
    \centering
    \begin{subfigure}[t]{\threefigwidth\textwidth}
        \centering
        \fbox{\includegraphics[width=\linewidth]{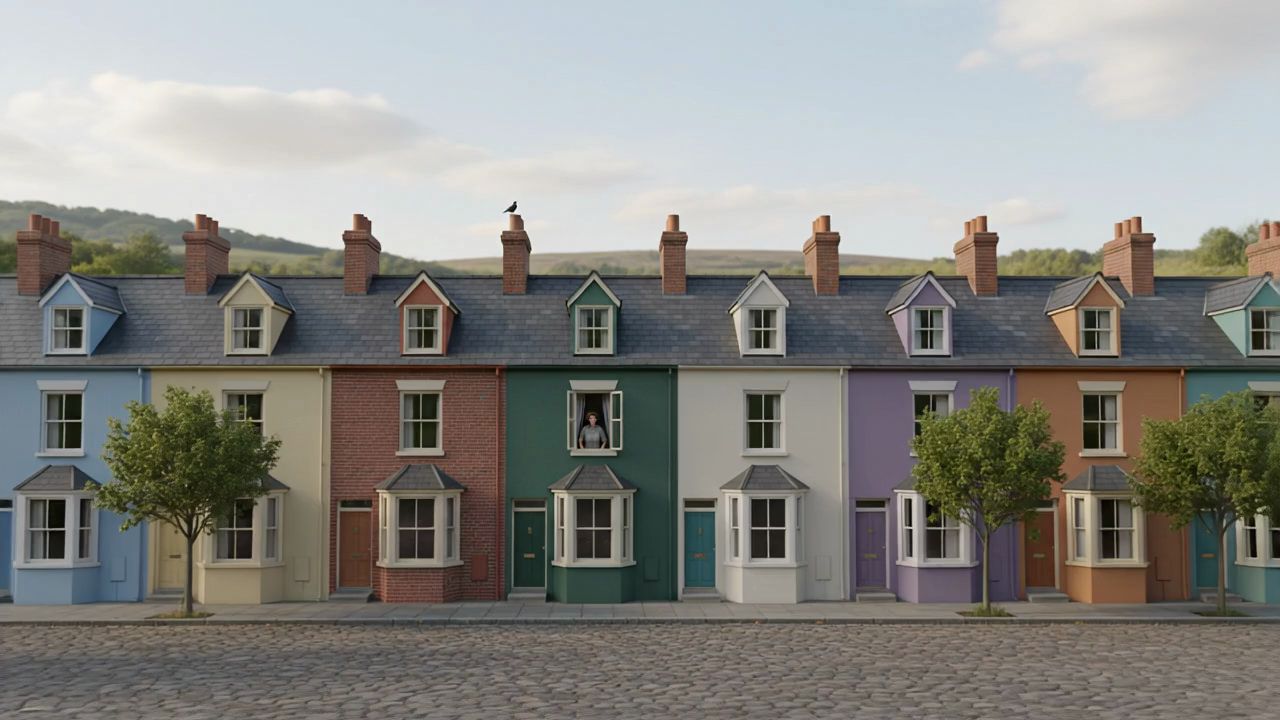}}
    \end{subfigure}
    \hfill
    \begin{subfigure}[t]{\threefigwidth\textwidth}
        \centering
        \fbox{\includegraphics[width=\linewidth]{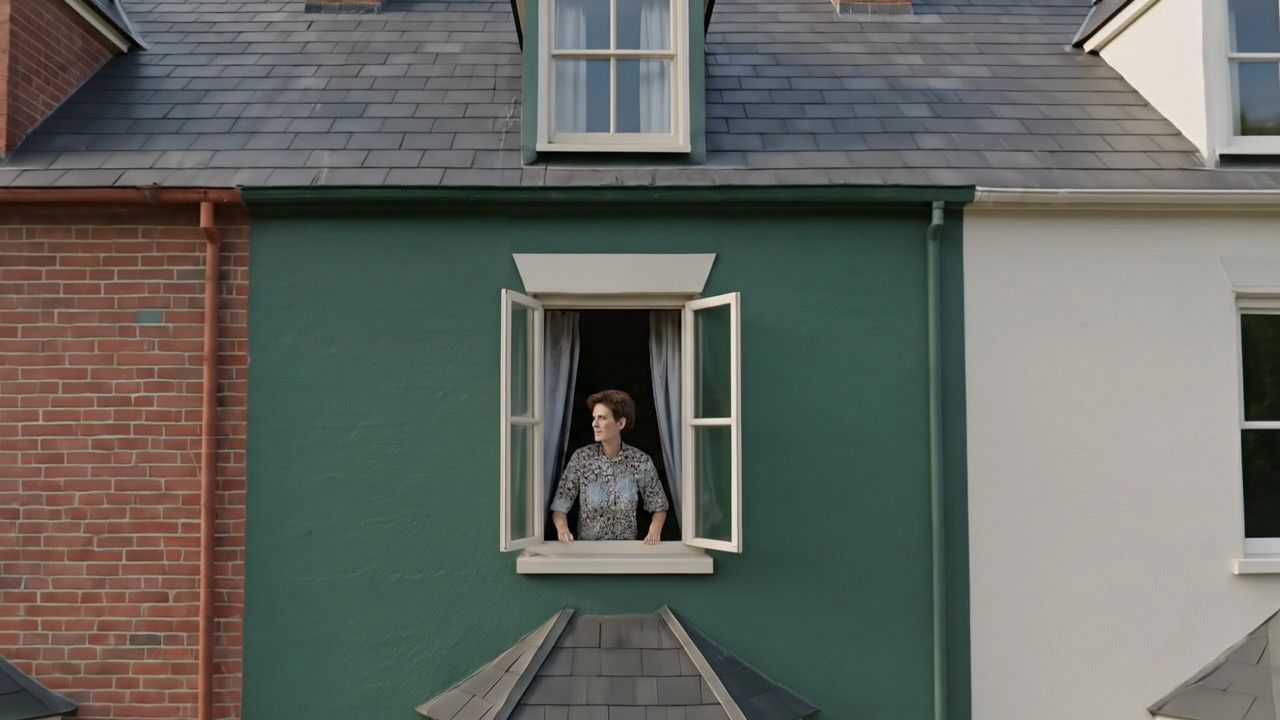}}
    \end{subfigure}
    \hfill    
    \begin{subfigure}[t]{\threefigwidth\textwidth}
        \centering
        \fbox{\includegraphics[width=\linewidth]{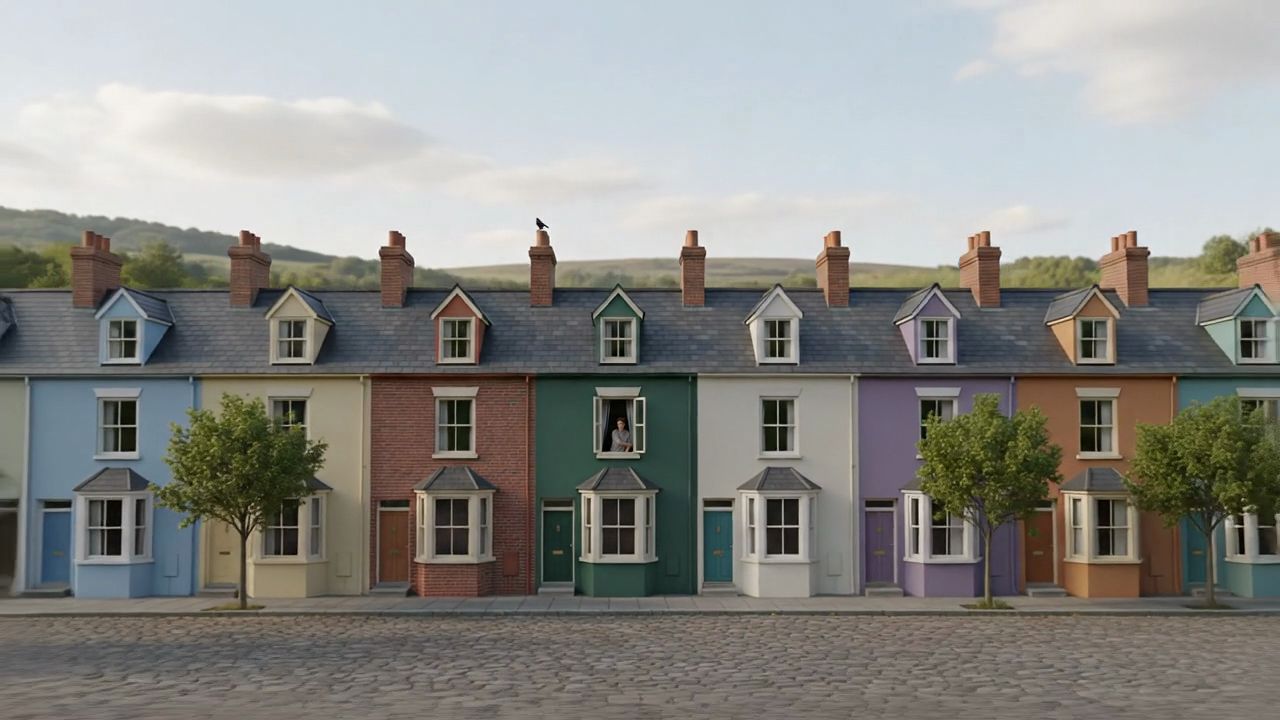}}
    \end{subfigure}
    \caption{\textbf{Memory of world states}.
    Prompt: \textit{``The camera zooms in to give a close up of the person looking out the window, then zooms back out to return to the original view.''} Success rate:~1.0.}
    \label{fig:working_memory}
\end{figure}

\subsection{\textcolor{manipulationcol}{\textbf{Manipulation}}}
\label{app:_manipulation_qualitative}

\begin{figure}[H]
    \centering
    \begin{subfigure}[t]{\threefigwidth\textwidth}
        \centering
        \fbox{\includegraphics[width=\linewidth]{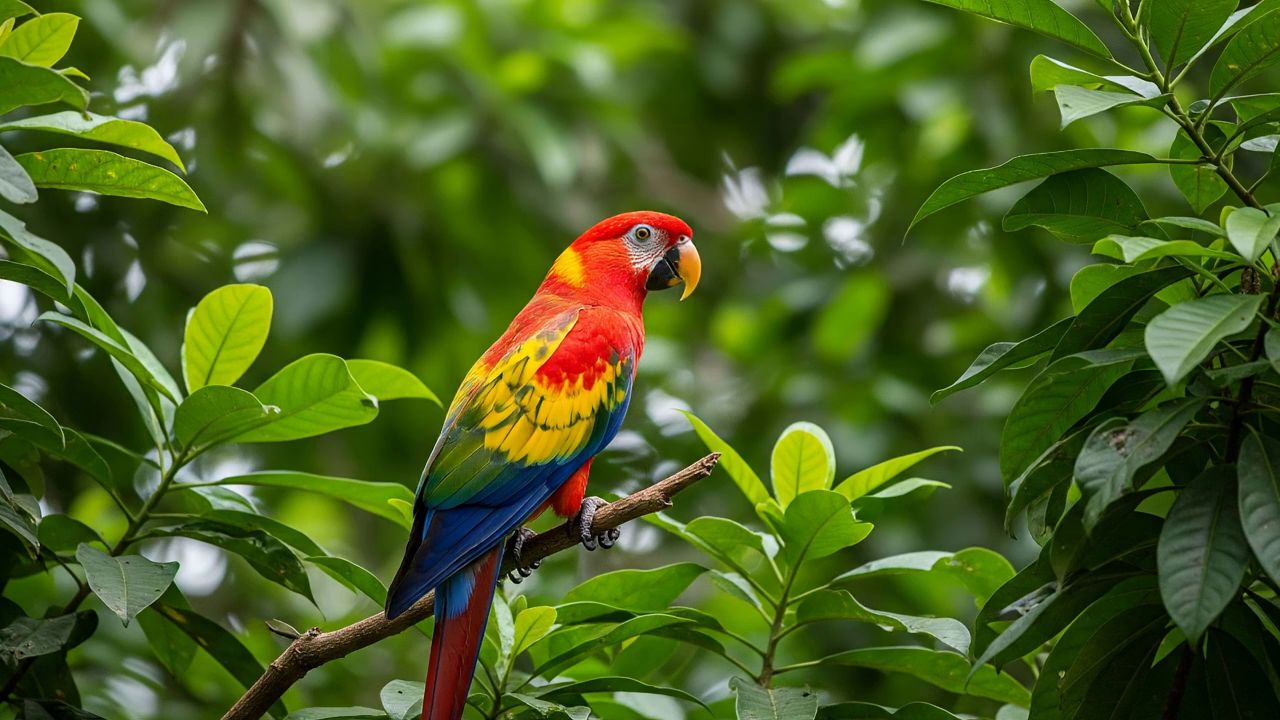}}
    \end{subfigure}
    \hfill
    \begin{subfigure}[t]{\threefigwidth\textwidth}
        \centering
        \fbox{\includegraphics[width=\linewidth]{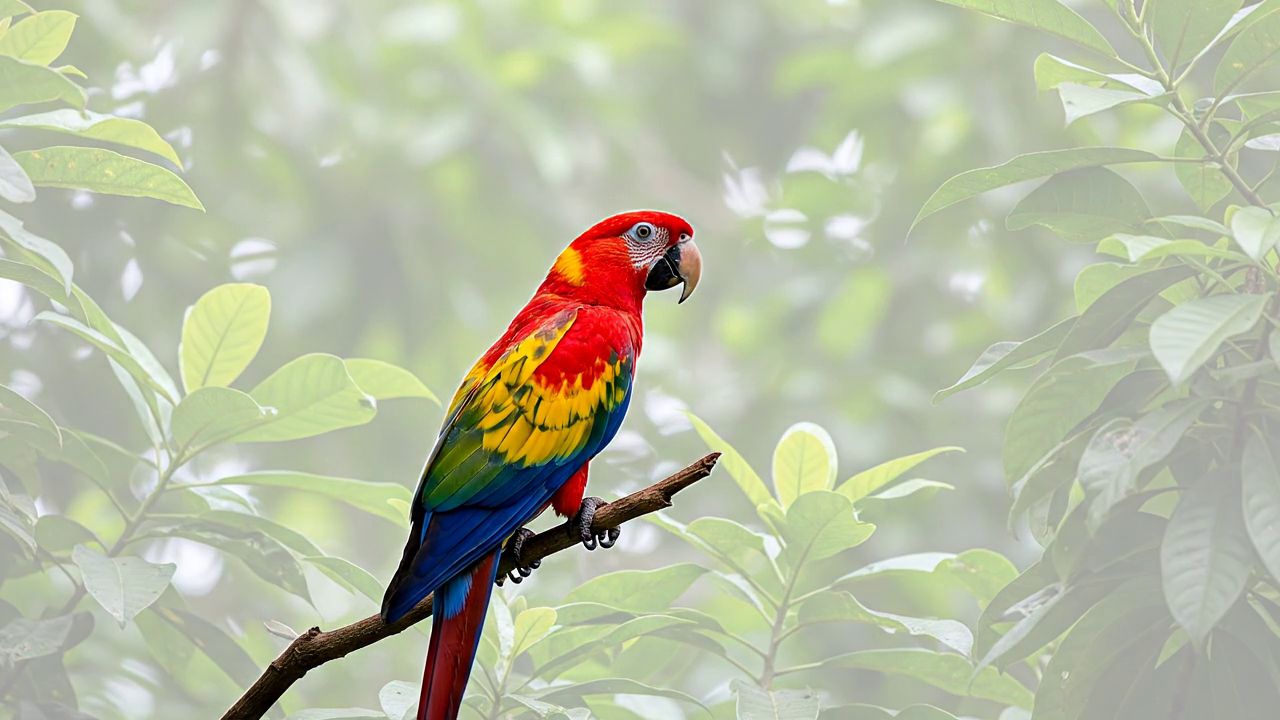}}
    \end{subfigure}
    \hfill    
    \begin{subfigure}[t]{\threefigwidth\textwidth}
        \centering
        \fbox{\includegraphics[width=\linewidth]{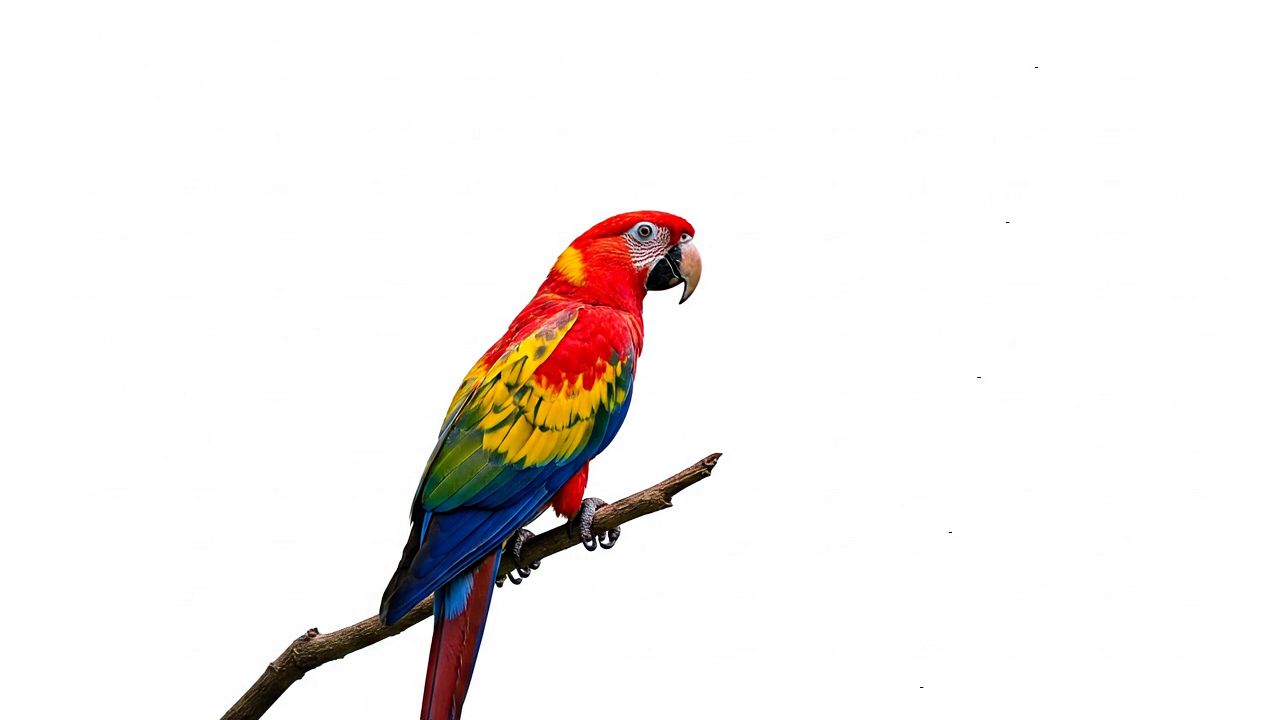}}
    \end{subfigure}
    \caption{\textbf{Background removal}. Prompt: \textit{``The background changes to white. Static camera perspective, no zoom or pan.''} Success rate:~0.83.}
    \label{fig:background_removal}
\end{figure}

\begin{figure}[H]
    \centering
    \begin{subfigure}[t]{\threefigwidth\textwidth}
        \centering
        \fbox{\includegraphics[width=\linewidth]{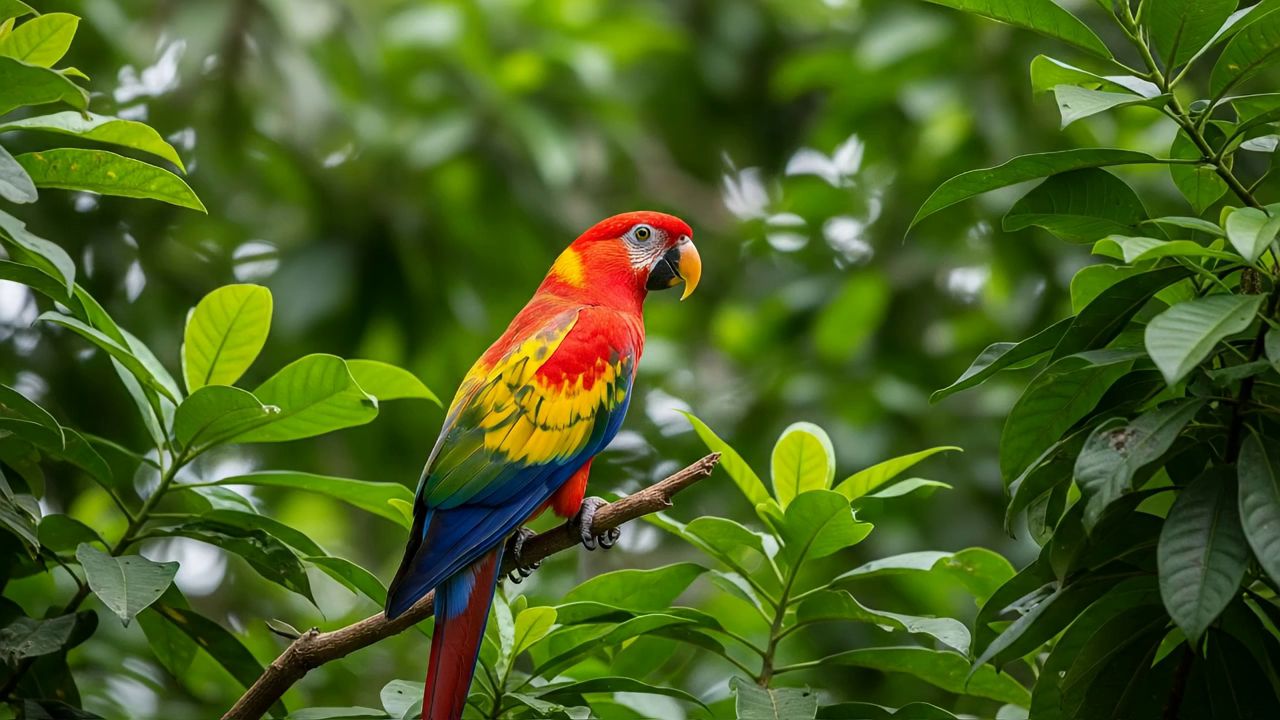}}
    \end{subfigure}
    \hfill
    \begin{subfigure}[t]{\threefigwidth\textwidth}
        \centering
        \fbox{\includegraphics[width=\linewidth]{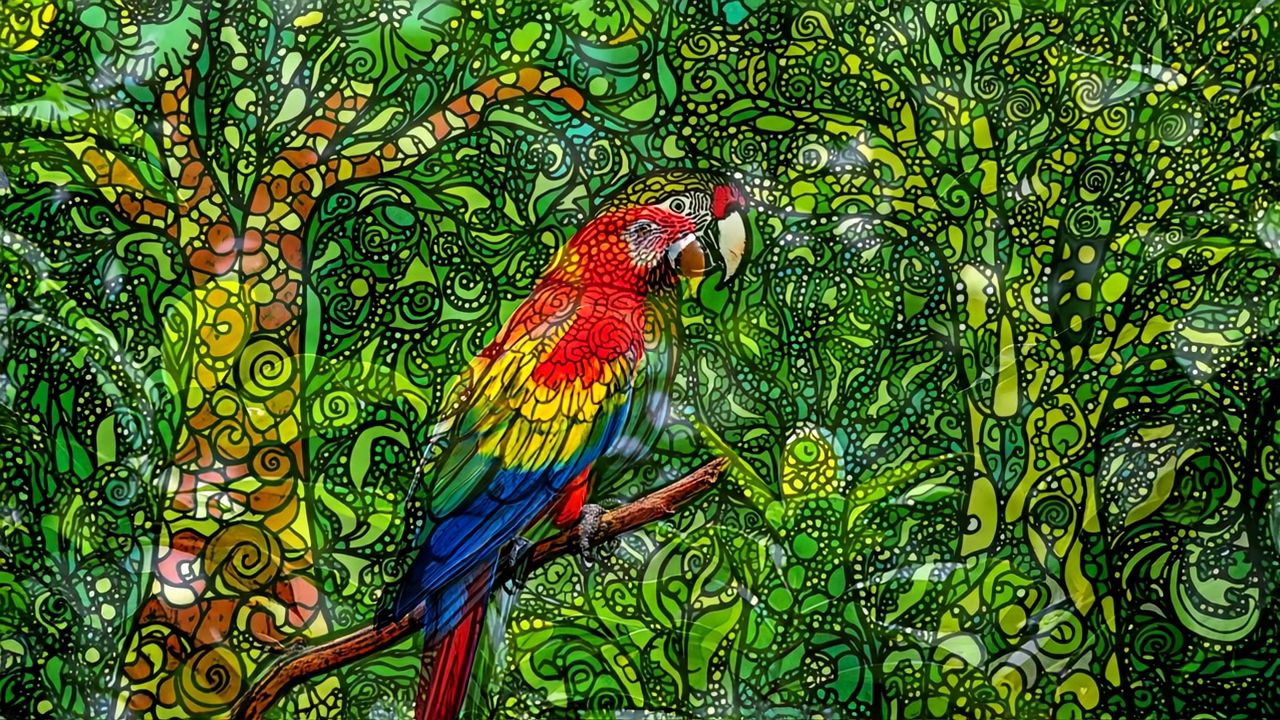}}
    \end{subfigure}
    \hfill    
    \begin{subfigure}[t]{\threefigwidth\textwidth}
        \centering
        \fbox{\includegraphics[width=\linewidth]{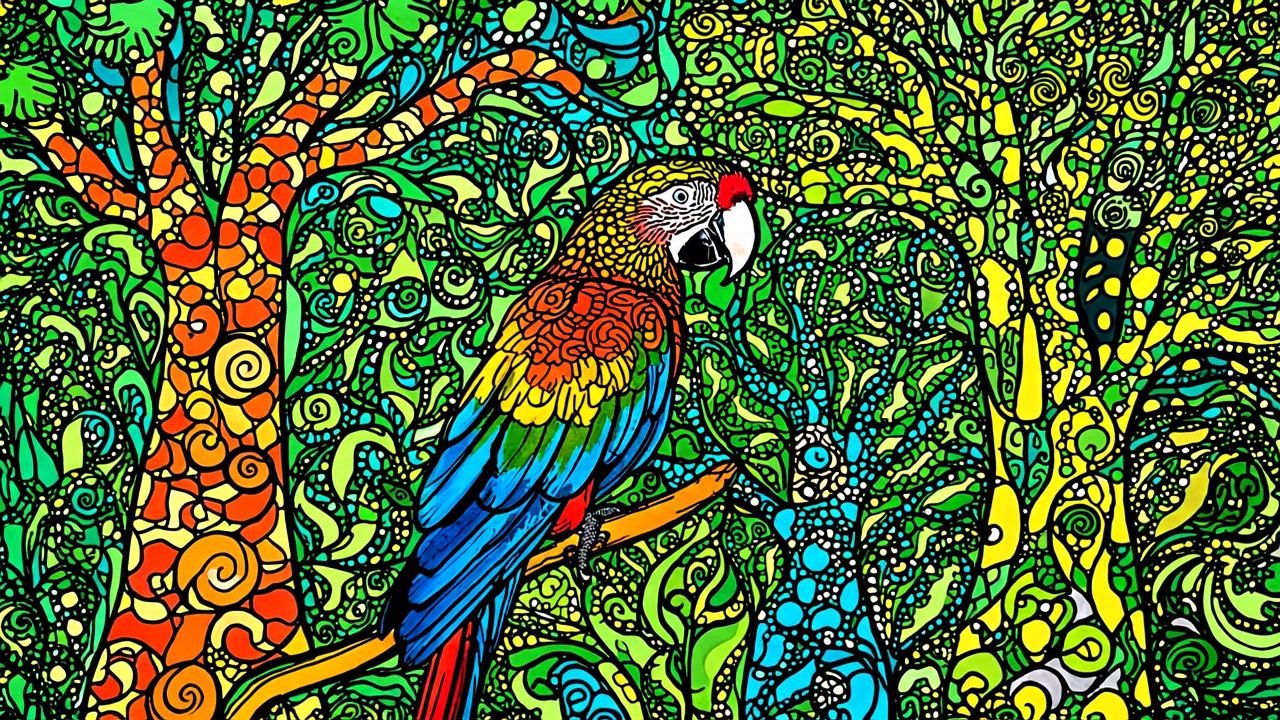}}
    \end{subfigure}
    \caption{\textbf{Style transfer}. Prompt: \textit{``The scene transforms into the style of a Hundertwasser painting, without changing perspective or orientation; the macaw does not move. Static camera perspective, no zoom or pan.''} Success rate:~0.75.}
    \label{fig:style_transfer}
\end{figure}

\begin{figure}[H]
    \centering
    \begin{subfigure}[t]{\threefigwidth\textwidth}
        \centering
        \fbox{\includegraphics[width=\linewidth]{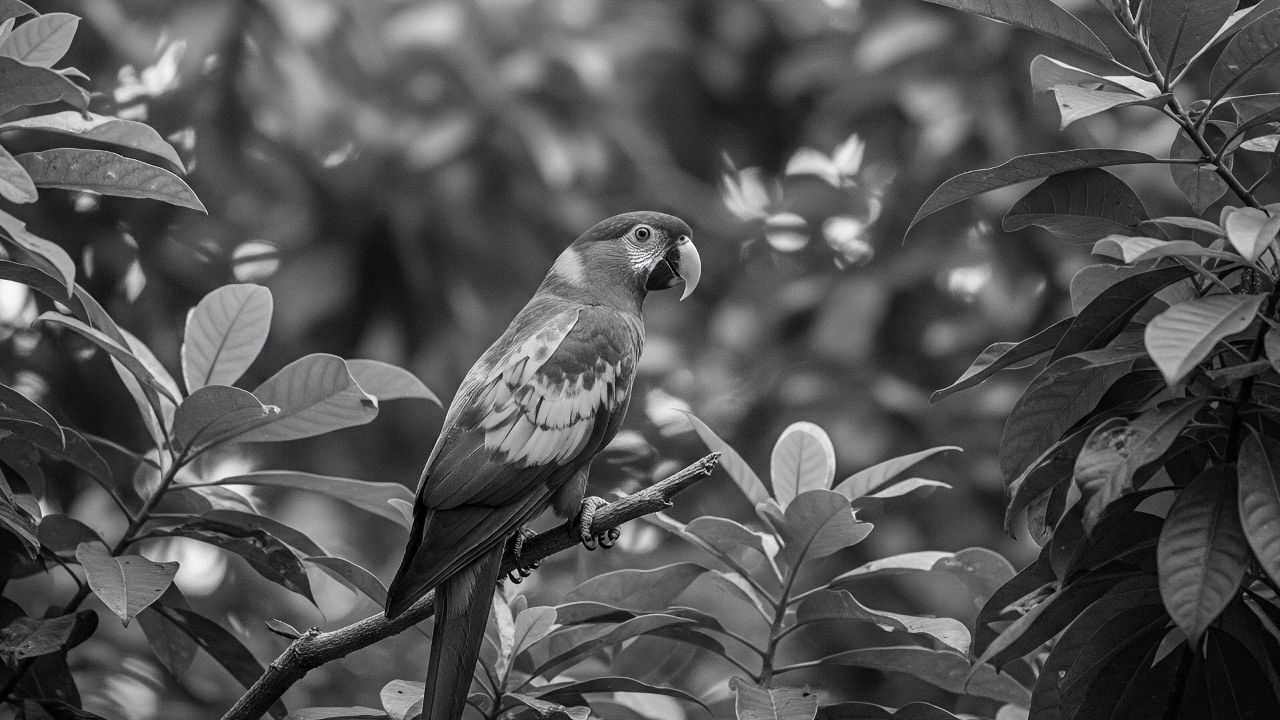}}
    \end{subfigure}
    \hfill
    \begin{subfigure}[t]{\threefigwidth\textwidth}
        \centering
        \fbox{\includegraphics[width=\linewidth]{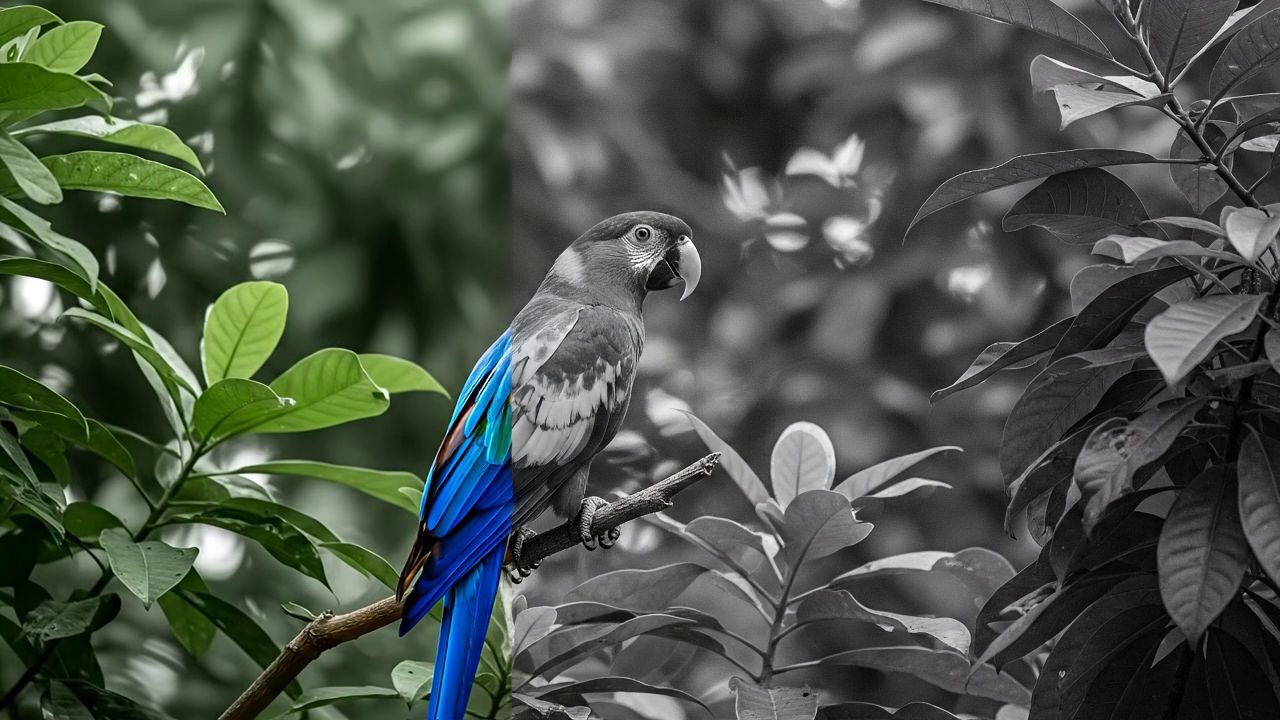}}
    \end{subfigure}
    \hfill    
    \begin{subfigure}[t]{\threefigwidth\textwidth}
        \centering
        \fbox{\includegraphics[width=\linewidth]{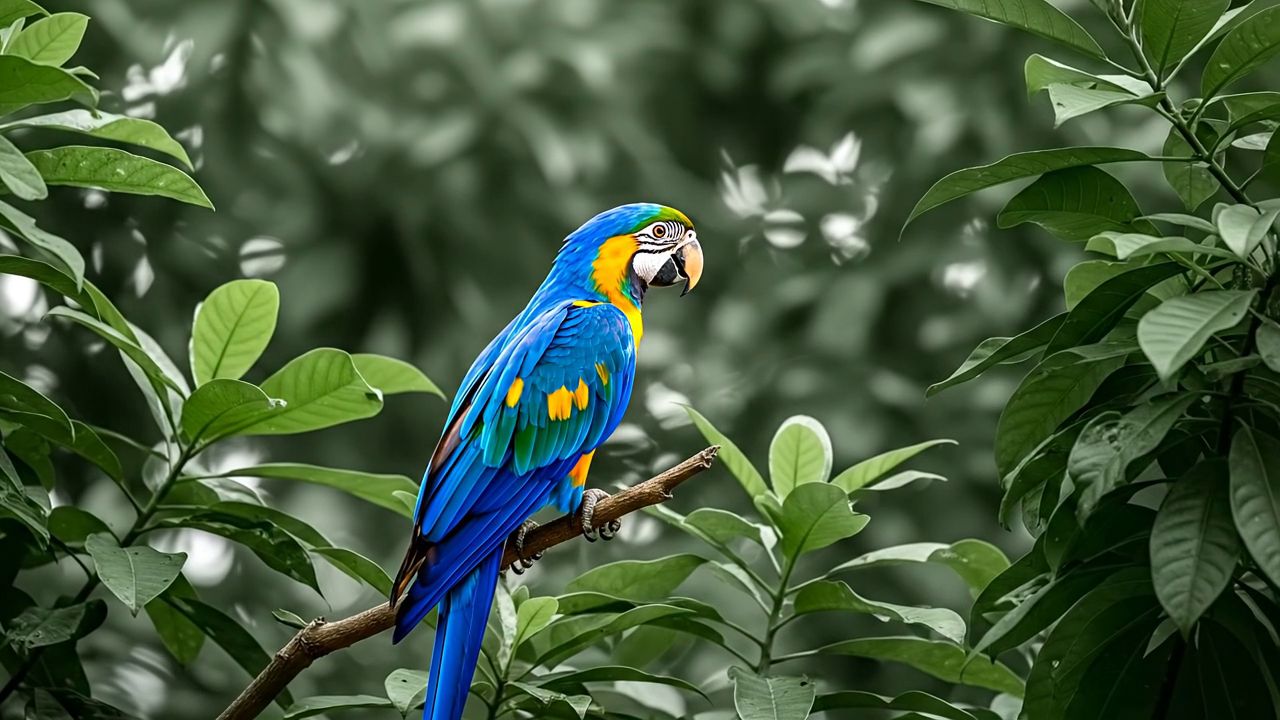}}
    \end{subfigure}
    \caption{\textbf{Colorization}. Prompt: \textit{``Perform colorization on this image. Static camera perspective, no zoom or pan.''} Success rate:~0.08.}
    \label{fig:colorization}
\end{figure}

\begin{figure}[H]
    \centering
    \begin{subfigure}[t]{\threefigwidth\textwidth}
        \centering
        \fbox{\includegraphics[width=\linewidth]{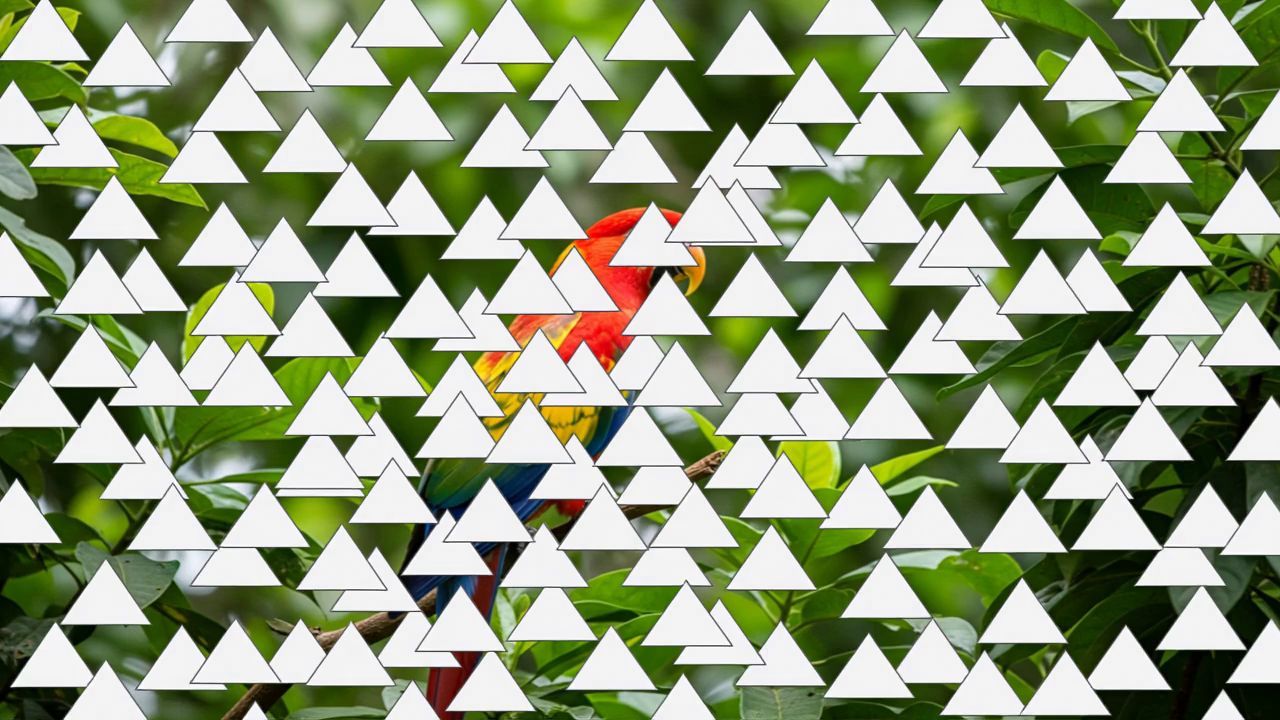}}
    \end{subfigure}
    \hfill
    \begin{subfigure}[t]{\threefigwidth\textwidth}
        \centering
        \fbox{\includegraphics[width=\linewidth]{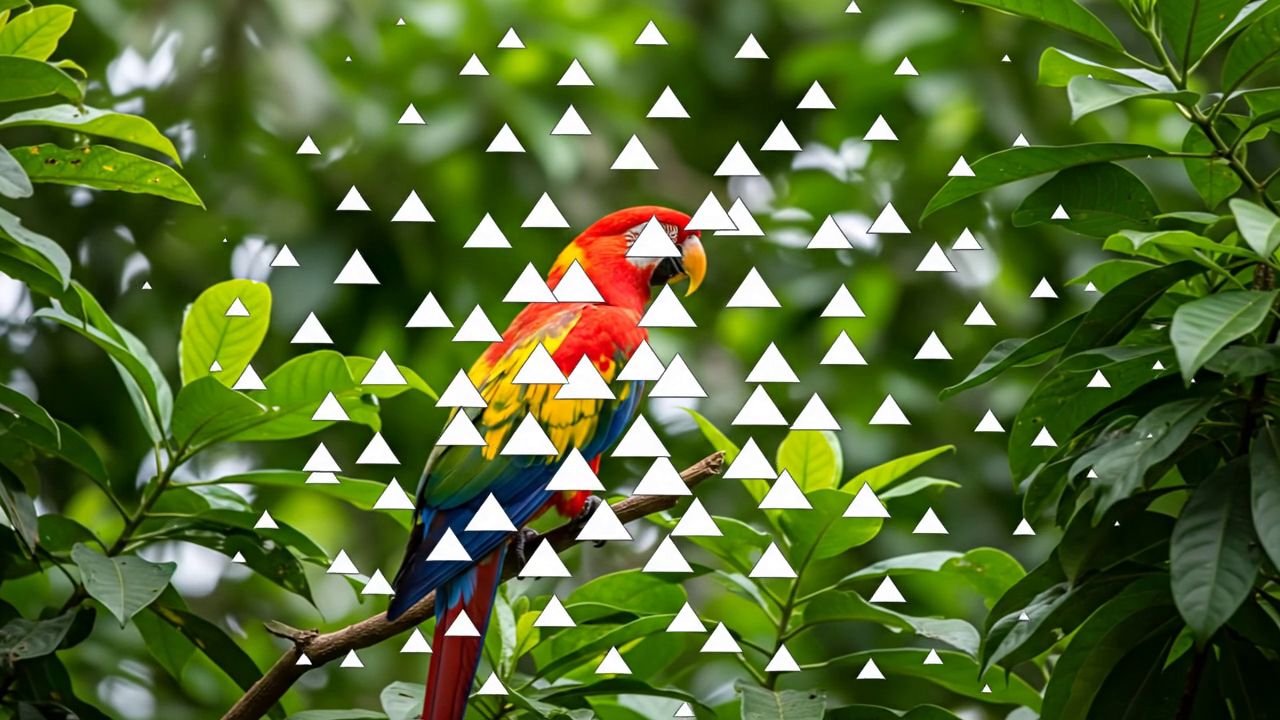}}
    \end{subfigure}
    \hfill    
    \begin{subfigure}[t]{\threefigwidth\textwidth}
        \centering
        \fbox{\includegraphics[width=\linewidth]{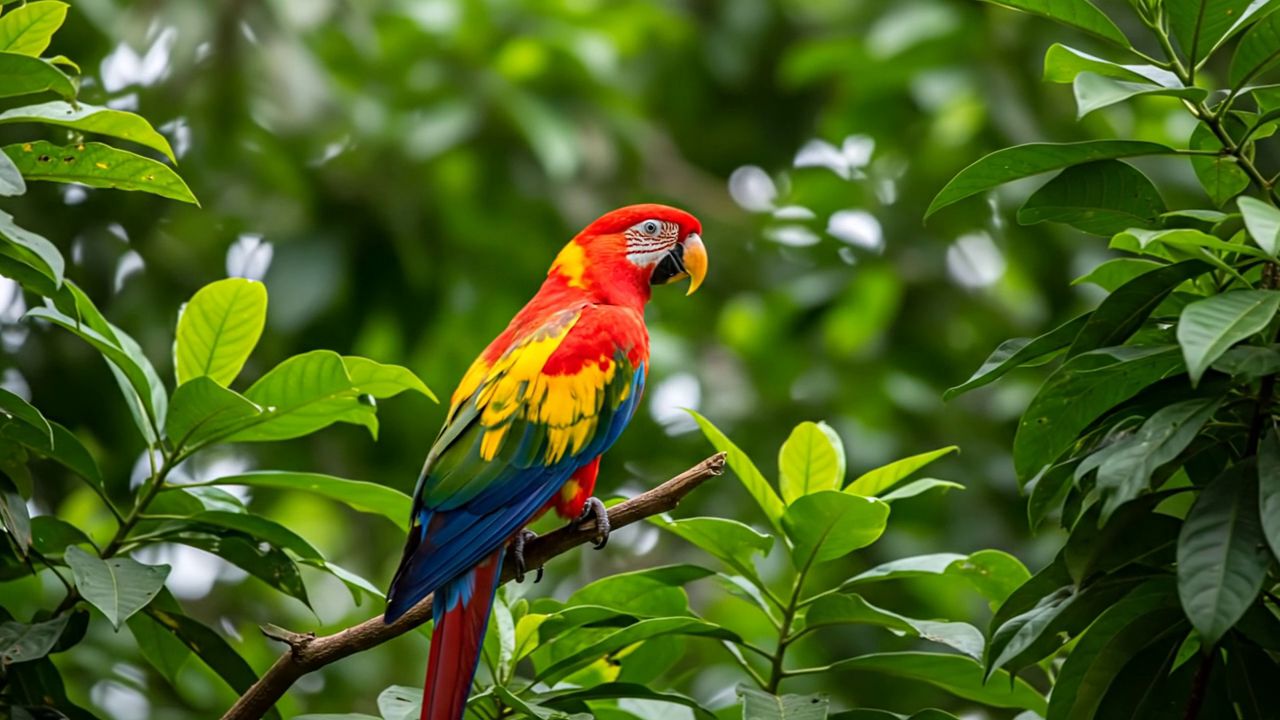}}
    \end{subfigure}
    \caption{\textbf{Inpainting}. Prompt: \textit{``The white triangles become smaller and smaller, then disappear altogether. Static camera perspective, no zoom or pan.''} Success rate:~1.0.}
    \label{fig:inpainting}
\end{figure}

\begin{figure}[H]
    \centering
    \begin{subfigure}[t]{\threefigwidth\textwidth}
        \centering
        \fbox{\includegraphics[width=\linewidth]{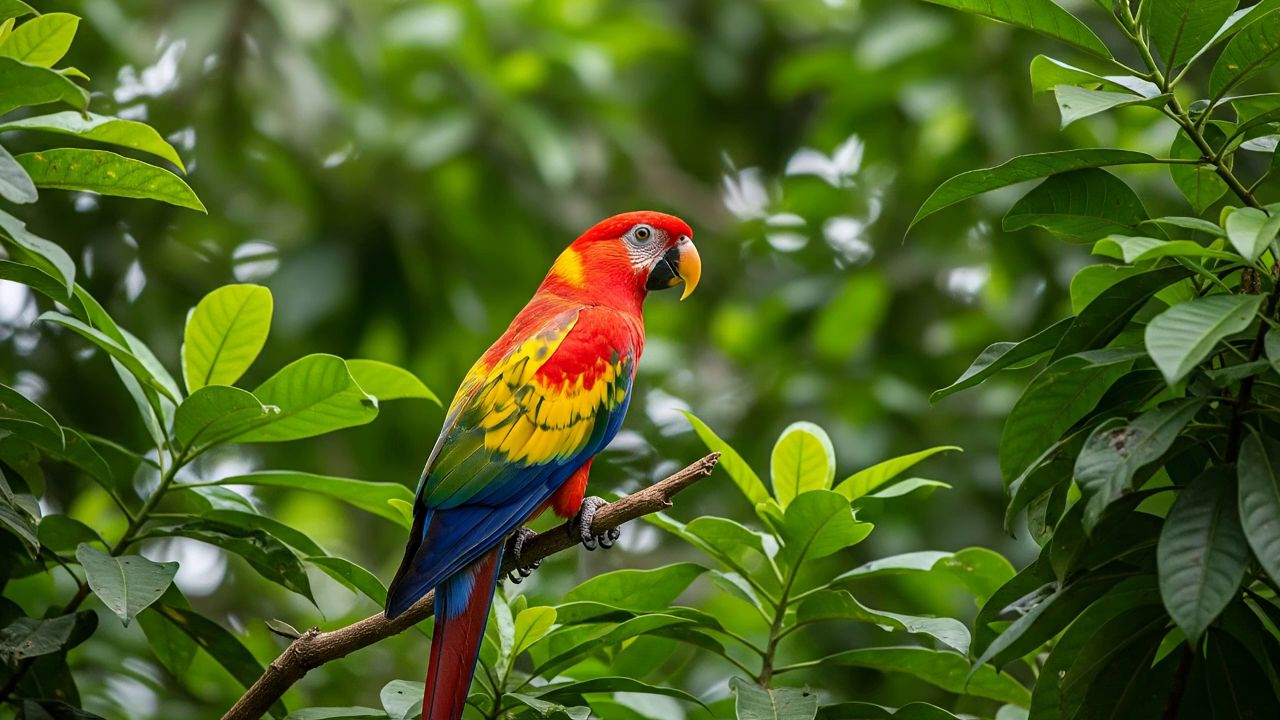}}
    \end{subfigure}
    \hfill
    \begin{subfigure}[t]{\threefigwidth\textwidth}
        \centering
        \fbox{\includegraphics[width=\linewidth]{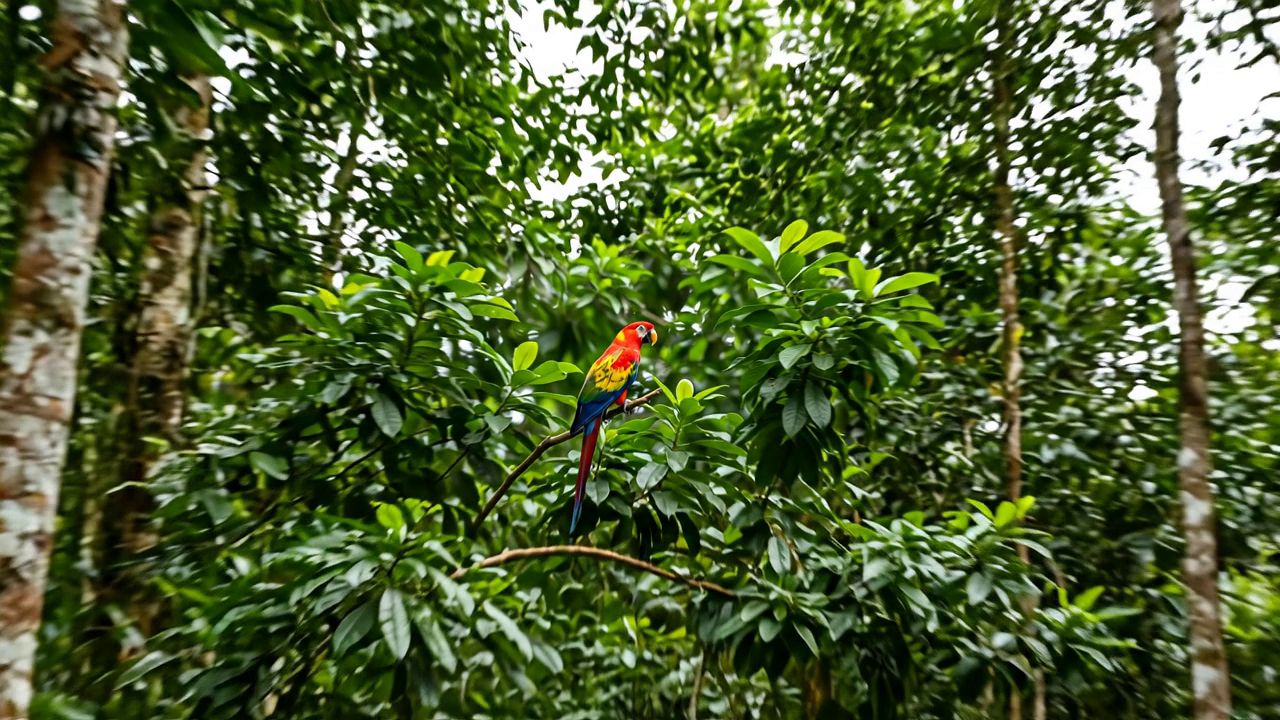}}
    \end{subfigure}
    \hfill    
    \begin{subfigure}[t]{\threefigwidth\textwidth}
        \centering
        \fbox{\includegraphics[width=\linewidth]{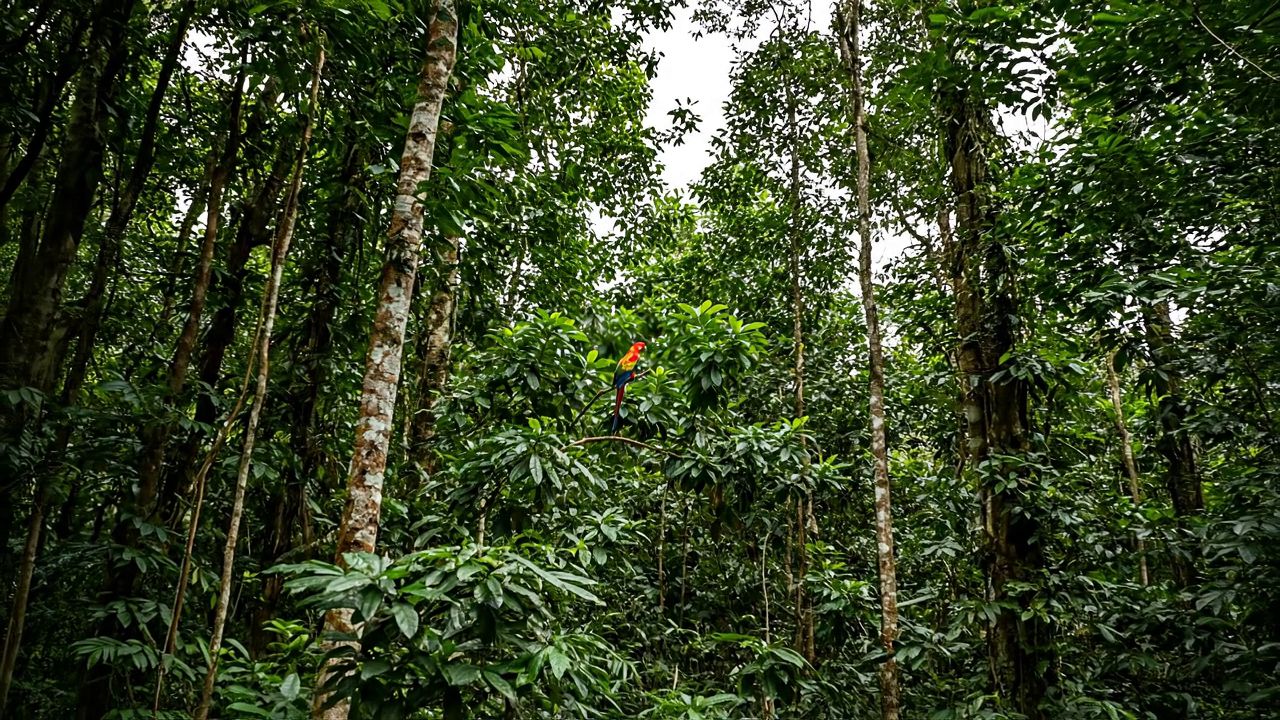}}
    \end{subfigure}
    \caption{\textbf{Outpainting}. Prompt: \textit{``Rapidly zoom out of this static image, revealing what's around it. The camera just zooms back, while the scene itself and everything in it does not move or change at all, it's a static image.''} Success rate:~1.0.}
    \label{fig:outpainting}
\end{figure}

\begin{figure}[H]
    \centering
    \begin{subfigure}[t]{\threefigwidth\textwidth}
        \centering
        \fbox{\includegraphics[width=\linewidth]{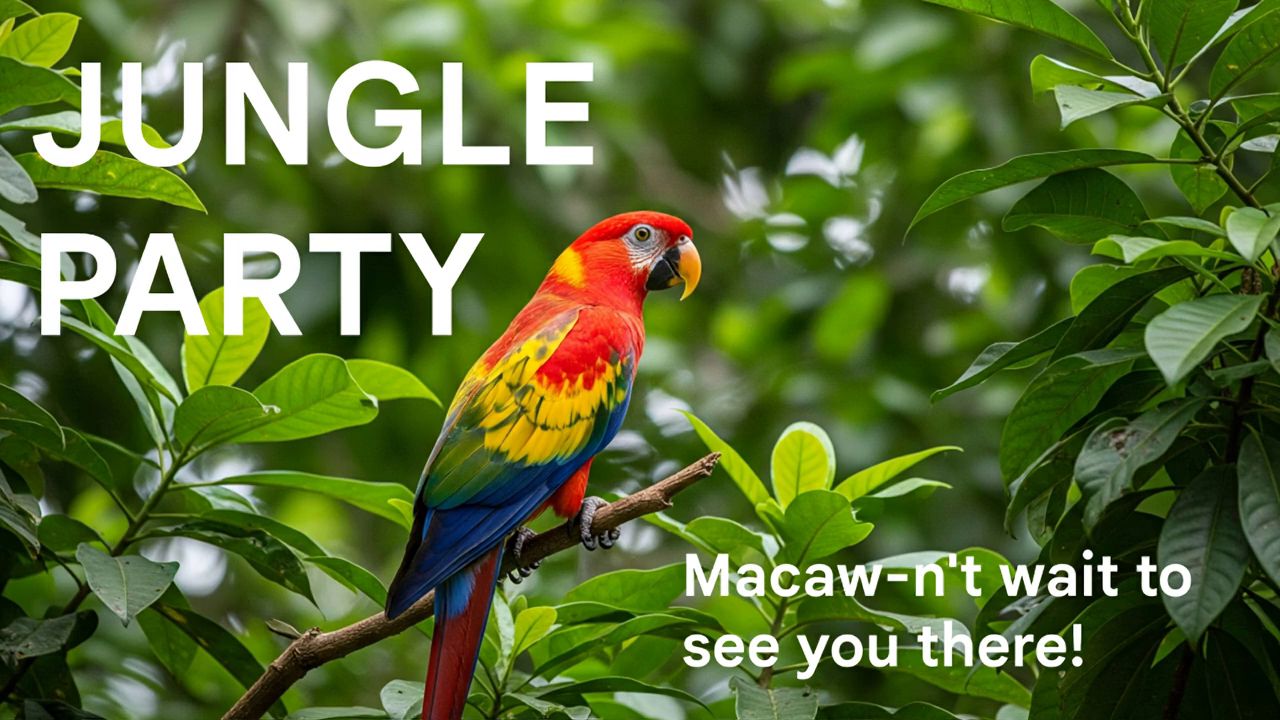}}
    \end{subfigure}
    \hfill
    \begin{subfigure}[t]{\threefigwidth\textwidth}
        \centering
        \fbox{\includegraphics[width=\linewidth]{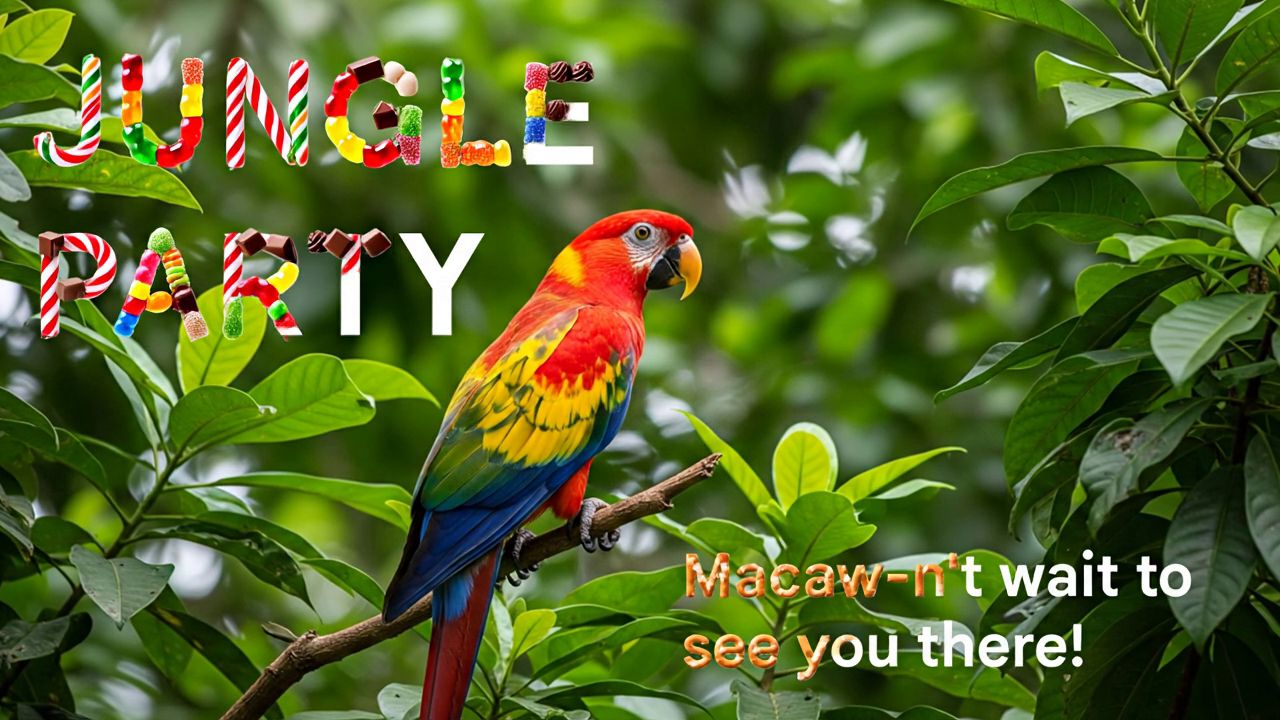}}
    \end{subfigure}
    \hfill
    \begin{subfigure}[t]{\threefigwidth\textwidth}
        \centering
        \fbox{\includegraphics[width=\linewidth]{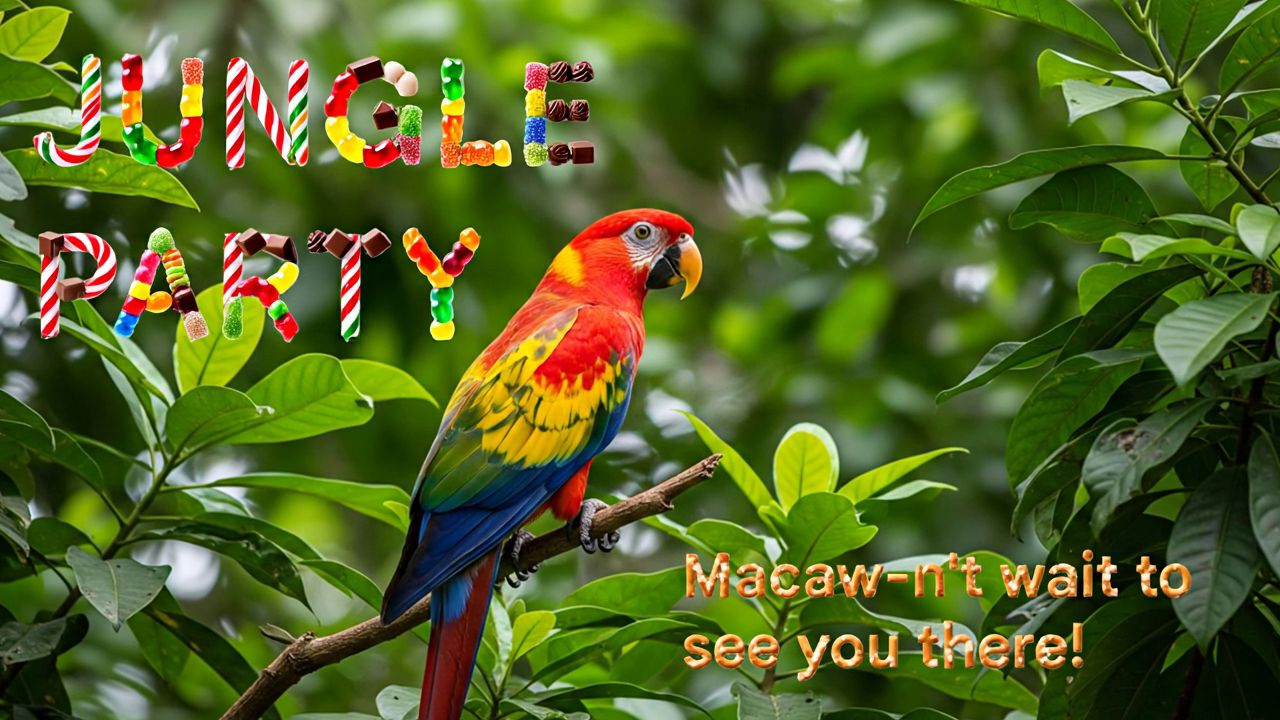}}
    \end{subfigure}
    \caption{\textbf{Text manipulation}. Prompt: \textit{``Animation of the text rapidly changing so that it is made out of different types of candy (top left text) and pretzel sticks (bottom right text). Static camera perspective, no zoom or pan.''} Success rate:~0.33.}
    \label{fig:text_manipulation}
\end{figure}

\begin{figure}[H]
    \centering
    \begin{subfigure}[t]{\threefigwidth\textwidth}
        \centering
        \fbox{\includegraphics[width=\linewidth]{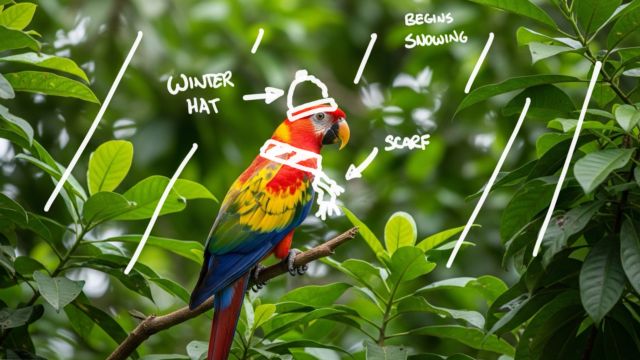}}
    \end{subfigure}
    \hfill
    \begin{subfigure}[t]{\threefigwidth\textwidth}
        \centering
        \fbox{\includegraphics[width=\linewidth]{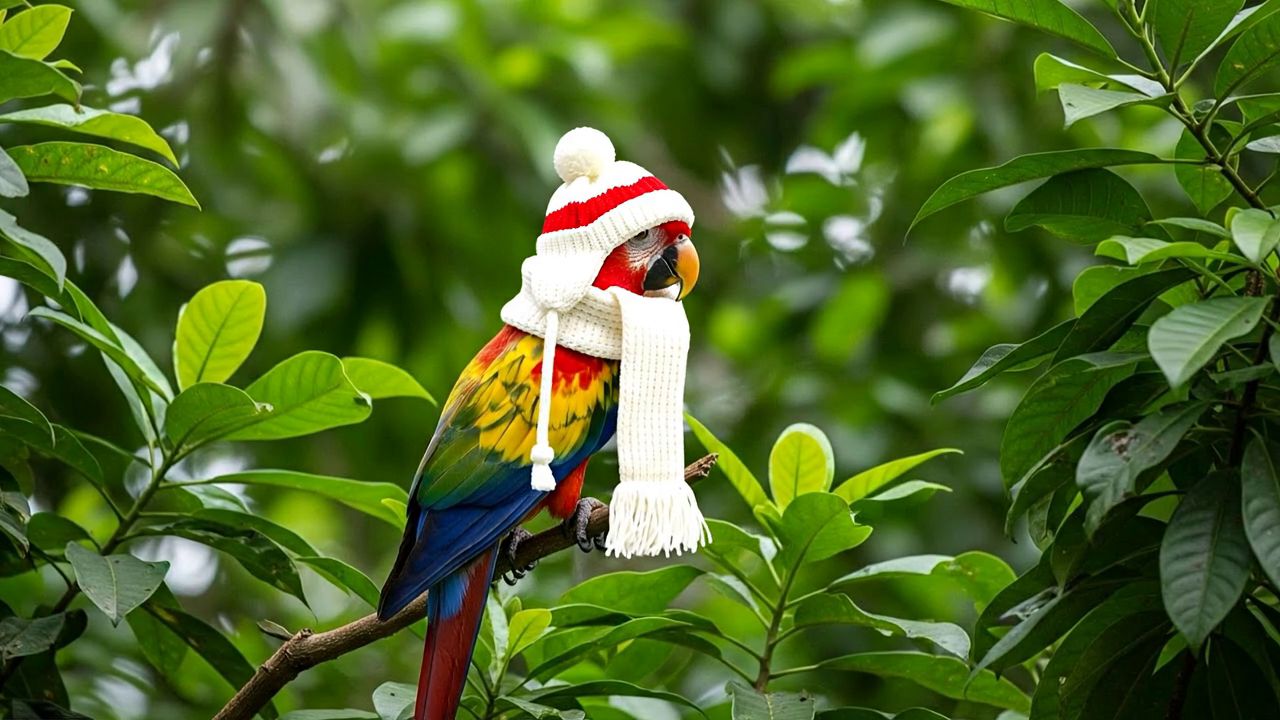}}
    \end{subfigure}
    \hfill
    \begin{subfigure}[t]{\threefigwidth\textwidth}
        \centering
        \fbox{\includegraphics[width=\linewidth]{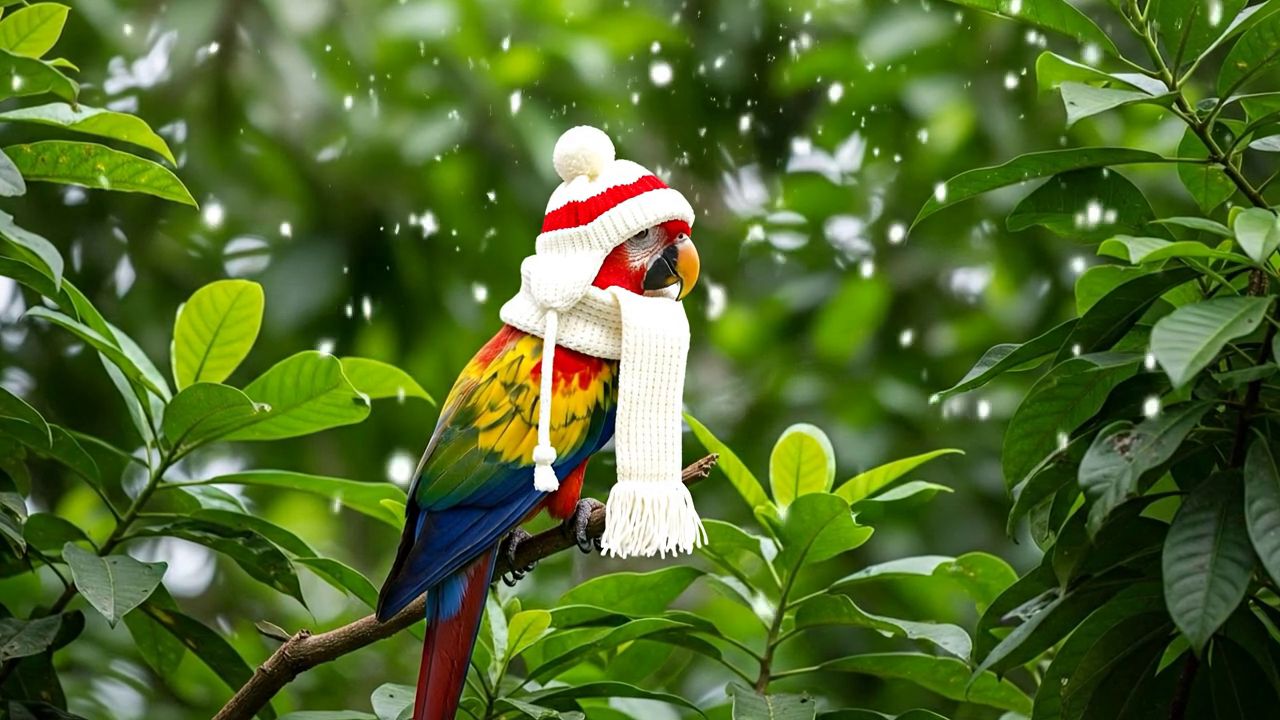}}
    \end{subfigure}
    \caption{\textbf{Image editing with doodles}. Prompt: \textit{``Changes happen instantly.''} Success rate:~1.0.}
    \label{fig:doodle}
\end{figure}

\begin{figure}[H]
    \centering
    \begin{subfigure}[t]{\threefigwidth\textwidth}
        \centering
        \fbox{\includegraphics[width=\linewidth]{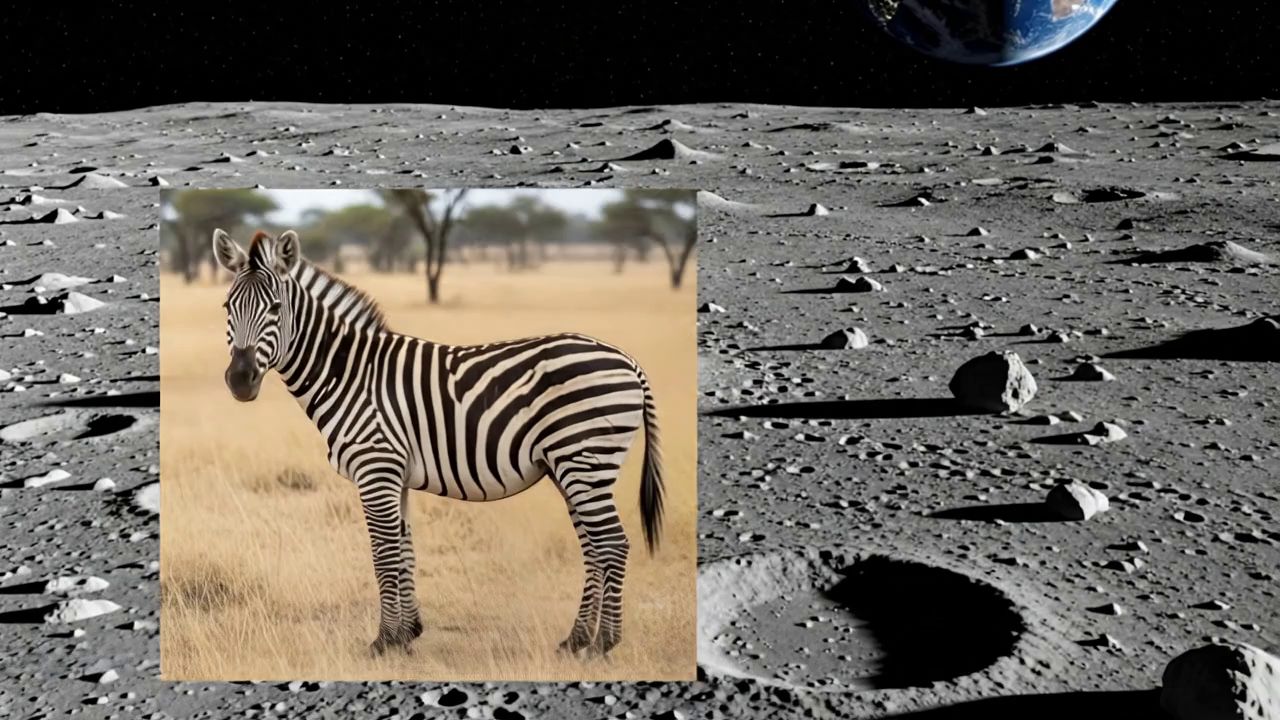}}
    \end{subfigure}
    \hfill
    \begin{subfigure}[t]{\threefigwidth\textwidth}
        \centering
        \fbox{\includegraphics[width=\linewidth]{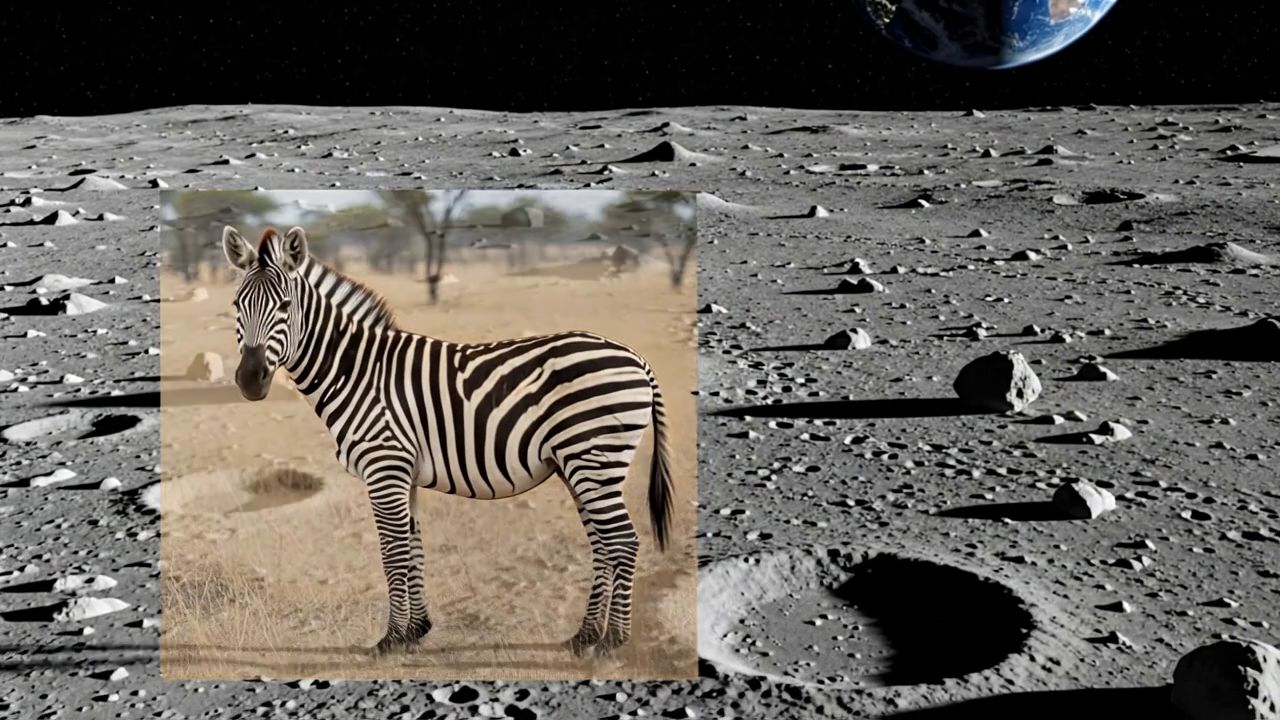}}
    \end{subfigure}
    \hfill    
    \begin{subfigure}[t]{\threefigwidth\textwidth}
        \centering
        \fbox{\includegraphics[width=\linewidth]{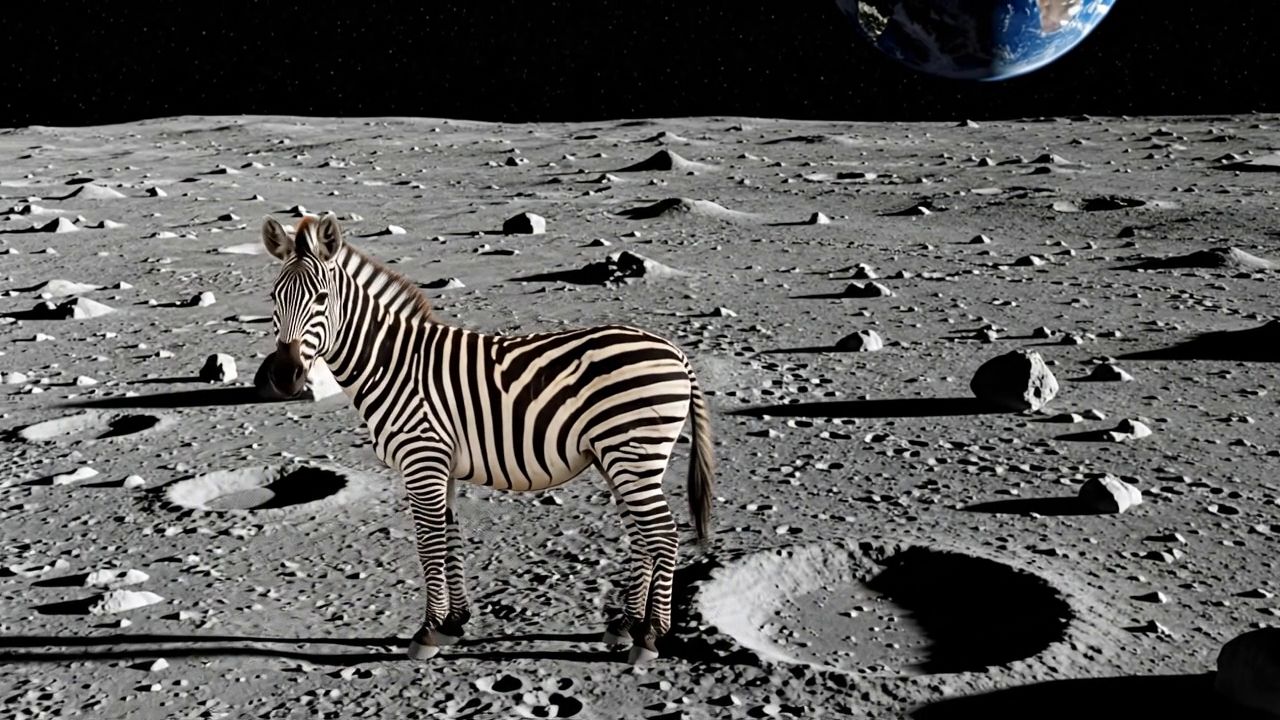}}
    \end{subfigure}
    \caption{\textbf{Scene composition}.
    Prompt: \textit{``A smooth animation blends the zebra naturally into the scene, removing the background of the zebra image, so that the angle, lighting, and shading look realistic. The final scene perfectly incorporates the zebra into the scene.''} Success rate:~0.75.}
    \label{fig:scene_composition}
\end{figure}

\begin{figure}[H]
    \centering
    \begin{subfigure}[t]{\threefigwidth\textwidth}
        \centering
        \fbox{\includegraphics[width=\linewidth]{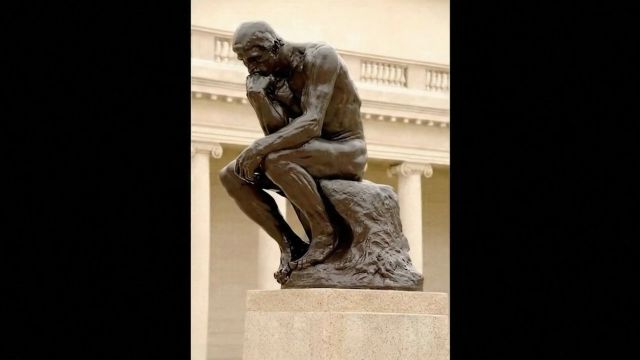}}
    \end{subfigure}
    \hfill
    \begin{subfigure}[t]{\threefigwidth\textwidth}
        \centering
        \fbox{\includegraphics[width=\linewidth]{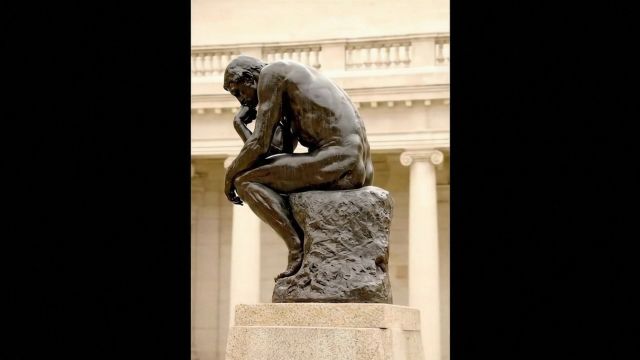}}
    \end{subfigure}
    \hfill
    \begin{subfigure}[t]{\threefigwidth\textwidth}
        \centering
        \fbox{\includegraphics[width=\linewidth]{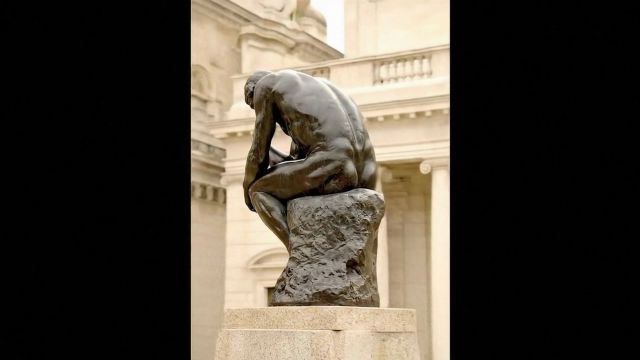}}
    \end{subfigure}
    \caption{\textbf{Single-image novel view synthesis}. Prompt: \textit{``Create a smooth, realistic animation where the camera seems to rotate around the object showing the object from all the sides. Do not change anything else. No zoom. No pan.''} Success rate:~0.92. Image source:  \cite{deitke2023objaverse}.}
    \label{fig:view_synthesis}
\end{figure}

\begin{figure}[H]
    \centering
    \begin{subfigure}[t]{\threefigwidth\textwidth}
        \centering
        \fbox{\includegraphics[width=\linewidth]{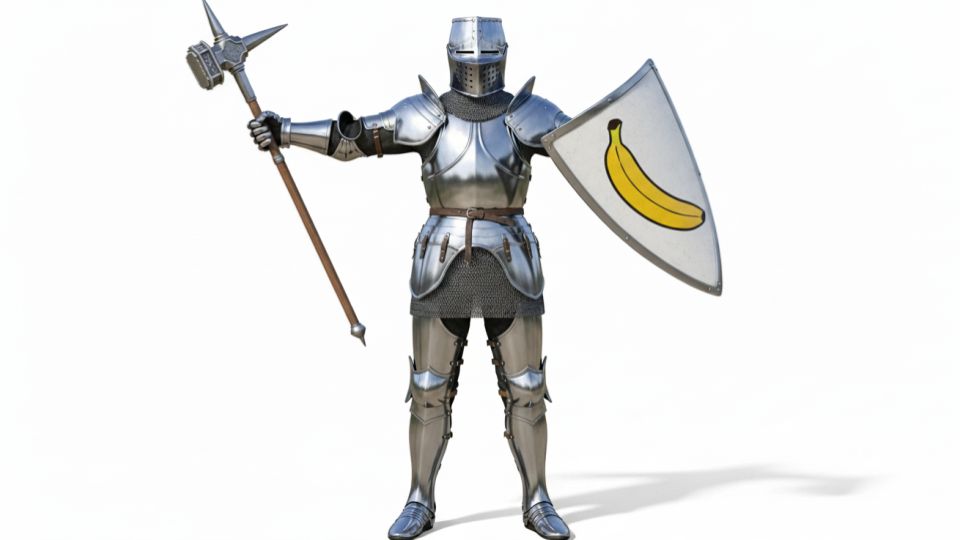}}
    \end{subfigure}
    \hfill
    \begin{subfigure}[t]{\threefigwidth\textwidth}
        \centering
        \fbox{\includegraphics[width=\linewidth]{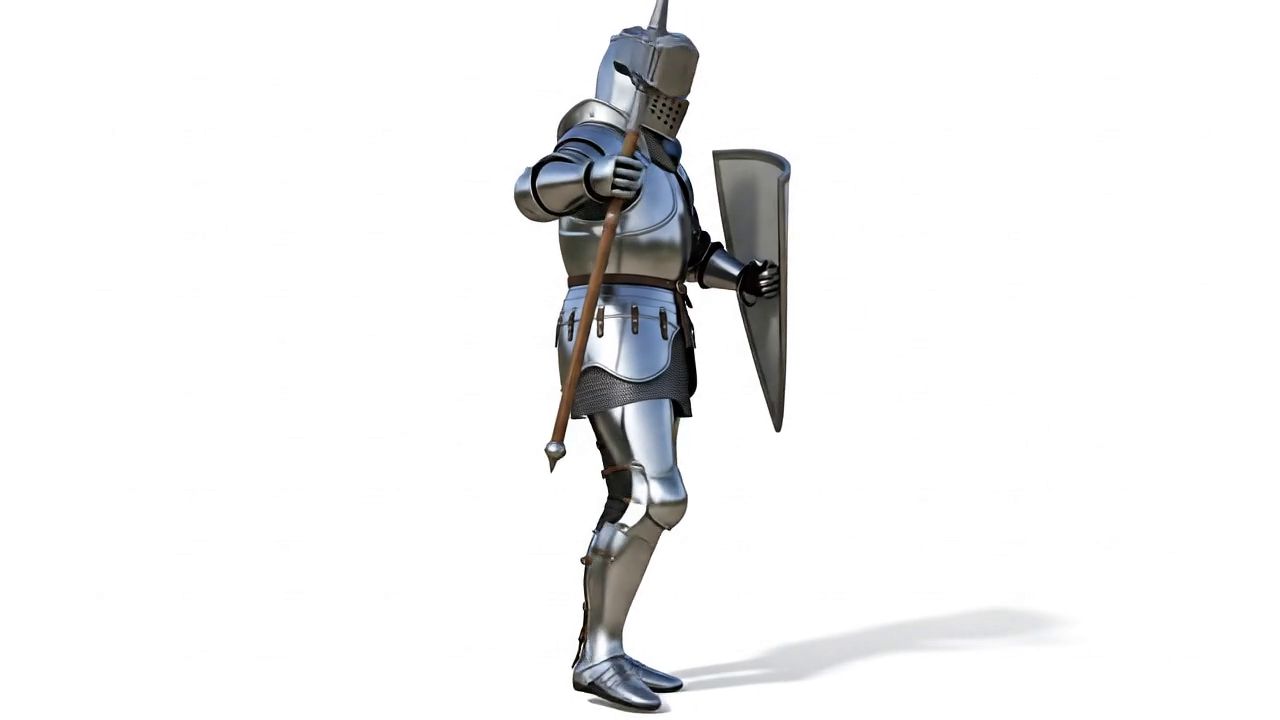}}
    \end{subfigure}
    \hfill
    \begin{subfigure}[t]{\threefigwidth\textwidth}
        \centering
        \fbox{\includegraphics[width=\linewidth]{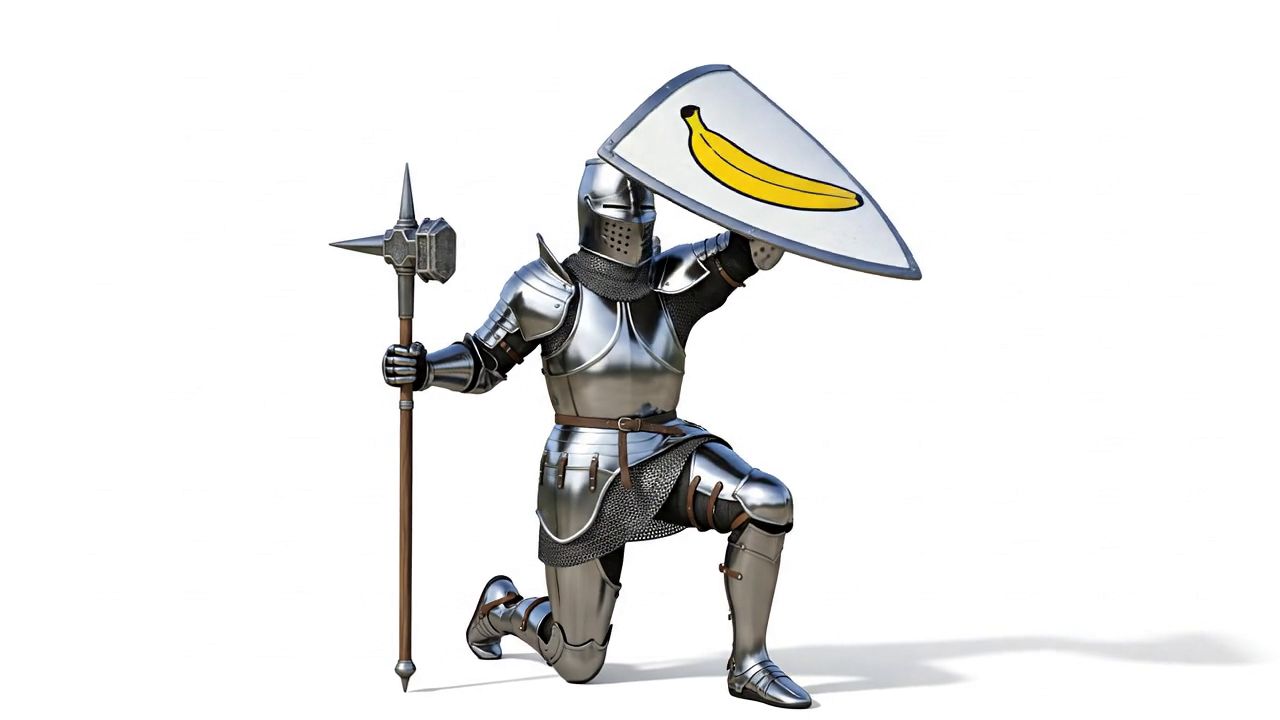}}
    \end{subfigure}
    \caption{\textbf{3D-aware reposing}. Prompt: \textit{``The knight turns to face to the right and drops on one knee, lifting the shield above his head to protect himself and resting the hilt of his weapon on the ground.''} Success rate:~0.83.}
    \label{fig:reposing}
\end{figure}

\begin{figure}[H]
    \centering
    \begin{subfigure}[t]{\threefigwidth\textwidth}
        \centering
        \fbox{\includegraphics[width=\linewidth]{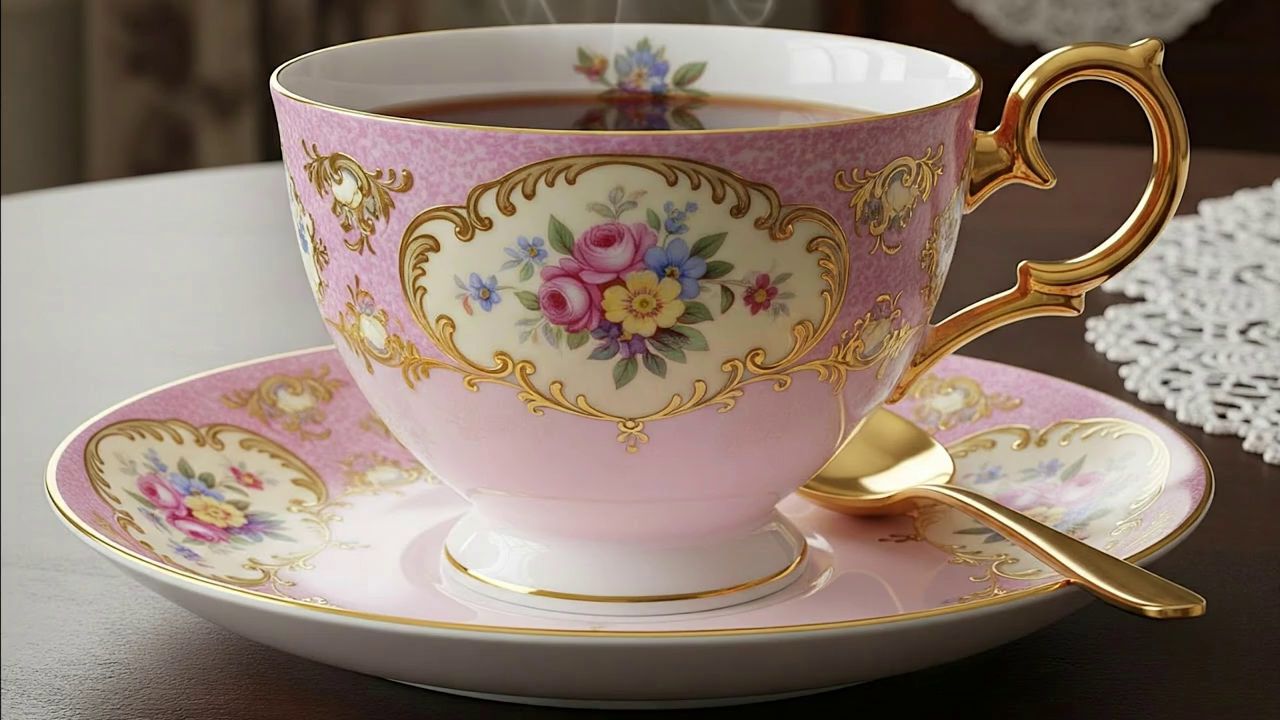}}
    \end{subfigure}
    \hfill
    \begin{subfigure}[t]{\threefigwidth\textwidth}
        \centering
        \fbox{\includegraphics[width=\linewidth]{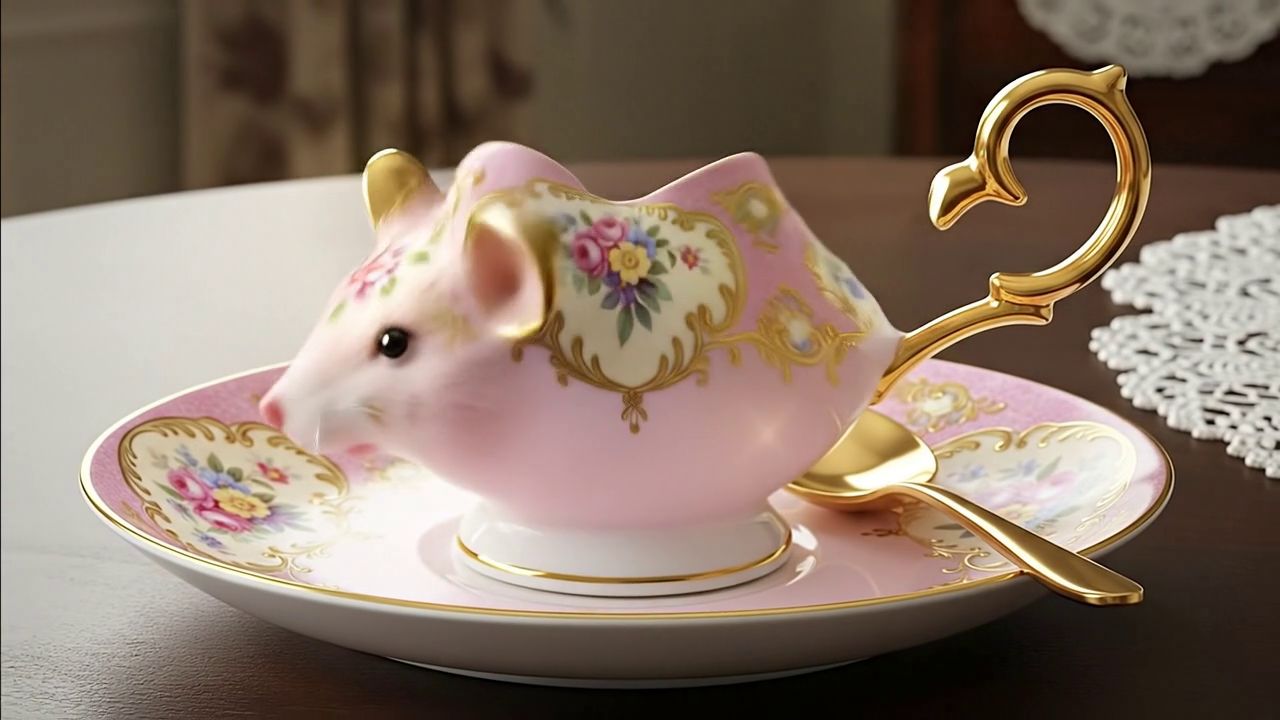}}
    \end{subfigure}
    \hfill
    \begin{subfigure}[t]{\threefigwidth\textwidth}
        \centering
        \fbox{\includegraphics[width=\linewidth]{figures-compressed/transfiguration_Magical_Spell_frame191.jpg}}
    \end{subfigure}
    \caption{\textbf{Transfiguration}.
    Prompt: \textit{``A magical spell smoothly transforms the structure of the teacup into a mouse.''} Success rate:~0.17.}
    \label{fig:transfiguration}
\end{figure}

\begin{figure}[H]
    \centering
    \begin{subfigure}[t]{\threefigwidth\textwidth}
        \centering
        \fbox{\includegraphics[width=\linewidth]{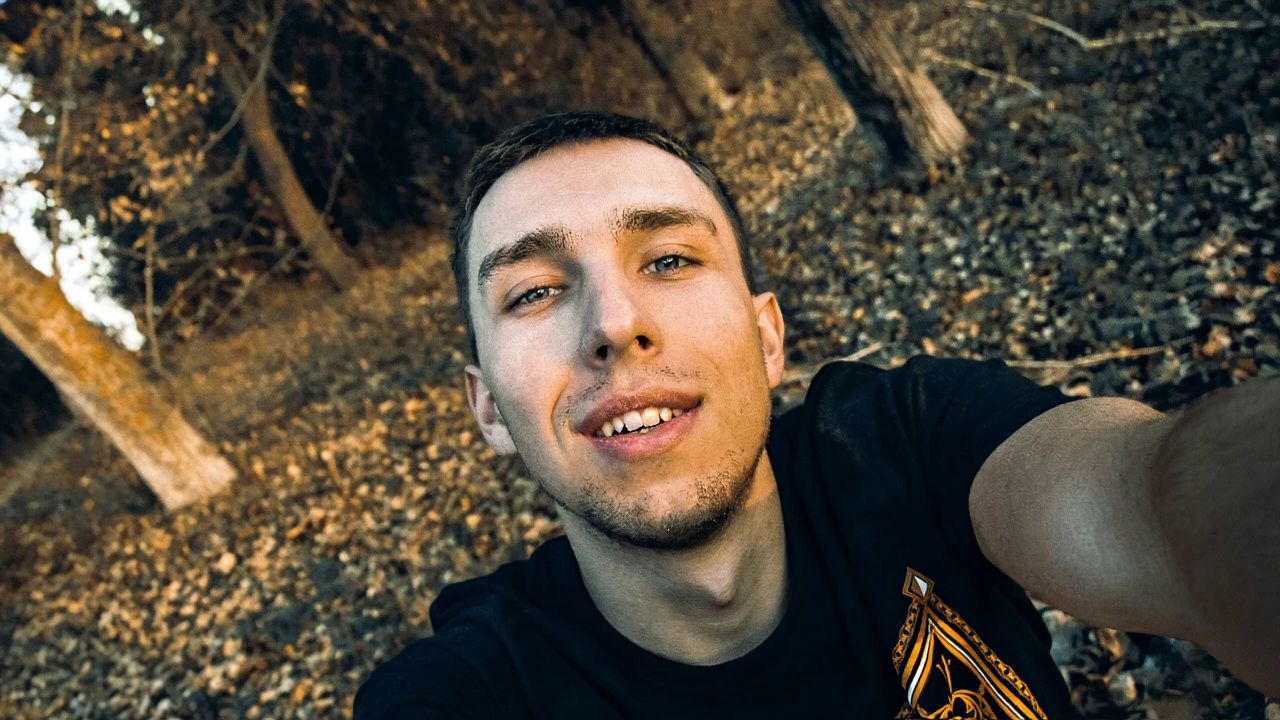}}
    \end{subfigure}
    \hfill
    \begin{subfigure}[t]{\threefigwidth\textwidth}
        \centering
        \fbox{\includegraphics[width=\linewidth]{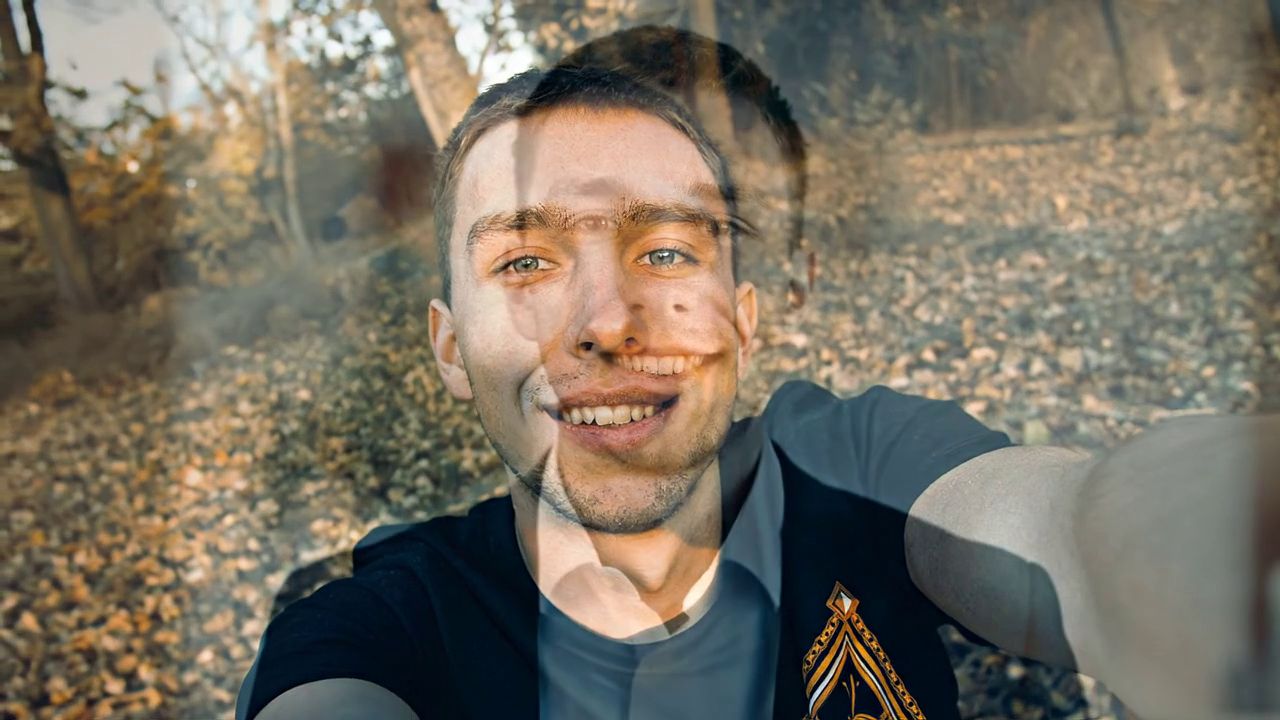}}
    \end{subfigure}
    \hfill    
    \begin{subfigure}[t]{\threefigwidth\textwidth}
        \centering
        \fbox{\includegraphics[width=\linewidth]{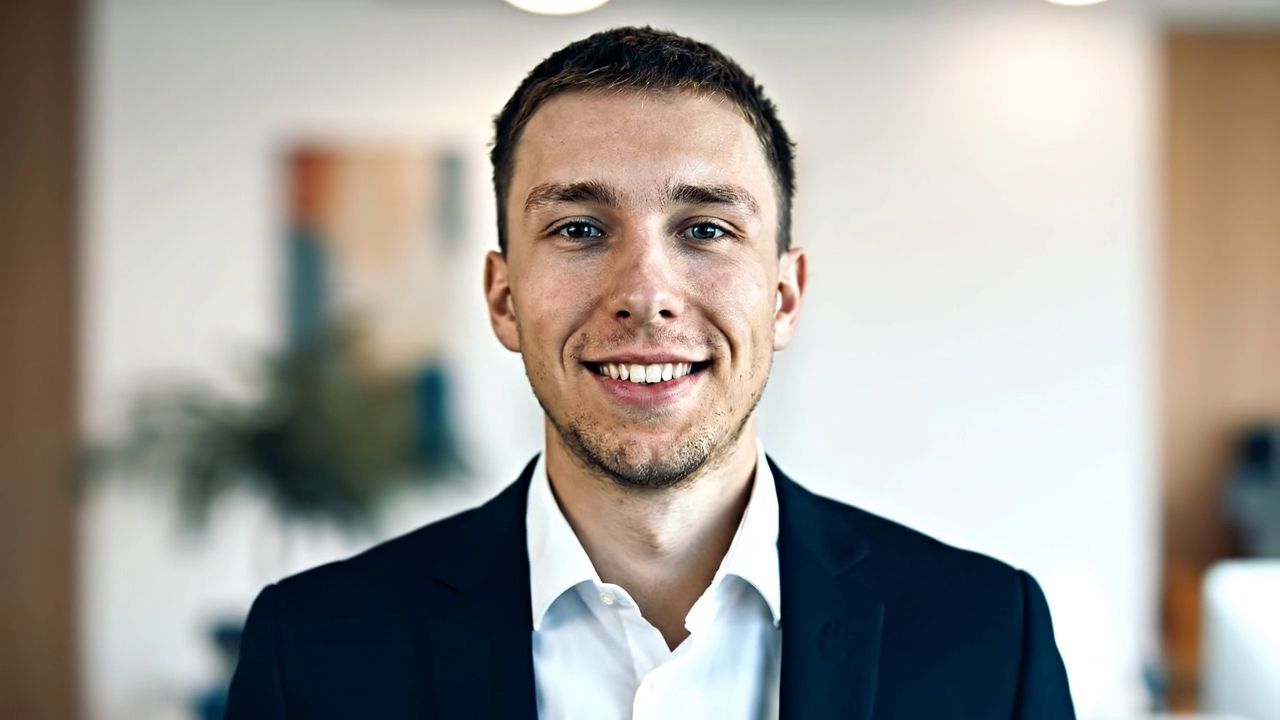}}
    \end{subfigure}
    \caption{\textbf{Professional headshot generation}. Prompt: \textit{``Turn this selfie into a professional headshot for LinkedIn.''} Success rate:~0.42. Image credit: photo by George Pisarevsky on Unsplash. }
    \label{fig:linkedin_headshot}
\end{figure}

\begin{figure}[H]
    \begin{subfigure}[t]{\threefigwidth\textwidth}
        \centering
        \fbox{\includegraphics[width=\linewidth]{figures-compressed/robot-jar_frame_0001.jpg}}
    \end{subfigure}
    \hfill
    \begin{subfigure}[t]{\threefigwidth\textwidth}
        \centering
        \fbox{\includegraphics[width=\linewidth]{figures-compressed/robot-jar_frame_0151.jpg}}
    \end{subfigure}
    \hfill    
    \begin{subfigure}[t]{\threefigwidth\textwidth}
        \centering
        \fbox{\includegraphics[width=\linewidth]{figures-compressed/robot-jar_frame_0192.jpg}}
    \end{subfigure}

    \vspace{0.2cm}
    
    \begin{subfigure}[t]{\threefigwidth\textwidth}
        \centering
        \fbox{\includegraphics[width=\linewidth]{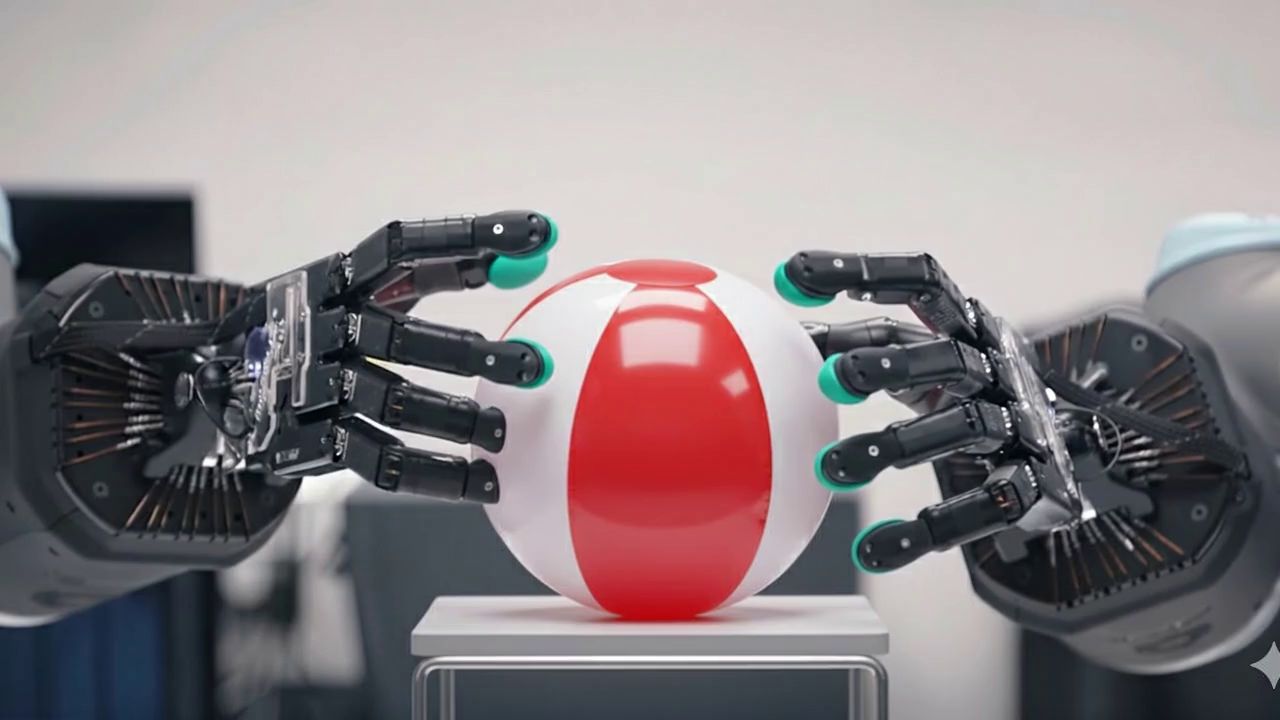}}
    \end{subfigure}
    \hfill
    \begin{subfigure}[t]{\threefigwidth\textwidth}
        \centering
        \fbox{\includegraphics[width=\linewidth]{figures-compressed/robot-ball_frame_0052.jpg}}
    \end{subfigure}
    \hfill    
    \begin{subfigure}[t]{\threefigwidth\textwidth}
        \centering
        \fbox{\includegraphics[width=\linewidth]{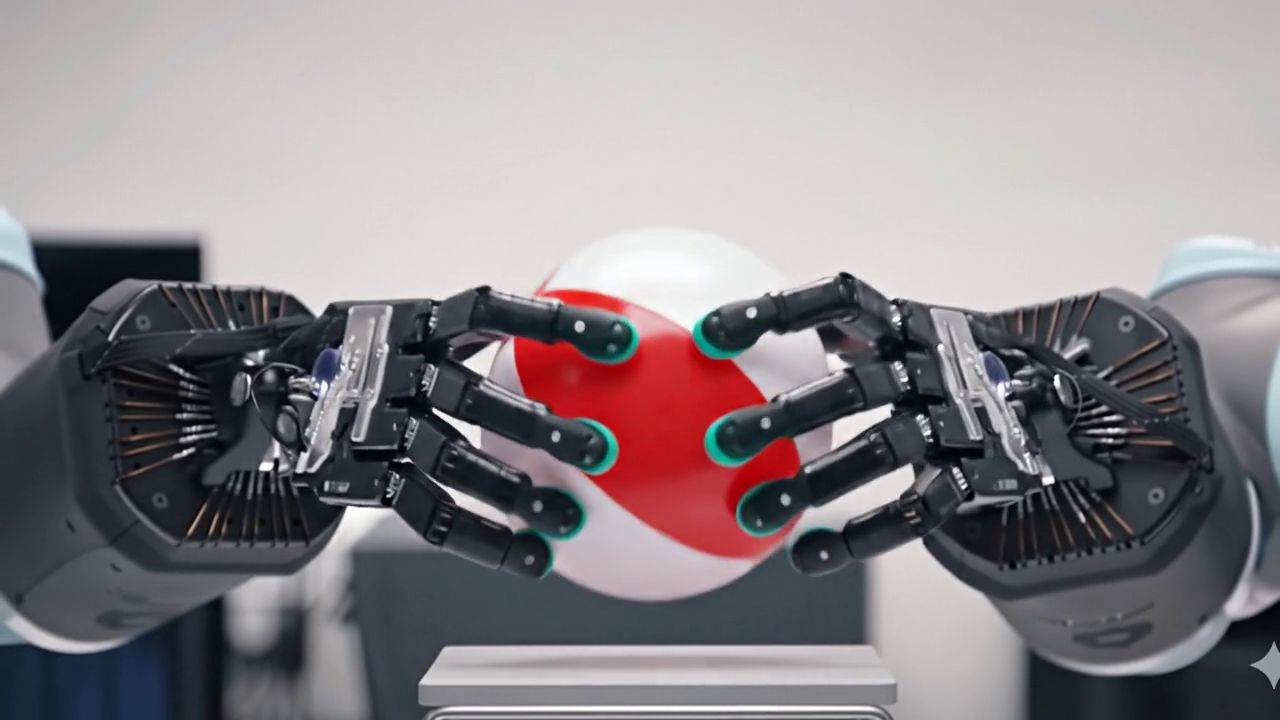}}
    \end{subfigure}

    \vspace{0.2cm}

    \centering
    \begin{subfigure}[t]{0.195\textwidth}
        \centering
        \fbox{\includegraphics[width=\linewidth]{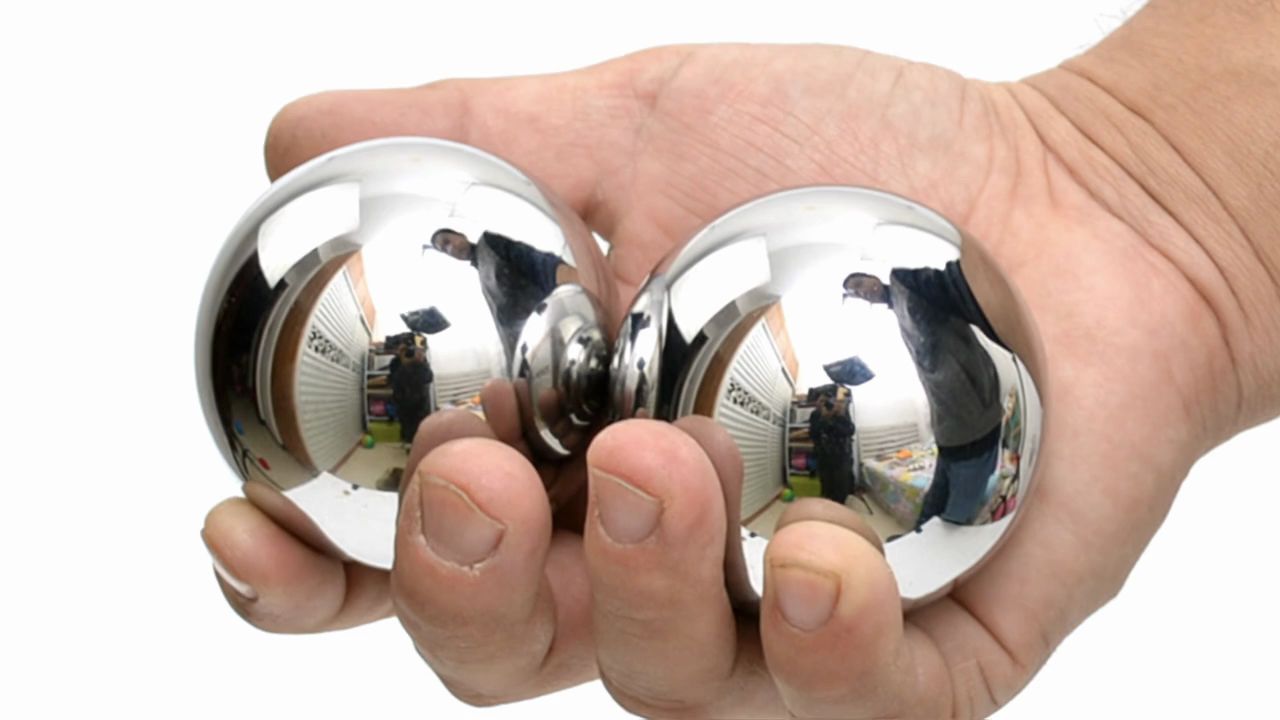}}
    \end{subfigure}
    \hfill
    \begin{subfigure}[t]{0.195\textwidth}
        \centering
        \fbox{\includegraphics[width=\linewidth]{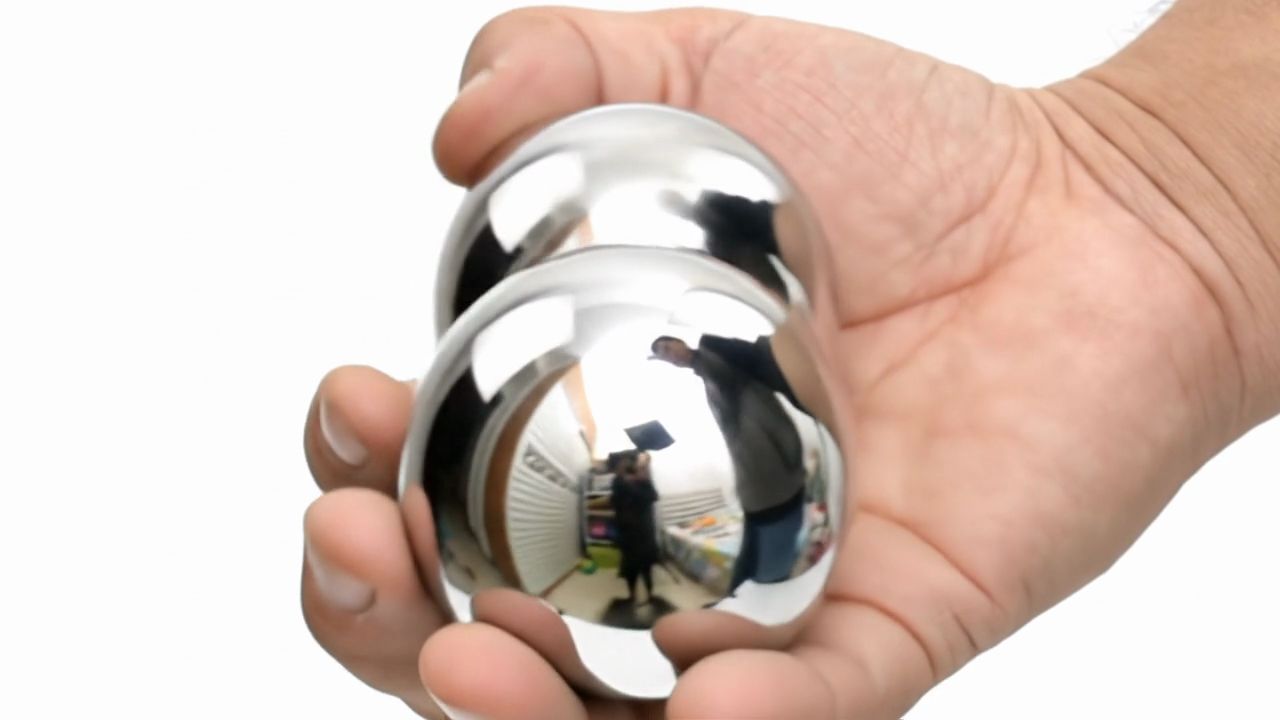}}
    \end{subfigure}
    \hfill
    \begin{subfigure}[t]{0.195\textwidth}
        \centering
        \fbox{\includegraphics[width=\linewidth]{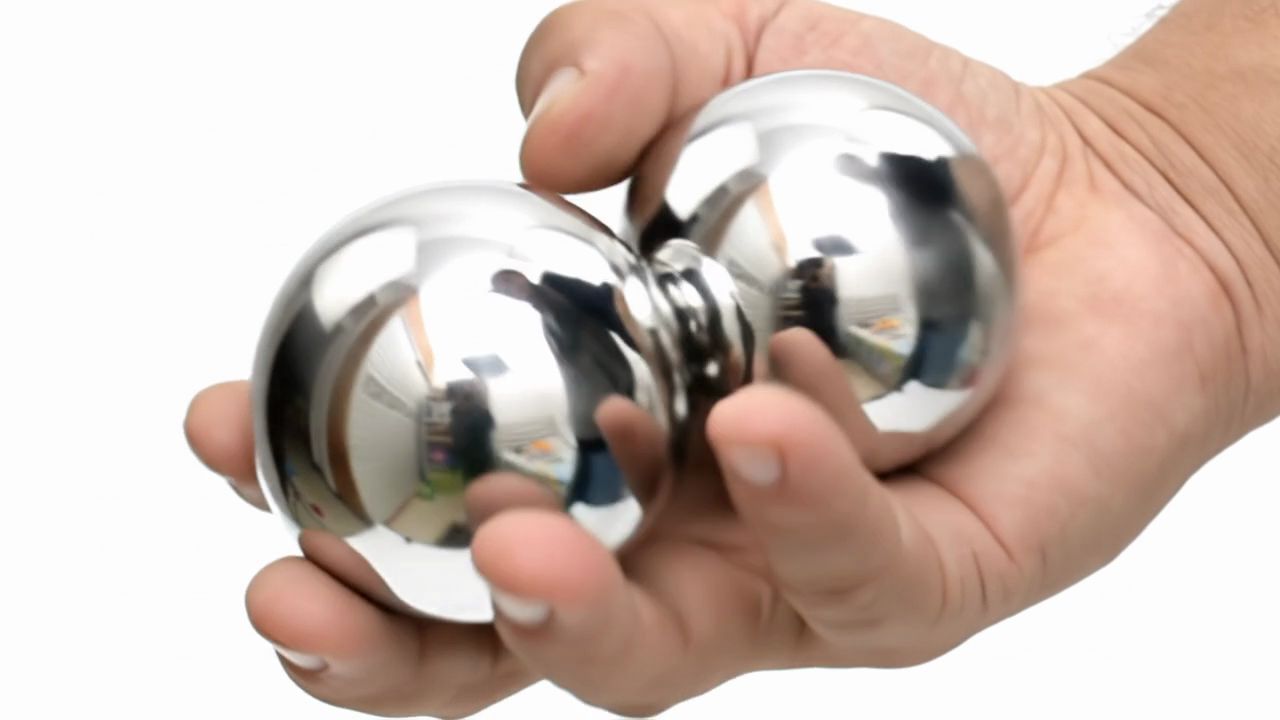}}
    \end{subfigure}
    \begin{subfigure}[t]{0.195\textwidth}
        \centering
        \fbox{\includegraphics[width=\linewidth]{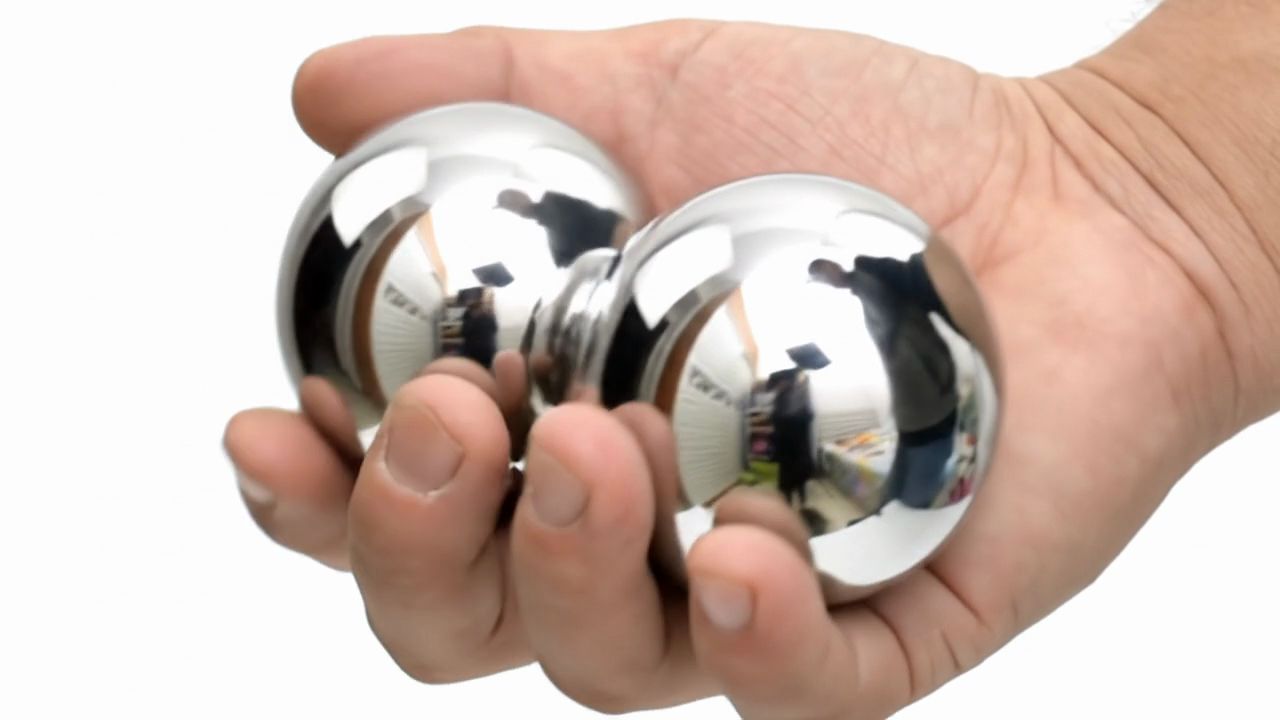}}
    \end{subfigure}
    \begin{subfigure}[t]{0.195\textwidth}
        \centering
        \fbox{\includegraphics[width=\linewidth]{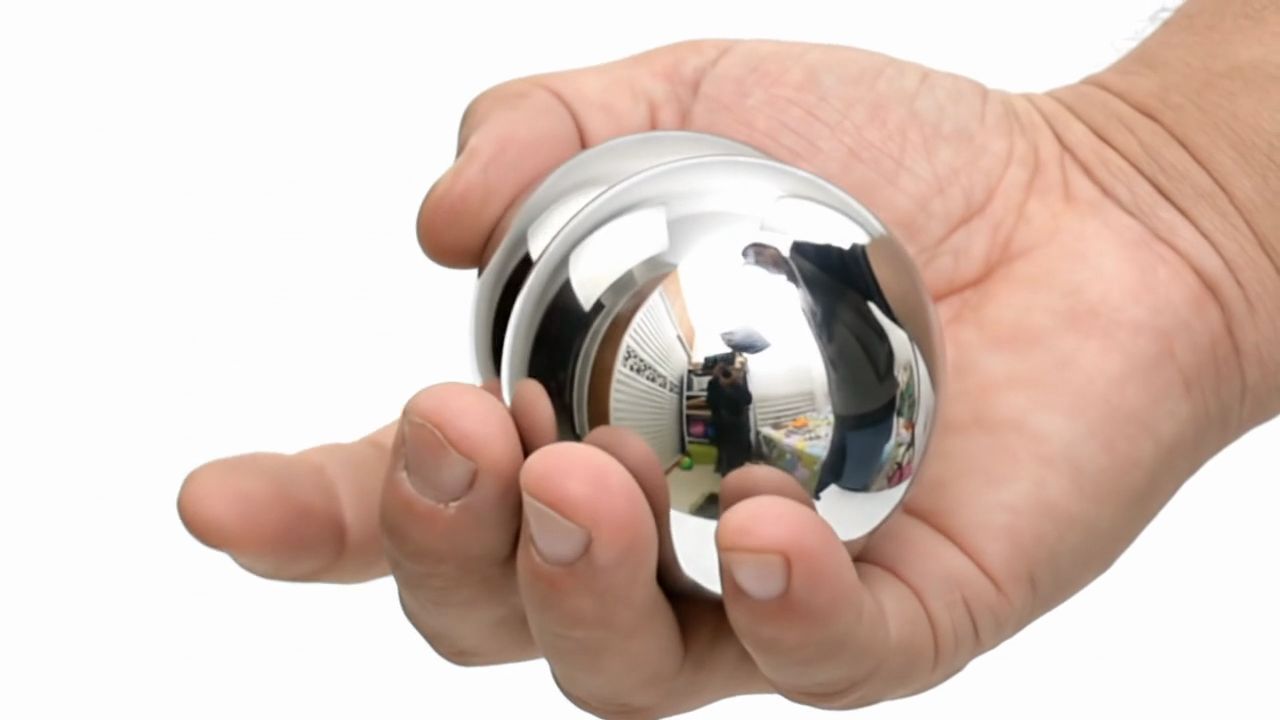}}
    \end{subfigure}
    \caption{\textbf{Dexterous manipulation}.
    \textbf{Jar opening} (top). Prompt: \textit{``Use common sense and have the two robot hands attached to robot arms open the jar, like how a human would.''} Success rate:~1.0.
    \textbf{Throwing and catching} (middle). Prompt: \textit{``Use common sense and have the two robot hands attached to robot arms throw the ball in the air, the ball goes up off the screen, hands move to positions to catch the ball, and catch the falling ball, like how a human would.''} Success rate:~1.0.
    \textbf{Rotating Baoding balls} (bottom). Prompt: \textit{``A human hand holds two metal Baoding balls. The fingers, including the thumb, index, and middle finger, skillfully manipulate the balls, causing them to rotate smoothly like two planets orbiting around each other and continuously in the palm, one ball circling the other in a fluid motion.''} Success rate:~0.08.}
    \label{fig:hand-dexterity}
\end{figure}

\begin{figure}[H]
    \centering
    \begin{subfigure}[t]{\threefigwidth\textwidth}
        \centering
        \fbox{\includegraphics[width=\linewidth]{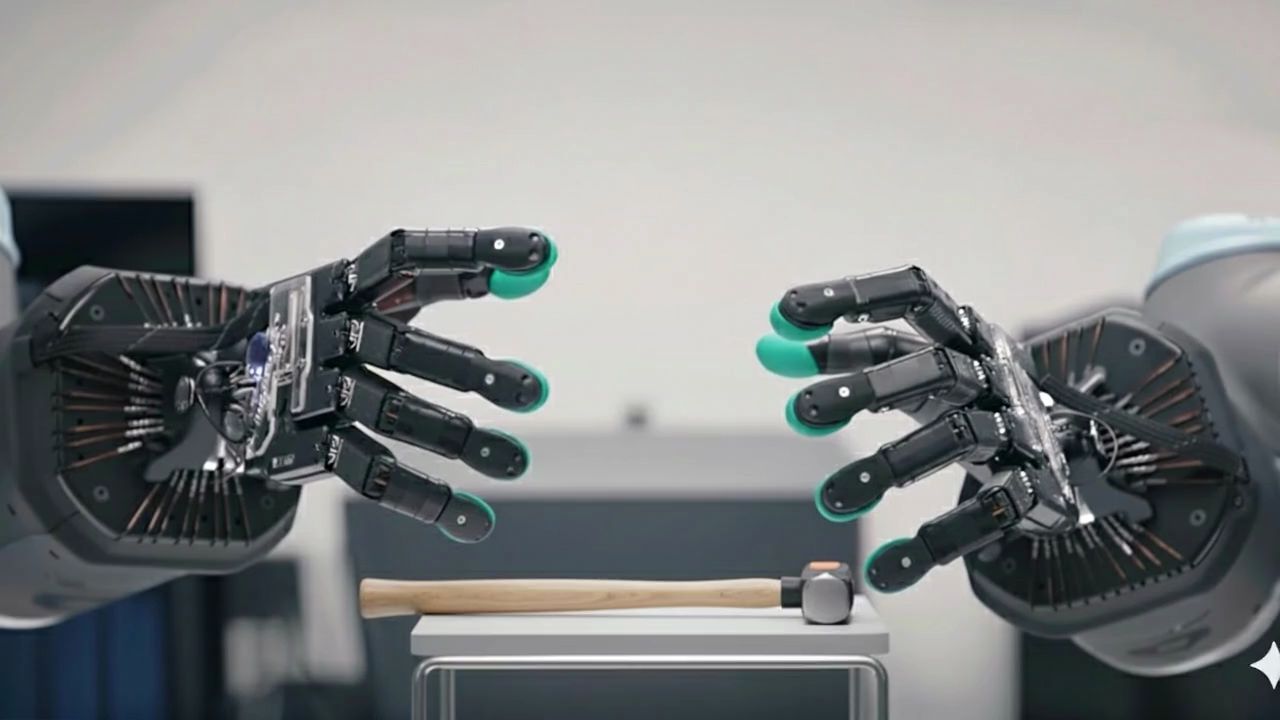}}
    \end{subfigure}
    \hfill
    \begin{subfigure}[t]{\threefigwidth\textwidth}
        \centering
        \fbox{\includegraphics[width=\linewidth]{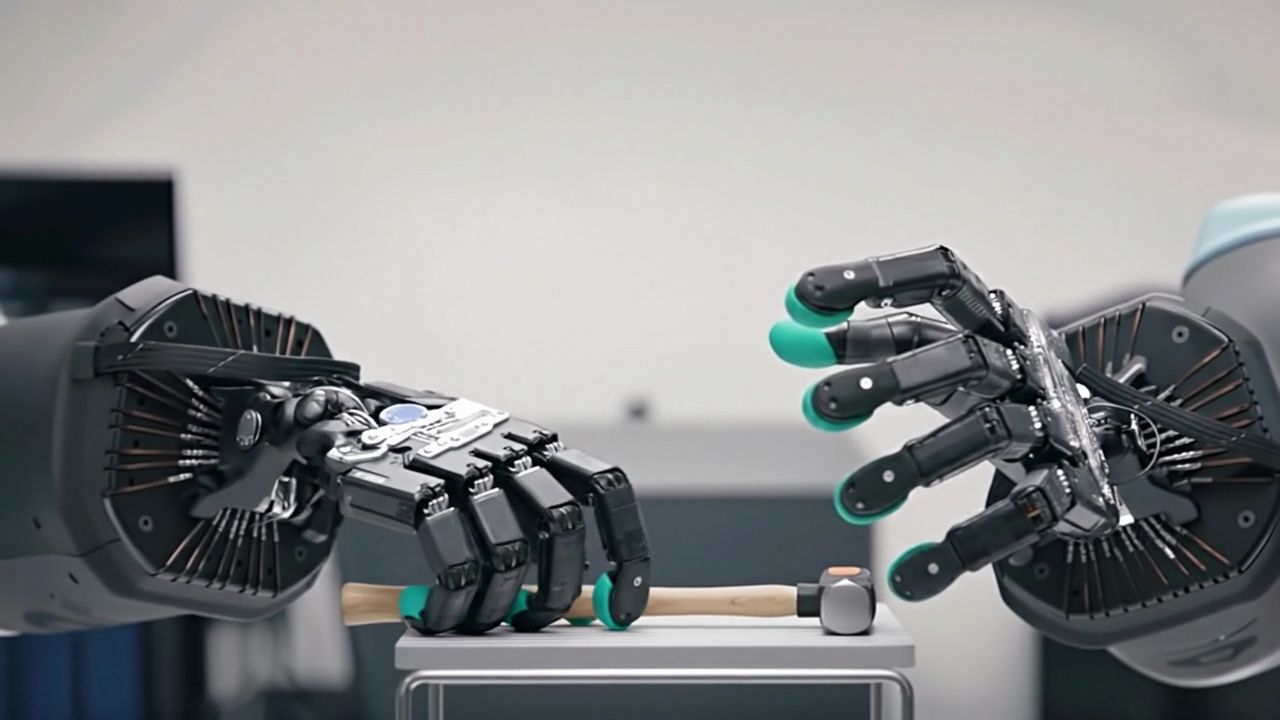}}
    \end{subfigure}
    \hfill    
    \begin{subfigure}[t]{\threefigwidth\textwidth}
        \centering
        \fbox{\includegraphics[width=\linewidth]{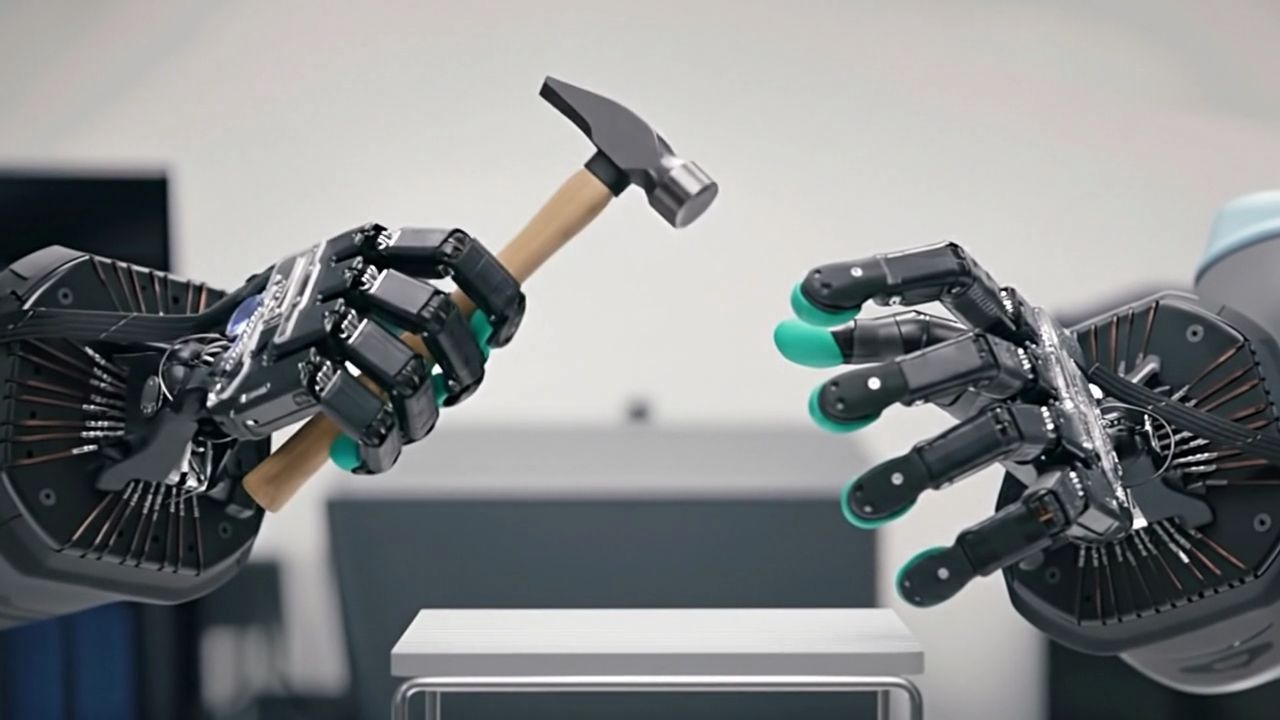}}
    \end{subfigure}
    \caption{\textbf{Affordance recognition}. Prompt: \textit{``The robot hands mounted on robot arms pick up the hammer, naturally like how a human would.''} Success rate:~0.5.}
    \label{fig:affordance}
\end{figure}

\begin{figure}[H]
    \centering
    \begin{subfigure}[t]{\threefigwidth\textwidth}
        \centering
        \fbox{\includegraphics[width=\linewidth]{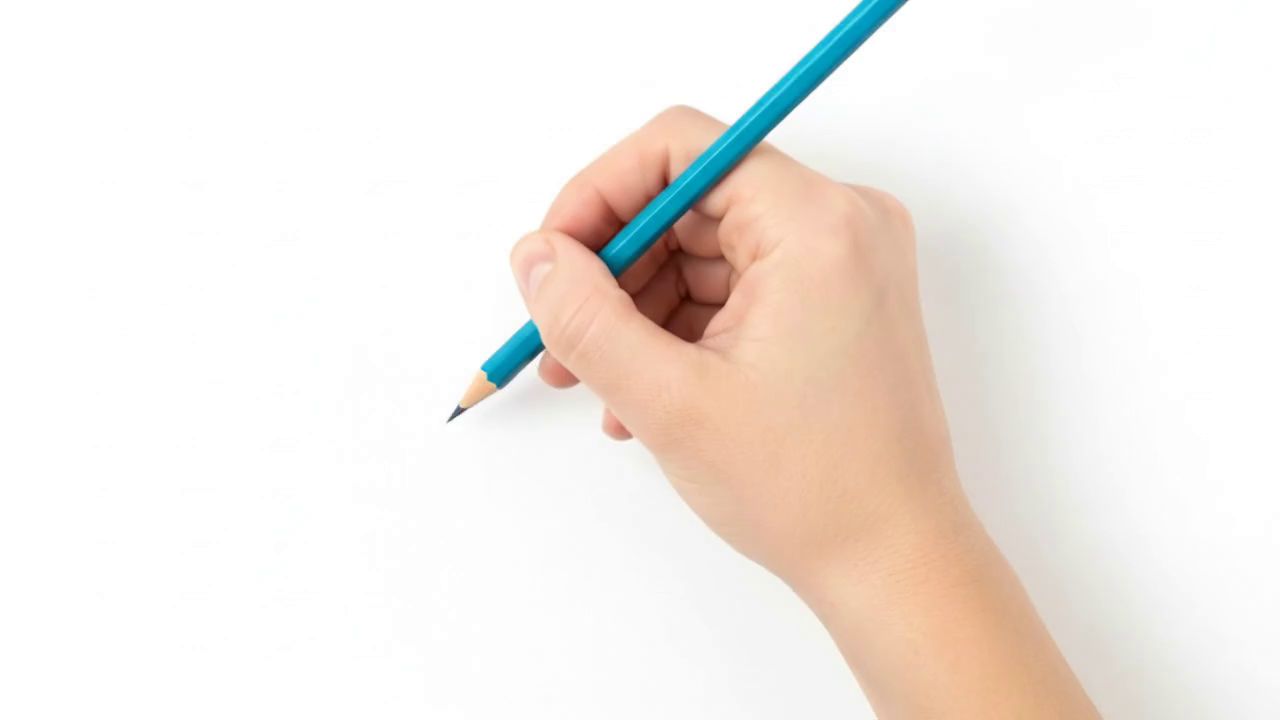}}
    \end{subfigure}
    \hfill
    \begin{subfigure}[t]{\threefigwidth\textwidth}
        \centering
        \fbox{\includegraphics[width=\linewidth]{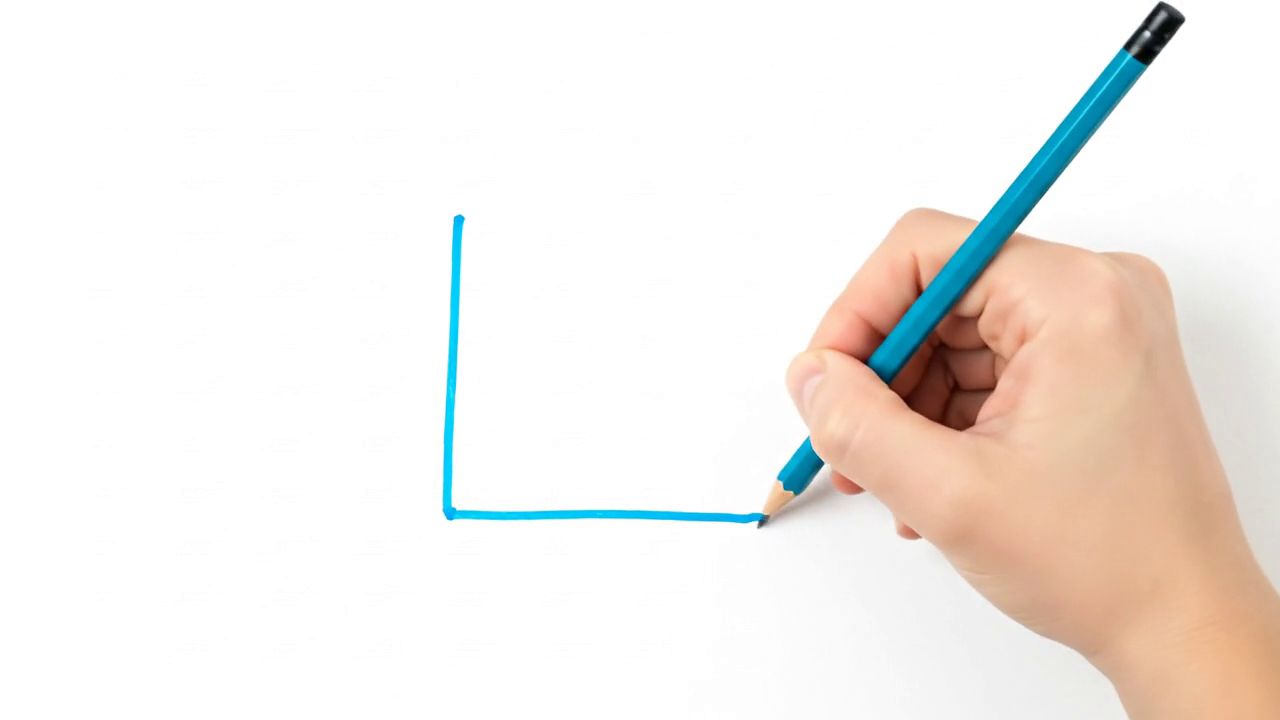}}
    \end{subfigure}
    \hfill
    \begin{subfigure}[t]{\threefigwidth\textwidth}
        \centering
        \fbox{\includegraphics[width=\linewidth]{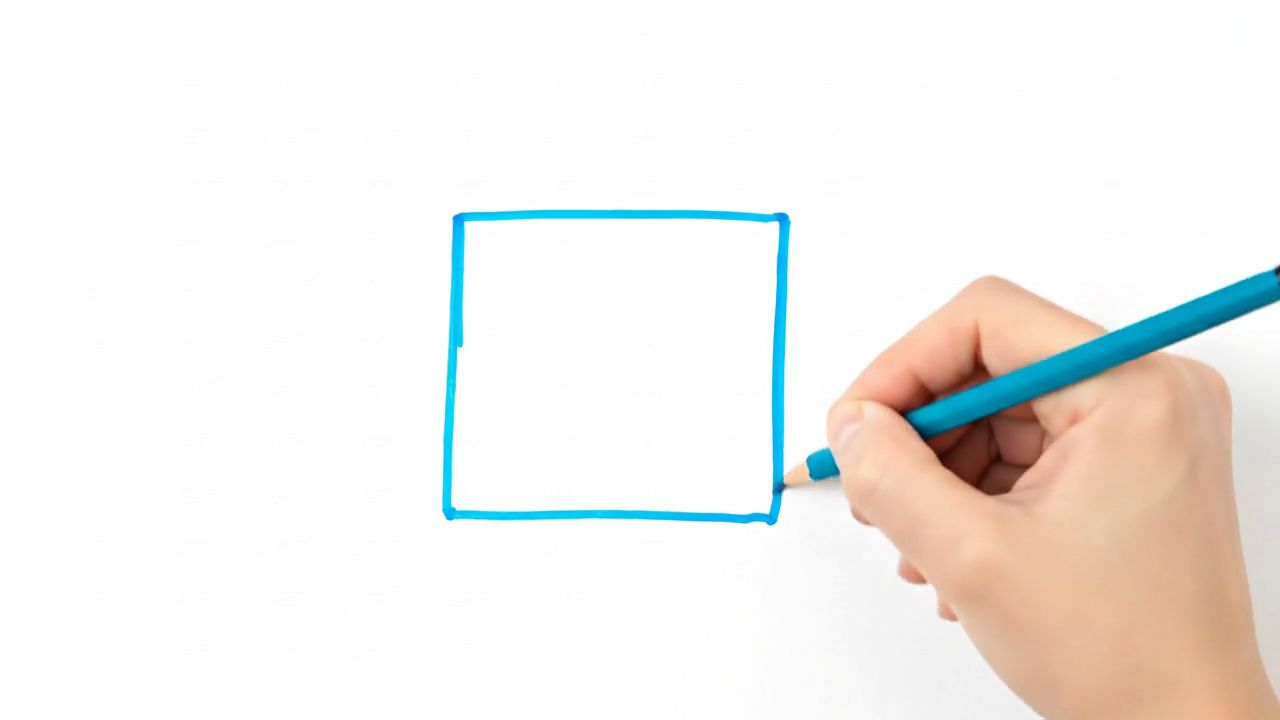}}
    \end{subfigure}
    \caption{\textbf{Drawing}.
    Prompt: \textit{``A person draws a square. Static camera, no pan, no zoom, no dolly.''} Success rate: 0.33.}
    \label{fig:drawing}
\end{figure}

\begin{figure}[H]
    \centering
    \begin{subfigure}[t]{\threefigwidth\textwidth}
        \centering
        \fbox{\includegraphics[width=\linewidth]{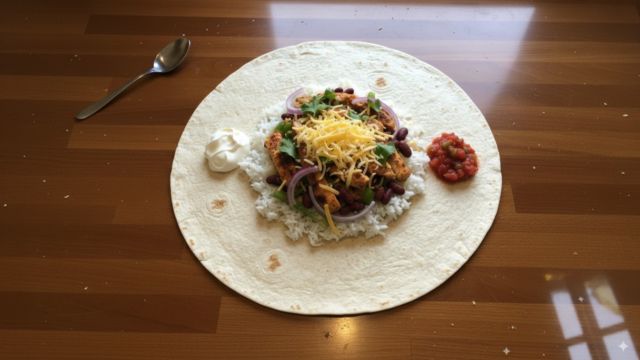}}
    \end{subfigure}
    \hfill
    \begin{subfigure}[t]{\threefigwidth\textwidth}
        \centering
        \fbox{\includegraphics[width=\linewidth]{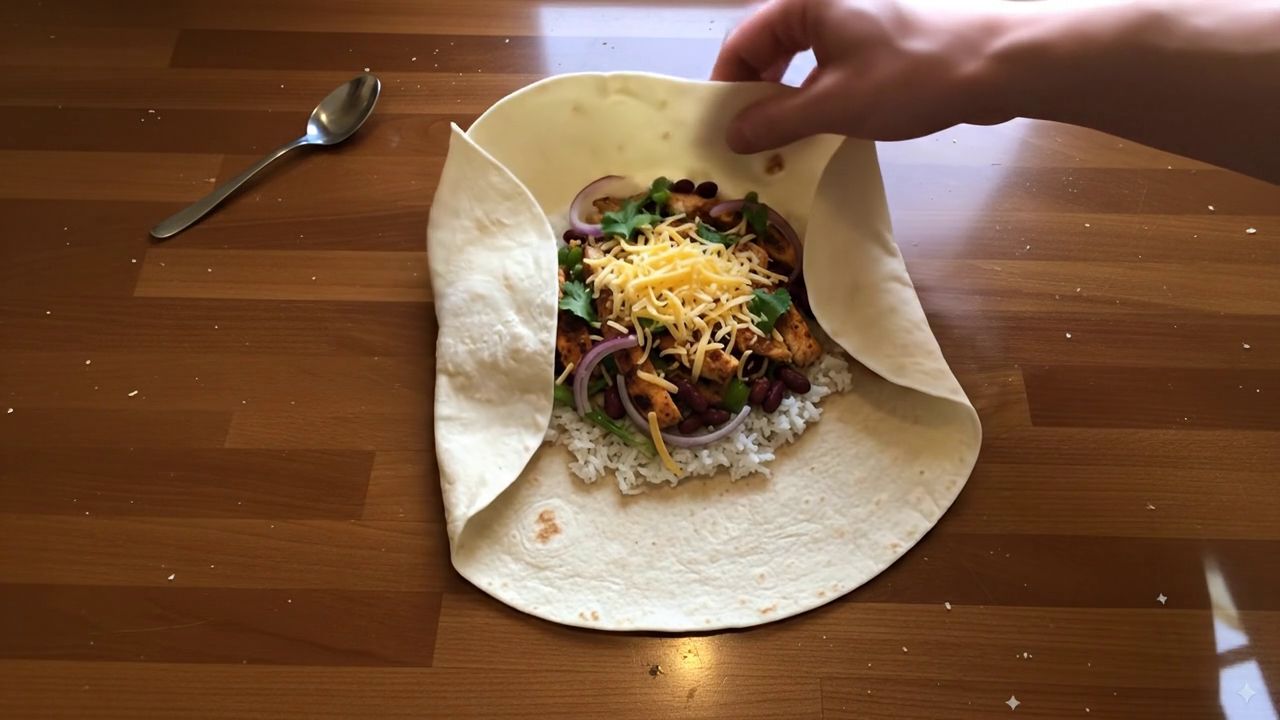}}
    \end{subfigure}
    \hfill
    \begin{subfigure}[t]{\threefigwidth\textwidth}
        \centering
        \fbox{\includegraphics[width=\linewidth]{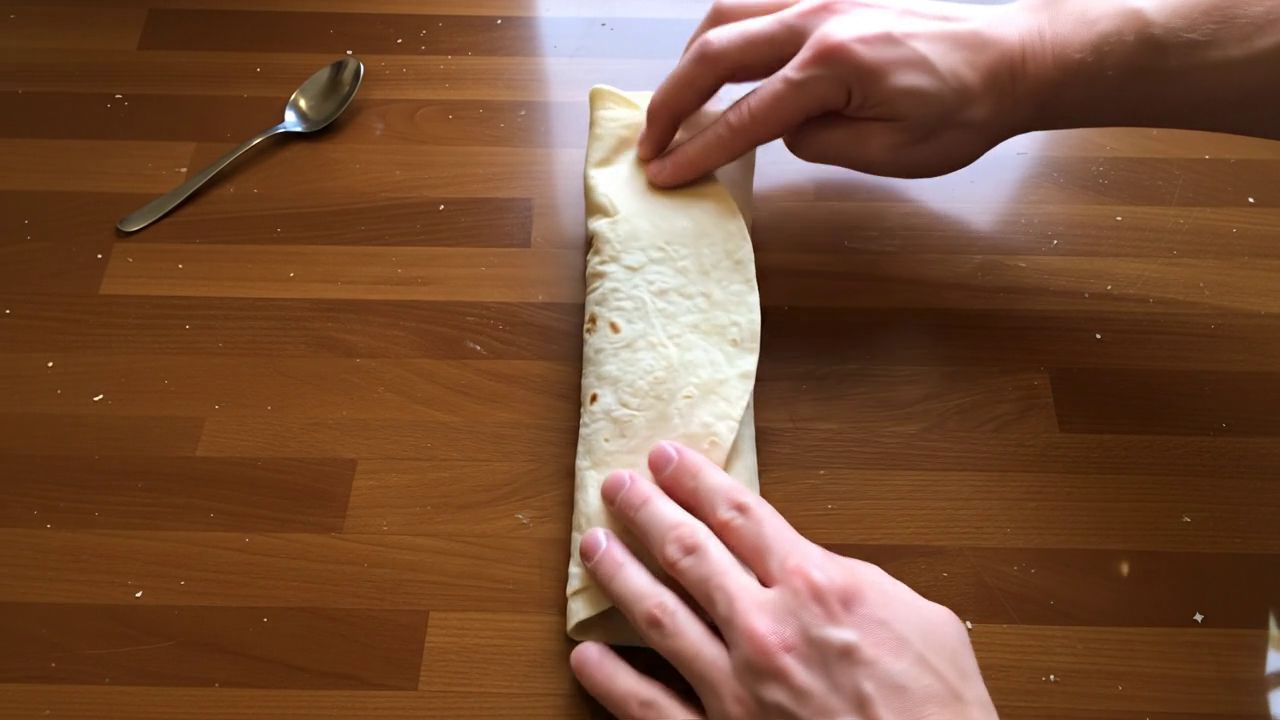}}
    \end{subfigure}
    \caption{\textbf{Visual instruction generation}. Prompt: \textit{``A montage clearly showing each step to roll a burrito.''} Success rate:~0.25. Inspiration: \cite{yang2024video} and \href{https://www.reddit.com/r/ChatGPT/comments/1mzl4rx/asked_chatgpt_to_show_me_how_to_roll_a_wrap/}{Reddit}.}
    \label{fig:visual-explanation}
\end{figure}

\subsection{\textcolor{reasoningcol}{\textbf{Reasoning}}}
\label{app:_reasoning_qualitative}

\begin{figure}[H]
    \centering
    \begin{subfigure}[t]{\threefigwidth\textwidth}
        \centering
        \fbox{\includegraphics[width=\linewidth]{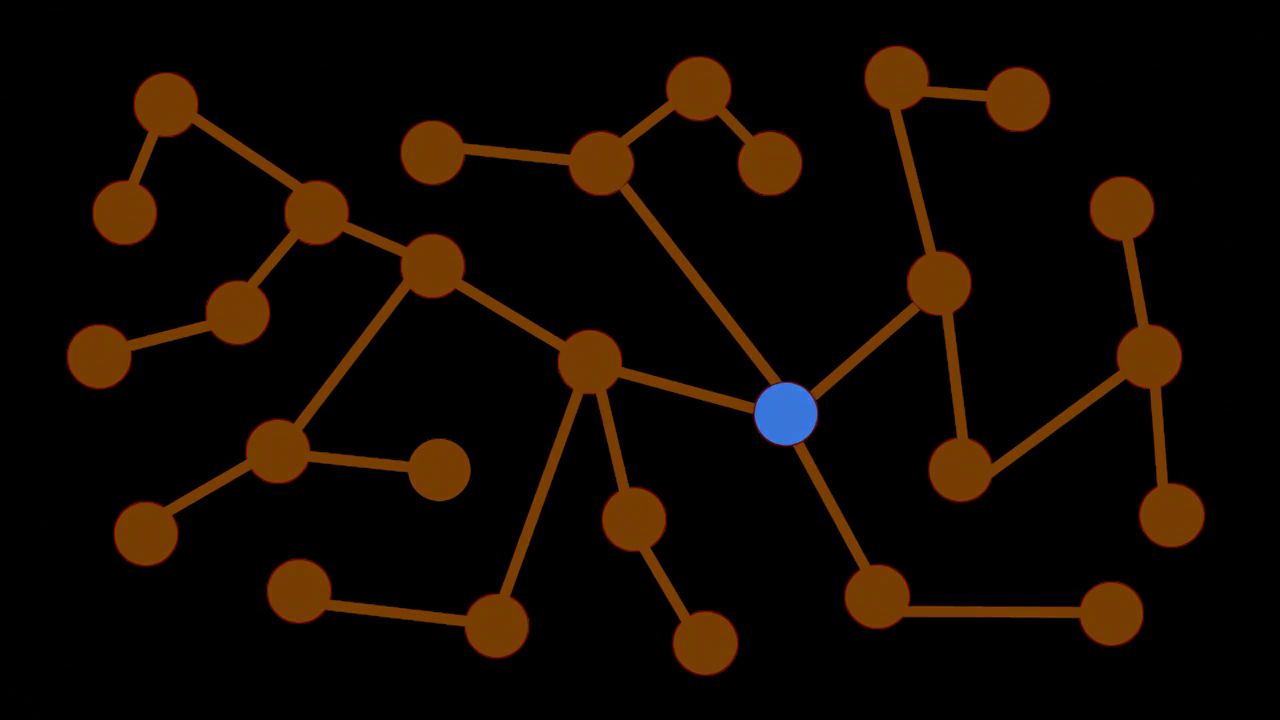}}
    \end{subfigure}
    \hfill
    \begin{subfigure}[t]{\threefigwidth\textwidth}
        \centering
        \fbox{\includegraphics[width=\linewidth]{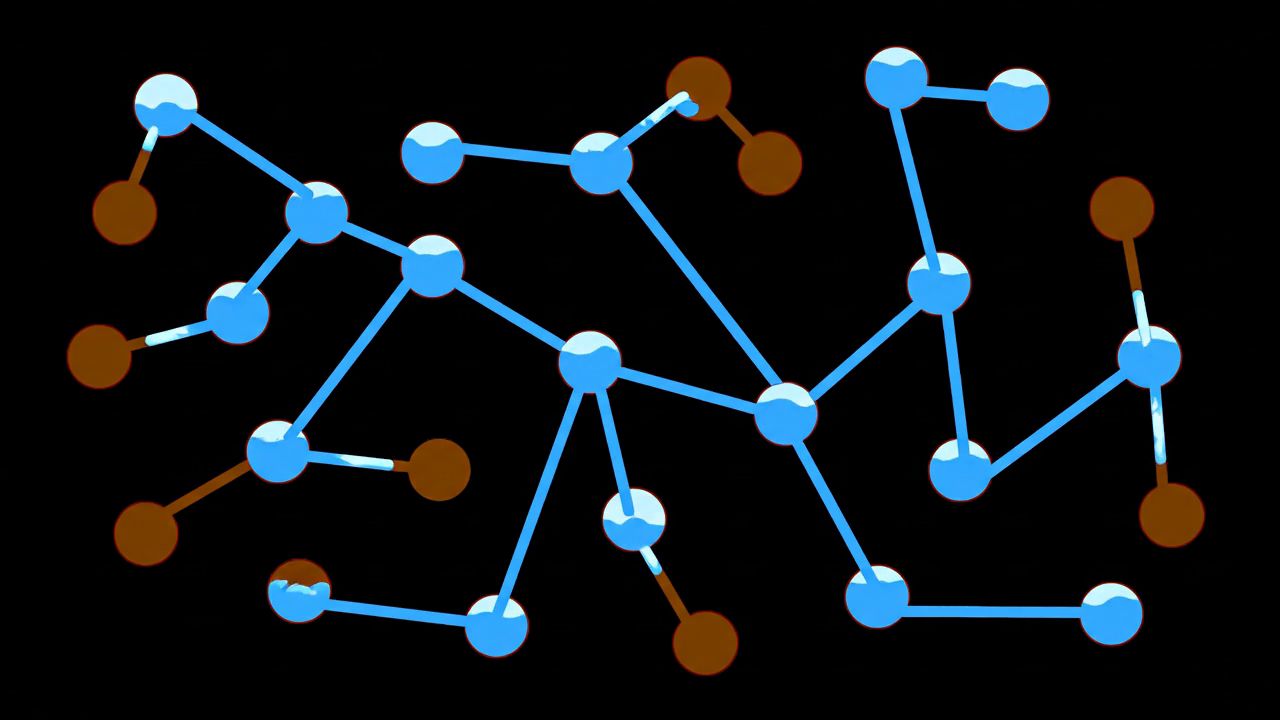}}
    \end{subfigure}
    \hfill    
    \begin{subfigure}[t]{\threefigwidth\textwidth}
        \centering
        \fbox{\includegraphics[width=\linewidth]{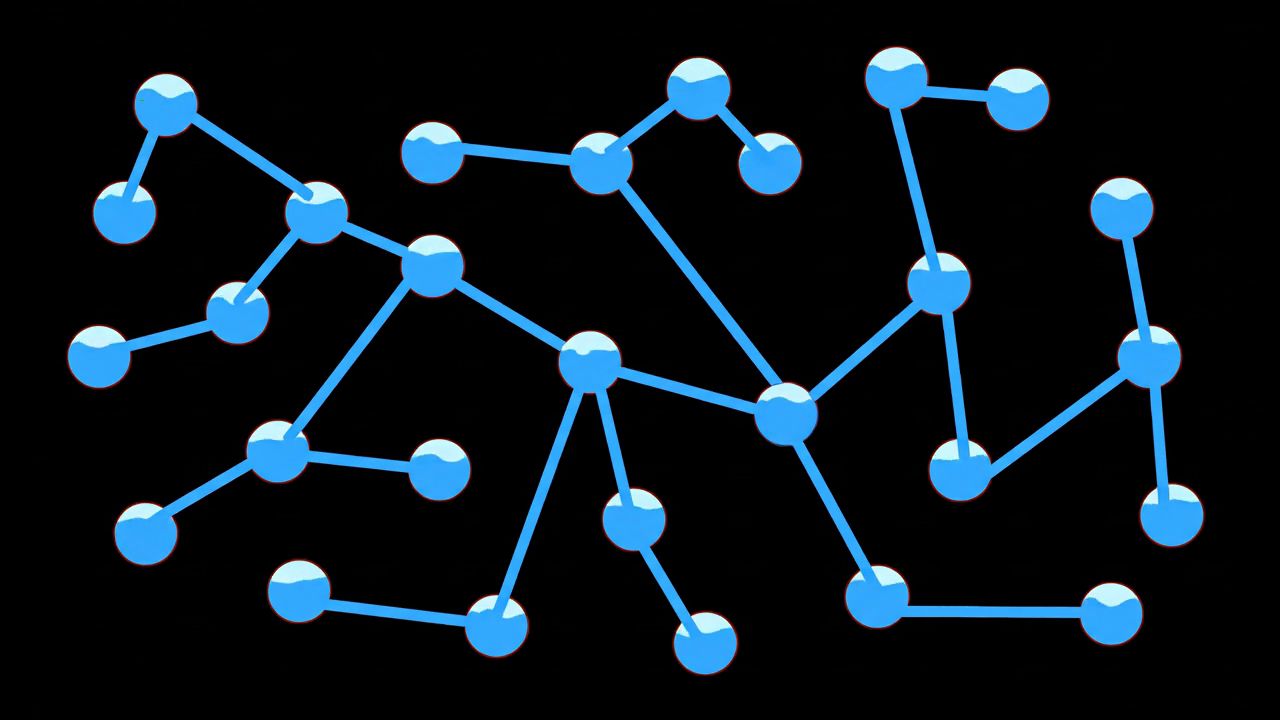}}
    \end{subfigure}
    \caption{\textbf{Graph traversal}. Prompt: \textit{``Starting from the blue well, an unlimited supply of blue water moves through the connected channel system without spilling into the black area.''} Success rate:~0.08.}
    \label{fig:graph_traversal}
\end{figure}

\begin{figure}[H]
    \centering
    \begin{subfigure}[t]{\threefigwidth\textwidth}
        \centering
        \fbox{\includegraphics[width=\linewidth]{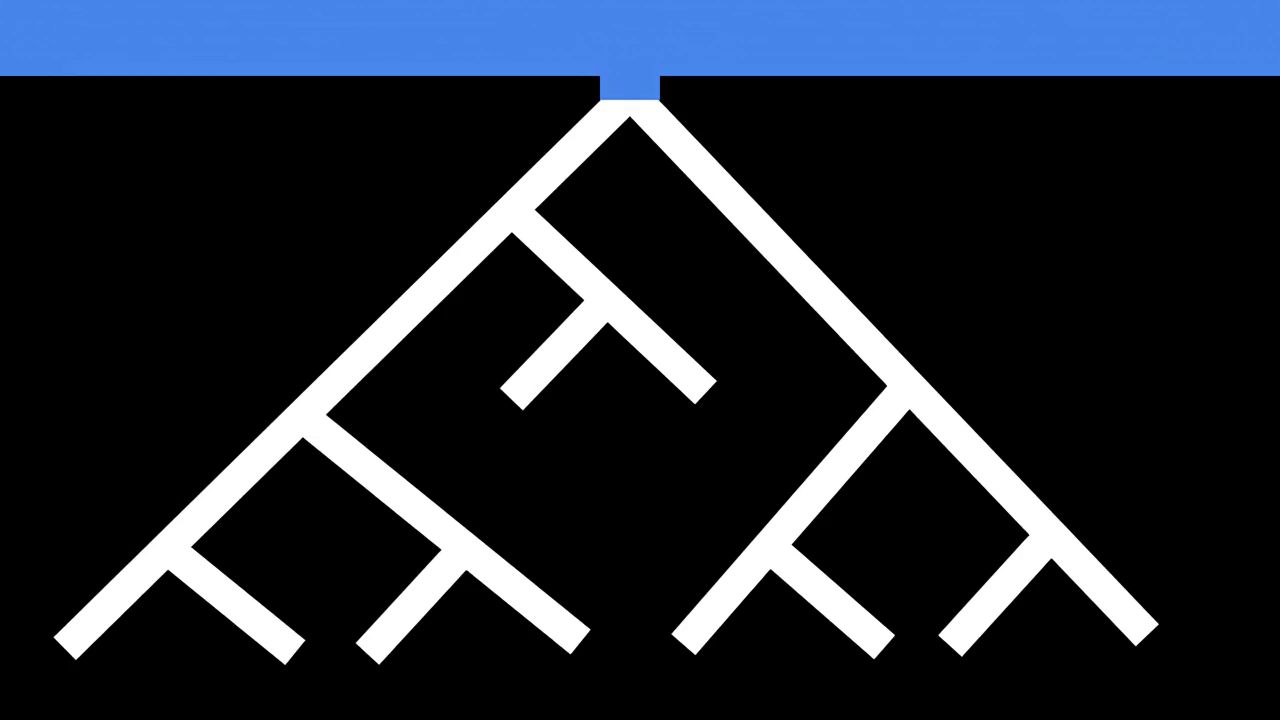}}
    \end{subfigure}
    \hfill
    \begin{subfigure}[t]{\threefigwidth\textwidth}
        \centering
        \fbox{\includegraphics[width=\linewidth]{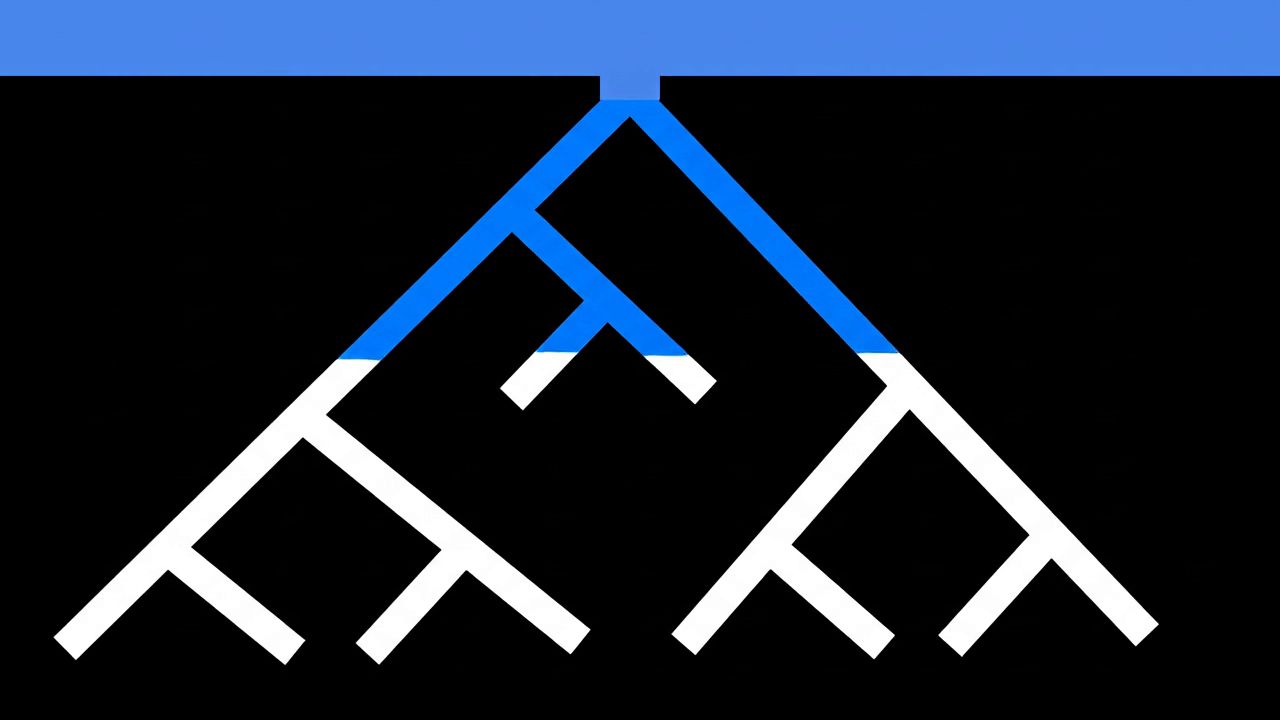}}
    \end{subfigure}
    \hfill    
    \begin{subfigure}[t]{\threefigwidth\textwidth}
        \centering
        \fbox{\includegraphics[width=\linewidth]{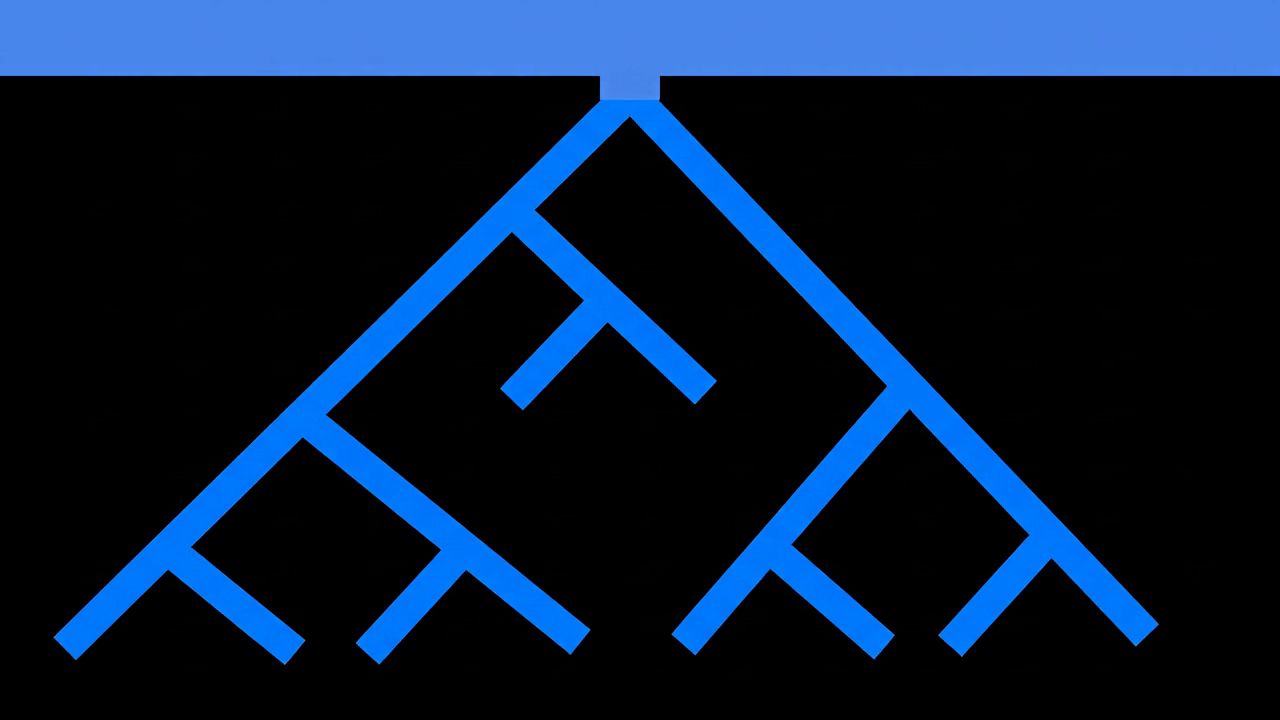}}
    \end{subfigure}
    \caption{\textbf{Tree BFS}. Prompt: \textit{``From the blue water basin, an unlimited supply of water flows at constant speed into the cave system until all caves are filled. Static camera perspective, no zoom or pan.''} Success rate:~0.17.}
    \label{fig:tree_bfs}
\end{figure}

\begin{figure}[H]
    \centering
    \begin{subfigure}[t]{0.24\textwidth}
        \centering
        \fbox{\includegraphics[width=\linewidth]{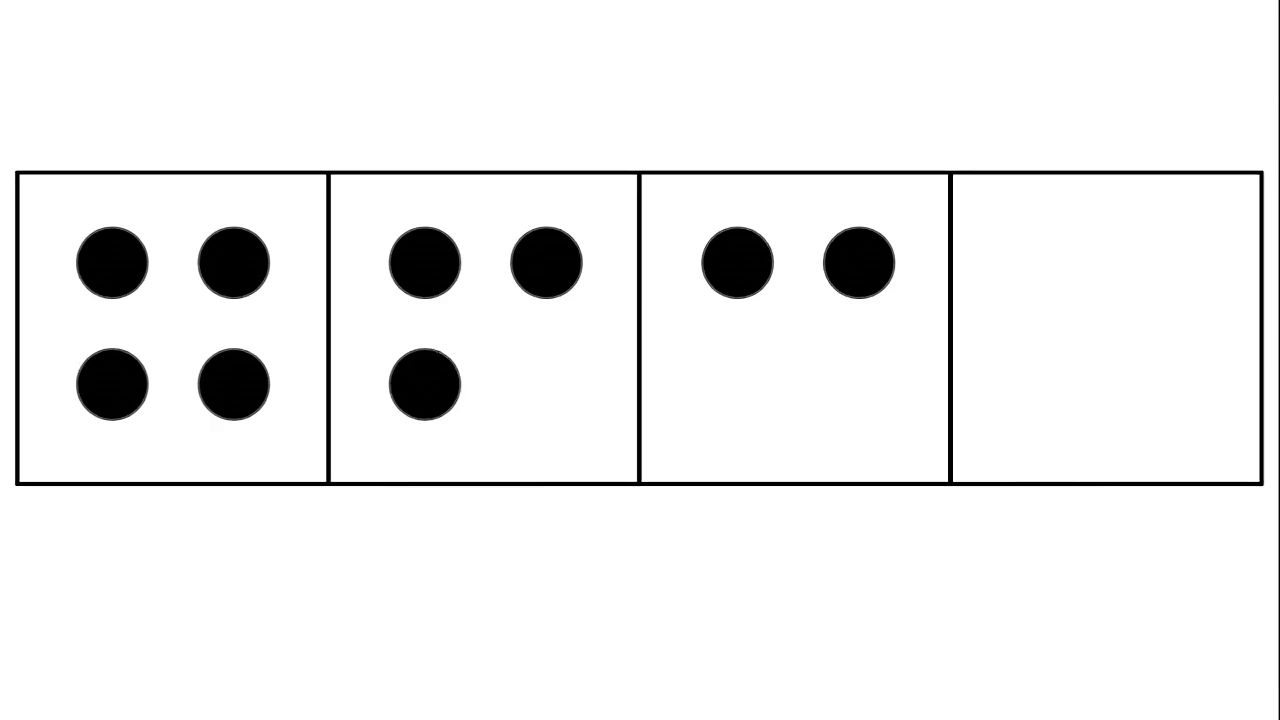}}
    \end{subfigure}
    \hfill
    \begin{subfigure}[t]{0.24\textwidth}
        \centering
        \fbox{\includegraphics[width=\linewidth]{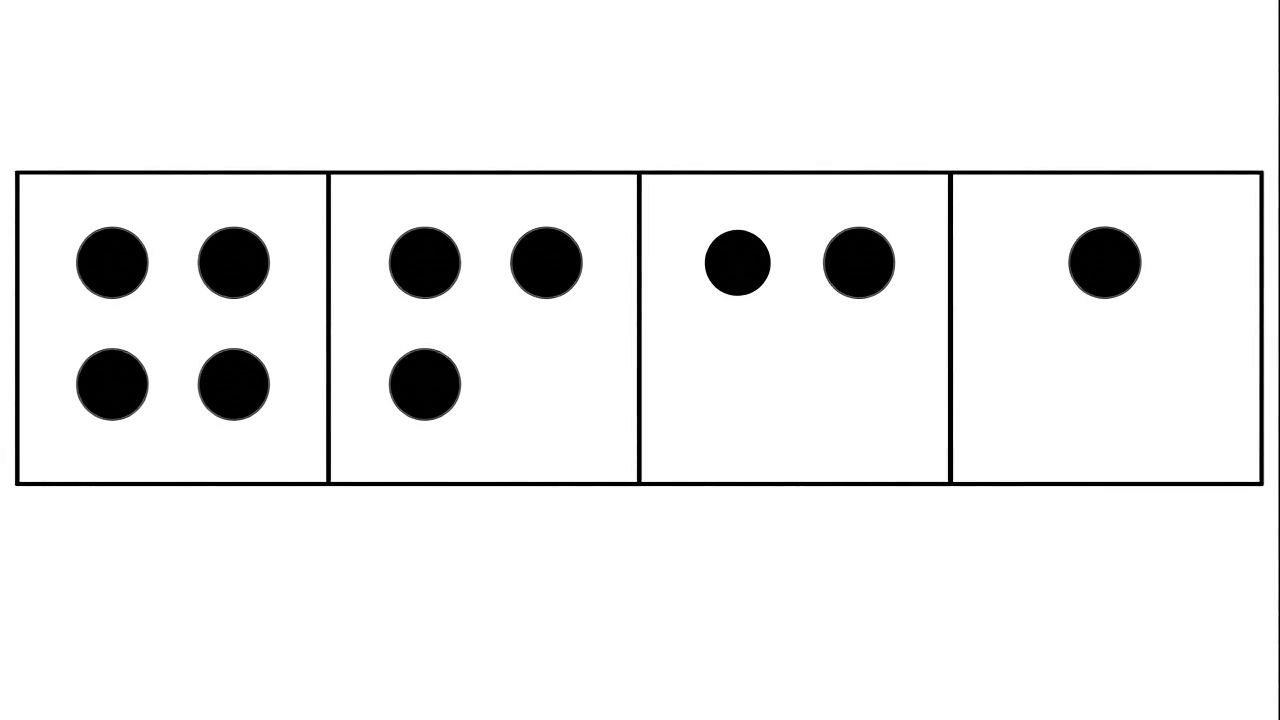}}
    \end{subfigure}
    \hspace{0.2cm}
    \begin{subfigure}[t]{0.24\textwidth}
        \centering
        \fbox{\includegraphics[width=\linewidth]{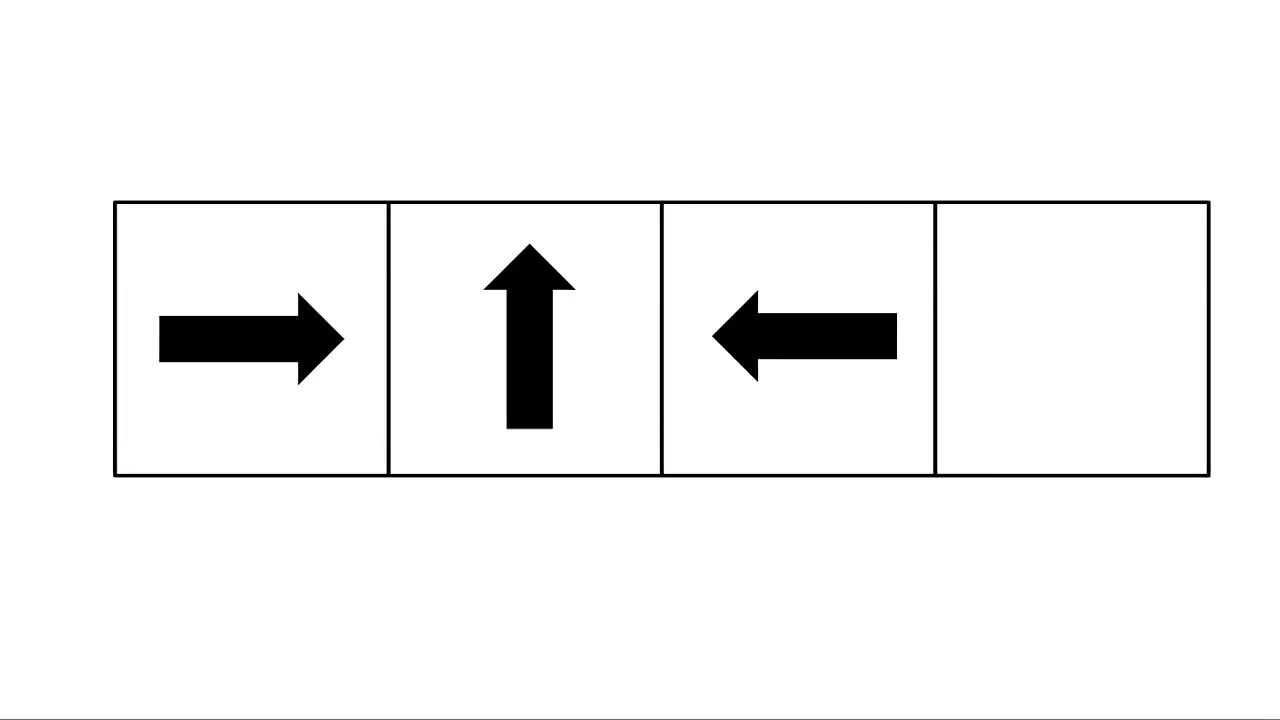}}
    \end{subfigure}
    \hfill
    \begin{subfigure}[t]{0.24\textwidth}
        \centering
        \fbox{\includegraphics[width=\linewidth]{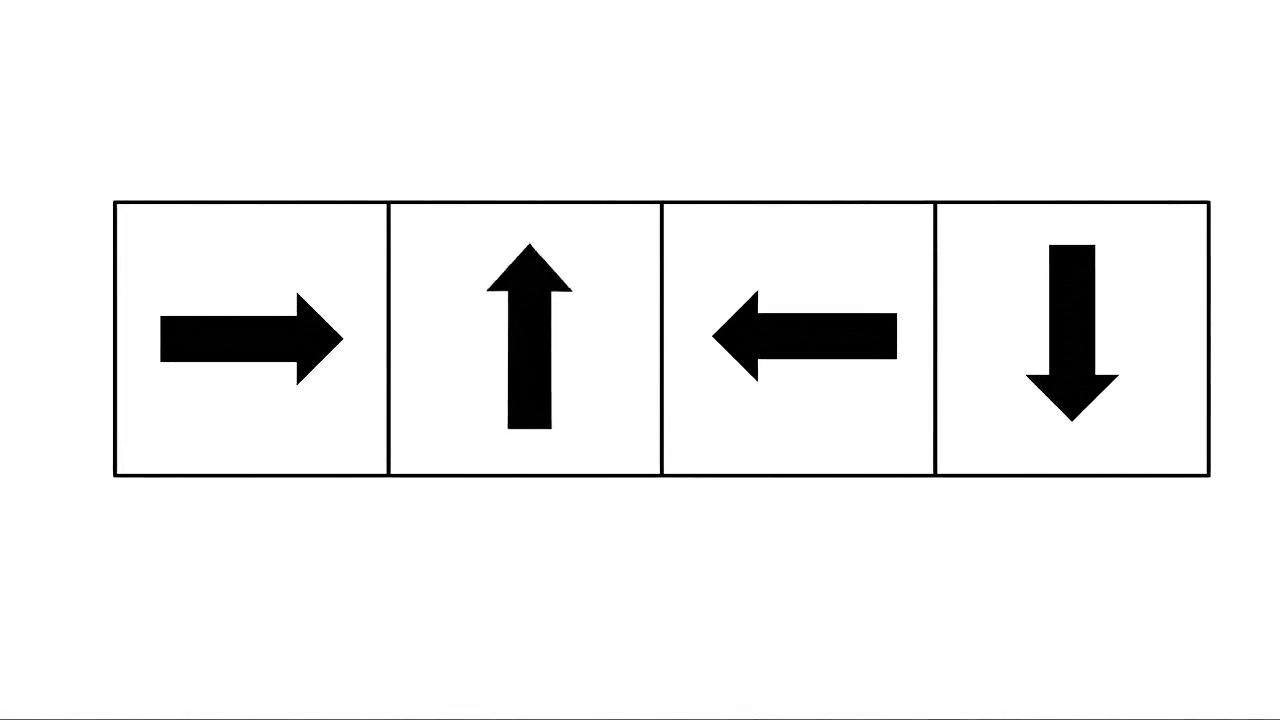}}
    \end{subfigure}

    \vspace{0.2cm}
    
    \begin{subfigure}[t]{0.24\textwidth}
        \centering
        \fbox{\includegraphics[width=\linewidth]{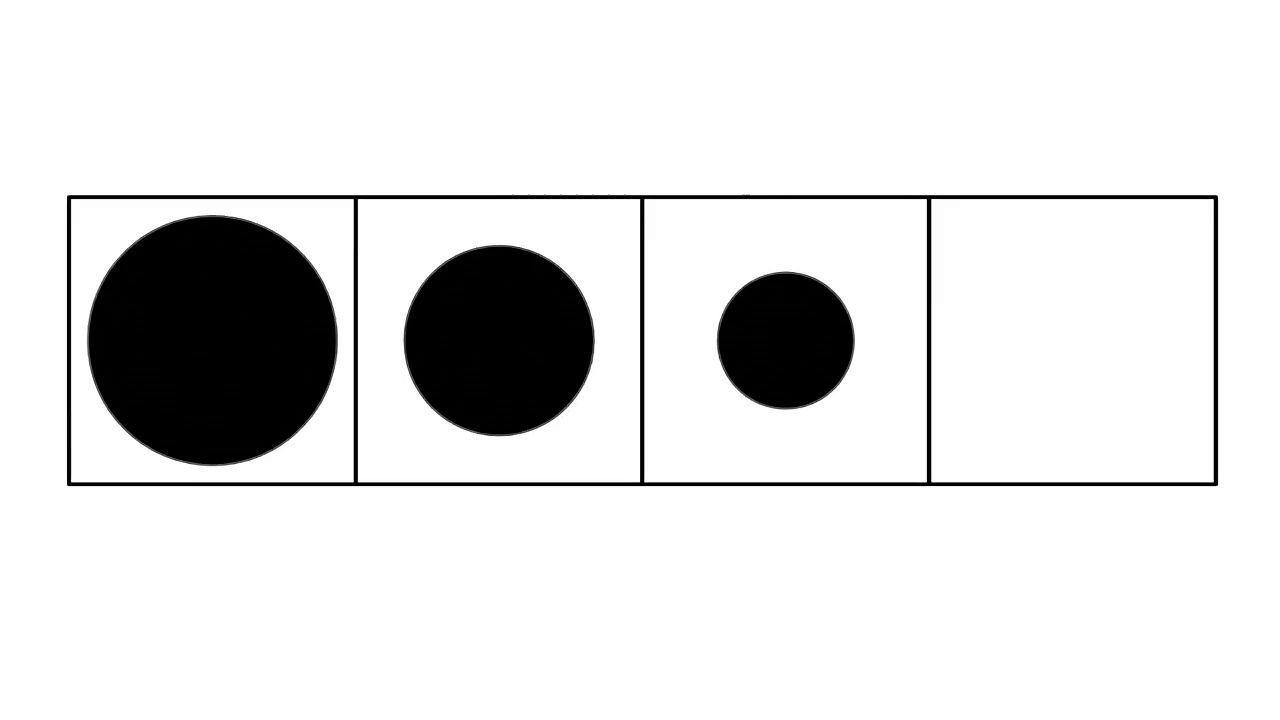}}
    \end{subfigure}
    \hfill
    \begin{subfigure}[t]{0.24\textwidth}
        \centering
        \fbox{\includegraphics[width=\linewidth]{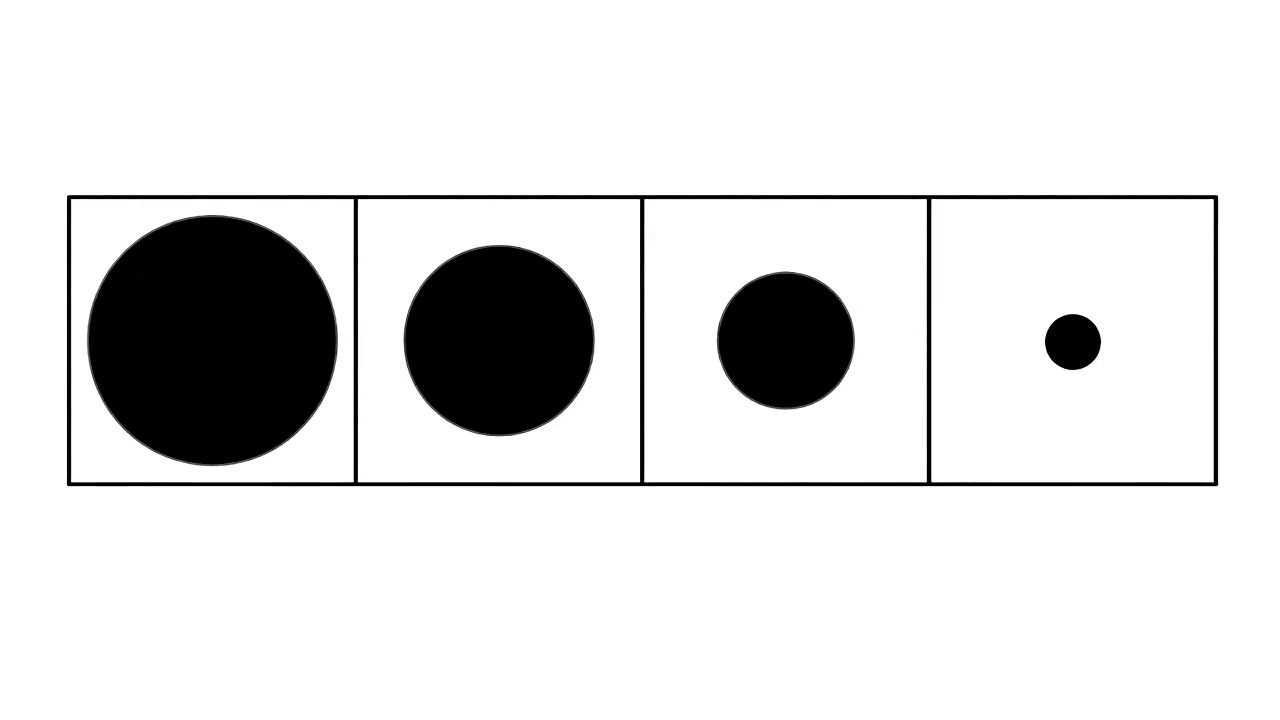}}
    \end{subfigure}
    \hspace{0.2cm}
    \begin{subfigure}[t]{0.24\textwidth}
        \centering
        \fbox{\includegraphics[width=\linewidth]{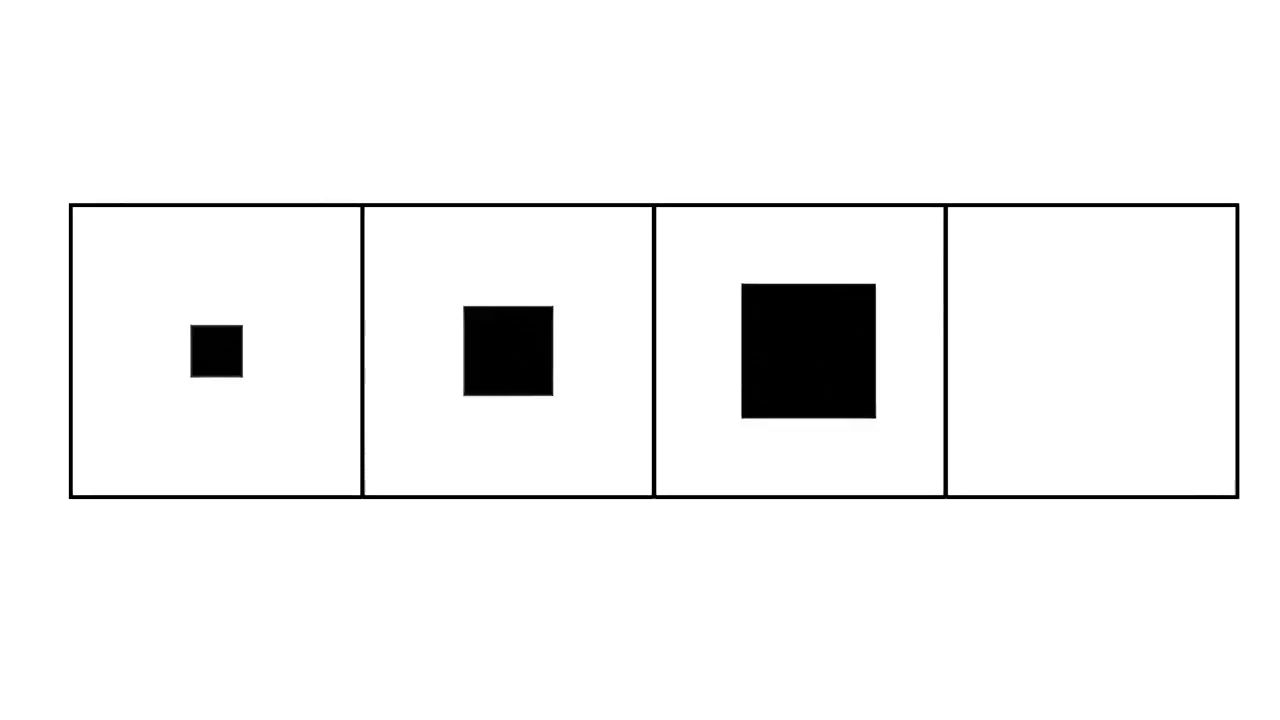}}
    \end{subfigure}
    \hfill
    \begin{subfigure}[t]{0.24\textwidth}
        \centering
        \fbox{\includegraphics[width=\linewidth]{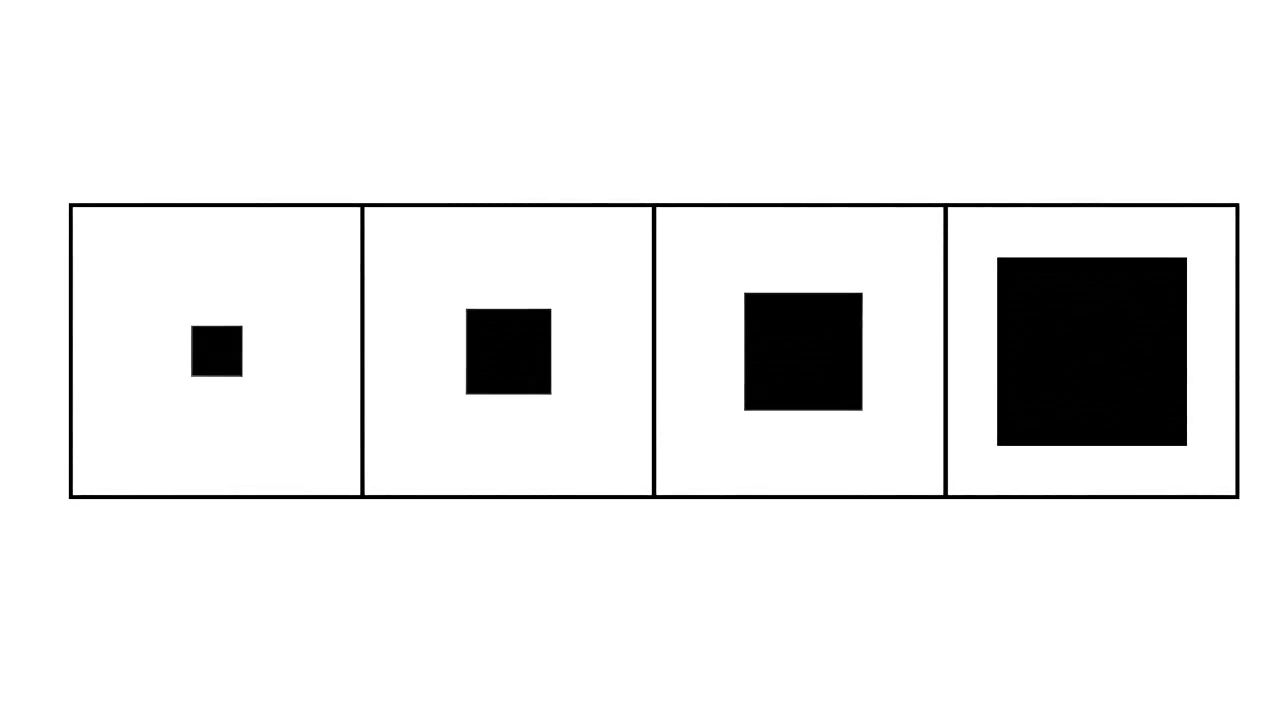}}
    \end{subfigure}
    \caption{\textbf{Sequence completion} inspired by Raven's progressive matrices.
    Each of the four pairs shows input (left) and generated output (right).
    Prompt: \textit{``Draw the figure that completes the pattern in the rightmost box. The images in the boxes are static. Do not modify the existing images, only draw in the empty box. Static camera, no zoom, no pan, no dolly.''} Success rate: 0.33 for dots, 1.0 for arrows, 0.75 for shrinking circles, 0.83 for growing squares.}
    \label{fig:raven_logic}
\end{figure}

\begin{figure}[H]
    \centering
    \begin{subfigure}[t]{\threefigwidth\textwidth}
        \centering
        \fbox{\includegraphics[width=\linewidth]{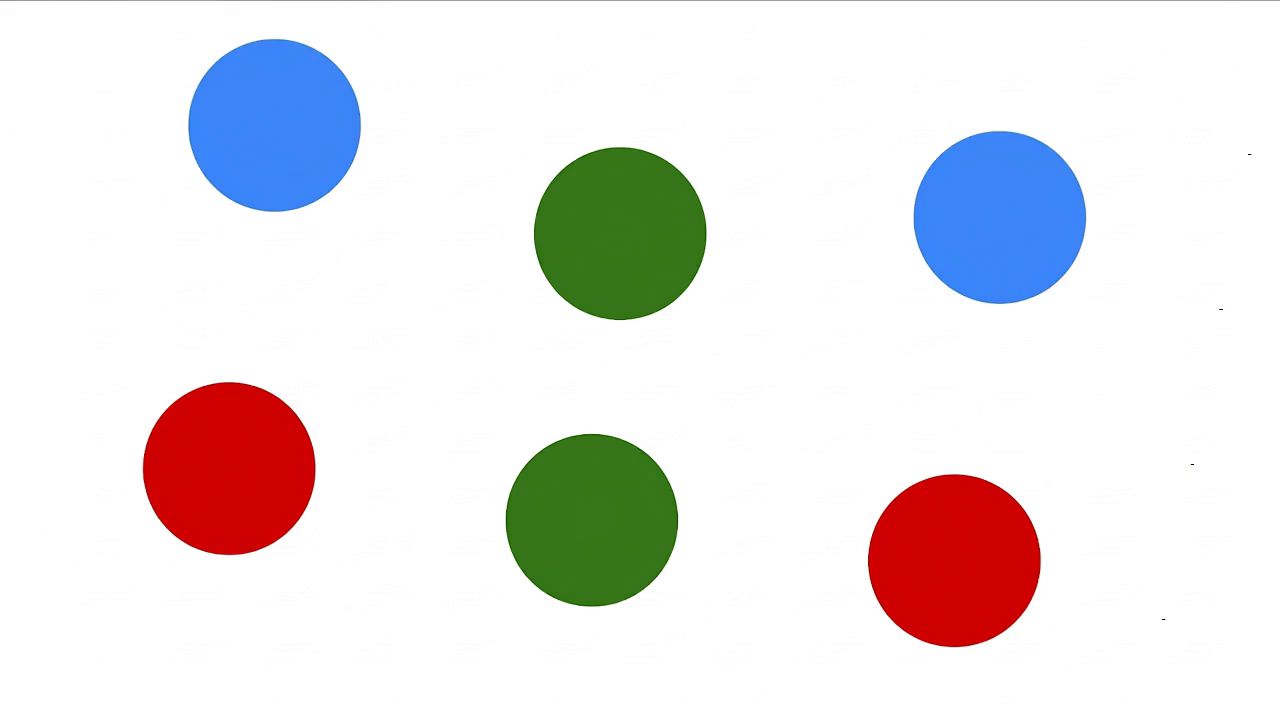}}
    \end{subfigure}
    \hfill
    \begin{subfigure}[t]{\threefigwidth\textwidth}
        \centering
        \fbox{\includegraphics[width=\linewidth]{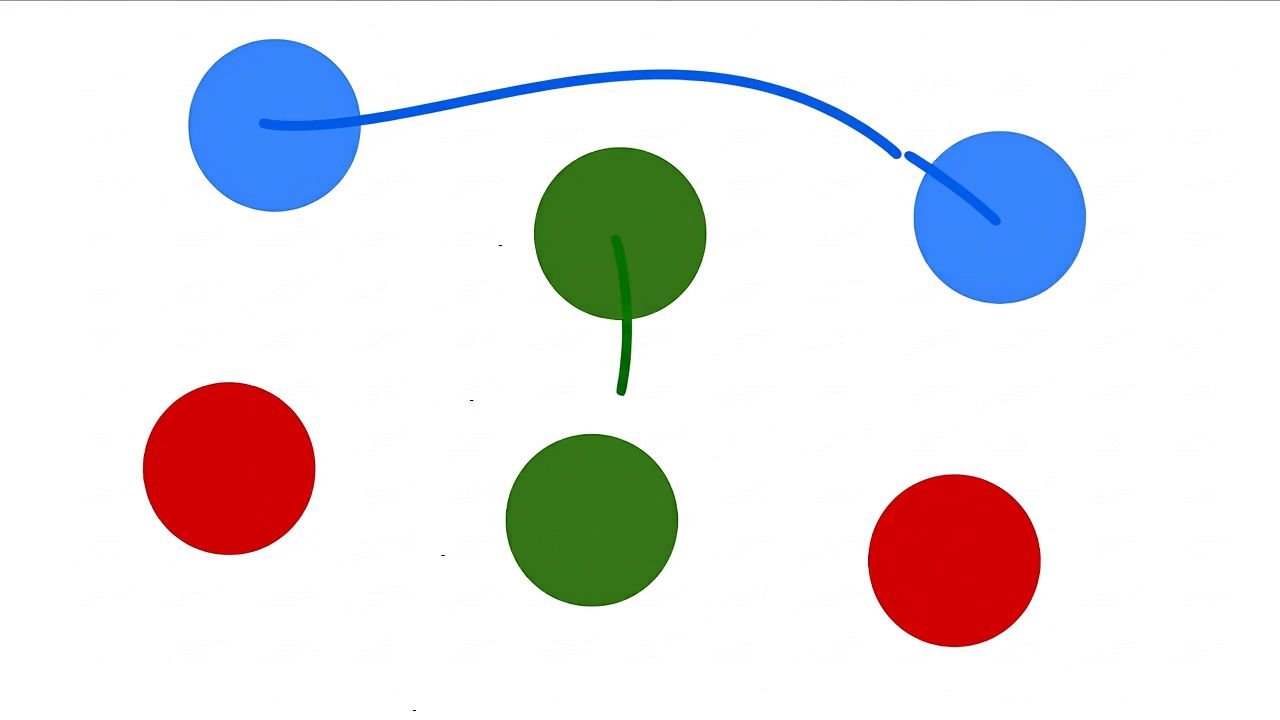}}
    \end{subfigure}
    \hfill    
    \begin{subfigure}[t]{\threefigwidth\textwidth}
        \centering
        \fbox{\includegraphics[width=\linewidth]{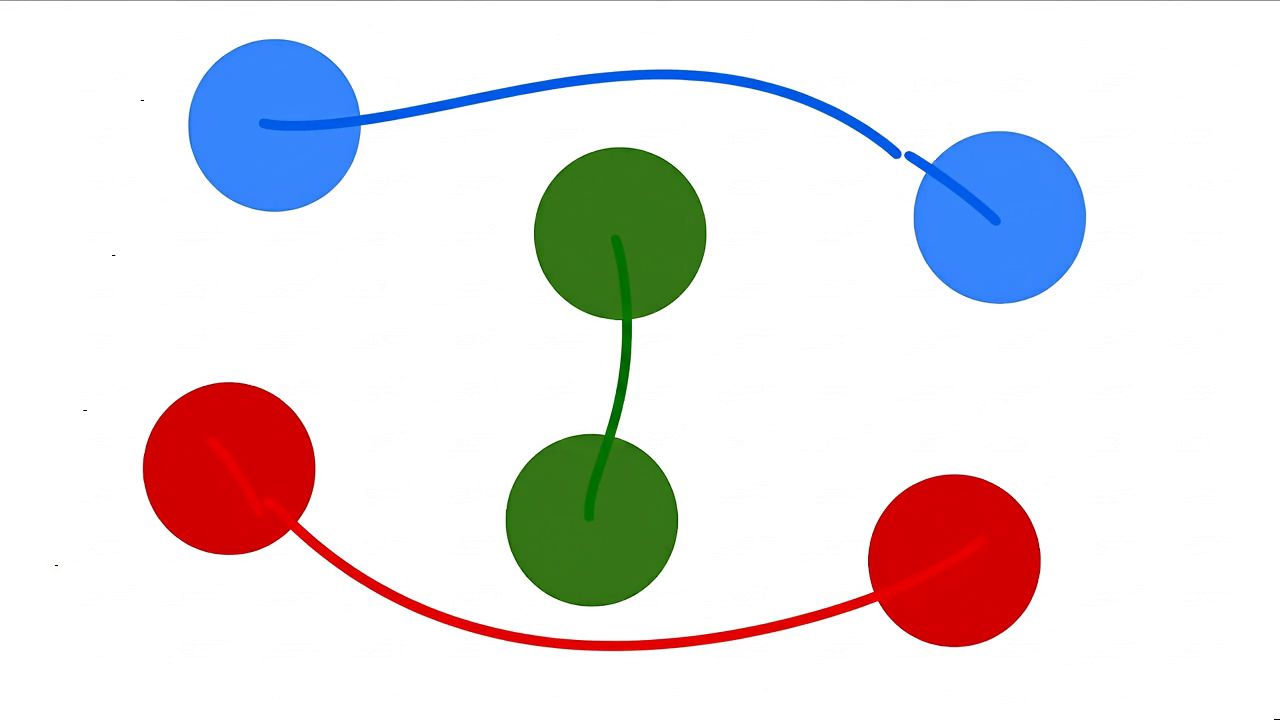}}
    \end{subfigure}
    \caption{\textbf{Connecting colors}.
    Prompt: \textit{``Draw three curves, one connecting each pair of circles of the same color.''} Success rate:~0.25.}
    \label{fig:connecting_colors}
\end{figure}

\begin{figure}[H]
    \centering
    \begin{subfigure}[t]{\fourfigwidth\textwidth}
        \centering
        \fbox{\includegraphics[width=\linewidth]{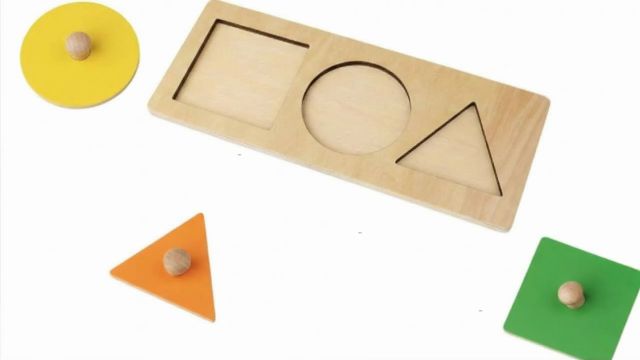}}
    \end{subfigure}
    \hfill
    \begin{subfigure}[t]{\fourfigwidth\textwidth}
        \centering
        \fbox{\includegraphics[width=\linewidth]{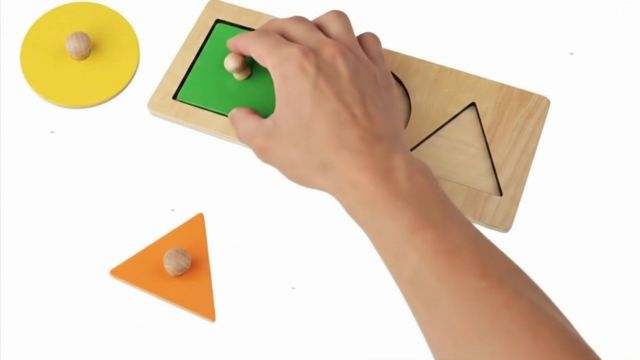}}
    \end{subfigure}
    \hfill
    \begin{subfigure}[t]{\fourfigwidth\textwidth}
        \centering
        \fbox{\includegraphics[width=\linewidth]{figures-compressed/spatial-reasoning_frame_0163.jpg}}
    \end{subfigure}
        \hfill
    \begin{subfigure}[t]{\fourfigwidth\textwidth}
        \centering
        \fbox{\includegraphics[width=\linewidth]{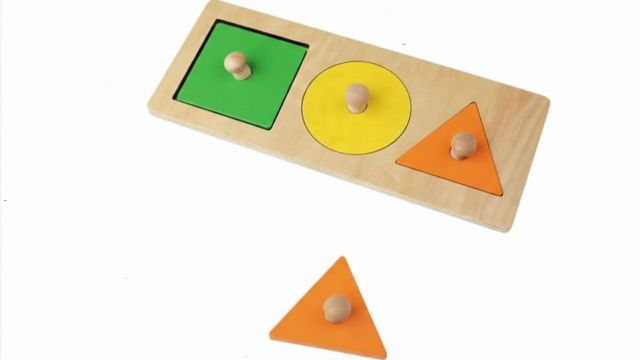}}
    \end{subfigure}
    \caption{\textbf{Shape fitting}. Prompt: \textit{``The scene shows three colored pieces, and a wooden panel with three holes. Each colored piece fits into one and only one hole. A hand grabs each colored piece and puts it into an empty hole that has the exact same shape - if it doesn't fit, the hand tries another hole. All the objects must be placed in their respective holes.''} Success rate:~0.25.}
    \label{fig:shape-fitting}
\end{figure}

\begin{figure}[H]
    \centering
    \begin{subfigure}[t]{0.16\textwidth}
        \centering
        \fbox{\includegraphics[width=\linewidth]{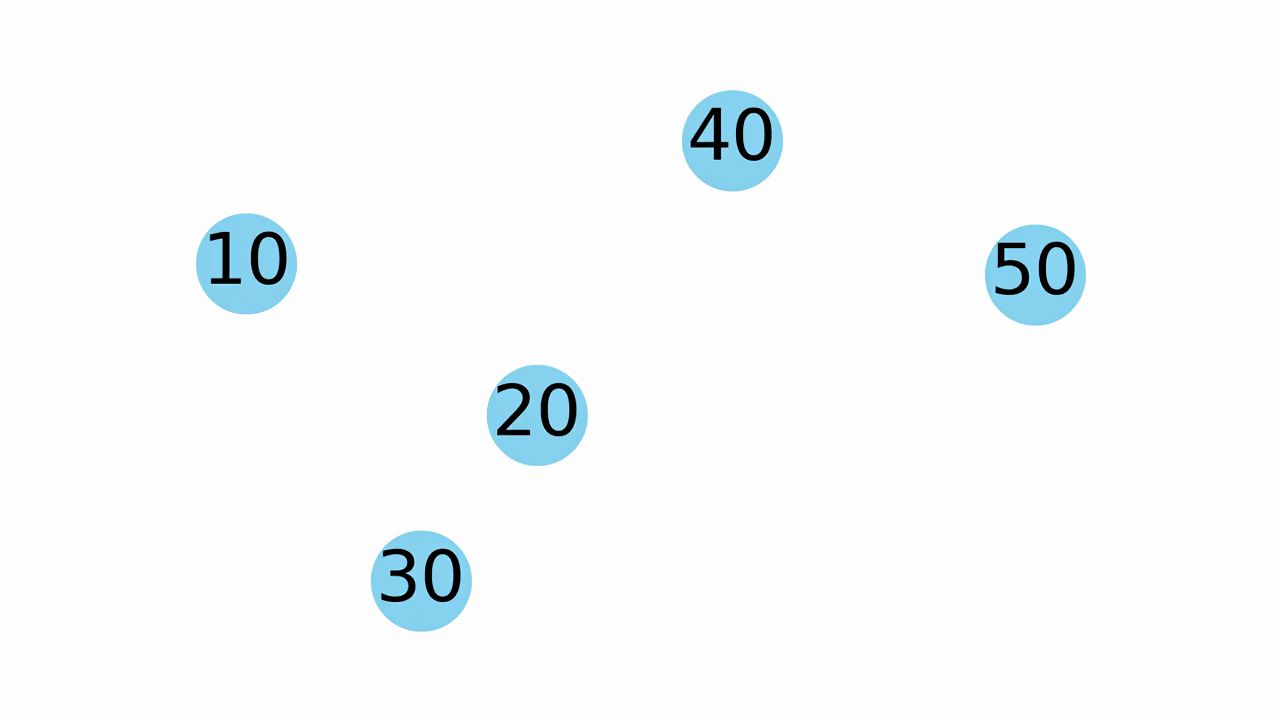}}
    \end{subfigure}
    \hfill
    \begin{subfigure}[t]{0.16\textwidth}
        \centering
        \fbox{\includegraphics[width=\linewidth]{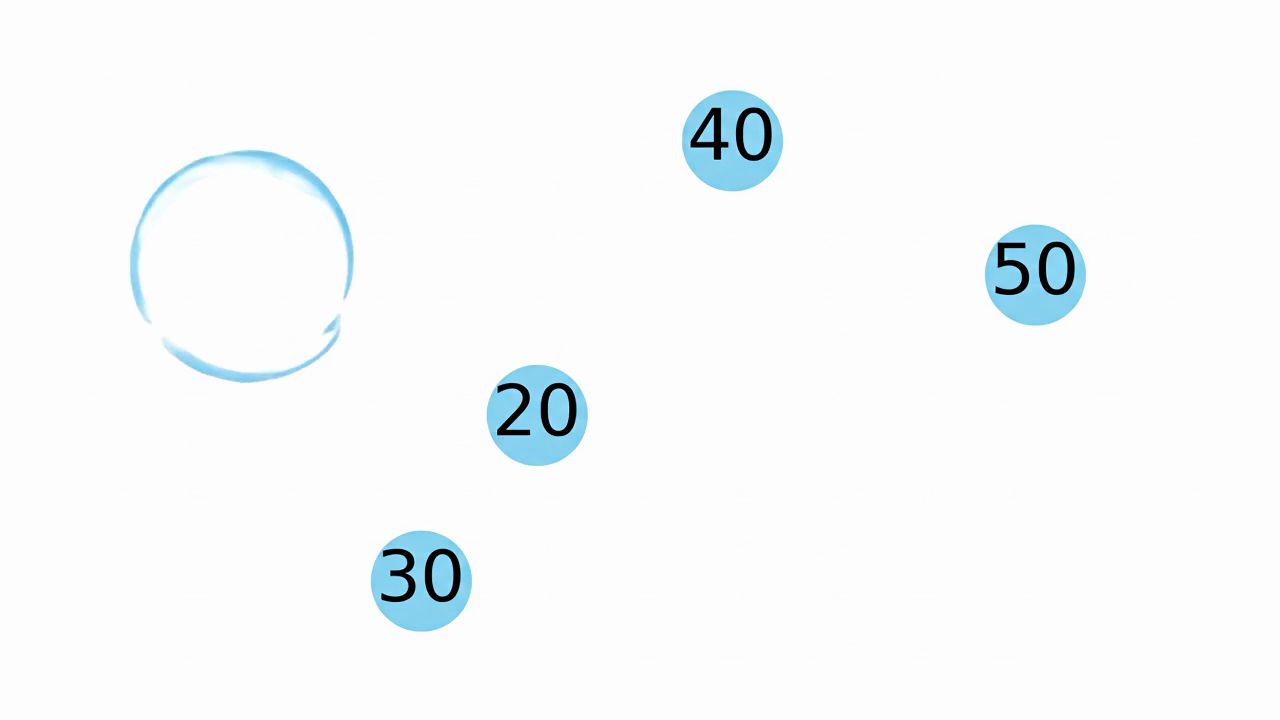}}
    \end{subfigure}
    \hfill    
    \begin{subfigure}[t]{0.16\textwidth}
        \centering
        \fbox{\includegraphics[width=\linewidth]{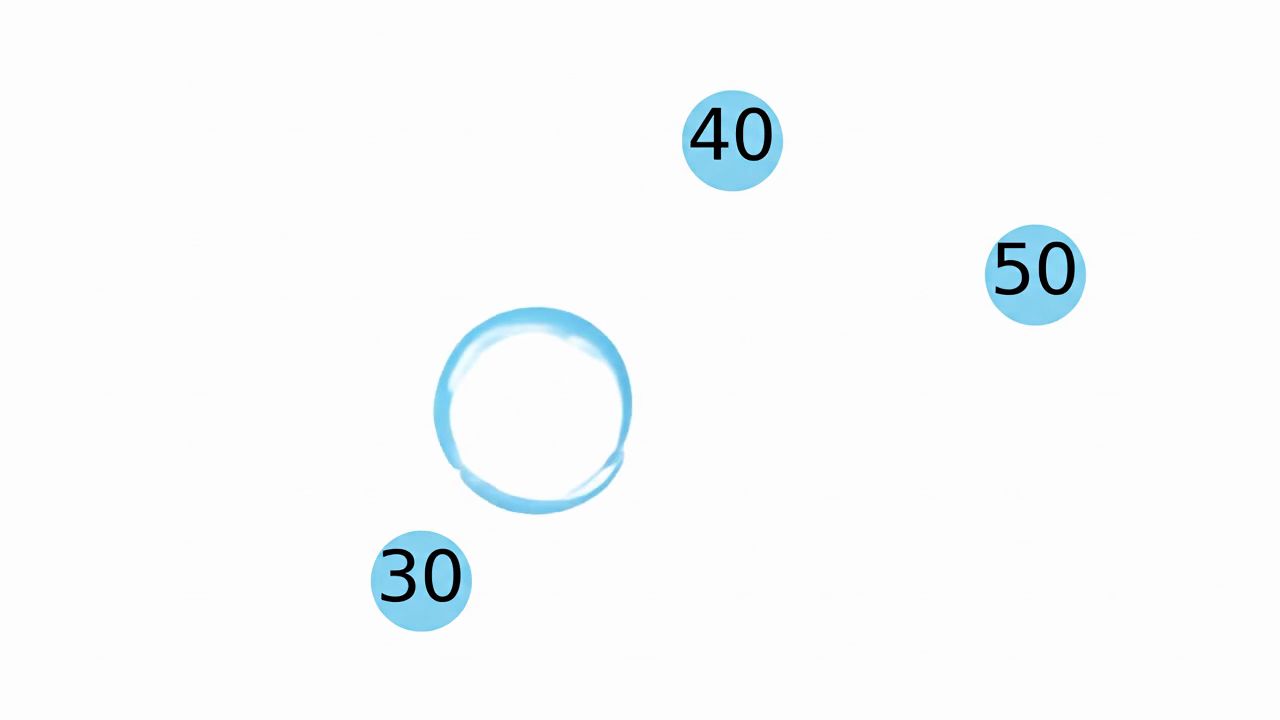}}
    \end{subfigure}    
    \begin{subfigure}[t]{0.16\textwidth}
        \centering
        \fbox{\includegraphics[width=\linewidth]{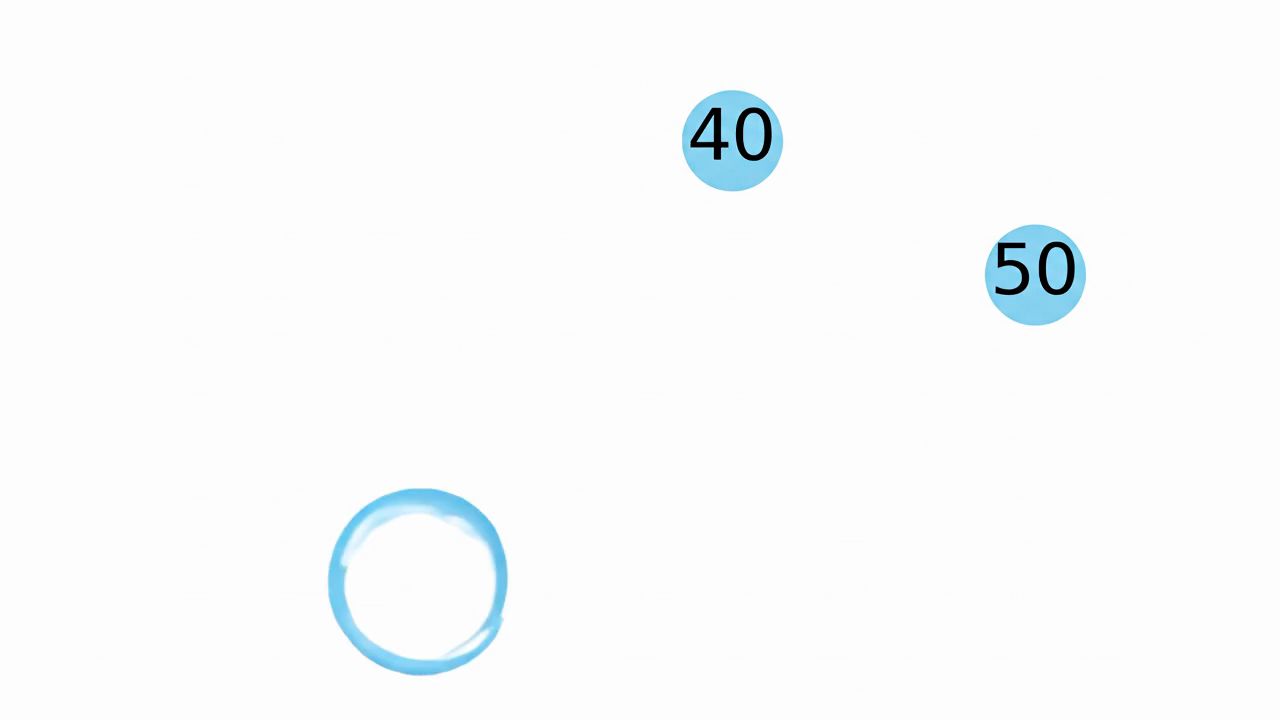}}
    \end{subfigure}
    \hfill    
    \begin{subfigure}[t]{0.16\textwidth}
        \centering
        \fbox{\includegraphics[width=\linewidth]{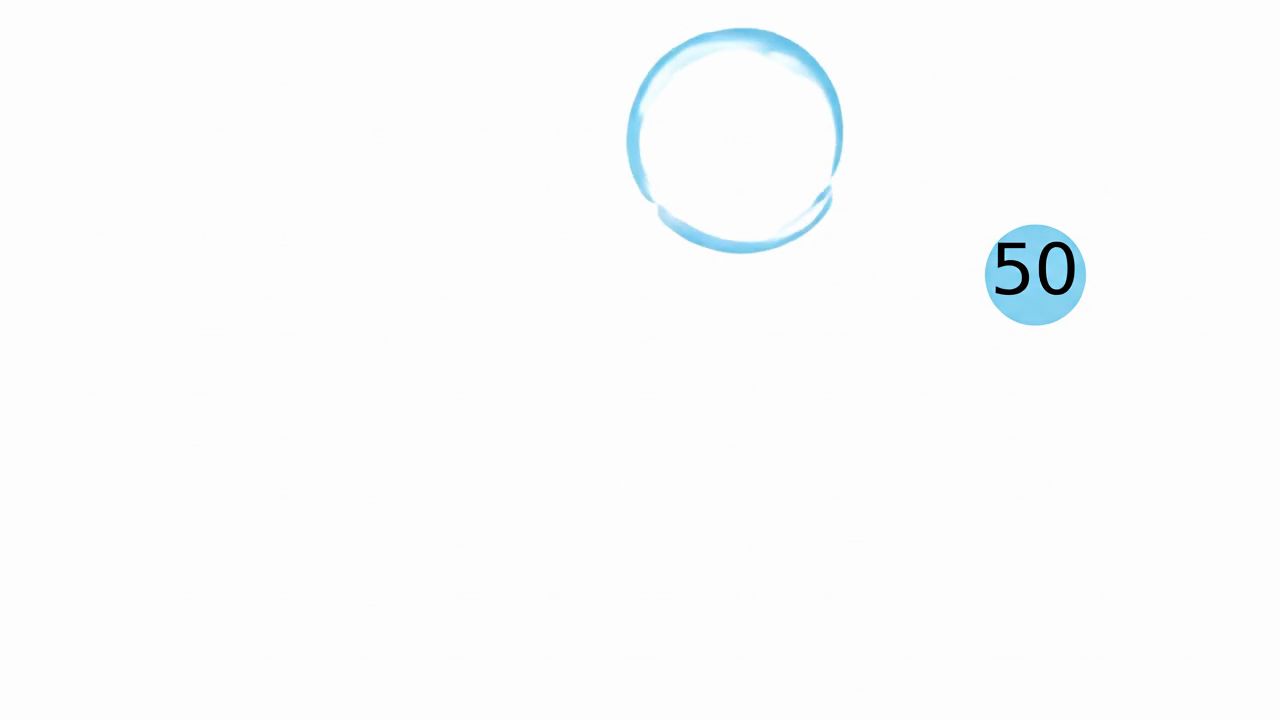}}
    \end{subfigure}
    \hfill    
    \begin{subfigure}[t]{0.16\textwidth}
        \centering
        \fbox{\includegraphics[width=\linewidth]{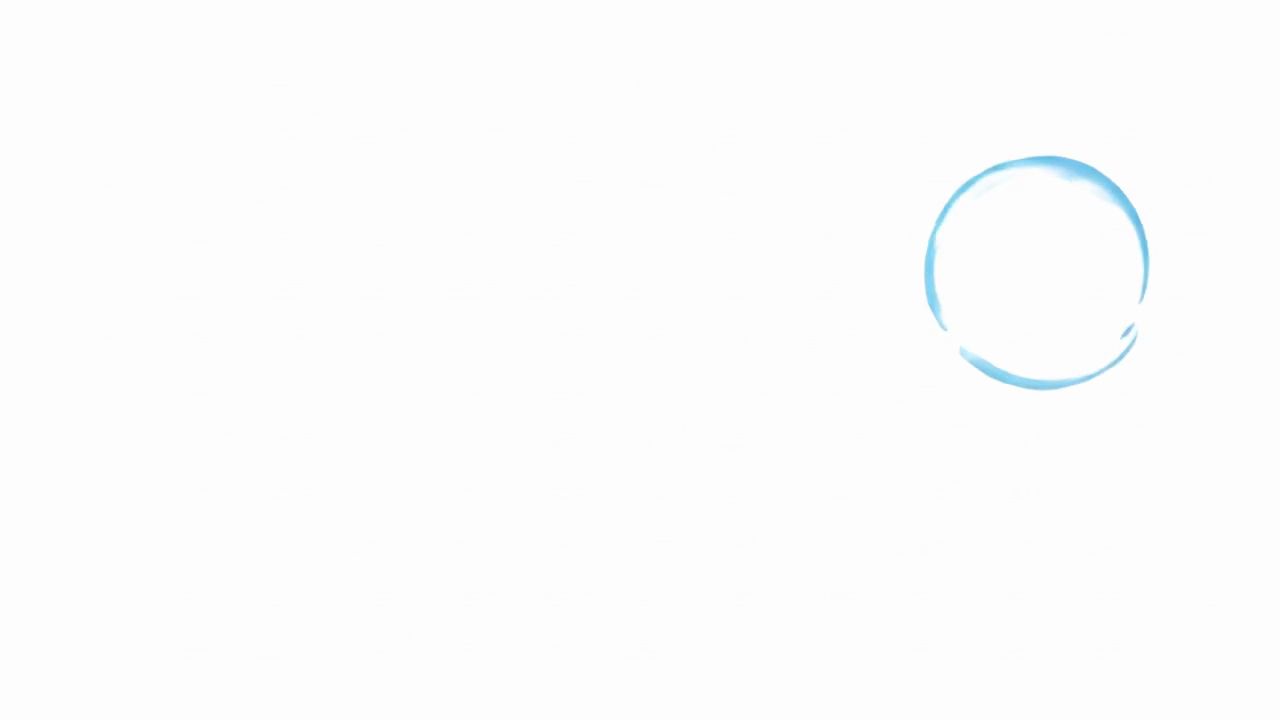}}
    \end{subfigure}
    \caption{\textbf{Sorting numbers}.
    Prompt: \textit{``The video starts with some numbered bubbles. The bubbles pop and disappear one at a time, in numeric order, starting from the one with the smallest number.''} Success rate:~0.08.}
    \label{fig:number_sort}
\end{figure}

\begin{figure}[H]
    \centering
    \begin{subfigure}[t]{0.195\textwidth}
        \centering
        \fbox{\includegraphics[width=\linewidth]{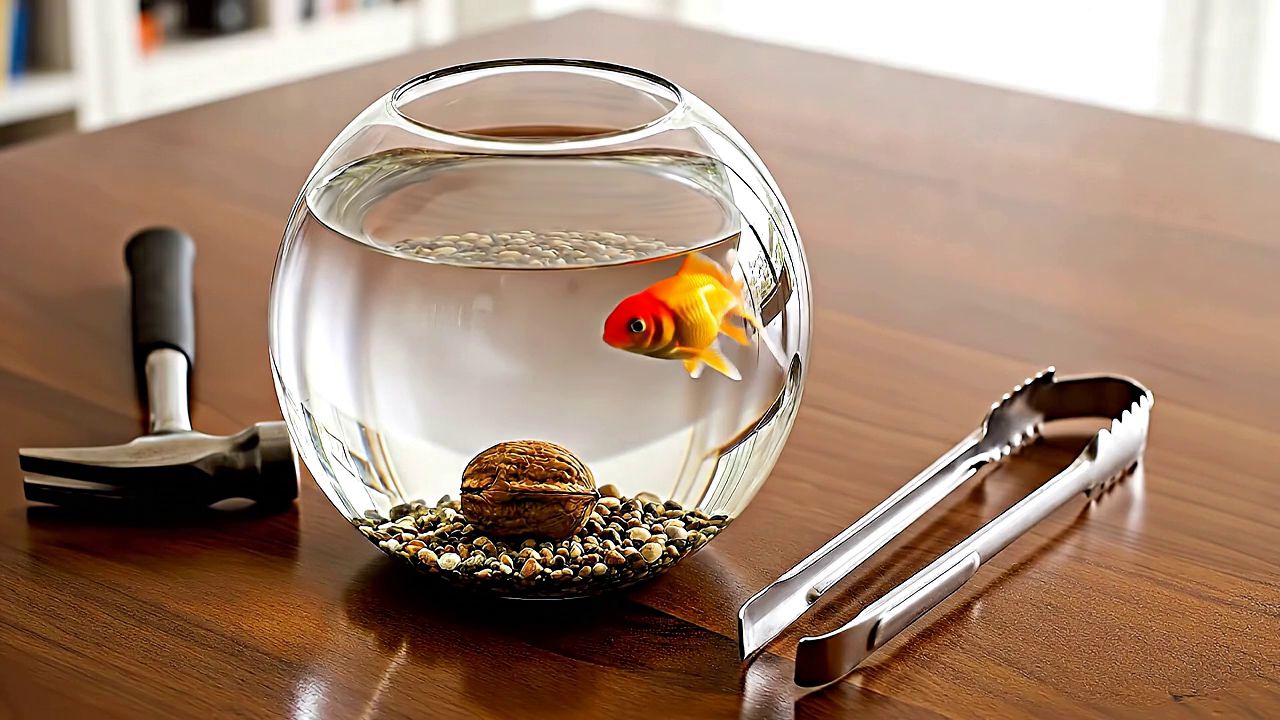}}
    \end{subfigure}
    \hfill
    \begin{subfigure}[t]{0.195\textwidth}
        \centering
        \fbox{\includegraphics[width=\linewidth]{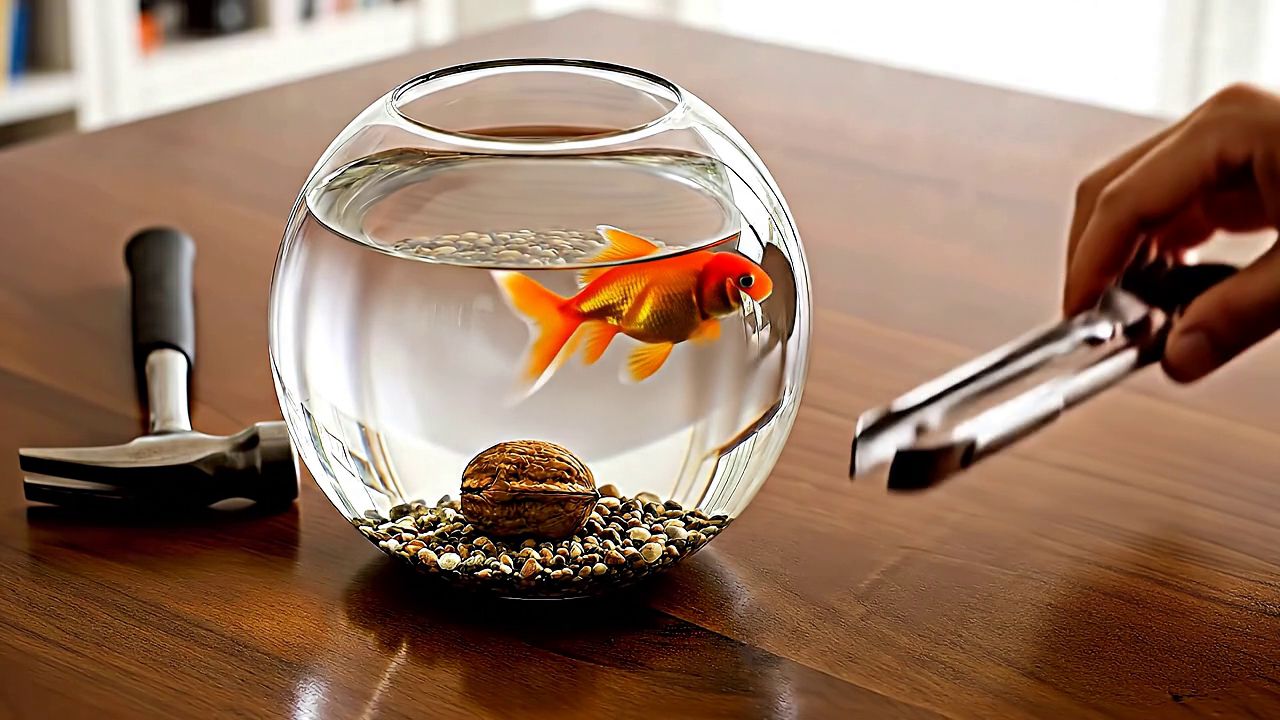}}
    \end{subfigure}
    \hfill
    \begin{subfigure}[t]{0.195\textwidth}
        \centering
        \fbox{\includegraphics[width=\linewidth]{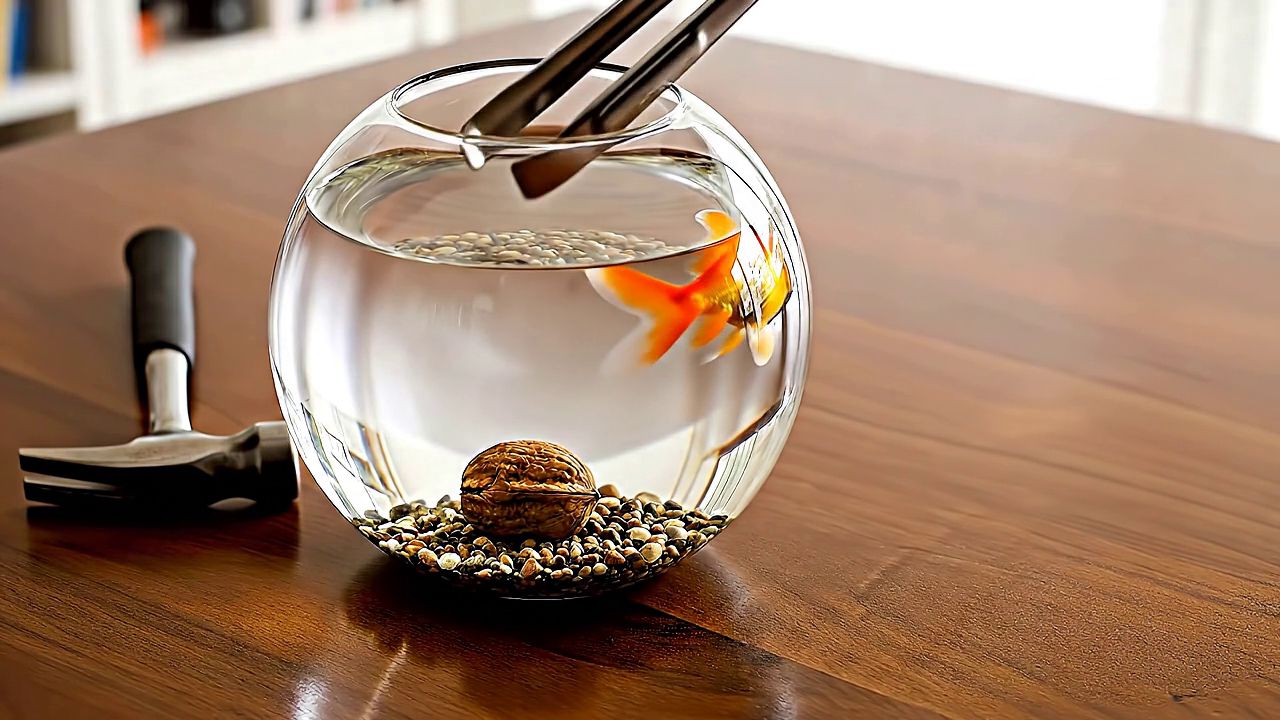}}
    \end{subfigure}
    \begin{subfigure}[t]{0.195\textwidth}
        \centering
        \fbox{\includegraphics[width=\linewidth]{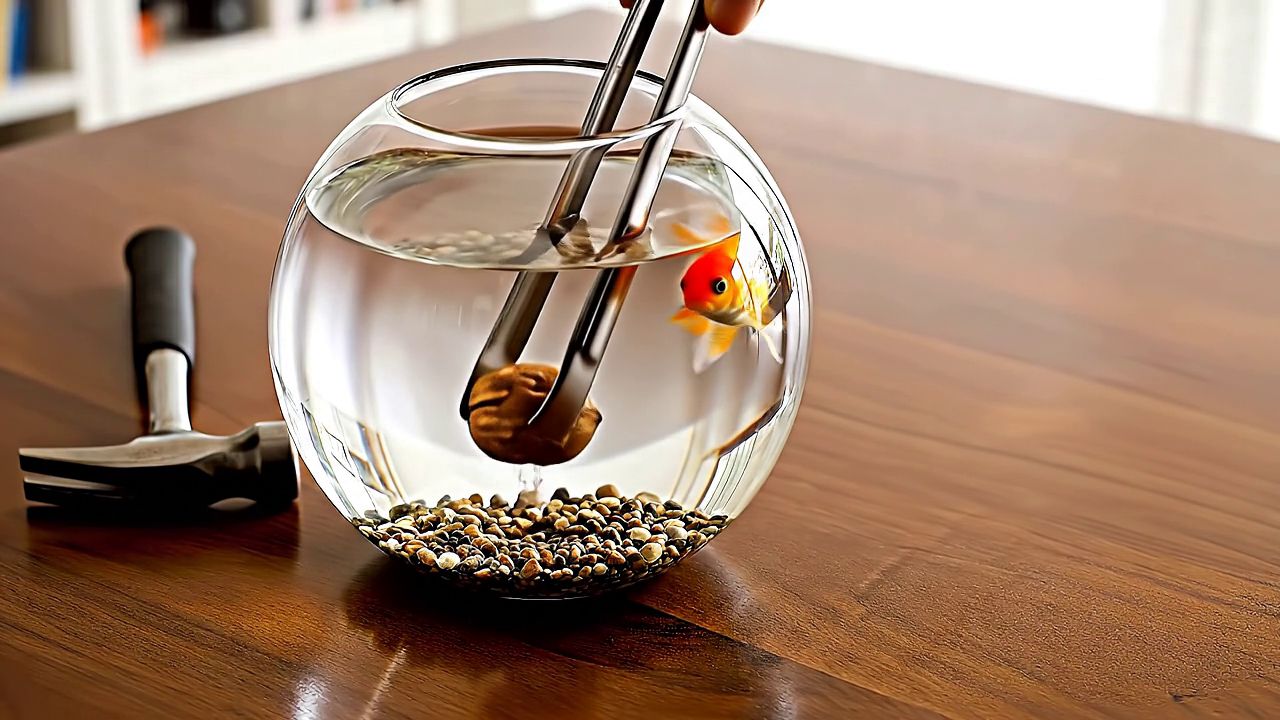}}
    \end{subfigure}
    \begin{subfigure}[t]{0.195\textwidth}
        \centering
        \fbox{\includegraphics[width=\linewidth]{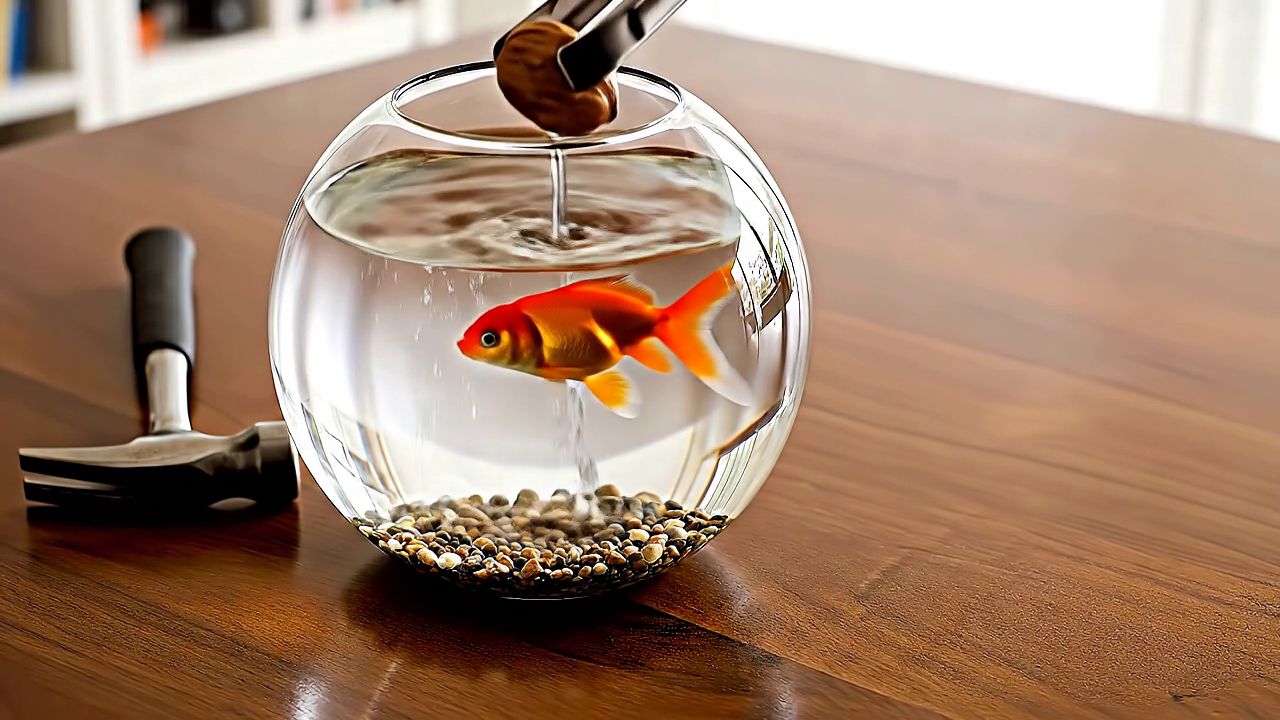}}
    \end{subfigure}
    \caption{\textbf{Tool use}.
    Prompt: \textit{``A person retrieves the walnut from the aquarium.''} Success rate:~0.92 (retrieval via tool) and 0.08 (retrieval via tool without intersecting the glass).}
    \label{fig:tool-use}
\end{figure}

\begin{figure}[H]
    \centering
    \begin{subfigure}[t]{\threefigwidth\textwidth}
        \centering
        \fbox{\includegraphics[width=\linewidth]{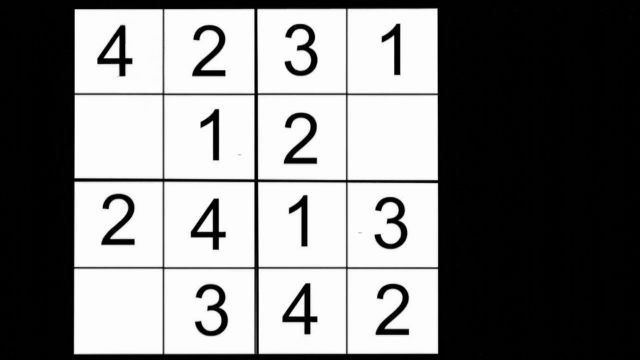}}
    \end{subfigure}
    \hfill
    \begin{subfigure}[t]{\threefigwidth\textwidth}
        \centering
        \fbox{\includegraphics[width=\linewidth]{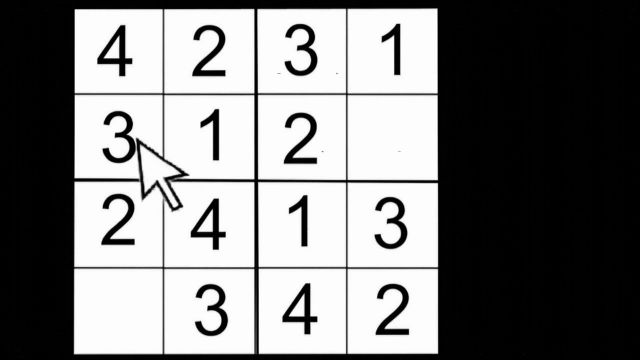}}
    \end{subfigure}
    \hfill
    \begin{subfigure}[t]{\threefigwidth\textwidth}
        \centering
        \fbox{\includegraphics[width=\linewidth]{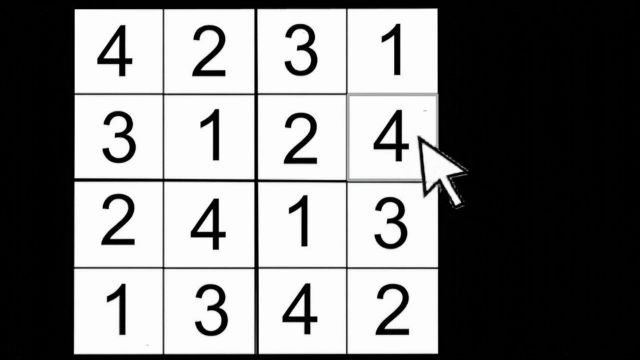}}
    \end{subfigure}
    \caption{\textbf{Simple Sudoku completion}. Prompt: \textit{``Create a static, smooth, animation that solves the given 4x4 sudoku. Enter the missing numbers one by one. Do not change anything else in the picture. Only fill the numbers in the empty cells so the sudoku is solved properly. A cursor moves and fills the correct number in the empty boxes.''} Success rate:~0.67.}
    \label{fig:sudoku}
\end{figure}

\begin{figure}[H]
    \centering
    \begin{subfigure}[t]{0.195\textwidth}
        \centering
        \fbox{\includegraphics[width=\linewidth]{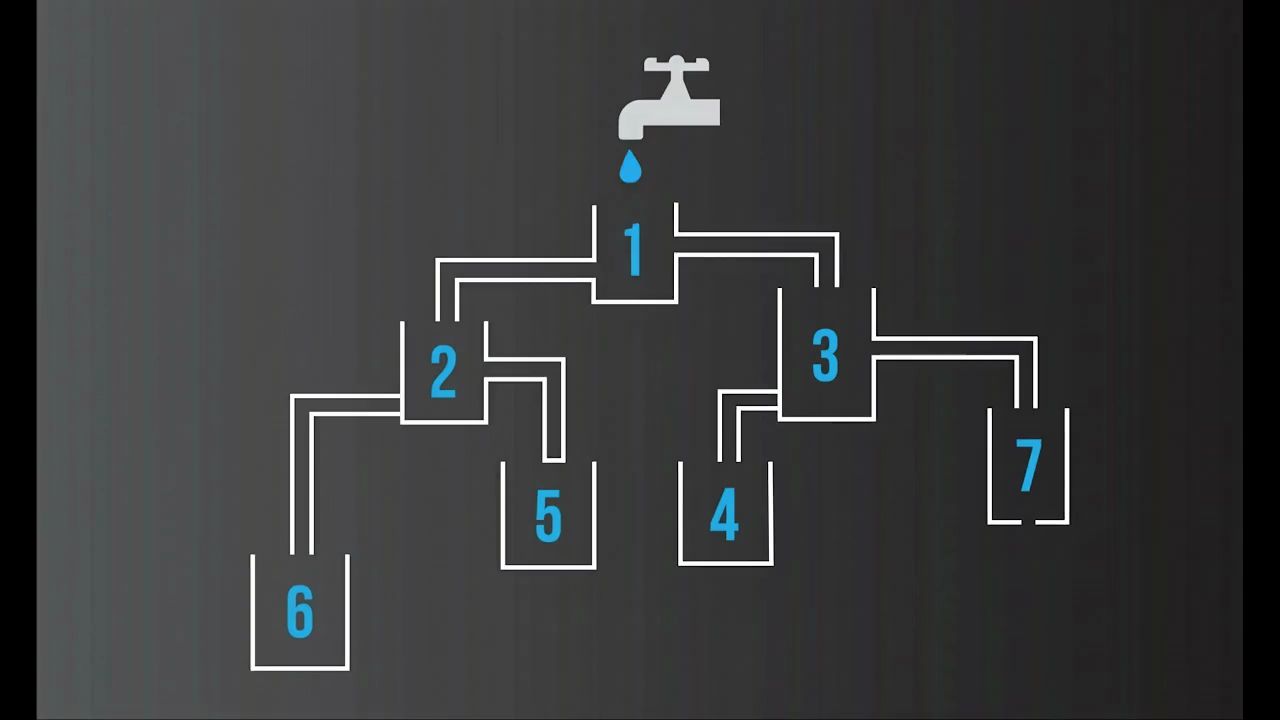}}
    \end{subfigure}
    \hfill
    \begin{subfigure}[t]{0.195\textwidth}
        \centering
        \fbox{\includegraphics[width=\linewidth]{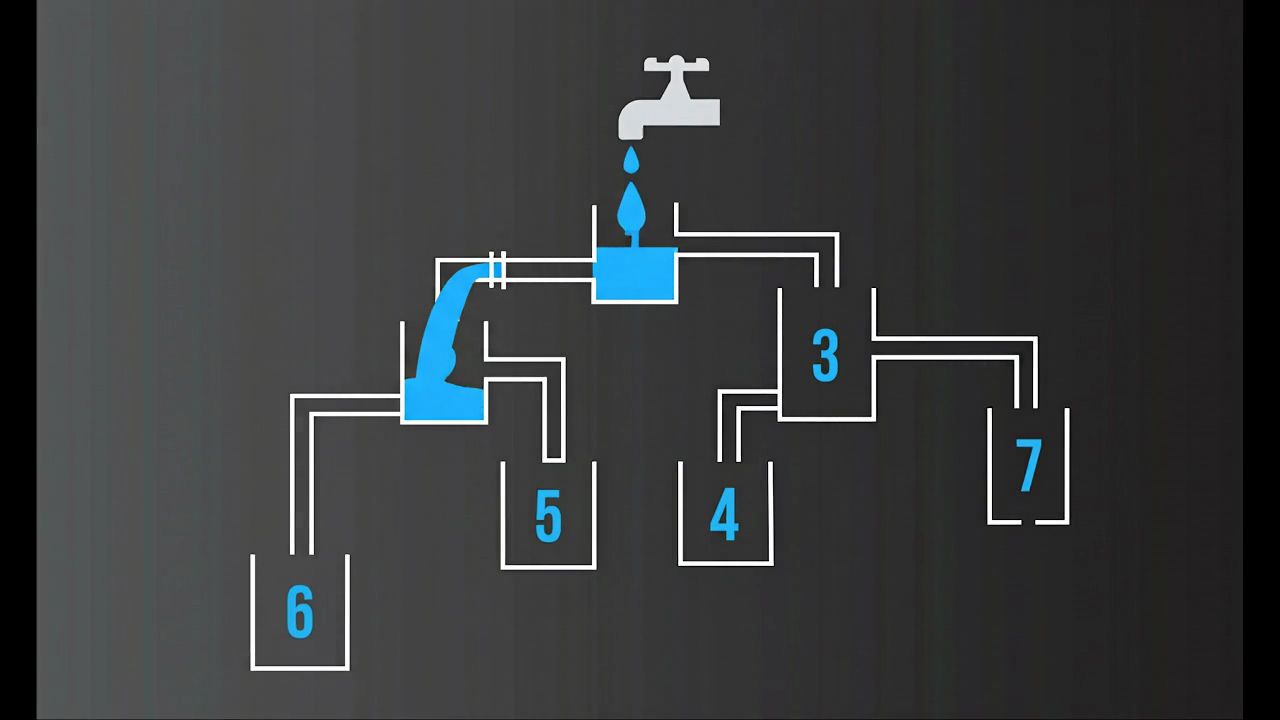}}
    \end{subfigure}
    \hfill
    \begin{subfigure}[t]{0.195\textwidth}
        \centering
        \fbox{\includegraphics[width=\linewidth]{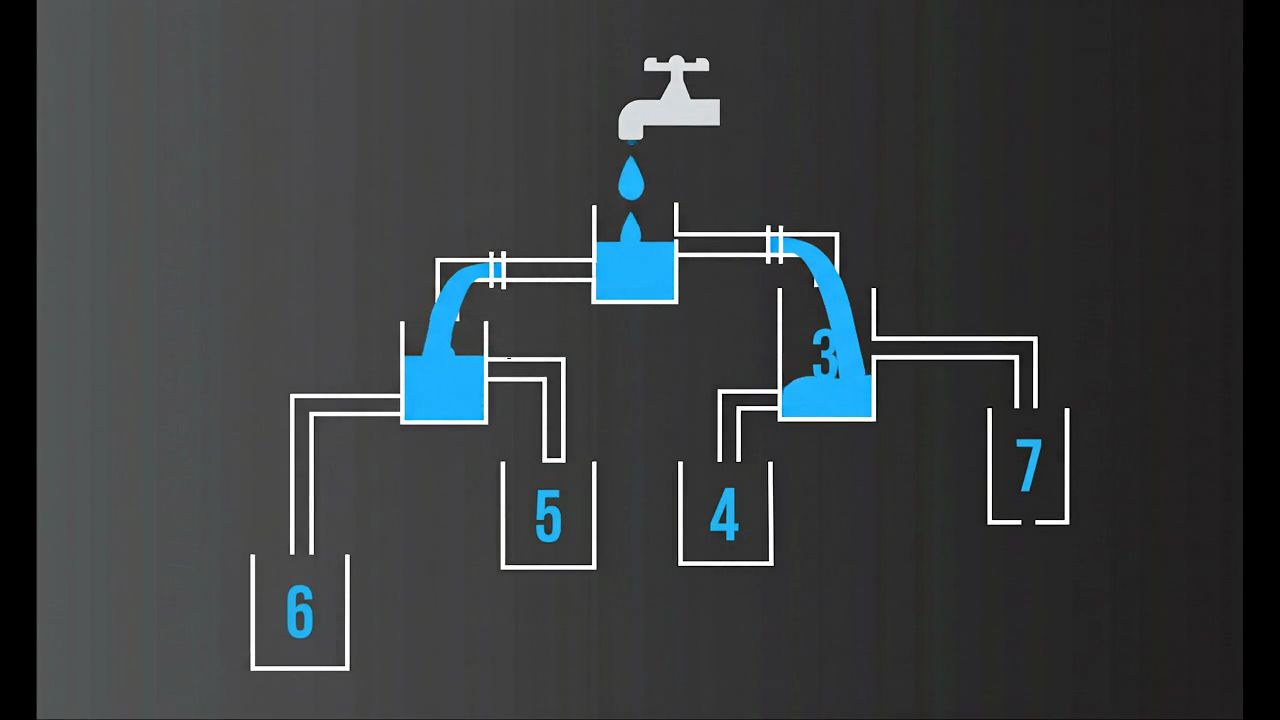}}
    \end{subfigure}
    \begin{subfigure}[t]{0.195\textwidth}
        \centering
        \fbox{\includegraphics[width=\linewidth]{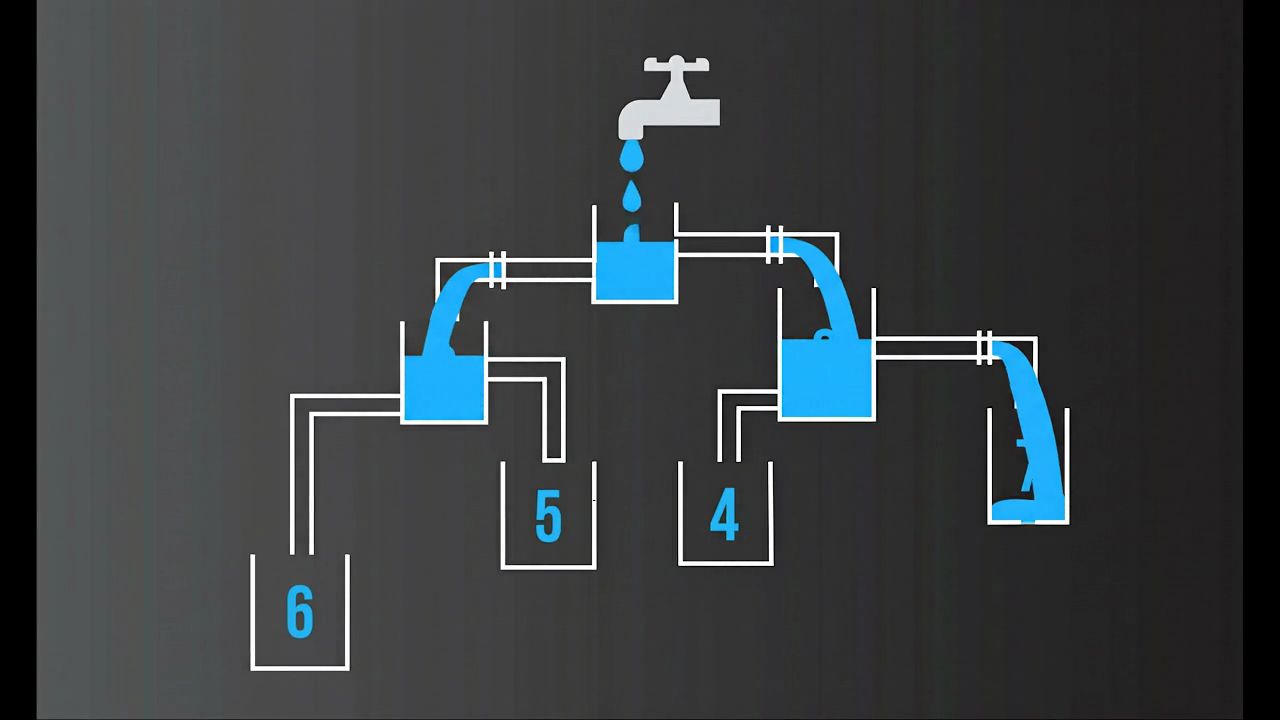}}
    \end{subfigure}
    \begin{subfigure}[t]{0.195\textwidth}
        \centering
        \fbox{\includegraphics[width=\linewidth]{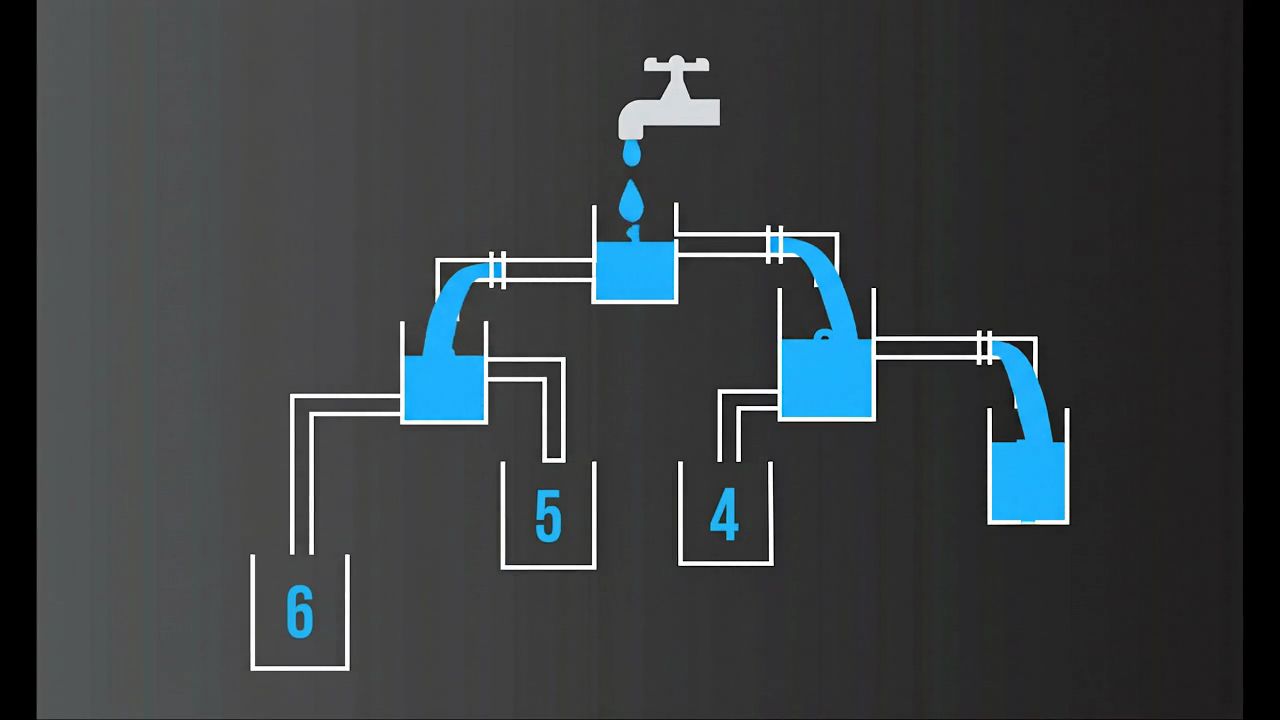}}
    \end{subfigure}
    \caption{\textbf{Water puzzle solving}.
    Prompt: \textit{``The tap is turned on and water starts flowing rapidly filling the containers. Create a smooth, static animation showing the containers getting filled with water in the correct order.''} (note: not all containers can be filled since some pipes are closed off---Veo fills the correct containers, in the right order.) Success rate:~0.5.}
    \label{fig:water-puzzle}
\end{figure}

\begin{figure}[H]
    \centering
    \begin{subfigure}[t]{\threefigwidth\textwidth}
        \centering
        \fbox{\includegraphics[width=\linewidth]{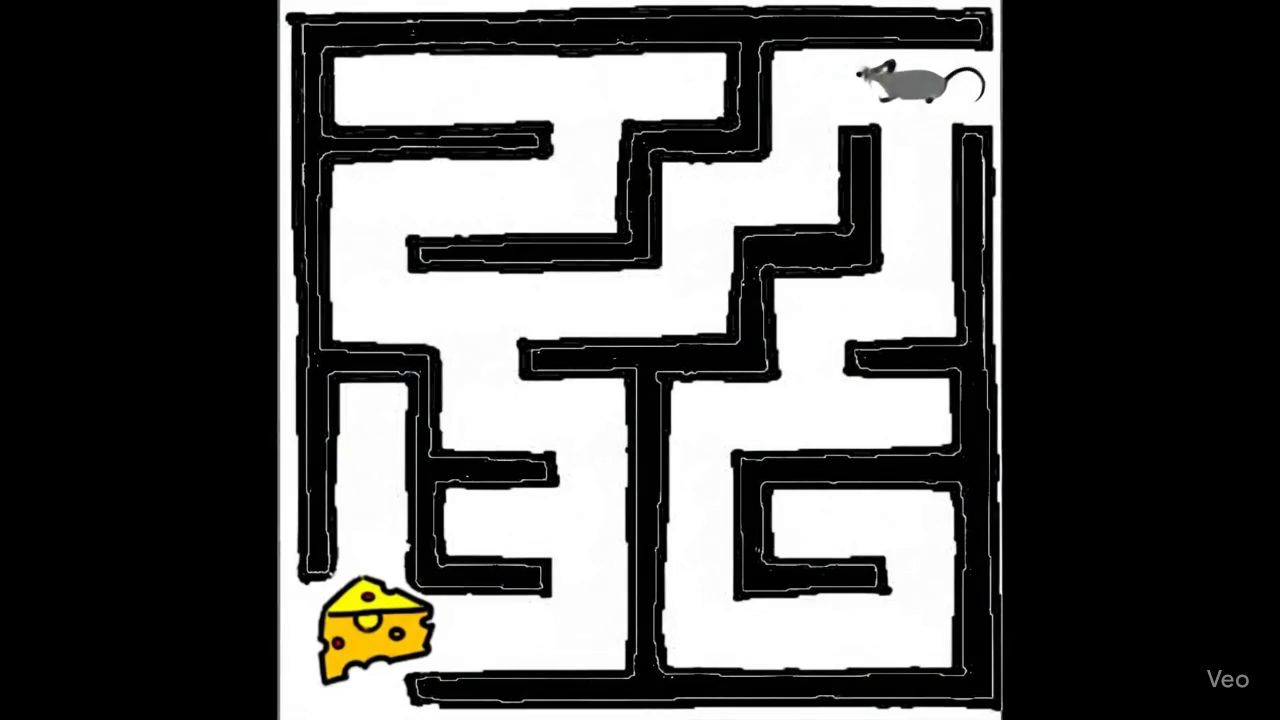}}
    \end{subfigure}
    \hfill
    \begin{subfigure}[t]{\threefigwidth\textwidth}
        \centering
        \fbox{\includegraphics[width=\linewidth]{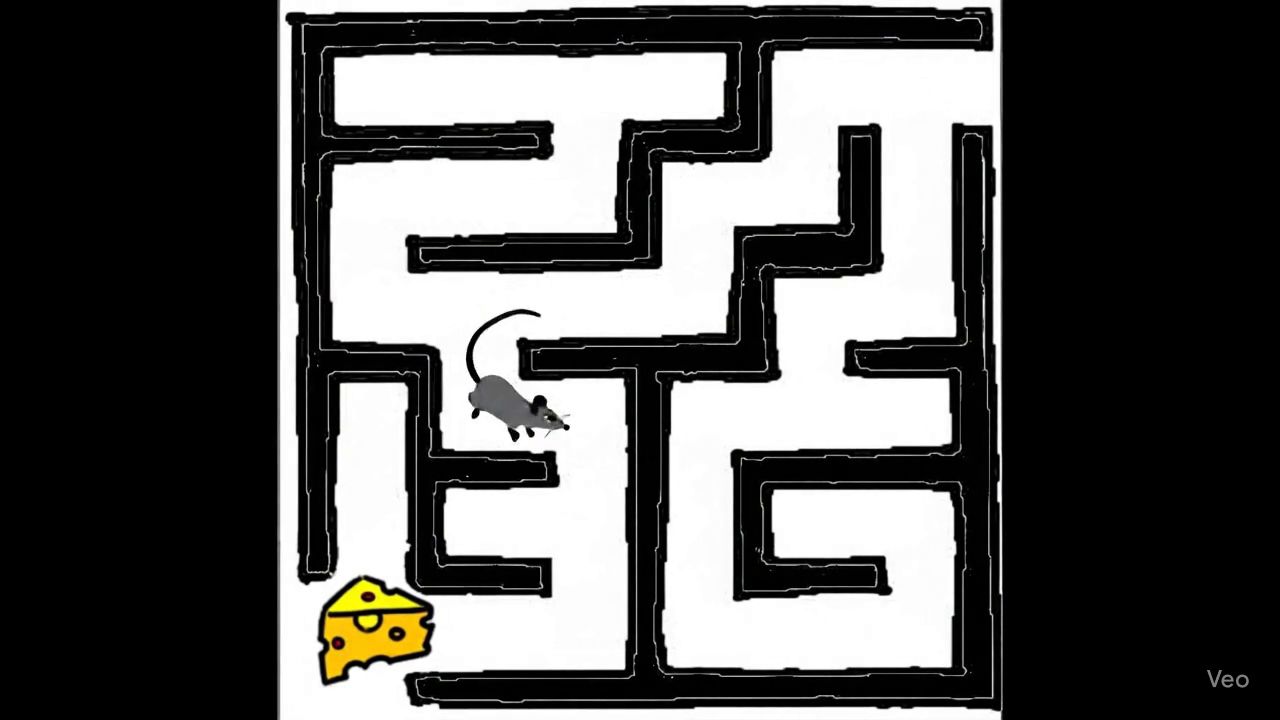}}
    \end{subfigure}
    \hfill    
    \begin{subfigure}[t]{\threefigwidth\textwidth}
        \centering
        \fbox{\includegraphics[width=\linewidth]{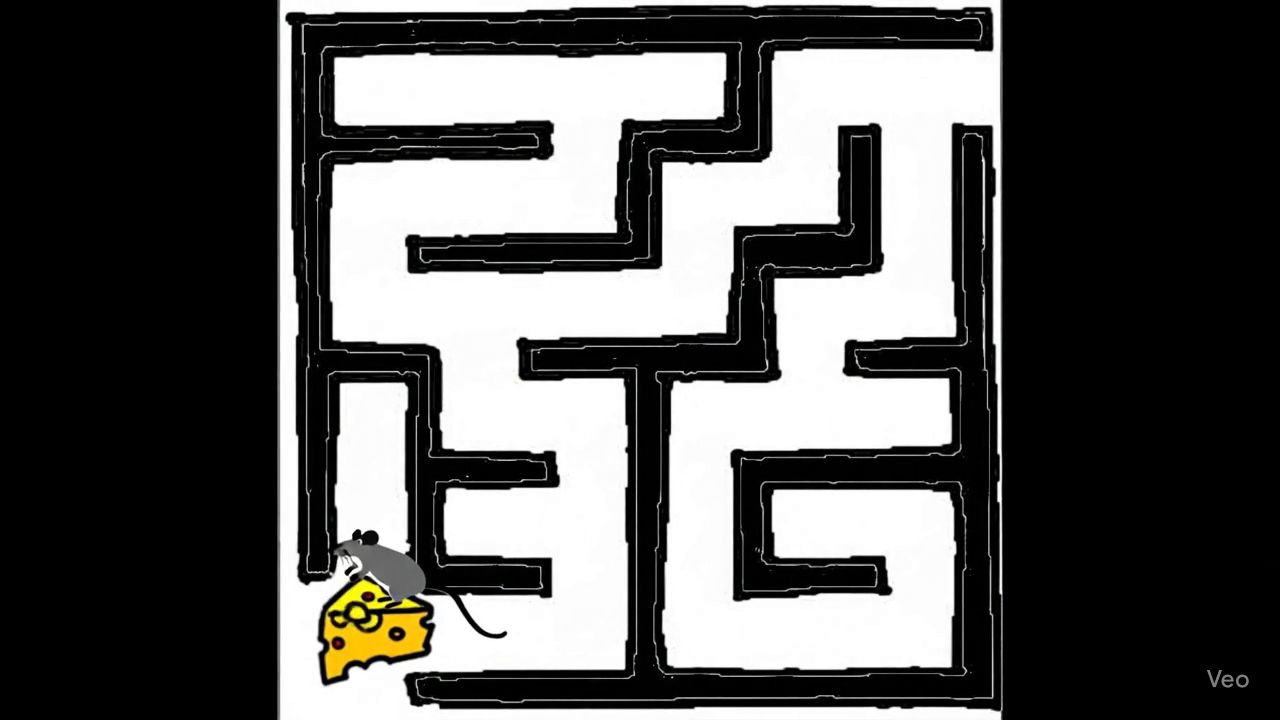}}
    \end{subfigure}
    \caption{\textbf{Maze solving}. Prompt: \textit{``Without crossing any black boundary, the grey mouse from the corner skillfully navigates the maze by walking around until it finds the yellow cheese.''} Success rate:~0.17.}
    \label{fig:maze_mouse}
\end{figure}

\begin{figure}[H]
    \centering
    \begin{subfigure}[t]{\threefigwidth\textwidth}
        \centering
        \fbox{\includegraphics[width=\linewidth]{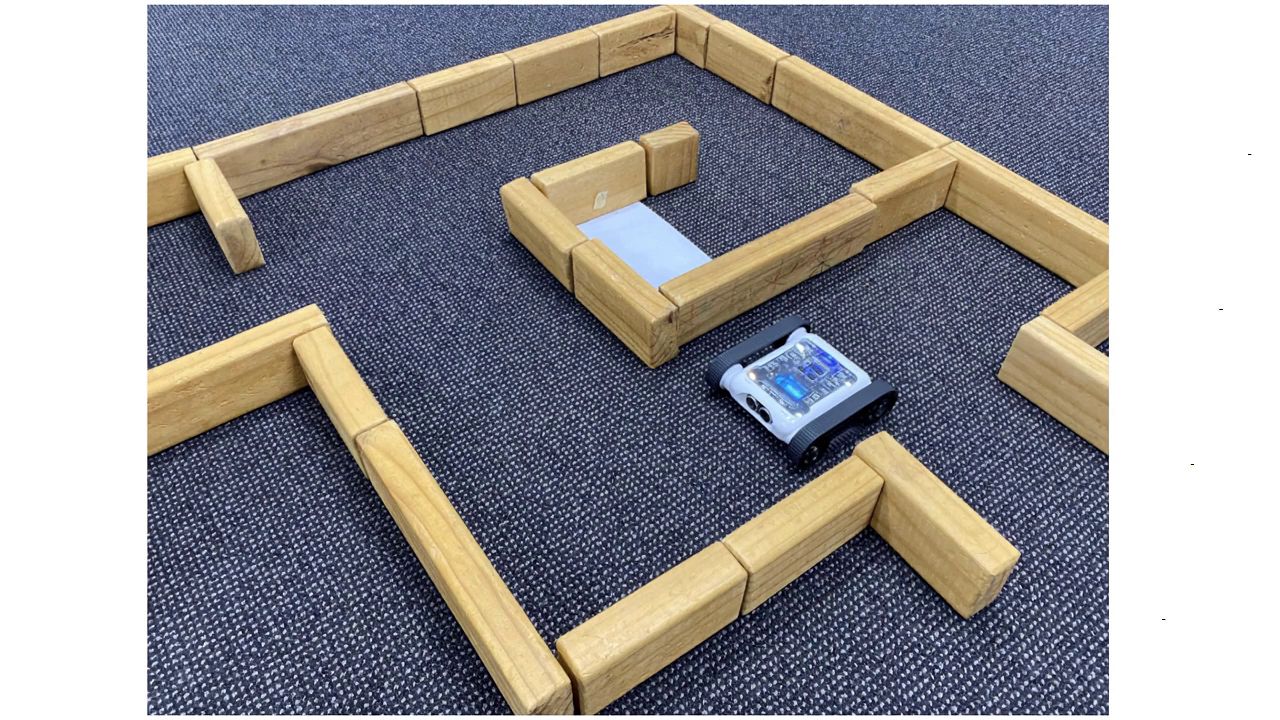}}
    \end{subfigure}
    \hfill
    \begin{subfigure}[t]{\threefigwidth\textwidth}
        \centering
        \fbox{\includegraphics[width=\linewidth]{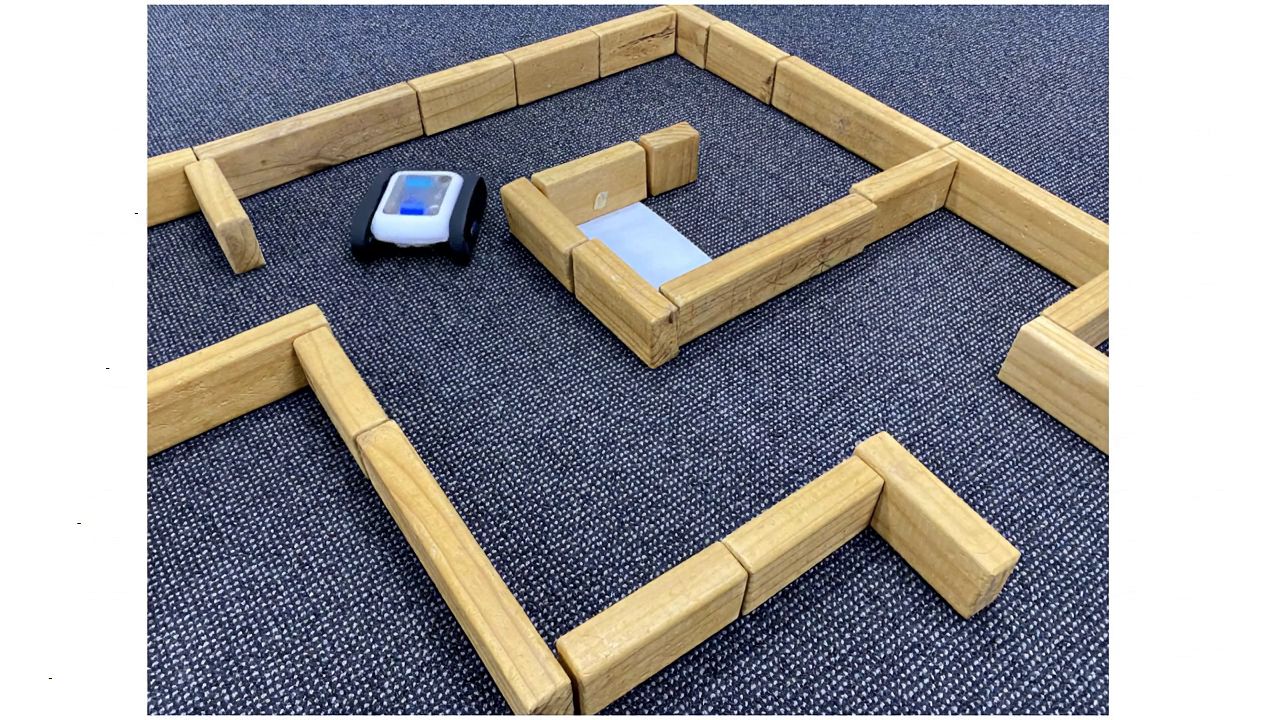}}
    \end{subfigure}
    \hfill    
    \begin{subfigure}[t]{\threefigwidth\textwidth}
        \centering
        \fbox{\includegraphics[width=\linewidth]{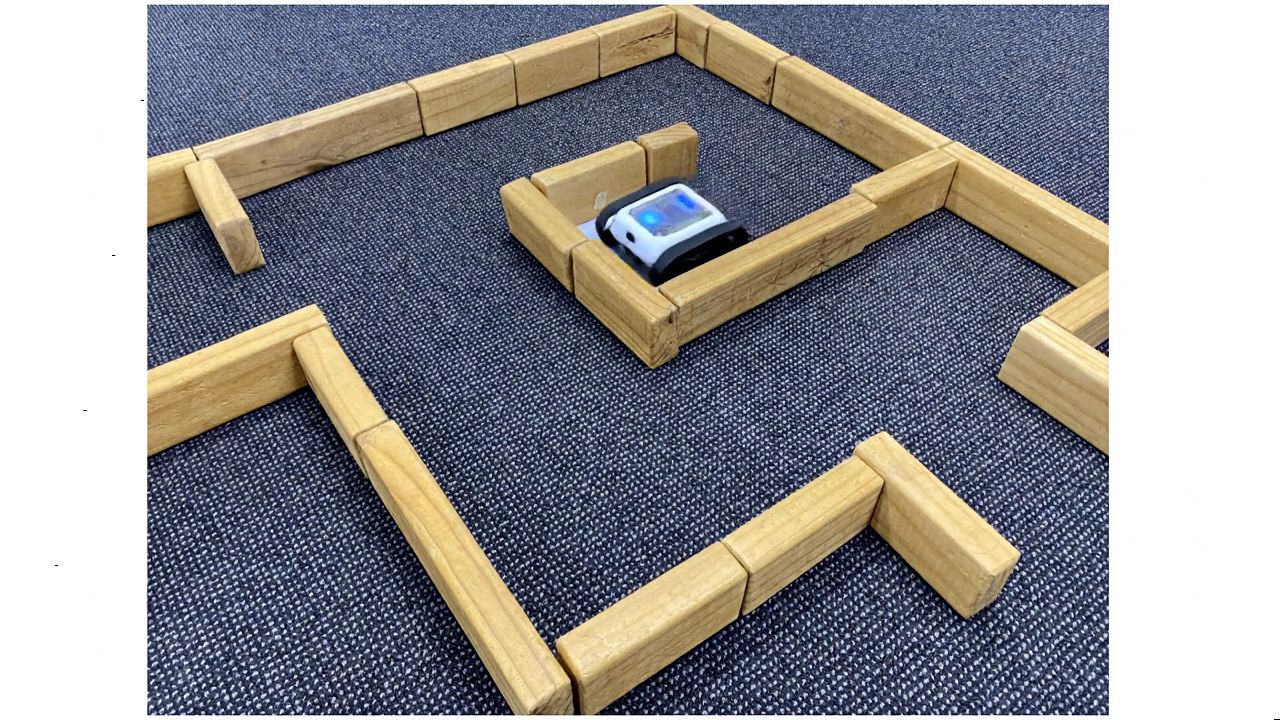}}
    \end{subfigure}
    \caption{\textbf{Robot navigation}. Prompt: \textit{``The robot drives to the blue area. Static camera perspective, no movement no zoom no scan no pan.''} Success rate:~0.58. Image credit: \href{https://micromelon.com.au/resources/creating-a-maze-unit}{Micromelon Robotics website} with permission from Tim Hadwen.}
    \label{fig:robot_navigation}
\end{figure}

\begin{figure}[H]
    \centering
    \begin{subfigure}[t]{\threefigwidth\textwidth}
        \centering
        \fbox{\includegraphics[width=\linewidth]{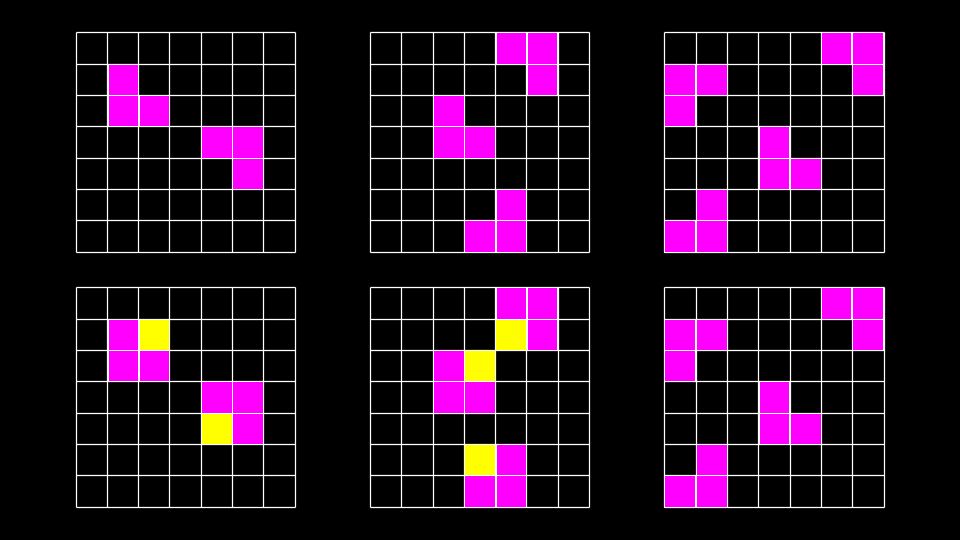}}
    \end{subfigure}
    \hfill
    \begin{subfigure}[t]{\threefigwidth\textwidth}
        \centering
        \fbox{\includegraphics[width=\linewidth]{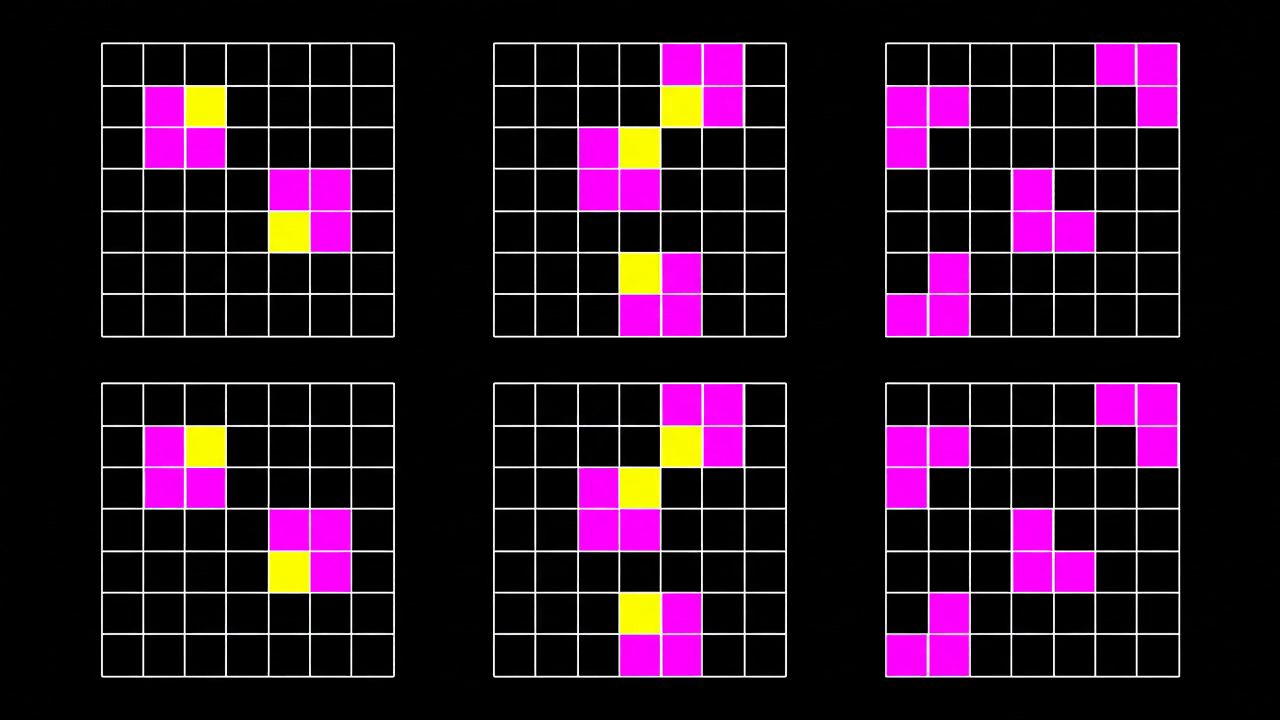}}
    \end{subfigure}
    \hfill    
    \begin{subfigure}[t]{\threefigwidth\textwidth}
        \centering
        \fbox{\includegraphics[width=\linewidth]{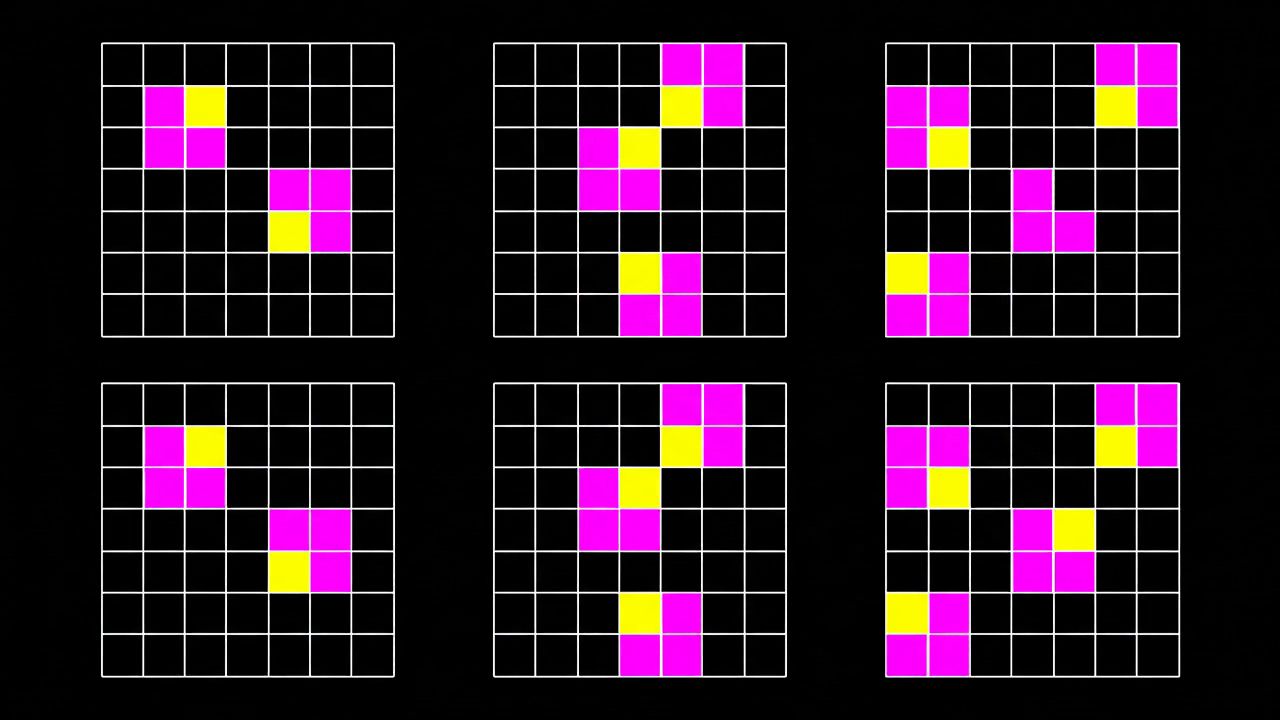}}
    \end{subfigure}
    \caption{\textbf{Rule extrapolation} inspired by ARC-AGI~\citep{chollet2019measure}. Prompt: \textit{``Modify the lower-right grid to adhere to the rule established by the other grids. You can fill cells, clear cells, or change a cell's color. Only modify the lower-right grid, don't modify any of the other grids. Static scene, no zoom, no pan, no dolly.''} Success rate:~0.08. While \veothree{} doesn't follow the prompt perfectly, the output grid (bottom right) is completed correctly.}
    \label{fig:arc-agi}
\end{figure}